%% file: main.tex
\PassOptionsToPackage{table}{xcolor}
\documentclass{article}

\usepackage{iclr2027_conference,times}

\input{custom_templates/math}

\input{style_iclr2027}

\usepackage{graphicx}
\usepackage{tikz}
\usepackage{multirow}
\usepackage{fontawesome5}
\usepackage{enumitem}
\usetikzlibrary{arrows.meta,calc,patterns,positioning,shapes.geometric}

\input{vmr-symbols-vecbold.tex} 

\title{\centering LinQuant: Functional Invariances for\\Low-Precision Linear Recurrences}

\author{%
\parbox[t]{\dimexpr\textwidth-2\tabcolsep\relax}{\centering
Neehal Tumma\textsuperscript{\normalfont *\,1}\quad
Philipp Nazari\textsuperscript{\normalfont *\,2\,3\,4\,5}\quad
Rohan Cherukuri\textsuperscript{\normalfont 1}\\[0.4em]
Daniela Rus\textsuperscript{\normalfont 1}\quad
T. Konstantin Rusch\textsuperscript{\normalfont 2\,4\,5}\\[0.5em]
\normalfont\small
\textsuperscript{1}MIT CSAIL\quad
\textsuperscript{2}Max Planck Institute for Intelligent Systems\\
\textsuperscript{3}ETH Z\"urich\quad
\textsuperscript{4}T\"ubingen AI Center\quad
\textsuperscript{5}ELLIS Institute T\"ubingen
}}

\iclrfinalcopy
\begin{document}

\maketitle
\begingroup
\renewcommand{\thefootnote}{}
\begin{NoHyper}
\footnotetext{\textsuperscript{*}Equal contribution.\quad Correspondence: \texttt{ntumma@mit.edu}.}
\end{NoHyper}
\endgroup
\fancyhead{}
\fancyhead[L]{\footnotesize\normalfont LinQuant: Functional Invariances for Low-Precision Linear Recurrences}
\renewcommand{\headrulewidth}{0.4pt}

\begin{abstract}
Linear recurrences are emerging as practical alternatives to softmax attention in large language models, appearing in both standalone and hybrid architectures. However, despite their rapid emergence, low-precision inference remains much less developed for these models than for softmax attention. During prefill, chunkwise-parallel algorithms cast much of the recurrence as dense matrix multiplications, suggesting a natural path to low-precision execution. Simply quantizing these operations in their native form, however, can be fragile. In this work, we identify functional invariances that preserve the full-precision computation while making the operands more amenable to quantization. Building on this observation, we introduce LinQuant, a framework for low-precision recurrent prefill based on seven such invariances, including rotations, diagonal rescalings, and alternative placements of recurrent decay and write factors. We derive when these transformations remain valid across common recurrence families spanning linear, gated, and delta-rule updates, and show that they provide complementary degrees of freedom for improving low-precision accuracy. Finally, we translate this framework to production by augmenting SSDi8, an existing low-precision prefill pipeline for Mamba-2, with the applicable invariances. LinQuant improves perplexity at matched latency on Mamba-2 {1.3B, 2.7B, and 8B} and Nemotron-H 8B, demonstrating the practical utility of invariance-based quantization for recurrent models.
\end{abstract}

\vspace{-1em}
{\centering\small
\href{https://github.com/ntumm120/LinQuant}{{\faGithub\enspace\nolinkurl{github.com/ntumm120/LinQuant}}}\par}

\input{main_material/1_introduction}
\input{main_material/2_background_and_preliminaries}
\input{main_material/3_method}
\input{main_material/4_experiments}
\input{main_material/5_conclusion}

\newpage

\section*{Acknowledgements}
This work was supported by the Hector Foundation, Max Planck ETH Center for Learning Systems, and the Department of the Air Force Artificial Intelligence Accelerator and was accomplished under Cooperative Agreement Number FA8750-19-2-1000. The views and conclusions contained in this document are those of the authors and should not be interpreted as representing the official policies, either expressed or implied, of the Department of the Air Force or the U.S. Government. The U.S. Government is authorized to reproduce and distribute reprints for Government purposes notwithstanding any copyright notation herein.

\bibliographystyle{iclr2027_conference}
\bibliography{iclr2027_conference}

\clearpage
\appendix

\input{supplementary_material/A_extended_related_work}
\input{supplementary_material/B_experimental_setup}
\input{supplementary_material/C_invariances}
\input{supplementary_material/D_kernels}
\input{supplementary_material/E_zoo_results}
\clearpage
\input{supplementary_material/F_production_results}

\end{document}

%% file: custom_templates/math.tex
\usepackage{amsmath}
\usepackage{amssymb}
\usepackage{amsfonts}
\usepackage{mathtools}
\usepackage{bm}
\usepackage{bbm}
\usepackage{xcolor}
\usepackage{algorithm}
\usepackage{algpseudocode}

\definecolor{chunksumcolor}{HTML}{045275}
\definecolor{chunkstatecolor}{HTML}{089099}
\definecolor{chunkstatepassingcolor}{HTML}{F0746E}
\definecolor{chunkbmmcolor}{HTML}{DC3977}
\definecolor{chunkscancolor}{HTML}{7C1D6F}
\definecolor{highlightcolor}{HTML}{7CCBA2}
\definecolor{highlightcolor2}{HTML}{FCDE9C}
\definecolor{highlightcolor3}{HTML}{F7B2AD}

\newcommand{\bb}[1]{\bm{#1}}
\newcommand{\mat}[1]{\bm{\mathrm{#1}}}
\newcommand{\R}{\mathbb{R}}
\newcommand{\T}{^{\mathsf T}}

\DeclareMathOperator{\diag}{diag}
\DeclareMathOperator{\tril}{tril}
\DeclareMathOperator{\clip}{clip}

\definecolor{decaycol}{RGB}{0,114,178}
\definecolor{gatecol}{RGB}{230,159,0}
\definecolor{deltacol}{RGB}{213,94,0}
\definecolor{rotcol}{HTML}{089099}
\definecolor{rescalecol}{HTML}{DC3977}

\newcommand{\DecayTerm}[1]{{\color{decaycol}#1}}
\newcommand{\WriteGateTerm}[1]{{\color{gatecol}#1}}
\newcommand{\DeltaTerm}[1]{{\color{deltacol}#1}}
\newcommand{\RotTerm}[1]{{\color{rotcol}#1}}
\newcommand{\RescaleTerm}[1]{{\color{rescalecol}#1}}

\newcommand{\Dtail}{\mat{\Gamma}^{\rightarrow}}
\newcommand{\Dpref}{\mat{\Gamma}^{\leftarrow}}


%% file: style_iclr2027.tex
\usepackage[table]{xcolor}
\usepackage{microtype}
\usepackage{xurl}
\usepackage{url}
\usepackage{booktabs}
\usepackage{makecell}
\usepackage{array}
\usepackage{graphicx}
\usepackage{nicefrac}
\usepackage{caption}
\usepackage{subcaption}
\usepackage{tabularx}
\usepackage{adjustbox}
\usepackage{float}
\usepackage[normalem]{ulem}
\usepackage{wrapfig}
\usepackage{needspace}

\usepackage{listings}
\usepackage{mdframed}
\usepackage[many]{tcolorbox}
\tcbuselibrary{breakable,xparse,skins}

\usepackage{hyperref}

\definecolor{mydarkblue}{rgb}{0,0.08,0.45}
\hypersetup{
  bookmarks=true,
  colorlinks=true,
  citecolor=mydarkblue,
  urlcolor=mydarkblue,
  linkcolor=mydarkblue
}

\newtcolorbox{example}[0]{breakable, enhanced, sharp corners, frame hidden, colback=green!5}
\newtcolorbox{note}{
    colback=black!10!white,
    colframe=white,
    boxrule=0pt,
    left=1pt,
    right=1pt,
    top=1pt,
    bottom=1pt,
    breakable,
}

\definecolor{codegreen}{rgb}{0,0.6,0}
\definecolor{codegray}{rgb}{0.5,0.5,0.5}
\definecolor{codepurple}{rgb}{0.58,0,0.82}
\definecolor{ghostwhite}{rgb}{0.97,0.97,1.0}

\lstdefinestyle{mystyle}{
    backgroundcolor=\color{ghostwhite},
    commentstyle=\color{codegreen},
    keywordstyle=\color{magenta},
    numberstyle=\tiny\color{codegray},
    stringstyle=\color{codepurple},
    basicstyle=\ttfamily\footnotesize,
    breakatwhitespace=false,
    breaklines=true,
    captionpos=b,
    keepspaces=true,
    numbers=left,
    numbersep=5pt,
    showspaces=false,
    showstringspaces=false,
    showtabs=false,
    tabsize=2
}


%% file: vmr-symbols-vecbold.tex
\usepackage{amssymb}
\usepackage{amsfonts}
\usepackage{mathrsfs}
\usepackage{xspace}
\usepackage{bm}
\usepackage{upgreek}

\newcommand{\safemath}[2]{\newcommand{#1}{\ensuremath{#2}\xspace}}

\safemath{\bma}{\mathbf{a}}
\safemath{\bmb}{\mathbf{b}}
\safemath{\bmc}{\mathbf{c}}
\safemath{\bmd}{\mathbf{d}}
\safemath{\bme}{\mathbf{e}}
\safemath{\bmf}{\mathbf{f}}
\safemath{\bmg}{\mathbf{g}}
\safemath{\bmh}{\mathbf{h}}
\safemath{\bmi}{\mathbf{i}}
\safemath{\bmj}{\mathbf{j}}
\safemath{\bmk}{\mathbf{k}}
\safemath{\bml}{\mathbf{l}}
\safemath{\bmm}{\mathbf{m}}
\safemath{\bmn}{\mathbf{n}}
\safemath{\bmo}{\mathbf{o}}
\safemath{\bmp}{\mathbf{p}}
\safemath{\bmq}{\mathbf{q}}
\safemath{\bmr}{\mathbf{r}}
\safemath{\bms}{\mathbf{s}}
\safemath{\bmt}{\mathbf{t}}
\safemath{\bmu}{\mathbf{u}}
\safemath{\bmv}{\mathbf{v}}
\safemath{\bmw}{\mathbf{w}}
\safemath{\bmx}{\mathbf{x}}
\safemath{\bmy}{\mathbf{y}}
\safemath{\bmz}{\mathbf{z}}
\safemath{\bmzero}{\mathbf{0}}
\safemath{\bmone}{\mathbf{1}}

\safemath{\bmbeta}{\mathbf{beta}}

\bmdefine{\biad}{a}
\bmdefine{\bibd}{b}
\bmdefine{\bicd}{c}
\bmdefine{\bidd}{d}
\bmdefine{\bied}{e}
\bmdefine{\bifd}{f}
\bmdefine{\bigd}{g}
\bmdefine{\bihd}{h}
\bmdefine{\biid}{i}
\bmdefine{\bijd}{j}
\bmdefine{\bikd}{k}
\bmdefine{\bild}{l}
\bmdefine{\bimd}{m}
\bmdefine{\bind}{n}
\bmdefine{\biod}{o}
\bmdefine{\bipd}{p}
\bmdefine{\biqd}{q}
\bmdefine{\bird}{r}
\bmdefine{\bisd}{s}
\bmdefine{\bitd}{t}
\bmdefine{\biud}{u}
\bmdefine{\bivd}{v}
\bmdefine{\biwd}{w}
\bmdefine{\bixd}{x}
\bmdefine{\biyd}{y}
\bmdefine{\bizd}{z}

\bmdefine{\bixid}{\xi}
\bmdefine{\bilambdad}{\lambda}
\bmdefine{\bimud}{\mu}
\bmdefine{\bithetad}{\theta}
\bmdefine{\biphid}{\phi}
\bmdefine{\bialphad}{\alpha}

\safemath{\bmia}{\biad}
\safemath{\bmib}{\bibd}
\safemath{\bmic}{\bicd}
\safemath{\bmid}{\bidd}
\safemath{\bmie}{\bied}
\safemath{\bmif}{\bifd}
\safemath{\bmig}{\bigd}
\safemath{\bmih}{\bihd}
\safemath{\bmii}{\biid}
\safemath{\bmij}{\bijd}
\safemath{\bmik}{\bikd}
\safemath{\bmil}{\bild}
\safemath{\bmim}{\bimd}
\safemath{\bmin}{\bind}
\safemath{\bmio}{\biod}
\safemath{\bmip}{\bipd}
\safemath{\bmiq}{\biqd}
\safemath{\bmir}{\bird}
\safemath{\bmis}{\bisd}
\safemath{\bmit}{\bitd}
\safemath{\bmiu}{\biud}
\safemath{\bmiv}{\bivd}
\safemath{\bmiw}{\biwd}
\safemath{\bmix}{\bixd}
\safemath{\bmiy}{\biyd}
\safemath{\bmiz}{\bizd}

\safemath{\bmxi}{\bixid}
\safemath{\bmlambda}{\bilambdad}
\safemath{\bmalpha}{\bialphad}
\safemath{\bmmu}{\bimud}
\safemath{\bmtheta}{\bithetad}
\safemath{\bmphi}{\biphid}

\safemath{\bA}{\mathbf{A}}
\safemath{\bB}{\mathbf{B}}
\safemath{\bC}{\mathbf{C}}
\safemath{\bD}{\mathbf{D}}
\safemath{\bE}{\mathbf{E}}
\safemath{\bF}{\mathbf{F}}
\safemath{\bG}{\mathbf{G}}
\safemath{\bH}{\mathbf{H}}
\safemath{\bI}{\mathbf{I}}
\safemath{\bJ}{\mathbf{J}}
\safemath{\bK}{\mathbf{K}}
\safemath{\bL}{\mathbf{L}}
\safemath{\bM}{\mathbf{M}}
\safemath{\bN}{\mathbf{N}}
\safemath{\bO}{\mathbf{O}}
\safemath{\bP}{\mathbf{P}}
\safemath{\bQ}{\mathbf{Q}}
\safemath{\bR}{\mathbf{R}}
\safemath{\bS}{\mathbf{S}}
\safemath{\bT}{\mathbf{T}}
\safemath{\bU}{\mathbf{U}}
\safemath{\bV}{\mathbf{V}}
\safemath{\bW}{\mathbf{W}}
\safemath{\bX}{\mathbf{X}}
\safemath{\bY}{\mathbf{Y}}
\safemath{\bZ}{\mathbf{Z}}

\safemath{\bZero}{\mathbf{0}}
\safemath{\bOne}{\mathbf{1}}
\safemath{\bDelta}{\mathbf{\Delta}}
\safemath{\bLambda}{\mathbf{\UpLambda}}
\safemath{\bPhi}{\mathbf{\Upphi}}
\safemath{\bSigma}{\mathbf{\Upsigma}}
\safemath{\bOmega}{\mathbf{\Upomega}}
\safemath{\bTheta}{\mathbf{\Uptheta}}

\bmdefine{\biAd}{A}
\bmdefine{\biBd}{B}
\bmdefine{\biCd}{C}
\bmdefine{\biDd}{D}
\bmdefine{\biEd}{E}
\bmdefine{\biFd}{F}
\bmdefine{\biGd}{G}
\bmdefine{\biHd}{H}
\bmdefine{\biId}{I}
\bmdefine{\biJd}{J}
\bmdefine{\biKd}{K}
\bmdefine{\biLd}{L}
\bmdefine{\biMd}{M}
\bmdefine{\biOd}{N}
\bmdefine{\biPd}{O}
\bmdefine{\biQd}{P}
\bmdefine{\biRd}{R}
\bmdefine{\biSd}{S}
\bmdefine{\biTd}{T}
\bmdefine{\biUd}{U}
\bmdefine{\biVd}{V}
\bmdefine{\biWd}{W}
\bmdefine{\biXd}{X}
\bmdefine{\biYd}{Y}
\bmdefine{\biZd}{Z}

\bmdefine{\biDelta}{\Delta}
\bmdefine{\biLambda}{\Lambda}
\bmdefine{\biPhi}{\Phi}
\bmdefine{\biSigma}{\Sigma}
\bmdefine{\biOmega}{\Omega}
\bmdefine{\biTheta}{\Theta}

\safemath{\bimA}{\biAd}
\safemath{\bimB}{\biBd}
\safemath{\bimC}{\biCd}
\safemath{\bimD}{\biDd}
\safemath{\bimE}{\biEd}
\safemath{\bimF}{\biFd}
\safemath{\bimG}{\biGd}
\safemath{\bimH}{\biHd}
\safemath{\bimI}{\biId}
\safemath{\bimJ}{\biJd}
\safemath{\bimK}{\biKd}
\safemath{\bimL}{\biLd}
\safemath{\bimM}{\biMd}
\safemath{\bimN}{\biNd}
\safemath{\bimO}{\biOd}
\safemath{\bimP}{\biPd}
\safemath{\bimQ}{\biQd}
\safemath{\bimR}{\biRd}
\safemath{\bimS}{\biSd}
\safemath{\bimT}{\biTd}
\safemath{\bimU}{\biUd}
\safemath{\bimV}{\biVd}
\safemath{\bimW}{\biWd}
\safemath{\bimX}{\biXd}
\safemath{\bimY}{\biYd}
\safemath{\bimZ}{\biZd}

\safemath{\bimDelta}{\biDelta}
\safemath{\bimLambda}{\biLambda}
\safemath{\bimPhi}{\biPhi}
\safemath{\bimSigma}{\biSigma}
\safemath{\bimOmega}{\biOmega}
\safemath{\bimTheta}{\biTheta}

\safemath{\setA}{\mathcal{A}}
\safemath{\setB}{\mathcal{B}}
\safemath{\setC}{\mathcal{C}}
\safemath{\setD}{\mathcal{D}}
\safemath{\setE}{\mathcal{E}}
\safemath{\setF}{\mathcal{F}}
\safemath{\setG}{\mathcal{G}}
\safemath{\setH}{\mathcal{H}}
\safemath{\setI}{\mathcal{I}}
\safemath{\setJ}{\mathcal{J}}
\safemath{\setK}{\mathcal{K}}
\safemath{\setL}{\mathcal{L}}
\safemath{\setM}{\mathcal{M}}
\safemath{\setN}{\mathcal{N}}
\safemath{\setO}{\mathcal{O}}
\safemath{\setP}{\mathcal{P}}
\safemath{\setQ}{\mathcal{Q}}
\safemath{\setR}{\mathcal{R}}
\safemath{\setS}{\mathcal{S}}
\safemath{\setT}{\mathcal{T}}
\safemath{\setU}{\mathcal{U}}
\safemath{\setV}{\mathcal{V}}
\safemath{\setW}{\mathcal{W}}
\safemath{\setX}{\mathcal{X}}
\safemath{\setY}{\mathcal{Y}}
\safemath{\setZ}{\mathcal{Z}}
\safemath{\emptySet}{\varnothing}

\safemath{\colA}{\mathscr{A}}
\safemath{\colB}{\mathscr{B}}
\safemath{\colC}{\mathscr{C}}
\safemath{\colD}{\mathscr{D}}
\safemath{\colE}{\mathscr{E}}
\safemath{\colF}{\mathscr{F}}
\safemath{\colG}{\mathscr{G}}
\safemath{\colH}{\mathscr{H}}
\safemath{\colI}{\mathscr{I}}
\safemath{\colJ}{\mathscr{J}}
\safemath{\colK}{\mathscr{K}}
\safemath{\colL}{\mathscr{L}}
\safemath{\colM}{\mathscr{M}}
\safemath{\colN}{\mathscr{N}}
\safemath{\colO}{\mathscr{O}}
\safemath{\colP}{\mathscr{P}}
\safemath{\colQ}{\mathscr{Q}}
\safemath{\colR}{\mathscr{R}}
\safemath{\colS}{\mathscr{S}}
\safemath{\colT}{\mathscr{T}}
\safemath{\colU}{\mathscr{U}}
\safemath{\colV}{\mathscr{V}}
\safemath{\colW}{\mathscr{W}}
\safemath{\colX}{\mathscr{X}}
\safemath{\colY}{\mathscr{Y}}
\safemath{\colZ}{\mathscr{Z}}

\safemath{\opA}{\mathbb{A}}
\safemath{\opB}{\mathbb{B}}
\safemath{\opC}{\mathbb{C}}
\safemath{\opD}{\mathbb{D}}
\safemath{\opE}{\mathbb{E}}
\safemath{\opF}{\mathbb{F}}
\safemath{\opG}{\mathbb{G}}
\safemath{\opH}{\mathbb{H}}
\safemath{\opI}{\mathbb{I}}
\safemath{\opJ}{\mathbb{J}}
\safemath{\opK}{\mathbb{K}}
\safemath{\opL}{\mathbb{L}}
\safemath{\opM}{\mathbb{M}}
\safemath{\opN}{\mathbb{N}}
\safemath{\opO}{\mathbb{O}}
\safemath{\opP}{\mathbb{P}}
\safemath{\opQ}{\mathbb{Q}}
\safemath{\opR}{\mathbb{R}}
\safemath{\opS}{\mathbb{S}}
\safemath{\opT}{\mathbb{T}}
\safemath{\opU}{\mathbb{U}}
\safemath{\opV}{\mathbb{V}}
\safemath{\opW}{\mathbb{W}}
\safemath{\opX}{\mathbb{X}}
\safemath{\opY}{\mathbb{Y}}
\safemath{\opZ}{\mathbb{Z}}
\safemath{\opZero}{\mathbb{O}}
\safemath{\identityop}{\opI}

\safemath{\veca}{\bma}
\safemath{\vecb}{\bmb}
\safemath{\vecc}{\bmc}
\safemath{\vecd}{\bmd}
\safemath{\vece}{\bme}
\safemath{\vecf}{\bmf}
\safemath{\vecg}{\bmg}
\safemath{\vech}{\bmh}
\safemath{\veci}{\bmi}
\safemath{\vecj}{\bmj}
\safemath{\veck}{\bmk}
\safemath{\vecl}{\bml}
\safemath{\vecm}{\bmm}
\safemath{\vecn}{\bmn}
\safemath{\veco}{\bmo}
\safemath{\vecp}{\bmmp}
\safemath{\vecq}{\bmq}
\safemath{\vecr}{\bmr}
\safemath{\vecs}{\bms}
\safemath{\vect}{\bmt}
\safemath{\vecu}{\bmu}
\safemath{\vecv}{\bmv}
\safemath{\vecw}{\bmw}
\safemath{\vecx}{\bmx}
\safemath{\vecy}{\bmy}
\safemath{\vecz}{\bmz}

\safemath{\veczero}{\bmzero}
\safemath{\vecone}{\bmone}
\safemath{\vecxi}{\bmxi}
\safemath{\veclambda}{\bmlambda}
\safemath{\vecmu}{\bmmu}
\safemath{\vectheta}{\bmtheta}
\safemath{\vecphi}{\bmphi}

\safemath{\matA}{\bA}
\safemath{\matB}{\bB}
\safemath{\matC}{\bC}
\safemath{\matD}{\bD}
\safemath{\matE}{\bE}
\safemath{\matF}{\bF}
\safemath{\matG}{\bG}
\safemath{\matH}{\bH}
\safemath{\matI}{\bI}
\safemath{\matJ}{\bJ}
\safemath{\matK}{\bK}
\safemath{\matL}{\bL}
\safemath{\matM}{\bM}
\safemath{\matN}{\bN}
\safemath{\matO}{\bO}
\safemath{\matP}{\bP}
\safemath{\matQ}{\bQ}
\safemath{\matR}{\bR}
\safemath{\matS}{\bS}
\safemath{\matT}{\bT}
\safemath{\matU}{\bU}
\safemath{\matV}{\bV}
\safemath{\matW}{\bW}
\safemath{\matX}{\bX}
\safemath{\matY}{\bY}
\safemath{\matZ}{\bZ}
\safemath{\matzero}{\bmzero}

\safemath{\matDelta}{\bDelta}
\safemath{\matLambda}{\bLambda}
\safemath{\matPhi}{\bPhi}
\safemath{\matSigma}{\bSigma}
\safemath{\matOmega}{\bOmega}
\safemath{\matTheta}{\bTheta}

\safemath{\matidentity}{\matI}
\safemath{\matone}{\matO}

\safemath{\rnda}{A}
\safemath{\rndb}{B}
\safemath{\rndc}{C}
\safemath{\rndd}{D}
\safemath{\rnde}{E}
\safemath{\rndf}{F}
\safemath{\rndg}{G}
\safemath{\rndh}{H}
\safemath{\rndi}{I}
\safemath{\rndj}{J}
\safemath{\rndk}{K}
\safemath{\rndl}{L}
\safemath{\rndm}{M}
\safemath{\rndn}{N}
\safemath{\rndo}{O}
\safemath{\rndp}{P}
\safemath{\rndq}{Q}
\safemath{\rndr}{R}
\safemath{\rnds}{S}
\safemath{\rndt}{T}
\safemath{\rndu}{U}
\safemath{\rndv}{V}
\safemath{\rndw}{W}
\safemath{\rndx}{X}
\safemath{\rndy}{Y}
\safemath{\rndz}{Z}

\safemath{\rveca}{\bimA}
\safemath{\rvecb}{\bimB}
\safemath{\rvecc}{\bimC}
\safemath{\rvecd}{\bimD}
\safemath{\rvece}{\bimE}
\safemath{\rvecf}{\bimF}
\safemath{\rvecg}{\bimG}
\safemath{\rvech}{\bimH}
\safemath{\rveci}{\bimI}
\safemath{\rvecj}{\bimJ}
\safemath{\rveck}{\bimK}
\safemath{\rvecl}{\bimL}
\safemath{\rvecm}{\bimM}
\safemath{\rvecn}{\bimN}
\safemath{\rveco}{\bomO}
\safemath{\rvecp}{\bimP}
\safemath{\rvecq}{\bimQ}
\safemath{\rvecr}{\bimR}
\safemath{\rvecs}{\bimS}
\safemath{\rvect}{\bimT}
\safemath{\rvecu}{\bimU}
\safemath{\rvecv}{\bimV}
\safemath{\rvecw}{\bimW}
\safemath{\rvecx}{\bimX}
\safemath{\rvecy}{\bimY}
\safemath{\rvecz}{\bimZ}

\safemath{\rvecxi}{\bmxi}
\safemath{\rveclambda}{\bmlambda}
\safemath{\rvecmu}{\bmmu}
\safemath{\rvectheta}{\bmtheta}
\safemath{\rvecphi}{\bmphi}

\safemath{\rmatA}{\bimA}
\safemath{\rmatB}{\bimB}
\safemath{\rmatC}{\bimC}
\safemath{\rmatD}{\bimD}
\safemath{\rmatE}{\bimE}
\safemath{\rmatF}{\bimF}
\safemath{\rmatG}{\bimG}
\safemath{\rmatH}{\bimH}
\safemath{\rmatI}{\bimI}
\safemath{\rmatJ}{\bimJ}
\safemath{\rmatK}{\bimK}
\safemath{\rmatL}{\bimL}
\safemath{\rmatM}{\bimM}
\safemath{\rmatN}{\bimN}
\safemath{\rmatO}{\bimO}
\safemath{\rmatP}{\bimP}
\safemath{\rmatQ}{\bimQ}
\safemath{\rmatR}{\bimR}
\safemath{\rmatS}{\bimS}
\safemath{\rmatT}{\bimT}
\safemath{\rmatU}{\bimU}
\safemath{\rmatV}{\bimV}
\safemath{\rmatW}{\bimW}
\safemath{\rmatX}{\bimX}
\safemath{\rmatY}{\bimY}
\safemath{\rmatZ}{\bimZ}

\safemath{\rmatDelta}{\bimDelta}
\safemath{\rmatLambda}{\bimLambda}
\safemath{\rmatPhi}{\bimPhi}
\safemath{\rmatSigma}{\bimSigma}
\safemath{\rmatOmega}{\bimOmega}
\safemath{\rmatTheta}{\bimTheta}

%% file: main_material/1_introduction.tex
\section{Introduction}
Linear recurrences offer an efficient, sub-quadratic alternative to softmax attention for sequence modeling~\citep{linear_attention,dao2024transformers,gdn,kda}. They summarize past tokens in a fixed-size state, avoiding the growing key-value cache of softmax attention, and are increasingly used in both standalone and hybrid language models~\citep{nemotron3,lfm2,rwkv7,xlstm}. For prefill, these models are typically evaluated using chunkwise-parallel algorithms, which process blocks of tokens in parallel through dense matrix multiplications while propagating a recurrent state between them~\citep{dao2024transformers,fla}. This mirrors the blockwise organization of efficient attention kernels such as FlashAttention and makes recurrent prefill naturally amenable to low-precision execution.\looseness-1

Low-precision attention has been studied extensively, with optimized integer and floating-point kernels now forming a mature part of the Transformer inference stack~\citep{sageattention,fa3}. In contrast, low-precision kernels for linear recurrences remain largely unexplored. Prior work has studied quantization of recurrent models~\citep{mambaquant,quamba2}, largely targeting the projections and activations surrounding the Mamba-2 recurrence. SSDi8, on the other hand, executes the recurrent computation in low precision, demonstrating an efficient INT8 implementation specific to Mamba-2~\citep{kimssdi8}. We instead study the space more generally, investigating low-precision computation across the design space of modern linear recurrent architectures.\looseness-1 

Executing sequence mixers in low-precision can come at the cost of performance. One powerful lever to mitigate this {is} \emph{functional invariances}. Such transformations preserve the full-precision operator while changing the numerics presented to a quantizer. {Figure~\ref{fig:write_error_sweep}} {in the appendix} illustrates this principle through write-factor absorption. We distinguish two broad classes. \emph{Mixer-agnostic} invariances act on the residual stream or individual linear layers and do not depend on the algebra of the sequence operator~\citep{smoothquant,quarot,spinquant}. Because these transformations are largely independent of the mixer, they transfer naturally to recurrent architectures and have already been adopted by methods such as MambaQuant and Quamba2~\citep{mambaquant,quamba2}. \emph{Mixer-specific} invariances instead rely on the sequence operator itself, and can be classified as either \emph{mixer-external} or \emph{mixer-internal}. The former materialize as changes of basis and can be applied outside of the sequence-mixer, while the latter are functions of a {recurrence's} algorithmic realization (Figure~\ref{fig:architecture-overview}). Transformer PTQ has exploited such invariances within softmax attention, for example through transformations of its query, key, and value representations~\citep{quarot,spinquant}. In contrast, the mixer-specific invariances of linear recurrences have yet to be characterized.\looseness-1

Linear recurrences present a qualitatively different invariance structure from softmax attention. Their transformations must remain consistent with a state that is updated and carried across chunks, while decay, gating, and delta-rule updates impose additional algebraic constraints. At the same time, these same mechanisms create new degrees of freedom that have no direct attention analogue. We therefore ask which attention-inspired invariances survive in recurrences, what new restrictions arise from the recurrent computation, and what additional invariances are unique to linear recurrences. Our main contributions are summarized as follows:

\begin{itemize}[leftmargin=15pt, labelindent=0pt, parsep=1pt, topsep=1pt, partopsep=1pt, itemsep=2pt]
\item We introduce LinQuant, a framework for computing low-precision linear recurrences based on both mixer-agnostic and mixer-specific functional invariances, and characterize their applicability across common recurrent architectures.\looseness-1
\item Across a controlled set of linear recurrences, we show how recurrence structure, query-key normalization, quantization scale granularity, and state-storage basis determine which functionally equivalent realization is most robust to low precision.
\item We implement LinQuant in INT8 chunkwise kernels and fuse query--key transformations into SSDi8 without increasing measured prefill latency.
\item We apply LinQuant to production W4A8 models with INT8 recurrent states, improving over SSDi8 on Mamba-2 {1.3B, 2.7B, and 8B} and Nemotron-H 8B.
\end{itemize}

%% file: main_material/2_background_and_preliminaries.tex
\begin{table}[t!]
\centering
\caption{Applicability of the seven invariances introduced in Section~\ref{sec:method} to common linear recurrences. \(\times\): not applicable. \(\checkmark\): applicable. \((\checkmark)\): applicable under constraints.}
\label{tab:predictions}
\footnotesize
\setlength{\tabcolsep}{4pt}
\begin{tabular}{@{}r@{\hspace{0.65em}}l@{\hspace{1.4em}}cccccc@{}}
\toprule
\multicolumn{1}{@{}r}{Nr.} & Transformation & LA & Mamba-2 & GLA & DN & GDN & KDA \\
\midrule
(1/7) & \(\RotTerm{\mat R_{QK}}\) & \(\checkmark\) & \(\checkmark\) & \((\checkmark)\) & \(\checkmark\) & \(\checkmark\) & \((\checkmark)\) \\
(2/7) & \(\RescaleTerm{\mat D_{QK}}\) & \(\checkmark\) & \(\checkmark\) & \(\checkmark\) & \(\times\) & \(\times\) & \(\times\) \\
(3/7) & \(\RotTerm{\mat R_V}\) & \(\checkmark\) & \(\checkmark\) & \(\checkmark\) & \(\checkmark\) & \(\checkmark\) & \(\checkmark\) \\
(4/7) & \(\RescaleTerm{\mat D_V}\) & \(\checkmark\) & \(\checkmark\) & \(\checkmark\) & \(\checkmark\) & \(\checkmark\) & \(\checkmark\) \\
\addlinespace[3pt]
(5/7) & \(\big(\DecayTerm{(\Dtail_{[i]})^\eta},\;\DecayTerm{(\Dtail_{[i]})^{1-\eta}}\big)\) & \(\times\) & \(\checkmark\) & \((\checkmark)\) & \(\times\) & \(\checkmark\) & \((\checkmark)\) \\[3pt]
(6/7) & \(\big(\DecayTerm{\mat E_{[i]}^\eta},\;\DecayTerm{\mat E_{[i]}^{1-\eta}}\big)\) & \(\times\) & \(\checkmark\) & \(\checkmark\) & \(\times\) & \(\checkmark\) & \(\checkmark\) \\[3pt]
(7/7) & \(\big(\WriteGateTerm{\mat B_{[i]}^\xi},\;\WriteGateTerm{\mat B_{[i]}^{1-\xi}}\big)\) & \((\checkmark)\) & \((\checkmark)\) & \((\checkmark)\) & \(\checkmark\) & \(\checkmark\) & \(\checkmark\) \\
\bottomrule
\end{tabular}
\end{table}

\section{Background and preliminaries}
\label{sec:background}

\subsection{Linear Recurrences}
\label{sec:sequence-mixers}

Causal softmax attention~\citep{attention} computes
\begin{equation*}
    \bmo_t
    =
    \bV_{1:t}^{\top}
    \operatorname{softmax}
    \left(
        \bK_{1:t}\bmq_t/\sqrt{D_k}
    \right),
\end{equation*}
from queries and keys
$\mathbf q_t,\mathbf k_t\in\mathbb R^{D_k}$
and values
$\mathbf v_t\in\mathbb R^{D_v}$,
where $D_k$ and $D_v$ denote the \emph{state} (key/query) and
\emph{value} dimensions, respectively.
Linear attention replaces the nonlinear softmax operator with an
unnormalized linear kernel~\citep{linear_attention}. In its simplest form,
associativity allows past keys and values to be summarized in a fixed-size
matrix-valued state
$\bS_t\in\mathbb R^{D_v\times D_k}$,
\begin{equation*}
    \bS_t
    =
    \bS_{t-1}
    +
    \bmv_t\bmk_t^\top,
    \qquad
    \bmo_t
    =
    \bS_t\bmq_t.
\end{equation*}

Modern linear recurrences extend this basic update with decay, write gating, and delta-rule corrections~\citep{dao2024transformers,gla,yang2024gated,kda}. We capture these axes through the common recurrence
\begin{equation}
\label{eq:zoo-recurrence}
    \bS_t
    =
    \bS_{t-1}\,
    \DecayTerm{\mathbf \Lambda_t}\,
    \DeltaTerm{\bigl(\bI-\delta\,\WriteGateTerm{\beta_t}\,\bmk_t\bmk_t^\top\bigr)}
    +
    \WriteGateTerm{\beta_t}\,\bmv_t\bmk_t^\top,
\end{equation}
where {$\WriteGateTerm{\beta_t}>0$ ($\WriteGateTerm{\beta_t}\in(0,2)$ for the delta rule), $\|\DecayTerm{\mat\Lambda_t}\|\leq 1$}, and
$\delta\in\{0,1\}$ is fixed for a given recurrence.
Equation~\ref{eq:zoo-recurrence} separates three structural choices.

\DecayTerm{\textbf{Decay}} controls how the previous state is forgotten before new information is written. A scalar gate $\DecayTerm{\mathbf\Lambda_t=\alpha_t\mathbf I}$
uses one decay rate per recurrent head~\citep{dao2024transformers,gdn},
whereas a diagonal gate $\DecayTerm{\mathbf\Lambda_t=\operatorname{diag}(\bm\alpha_t)}$ allows the decay to vary across state channels~\citep{gla,kda}.
\WriteGateTerm{\textbf{Write gating}} scales the new association $\bmv_t\bmk_t^\top$. \DeltaTerm{\textbf{Delta correction}}~\citep{deltarule,deltanet}
removes the component of the existing state aligned with the current key before writing the new association. Together, these choices organize the space of modern recurrences as shown in Table~\ref{tab:mixer-zoo} in Appendix~\ref{app:zoo-details}. We later instantiate this taxonomy as a controlled \emph{mixer zoo}, varying the recurrent update while holding the surrounding model architecture and training setup fixed.

\subsection{Chunkwise-parallel computation}
\label{sec:background-chunkwise}

Chunkwise implementations parallelize Equation~\eqref{eq:zoo-recurrence} within chunks of \(C\) tokens~\citep{dao2024transformers,fla,pdn}, while processing chunks sequentially. Chunk \([i]\) contains positions \(iC+1,\ldots,(i+1)C\), with incoming state \(\mat S_{[i]}:=\mat S_{iC}\), queries/keys \(\mat Q_{[i]},\mat K_{[i]}\in\R^{C\times D_k}\), and values/outputs \(\mat V_{[i]},\mat O_{[i]}\in\R^{C\times D_v}\). Indices \(r,s\) denote positions within chunk \([i]\), and \(n\) denotes a channel. For example, \(\DecayTerm{\alpha_{[i]}^r}:=\DecayTerm{\alpha_{iC+r}}\) and \(\WriteGateTerm{\beta_{[i]}^r}:=\WriteGateTerm{\beta_{iC+r}}\). We collect the write gates in \({\mat B_{[i]}:=\diag(\beta_{[i]}^1,\ldots,\beta_{[i]}^C)}\in\R^{C\times C}\). Throughout this work, we also refer to matrix multiplications as ``contractions''. In the following, we lay out chunkwise parallel formulations for the recurrences covered by Equation~\eqref{eq:zoo-recurrence}.

\DecayTerm{\textbf{Scalar decay.}}
For no delta correction ($\delta = 0$) and scalar decay \(\DecayTerm{\mat\Lambda_t=\alpha_t\mat I}\), define the cumulative decay \(\DecayTerm{\gamma_{[i]}^{s\to r}}:=\DecayTerm{\prod_{u=s+1}^{r}\alpha_{[i]}^u}\) and \(\DecayTerm{\gamma_{[i]}^r}:=\DecayTerm{\gamma_{[i]}^{0\to r}}\). The prefix and tail vectors \(\DecayTerm{\Dpref_{[i]}},\DecayTerm{\Dtail_{[i]}}\in\R^C\) have entries \(\DecayTerm{(\Dpref_{[i]})_r}=\DecayTerm{\gamma_{[i]}^r}\) and \(\DecayTerm{(\Dtail_{[i]})_r}=\DecayTerm{\gamma_{[i]}^{r\to C}}\). {Define the causal decay matrix \(\mat\Gamma_{[i]}\in\R^{C\times C}\), with entries \((\mat\Gamma_{[i]})_{rs}=\gamma_{[i]}^{s\to r}\) for \(s\le r\) and zero otherwise.} The chunkwise parallel realization then reads~\citep{dao2024transformers}
\begin{align*}
    \mat S_{[i+1]}&=\DecayTerm{\gamma_{[i]}^C}\mat S_{[i]}+\mat V_{[i]}^\top\WriteGateTerm{\mat B_{[i]}}(\DecayTerm{\Dtail_{[i]}}\odot\mat K_{[i]}),\\
    \mat O_{[i]}&=(\DecayTerm{\Dpref_{[i]}}\odot\mat Q_{[i]})\mat S_{[i]}^\top+{((\mat Q_{[i]}\mat K_{[i]}^\top)\odot\mat\Gamma_{[i]})}\WriteGateTerm{\mat B_{[i]}}\mat V_{[i]}.
\end{align*}
\DecayTerm{\textbf{Diagonal decay.}}
In the absence of delta correction ($\delta=0$) but with diagonal decay  \(\DecayTerm{\mat\Lambda_t=\diag(\bb\alpha_t)}\), define a channel-wise diagonal decay \(\DecayTerm{\gamma_{[i],n}^{s\to r}}:=\DecayTerm{\prod_{u=s+1}^{r}\alpha_{[i],n}^u}\) and \(\DecayTerm{\gamma_{[i],n}^r}:=\DecayTerm{\gamma_{[i],n}^{0\to r}}\). The stacked prefix- and tail-decay now {are} \(\DecayTerm{\Dpref_{[i]}},\DecayTerm{\Dtail_{[i]}}\in\R^{C\times D_k}\), with entries \(\DecayTerm{(\Dpref_{[i]})_{rn}}=\DecayTerm{\gamma_{[i],n}^r}\) and \(\DecayTerm{(\Dtail_{[i]})_{rn}}=\DecayTerm{\gamma_{[i],n}^{r\to C}}\). {For positive decay gates,} the chunkwise parallel form then reads~\citep{gla,kda}
\begin{align*}
    \mat S_{[i+1]}&=\mat S_{[i]}\DecayTerm{\diag(\bm{\gamma}_{[i]}^C)}+(\WriteGateTerm{\mat B_{[i]}}\mat V_{[i]})^\top(\DecayTerm{\Dtail_{[i]}}\odot\mat K_{[i]}),\\
    \mat O_{[i]}&=(\DecayTerm{\Dpref_{[i]}}\odot\mat Q_{[i]})\mat S_{[i]}^\top
    +\tril((\DecayTerm{\Dpref_{[i]}}\odot\mat Q_{[i]})
    ({\mat K_{[i]}}\oslash{\DecayTerm{\Dpref_{[i]}}})^\top)
    \WriteGateTerm{\mat B_{[i]}}\mat V_{[i]}.
\end{align*}
\DeltaTerm{\textbf{Delta correction.}}
For delta correction ($\delta=1$) without decay, the chunkwise parallel form reads~\citep{deltanet}
\begin{align*}
    \mat S_{[i+1]}&=\mat S_{[i]}+\DeltaTerm{{\mat F}_{[i]}}^\top\mat K_{[i]},
    \\
    \mat O_{[i]}&=\mat Q_{[i]}\mat S_{[i]}^\top+\tril(\mat Q_{[i]}\mat K_{[i]}^\top)\DeltaTerm{{\mat F}_{[i]}},
\end{align*}
where \({\mat F}_{[i]}\) are delta-corrected \emph{pseudo-values} generated by
\begin{equation*}
    \DeltaTerm{{\mat F}_{[i]}}
    =(\mat I+\DeltaTerm{\tril(
    \WriteGateTerm{\mat B_{[i]}}
    \mat K_{[i]}\mat K_{[i]}^\top,-1)})^{-1}
    \WriteGateTerm{\mat B_{[i]}}(\mat V_{[i]}-\DeltaTerm{\mat K_{[i]}}\mat S_{[i]}^\top).
\end{equation*}
Appendix~\ref{app:quantized-forms} gives the decaying delta forms~\citep{gdn,kda}.

%% file: main_material/3_method.tex
\begin{figure*}[t!]
\centering
\resizebox{\textwidth}{!}{%
\input{figures/figure_one}
}
\caption{Architectural overview of a recurrent model at three resolutions: the model, recurrent layer, and chunkwise recurrence. Axis transformations and decay/write-factor allocations change the operands exposed to quantization. Applicability depends on the recurrence, as detailed in Section~\ref{sec:method}. Style inspired by~\citep[Figure 1]{spinquant}.}
\label{fig:architecture-overview}
\end{figure*}
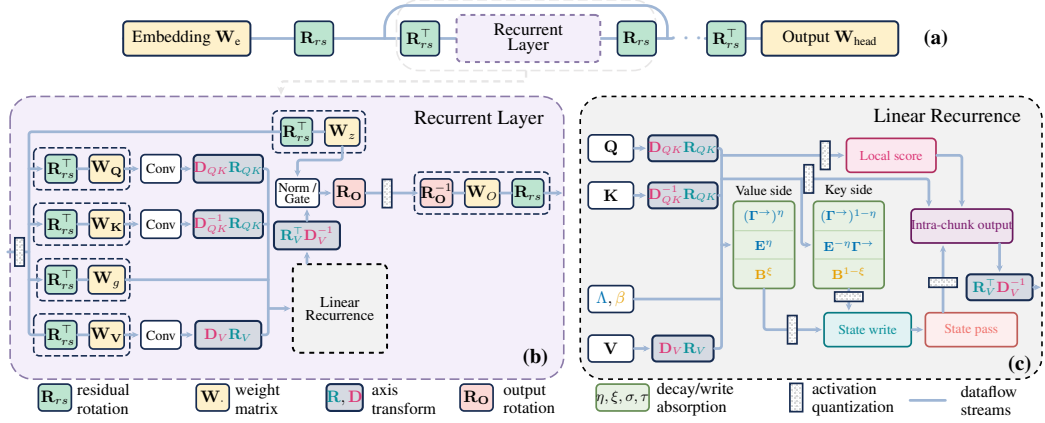

\section{LinQuant: Functional invariances for low-precision recurrences}
\label{sec:method}

The chunkwise parallel forms in Section~\ref{sec:background-chunkwise} expose a set of matrix multiplications and a state carried between chunks. Importantly, these chunkwise forms are not unique: the same full-precision recurrence can be expressed through functionally equivalent realizations that expose different quantities to low precision. LinQuant exploits this freedom. We first define the low-precision realization used throughout the paper, then derive seven functional invariances: four axis transformations and three absorptions. Table~\ref{tab:predictions} summarizes their applicability; proofs are in Appendix~\ref{app:invariance-proofs}.

\subsection{Low-precision recurrence realization}
\label{sec:method:quantization}

Across the recurrences in Table~\ref{tab:mixer-zoo}, we find that the chunkwise computation can be organized into four matrix contractions: a state write, state read, local score, and intra-chunk output. Together with a recurrent state pass between chunks, this decomposition defines the low-precision boundaries we study. Consider for example the recurrence underlying Mamba-2. Using the notation of Section~\ref{sec:background-chunkwise} and {writing $\mathcal Q$ for quantization, with scale factors implicit}, we consider the following low-precision chunkwise realization:
\begin{equation*}
\begin{aligned}
    \widehat{\mat H}_{[i]}
    &=\mathcal Q(\WriteGateTerm{\mathbf B_{[i]}}\mat V_{[i]})^\top
      \mathcal Q(\DecayTerm{\Dtail_{[i]}}\odot\mat K_{[i]}),
    &&\text{(state write)},\\
    \widehat{\mat S}_{[i+1]}
    &= {\mathcal Q(\gamma_{[i]}^C\mat S_{[i]}+\widehat{\mat H}_{[i]})},
    &&\text{(state pass)},\\
    \widehat{\mat O}^{\mathrm{inter}}_{[i]}
    &={\DecayTerm{\Dpref_{[i]}}\odot\big(\mathcal Q(\mat Q_{[i]})\widehat{\mat S}_{[i]}^\top\big)},
    &&\text{(state read)},\\
    \widehat{\mat A}_{[i]}
    &=(\mathcal Q(\mat Q_{[i]})
      \mathcal Q(\mat K_{[i]})^\top)
      \odot\DecayTerm{\mat\Gamma_{[i]}},
    &&\text{(local score)},\\
    \widehat{\mat O}^{\mathrm{intra}}_{[i]}
    &=\widehat{\mat A}_{[i]}
      \mathcal Q(\WriteGateTerm{\mathbf B_{[i]}}\mat V_{[i]}),
    &&\text{(intra-chunk output)}.
\end{aligned}
\end{equation*}
For models that employ the delta-rule, the Gram and WY solve are executed in full precision. {The incoming-state correction used to form pseudo-values is an additional quantized contraction.} Delta-rule writes and intra-chunk outputs replace the gated values $\WriteGateTerm{\mathbf B_{[i]}}\mathbf V_{[i]}$ by the pseudo-values $\mat F_{[i]}$ computed for the corresponding recurrence. Appendix~\ref{app:quantized-forms} provides more details.

LinQuant exploits the fact that functional invariances of these operators generally do not commute with quantization $\mathcal Q$. These functional invariances fall into two classes: \emph{changes of basis}, which change the representation along the key or value axes, and \emph{absorptions}, which redistribute decay or write factors within an equivalent chunkwise computation. Appendix~\ref{app:boundaries} studies sensitivity to the stored-state and write-side quantization boundaries.

\subsection{{Mixer-external invariances: changes of basis}}
\label{sec:change-of-basis}

Both the key- and the value axis of the recurrent state $\mat S_t\in\mathbb{R}^{D_v\times D_k}$ expose a choice of basis which leaves the full precision recurrence unchanged. We consider two such transformations along each axis: orthogonal rotations and positive diagonal rescalings. Notably, changes of basis along the key axis must remain compatible with the recurrent dynamics, whereas the value axis admits the same equivariance across all recurrences {in Equation~\eqref{eq:zoo-recurrence}}.

\textbf{(1/7) Key-side rotation.}
A joint orthogonal rotation of queries and keys,
\[
    (\bb q_t,\bb k_t)
    \rightarrow
    (\RotTerm{\mat R_{QK}}\bb q_t,\RotTerm{\mat R_{QK}}\bb k_t),
    \qquad
    \RotTerm{\mat R_{QK}}\in\mathcal O(D_k),
\]
preserves query/key inner products and is a standard invariance of full precision softmax attention exploited by rotation-based PTQ methods such as QuaRot and SpinQuant~\citep{quarot,spinquant}. In low precision, random rotations spread energy across channels, reducing outlier concentration and improving quantization robustness.

For a linear recurrence, the invariance holds if $\RotTerm{\mat R_{QK}}\DecayTerm{\mat\Lambda_t} = \DecayTerm{\mat\Lambda_t}\RotTerm{\mat R_{QK}}$~\citep{nazari2026key}, and implicitly applies a change of basis to the {state's} key axis, which transforms like $\mat S_t\rightarrow \mat S_t\RotTerm{\mat R_{QK}^{\top}}$. Invariance thus holds with no decay or scalar decay. With diagonal decay, a general rotation does not commute with $\DecayTerm{\mat\Lambda_t}$, so the recurrent state cannot be carried in a globally rotated basis. {However, we can still rotate queries and keys together after applying their decay factors when computing within-chunk scores. The recurrent state remains unrotated (Appendix~\ref{app:quantized-forms}).}

In this work, we use random signed Hadamard rotations. Recurrent models may apply per-channel causal convolutions and nonlinearities between their input projections and the recurrence, which generally do not commute with a dense rotation. We thus apply $\RotTerm{\mat R_{QK}}$ online after these operations.

\textbf{(2/7) Key-side rescaling.}
The key axis also admits a reciprocal diagonal transformation,
\[
    (\bb q_t,\bb k_t)
    \rightarrow
    (\RescaleTerm{\mat D_{QK}}\bb q_t,\RescaleTerm{\mat D_{QK}^{-1}}\bb k_t),
    \qquad
    \RescaleTerm{\mat D_{QK}}\succ 0
    \text{ diagonal},
\]
which likewise preserves query-key inner products. Related rescalings underlie SmoothAttention in QServe and, more broadly, methods such as SmoothQuant~\citep{qserve,smoothquant}.

For a linear recurrence, the state implicitly transforms like $\mat S_t\rightarrow \mat S_t\RescaleTerm{\mat D_{QK}^{-1}}$. Unlike a dense rotation, $\RescaleTerm{\mat D_{QK}}$ commutes with diagonal decay, making the transformation valid for linear writes under no, scalar, or diagonal decay. It does not, however, preserve delta corrected writes.

$\RescaleTerm{\mathbf D_{QK}}$ transports magnitude between queries and keys. In this work, we parameterize $\RescaleTerm{\mat D_{QK}}=\diag(\bb d)$ by $d_n=(\max_t |k_{t,n}|/\max_t |q_{t,n}|)^\kappa$, where the maxima are computed over calibration tokens within each head. The exponent $\kappa\in[0,1]$ controls the rescaling strength. Specifically, $\kappa=\tfrac12$ balances the magnitudes of queries and keys. We leave other parametrizations to future work.

\textbf{(3/7) Value-side rotation.}
The value axis admits an analogous orthogonal transformation,
\[
    \bb v_t\rightarrow \RotTerm{\mat R_V}\bb v_t,
    \qquad
    \RotTerm{\mat R_V}\in\mathcal O(D_v).
\]
In softmax attention, value rotations are used for KV-cache quantization~\citep{spinquant}. The operator, however, is not invariant to value rotations, which have to be undone after the softmax. Linear recurrences exhibit the same equivariance. Intuitively, this is because the recurrence acts independently along the value axis, which is implicitly rotated like $\mathbf S_t \to \RotTerm{\mathbf R_V} \mathbf S_t$.

We note that, compared to softmax attention, the recurrent setting reveals another degree of freedom. The state can, in between chunks, either be stored in its canonical basis, $\mathcal Q(\mathbf S_t)$, or in a rotated basis, $\mathcal Q(\RotTerm{\mathbf R_V} \mathbf S_t)$. These realizations are equivalent in full precision but expose different representations to $\mathcal Q$. Rotated storage is what the rotation produces by itself, whereas canonical storage returns each chunk's write to the original basis before it joins the carried state. This comes at the cost of an inverse transform per chunk, and keeps the state in the basis whose per-channel scales fit it.

\textbf{(4/7) Value-side rescaling.}
Finally, the value axis admits the positive diagonal analogue
\[
    \bb v_t\rightarrow \RescaleTerm{\mat D_V}\bb v_t,
    \qquad
    \RescaleTerm{\mat D_V}\succ 0
    \text{ diagonal}.
\]
The recurrent state transforms accordingly as $\mat S_t \rightarrow \RescaleTerm{\mat D_V}\mat S_t$, {with the inverse rescaling applied to the recurrence output before downstream normalization or other operations that do not commute with it. If a linear output projection follows the recurrence directly, it can absorb the inverse rescaling.} As with $\RotTerm{\mat R_V}$, this transformation is valid across all recurrence families considered here. 


In this work, we parametrize $\RescaleTerm{\mat D_V}=\diag(\bb d)$ by $d_p=\bigl((\prod_{j=1}^{D_v}\max_t |v_{t,j}|)^{1/D_v} /\max_t |v_{t,p}|\bigr)^\kappa$, where the maxima are computed over calibration tokens and the product runs over value channels within each head. This parametrization moves range between value channels. At $\kappa=1$, the ranges of the {value channels} are equalized.

\subsection{{Mixer-internal invariances: absorptions}}

The first four invariances discussed in the previous section change the basis along the two axes of the recurrent state. The remaining three redistribute multiplicative factors within a chunkwise parallel implementation of the recurrence. In particular, decay and write factors can often be divided between the two sides of a matrix product without changing the full-precision operator. 

\textbf{(5/7) Scalar-decay absorption.}
For scalar decay, the tail-decay $\DecayTerm{\Dtail_{[i]}}$ can be split between the value and key sides of the state write:
\[
    \mat H_{[i]}
    =
    (
        \DecayTerm{(\Dtail_{[i]})^\eta}
        \odot
        (\WriteGateTerm{\mat B_{[i]}}\mat V_{[i]})
    )^\top
    (
        \DecayTerm{(\Dtail_{[i]})^{1-\eta}}
        \odot
        \mat K_{[i]}
    ),
    \qquad
    \eta\in[0,1].
\]
Thus, $\eta$ continuously interpolates between equivalent realizations of the same write. At $\eta=0$, all tail decay remains on the keys, which we call \emph{key-side decay absorption}. At $\eta=1$, it moves entirely to the values, giving \emph{value-side decay absorption}, the endpoint used by SSDi8~\citep{kimssdi8}. Intermediate values split the decay between the two sides. The same split applies to delta-rule writes, replacing $\WriteGateTerm{\mat B_{[i]}}\mat V_{[i]}$ by the corresponding pseudo-values $\mat F_{[i]}$.

\textbf{(6/7) Diagonal-decay absorption.}
Diagonal decay varies across state channels and therefore cannot, in general, be transferred entirely to the value side. Nevertheless, we can factor each token's decay into a scalar magnitude and a relative channelwise pattern. This decomposition is not unique: any positive scalar factor yields an exact redistribution. Here, we use the channelwise maximum, and leave alternative choices to future work.

For diagonal tail decay $\DecayTerm{\Dtail_{[i]}}$, define the per-token scalar magnitude and normalized decay
\[
    \DecayTerm{e_{[i]}^r}
    :=
    \max_n
    \DecayTerm{(\Dtail_{[i]})_{rn}},
    \qquad
    \DecayTerm{\mat E_{[i]}}
    :=
    \DecayTerm{\diag(e_{[i]}^1,\ldots,e_{[i]}^C)}, \qquad \DecayTerm{\overline{\mat D}^{\mathrm{tail}}_{[i]}}
    :=
    \DecayTerm{\mat E_{[i]}^{-1}\Dtail_{[i]}}.
\]
Then the state write can be expressed as
\[
    \mat H_{[i]}
    =
    (
        \DecayTerm{\mat E_{[i]}^\eta}
        \WriteGateTerm{\mat B_{[i]}}\mat V_{[i]}
    )^\top
    (
        \DecayTerm{\mat E_{[i]}^{1-\eta}
        \overline{\mat D}^{\mathrm{tail}}_{[i]}}
        \odot
        \mat K_{[i]}
    ),
    \qquad
    \eta\in[0,1].
\]
Here, $\eta$ moves the magnitude of the diagonal decay. At $\eta=0$, the standard key-side form is recovered. At $\eta=1$, each token's scalar decay magnitude moves to the value side, while the relative channel-dependent decay necessarily remains on the key.

\textbf{(7/7) Write-factor absorption.}
For a linear write, the write factor admits an analogous split:
\[
    \mat H_{[i]}
    =
    (
        \WriteGateTerm{\mat B_{[i]}^\xi} \mat V_{[i]}
    )^\top
    (
        \DecayTerm{\Dtail_{[i]}}
        \odot
        \WriteGateTerm{\mat B_{[i]}^{1-\xi}}\mat K_{[i]}
    ),
    \qquad
    \xi\in[0,1].
\]
The endpoints $\xi=1$ and $\xi=0$ place the write factor entirely on the value and key sides, respectively.

For delta-corrected writes, the write gate also enters the triangular system used to construct the pseudo-values. Without decay, split it between the two key factors of the Gram matrix:
\[
    \mat L_{[i]}^{(\sigma)}
    :=
    \operatorname{tril}
    (
        (\WriteGateTerm{\mat B_{[i]}^\sigma}\mat K_{[i]})
        (\WriteGateTerm{\mat B_{[i]}^{1-\sigma}}\mat K_{[i]})^\top,
        -1
    ),
    \qquad
    \sigma\in[0,1].
\]
Using $\WriteGateTerm{\mat B_{[i]}^\sigma}$ on the corresponding right-hand side yields transformed pseudo-values $\mat F_{[i]}^{(\sigma)}$ satisfying
\[
    \mat F_{[i]}^{(\sigma)}
    =
    \WriteGateTerm{\mat B_{[i]}^{\sigma-1}}\mat F_{[i]},
\]
where $\mat F_{[i]}$ denotes the pseudo-values produced by the standard formulation in Section~\ref{sec:background-chunkwise}. A second parameter $\tau$ splits the complementary factor $\WriteGateTerm{\mat B_{[i]}^{1-\sigma}}$ between the transformed write operands:
\begin{align*}
    \mat V_{[i]}^{(\sigma,\tau)}
    &=\WriteGateTerm{\mat B_{[i]}^{(1-\sigma)(1-\tau)}}\mat F_{[i]}^{(\sigma)},
    \qquad
    \mat K_{[i]}^{(\sigma,\tau)}
    =\WriteGateTerm{\mat B_{[i]}^{(1-\sigma)\tau}}\mat K_{[i]},\\
    \mat H_{[i]}
    &=\bigl(\mat V_{[i]}^{(\sigma,\tau)}\bigr)^\top\mat K_{[i]}^{(\sigma,\tau)}
    =\mat F_{[i]}^\top\mat K_{[i]},
    \qquad \tau\in[0,1].
\end{align*}
Thus, $\sigma$ determines how the write factor is distributed inside the Gram construction, while $\tau$ determines how the remaining compensation is distributed in the subsequent write. At $\tau=0$, the compensation remains on the pseudo-values. At $\tau=1$, it moves to the keys. Setting $\sigma=1$ recovers the standard triangular formulation, while $\sigma=1/2$ places square-root write factors on the two Gram operands. Appendix~\ref{app:invariance-proofs} proves these equivalences and extends the construction to decayed delta-rule recurrences.



Together, the four changes of basis and three absorptions define seven functional degrees of freedom for realizing the same recurrent computation. Their applicability depends on the recurrence structure, while their numerical behavior depends on the resulting low-precision representation, which we explore in the following section.

%% file: figures/figure_one.tex
\definecolor{navy}{HTML}{172C51}
\definecolor{cream}{HTML}{FEF1CB}
\definecolor{rotpurple}{HTML}{D3A6D3}
\definecolor{onlineblue}{HTML}{D6DAE2}
\definecolor{lightdash}{HTML}{E7E6E6}
\definecolor{textgray}{HTML}{777777}
\definecolor{rotline}{HTML}{D3A6D3}
\definecolor{actline}{HTML}{B2AED2}
\definecolor{qkline}{HTML}{9FB6D5}
\definecolor{origline}{HTML}{CFCFCF}
\definecolor{kvhatch}{HTML}{A3B0C4}
\definecolor{absorbgreen}{HTML}{7FB069}
\definecolor{checkA}{HTML}{F6F6F6}
\definecolor{checkB}{HTML}{D7D7D7}
\definecolor{checkC}{HTML}{B5BDC6}
\definecolor{lightpurple}{HTML}{F4EFFA}
\definecolor{darkpurple}{HTML}{8B7AA8}

\tikzset{
  every node/.style={text=black},
  smallnode/.style={draw=navy, line width=.36mm, rounded corners=.9mm, align=center,
    inner sep=0pt, minimum height=6.1mm, font=\fontsize{9.7}{10.1}\selectfont},
  opnode/.style={smallnode, fill=cream},
  matchnote/.style={text=black, font=\fontsize{10.6}{11.0}\selectfont},
  rnode/.style={smallnode, fill=highlightcolor!50},
  rnode2/.style={smallnode, fill=highlightcolor3!50},
  qknode/.style={smallnode, fill=onlineblue, solid, line width=.44mm},
  absorbnode/.style={smallnode, fill=absorbgreen!18, draw=absorbgreen!80!black},
  opplain/.style={smallnode, fill=white},
  kernel/.style={smallnode, minimum height=7.2mm, font=\fontsize{8.2}{8.5}\selectfont},
  kstate/.style={kernel, fill=chunkstatecolor!12, draw=chunkstatecolor!85!black},
  kbmm/.style={kernel, fill=chunkbmmcolor!12, draw=chunkbmmcolor!80!black},
  kpass/.style={kernel, fill=chunkstatepassingcolor!12, draw=chunkstatepassingcolor!80!black},
  kscan/.style={kernel, fill=chunkscancolor!12, draw=chunkscancolor!85!black},
  kcumsum/.style={kernel, fill=chunksumcolor!12, draw=chunksumcolor!85!black},
  topnet/.style={draw=navy, line width=.42mm, rounded corners=1.2mm, fill=white,
    align=center, inner sep=0pt, font=\fontsize{10.6}{11.0}\selectfont},
  topwide/.style={draw=navy, line width=.42mm, rounded corners=.9mm, fill=cream,
    align=center, inner sep=0pt, font=\fontsize{10.3}{10.8}\selectfont},
  mergebox/.style={draw=navy, dashed, dash pattern=on 1.35mm off .55mm,
    line width=.34mm, rounded corners=1.25mm, fill=none},
  outerbox/.style={draw=lightdash, dashed, dash pattern=on 2.1mm off .95mm,
    line width=.34mm, rounded corners=4.8mm, fill=none},
  flow/.style={-{Stealth[length=1.5mm,width=1.35mm]}, shorten >=.25mm,
    line width=.58mm, line cap=round, line join=round},
  rflow/.style={flow, rounded corners=.6mm},
  rflowplain/.style={line width=.58mm, line cap=round, line join=round, rounded corners=.6mm},
  flowplain/.style={line width=.58mm, line cap=round, line join=round},
  panel/.style={font=\fontsize{12.6}{13.0}\bfseries},
  note/.style={text=textgray, font=\fontsize{9.4}{9.8}\selectfont},
  tinynote/.style={text=textgray, align=center, font=\fontsize{7.4}{7.8}\selectfont},
  legtext/.style={align=left, anchor=west, font=\fontsize{8.3}{8.7}\selectfont},
  streamlabel/.style={anchor=east, font=\fontsize{8.7}{9.0}\selectfont},
  circop/.style={draw=black, fill=white, circle, line width=.38mm, minimum size=4.4mm, inner sep=0pt}
}

\newcommand{\actbar}[4]{%
  \pgfmathtruncatemacro{\nx}{max(int(#3/0.55),1)}%
  \pgfmathtruncatemacro{\ny}{max(int(#4/0.55),1)}%
  \pgfmathtruncatemacro{\mx}{max(int(#3/0.70),1)}%
  \pgfmathtruncatemacro{\my}{max(int(#4/1.10),1)}%
  \begin{scope}
    \clip (#1,#2) rectangle ++(#3,#4);
    \fill[white] (#1,#2) rectangle ++(#3,#4);
    \foreach \i in {0,...,\nx}{
      \foreach \j in {0,...,\ny}{
        \pgfmathsetmacro{\xx}{#1+0.33+\i*0.55}
        \pgfmathsetmacro{\yy}{#2+0.22+\j*0.55}
        \pgfmathtruncatemacro{\parity}{mod(\i+\j,2)}
        \ifnum\parity=0
          \fill[checkB] (\xx,\yy) rectangle ++(.32,.32);
        \else
          \fill[checkA] (\xx,\yy) rectangle ++(.32,.32);
        \fi
      }
    }
    \foreach \i in {0,...,\mx}{
      \foreach \j in {0,...,\my}{
        \pgfmathsetmacro{\xx}{#1+0.43+\i*0.70}
        \pgfmathsetmacro{\yy}{#2+0.48+\j*1.10}
        \fill[checkC,opacity=.32] (\xx,\yy) rectangle ++(.18,.18);
      }
    }
  \end{scope}
  \draw[navy, line width=.36mm] (#1,#2) rectangle ++(#3,#4);
}

\newcommand{\intbar}[4]{%
  \path[draw=navy, line width=.36mm, fill=white, pattern=north east lines, pattern color=kvhatch]
    (#1,#2) rectangle ++(#3,#4);
}

\begin{tikzpicture}[x=1mm,y=-1mm]
\begin{scope}[scale=1.0]
\path[use as bounding box] (0,0) rectangle (204.8,95.0);


\begin{scope}[xshift=7mm]

\draw[flowplain, color=qkline] (2.0,13.2) -- (118.5,13.2);
\draw[flowplain, color=qkline] (125.5,13.2) -- (162.0,13.2);

\node[topwide, minimum width=27.0mm, minimum height=7.3mm] at (15.5,13.2)
  {Embedding $\mat W_{\text{e}}$};

\node[rnode, minimum width=8.2mm, minimum height=7.3mm] at (42.5,13.2)
  {$\mat R_{rs}$};

\draw[outerbox] (60.0,5.0) rectangle (115.0,20.0) coordinate[midway] (abox);

\draw[flowplain, color=qkline, rounded corners=4.8mm]
  (57.5,13.2)--(57.5,6.2)--(117.5,6.2)--(117.5,13.2);

\node[rnode, minimum width=8.2mm, minimum height=7.3mm] at (65.0,13.2)
  {$\mat R_{rs}^\top$};

\node[topnet, minimum width=30.5mm, minimum height=10.0mm, draw=darkpurple, fill=lightpurple, dashed] at (88.0,13.2)
  {Recurrent\\[-.7mm]Layer};

\node[rnode, minimum width=8.2mm, minimum height=7.3mm] at (111.0,13.2)
  {$\mat R_{rs}$};

\node[text=qkline, font=\fontsize{16}{16}\selectfont] at (122.0,13.2)
  {$\cdots$};

\node[rnode, minimum width=8.2mm, minimum height=7.3mm] at (130.0,13.2)
  {$\mat R_{rs}^\top$};

\node[topwide, minimum width=29.0mm, minimum height=7.3mm] at (152.0,13.2)
  {Output $\mat {W}_{\text{head}}$};



\coordinate (aboxSouth) at (87.5, 20.0);

\node[panel] at (174.5,13.2) {(a)};

\end{scope}
\begin{scope}[shift={(-1,10)}]

\begin{scope}[xshift=-13mm]
\draw[outerbox, draw=darkpurple, fill=lightpurple] (-1.0,15.5) rectangle (116.5,75.5) coordinate[midway] (bbox);
\node[matchnote, anchor=east, font=\large] at (113.5,20.5) {Recurrent Layer};
\node[panel] at (110.5,71.0) {(b)};

\coordinate (bboxNorth) at (56.5, 15.5);

\draw[-{Stealth[length=1.8mm,width=1.6mm]}, shorten >=.3mm,
      color=lightdash, line width=.5mm,
      dashed, dash pattern=on 1.5mm off 1mm,
      rounded corners=.6mm]
  (aboxSouth) |- (72, 12.3) -| (bboxNorth);

\begin{scope}[xshift=-6mm, yshift=-5mm]

\draw[flowplain, color=qkline] (4.5,43.5) -- (9.0,43.5);

\draw[mergebox] (10.7,49.4-4.45) rectangle (30.4,49.4+4.45);
\node[rnode, minimum width=6.7mm, minimum height=6.0mm] (rGam) at (15.8,49.4) {$\mat R_{rs}^\top$};
\node[opnode, minimum width=7.8mm, minimum height=6.0mm] (wGam) at (25.5,49.4) {$\mat W_g$};
\draw[flow, color=qkline] (9,49.4)--(rGam.west);
\draw[flowplain, color=qkline] (rGam.east)--(wGam.west);

\foreach \y/\w/\lab in {25.85/W_Q/{\mat W_{\mat Q}},37.6/W_K/{\mat W_{\mat K}},61.1/W_V/{\mat W_{\mat V}}}{
  \draw[mergebox] (10.,\y-4.45) rectangle (30.4,\y+4.45);
  \node[rnode, minimum width=6.7mm, minimum height=6.0mm] (r\w) at (15.8,\y) {$\mat R_{rs}^\top$};
  \node[opnode, minimum width=7.4mm, minimum height=6.0mm] (w\w) at (25.5,\y) {$\lab$};
  \node[opplain, minimum width=8.5mm, minimum height=6.0mm, font=\fontsize{8.4}{8.7}\selectfont] (conv\w) at (37.0,\y) {Conv};
  \draw[flowplain, color=qkline] (r\w.east)--(w\w.west);
  \draw[flow, color=qkline] (w\w.east)--(conv\w.west);
}
\draw[flow, color=qkline] (9,25.85)--(rW_Q.west);
\draw[flow, color=qkline] (9,37.6)--(rW_K.west);

\node[qknode, minimum width=14.0mm, minimum height=5.9mm, font=\fontsize{8.7}{9}\selectfont] (qRotB) at (51.3,25.85) {$\RescaleTerm{\mat D_{QK}}\RotTerm{\mat R_{QK}}$};
\node[qknode, minimum width=14.0mm, minimum height=5.9mm, font=\fontsize{8.7}{9}\selectfont] (kRotB) at (51.3,37.6) {$\RescaleTerm{\mat D_{QK}^{-1}}\RotTerm{\mat R_{QK}}$};

\node[
  topnet,
  draw=black,
  dashed,
  minimum width=20.0mm,
  minimum height=18.8mm,
  fill=black!5,
  font=\fontsize{9.1}{9.5}\selectfont
] (mix) at (75.0,55.5)
{Linear\\[-.5mm]Recurrence};

\node[opplain, minimum width=9.2mm, minimum height=6.0mm, font=\fontsize{8.1}{8.4}\selectfont] (gNorm) at (66.0,31) {Norm /\\[-.8mm]Gate};

\draw[mergebox] (60.7, 18-4.45) rectangle (80.4, 18+4.45);
\node[rnode, minimum width=6.7mm, minimum height=6.0mm] (rInv) at (65.8, 18) {$\mat R_{rs}^\top$};
\node[opnode, minimum width=7.4mm, minimum height=6.0mm] (wZ) at (75.5, 18) {$\mat W_z$};

\draw[rflowplain, color=qkline]
  (rInv.west)--(9,18)--(9,61.1)--(rW_V.west);
\draw[flowplain, color=qkline] (rInv.east)--(wZ.west);
\draw[rflow, color=qkline]
  (wZ.south)--(75.5,24)--(66,24)--(gNorm.north);

\draw[mergebox] (90.8, 31-4.45) rectangle (119.5, 31+4.45);

\node[rnode2, minimum width=6.7mm, minimum height=6.0mm] (rIn2) at (95.5, 31) {$\mat R_{\mat O}^{-1}$};
\node[opnode, minimum width=7.4mm, minimum height=6.0mm] (wOut) at (105.2, 31) {$\mat W_O$};
\node[rnode, minimum width=6.7mm, minimum height=6.0mm] (rOut) at (114.9, 31) {$\mat R_{rs}$};

\node[rnode2, minimum width=6.7mm, minimum height=6.0mm] (rIn1) at (77.0, 31) {$\mat R_{\mat O}$};

\draw[flow, color=qkline] (gNorm.east) -- (rIn1.west);
\draw[flow, color=qkline] (rIn1.east) -- (rIn2.west);
\draw[flowplain, color=qkline] (rIn2.east)--(wOut.west);
\draw[flowplain, color=qkline] (wOut.east)--(rOut.west);
\draw[flow, color=qkline] (rOut.east) -- (122.5,31);

\draw[flow, color=qkline] (convW_Q.east) -- (qRotB.west);
\draw[flow, color=qkline] (convW_K.east) -- (kRotB.west);

\coordinate (mixMerge) at ($(mix.west)+(-5.0mm,0)$);

\node[qknode, minimum width=14.0mm, minimum height=5.9mm, font=\fontsize{8.7}{9}\selectfont] (vRotB) at (51.3, 61.1) {$\RescaleTerm{\mat D_V}\RotTerm{\mat R_V}$};

\draw[flow, color=qkline] (convW_V.east) -- (vRotB.west);
\draw[rflowplain, color=qkline]
  (qRotB.east)--(mixMerge |- qRotB.east)
  --(mixMerge |- vRotB.east)--(vRotB.east);
\draw[flowplain, color=qkline] (kRotB.east)--(mixMerge |- kRotB.east);
\draw[flowplain, color=qkline] (wGam.east)--(mixMerge |- wGam.east);

\draw[flow, color=qkline] (mixMerge) -- (mix.west);

\node[qknode, minimum width=14.0mm, minimum height=5.9mm, font=\fontsize{8.7}{9}\selectfont]
  (vRotInv) at (68.0,40.0) {$\RotTerm{\mat R_V^\top}\RescaleTerm{\mat D_V^{-1}}$};

\draw[flow, color=qkline] (mix.north -| vRotInv.south)--(vRotInv.south);
\draw[flow, color=qkline] (vRotInv.north)--(vRotInv.north |- gNorm.south);

\actbar{6.0}{40.5}{2.0}{6.0}
\actbar{84.0}{28.1}{2.0}{6.0}

\end{scope}
\end{scope}
\draw[outerbox, draw=black, dashed, fill=black!5]
  (107,15.5) rectangle (205.0,75.5);
\begin{scope}[xshift=-4mm]
\node[matchnote, anchor=east, font=\large] at (205.8,19.5) {Linear Recurrence};

\node[opplain, minimum width=9.6mm, minimum height=5.7mm] (qIn) at (117.5,26.5) {$\mat Q$};
\node[opplain, minimum width=9.6mm, minimum height=5.7mm] (kIn) at (117.5,36.5) {$\mat K$};
\node[opplain, minimum width=9.6mm, minimum height=5.7mm] (aIn) at (117.5,58.5) {$\DecayTerm{\Lambda},\WriteGateTerm{\beta}$};
\node[opplain, minimum width=9.6mm, minimum height=5.7mm] (vIn) at (117.5,69.0) {$\mat V$};
\node[qknode, minimum width=13mm, minimum height=5.5mm, font=\fontsize{8.7}{9}\selectfont]
  (qRot) at (132.8,26.5) {$\RescaleTerm{\mat D_{QK}}\RotTerm{\mat R_{QK}}$};
\node[qknode, minimum width=13mm, minimum height=5.5mm, font=\fontsize{8.7}{9}\selectfont]
  (kRot) at (132.8,36.5) {$\RescaleTerm{\mat D_{QK}^{-1}}\RotTerm{\mat R_{QK}}$};
\node[qknode, minimum width=13mm, minimum height=5.5mm, font=\fontsize{8.7}{9}\selectfont]
  (vRot) at (132.8,69.0) {$\RescaleTerm{\mat D_V}\RotTerm{\mat R_V}$};
\draw[flow,color=qkline] (qIn.east)--(qRot.west);
\draw[flow,color=qkline] (kIn.east)--(kRot.west);
\draw[flow,color=qkline] (vIn.east)--(vRot.west);

\node[font=\fontsize{8.3}{8.6}\selectfont] at (150,35.5) {Value side};
\node[font=\fontsize{8.3}{8.6}\selectfont] at (168,35.5) {Key side};
\node[absorbnode, minimum width=13mm, minimum height=18.5mm,
      font=\fontsize{8.3}{8.6}\selectfont] (vFactors) at (150,47) {};
\node[absorbnode, minimum width=15mm, minimum height=18.5mm,
      font=\fontsize{8.3}{8.6}\selectfont] (kFactors) at (168,47) {};
\node[font=\fontsize{8.3}{8.6}\selectfont] at (150,41) {$\DecayTerm{(\Dtail)^\eta}$};
\node[font=\fontsize{8.3}{8.6}\selectfont] at (168,41) {$\DecayTerm{(\Dtail)^{1-\eta}}$};
\node[font=\fontsize{8.3}{8.6}\selectfont] at (150,47) {$\DecayTerm{\mat E^\eta}$};
\node[font=\fontsize{8.3}{8.6}\selectfont] at (168,47) {$\DecayTerm{\mat E^{-\eta}\Dtail}$};
\node[font=\fontsize{8.3}{8.6}\selectfont] at (150,53) {$\WriteGateTerm{\mat B^\xi}$};
\node[font=\fontsize{8.3}{8.6}\selectfont] at (168,53) {$\WriteGateTerm{\mat B^{1-\xi}}$};
\draw[absorbgreen!55, line width=.18mm] (143.8,44)--(156.2,44);
\draw[absorbgreen!55, line width=.18mm] (160.8,44)--(175.2,44);
\draw[absorbgreen!55, line width=.18mm] (143.8,50)--(156.2,50);
\draw[absorbgreen!55, line width=.18mm] (160.8,50)--(175.2,50);

\draw[rflowplain,color=qkline]
  (qRot.east)--(141,26.5)--(141,69)--(vRot.east);
\draw[flowplain,color=qkline] (kRot.east)--(141,36.5);
\draw[flowplain,color=qkline] (aIn.east)--(141,58.5);
\draw[flow,color=qkline] (141,47)--(vFactors.west);
\draw[rflow,color=qkline] (158,33)--(158,47)--(kFactors.west);

\node[kbmm, minimum width=19.0mm] (bmm) at (177.0,28.0)
  {\textcolor{chunkbmmcolor}{Local score}};
\node[kstate, minimum width=18.5mm] (cstate) at (172.0,65.0)
  {\textcolor{chunkstatecolor}{State write}};
\node[kpass, minimum width=19.5mm] (spass) at (194.0,65.0)
  {\textcolor{chunkstatepassingcolor}{State pass}};
\node[kscan, minimum width=19.0mm] (scan) at (192.0,43.0)
  {\textcolor{chunkscancolor}{Intra-chunk output}};
\node[qknode, minimum width=14.0mm, minimum height=5.5mm, font=\fontsize{8.7}{9}\selectfont]
  (vRotInvc) at (200.0,56.0) {$\RotTerm{\mat R_V^\top}\RescaleTerm{\mat D_V^{-1}}$};

\draw[flow,color=qkline] (141,28)--(bmm.west);
\draw[rflow,color=qkline]
  (141,33)--(185,33)--(185,39.2);
\draw[rflow,color=qkline] (vFactors.south) |- (cstate.west);
\draw[flow,color=qkline] (kFactors.south)--(kFactors.south |- cstate.north);
\draw[rflow,color=qkline]
  (bmm.east)--(192,28)--(scan.north);
\draw[flow,color=qkline] (cstate.east)--(spass.west);
\draw[flow,color=qkline]
  (188,61.2)--(188,46.8);
\draw[flow,color=qkline]
  (scan.south -| vRotInvc.north)--(vRotInvc.north);
\draw[flow,color=qkline] (vRotInvc.east)--(209,56);

\actbar{162}{25}{2}{6}
\actbar{158.5}{30}{2}{6}
\actbar{155}{62}{2}{6}
\actbar{165}{57}{6}{2}
\actbar{185}{54}{6}{2}

\node[panel,anchor=east] at (208,72.5) {(c)};
\end{scope}
\end{scope}
\begin{scope}[xshift=-12mm, yshift=-16mm]
\node[rnode, minimum width=7.0mm, minimum height=5.7mm] at (6.5,73.4) {$\mat R_{rs}$};
\node[legtext, font=\normalsize] at (10,73.7) {residual\\rotation};
\node[opnode, minimum width=7.0mm, minimum height=5.7mm] at (40.0,73.4) {$\mat W_{\cdot}$};
\node[legtext, font=\normalsize] at (43.5,73.7) {weight \\ matrix};
\node[qknode, minimum width=7.0mm, minimum height=5.7mm] at (68.0,73.4) {$\RotTerm{\mat R},\RescaleTerm{\mat D}$};
\node[legtext, font=\normalsize] at (72.5,73.7) {axis \\transform};
\node[rnode2, minimum width=7.0mm, minimum height=5.7mm] at (96.0,73.4) {$\mat R_{\mat O}$};
\node[legtext, font=\normalsize] at (100.5,73.7) {output\\rotation};
\node[absorbnode, minimum width=9.8mm, minimum height=5.7mm, font=\small]
  at (127.0,73.4) {$\eta,\xi,\sigma,\tau$};
\node[legtext, font=\normalsize] at (133.0,73.7) {decay/write\\absorption};
\actbar{162.5}{70.4}{2.7}{6.0}
\node[legtext, font=\normalsize] at (166.0,73.7) {activation\\quantization};
\draw[flowplain, color=qkline] (188.0,74.0) -- (195.5,74.0);
\node[legtext, font=\normalsize] at (197.5,74.1) {dataflow \\ streams};
\end{scope}

\end{scope}

\end{tikzpicture}

%% file: main_material/4_experiments.tex
\begin{figure}[t!]
\centering
\begin{subfigure}[t]{0.245\linewidth}\vskip 0pt\centering\input{figures/pgf/fig_rqk_decay.pgf}\caption{Recurrence structure.}\label{fig:key_structure}\end{subfigure}\hfill
\begin{subfigure}[t]{0.245\linewidth}\vskip 0pt\centering\input{figures/pgf/fig_rqk_norm.pgf}\caption{QK norm.}\label{fig:key_norm}\end{subfigure}\hfill
\begin{subfigure}[t]{0.245\linewidth}\vskip 0pt\centering\input{figures/pgf/fig_geometry_key.pgf}\caption{Scale granularity.}\label{fig:key_geometry}\end{subfigure}\hfill
\begin{subfigure}[t]{0.245\linewidth}\vskip 0pt\centering\input{figures/pgf/fig_sweep_kappa.pgf}\caption{Rescaling exponent.}\label{fig:key_kappa}\end{subfigure}
\caption{{Key-axis transformations at INT6.} (\subref{fig:key_structure})~{Recovered excess NLL by decay and write rule}. (\subref{fig:key_norm})~Excess NLL with and without QK normalization ($\dagger$). Panels (\subref{fig:key_structure},\subref{fig:key_norm}) use $\langle\mathrm{T,C}\rangle$ and $\kappa=1/2$ for $\RescaleTerm{\mat D_{QK}}$. (\subref{fig:key_geometry})~Recovery under $\langle\mathrm{H,C}\rangle$ and its change under $\langle\mathrm{T,C}\rangle$, with selected $\kappa$. (\subref{fig:key_kappa})~$\kappa$ sweep, one curve per model. \looseness-1}
\label{fig:key}
\vspace{-1em}
\end{figure}

\section{Experiments}
\label{sec:experiments}
In this section, we first evaluate how the invariances discussed in Section~\ref{sec:method} affect quantization error in a controlled model suite (Section~\ref{sec:experiments:quality}). {We then translate these insights to a production setting, evaluating quality on pretrained Mamba-2 and Nemotron-H models and latency on Mamba-2 (Section~\ref{sec:experiments:production}).}

\subsection{Analyzing LinQuant}
\label{sec:experiments:quality}

\textbf{Model suite.}
To isolate the impact of each component that comprises the linear recurrence, we set up a controlled \emph{mixer zoo}. All models share the same backbone and training recipe. It consists of Linear Attention, Mamba-2, GLA, DeltaNet, Gated DeltaNet, and KDA at 340M and 1.3B parameters, trained with 15B and 100B tokens of SlimPajama, respectively. The 340M suite furthermore contains models that vary write gating $\WriteGateTerm{\beta_t}$ and QK normalization. Appendix~\ref{app:zoo-details} gives the full setup.\looseness-1

\textbf{Quantization settings.}
We evaluate how the identified invariances affect quantization error using WikiText-2 test NLL, comparing against full precision and a naive quantizer without invariance transforms. We use the signed Hadamard rotations and diagonal rescalings discussed in Section~\ref{sec:change-of-basis}, calibrated using WikiText-2. {Several normalized decaying and delta-rule models already incur little quantization error at INT8 with per-channel value scales (Table~\ref{tab:recovery}). We therefore use INT6 in the controlled study to amplify quantization effects and make the impact of the invariances easier to distinguish. INT6 uses a {restricted integer range} within our INT8 kernels; the production experiments use INT8 recurrent computation.} The end-of-chunk states are passed in FP32, while quantizing a copy to an INT8 readout. Appendix~\ref{app:zoo-details} gives calibration details, and Appendix~\ref{app:zoo-results} reports INT8, INT4, 1.3B, and zero-shot results.\looseness-1

\textbf{Scale geometries.}
The queries/read-keys and values can be quantized with different granularities, denoted by $\langle b_{QK}, b_{V} \rangle$. Here, $b_{QK}$ gives the scale-granularity for queries and read-keys, while $b_{V}$ gives that of the values. Both can be C (per-channel), H (per head) or T (per-tensor). The scale groupings considered are directly influenced by the chunkwise kernels. In a matrix contraction, output-channel scales can be applied after the product, whereas scales along the summed dimension must enter the contraction. The key axis is contracted in query/key products and state reads. In contrast, the value axis remains an output dimension and admits per-channel scales that can be incorporated in a kernel's output scaling without modifying the integer contraction or requiring an additional pass. We compare the three scale regimes $\langle\mathrm{H,C}\rangle$, $\langle\mathrm{T,C}\rangle$, and $\langle\mathrm{T,T}\rangle$. In the state write contraction, the key dimension stays an outer dimension. We exploit this by quantizing write keys per channel in $\langle\mathrm{H,C}\rangle$, where each contraction quantizes its own copy of the keys, so the write can take the finest scale it can absorb. In the other two regimes, write keys are quantized per tensor and reused by each contraction.\looseness-1

\begin{figure}[t!]
\centering
\begin{subfigure}[t]{0.245\linewidth}\vskip 0pt\centering\input{figures/pgf/fig_val_structure.pgf}\caption{Recurrence structure.}\label{fig:val_structure}\end{subfigure}\hfill
\begin{subfigure}[t]{0.245\linewidth}\vskip 0pt\centering\input{figures/pgf/fig_val_norm.pgf}\caption{QK norm.}\label{fig:val_norm}\end{subfigure}\hfill
\begin{subfigure}[t]{0.245\linewidth}\vskip 0pt\centering\input{figures/pgf/fig_geometry_value.pgf}\caption{Scale granularity.}\label{fig:val_geometry}\end{subfigure}\hfill
\begin{subfigure}[t]{0.245\linewidth}\vskip 0pt\centering\input{figures/pgf/fig_storage_diff.pgf}\caption{Storage basis.}\label{fig:val_storage}\end{subfigure}
\caption{{Value-axis transformations at INT6.} (\subref{fig:val_structure})~Recovered excess NLL under $\langle\mathrm{T,T}\rangle$. (\subref{fig:val_norm})~Excess NLL under $\langle\mathrm{T,T}\rangle$ with and without QK normalization ($\dagger$). (\subref{fig:val_geometry})~Recovery under $\langle\mathrm{T,C}\rangle$ and $\langle\mathrm{T,T}\rangle$. (\subref{fig:val_storage})~Recovery lost by storing the state rotated instead of canonically under $\RotTerm{\mat R_V}$, relative to naive.\looseness-1}
\label{fig:value}
\vspace{-1em}
\end{figure}
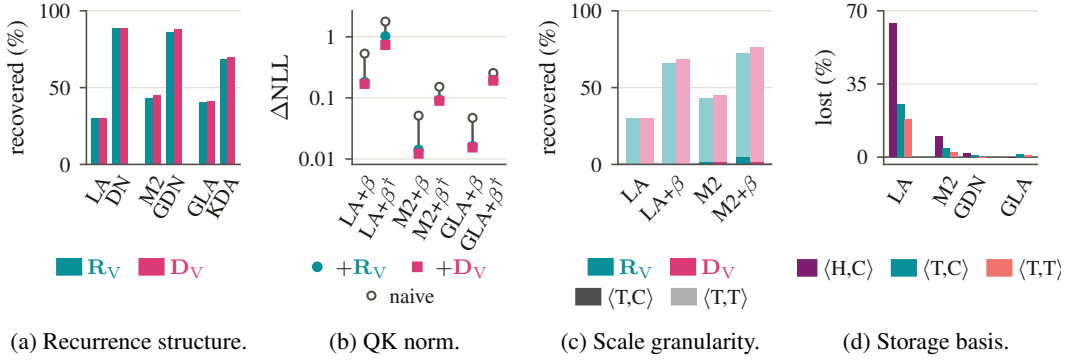

We next examine key-axis transformations (1-2), value-axis transformations (3-4), and absorptions (5-7) on quantization error under different scale-granularities, as well as the choice of storage basis.\looseness-1

\textbf{(1-2) Key-axis transformations.}
The benefit of $\RotTerm{\mat R_{QK}}$ depends on the recurrence structure. With no or scalar decay, it rotates the carried state along the key axis and reduces quantization error. With diagonal decay, it is restricted to local contractions and provides little benefit (Figure~\ref{fig:key_structure}). In contrast, $\RescaleTerm{\mat D_{QK}}$ commutes with diagonal decay and improves quantization error where applicable. QK normalization also reduces quantization error (Figure~\ref{fig:key_norm}). Coarsening key scales increases the benefit of $\RotTerm{\mat R_{QK}}$, while $\RescaleTerm{\mat D_{QK}}$ remains useful with per-head scales (Figure~\ref{fig:key_geometry}). The diagonal rescaling is generally best shared between queries and keys (Figure~\ref{fig:key_kappa}).\looseness-1 

\begin{wrapfigure}[17]{r}{0.50\textwidth}
\vspace{-\intextsep}
\centering
\begin{subfigure}[t]{0.47\linewidth}\vskip 0pt\centering\input{figures/pgf/fig_abs_structure.pgf}\caption{Write gate.}\label{fig:abs_structure}\end{subfigure}\hfill
\begin{subfigure}[t]{0.47\linewidth}\vskip 0pt\centering\input{figures/pgf/fig_decay_side.pgf}\caption{Decay absorption.}\label{fig:abs_decay}\end{subfigure}
\caption{{Absorptions at INT6 under $\langle\mathrm{T,T}\rangle$.} (\subref{fig:abs_structure})~Recovered excess NLL from the selected write-factor allocation ($\xi$ or $(\sigma,\tau)$). (\subref{fig:abs_decay})~NLL increase from placing decay on keys ($\eta=0$) rather than values ($\eta=1$), relative to $\eta=1$.\looseness-1}
\label{fig:absorption}
\end{wrapfigure}
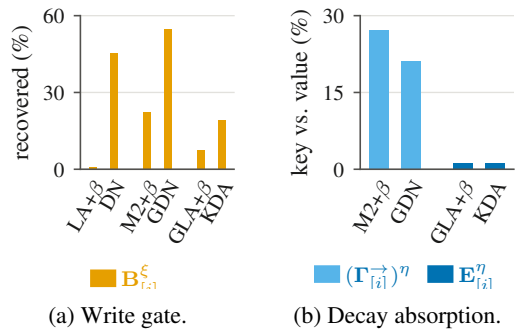

\textbf{(3-4) Value-axis transformations.}
Both $\RotTerm{\mat R_V}$ and $\RescaleTerm{\mat D_V}$ apply to every recurrence. They are most useful with per-tensor value scales, where they recover a large portion of the excess NLL (Figure~\ref{fig:val_structure}). The two transformations yield similar gains as value scales are coarsened (Figure~\ref{fig:val_geometry}). Only $\RotTerm{\mat R_V}$ introduces a choice of state basis: quantizing the state in the rotated basis incurs more error than quantizing it in the canonical basis (Figure~\ref{fig:val_storage}). While the two transformations give similar quality, the cost of canonical storage favors $\RescaleTerm{\mat D_V}$, which folds into the quantization scales.\looseness-1

\textbf{(5-7) Absorptions.}
We study absorptions in the coarse-scale regime. Under $\langle\mathrm{T,T}\rangle$, write-gate absorption recovers a significant portion of the excess NLL (Figure~\ref{fig:abs_structure}). {The preferred write-gate placement also depends on scale granularity. For gated linear writes, under $\langle\mathrm{T,C}\rangle$ keeping the gate on values ($\xi=1$) is preferred, while the other regimes can favor splitting it between keys and values (Figure~\ref{fig:response_rho}). For delta writes under $\langle\mathrm{T,T}\rangle$, the selected settings use $\sigma\in\{0,\tfrac12\}$ and $\tau=1$ (Figure~\ref{fig:response_sigtau}). Compared with the standard WY placement $(\sigma,\tau)=(1,1)$ used in Flash Linear Attention~\citep{fla}, this moves more of the gate factor onto the transformed keys. The standard floating-point placement is therefore not always best after quantization.} Decay placement matters most for scalar decay: with per-tensor value scales, moving scalar decay from values to keys increases excess NLL. By contrast, diagonal-decay absorption provides little gain (Figure~\ref{fig:abs_decay}). We conjecture that variation across the diagonal decay, which remains on the keys, matters more than its overall magnitude and leave other ways of extracting a factor to future work.\looseness-1

\begingroup
\textbf{Combined transformations.}
Combining transformations reduces quantization error more than applying them individually (Figure~\ref{fig:heatmap} and Tables~\ref{tab:zoo-quality-340m} and~\ref{tab:zoo-quality-1p3b}). At INT6 under $\langle\mathrm{T,T}\rangle$, the selected combination recovers roughly twice as much excess NLL as the best single transformation for both 340M Mamba-2 and LA (Table~\ref{tab:zoo-quality-340m}).
\par
\endgroup

\begingroup
\begin{samepage}
\textbf{Downstream accuracy.}
The NLL improvements also translate to downstream accuracy for LA. At INT8 under $\langle\mathrm{T,C}\rangle$, combining transformations largely closes the gap to full precision for 340M LA on both LAMBADA and the six-task average (Table~\ref{tab:zeroshot}). At 1.3B, {LA$+\WriteGateTerm{\beta}$} recovers most of the LAMBADA accuracy lost to naive quantization, though a gap to full precision remains (Table~\ref{tab:zeroshot-1p3b}).
\par
\end{samepage}
\endgroup

\textbf{Kernels.}
{We implement each model's applicable invariances in a single INT8 Triton chunk kernel (Appendix~\ref{app:zoo-kernels}). Rescalings fold into the quantization scales, and absorptions redistribute factors between existing operands. Canonical storage under $\mat R_V$ requires inverse transforms of the state write and local output (Figure~\ref{fig:overhead_rv}). These kernel timings exclude forward rotations performed outside the kernel; the production timings below cover end-to-end prefill.}\looseness-1

\subsection{LinQuant in a production setting}
\label{sec:experiments:production}

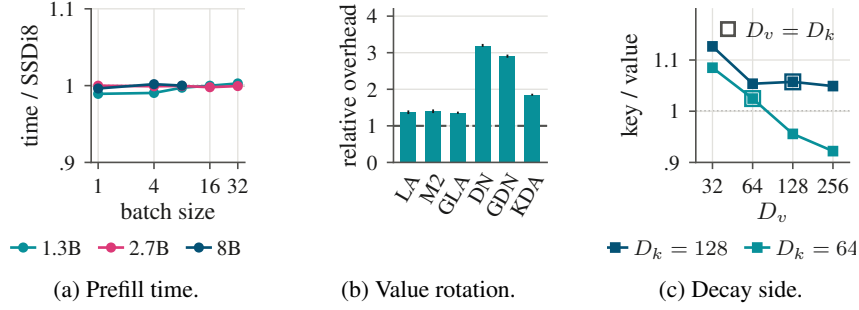
\begin{figure}[t]
\centering
\begin{subfigure}[t]{0.26\linewidth}\vskip 0pt\centering\input{figures/pgf/fig_prefill.pgf}\caption{Prefill time.}\label{fig:overhead_prefill}\end{subfigure}\hspace{0.02\linewidth}
\begin{subfigure}[t]{0.26\linewidth}\vskip 0pt\centering\input{figures/pgf/fig_kernels.pgf}\caption{Value rotation.}\label{fig:overhead_rv}\end{subfigure}\hspace{0.02\linewidth}
\begin{subfigure}[t]{0.26\linewidth}\vskip 0pt\centering\input{figures/pgf/fig_crossover.pgf}\caption{Decay side.}\label{fig:crossover}\end{subfigure}
\caption{(\subref{fig:overhead_prefill})~End-to-end prefill time of LinQuant ({$\RotTerm{\mat R_{QK}}$, value-side decay}) relative to the matched SSDi8 setting, production Mamba-2 (A5000, $T=2048$). (\subref{fig:overhead_rv})~INT8 zoo kernel with $\RotTerm{\mat R_V}$ and canonical storage relative to naive INT8 (A5000, $B=H=8$, $T=2048$, $D_k=D_v=128$). (\subref{fig:crossover})~Production recurrence latency with decay on keys relative to values (static scales, 64 heads, batch 8).\looseness-1}
\label{fig:overhead}
\vspace{-1em}
\end{figure}

\begin{wraptable}[15]{R}{0.48\textwidth}
\vspace{-\intextsep}
\centering
\small
\captionsetup{skip=4pt}
\setlength{\tabcolsep}{1.5pt}
\caption{Production WikiText-2 perplexity. SSDi8 uses value-side decay absorption. The combined transformation is evaluated with decay on either operand. More results in Table~\ref{tab:production-full}.\looseness-1}
\label{tab:production}
\begin{tabular}{@{}lrrr>{}rr@{}}
\toprule
 & & & $\RotTerm{\mat R_{QK}}$ & \multicolumn{2}{c@{}}{$\RotTerm{\mat R_{QK}}+\RescaleTerm{\mat D_{QK}}$} \\
\cmidrule(l){4-6}
Model & FP16 & SSDi8 & value & value & key \\
\midrule
\multicolumn{6}{@{}l}{\emph{Mamba-2}} \\
\quad 1.3B & 10.420 & 10.928 & 10.860 & 10.849 & 10.830 \\
\quad 2.7B & 9.060 & 9.441 & 9.379 & 9.383 & 9.372 \\
\quad 8B & 7.250 & 7.620 & 7.456 & 7.435 & 7.437 \\
\midrule\multicolumn{6}{@{}l}{\emph{Nemotron-H}} \\
\quad Base & 6.030 & 6.338 & 6.186 & 6.178 & 6.175 \\
\quad Reasoning & 8.423 & 8.921 & 8.666 & 8.650 & 8.670 \\
\bottomrule
\end{tabular}
\end{wraptable}

\textbf{Production setup.}
We next apply LinQuant to production-scale pretrained Mamba-2 models and Nemotron-H~\citep{blakeman2025nemotron}. Our starting point is SSDi8~\citep{kimssdi8}, a low-precision recurrent prefill pipeline for Mamba-2, which we augment with the applicable LinQuant invariances. The production Mamba-2 recurrence is most analogous to the no-$\beta$, no-QK-normalization Mamba-2 model in our mixer zoo. The controlled study therefore points primarily to the key-side changes of basis, $\RotTerm{\mat R_{QK}}$ and $\RescaleTerm{\mat D_{QK}}$, while its fine-grained value scales suggest little need for a value-side change of basis. We add these key-side transformations to SSDi8's W4A8 projection and INT8-recurrence pipeline and evaluate both equivalent placements of scalar decay. Table~\ref{tab:production} reports quality. Setup and further results are in Appendices~\ref{app:experimental-details} and~\ref{app:production-ablations}.\looseness-1

\textbf{Key-side transformations transfer.}
Across every reported Mamba-2 and Nemotron-H model, $\RotTerm{\mat R_{QK}}$ improves perplexity over the matched SSDi8 setting, while $\RescaleTerm{\mat D_{QK}}$ provides an additional gain in most configurations (Table~\ref{tab:production}). {Both $\RescaleTerm{\mat D_{QK}}$ and $\RotTerm{\mat R_{QK}}$ are fused into the operand-preparation pass. Query--key rotation adds no measurable end-to-end prefill latency on the evaluated Mamba-2 models (Figure~\ref{fig:overhead_prefill}), while fused rescaling leaves the measured Mamba-2 1.3B layer latency unchanged (Table~\ref{tab:production-rescaling-controls}).} Full results are reported in Appendix~\ref{app:production-ablations}. The SSDi8 pipeline already uses per-channel value scales, matching the controlled setting in which $\RotTerm{\mat R_V}$ and $\RescaleTerm{\mat D_V}$ provide little benefit. Rotated state storage performs worse, while canonical storage does not improve over the unrotated baseline (Appendix~\ref{app:value-rotation}). Since canonical $\RotTerm{\mat R_V}$ adds overhead (Figure~\ref{fig:overhead_rv}), we apply neither value-side transformation in the production configuration.\looseness-1

\textbf{Kernels.}
The optimized SSDi8 kernels quantize operands once and reuse their INT8 representations across multiple contractions. \citet{kimssdi8} place scalar decay on the value side, motivated by concerns that key-side decay would require per-position scales and materializing per-head keys. {Our evaluated key-side configurations achieve comparable perplexity to the value-side configurations (Appendix~\ref{app:production-ablations}), and per-head decay can be applied in GPU registers without materializing the expanded tensor (Appendix~\ref{app:decay-absorption-kernel}). Their runtime also depends on operand reuse.} Moving decay onto an operand after its INT8 representation has been formed requires rereading and requantizing the floating-point tensor. The cost therefore depends on which operand is affected, its dimension, and how often its quantized representation is reused. Figure~\ref{fig:crossover} shows that value-side absorption is faster for the evaluated Mamba-2 shapes, although the gap narrows and can reverse as the value dimension $D_v$ increases. We therefore retain value-side absorption for its lower runtime on the target Mamba-2 geometries. \looseness-1

%% file: figures/pgf/fig_rqk_decay.pgf
\providecommand{\mathdefault}[1]{#1}%
\begingroup%
\makeatletter%
\begin{pgfpicture}%
\pgfpathrectangle{\pgfpointorigin}{\pgfqpoint{1.370000in}{1.640000in}}%
\pgfusepath{use as bounding box, clip}%
\begin{pgfscope}%
\pgfsetbuttcap%
\pgfsetmiterjoin%
\definecolor{currentfill}{rgb}{1.000000,1.000000,1.000000}%
\pgfsetfillcolor{currentfill}%
\pgfsetlinewidth{0.000000pt}%
\definecolor{currentstroke}{rgb}{1.000000,1.000000,1.000000}%
\pgfsetstrokecolor{currentstroke}%
\pgfsetdash{}{0pt}%
\pgfpathmoveto{\pgfqpoint{0.000000in}{0.000000in}}%
\pgfpathlineto{\pgfqpoint{1.370000in}{0.000000in}}%
\pgfpathlineto{\pgfqpoint{1.370000in}{1.640000in}}%
\pgfpathlineto{\pgfqpoint{0.000000in}{1.640000in}}%
\pgfpathlineto{\pgfqpoint{0.000000in}{0.000000in}}%
\pgfpathclose%
\pgfusepath{fill}%
\end{pgfscope}%
\begin{pgfscope}%
\pgfsetbuttcap%
\pgfsetmiterjoin%
\definecolor{currentfill}{rgb}{1.000000,1.000000,1.000000}%
\pgfsetfillcolor{currentfill}%
\pgfsetlinewidth{0.000000pt}%
\definecolor{currentstroke}{rgb}{0.000000,0.000000,0.000000}%
\pgfsetstrokecolor{currentstroke}%
\pgfsetstrokeopacity{0.000000}%
\pgfsetdash{}{0pt}%
\pgfpathmoveto{\pgfqpoint{0.450000in}{0.780000in}}%
\pgfpathlineto{\pgfqpoint{1.250000in}{0.780000in}}%
\pgfpathlineto{\pgfqpoint{1.250000in}{1.580000in}}%
\pgfpathlineto{\pgfqpoint{0.450000in}{1.580000in}}%
\pgfpathlineto{\pgfqpoint{0.450000in}{0.780000in}}%
\pgfpathclose%
\pgfusepath{fill}%
\end{pgfscope}%
\begin{pgfscope}%
\pgfpathrectangle{\pgfqpoint{0.450000in}{0.780000in}}{\pgfqpoint{0.800000in}{0.800000in}}%
\pgfusepath{clip}%
\pgfsetrectcap%
\pgfsetroundjoin%
\pgfsetlinewidth{0.501875pt}%
\definecolor{currentstroke}{rgb}{0.886275,0.882353,0.866667}%
\pgfsetstrokecolor{currentstroke}%
\pgfsetdash{}{0pt}%
\pgfpathmoveto{\pgfqpoint{0.450000in}{0.780000in}}%
\pgfpathlineto{\pgfqpoint{1.250000in}{0.780000in}}%
\pgfusepath{stroke}%
\end{pgfscope}%
\begin{pgfscope}%
\pgfsetbuttcap%
\pgfsetroundjoin%
\definecolor{currentfill}{rgb}{0.321569,0.317647,0.305882}%
\pgfsetfillcolor{currentfill}%
\pgfsetlinewidth{0.602250pt}%
\definecolor{currentstroke}{rgb}{0.321569,0.317647,0.305882}%
\pgfsetstrokecolor{currentstroke}%
\pgfsetdash{}{0pt}%
\pgfsys@defobject{currentmarker}{\pgfqpoint{-0.034722in}{0.000000in}}{\pgfqpoint{-0.000000in}{0.000000in}}{%
\pgfpathmoveto{\pgfqpoint{-0.000000in}{0.000000in}}%
\pgfpathlineto{\pgfqpoint{-0.034722in}{0.000000in}}%
\pgfusepath{stroke,fill}%
}%
\begin{pgfscope}%
\pgfsys@transformshift{0.450000in}{0.780000in}%
\pgfsys@useobject{currentmarker}{}%
\end{pgfscope}%
\end{pgfscope}%
\begin{pgfscope}%
\definecolor{textcolor}{rgb}{0.101961,0.101961,0.098039}%
\pgfsetstrokecolor{textcolor}%
\pgfsetfillcolor{textcolor}%
\pgftext[x=0.311111in, y=0.742112in, left, base]{\color{textcolor}\rmfamily\fontsize{8.000000}{9.600000}\selectfont 0}%
\end{pgfscope}%
\begin{pgfscope}%
\pgfpathrectangle{\pgfqpoint{0.450000in}{0.780000in}}{\pgfqpoint{0.800000in}{0.800000in}}%
\pgfusepath{clip}%
\pgfsetrectcap%
\pgfsetroundjoin%
\pgfsetlinewidth{0.501875pt}%
\definecolor{currentstroke}{rgb}{0.886275,0.882353,0.866667}%
\pgfsetstrokecolor{currentstroke}%
\pgfsetdash{}{0pt}%
\pgfpathmoveto{\pgfqpoint{0.450000in}{1.180000in}}%
\pgfpathlineto{\pgfqpoint{1.250000in}{1.180000in}}%
\pgfusepath{stroke}%
\end{pgfscope}%
\begin{pgfscope}%
\pgfsetbuttcap%
\pgfsetroundjoin%
\definecolor{currentfill}{rgb}{0.321569,0.317647,0.305882}%
\pgfsetfillcolor{currentfill}%
\pgfsetlinewidth{0.602250pt}%
\definecolor{currentstroke}{rgb}{0.321569,0.317647,0.305882}%
\pgfsetstrokecolor{currentstroke}%
\pgfsetdash{}{0pt}%
\pgfsys@defobject{currentmarker}{\pgfqpoint{-0.034722in}{0.000000in}}{\pgfqpoint{-0.000000in}{0.000000in}}{%
\pgfpathmoveto{\pgfqpoint{-0.000000in}{0.000000in}}%
\pgfpathlineto{\pgfqpoint{-0.034722in}{0.000000in}}%
\pgfusepath{stroke,fill}%
}%
\begin{pgfscope}%
\pgfsys@transformshift{0.450000in}{1.180000in}%
\pgfsys@useobject{currentmarker}{}%
\end{pgfscope}%
\end{pgfscope}%
\begin{pgfscope}%
\definecolor{textcolor}{rgb}{0.101961,0.101961,0.098039}%
\pgfsetstrokecolor{textcolor}%
\pgfsetfillcolor{textcolor}%
\pgftext[x=0.255556in, y=1.142112in, left, base]{\color{textcolor}\rmfamily\fontsize{8.000000}{9.600000}\selectfont 50}%
\end{pgfscope}%
\begin{pgfscope}%
\pgfpathrectangle{\pgfqpoint{0.450000in}{0.780000in}}{\pgfqpoint{0.800000in}{0.800000in}}%
\pgfusepath{clip}%
\pgfsetrectcap%
\pgfsetroundjoin%
\pgfsetlinewidth{0.501875pt}%
\definecolor{currentstroke}{rgb}{0.886275,0.882353,0.866667}%
\pgfsetstrokecolor{currentstroke}%
\pgfsetdash{}{0pt}%
\pgfpathmoveto{\pgfqpoint{0.450000in}{1.580000in}}%
\pgfpathlineto{\pgfqpoint{1.250000in}{1.580000in}}%
\pgfusepath{stroke}%
\end{pgfscope}%
\begin{pgfscope}%
\pgfsetbuttcap%
\pgfsetroundjoin%
\definecolor{currentfill}{rgb}{0.321569,0.317647,0.305882}%
\pgfsetfillcolor{currentfill}%
\pgfsetlinewidth{0.602250pt}%
\definecolor{currentstroke}{rgb}{0.321569,0.317647,0.305882}%
\pgfsetstrokecolor{currentstroke}%
\pgfsetdash{}{0pt}%
\pgfsys@defobject{currentmarker}{\pgfqpoint{-0.034722in}{0.000000in}}{\pgfqpoint{-0.000000in}{0.000000in}}{%
\pgfpathmoveto{\pgfqpoint{-0.000000in}{0.000000in}}%
\pgfpathlineto{\pgfqpoint{-0.034722in}{0.000000in}}%
\pgfusepath{stroke,fill}%
}%
\begin{pgfscope}%
\pgfsys@transformshift{0.450000in}{1.580000in}%
\pgfsys@useobject{currentmarker}{}%
\end{pgfscope}%
\end{pgfscope}%
\begin{pgfscope}%
\definecolor{textcolor}{rgb}{0.101961,0.101961,0.098039}%
\pgfsetstrokecolor{textcolor}%
\pgfsetfillcolor{textcolor}%
\pgftext[x=0.200000in, y=1.542112in, left, base]{\color{textcolor}\rmfamily\fontsize{8.000000}{9.600000}\selectfont 100}%
\end{pgfscope}%
\begin{pgfscope}%
\definecolor{textcolor}{rgb}{0.101961,0.101961,0.098039}%
\pgfsetstrokecolor{textcolor}%
\pgfsetfillcolor{textcolor}%
\pgftext[x=0.144444in,y=1.180000in,,bottom,rotate=90.000000]{\color{textcolor}\rmfamily\fontsize{9.000000}{10.800000}\selectfont recovered (\%)}%
\end{pgfscope}%
\begin{pgfscope}%
\pgfsetbuttcap%
\pgfsetmiterjoin%
\definecolor{currentfill}{rgb}{0.031373,0.564706,0.600000}%
\pgfsetfillcolor{currentfill}%
\pgfsetlinewidth{0.000000pt}%
\definecolor{currentstroke}{rgb}{0.000000,0.000000,0.000000}%
\pgfsetstrokecolor{currentstroke}%
\pgfsetstrokeopacity{0.000000}%
\pgfsetdash{}{0pt}%
\pgfpathmoveto{\pgfqpoint{0.473784in}{0.780000in}}%
\pgfpathlineto{\pgfqpoint{0.514865in}{0.780000in}}%
\pgfpathlineto{\pgfqpoint{0.514865in}{1.194913in}}%
\pgfpathlineto{\pgfqpoint{0.473784in}{1.194913in}}%
\pgfpathlineto{\pgfqpoint{0.473784in}{0.780000in}}%
\pgfpathclose%
\pgfusepath{fill}%
\end{pgfscope}%
\begin{pgfscope}%
\pgfsetbuttcap%
\pgfsetmiterjoin%
\definecolor{currentfill}{rgb}{0.862745,0.223529,0.466667}%
\pgfsetfillcolor{currentfill}%
\pgfsetlinewidth{0.000000pt}%
\definecolor{currentstroke}{rgb}{0.000000,0.000000,0.000000}%
\pgfsetstrokecolor{currentstroke}%
\pgfsetstrokeopacity{0.000000}%
\pgfsetdash{}{0pt}%
\pgfpathmoveto{\pgfqpoint{0.514865in}{0.780000in}}%
\pgfpathlineto{\pgfqpoint{0.555946in}{0.780000in}}%
\pgfpathlineto{\pgfqpoint{0.555946in}{1.316229in}}%
\pgfpathlineto{\pgfqpoint{0.514865in}{1.316229in}}%
\pgfpathlineto{\pgfqpoint{0.514865in}{0.780000in}}%
\pgfpathclose%
\pgfusepath{fill}%
\end{pgfscope}%
\begin{pgfscope}%
\pgfsetbuttcap%
\pgfsetmiterjoin%
\definecolor{currentfill}{rgb}{0.031373,0.564706,0.600000}%
\pgfsetfillcolor{currentfill}%
\pgfsetlinewidth{0.000000pt}%
\definecolor{currentstroke}{rgb}{0.000000,0.000000,0.000000}%
\pgfsetstrokecolor{currentstroke}%
\pgfsetstrokeopacity{0.000000}%
\pgfsetdash{}{0pt}%
\pgfpathmoveto{\pgfqpoint{0.602432in}{0.780000in}}%
\pgfpathlineto{\pgfqpoint{0.643514in}{0.780000in}}%
\pgfpathlineto{\pgfqpoint{0.643514in}{1.254605in}}%
\pgfpathlineto{\pgfqpoint{0.602432in}{1.254605in}}%
\pgfpathlineto{\pgfqpoint{0.602432in}{0.780000in}}%
\pgfpathclose%
\pgfusepath{fill}%
\end{pgfscope}%
\begin{pgfscope}%
\pgfsetbuttcap%
\pgfsetmiterjoin%
\definecolor{currentfill}{rgb}{0.031373,0.564706,0.600000}%
\pgfsetfillcolor{currentfill}%
\pgfsetlinewidth{0.000000pt}%
\definecolor{currentstroke}{rgb}{0.000000,0.000000,0.000000}%
\pgfsetstrokecolor{currentstroke}%
\pgfsetstrokeopacity{0.000000}%
\pgfsetdash{}{0pt}%
\pgfpathmoveto{\pgfqpoint{0.754865in}{0.780000in}}%
\pgfpathlineto{\pgfqpoint{0.795946in}{0.780000in}}%
\pgfpathlineto{\pgfqpoint{0.795946in}{1.034522in}}%
\pgfpathlineto{\pgfqpoint{0.754865in}{1.034522in}}%
\pgfpathlineto{\pgfqpoint{0.754865in}{0.780000in}}%
\pgfpathclose%
\pgfusepath{fill}%
\end{pgfscope}%
\begin{pgfscope}%
\pgfsetbuttcap%
\pgfsetmiterjoin%
\definecolor{currentfill}{rgb}{0.862745,0.223529,0.466667}%
\pgfsetfillcolor{currentfill}%
\pgfsetlinewidth{0.000000pt}%
\definecolor{currentstroke}{rgb}{0.000000,0.000000,0.000000}%
\pgfsetstrokecolor{currentstroke}%
\pgfsetstrokeopacity{0.000000}%
\pgfsetdash{}{0pt}%
\pgfpathmoveto{\pgfqpoint{0.795946in}{0.780000in}}%
\pgfpathlineto{\pgfqpoint{0.837027in}{0.780000in}}%
\pgfpathlineto{\pgfqpoint{0.837027in}{1.293197in}}%
\pgfpathlineto{\pgfqpoint{0.795946in}{1.293197in}}%
\pgfpathlineto{\pgfqpoint{0.795946in}{0.780000in}}%
\pgfpathclose%
\pgfusepath{fill}%
\end{pgfscope}%
\begin{pgfscope}%
\pgfsetbuttcap%
\pgfsetmiterjoin%
\definecolor{currentfill}{rgb}{0.031373,0.564706,0.600000}%
\pgfsetfillcolor{currentfill}%
\pgfsetlinewidth{0.000000pt}%
\definecolor{currentstroke}{rgb}{0.000000,0.000000,0.000000}%
\pgfsetstrokecolor{currentstroke}%
\pgfsetstrokeopacity{0.000000}%
\pgfsetdash{}{0pt}%
\pgfpathmoveto{\pgfqpoint{0.883514in}{0.780000in}}%
\pgfpathlineto{\pgfqpoint{0.924595in}{0.780000in}}%
\pgfpathlineto{\pgfqpoint{0.924595in}{1.330286in}}%
\pgfpathlineto{\pgfqpoint{0.883514in}{1.330286in}}%
\pgfpathlineto{\pgfqpoint{0.883514in}{0.780000in}}%
\pgfpathclose%
\pgfusepath{fill}%
\end{pgfscope}%
\begin{pgfscope}%
\pgfsetbuttcap%
\pgfsetmiterjoin%
\definecolor{currentfill}{rgb}{0.031373,0.564706,0.600000}%
\pgfsetfillcolor{currentfill}%
\pgfsetlinewidth{0.000000pt}%
\definecolor{currentstroke}{rgb}{0.000000,0.000000,0.000000}%
\pgfsetstrokecolor{currentstroke}%
\pgfsetstrokeopacity{0.000000}%
\pgfsetdash{}{0pt}%
\pgfpathmoveto{\pgfqpoint{1.035946in}{0.780000in}}%
\pgfpathlineto{\pgfqpoint{1.077027in}{0.780000in}}%
\pgfpathlineto{\pgfqpoint{1.077027in}{0.709706in}}%
\pgfpathlineto{\pgfqpoint{1.035946in}{0.709706in}}%
\pgfpathlineto{\pgfqpoint{1.035946in}{0.780000in}}%
\pgfpathclose%
\pgfusepath{fill}%
\end{pgfscope}%
\begin{pgfscope}%
\pgfsetbuttcap%
\pgfsetmiterjoin%
\definecolor{currentfill}{rgb}{0.862745,0.223529,0.466667}%
\pgfsetfillcolor{currentfill}%
\pgfsetlinewidth{0.000000pt}%
\definecolor{currentstroke}{rgb}{0.000000,0.000000,0.000000}%
\pgfsetstrokecolor{currentstroke}%
\pgfsetstrokeopacity{0.000000}%
\pgfsetdash{}{0pt}%
\pgfpathmoveto{\pgfqpoint{1.077027in}{0.780000in}}%
\pgfpathlineto{\pgfqpoint{1.118108in}{0.780000in}}%
\pgfpathlineto{\pgfqpoint{1.118108in}{1.287123in}}%
\pgfpathlineto{\pgfqpoint{1.077027in}{1.287123in}}%
\pgfpathlineto{\pgfqpoint{1.077027in}{0.780000in}}%
\pgfpathclose%
\pgfusepath{fill}%
\end{pgfscope}%
\begin{pgfscope}%
\pgfsetbuttcap%
\pgfsetmiterjoin%
\definecolor{currentfill}{rgb}{0.031373,0.564706,0.600000}%
\pgfsetfillcolor{currentfill}%
\pgfsetlinewidth{0.000000pt}%
\definecolor{currentstroke}{rgb}{0.000000,0.000000,0.000000}%
\pgfsetstrokecolor{currentstroke}%
\pgfsetstrokeopacity{0.000000}%
\pgfsetdash{}{0pt}%
\pgfpathmoveto{\pgfqpoint{1.164595in}{0.780000in}}%
\pgfpathlineto{\pgfqpoint{1.205676in}{0.780000in}}%
\pgfpathlineto{\pgfqpoint{1.205676in}{0.771146in}}%
\pgfpathlineto{\pgfqpoint{1.164595in}{0.771146in}}%
\pgfpathlineto{\pgfqpoint{1.164595in}{0.780000in}}%
\pgfpathclose%
\pgfusepath{fill}%
\end{pgfscope}%
\begin{pgfscope}%
\pgfsetrectcap%
\pgfsetmiterjoin%
\pgfsetlinewidth{0.602250pt}%
\definecolor{currentstroke}{rgb}{0.321569,0.317647,0.305882}%
\pgfsetstrokecolor{currentstroke}%
\pgfsetdash{}{0pt}%
\pgfpathmoveto{\pgfqpoint{0.450000in}{0.780000in}}%
\pgfpathlineto{\pgfqpoint{0.450000in}{1.580000in}}%
\pgfusepath{stroke}%
\end{pgfscope}%
\begin{pgfscope}%
\pgfsetrectcap%
\pgfsetmiterjoin%
\pgfsetlinewidth{0.602250pt}%
\definecolor{currentstroke}{rgb}{0.321569,0.317647,0.305882}%
\pgfsetstrokecolor{currentstroke}%
\pgfsetdash{}{0pt}%
\pgfpathmoveto{\pgfqpoint{0.450000in}{0.780000in}}%
\pgfpathlineto{\pgfqpoint{1.250000in}{0.780000in}}%
\pgfusepath{stroke}%
\end{pgfscope}%
\begin{pgfscope}%
\definecolor{textcolor}{rgb}{0.101961,0.101961,0.098039}%
\pgfsetstrokecolor{textcolor}%
\pgfsetfillcolor{textcolor}%
\pgftext[x=0.514865in,y=0.731389in,right,top,rotate=60.000000]{\color{textcolor}\rmfamily\fontsize{8.000000}{9.600000}\selectfont LA}%
\end{pgfscope}%
\begin{pgfscope}%
\definecolor{textcolor}{rgb}{0.101961,0.101961,0.098039}%
\pgfsetstrokecolor{textcolor}%
\pgfsetfillcolor{textcolor}%
\pgftext[x=0.622973in,y=0.731389in,right,top,rotate=60.000000]{\color{textcolor}\rmfamily\fontsize{8.000000}{9.600000}\selectfont DN}%
\end{pgfscope}%
\begin{pgfscope}%
\definecolor{textcolor}{rgb}{0.101961,0.101961,0.098039}%
\pgfsetstrokecolor{textcolor}%
\pgfsetfillcolor{textcolor}%
\pgftext[x=0.795946in,y=0.731389in,right,top,rotate=60.000000]{\color{textcolor}\rmfamily\fontsize{8.000000}{9.600000}\selectfont M2}%
\end{pgfscope}%
\begin{pgfscope}%
\definecolor{textcolor}{rgb}{0.101961,0.101961,0.098039}%
\pgfsetstrokecolor{textcolor}%
\pgfsetfillcolor{textcolor}%
\pgftext[x=0.904054in,y=0.731389in,right,top,rotate=60.000000]{\color{textcolor}\rmfamily\fontsize{8.000000}{9.600000}\selectfont GDN}%
\end{pgfscope}%
\begin{pgfscope}%
\definecolor{textcolor}{rgb}{0.101961,0.101961,0.098039}%
\pgfsetstrokecolor{textcolor}%
\pgfsetfillcolor{textcolor}%
\pgftext[x=1.077027in,y=0.731389in,right,top,rotate=60.000000]{\color{textcolor}\rmfamily\fontsize{8.000000}{9.600000}\selectfont GLA}%
\end{pgfscope}%
\begin{pgfscope}%
\definecolor{textcolor}{rgb}{0.101961,0.101961,0.098039}%
\pgfsetstrokecolor{textcolor}%
\pgfsetfillcolor{textcolor}%
\pgftext[x=1.185135in,y=0.731389in,right,top,rotate=60.000000]{\color{textcolor}\rmfamily\fontsize{8.000000}{9.600000}\selectfont KDA}%
\end{pgfscope}%
\begin{pgfscope}%
\pgfsetbuttcap%
\pgfsetmiterjoin%
\definecolor{currentfill}{rgb}{0.031373,0.564706,0.600000}%
\pgfsetfillcolor{currentfill}%
\pgfsetlinewidth{1.003750pt}%
\definecolor{currentstroke}{rgb}{0.031373,0.564706,0.600000}%
\pgfsetstrokecolor{currentstroke}%
\pgfsetdash{}{0pt}%
\pgfpathmoveto{\pgfqpoint{0.219801in}{0.207778in}}%
\pgfpathlineto{\pgfqpoint{0.342023in}{0.207778in}}%
\pgfpathlineto{\pgfqpoint{0.342023in}{0.285556in}}%
\pgfpathlineto{\pgfqpoint{0.219801in}{0.285556in}}%
\pgfpathlineto{\pgfqpoint{0.219801in}{0.207778in}}%
\pgfpathclose%
\pgfusepath{stroke,fill}%
\end{pgfscope}%
\begin{pgfscope}%
\definecolor{textcolor}{rgb}{0.101961,0.101961,0.098039}%
\pgfsetstrokecolor{textcolor}%
\pgfsetfillcolor{textcolor}%
\pgftext[x=0.375356in,y=0.207778in,left,base]{\color{textcolor}\rmfamily\fontsize{8.000000}{9.600000}\selectfont \(\displaystyle \RotTerm{\mathbf{R}_{\mathrm{QK}}}\)}%
\end{pgfscope}%
\begin{pgfscope}%
\pgfsetbuttcap%
\pgfsetmiterjoin%
\definecolor{currentfill}{rgb}{0.862745,0.223529,0.466667}%
\pgfsetfillcolor{currentfill}%
\pgfsetlinewidth{1.003750pt}%
\definecolor{currentstroke}{rgb}{0.862745,0.223529,0.466667}%
\pgfsetstrokecolor{currentstroke}%
\pgfsetdash{}{0pt}%
\pgfpathmoveto{\pgfqpoint{0.728210in}{0.207778in}}%
\pgfpathlineto{\pgfqpoint{0.850432in}{0.207778in}}%
\pgfpathlineto{\pgfqpoint{0.850432in}{0.285556in}}%
\pgfpathlineto{\pgfqpoint{0.728210in}{0.285556in}}%
\pgfpathlineto{\pgfqpoint{0.728210in}{0.207778in}}%
\pgfpathclose%
\pgfusepath{stroke,fill}%
\end{pgfscope}%
\begin{pgfscope}%
\definecolor{textcolor}{rgb}{0.101961,0.101961,0.098039}%
\pgfsetstrokecolor{textcolor}%
\pgfsetfillcolor{textcolor}%
\pgftext[x=0.883765in,y=0.207778in,left,base]{\color{textcolor}\rmfamily\fontsize{8.000000}{9.600000}\selectfont \(\displaystyle \RescaleTerm{\mathbf{D}_{\mathrm{QK}}}\)}%
\end{pgfscope}%
\end{pgfpicture}%
\makeatother%
\endgroup%

%% file: figures/pgf/fig_rqk_norm.pgf
\providecommand{\mathdefault}[1]{#1}%
\begingroup%
\makeatletter%
\begin{pgfpicture}%
\pgfpathrectangle{\pgfpointorigin}{\pgfqpoint{1.370000in}{1.640000in}}%
\pgfusepath{use as bounding box, clip}%
\begin{pgfscope}%
\pgfsetbuttcap%
\pgfsetmiterjoin%
\definecolor{currentfill}{rgb}{1.000000,1.000000,1.000000}%
\pgfsetfillcolor{currentfill}%
\pgfsetlinewidth{0.000000pt}%
\definecolor{currentstroke}{rgb}{1.000000,1.000000,1.000000}%
\pgfsetstrokecolor{currentstroke}%
\pgfsetdash{}{0pt}%
\pgfpathmoveto{\pgfqpoint{0.000000in}{0.000000in}}%
\pgfpathlineto{\pgfqpoint{1.370000in}{0.000000in}}%
\pgfpathlineto{\pgfqpoint{1.370000in}{1.640000in}}%
\pgfpathlineto{\pgfqpoint{0.000000in}{1.640000in}}%
\pgfpathlineto{\pgfqpoint{0.000000in}{0.000000in}}%
\pgfpathclose%
\pgfusepath{fill}%
\end{pgfscope}%
\begin{pgfscope}%
\pgfsetbuttcap%
\pgfsetmiterjoin%
\definecolor{currentfill}{rgb}{1.000000,1.000000,1.000000}%
\pgfsetfillcolor{currentfill}%
\pgfsetlinewidth{0.000000pt}%
\definecolor{currentstroke}{rgb}{0.000000,0.000000,0.000000}%
\pgfsetstrokecolor{currentstroke}%
\pgfsetstrokeopacity{0.000000}%
\pgfsetdash{}{0pt}%
\pgfpathmoveto{\pgfqpoint{0.450000in}{0.780000in}}%
\pgfpathlineto{\pgfqpoint{1.250000in}{0.780000in}}%
\pgfpathlineto{\pgfqpoint{1.250000in}{1.580000in}}%
\pgfpathlineto{\pgfqpoint{0.450000in}{1.580000in}}%
\pgfpathlineto{\pgfqpoint{0.450000in}{0.780000in}}%
\pgfpathclose%
\pgfusepath{fill}%
\end{pgfscope}%
\begin{pgfscope}%
\definecolor{textcolor}{rgb}{0.101961,0.101961,0.098039}%
\pgfsetstrokecolor{textcolor}%
\pgfsetfillcolor{textcolor}%
\pgftext[x=0.514865in,y=0.731389in,right,top,rotate=60.000000]{\color{textcolor}\rmfamily\fontsize{8.000000}{9.600000}\selectfont LA+\(\displaystyle \beta\)}%
\end{pgfscope}%
\begin{pgfscope}%
\definecolor{textcolor}{rgb}{0.101961,0.101961,0.098039}%
\pgfsetstrokecolor{textcolor}%
\pgfsetfillcolor{textcolor}%
\pgftext[x=0.622973in,y=0.731389in,right,top,rotate=60.000000]{\color{textcolor}\rmfamily\fontsize{8.000000}{9.600000}\selectfont LA+\(\displaystyle \beta^\dagger\)}%
\end{pgfscope}%
\begin{pgfscope}%
\definecolor{textcolor}{rgb}{0.101961,0.101961,0.098039}%
\pgfsetstrokecolor{textcolor}%
\pgfsetfillcolor{textcolor}%
\pgftext[x=0.795946in,y=0.731389in,right,top,rotate=60.000000]{\color{textcolor}\rmfamily\fontsize{8.000000}{9.600000}\selectfont M2+\(\displaystyle \beta\)}%
\end{pgfscope}%
\begin{pgfscope}%
\definecolor{textcolor}{rgb}{0.101961,0.101961,0.098039}%
\pgfsetstrokecolor{textcolor}%
\pgfsetfillcolor{textcolor}%
\pgftext[x=0.904054in,y=0.731389in,right,top,rotate=60.000000]{\color{textcolor}\rmfamily\fontsize{8.000000}{9.600000}\selectfont M2+\(\displaystyle \beta^\dagger\)}%
\end{pgfscope}%
\begin{pgfscope}%
\definecolor{textcolor}{rgb}{0.101961,0.101961,0.098039}%
\pgfsetstrokecolor{textcolor}%
\pgfsetfillcolor{textcolor}%
\pgftext[x=1.077027in,y=0.731389in,right,top,rotate=60.000000]{\color{textcolor}\rmfamily\fontsize{8.000000}{9.600000}\selectfont GLA+\(\displaystyle \beta\)}%
\end{pgfscope}%
\begin{pgfscope}%
\definecolor{textcolor}{rgb}{0.101961,0.101961,0.098039}%
\pgfsetstrokecolor{textcolor}%
\pgfsetfillcolor{textcolor}%
\pgftext[x=1.185135in,y=0.731389in,right,top,rotate=60.000000]{\color{textcolor}\rmfamily\fontsize{8.000000}{9.600000}\selectfont GLA+\(\displaystyle \beta^\dagger\)}%
\end{pgfscope}%
\begin{pgfscope}%
\pgfpathrectangle{\pgfqpoint{0.450000in}{0.780000in}}{\pgfqpoint{0.800000in}{0.800000in}}%
\pgfusepath{clip}%
\pgfsetrectcap%
\pgfsetroundjoin%
\pgfsetlinewidth{0.501875pt}%
\definecolor{currentstroke}{rgb}{0.886275,0.882353,0.866667}%
\pgfsetstrokecolor{currentstroke}%
\pgfsetdash{}{0pt}%
\pgfpathmoveto{\pgfqpoint{0.450000in}{0.954957in}}%
\pgfpathlineto{\pgfqpoint{1.250000in}{0.954957in}}%
\pgfusepath{stroke}%
\end{pgfscope}%
\begin{pgfscope}%
\pgfsetbuttcap%
\pgfsetroundjoin%
\definecolor{currentfill}{rgb}{0.321569,0.317647,0.305882}%
\pgfsetfillcolor{currentfill}%
\pgfsetlinewidth{0.602250pt}%
\definecolor{currentstroke}{rgb}{0.321569,0.317647,0.305882}%
\pgfsetstrokecolor{currentstroke}%
\pgfsetdash{}{0pt}%
\pgfsys@defobject{currentmarker}{\pgfqpoint{-0.034722in}{0.000000in}}{\pgfqpoint{-0.000000in}{0.000000in}}{%
\pgfpathmoveto{\pgfqpoint{-0.000000in}{0.000000in}}%
\pgfpathlineto{\pgfqpoint{-0.034722in}{0.000000in}}%
\pgfusepath{stroke,fill}%
}%
\begin{pgfscope}%
\pgfsys@transformshift{0.450000in}{0.954957in}%
\pgfsys@useobject{currentmarker}{}%
\end{pgfscope}%
\end{pgfscope}%
\begin{pgfscope}%
\definecolor{textcolor}{rgb}{0.101961,0.101961,0.098039}%
\pgfsetstrokecolor{textcolor}%
\pgfsetfillcolor{textcolor}%
\pgftext[x=0.172222in, y=0.917069in, left, base]{\color{textcolor}\rmfamily\fontsize{8.000000}{9.600000}\selectfont 0.01}%
\end{pgfscope}%
\begin{pgfscope}%
\pgfpathrectangle{\pgfqpoint{0.450000in}{0.780000in}}{\pgfqpoint{0.800000in}{0.800000in}}%
\pgfusepath{clip}%
\pgfsetrectcap%
\pgfsetroundjoin%
\pgfsetlinewidth{0.501875pt}%
\definecolor{currentstroke}{rgb}{0.886275,0.882353,0.866667}%
\pgfsetstrokecolor{currentstroke}%
\pgfsetdash{}{0pt}%
\pgfpathmoveto{\pgfqpoint{0.450000in}{1.275651in}}%
\pgfpathlineto{\pgfqpoint{1.250000in}{1.275651in}}%
\pgfusepath{stroke}%
\end{pgfscope}%
\begin{pgfscope}%
\pgfsetbuttcap%
\pgfsetroundjoin%
\definecolor{currentfill}{rgb}{0.321569,0.317647,0.305882}%
\pgfsetfillcolor{currentfill}%
\pgfsetlinewidth{0.602250pt}%
\definecolor{currentstroke}{rgb}{0.321569,0.317647,0.305882}%
\pgfsetstrokecolor{currentstroke}%
\pgfsetdash{}{0pt}%
\pgfsys@defobject{currentmarker}{\pgfqpoint{-0.034722in}{0.000000in}}{\pgfqpoint{-0.000000in}{0.000000in}}{%
\pgfpathmoveto{\pgfqpoint{-0.000000in}{0.000000in}}%
\pgfpathlineto{\pgfqpoint{-0.034722in}{0.000000in}}%
\pgfusepath{stroke,fill}%
}%
\begin{pgfscope}%
\pgfsys@transformshift{0.450000in}{1.275651in}%
\pgfsys@useobject{currentmarker}{}%
\end{pgfscope}%
\end{pgfscope}%
\begin{pgfscope}%
\definecolor{textcolor}{rgb}{0.101961,0.101961,0.098039}%
\pgfsetstrokecolor{textcolor}%
\pgfsetfillcolor{textcolor}%
\pgftext[x=0.227778in, y=1.237763in, left, base]{\color{textcolor}\rmfamily\fontsize{8.000000}{9.600000}\selectfont 0.1}%
\end{pgfscope}%
\begin{pgfscope}%
\definecolor{textcolor}{rgb}{0.101961,0.101961,0.098039}%
\pgfsetstrokecolor{textcolor}%
\pgfsetfillcolor{textcolor}%
\pgftext[x=0.116667in,y=1.180000in,,bottom,rotate=90.000000]{\color{textcolor}\rmfamily\fontsize{9.000000}{10.800000}\selectfont \(\displaystyle \Delta\)NLL}%
\end{pgfscope}%
\begin{pgfscope}%
\pgfpathrectangle{\pgfqpoint{0.450000in}{0.780000in}}{\pgfqpoint{0.800000in}{0.800000in}}%
\pgfusepath{clip}%
\pgfsetrectcap%
\pgfsetroundjoin%
\pgfsetlinewidth{0.903375pt}%
\definecolor{currentstroke}{rgb}{0.321569,0.317647,0.305882}%
\pgfsetstrokecolor{currentstroke}%
\pgfsetdash{}{0pt}%
\pgfpathmoveto{\pgfqpoint{0.514865in}{1.172001in}}%
\pgfpathlineto{\pgfqpoint{0.514865in}{1.343202in}}%
\pgfusepath{stroke}%
\end{pgfscope}%
\begin{pgfscope}%
\pgfpathrectangle{\pgfqpoint{0.450000in}{0.780000in}}{\pgfqpoint{0.800000in}{0.800000in}}%
\pgfusepath{clip}%
\pgfsetrectcap%
\pgfsetroundjoin%
\pgfsetlinewidth{0.903375pt}%
\definecolor{currentstroke}{rgb}{0.321569,0.317647,0.305882}%
\pgfsetstrokecolor{currentstroke}%
\pgfsetdash{}{0pt}%
\pgfpathmoveto{\pgfqpoint{0.622973in}{1.381928in}}%
\pgfpathlineto{\pgfqpoint{0.622973in}{1.523529in}}%
\pgfusepath{stroke}%
\end{pgfscope}%
\begin{pgfscope}%
\pgfpathrectangle{\pgfqpoint{0.450000in}{0.780000in}}{\pgfqpoint{0.800000in}{0.800000in}}%
\pgfusepath{clip}%
\pgfsetrectcap%
\pgfsetroundjoin%
\pgfsetlinewidth{0.903375pt}%
\definecolor{currentstroke}{rgb}{0.321569,0.317647,0.305882}%
\pgfsetstrokecolor{currentstroke}%
\pgfsetdash{}{0pt}%
\pgfpathmoveto{\pgfqpoint{0.795946in}{0.836471in}}%
\pgfpathlineto{\pgfqpoint{0.795946in}{0.971794in}}%
\pgfusepath{stroke}%
\end{pgfscope}%
\begin{pgfscope}%
\pgfpathrectangle{\pgfqpoint{0.450000in}{0.780000in}}{\pgfqpoint{0.800000in}{0.800000in}}%
\pgfusepath{clip}%
\pgfsetrectcap%
\pgfsetroundjoin%
\pgfsetlinewidth{0.903375pt}%
\definecolor{currentstroke}{rgb}{0.321569,0.317647,0.305882}%
\pgfsetstrokecolor{currentstroke}%
\pgfsetdash{}{0pt}%
\pgfpathmoveto{\pgfqpoint{0.904054in}{0.909227in}}%
\pgfpathlineto{\pgfqpoint{0.904054in}{1.256660in}}%
\pgfusepath{stroke}%
\end{pgfscope}%
\begin{pgfscope}%
\pgfpathrectangle{\pgfqpoint{0.450000in}{0.780000in}}{\pgfqpoint{0.800000in}{0.800000in}}%
\pgfusepath{clip}%
\pgfsetrectcap%
\pgfsetroundjoin%
\pgfsetlinewidth{0.903375pt}%
\definecolor{currentstroke}{rgb}{0.321569,0.317647,0.305882}%
\pgfsetstrokecolor{currentstroke}%
\pgfsetdash{}{0pt}%
\pgfpathmoveto{\pgfqpoint{1.077027in}{0.903290in}}%
\pgfpathlineto{\pgfqpoint{1.077027in}{1.023243in}}%
\pgfusepath{stroke}%
\end{pgfscope}%
\begin{pgfscope}%
\pgfpathrectangle{\pgfqpoint{0.450000in}{0.780000in}}{\pgfqpoint{0.800000in}{0.800000in}}%
\pgfusepath{clip}%
\pgfsetrectcap%
\pgfsetroundjoin%
\pgfsetlinewidth{0.903375pt}%
\definecolor{currentstroke}{rgb}{0.321569,0.317647,0.305882}%
\pgfsetstrokecolor{currentstroke}%
\pgfsetdash{}{0pt}%
\pgfpathmoveto{\pgfqpoint{1.185135in}{1.199508in}}%
\pgfpathlineto{\pgfqpoint{1.185135in}{1.371242in}}%
\pgfusepath{stroke}%
\end{pgfscope}%
\begin{pgfscope}%
\pgfsetrectcap%
\pgfsetmiterjoin%
\pgfsetlinewidth{0.602250pt}%
\definecolor{currentstroke}{rgb}{0.321569,0.317647,0.305882}%
\pgfsetstrokecolor{currentstroke}%
\pgfsetdash{}{0pt}%
\pgfpathmoveto{\pgfqpoint{0.450000in}{0.780000in}}%
\pgfpathlineto{\pgfqpoint{0.450000in}{1.580000in}}%
\pgfusepath{stroke}%
\end{pgfscope}%
\begin{pgfscope}%
\pgfsetrectcap%
\pgfsetmiterjoin%
\pgfsetlinewidth{0.602250pt}%
\definecolor{currentstroke}{rgb}{0.321569,0.317647,0.305882}%
\pgfsetstrokecolor{currentstroke}%
\pgfsetdash{}{0pt}%
\pgfpathmoveto{\pgfqpoint{0.450000in}{0.780000in}}%
\pgfpathlineto{\pgfqpoint{1.250000in}{0.780000in}}%
\pgfusepath{stroke}%
\end{pgfscope}%
\begin{pgfscope}%
\pgfpathrectangle{\pgfqpoint{0.450000in}{0.780000in}}{\pgfqpoint{0.800000in}{0.800000in}}%
\pgfusepath{clip}%
\pgfsetbuttcap%
\pgfsetroundjoin%
\definecolor{currentfill}{rgb}{1.000000,1.000000,1.000000}%
\pgfsetfillcolor{currentfill}%
\pgfsetlinewidth{0.803000pt}%
\definecolor{currentstroke}{rgb}{0.321569,0.317647,0.305882}%
\pgfsetstrokecolor{currentstroke}%
\pgfsetdash{}{0pt}%
\pgfsys@defobject{currentmarker}{\pgfqpoint{-0.020833in}{-0.020833in}}{\pgfqpoint{0.020833in}{0.020833in}}{%
\pgfpathmoveto{\pgfqpoint{0.000000in}{-0.020833in}}%
\pgfpathcurveto{\pgfqpoint{0.005525in}{-0.020833in}}{\pgfqpoint{0.010825in}{-0.018638in}}{\pgfqpoint{0.014731in}{-0.014731in}}%
\pgfpathcurveto{\pgfqpoint{0.018638in}{-0.010825in}}{\pgfqpoint{0.020833in}{-0.005525in}}{\pgfqpoint{0.020833in}{0.000000in}}%
\pgfpathcurveto{\pgfqpoint{0.020833in}{0.005525in}}{\pgfqpoint{0.018638in}{0.010825in}}{\pgfqpoint{0.014731in}{0.014731in}}%
\pgfpathcurveto{\pgfqpoint{0.010825in}{0.018638in}}{\pgfqpoint{0.005525in}{0.020833in}}{\pgfqpoint{0.000000in}{0.020833in}}%
\pgfpathcurveto{\pgfqpoint{-0.005525in}{0.020833in}}{\pgfqpoint{-0.010825in}{0.018638in}}{\pgfqpoint{-0.014731in}{0.014731in}}%
\pgfpathcurveto{\pgfqpoint{-0.018638in}{0.010825in}}{\pgfqpoint{-0.020833in}{0.005525in}}{\pgfqpoint{-0.020833in}{0.000000in}}%
\pgfpathcurveto{\pgfqpoint{-0.020833in}{-0.005525in}}{\pgfqpoint{-0.018638in}{-0.010825in}}{\pgfqpoint{-0.014731in}{-0.014731in}}%
\pgfpathcurveto{\pgfqpoint{-0.010825in}{-0.018638in}}{\pgfqpoint{-0.005525in}{-0.020833in}}{\pgfqpoint{0.000000in}{-0.020833in}}%
\pgfpathlineto{\pgfqpoint{0.000000in}{-0.020833in}}%
\pgfpathclose%
\pgfusepath{stroke,fill}%
}%
\begin{pgfscope}%
\pgfsys@transformshift{0.514865in}{1.343202in}%
\pgfsys@useobject{currentmarker}{}%
\end{pgfscope}%
\end{pgfscope}%
\begin{pgfscope}%
\pgfpathrectangle{\pgfqpoint{0.450000in}{0.780000in}}{\pgfqpoint{0.800000in}{0.800000in}}%
\pgfusepath{clip}%
\pgfsetbuttcap%
\pgfsetroundjoin%
\definecolor{currentfill}{rgb}{0.031373,0.564706,0.600000}%
\pgfsetfillcolor{currentfill}%
\pgfsetlinewidth{0.803000pt}%
\definecolor{currentstroke}{rgb}{0.031373,0.564706,0.600000}%
\pgfsetstrokecolor{currentstroke}%
\pgfsetdash{}{0pt}%
\pgfsys@defobject{currentmarker}{\pgfqpoint{-0.020833in}{-0.020833in}}{\pgfqpoint{0.020833in}{0.020833in}}{%
\pgfpathmoveto{\pgfqpoint{0.000000in}{-0.020833in}}%
\pgfpathcurveto{\pgfqpoint{0.005525in}{-0.020833in}}{\pgfqpoint{0.010825in}{-0.018638in}}{\pgfqpoint{0.014731in}{-0.014731in}}%
\pgfpathcurveto{\pgfqpoint{0.018638in}{-0.010825in}}{\pgfqpoint{0.020833in}{-0.005525in}}{\pgfqpoint{0.020833in}{0.000000in}}%
\pgfpathcurveto{\pgfqpoint{0.020833in}{0.005525in}}{\pgfqpoint{0.018638in}{0.010825in}}{\pgfqpoint{0.014731in}{0.014731in}}%
\pgfpathcurveto{\pgfqpoint{0.010825in}{0.018638in}}{\pgfqpoint{0.005525in}{0.020833in}}{\pgfqpoint{0.000000in}{0.020833in}}%
\pgfpathcurveto{\pgfqpoint{-0.005525in}{0.020833in}}{\pgfqpoint{-0.010825in}{0.018638in}}{\pgfqpoint{-0.014731in}{0.014731in}}%
\pgfpathcurveto{\pgfqpoint{-0.018638in}{0.010825in}}{\pgfqpoint{-0.020833in}{0.005525in}}{\pgfqpoint{-0.020833in}{0.000000in}}%
\pgfpathcurveto{\pgfqpoint{-0.020833in}{-0.005525in}}{\pgfqpoint{-0.018638in}{-0.010825in}}{\pgfqpoint{-0.014731in}{-0.014731in}}%
\pgfpathcurveto{\pgfqpoint{-0.010825in}{-0.018638in}}{\pgfqpoint{-0.005525in}{-0.020833in}}{\pgfqpoint{0.000000in}{-0.020833in}}%
\pgfpathlineto{\pgfqpoint{0.000000in}{-0.020833in}}%
\pgfpathclose%
\pgfusepath{stroke,fill}%
}%
\begin{pgfscope}%
\pgfsys@transformshift{0.514865in}{1.197659in}%
\pgfsys@useobject{currentmarker}{}%
\end{pgfscope}%
\end{pgfscope}%
\begin{pgfscope}%
\pgfpathrectangle{\pgfqpoint{0.450000in}{0.780000in}}{\pgfqpoint{0.800000in}{0.800000in}}%
\pgfusepath{clip}%
\pgfsetbuttcap%
\pgfsetmiterjoin%
\definecolor{currentfill}{rgb}{0.862745,0.223529,0.466667}%
\pgfsetfillcolor{currentfill}%
\pgfsetlinewidth{0.803000pt}%
\definecolor{currentstroke}{rgb}{0.862745,0.223529,0.466667}%
\pgfsetstrokecolor{currentstroke}%
\pgfsetdash{}{0pt}%
\pgfsys@defobject{currentmarker}{\pgfqpoint{-0.020833in}{-0.020833in}}{\pgfqpoint{0.020833in}{0.020833in}}{%
\pgfpathmoveto{\pgfqpoint{-0.020833in}{-0.020833in}}%
\pgfpathlineto{\pgfqpoint{0.020833in}{-0.020833in}}%
\pgfpathlineto{\pgfqpoint{0.020833in}{0.020833in}}%
\pgfpathlineto{\pgfqpoint{-0.020833in}{0.020833in}}%
\pgfpathlineto{\pgfqpoint{-0.020833in}{-0.020833in}}%
\pgfpathclose%
\pgfusepath{stroke,fill}%
}%
\begin{pgfscope}%
\pgfsys@transformshift{0.514865in}{1.172001in}%
\pgfsys@useobject{currentmarker}{}%
\end{pgfscope}%
\end{pgfscope}%
\begin{pgfscope}%
\pgfpathrectangle{\pgfqpoint{0.450000in}{0.780000in}}{\pgfqpoint{0.800000in}{0.800000in}}%
\pgfusepath{clip}%
\pgfsetbuttcap%
\pgfsetroundjoin%
\definecolor{currentfill}{rgb}{1.000000,1.000000,1.000000}%
\pgfsetfillcolor{currentfill}%
\pgfsetlinewidth{0.803000pt}%
\definecolor{currentstroke}{rgb}{0.321569,0.317647,0.305882}%
\pgfsetstrokecolor{currentstroke}%
\pgfsetdash{}{0pt}%
\pgfsys@defobject{currentmarker}{\pgfqpoint{-0.020833in}{-0.020833in}}{\pgfqpoint{0.020833in}{0.020833in}}{%
\pgfpathmoveto{\pgfqpoint{0.000000in}{-0.020833in}}%
\pgfpathcurveto{\pgfqpoint{0.005525in}{-0.020833in}}{\pgfqpoint{0.010825in}{-0.018638in}}{\pgfqpoint{0.014731in}{-0.014731in}}%
\pgfpathcurveto{\pgfqpoint{0.018638in}{-0.010825in}}{\pgfqpoint{0.020833in}{-0.005525in}}{\pgfqpoint{0.020833in}{0.000000in}}%
\pgfpathcurveto{\pgfqpoint{0.020833in}{0.005525in}}{\pgfqpoint{0.018638in}{0.010825in}}{\pgfqpoint{0.014731in}{0.014731in}}%
\pgfpathcurveto{\pgfqpoint{0.010825in}{0.018638in}}{\pgfqpoint{0.005525in}{0.020833in}}{\pgfqpoint{0.000000in}{0.020833in}}%
\pgfpathcurveto{\pgfqpoint{-0.005525in}{0.020833in}}{\pgfqpoint{-0.010825in}{0.018638in}}{\pgfqpoint{-0.014731in}{0.014731in}}%
\pgfpathcurveto{\pgfqpoint{-0.018638in}{0.010825in}}{\pgfqpoint{-0.020833in}{0.005525in}}{\pgfqpoint{-0.020833in}{0.000000in}}%
\pgfpathcurveto{\pgfqpoint{-0.020833in}{-0.005525in}}{\pgfqpoint{-0.018638in}{-0.010825in}}{\pgfqpoint{-0.014731in}{-0.014731in}}%
\pgfpathcurveto{\pgfqpoint{-0.010825in}{-0.018638in}}{\pgfqpoint{-0.005525in}{-0.020833in}}{\pgfqpoint{0.000000in}{-0.020833in}}%
\pgfpathlineto{\pgfqpoint{0.000000in}{-0.020833in}}%
\pgfpathclose%
\pgfusepath{stroke,fill}%
}%
\begin{pgfscope}%
\pgfsys@transformshift{0.622973in}{1.523529in}%
\pgfsys@useobject{currentmarker}{}%
\end{pgfscope}%
\end{pgfscope}%
\begin{pgfscope}%
\pgfpathrectangle{\pgfqpoint{0.450000in}{0.780000in}}{\pgfqpoint{0.800000in}{0.800000in}}%
\pgfusepath{clip}%
\pgfsetbuttcap%
\pgfsetroundjoin%
\definecolor{currentfill}{rgb}{0.031373,0.564706,0.600000}%
\pgfsetfillcolor{currentfill}%
\pgfsetlinewidth{0.803000pt}%
\definecolor{currentstroke}{rgb}{0.031373,0.564706,0.600000}%
\pgfsetstrokecolor{currentstroke}%
\pgfsetdash{}{0pt}%
\pgfsys@defobject{currentmarker}{\pgfqpoint{-0.020833in}{-0.020833in}}{\pgfqpoint{0.020833in}{0.020833in}}{%
\pgfpathmoveto{\pgfqpoint{0.000000in}{-0.020833in}}%
\pgfpathcurveto{\pgfqpoint{0.005525in}{-0.020833in}}{\pgfqpoint{0.010825in}{-0.018638in}}{\pgfqpoint{0.014731in}{-0.014731in}}%
\pgfpathcurveto{\pgfqpoint{0.018638in}{-0.010825in}}{\pgfqpoint{0.020833in}{-0.005525in}}{\pgfqpoint{0.020833in}{0.000000in}}%
\pgfpathcurveto{\pgfqpoint{0.020833in}{0.005525in}}{\pgfqpoint{0.018638in}{0.010825in}}{\pgfqpoint{0.014731in}{0.014731in}}%
\pgfpathcurveto{\pgfqpoint{0.010825in}{0.018638in}}{\pgfqpoint{0.005525in}{0.020833in}}{\pgfqpoint{0.000000in}{0.020833in}}%
\pgfpathcurveto{\pgfqpoint{-0.005525in}{0.020833in}}{\pgfqpoint{-0.010825in}{0.018638in}}{\pgfqpoint{-0.014731in}{0.014731in}}%
\pgfpathcurveto{\pgfqpoint{-0.018638in}{0.010825in}}{\pgfqpoint{-0.020833in}{0.005525in}}{\pgfqpoint{-0.020833in}{0.000000in}}%
\pgfpathcurveto{\pgfqpoint{-0.020833in}{-0.005525in}}{\pgfqpoint{-0.018638in}{-0.010825in}}{\pgfqpoint{-0.014731in}{-0.014731in}}%
\pgfpathcurveto{\pgfqpoint{-0.010825in}{-0.018638in}}{\pgfqpoint{-0.005525in}{-0.020833in}}{\pgfqpoint{0.000000in}{-0.020833in}}%
\pgfpathlineto{\pgfqpoint{0.000000in}{-0.020833in}}%
\pgfpathclose%
\pgfusepath{stroke,fill}%
}%
\begin{pgfscope}%
\pgfsys@transformshift{0.622973in}{1.515126in}%
\pgfsys@useobject{currentmarker}{}%
\end{pgfscope}%
\end{pgfscope}%
\begin{pgfscope}%
\pgfpathrectangle{\pgfqpoint{0.450000in}{0.780000in}}{\pgfqpoint{0.800000in}{0.800000in}}%
\pgfusepath{clip}%
\pgfsetbuttcap%
\pgfsetmiterjoin%
\definecolor{currentfill}{rgb}{0.862745,0.223529,0.466667}%
\pgfsetfillcolor{currentfill}%
\pgfsetlinewidth{0.803000pt}%
\definecolor{currentstroke}{rgb}{0.862745,0.223529,0.466667}%
\pgfsetstrokecolor{currentstroke}%
\pgfsetdash{}{0pt}%
\pgfsys@defobject{currentmarker}{\pgfqpoint{-0.020833in}{-0.020833in}}{\pgfqpoint{0.020833in}{0.020833in}}{%
\pgfpathmoveto{\pgfqpoint{-0.020833in}{-0.020833in}}%
\pgfpathlineto{\pgfqpoint{0.020833in}{-0.020833in}}%
\pgfpathlineto{\pgfqpoint{0.020833in}{0.020833in}}%
\pgfpathlineto{\pgfqpoint{-0.020833in}{0.020833in}}%
\pgfpathlineto{\pgfqpoint{-0.020833in}{-0.020833in}}%
\pgfpathclose%
\pgfusepath{stroke,fill}%
}%
\begin{pgfscope}%
\pgfsys@transformshift{0.622973in}{1.381928in}%
\pgfsys@useobject{currentmarker}{}%
\end{pgfscope}%
\end{pgfscope}%
\begin{pgfscope}%
\pgfpathrectangle{\pgfqpoint{0.450000in}{0.780000in}}{\pgfqpoint{0.800000in}{0.800000in}}%
\pgfusepath{clip}%
\pgfsetbuttcap%
\pgfsetroundjoin%
\definecolor{currentfill}{rgb}{1.000000,1.000000,1.000000}%
\pgfsetfillcolor{currentfill}%
\pgfsetlinewidth{0.803000pt}%
\definecolor{currentstroke}{rgb}{0.321569,0.317647,0.305882}%
\pgfsetstrokecolor{currentstroke}%
\pgfsetdash{}{0pt}%
\pgfsys@defobject{currentmarker}{\pgfqpoint{-0.020833in}{-0.020833in}}{\pgfqpoint{0.020833in}{0.020833in}}{%
\pgfpathmoveto{\pgfqpoint{0.000000in}{-0.020833in}}%
\pgfpathcurveto{\pgfqpoint{0.005525in}{-0.020833in}}{\pgfqpoint{0.010825in}{-0.018638in}}{\pgfqpoint{0.014731in}{-0.014731in}}%
\pgfpathcurveto{\pgfqpoint{0.018638in}{-0.010825in}}{\pgfqpoint{0.020833in}{-0.005525in}}{\pgfqpoint{0.020833in}{0.000000in}}%
\pgfpathcurveto{\pgfqpoint{0.020833in}{0.005525in}}{\pgfqpoint{0.018638in}{0.010825in}}{\pgfqpoint{0.014731in}{0.014731in}}%
\pgfpathcurveto{\pgfqpoint{0.010825in}{0.018638in}}{\pgfqpoint{0.005525in}{0.020833in}}{\pgfqpoint{0.000000in}{0.020833in}}%
\pgfpathcurveto{\pgfqpoint{-0.005525in}{0.020833in}}{\pgfqpoint{-0.010825in}{0.018638in}}{\pgfqpoint{-0.014731in}{0.014731in}}%
\pgfpathcurveto{\pgfqpoint{-0.018638in}{0.010825in}}{\pgfqpoint{-0.020833in}{0.005525in}}{\pgfqpoint{-0.020833in}{0.000000in}}%
\pgfpathcurveto{\pgfqpoint{-0.020833in}{-0.005525in}}{\pgfqpoint{-0.018638in}{-0.010825in}}{\pgfqpoint{-0.014731in}{-0.014731in}}%
\pgfpathcurveto{\pgfqpoint{-0.010825in}{-0.018638in}}{\pgfqpoint{-0.005525in}{-0.020833in}}{\pgfqpoint{0.000000in}{-0.020833in}}%
\pgfpathlineto{\pgfqpoint{0.000000in}{-0.020833in}}%
\pgfpathclose%
\pgfusepath{stroke,fill}%
}%
\begin{pgfscope}%
\pgfsys@transformshift{0.795946in}{0.971794in}%
\pgfsys@useobject{currentmarker}{}%
\end{pgfscope}%
\end{pgfscope}%
\begin{pgfscope}%
\pgfpathrectangle{\pgfqpoint{0.450000in}{0.780000in}}{\pgfqpoint{0.800000in}{0.800000in}}%
\pgfusepath{clip}%
\pgfsetbuttcap%
\pgfsetroundjoin%
\definecolor{currentfill}{rgb}{0.031373,0.564706,0.600000}%
\pgfsetfillcolor{currentfill}%
\pgfsetlinewidth{0.803000pt}%
\definecolor{currentstroke}{rgb}{0.031373,0.564706,0.600000}%
\pgfsetstrokecolor{currentstroke}%
\pgfsetdash{}{0pt}%
\pgfsys@defobject{currentmarker}{\pgfqpoint{-0.020833in}{-0.020833in}}{\pgfqpoint{0.020833in}{0.020833in}}{%
\pgfpathmoveto{\pgfqpoint{0.000000in}{-0.020833in}}%
\pgfpathcurveto{\pgfqpoint{0.005525in}{-0.020833in}}{\pgfqpoint{0.010825in}{-0.018638in}}{\pgfqpoint{0.014731in}{-0.014731in}}%
\pgfpathcurveto{\pgfqpoint{0.018638in}{-0.010825in}}{\pgfqpoint{0.020833in}{-0.005525in}}{\pgfqpoint{0.020833in}{0.000000in}}%
\pgfpathcurveto{\pgfqpoint{0.020833in}{0.005525in}}{\pgfqpoint{0.018638in}{0.010825in}}{\pgfqpoint{0.014731in}{0.014731in}}%
\pgfpathcurveto{\pgfqpoint{0.010825in}{0.018638in}}{\pgfqpoint{0.005525in}{0.020833in}}{\pgfqpoint{0.000000in}{0.020833in}}%
\pgfpathcurveto{\pgfqpoint{-0.005525in}{0.020833in}}{\pgfqpoint{-0.010825in}{0.018638in}}{\pgfqpoint{-0.014731in}{0.014731in}}%
\pgfpathcurveto{\pgfqpoint{-0.018638in}{0.010825in}}{\pgfqpoint{-0.020833in}{0.005525in}}{\pgfqpoint{-0.020833in}{0.000000in}}%
\pgfpathcurveto{\pgfqpoint{-0.020833in}{-0.005525in}}{\pgfqpoint{-0.018638in}{-0.010825in}}{\pgfqpoint{-0.014731in}{-0.014731in}}%
\pgfpathcurveto{\pgfqpoint{-0.010825in}{-0.018638in}}{\pgfqpoint{-0.005525in}{-0.020833in}}{\pgfqpoint{0.000000in}{-0.020833in}}%
\pgfpathlineto{\pgfqpoint{0.000000in}{-0.020833in}}%
\pgfpathclose%
\pgfusepath{stroke,fill}%
}%
\begin{pgfscope}%
\pgfsys@transformshift{0.795946in}{0.861454in}%
\pgfsys@useobject{currentmarker}{}%
\end{pgfscope}%
\end{pgfscope}%
\begin{pgfscope}%
\pgfpathrectangle{\pgfqpoint{0.450000in}{0.780000in}}{\pgfqpoint{0.800000in}{0.800000in}}%
\pgfusepath{clip}%
\pgfsetbuttcap%
\pgfsetmiterjoin%
\definecolor{currentfill}{rgb}{0.862745,0.223529,0.466667}%
\pgfsetfillcolor{currentfill}%
\pgfsetlinewidth{0.803000pt}%
\definecolor{currentstroke}{rgb}{0.862745,0.223529,0.466667}%
\pgfsetstrokecolor{currentstroke}%
\pgfsetdash{}{0pt}%
\pgfsys@defobject{currentmarker}{\pgfqpoint{-0.020833in}{-0.020833in}}{\pgfqpoint{0.020833in}{0.020833in}}{%
\pgfpathmoveto{\pgfqpoint{-0.020833in}{-0.020833in}}%
\pgfpathlineto{\pgfqpoint{0.020833in}{-0.020833in}}%
\pgfpathlineto{\pgfqpoint{0.020833in}{0.020833in}}%
\pgfpathlineto{\pgfqpoint{-0.020833in}{0.020833in}}%
\pgfpathlineto{\pgfqpoint{-0.020833in}{-0.020833in}}%
\pgfpathclose%
\pgfusepath{stroke,fill}%
}%
\begin{pgfscope}%
\pgfsys@transformshift{0.795946in}{0.836471in}%
\pgfsys@useobject{currentmarker}{}%
\end{pgfscope}%
\end{pgfscope}%
\begin{pgfscope}%
\pgfpathrectangle{\pgfqpoint{0.450000in}{0.780000in}}{\pgfqpoint{0.800000in}{0.800000in}}%
\pgfusepath{clip}%
\pgfsetbuttcap%
\pgfsetroundjoin%
\definecolor{currentfill}{rgb}{1.000000,1.000000,1.000000}%
\pgfsetfillcolor{currentfill}%
\pgfsetlinewidth{0.803000pt}%
\definecolor{currentstroke}{rgb}{0.321569,0.317647,0.305882}%
\pgfsetstrokecolor{currentstroke}%
\pgfsetdash{}{0pt}%
\pgfsys@defobject{currentmarker}{\pgfqpoint{-0.020833in}{-0.020833in}}{\pgfqpoint{0.020833in}{0.020833in}}{%
\pgfpathmoveto{\pgfqpoint{0.000000in}{-0.020833in}}%
\pgfpathcurveto{\pgfqpoint{0.005525in}{-0.020833in}}{\pgfqpoint{0.010825in}{-0.018638in}}{\pgfqpoint{0.014731in}{-0.014731in}}%
\pgfpathcurveto{\pgfqpoint{0.018638in}{-0.010825in}}{\pgfqpoint{0.020833in}{-0.005525in}}{\pgfqpoint{0.020833in}{0.000000in}}%
\pgfpathcurveto{\pgfqpoint{0.020833in}{0.005525in}}{\pgfqpoint{0.018638in}{0.010825in}}{\pgfqpoint{0.014731in}{0.014731in}}%
\pgfpathcurveto{\pgfqpoint{0.010825in}{0.018638in}}{\pgfqpoint{0.005525in}{0.020833in}}{\pgfqpoint{0.000000in}{0.020833in}}%
\pgfpathcurveto{\pgfqpoint{-0.005525in}{0.020833in}}{\pgfqpoint{-0.010825in}{0.018638in}}{\pgfqpoint{-0.014731in}{0.014731in}}%
\pgfpathcurveto{\pgfqpoint{-0.018638in}{0.010825in}}{\pgfqpoint{-0.020833in}{0.005525in}}{\pgfqpoint{-0.020833in}{0.000000in}}%
\pgfpathcurveto{\pgfqpoint{-0.020833in}{-0.005525in}}{\pgfqpoint{-0.018638in}{-0.010825in}}{\pgfqpoint{-0.014731in}{-0.014731in}}%
\pgfpathcurveto{\pgfqpoint{-0.010825in}{-0.018638in}}{\pgfqpoint{-0.005525in}{-0.020833in}}{\pgfqpoint{0.000000in}{-0.020833in}}%
\pgfpathlineto{\pgfqpoint{0.000000in}{-0.020833in}}%
\pgfpathclose%
\pgfusepath{stroke,fill}%
}%
\begin{pgfscope}%
\pgfsys@transformshift{0.904054in}{1.256660in}%
\pgfsys@useobject{currentmarker}{}%
\end{pgfscope}%
\end{pgfscope}%
\begin{pgfscope}%
\pgfpathrectangle{\pgfqpoint{0.450000in}{0.780000in}}{\pgfqpoint{0.800000in}{0.800000in}}%
\pgfusepath{clip}%
\pgfsetbuttcap%
\pgfsetroundjoin%
\definecolor{currentfill}{rgb}{0.031373,0.564706,0.600000}%
\pgfsetfillcolor{currentfill}%
\pgfsetlinewidth{0.803000pt}%
\definecolor{currentstroke}{rgb}{0.031373,0.564706,0.600000}%
\pgfsetstrokecolor{currentstroke}%
\pgfsetdash{}{0pt}%
\pgfsys@defobject{currentmarker}{\pgfqpoint{-0.020833in}{-0.020833in}}{\pgfqpoint{0.020833in}{0.020833in}}{%
\pgfpathmoveto{\pgfqpoint{0.000000in}{-0.020833in}}%
\pgfpathcurveto{\pgfqpoint{0.005525in}{-0.020833in}}{\pgfqpoint{0.010825in}{-0.018638in}}{\pgfqpoint{0.014731in}{-0.014731in}}%
\pgfpathcurveto{\pgfqpoint{0.018638in}{-0.010825in}}{\pgfqpoint{0.020833in}{-0.005525in}}{\pgfqpoint{0.020833in}{0.000000in}}%
\pgfpathcurveto{\pgfqpoint{0.020833in}{0.005525in}}{\pgfqpoint{0.018638in}{0.010825in}}{\pgfqpoint{0.014731in}{0.014731in}}%
\pgfpathcurveto{\pgfqpoint{0.010825in}{0.018638in}}{\pgfqpoint{0.005525in}{0.020833in}}{\pgfqpoint{0.000000in}{0.020833in}}%
\pgfpathcurveto{\pgfqpoint{-0.005525in}{0.020833in}}{\pgfqpoint{-0.010825in}{0.018638in}}{\pgfqpoint{-0.014731in}{0.014731in}}%
\pgfpathcurveto{\pgfqpoint{-0.018638in}{0.010825in}}{\pgfqpoint{-0.020833in}{0.005525in}}{\pgfqpoint{-0.020833in}{0.000000in}}%
\pgfpathcurveto{\pgfqpoint{-0.020833in}{-0.005525in}}{\pgfqpoint{-0.018638in}{-0.010825in}}{\pgfqpoint{-0.014731in}{-0.014731in}}%
\pgfpathcurveto{\pgfqpoint{-0.010825in}{-0.018638in}}{\pgfqpoint{-0.005525in}{-0.020833in}}{\pgfqpoint{0.000000in}{-0.020833in}}%
\pgfpathlineto{\pgfqpoint{0.000000in}{-0.020833in}}%
\pgfpathclose%
\pgfusepath{stroke,fill}%
}%
\begin{pgfscope}%
\pgfsys@transformshift{0.904054in}{0.909227in}%
\pgfsys@useobject{currentmarker}{}%
\end{pgfscope}%
\end{pgfscope}%
\begin{pgfscope}%
\pgfpathrectangle{\pgfqpoint{0.450000in}{0.780000in}}{\pgfqpoint{0.800000in}{0.800000in}}%
\pgfusepath{clip}%
\pgfsetbuttcap%
\pgfsetmiterjoin%
\definecolor{currentfill}{rgb}{0.862745,0.223529,0.466667}%
\pgfsetfillcolor{currentfill}%
\pgfsetlinewidth{0.803000pt}%
\definecolor{currentstroke}{rgb}{0.862745,0.223529,0.466667}%
\pgfsetstrokecolor{currentstroke}%
\pgfsetdash{}{0pt}%
\pgfsys@defobject{currentmarker}{\pgfqpoint{-0.020833in}{-0.020833in}}{\pgfqpoint{0.020833in}{0.020833in}}{%
\pgfpathmoveto{\pgfqpoint{-0.020833in}{-0.020833in}}%
\pgfpathlineto{\pgfqpoint{0.020833in}{-0.020833in}}%
\pgfpathlineto{\pgfqpoint{0.020833in}{0.020833in}}%
\pgfpathlineto{\pgfqpoint{-0.020833in}{0.020833in}}%
\pgfpathlineto{\pgfqpoint{-0.020833in}{-0.020833in}}%
\pgfpathclose%
\pgfusepath{stroke,fill}%
}%
\begin{pgfscope}%
\pgfsys@transformshift{0.904054in}{1.166228in}%
\pgfsys@useobject{currentmarker}{}%
\end{pgfscope}%
\end{pgfscope}%
\begin{pgfscope}%
\pgfpathrectangle{\pgfqpoint{0.450000in}{0.780000in}}{\pgfqpoint{0.800000in}{0.800000in}}%
\pgfusepath{clip}%
\pgfsetbuttcap%
\pgfsetroundjoin%
\definecolor{currentfill}{rgb}{1.000000,1.000000,1.000000}%
\pgfsetfillcolor{currentfill}%
\pgfsetlinewidth{0.803000pt}%
\definecolor{currentstroke}{rgb}{0.321569,0.317647,0.305882}%
\pgfsetstrokecolor{currentstroke}%
\pgfsetdash{}{0pt}%
\pgfsys@defobject{currentmarker}{\pgfqpoint{-0.020833in}{-0.020833in}}{\pgfqpoint{0.020833in}{0.020833in}}{%
\pgfpathmoveto{\pgfqpoint{0.000000in}{-0.020833in}}%
\pgfpathcurveto{\pgfqpoint{0.005525in}{-0.020833in}}{\pgfqpoint{0.010825in}{-0.018638in}}{\pgfqpoint{0.014731in}{-0.014731in}}%
\pgfpathcurveto{\pgfqpoint{0.018638in}{-0.010825in}}{\pgfqpoint{0.020833in}{-0.005525in}}{\pgfqpoint{0.020833in}{0.000000in}}%
\pgfpathcurveto{\pgfqpoint{0.020833in}{0.005525in}}{\pgfqpoint{0.018638in}{0.010825in}}{\pgfqpoint{0.014731in}{0.014731in}}%
\pgfpathcurveto{\pgfqpoint{0.010825in}{0.018638in}}{\pgfqpoint{0.005525in}{0.020833in}}{\pgfqpoint{0.000000in}{0.020833in}}%
\pgfpathcurveto{\pgfqpoint{-0.005525in}{0.020833in}}{\pgfqpoint{-0.010825in}{0.018638in}}{\pgfqpoint{-0.014731in}{0.014731in}}%
\pgfpathcurveto{\pgfqpoint{-0.018638in}{0.010825in}}{\pgfqpoint{-0.020833in}{0.005525in}}{\pgfqpoint{-0.020833in}{0.000000in}}%
\pgfpathcurveto{\pgfqpoint{-0.020833in}{-0.005525in}}{\pgfqpoint{-0.018638in}{-0.010825in}}{\pgfqpoint{-0.014731in}{-0.014731in}}%
\pgfpathcurveto{\pgfqpoint{-0.010825in}{-0.018638in}}{\pgfqpoint{-0.005525in}{-0.020833in}}{\pgfqpoint{0.000000in}{-0.020833in}}%
\pgfpathlineto{\pgfqpoint{0.000000in}{-0.020833in}}%
\pgfpathclose%
\pgfusepath{stroke,fill}%
}%
\begin{pgfscope}%
\pgfsys@transformshift{1.077027in}{1.011816in}%
\pgfsys@useobject{currentmarker}{}%
\end{pgfscope}%
\end{pgfscope}%
\begin{pgfscope}%
\pgfpathrectangle{\pgfqpoint{0.450000in}{0.780000in}}{\pgfqpoint{0.800000in}{0.800000in}}%
\pgfusepath{clip}%
\pgfsetbuttcap%
\pgfsetroundjoin%
\definecolor{currentfill}{rgb}{0.031373,0.564706,0.600000}%
\pgfsetfillcolor{currentfill}%
\pgfsetlinewidth{0.803000pt}%
\definecolor{currentstroke}{rgb}{0.031373,0.564706,0.600000}%
\pgfsetstrokecolor{currentstroke}%
\pgfsetdash{}{0pt}%
\pgfsys@defobject{currentmarker}{\pgfqpoint{-0.020833in}{-0.020833in}}{\pgfqpoint{0.020833in}{0.020833in}}{%
\pgfpathmoveto{\pgfqpoint{0.000000in}{-0.020833in}}%
\pgfpathcurveto{\pgfqpoint{0.005525in}{-0.020833in}}{\pgfqpoint{0.010825in}{-0.018638in}}{\pgfqpoint{0.014731in}{-0.014731in}}%
\pgfpathcurveto{\pgfqpoint{0.018638in}{-0.010825in}}{\pgfqpoint{0.020833in}{-0.005525in}}{\pgfqpoint{0.020833in}{0.000000in}}%
\pgfpathcurveto{\pgfqpoint{0.020833in}{0.005525in}}{\pgfqpoint{0.018638in}{0.010825in}}{\pgfqpoint{0.014731in}{0.014731in}}%
\pgfpathcurveto{\pgfqpoint{0.010825in}{0.018638in}}{\pgfqpoint{0.005525in}{0.020833in}}{\pgfqpoint{0.000000in}{0.020833in}}%
\pgfpathcurveto{\pgfqpoint{-0.005525in}{0.020833in}}{\pgfqpoint{-0.010825in}{0.018638in}}{\pgfqpoint{-0.014731in}{0.014731in}}%
\pgfpathcurveto{\pgfqpoint{-0.018638in}{0.010825in}}{\pgfqpoint{-0.020833in}{0.005525in}}{\pgfqpoint{-0.020833in}{0.000000in}}%
\pgfpathcurveto{\pgfqpoint{-0.020833in}{-0.005525in}}{\pgfqpoint{-0.018638in}{-0.010825in}}{\pgfqpoint{-0.014731in}{-0.014731in}}%
\pgfpathcurveto{\pgfqpoint{-0.010825in}{-0.018638in}}{\pgfqpoint{-0.005525in}{-0.020833in}}{\pgfqpoint{0.000000in}{-0.020833in}}%
\pgfpathlineto{\pgfqpoint{0.000000in}{-0.020833in}}%
\pgfpathclose%
\pgfusepath{stroke,fill}%
}%
\begin{pgfscope}%
\pgfsys@transformshift{1.077027in}{1.023243in}%
\pgfsys@useobject{currentmarker}{}%
\end{pgfscope}%
\end{pgfscope}%
\begin{pgfscope}%
\pgfpathrectangle{\pgfqpoint{0.450000in}{0.780000in}}{\pgfqpoint{0.800000in}{0.800000in}}%
\pgfusepath{clip}%
\pgfsetbuttcap%
\pgfsetmiterjoin%
\definecolor{currentfill}{rgb}{0.862745,0.223529,0.466667}%
\pgfsetfillcolor{currentfill}%
\pgfsetlinewidth{0.803000pt}%
\definecolor{currentstroke}{rgb}{0.862745,0.223529,0.466667}%
\pgfsetstrokecolor{currentstroke}%
\pgfsetdash{}{0pt}%
\pgfsys@defobject{currentmarker}{\pgfqpoint{-0.020833in}{-0.020833in}}{\pgfqpoint{0.020833in}{0.020833in}}{%
\pgfpathmoveto{\pgfqpoint{-0.020833in}{-0.020833in}}%
\pgfpathlineto{\pgfqpoint{0.020833in}{-0.020833in}}%
\pgfpathlineto{\pgfqpoint{0.020833in}{0.020833in}}%
\pgfpathlineto{\pgfqpoint{-0.020833in}{0.020833in}}%
\pgfpathlineto{\pgfqpoint{-0.020833in}{-0.020833in}}%
\pgfpathclose%
\pgfusepath{stroke,fill}%
}%
\begin{pgfscope}%
\pgfsys@transformshift{1.077027in}{0.903290in}%
\pgfsys@useobject{currentmarker}{}%
\end{pgfscope}%
\end{pgfscope}%
\begin{pgfscope}%
\pgfpathrectangle{\pgfqpoint{0.450000in}{0.780000in}}{\pgfqpoint{0.800000in}{0.800000in}}%
\pgfusepath{clip}%
\pgfsetbuttcap%
\pgfsetroundjoin%
\definecolor{currentfill}{rgb}{1.000000,1.000000,1.000000}%
\pgfsetfillcolor{currentfill}%
\pgfsetlinewidth{0.803000pt}%
\definecolor{currentstroke}{rgb}{0.321569,0.317647,0.305882}%
\pgfsetstrokecolor{currentstroke}%
\pgfsetdash{}{0pt}%
\pgfsys@defobject{currentmarker}{\pgfqpoint{-0.020833in}{-0.020833in}}{\pgfqpoint{0.020833in}{0.020833in}}{%
\pgfpathmoveto{\pgfqpoint{0.000000in}{-0.020833in}}%
\pgfpathcurveto{\pgfqpoint{0.005525in}{-0.020833in}}{\pgfqpoint{0.010825in}{-0.018638in}}{\pgfqpoint{0.014731in}{-0.014731in}}%
\pgfpathcurveto{\pgfqpoint{0.018638in}{-0.010825in}}{\pgfqpoint{0.020833in}{-0.005525in}}{\pgfqpoint{0.020833in}{0.000000in}}%
\pgfpathcurveto{\pgfqpoint{0.020833in}{0.005525in}}{\pgfqpoint{0.018638in}{0.010825in}}{\pgfqpoint{0.014731in}{0.014731in}}%
\pgfpathcurveto{\pgfqpoint{0.010825in}{0.018638in}}{\pgfqpoint{0.005525in}{0.020833in}}{\pgfqpoint{0.000000in}{0.020833in}}%
\pgfpathcurveto{\pgfqpoint{-0.005525in}{0.020833in}}{\pgfqpoint{-0.010825in}{0.018638in}}{\pgfqpoint{-0.014731in}{0.014731in}}%
\pgfpathcurveto{\pgfqpoint{-0.018638in}{0.010825in}}{\pgfqpoint{-0.020833in}{0.005525in}}{\pgfqpoint{-0.020833in}{0.000000in}}%
\pgfpathcurveto{\pgfqpoint{-0.020833in}{-0.005525in}}{\pgfqpoint{-0.018638in}{-0.010825in}}{\pgfqpoint{-0.014731in}{-0.014731in}}%
\pgfpathcurveto{\pgfqpoint{-0.010825in}{-0.018638in}}{\pgfqpoint{-0.005525in}{-0.020833in}}{\pgfqpoint{0.000000in}{-0.020833in}}%
\pgfpathlineto{\pgfqpoint{0.000000in}{-0.020833in}}%
\pgfpathclose%
\pgfusepath{stroke,fill}%
}%
\begin{pgfscope}%
\pgfsys@transformshift{1.185135in}{1.365267in}%
\pgfsys@useobject{currentmarker}{}%
\end{pgfscope}%
\end{pgfscope}%
\begin{pgfscope}%
\pgfpathrectangle{\pgfqpoint{0.450000in}{0.780000in}}{\pgfqpoint{0.800000in}{0.800000in}}%
\pgfusepath{clip}%
\pgfsetbuttcap%
\pgfsetroundjoin%
\definecolor{currentfill}{rgb}{0.031373,0.564706,0.600000}%
\pgfsetfillcolor{currentfill}%
\pgfsetlinewidth{0.803000pt}%
\definecolor{currentstroke}{rgb}{0.031373,0.564706,0.600000}%
\pgfsetstrokecolor{currentstroke}%
\pgfsetdash{}{0pt}%
\pgfsys@defobject{currentmarker}{\pgfqpoint{-0.020833in}{-0.020833in}}{\pgfqpoint{0.020833in}{0.020833in}}{%
\pgfpathmoveto{\pgfqpoint{0.000000in}{-0.020833in}}%
\pgfpathcurveto{\pgfqpoint{0.005525in}{-0.020833in}}{\pgfqpoint{0.010825in}{-0.018638in}}{\pgfqpoint{0.014731in}{-0.014731in}}%
\pgfpathcurveto{\pgfqpoint{0.018638in}{-0.010825in}}{\pgfqpoint{0.020833in}{-0.005525in}}{\pgfqpoint{0.020833in}{0.000000in}}%
\pgfpathcurveto{\pgfqpoint{0.020833in}{0.005525in}}{\pgfqpoint{0.018638in}{0.010825in}}{\pgfqpoint{0.014731in}{0.014731in}}%
\pgfpathcurveto{\pgfqpoint{0.010825in}{0.018638in}}{\pgfqpoint{0.005525in}{0.020833in}}{\pgfqpoint{0.000000in}{0.020833in}}%
\pgfpathcurveto{\pgfqpoint{-0.005525in}{0.020833in}}{\pgfqpoint{-0.010825in}{0.018638in}}{\pgfqpoint{-0.014731in}{0.014731in}}%
\pgfpathcurveto{\pgfqpoint{-0.018638in}{0.010825in}}{\pgfqpoint{-0.020833in}{0.005525in}}{\pgfqpoint{-0.020833in}{0.000000in}}%
\pgfpathcurveto{\pgfqpoint{-0.020833in}{-0.005525in}}{\pgfqpoint{-0.018638in}{-0.010825in}}{\pgfqpoint{-0.014731in}{-0.014731in}}%
\pgfpathcurveto{\pgfqpoint{-0.010825in}{-0.018638in}}{\pgfqpoint{-0.005525in}{-0.020833in}}{\pgfqpoint{0.000000in}{-0.020833in}}%
\pgfpathlineto{\pgfqpoint{0.000000in}{-0.020833in}}%
\pgfpathclose%
\pgfusepath{stroke,fill}%
}%
\begin{pgfscope}%
\pgfsys@transformshift{1.185135in}{1.371242in}%
\pgfsys@useobject{currentmarker}{}%
\end{pgfscope}%
\end{pgfscope}%
\begin{pgfscope}%
\pgfpathrectangle{\pgfqpoint{0.450000in}{0.780000in}}{\pgfqpoint{0.800000in}{0.800000in}}%
\pgfusepath{clip}%
\pgfsetbuttcap%
\pgfsetmiterjoin%
\definecolor{currentfill}{rgb}{0.862745,0.223529,0.466667}%
\pgfsetfillcolor{currentfill}%
\pgfsetlinewidth{0.803000pt}%
\definecolor{currentstroke}{rgb}{0.862745,0.223529,0.466667}%
\pgfsetstrokecolor{currentstroke}%
\pgfsetdash{}{0pt}%
\pgfsys@defobject{currentmarker}{\pgfqpoint{-0.020833in}{-0.020833in}}{\pgfqpoint{0.020833in}{0.020833in}}{%
\pgfpathmoveto{\pgfqpoint{-0.020833in}{-0.020833in}}%
\pgfpathlineto{\pgfqpoint{0.020833in}{-0.020833in}}%
\pgfpathlineto{\pgfqpoint{0.020833in}{0.020833in}}%
\pgfpathlineto{\pgfqpoint{-0.020833in}{0.020833in}}%
\pgfpathlineto{\pgfqpoint{-0.020833in}{-0.020833in}}%
\pgfpathclose%
\pgfusepath{stroke,fill}%
}%
\begin{pgfscope}%
\pgfsys@transformshift{1.185135in}{1.199508in}%
\pgfsys@useobject{currentmarker}{}%
\end{pgfscope}%
\end{pgfscope}%
\begin{pgfscope}%
\pgfsetbuttcap%
\pgfsetroundjoin%
\definecolor{currentfill}{rgb}{0.031373,0.564706,0.600000}%
\pgfsetfillcolor{currentfill}%
\pgfsetlinewidth{1.003750pt}%
\definecolor{currentstroke}{rgb}{0.031373,0.564706,0.600000}%
\pgfsetstrokecolor{currentstroke}%
\pgfsetdash{}{0pt}%
\pgfsys@defobject{currentmarker}{\pgfqpoint{-0.020833in}{-0.020833in}}{\pgfqpoint{0.020833in}{0.020833in}}{%
\pgfpathmoveto{\pgfqpoint{0.000000in}{-0.020833in}}%
\pgfpathcurveto{\pgfqpoint{0.005525in}{-0.020833in}}{\pgfqpoint{0.010825in}{-0.018638in}}{\pgfqpoint{0.014731in}{-0.014731in}}%
\pgfpathcurveto{\pgfqpoint{0.018638in}{-0.010825in}}{\pgfqpoint{0.020833in}{-0.005525in}}{\pgfqpoint{0.020833in}{0.000000in}}%
\pgfpathcurveto{\pgfqpoint{0.020833in}{0.005525in}}{\pgfqpoint{0.018638in}{0.010825in}}{\pgfqpoint{0.014731in}{0.014731in}}%
\pgfpathcurveto{\pgfqpoint{0.010825in}{0.018638in}}{\pgfqpoint{0.005525in}{0.020833in}}{\pgfqpoint{0.000000in}{0.020833in}}%
\pgfpathcurveto{\pgfqpoint{-0.005525in}{0.020833in}}{\pgfqpoint{-0.010825in}{0.018638in}}{\pgfqpoint{-0.014731in}{0.014731in}}%
\pgfpathcurveto{\pgfqpoint{-0.018638in}{0.010825in}}{\pgfqpoint{-0.020833in}{0.005525in}}{\pgfqpoint{-0.020833in}{0.000000in}}%
\pgfpathcurveto{\pgfqpoint{-0.020833in}{-0.005525in}}{\pgfqpoint{-0.018638in}{-0.010825in}}{\pgfqpoint{-0.014731in}{-0.014731in}}%
\pgfpathcurveto{\pgfqpoint{-0.010825in}{-0.018638in}}{\pgfqpoint{-0.005525in}{-0.020833in}}{\pgfqpoint{0.000000in}{-0.020833in}}%
\pgfpathlineto{\pgfqpoint{0.000000in}{-0.020833in}}%
\pgfpathclose%
\pgfusepath{stroke,fill}%
}%
\begin{pgfscope}%
\pgfsys@transformshift{0.189089in}{0.246667in}%
\pgfsys@useobject{currentmarker}{}%
\end{pgfscope}%
\end{pgfscope}%
\begin{pgfscope}%
\definecolor{textcolor}{rgb}{0.101961,0.101961,0.098039}%
\pgfsetstrokecolor{textcolor}%
\pgfsetfillcolor{textcolor}%
\pgftext[x=0.283534in,y=0.207778in,left,base]{\color{textcolor}\rmfamily\fontsize{8.000000}{9.600000}\selectfont \(\displaystyle +\RotTerm{\mathbf{R}_{\mathrm{QK}}}\)}%
\end{pgfscope}%
\begin{pgfscope}%
\pgfsetbuttcap%
\pgfsetmiterjoin%
\definecolor{currentfill}{rgb}{0.862745,0.223529,0.466667}%
\pgfsetfillcolor{currentfill}%
\pgfsetlinewidth{1.003750pt}%
\definecolor{currentstroke}{rgb}{0.862745,0.223529,0.466667}%
\pgfsetstrokecolor{currentstroke}%
\pgfsetdash{}{0pt}%
\pgfsys@defobject{currentmarker}{\pgfqpoint{-0.020833in}{-0.020833in}}{\pgfqpoint{0.020833in}{0.020833in}}{%
\pgfpathmoveto{\pgfqpoint{-0.020833in}{-0.020833in}}%
\pgfpathlineto{\pgfqpoint{0.020833in}{-0.020833in}}%
\pgfpathlineto{\pgfqpoint{0.020833in}{0.020833in}}%
\pgfpathlineto{\pgfqpoint{-0.020833in}{0.020833in}}%
\pgfpathlineto{\pgfqpoint{-0.020833in}{-0.020833in}}%
\pgfpathclose%
\pgfusepath{stroke,fill}%
}%
\begin{pgfscope}%
\pgfsys@transformshift{0.789321in}{0.246667in}%
\pgfsys@useobject{currentmarker}{}%
\end{pgfscope}%
\end{pgfscope}%
\begin{pgfscope}%
\definecolor{textcolor}{rgb}{0.101961,0.101961,0.098039}%
\pgfsetstrokecolor{textcolor}%
\pgfsetfillcolor{textcolor}%
\pgftext[x=0.883765in,y=0.207778in,left,base]{\color{textcolor}\rmfamily\fontsize{8.000000}{9.600000}\selectfont \(\displaystyle +\RescaleTerm{\mathbf{D}_{\mathrm{QK}}}\)}%
\end{pgfscope}%
\begin{pgfscope}%
\pgfsetbuttcap%
\pgfsetroundjoin%
\definecolor{currentfill}{rgb}{1.000000,1.000000,1.000000}%
\pgfsetfillcolor{currentfill}%
\pgfsetlinewidth{0.803000pt}%
\definecolor{currentstroke}{rgb}{0.321569,0.317647,0.305882}%
\pgfsetstrokecolor{currentstroke}%
\pgfsetdash{}{0pt}%
\pgfsys@defobject{currentmarker}{\pgfqpoint{-0.020833in}{-0.020833in}}{\pgfqpoint{0.020833in}{0.020833in}}{%
\pgfpathmoveto{\pgfqpoint{0.000000in}{-0.020833in}}%
\pgfpathcurveto{\pgfqpoint{0.005525in}{-0.020833in}}{\pgfqpoint{0.010825in}{-0.018638in}}{\pgfqpoint{0.014731in}{-0.014731in}}%
\pgfpathcurveto{\pgfqpoint{0.018638in}{-0.010825in}}{\pgfqpoint{0.020833in}{-0.005525in}}{\pgfqpoint{0.020833in}{0.000000in}}%
\pgfpathcurveto{\pgfqpoint{0.020833in}{0.005525in}}{\pgfqpoint{0.018638in}{0.010825in}}{\pgfqpoint{0.014731in}{0.014731in}}%
\pgfpathcurveto{\pgfqpoint{0.010825in}{0.018638in}}{\pgfqpoint{0.005525in}{0.020833in}}{\pgfqpoint{0.000000in}{0.020833in}}%
\pgfpathcurveto{\pgfqpoint{-0.005525in}{0.020833in}}{\pgfqpoint{-0.010825in}{0.018638in}}{\pgfqpoint{-0.014731in}{0.014731in}}%
\pgfpathcurveto{\pgfqpoint{-0.018638in}{0.010825in}}{\pgfqpoint{-0.020833in}{0.005525in}}{\pgfqpoint{-0.020833in}{0.000000in}}%
\pgfpathcurveto{\pgfqpoint{-0.020833in}{-0.005525in}}{\pgfqpoint{-0.018638in}{-0.010825in}}{\pgfqpoint{-0.014731in}{-0.014731in}}%
\pgfpathcurveto{\pgfqpoint{-0.010825in}{-0.018638in}}{\pgfqpoint{-0.005525in}{-0.020833in}}{\pgfqpoint{0.000000in}{-0.020833in}}%
\pgfpathlineto{\pgfqpoint{0.000000in}{-0.020833in}}%
\pgfpathclose%
\pgfusepath{stroke,fill}%
}%
\begin{pgfscope}%
\pgfsys@transformshift{0.550223in}{0.096667in}%
\pgfsys@useobject{currentmarker}{}%
\end{pgfscope}%
\end{pgfscope}%
\begin{pgfscope}%
\definecolor{textcolor}{rgb}{0.101961,0.101961,0.098039}%
\pgfsetstrokecolor{textcolor}%
\pgfsetfillcolor{textcolor}%
\pgftext[x=0.644667in,y=0.057778in,left,base]{\color{textcolor}\rmfamily\fontsize{8.000000}{9.600000}\selectfont naive}%
\end{pgfscope}%
\end{pgfpicture}%
\makeatother%
\endgroup%

%% file: figures/pgf/fig_geometry_key.pgf
\providecommand{\mathdefault}[1]{#1}%
\begingroup%
\makeatletter%
\begin{pgfpicture}%
\pgfpathrectangle{\pgfpointorigin}{\pgfqpoint{1.370000in}{1.640000in}}%
\pgfusepath{use as bounding box, clip}%
\begin{pgfscope}%
\pgfsetbuttcap%
\pgfsetmiterjoin%
\definecolor{currentfill}{rgb}{1.000000,1.000000,1.000000}%
\pgfsetfillcolor{currentfill}%
\pgfsetlinewidth{0.000000pt}%
\definecolor{currentstroke}{rgb}{1.000000,1.000000,1.000000}%
\pgfsetstrokecolor{currentstroke}%
\pgfsetdash{}{0pt}%
\pgfpathmoveto{\pgfqpoint{0.000000in}{0.000000in}}%
\pgfpathlineto{\pgfqpoint{1.370000in}{0.000000in}}%
\pgfpathlineto{\pgfqpoint{1.370000in}{1.640000in}}%
\pgfpathlineto{\pgfqpoint{0.000000in}{1.640000in}}%
\pgfpathlineto{\pgfqpoint{0.000000in}{0.000000in}}%
\pgfpathclose%
\pgfusepath{fill}%
\end{pgfscope}%
\begin{pgfscope}%
\pgfsetbuttcap%
\pgfsetmiterjoin%
\definecolor{currentfill}{rgb}{1.000000,1.000000,1.000000}%
\pgfsetfillcolor{currentfill}%
\pgfsetlinewidth{0.000000pt}%
\definecolor{currentstroke}{rgb}{0.000000,0.000000,0.000000}%
\pgfsetstrokecolor{currentstroke}%
\pgfsetstrokeopacity{0.000000}%
\pgfsetdash{}{0pt}%
\pgfpathmoveto{\pgfqpoint{0.450000in}{0.780000in}}%
\pgfpathlineto{\pgfqpoint{1.250000in}{0.780000in}}%
\pgfpathlineto{\pgfqpoint{1.250000in}{1.580000in}}%
\pgfpathlineto{\pgfqpoint{0.450000in}{1.580000in}}%
\pgfpathlineto{\pgfqpoint{0.450000in}{0.780000in}}%
\pgfpathclose%
\pgfusepath{fill}%
\end{pgfscope}%
\begin{pgfscope}%
\definecolor{textcolor}{rgb}{0.101961,0.101961,0.098039}%
\pgfsetstrokecolor{textcolor}%
\pgfsetfillcolor{textcolor}%
\pgftext[x=0.564286in,y=0.731389in,right,top,rotate=45.000000]{\color{textcolor}\rmfamily\fontsize{8.000000}{9.600000}\selectfont LA}%
\end{pgfscope}%
\begin{pgfscope}%
\definecolor{textcolor}{rgb}{0.101961,0.101961,0.098039}%
\pgfsetstrokecolor{textcolor}%
\pgfsetfillcolor{textcolor}%
\pgftext[x=0.754762in,y=0.731389in,right,top,rotate=45.000000]{\color{textcolor}\rmfamily\fontsize{8.000000}{9.600000}\selectfont LA+\(\displaystyle \beta\)}%
\end{pgfscope}%
\begin{pgfscope}%
\definecolor{textcolor}{rgb}{0.101961,0.101961,0.098039}%
\pgfsetstrokecolor{textcolor}%
\pgfsetfillcolor{textcolor}%
\pgftext[x=0.945238in,y=0.731389in,right,top,rotate=45.000000]{\color{textcolor}\rmfamily\fontsize{8.000000}{9.600000}\selectfont M2}%
\end{pgfscope}%
\begin{pgfscope}%
\definecolor{textcolor}{rgb}{0.101961,0.101961,0.098039}%
\pgfsetstrokecolor{textcolor}%
\pgfsetfillcolor{textcolor}%
\pgftext[x=1.135714in,y=0.731389in,right,top,rotate=45.000000]{\color{textcolor}\rmfamily\fontsize{8.000000}{9.600000}\selectfont M2+\(\displaystyle \beta\)}%
\end{pgfscope}%
\begin{pgfscope}%
\pgfpathrectangle{\pgfqpoint{0.450000in}{0.780000in}}{\pgfqpoint{0.800000in}{0.800000in}}%
\pgfusepath{clip}%
\pgfsetrectcap%
\pgfsetroundjoin%
\pgfsetlinewidth{0.501875pt}%
\definecolor{currentstroke}{rgb}{0.886275,0.882353,0.866667}%
\pgfsetstrokecolor{currentstroke}%
\pgfsetdash{}{0pt}%
\pgfpathmoveto{\pgfqpoint{0.450000in}{0.780000in}}%
\pgfpathlineto{\pgfqpoint{1.250000in}{0.780000in}}%
\pgfusepath{stroke}%
\end{pgfscope}%
\begin{pgfscope}%
\pgfsetbuttcap%
\pgfsetroundjoin%
\definecolor{currentfill}{rgb}{0.321569,0.317647,0.305882}%
\pgfsetfillcolor{currentfill}%
\pgfsetlinewidth{0.602250pt}%
\definecolor{currentstroke}{rgb}{0.321569,0.317647,0.305882}%
\pgfsetstrokecolor{currentstroke}%
\pgfsetdash{}{0pt}%
\pgfsys@defobject{currentmarker}{\pgfqpoint{-0.034722in}{0.000000in}}{\pgfqpoint{-0.000000in}{0.000000in}}{%
\pgfpathmoveto{\pgfqpoint{-0.000000in}{0.000000in}}%
\pgfpathlineto{\pgfqpoint{-0.034722in}{0.000000in}}%
\pgfusepath{stroke,fill}%
}%
\begin{pgfscope}%
\pgfsys@transformshift{0.450000in}{0.780000in}%
\pgfsys@useobject{currentmarker}{}%
\end{pgfscope}%
\end{pgfscope}%
\begin{pgfscope}%
\definecolor{textcolor}{rgb}{0.101961,0.101961,0.098039}%
\pgfsetstrokecolor{textcolor}%
\pgfsetfillcolor{textcolor}%
\pgftext[x=0.311111in, y=0.742112in, left, base]{\color{textcolor}\rmfamily\fontsize{8.000000}{9.600000}\selectfont 0}%
\end{pgfscope}%
\begin{pgfscope}%
\pgfpathrectangle{\pgfqpoint{0.450000in}{0.780000in}}{\pgfqpoint{0.800000in}{0.800000in}}%
\pgfusepath{clip}%
\pgfsetrectcap%
\pgfsetroundjoin%
\pgfsetlinewidth{0.501875pt}%
\definecolor{currentstroke}{rgb}{0.886275,0.882353,0.866667}%
\pgfsetstrokecolor{currentstroke}%
\pgfsetdash{}{0pt}%
\pgfpathmoveto{\pgfqpoint{0.450000in}{1.180000in}}%
\pgfpathlineto{\pgfqpoint{1.250000in}{1.180000in}}%
\pgfusepath{stroke}%
\end{pgfscope}%
\begin{pgfscope}%
\pgfsetbuttcap%
\pgfsetroundjoin%
\definecolor{currentfill}{rgb}{0.321569,0.317647,0.305882}%
\pgfsetfillcolor{currentfill}%
\pgfsetlinewidth{0.602250pt}%
\definecolor{currentstroke}{rgb}{0.321569,0.317647,0.305882}%
\pgfsetstrokecolor{currentstroke}%
\pgfsetdash{}{0pt}%
\pgfsys@defobject{currentmarker}{\pgfqpoint{-0.034722in}{0.000000in}}{\pgfqpoint{-0.000000in}{0.000000in}}{%
\pgfpathmoveto{\pgfqpoint{-0.000000in}{0.000000in}}%
\pgfpathlineto{\pgfqpoint{-0.034722in}{0.000000in}}%
\pgfusepath{stroke,fill}%
}%
\begin{pgfscope}%
\pgfsys@transformshift{0.450000in}{1.180000in}%
\pgfsys@useobject{currentmarker}{}%
\end{pgfscope}%
\end{pgfscope}%
\begin{pgfscope}%
\definecolor{textcolor}{rgb}{0.101961,0.101961,0.098039}%
\pgfsetstrokecolor{textcolor}%
\pgfsetfillcolor{textcolor}%
\pgftext[x=0.255556in, y=1.142112in, left, base]{\color{textcolor}\rmfamily\fontsize{8.000000}{9.600000}\selectfont 50}%
\end{pgfscope}%
\begin{pgfscope}%
\pgfpathrectangle{\pgfqpoint{0.450000in}{0.780000in}}{\pgfqpoint{0.800000in}{0.800000in}}%
\pgfusepath{clip}%
\pgfsetrectcap%
\pgfsetroundjoin%
\pgfsetlinewidth{0.501875pt}%
\definecolor{currentstroke}{rgb}{0.886275,0.882353,0.866667}%
\pgfsetstrokecolor{currentstroke}%
\pgfsetdash{}{0pt}%
\pgfpathmoveto{\pgfqpoint{0.450000in}{1.580000in}}%
\pgfpathlineto{\pgfqpoint{1.250000in}{1.580000in}}%
\pgfusepath{stroke}%
\end{pgfscope}%
\begin{pgfscope}%
\pgfsetbuttcap%
\pgfsetroundjoin%
\definecolor{currentfill}{rgb}{0.321569,0.317647,0.305882}%
\pgfsetfillcolor{currentfill}%
\pgfsetlinewidth{0.602250pt}%
\definecolor{currentstroke}{rgb}{0.321569,0.317647,0.305882}%
\pgfsetstrokecolor{currentstroke}%
\pgfsetdash{}{0pt}%
\pgfsys@defobject{currentmarker}{\pgfqpoint{-0.034722in}{0.000000in}}{\pgfqpoint{-0.000000in}{0.000000in}}{%
\pgfpathmoveto{\pgfqpoint{-0.000000in}{0.000000in}}%
\pgfpathlineto{\pgfqpoint{-0.034722in}{0.000000in}}%
\pgfusepath{stroke,fill}%
}%
\begin{pgfscope}%
\pgfsys@transformshift{0.450000in}{1.580000in}%
\pgfsys@useobject{currentmarker}{}%
\end{pgfscope}%
\end{pgfscope}%
\begin{pgfscope}%
\definecolor{textcolor}{rgb}{0.101961,0.101961,0.098039}%
\pgfsetstrokecolor{textcolor}%
\pgfsetfillcolor{textcolor}%
\pgftext[x=0.200000in, y=1.542112in, left, base]{\color{textcolor}\rmfamily\fontsize{8.000000}{9.600000}\selectfont 100}%
\end{pgfscope}%
\begin{pgfscope}%
\definecolor{textcolor}{rgb}{0.101961,0.101961,0.098039}%
\pgfsetstrokecolor{textcolor}%
\pgfsetfillcolor{textcolor}%
\pgftext[x=0.144444in,y=1.180000in,,bottom,rotate=90.000000]{\color{textcolor}\rmfamily\fontsize{9.000000}{10.800000}\selectfont recovered (\%)}%
\end{pgfscope}%
\begin{pgfscope}%
\pgfpathrectangle{\pgfqpoint{0.450000in}{0.780000in}}{\pgfqpoint{0.800000in}{0.800000in}}%
\pgfusepath{clip}%
\pgfsetrectcap%
\pgfsetroundjoin%
\pgfsetlinewidth{0.602250pt}%
\definecolor{currentstroke}{rgb}{0.321569,0.317647,0.305882}%
\pgfsetstrokecolor{currentstroke}%
\pgfsetdash{}{0pt}%
\pgfpathmoveto{\pgfqpoint{0.450000in}{0.780000in}}%
\pgfpathlineto{\pgfqpoint{1.250000in}{0.780000in}}%
\pgfusepath{stroke}%
\end{pgfscope}%
\begin{pgfscope}%
\pgfpathrectangle{\pgfqpoint{0.450000in}{0.780000in}}{\pgfqpoint{0.800000in}{0.800000in}}%
\pgfusepath{clip}%
\pgfsetbuttcap%
\pgfsetmiterjoin%
\definecolor{currentfill}{rgb}{0.564118,0.804118,0.820000}%
\pgfsetfillcolor{currentfill}%
\pgfsetlinewidth{0.000000pt}%
\definecolor{currentstroke}{rgb}{0.000000,0.000000,0.000000}%
\pgfsetstrokecolor{currentstroke}%
\pgfsetstrokeopacity{0.000000}%
\pgfsetdash{}{0pt}%
\pgfpathmoveto{\pgfqpoint{0.491905in}{1.141001in}}%
\pgfpathlineto{\pgfqpoint{0.564286in}{1.141001in}}%
\pgfpathlineto{\pgfqpoint{0.564286in}{1.194920in}}%
\pgfpathlineto{\pgfqpoint{0.491905in}{1.194920in}}%
\pgfpathlineto{\pgfqpoint{0.491905in}{1.141001in}}%
\pgfpathclose%
\pgfusepath{fill}%
\end{pgfscope}%
\begin{pgfscope}%
\pgfpathrectangle{\pgfqpoint{0.450000in}{0.780000in}}{\pgfqpoint{0.800000in}{0.800000in}}%
\pgfusepath{clip}%
\pgfsetbuttcap%
\pgfsetmiterjoin%
\definecolor{currentfill}{rgb}{0.564118,0.804118,0.820000}%
\pgfsetfillcolor{currentfill}%
\pgfsetlinewidth{0.000000pt}%
\definecolor{currentstroke}{rgb}{0.000000,0.000000,0.000000}%
\pgfsetstrokecolor{currentstroke}%
\pgfsetstrokeopacity{0.000000}%
\pgfsetdash{}{0pt}%
\pgfpathmoveto{\pgfqpoint{0.682381in}{1.176166in}}%
\pgfpathlineto{\pgfqpoint{0.754762in}{1.176166in}}%
\pgfpathlineto{\pgfqpoint{0.754762in}{1.298724in}}%
\pgfpathlineto{\pgfqpoint{0.682381in}{1.298724in}}%
\pgfpathlineto{\pgfqpoint{0.682381in}{1.176166in}}%
\pgfpathclose%
\pgfusepath{fill}%
\end{pgfscope}%
\begin{pgfscope}%
\pgfpathrectangle{\pgfqpoint{0.450000in}{0.780000in}}{\pgfqpoint{0.800000in}{0.800000in}}%
\pgfusepath{clip}%
\pgfsetbuttcap%
\pgfsetmiterjoin%
\definecolor{currentfill}{rgb}{0.564118,0.804118,0.820000}%
\pgfsetfillcolor{currentfill}%
\pgfsetlinewidth{0.000000pt}%
\definecolor{currentstroke}{rgb}{0.000000,0.000000,0.000000}%
\pgfsetstrokecolor{currentstroke}%
\pgfsetstrokeopacity{0.000000}%
\pgfsetdash{}{0pt}%
\pgfpathmoveto{\pgfqpoint{0.872857in}{0.871141in}}%
\pgfpathlineto{\pgfqpoint{0.945238in}{0.871141in}}%
\pgfpathlineto{\pgfqpoint{0.945238in}{1.034760in}}%
\pgfpathlineto{\pgfqpoint{0.872857in}{1.034760in}}%
\pgfpathlineto{\pgfqpoint{0.872857in}{0.871141in}}%
\pgfpathclose%
\pgfusepath{fill}%
\end{pgfscope}%
\begin{pgfscope}%
\pgfpathrectangle{\pgfqpoint{0.450000in}{0.780000in}}{\pgfqpoint{0.800000in}{0.800000in}}%
\pgfusepath{clip}%
\pgfsetbuttcap%
\pgfsetmiterjoin%
\definecolor{currentfill}{rgb}{0.938235,0.650588,0.760000}%
\pgfsetfillcolor{currentfill}%
\pgfsetlinewidth{0.000000pt}%
\definecolor{currentstroke}{rgb}{0.000000,0.000000,0.000000}%
\pgfsetstrokecolor{currentstroke}%
\pgfsetstrokeopacity{0.000000}%
\pgfsetdash{}{0pt}%
\pgfpathmoveto{\pgfqpoint{0.945238in}{1.283268in}}%
\pgfpathlineto{\pgfqpoint{1.017619in}{1.283268in}}%
\pgfpathlineto{\pgfqpoint{1.017619in}{1.293421in}}%
\pgfpathlineto{\pgfqpoint{0.945238in}{1.293421in}}%
\pgfpathlineto{\pgfqpoint{0.945238in}{1.283268in}}%
\pgfpathclose%
\pgfusepath{fill}%
\end{pgfscope}%
\begin{pgfscope}%
\pgfpathrectangle{\pgfqpoint{0.450000in}{0.780000in}}{\pgfqpoint{0.800000in}{0.800000in}}%
\pgfusepath{clip}%
\pgfsetbuttcap%
\pgfsetmiterjoin%
\definecolor{currentfill}{rgb}{0.564118,0.804118,0.820000}%
\pgfsetfillcolor{currentfill}%
\pgfsetlinewidth{0.000000pt}%
\definecolor{currentstroke}{rgb}{0.000000,0.000000,0.000000}%
\pgfsetstrokecolor{currentstroke}%
\pgfsetstrokeopacity{0.000000}%
\pgfsetdash{}{0pt}%
\pgfpathmoveto{\pgfqpoint{1.063333in}{1.194529in}}%
\pgfpathlineto{\pgfqpoint{1.135714in}{1.194529in}}%
\pgfpathlineto{\pgfqpoint{1.135714in}{1.215836in}}%
\pgfpathlineto{\pgfqpoint{1.063333in}{1.215836in}}%
\pgfpathlineto{\pgfqpoint{1.063333in}{1.194529in}}%
\pgfpathclose%
\pgfusepath{fill}%
\end{pgfscope}%
\begin{pgfscope}%
\pgfpathrectangle{\pgfqpoint{0.450000in}{0.780000in}}{\pgfqpoint{0.800000in}{0.800000in}}%
\pgfusepath{clip}%
\pgfsetbuttcap%
\pgfsetmiterjoin%
\definecolor{currentfill}{rgb}{0.031373,0.564706,0.600000}%
\pgfsetfillcolor{currentfill}%
\pgfsetlinewidth{0.000000pt}%
\definecolor{currentstroke}{rgb}{0.000000,0.000000,0.000000}%
\pgfsetstrokecolor{currentstroke}%
\pgfsetstrokeopacity{0.000000}%
\pgfsetdash{}{0pt}%
\pgfpathmoveto{\pgfqpoint{0.491905in}{0.780000in}}%
\pgfpathlineto{\pgfqpoint{0.564286in}{0.780000in}}%
\pgfpathlineto{\pgfqpoint{0.564286in}{1.141001in}}%
\pgfpathlineto{\pgfqpoint{0.491905in}{1.141001in}}%
\pgfpathlineto{\pgfqpoint{0.491905in}{0.780000in}}%
\pgfpathclose%
\pgfusepath{fill}%
\end{pgfscope}%
\begin{pgfscope}%
\pgfpathrectangle{\pgfqpoint{0.450000in}{0.780000in}}{\pgfqpoint{0.800000in}{0.800000in}}%
\pgfusepath{clip}%
\pgfsetbuttcap%
\pgfsetmiterjoin%
\definecolor{currentfill}{rgb}{0.862745,0.223529,0.466667}%
\pgfsetfillcolor{currentfill}%
\pgfsetlinewidth{0.000000pt}%
\definecolor{currentstroke}{rgb}{0.000000,0.000000,0.000000}%
\pgfsetstrokecolor{currentstroke}%
\pgfsetstrokeopacity{0.000000}%
\pgfsetdash{}{0pt}%
\pgfpathmoveto{\pgfqpoint{0.564286in}{0.780000in}}%
\pgfpathlineto{\pgfqpoint{0.636667in}{0.780000in}}%
\pgfpathlineto{\pgfqpoint{0.636667in}{1.421816in}}%
\pgfpathlineto{\pgfqpoint{0.564286in}{1.421816in}}%
\pgfpathlineto{\pgfqpoint{0.564286in}{0.780000in}}%
\pgfpathclose%
\pgfusepath{fill}%
\end{pgfscope}%
\begin{pgfscope}%
\pgfpathrectangle{\pgfqpoint{0.450000in}{0.780000in}}{\pgfqpoint{0.800000in}{0.800000in}}%
\pgfusepath{clip}%
\pgfsetbuttcap%
\pgfsetmiterjoin%
\definecolor{currentfill}{rgb}{0.031373,0.564706,0.600000}%
\pgfsetfillcolor{currentfill}%
\pgfsetlinewidth{0.000000pt}%
\definecolor{currentstroke}{rgb}{0.000000,0.000000,0.000000}%
\pgfsetstrokecolor{currentstroke}%
\pgfsetstrokeopacity{0.000000}%
\pgfsetdash{}{0pt}%
\pgfpathmoveto{\pgfqpoint{0.682381in}{0.780000in}}%
\pgfpathlineto{\pgfqpoint{0.754762in}{0.780000in}}%
\pgfpathlineto{\pgfqpoint{0.754762in}{1.176166in}}%
\pgfpathlineto{\pgfqpoint{0.682381in}{1.176166in}}%
\pgfpathlineto{\pgfqpoint{0.682381in}{0.780000in}}%
\pgfpathclose%
\pgfusepath{fill}%
\end{pgfscope}%
\begin{pgfscope}%
\pgfpathrectangle{\pgfqpoint{0.450000in}{0.780000in}}{\pgfqpoint{0.800000in}{0.800000in}}%
\pgfusepath{clip}%
\pgfsetbuttcap%
\pgfsetmiterjoin%
\definecolor{currentfill}{rgb}{0.862745,0.223529,0.466667}%
\pgfsetfillcolor{currentfill}%
\pgfsetlinewidth{0.000000pt}%
\definecolor{currentstroke}{rgb}{0.000000,0.000000,0.000000}%
\pgfsetstrokecolor{currentstroke}%
\pgfsetstrokeopacity{0.000000}%
\pgfsetdash{}{0pt}%
\pgfpathmoveto{\pgfqpoint{0.754762in}{0.780000in}}%
\pgfpathlineto{\pgfqpoint{0.827143in}{0.780000in}}%
\pgfpathlineto{\pgfqpoint{0.827143in}{1.382902in}}%
\pgfpathlineto{\pgfqpoint{0.754762in}{1.382902in}}%
\pgfpathlineto{\pgfqpoint{0.754762in}{0.780000in}}%
\pgfpathclose%
\pgfusepath{fill}%
\end{pgfscope}%
\begin{pgfscope}%
\pgfpathrectangle{\pgfqpoint{0.450000in}{0.780000in}}{\pgfqpoint{0.800000in}{0.800000in}}%
\pgfusepath{clip}%
\pgfsetbuttcap%
\pgfsetmiterjoin%
\definecolor{currentfill}{rgb}{0.031373,0.564706,0.600000}%
\pgfsetfillcolor{currentfill}%
\pgfsetlinewidth{0.000000pt}%
\definecolor{currentstroke}{rgb}{0.000000,0.000000,0.000000}%
\pgfsetstrokecolor{currentstroke}%
\pgfsetstrokeopacity{0.000000}%
\pgfsetdash{}{0pt}%
\pgfpathmoveto{\pgfqpoint{0.872857in}{0.780000in}}%
\pgfpathlineto{\pgfqpoint{0.945238in}{0.780000in}}%
\pgfpathlineto{\pgfqpoint{0.945238in}{0.871141in}}%
\pgfpathlineto{\pgfqpoint{0.872857in}{0.871141in}}%
\pgfpathlineto{\pgfqpoint{0.872857in}{0.780000in}}%
\pgfpathclose%
\pgfusepath{fill}%
\end{pgfscope}%
\begin{pgfscope}%
\pgfpathrectangle{\pgfqpoint{0.450000in}{0.780000in}}{\pgfqpoint{0.800000in}{0.800000in}}%
\pgfusepath{clip}%
\pgfsetbuttcap%
\pgfsetmiterjoin%
\definecolor{currentfill}{rgb}{0.862745,0.223529,0.466667}%
\pgfsetfillcolor{currentfill}%
\pgfsetlinewidth{0.000000pt}%
\definecolor{currentstroke}{rgb}{0.000000,0.000000,0.000000}%
\pgfsetstrokecolor{currentstroke}%
\pgfsetstrokeopacity{0.000000}%
\pgfsetdash{}{0pt}%
\pgfpathmoveto{\pgfqpoint{0.945238in}{0.780000in}}%
\pgfpathlineto{\pgfqpoint{1.017619in}{0.780000in}}%
\pgfpathlineto{\pgfqpoint{1.017619in}{1.283268in}}%
\pgfpathlineto{\pgfqpoint{0.945238in}{1.283268in}}%
\pgfpathlineto{\pgfqpoint{0.945238in}{0.780000in}}%
\pgfpathclose%
\pgfusepath{fill}%
\end{pgfscope}%
\begin{pgfscope}%
\pgfpathrectangle{\pgfqpoint{0.450000in}{0.780000in}}{\pgfqpoint{0.800000in}{0.800000in}}%
\pgfusepath{clip}%
\pgfsetbuttcap%
\pgfsetmiterjoin%
\definecolor{currentfill}{rgb}{0.031373,0.564706,0.600000}%
\pgfsetfillcolor{currentfill}%
\pgfsetlinewidth{0.000000pt}%
\definecolor{currentstroke}{rgb}{0.000000,0.000000,0.000000}%
\pgfsetstrokecolor{currentstroke}%
\pgfsetstrokeopacity{0.000000}%
\pgfsetdash{}{0pt}%
\pgfpathmoveto{\pgfqpoint{1.063333in}{0.780000in}}%
\pgfpathlineto{\pgfqpoint{1.135714in}{0.780000in}}%
\pgfpathlineto{\pgfqpoint{1.135714in}{1.194529in}}%
\pgfpathlineto{\pgfqpoint{1.063333in}{1.194529in}}%
\pgfpathlineto{\pgfqpoint{1.063333in}{0.780000in}}%
\pgfpathclose%
\pgfusepath{fill}%
\end{pgfscope}%
\begin{pgfscope}%
\pgfpathrectangle{\pgfqpoint{0.450000in}{0.780000in}}{\pgfqpoint{0.800000in}{0.800000in}}%
\pgfusepath{clip}%
\pgfsetbuttcap%
\pgfsetmiterjoin%
\definecolor{currentfill}{rgb}{0.862745,0.223529,0.466667}%
\pgfsetfillcolor{currentfill}%
\pgfsetlinewidth{0.000000pt}%
\definecolor{currentstroke}{rgb}{0.000000,0.000000,0.000000}%
\pgfsetstrokecolor{currentstroke}%
\pgfsetstrokeopacity{0.000000}%
\pgfsetdash{}{0pt}%
\pgfpathmoveto{\pgfqpoint{1.135714in}{0.780000in}}%
\pgfpathlineto{\pgfqpoint{1.208095in}{0.780000in}}%
\pgfpathlineto{\pgfqpoint{1.208095in}{1.344845in}}%
\pgfpathlineto{\pgfqpoint{1.135714in}{1.344845in}}%
\pgfpathlineto{\pgfqpoint{1.135714in}{0.780000in}}%
\pgfpathclose%
\pgfusepath{fill}%
\end{pgfscope}%
\begin{pgfscope}%
\pgfsetrectcap%
\pgfsetmiterjoin%
\pgfsetlinewidth{0.602250pt}%
\definecolor{currentstroke}{rgb}{0.321569,0.317647,0.305882}%
\pgfsetstrokecolor{currentstroke}%
\pgfsetdash{}{0pt}%
\pgfpathmoveto{\pgfqpoint{0.450000in}{0.780000in}}%
\pgfpathlineto{\pgfqpoint{0.450000in}{1.580000in}}%
\pgfusepath{stroke}%
\end{pgfscope}%
\begin{pgfscope}%
\pgfsetrectcap%
\pgfsetmiterjoin%
\pgfsetlinewidth{0.602250pt}%
\definecolor{currentstroke}{rgb}{0.321569,0.317647,0.305882}%
\pgfsetstrokecolor{currentstroke}%
\pgfsetdash{}{0pt}%
\pgfpathmoveto{\pgfqpoint{0.450000in}{0.780000in}}%
\pgfpathlineto{\pgfqpoint{1.250000in}{0.780000in}}%
\pgfusepath{stroke}%
\end{pgfscope}%
\begin{pgfscope}%
\pgfpathrectangle{\pgfqpoint{0.450000in}{0.780000in}}{\pgfqpoint{0.800000in}{0.800000in}}%
\pgfusepath{clip}%
\pgfsetbuttcap%
\pgfsetmiterjoin%
\definecolor{currentfill}{rgb}{1.000000,1.000000,1.000000}%
\pgfsetfillcolor{currentfill}%
\pgfsetlinewidth{0.000000pt}%
\definecolor{currentstroke}{rgb}{0.000000,0.000000,0.000000}%
\pgfsetstrokecolor{currentstroke}%
\pgfsetstrokeopacity{0.000000}%
\pgfsetdash{}{0pt}%
\pgfpathmoveto{\pgfqpoint{0.564286in}{1.330794in}}%
\pgfpathlineto{\pgfqpoint{0.636667in}{1.330794in}}%
\pgfpathlineto{\pgfqpoint{0.636667in}{1.421816in}}%
\pgfpathlineto{\pgfqpoint{0.564286in}{1.421816in}}%
\pgfpathlineto{\pgfqpoint{0.564286in}{1.330794in}}%
\pgfpathclose%
\pgfusepath{fill}%
\end{pgfscope}%
\begin{pgfscope}%
\pgfpathrectangle{\pgfqpoint{0.450000in}{0.780000in}}{\pgfqpoint{0.800000in}{0.800000in}}%
\pgfusepath{clip}%
\pgfsetbuttcap%
\pgfsetmiterjoin%
\definecolor{currentfill}{rgb}{1.000000,1.000000,1.000000}%
\pgfsetfillcolor{currentfill}%
\pgfsetlinewidth{0.000000pt}%
\definecolor{currentstroke}{rgb}{0.000000,0.000000,0.000000}%
\pgfsetstrokecolor{currentstroke}%
\pgfsetstrokeopacity{0.000000}%
\pgfsetdash{}{0pt}%
\pgfpathmoveto{\pgfqpoint{0.754762in}{1.346088in}}%
\pgfpathlineto{\pgfqpoint{0.827143in}{1.346088in}}%
\pgfpathlineto{\pgfqpoint{0.827143in}{1.382902in}}%
\pgfpathlineto{\pgfqpoint{0.754762in}{1.382902in}}%
\pgfpathlineto{\pgfqpoint{0.754762in}{1.346088in}}%
\pgfpathclose%
\pgfusepath{fill}%
\end{pgfscope}%
\begin{pgfscope}%
\pgfpathrectangle{\pgfqpoint{0.450000in}{0.780000in}}{\pgfqpoint{0.800000in}{0.800000in}}%
\pgfusepath{clip}%
\pgfsetbuttcap%
\pgfsetmiterjoin%
\definecolor{currentfill}{rgb}{1.000000,1.000000,1.000000}%
\pgfsetfillcolor{currentfill}%
\pgfsetlinewidth{0.000000pt}%
\definecolor{currentstroke}{rgb}{0.000000,0.000000,0.000000}%
\pgfsetstrokecolor{currentstroke}%
\pgfsetstrokeopacity{0.000000}%
\pgfsetdash{}{0pt}%
\pgfpathmoveto{\pgfqpoint{1.135714in}{1.278787in}}%
\pgfpathlineto{\pgfqpoint{1.208095in}{1.278787in}}%
\pgfpathlineto{\pgfqpoint{1.208095in}{1.344845in}}%
\pgfpathlineto{\pgfqpoint{1.135714in}{1.344845in}}%
\pgfpathlineto{\pgfqpoint{1.135714in}{1.278787in}}%
\pgfpathclose%
\pgfusepath{fill}%
\end{pgfscope}%
\begin{pgfscope}%
\pgfpathrectangle{\pgfqpoint{0.564286in}{1.330794in}}{\pgfqpoint{0.072381in}{0.091022in}}%
\pgfusepath{clip}%
\pgfsetrectcap%
\pgfsetroundjoin%
\pgfsetlinewidth{0.501875pt}%
\definecolor{currentstroke}{rgb}{0.862745,0.223529,0.466667}%
\pgfsetstrokecolor{currentstroke}%
\pgfsetdash{}{0pt}%
\pgfpathmoveto{\pgfqpoint{0.626667in}{1.320794in}}%
\pgfpathlineto{\pgfqpoint{0.636667in}{1.330794in}}%
\pgfusepath{stroke}%
\end{pgfscope}%
\begin{pgfscope}%
\pgfpathrectangle{\pgfqpoint{0.564286in}{1.330794in}}{\pgfqpoint{0.072381in}{0.091022in}}%
\pgfusepath{clip}%
\pgfsetrectcap%
\pgfsetroundjoin%
\pgfsetlinewidth{0.501875pt}%
\definecolor{currentstroke}{rgb}{0.862745,0.223529,0.466667}%
\pgfsetstrokecolor{currentstroke}%
\pgfsetdash{}{0pt}%
\pgfpathmoveto{\pgfqpoint{0.609524in}{1.320794in}}%
\pgfpathlineto{\pgfqpoint{0.636667in}{1.347937in}}%
\pgfusepath{stroke}%
\end{pgfscope}%
\begin{pgfscope}%
\pgfpathrectangle{\pgfqpoint{0.564286in}{1.330794in}}{\pgfqpoint{0.072381in}{0.091022in}}%
\pgfusepath{clip}%
\pgfsetrectcap%
\pgfsetroundjoin%
\pgfsetlinewidth{0.501875pt}%
\definecolor{currentstroke}{rgb}{0.862745,0.223529,0.466667}%
\pgfsetstrokecolor{currentstroke}%
\pgfsetdash{}{0pt}%
\pgfpathmoveto{\pgfqpoint{0.592381in}{1.320794in}}%
\pgfpathlineto{\pgfqpoint{0.636667in}{1.365080in}}%
\pgfusepath{stroke}%
\end{pgfscope}%
\begin{pgfscope}%
\pgfpathrectangle{\pgfqpoint{0.564286in}{1.330794in}}{\pgfqpoint{0.072381in}{0.091022in}}%
\pgfusepath{clip}%
\pgfsetrectcap%
\pgfsetroundjoin%
\pgfsetlinewidth{0.501875pt}%
\definecolor{currentstroke}{rgb}{0.862745,0.223529,0.466667}%
\pgfsetstrokecolor{currentstroke}%
\pgfsetdash{}{0pt}%
\pgfpathmoveto{\pgfqpoint{0.575238in}{1.320794in}}%
\pgfpathlineto{\pgfqpoint{0.636667in}{1.382222in}}%
\pgfusepath{stroke}%
\end{pgfscope}%
\begin{pgfscope}%
\pgfpathrectangle{\pgfqpoint{0.564286in}{1.330794in}}{\pgfqpoint{0.072381in}{0.091022in}}%
\pgfusepath{clip}%
\pgfsetrectcap%
\pgfsetroundjoin%
\pgfsetlinewidth{0.501875pt}%
\definecolor{currentstroke}{rgb}{0.862745,0.223529,0.466667}%
\pgfsetstrokecolor{currentstroke}%
\pgfsetdash{}{0pt}%
\pgfpathmoveto{\pgfqpoint{0.564286in}{1.326984in}}%
\pgfpathlineto{\pgfqpoint{0.636667in}{1.399365in}}%
\pgfusepath{stroke}%
\end{pgfscope}%
\begin{pgfscope}%
\pgfpathrectangle{\pgfqpoint{0.564286in}{1.330794in}}{\pgfqpoint{0.072381in}{0.091022in}}%
\pgfusepath{clip}%
\pgfsetrectcap%
\pgfsetroundjoin%
\pgfsetlinewidth{0.501875pt}%
\definecolor{currentstroke}{rgb}{0.862745,0.223529,0.466667}%
\pgfsetstrokecolor{currentstroke}%
\pgfsetdash{}{0pt}%
\pgfpathmoveto{\pgfqpoint{0.564286in}{1.344127in}}%
\pgfpathlineto{\pgfqpoint{0.636667in}{1.416508in}}%
\pgfusepath{stroke}%
\end{pgfscope}%
\begin{pgfscope}%
\pgfpathrectangle{\pgfqpoint{0.564286in}{1.330794in}}{\pgfqpoint{0.072381in}{0.091022in}}%
\pgfusepath{clip}%
\pgfsetrectcap%
\pgfsetroundjoin%
\pgfsetlinewidth{0.501875pt}%
\definecolor{currentstroke}{rgb}{0.862745,0.223529,0.466667}%
\pgfsetstrokecolor{currentstroke}%
\pgfsetdash{}{0pt}%
\pgfpathmoveto{\pgfqpoint{0.564286in}{1.361270in}}%
\pgfpathlineto{\pgfqpoint{0.634832in}{1.431816in}}%
\pgfusepath{stroke}%
\end{pgfscope}%
\begin{pgfscope}%
\pgfpathrectangle{\pgfqpoint{0.564286in}{1.330794in}}{\pgfqpoint{0.072381in}{0.091022in}}%
\pgfusepath{clip}%
\pgfsetrectcap%
\pgfsetroundjoin%
\pgfsetlinewidth{0.501875pt}%
\definecolor{currentstroke}{rgb}{0.862745,0.223529,0.466667}%
\pgfsetstrokecolor{currentstroke}%
\pgfsetdash{}{0pt}%
\pgfpathmoveto{\pgfqpoint{0.564286in}{1.378413in}}%
\pgfpathlineto{\pgfqpoint{0.617689in}{1.431816in}}%
\pgfusepath{stroke}%
\end{pgfscope}%
\begin{pgfscope}%
\pgfpathrectangle{\pgfqpoint{0.564286in}{1.330794in}}{\pgfqpoint{0.072381in}{0.091022in}}%
\pgfusepath{clip}%
\pgfsetrectcap%
\pgfsetroundjoin%
\pgfsetlinewidth{0.501875pt}%
\definecolor{currentstroke}{rgb}{0.862745,0.223529,0.466667}%
\pgfsetstrokecolor{currentstroke}%
\pgfsetdash{}{0pt}%
\pgfpathmoveto{\pgfqpoint{0.564286in}{1.395556in}}%
\pgfpathlineto{\pgfqpoint{0.600546in}{1.431816in}}%
\pgfusepath{stroke}%
\end{pgfscope}%
\begin{pgfscope}%
\pgfpathrectangle{\pgfqpoint{0.564286in}{1.330794in}}{\pgfqpoint{0.072381in}{0.091022in}}%
\pgfusepath{clip}%
\pgfsetrectcap%
\pgfsetroundjoin%
\pgfsetlinewidth{0.501875pt}%
\definecolor{currentstroke}{rgb}{0.862745,0.223529,0.466667}%
\pgfsetstrokecolor{currentstroke}%
\pgfsetdash{}{0pt}%
\pgfpathmoveto{\pgfqpoint{0.564286in}{1.412699in}}%
\pgfpathlineto{\pgfqpoint{0.583403in}{1.431816in}}%
\pgfusepath{stroke}%
\end{pgfscope}%
\begin{pgfscope}%
\pgfpathrectangle{\pgfqpoint{0.754762in}{1.346088in}}{\pgfqpoint{0.072381in}{0.036815in}}%
\pgfusepath{clip}%
\pgfsetrectcap%
\pgfsetroundjoin%
\pgfsetlinewidth{0.501875pt}%
\definecolor{currentstroke}{rgb}{0.862745,0.223529,0.466667}%
\pgfsetstrokecolor{currentstroke}%
\pgfsetdash{}{0pt}%
\pgfpathmoveto{\pgfqpoint{0.817143in}{1.336088in}}%
\pgfpathlineto{\pgfqpoint{0.827143in}{1.346088in}}%
\pgfusepath{stroke}%
\end{pgfscope}%
\begin{pgfscope}%
\pgfpathrectangle{\pgfqpoint{0.754762in}{1.346088in}}{\pgfqpoint{0.072381in}{0.036815in}}%
\pgfusepath{clip}%
\pgfsetrectcap%
\pgfsetroundjoin%
\pgfsetlinewidth{0.501875pt}%
\definecolor{currentstroke}{rgb}{0.862745,0.223529,0.466667}%
\pgfsetstrokecolor{currentstroke}%
\pgfsetdash{}{0pt}%
\pgfpathmoveto{\pgfqpoint{0.800000in}{1.336088in}}%
\pgfpathlineto{\pgfqpoint{0.827143in}{1.363231in}}%
\pgfusepath{stroke}%
\end{pgfscope}%
\begin{pgfscope}%
\pgfpathrectangle{\pgfqpoint{0.754762in}{1.346088in}}{\pgfqpoint{0.072381in}{0.036815in}}%
\pgfusepath{clip}%
\pgfsetrectcap%
\pgfsetroundjoin%
\pgfsetlinewidth{0.501875pt}%
\definecolor{currentstroke}{rgb}{0.862745,0.223529,0.466667}%
\pgfsetstrokecolor{currentstroke}%
\pgfsetdash{}{0pt}%
\pgfpathmoveto{\pgfqpoint{0.782857in}{1.336088in}}%
\pgfpathlineto{\pgfqpoint{0.827143in}{1.380373in}}%
\pgfusepath{stroke}%
\end{pgfscope}%
\begin{pgfscope}%
\pgfpathrectangle{\pgfqpoint{0.754762in}{1.346088in}}{\pgfqpoint{0.072381in}{0.036815in}}%
\pgfusepath{clip}%
\pgfsetrectcap%
\pgfsetroundjoin%
\pgfsetlinewidth{0.501875pt}%
\definecolor{currentstroke}{rgb}{0.862745,0.223529,0.466667}%
\pgfsetstrokecolor{currentstroke}%
\pgfsetdash{}{0pt}%
\pgfpathmoveto{\pgfqpoint{0.765714in}{1.336088in}}%
\pgfpathlineto{\pgfqpoint{0.822529in}{1.392902in}}%
\pgfusepath{stroke}%
\end{pgfscope}%
\begin{pgfscope}%
\pgfpathrectangle{\pgfqpoint{0.754762in}{1.346088in}}{\pgfqpoint{0.072381in}{0.036815in}}%
\pgfusepath{clip}%
\pgfsetrectcap%
\pgfsetroundjoin%
\pgfsetlinewidth{0.501875pt}%
\definecolor{currentstroke}{rgb}{0.862745,0.223529,0.466667}%
\pgfsetstrokecolor{currentstroke}%
\pgfsetdash{}{0pt}%
\pgfpathmoveto{\pgfqpoint{0.754762in}{1.342278in}}%
\pgfpathlineto{\pgfqpoint{0.805386in}{1.392902in}}%
\pgfusepath{stroke}%
\end{pgfscope}%
\begin{pgfscope}%
\pgfpathrectangle{\pgfqpoint{0.754762in}{1.346088in}}{\pgfqpoint{0.072381in}{0.036815in}}%
\pgfusepath{clip}%
\pgfsetrectcap%
\pgfsetroundjoin%
\pgfsetlinewidth{0.501875pt}%
\definecolor{currentstroke}{rgb}{0.862745,0.223529,0.466667}%
\pgfsetstrokecolor{currentstroke}%
\pgfsetdash{}{0pt}%
\pgfpathmoveto{\pgfqpoint{0.754762in}{1.359421in}}%
\pgfpathlineto{\pgfqpoint{0.788243in}{1.392902in}}%
\pgfusepath{stroke}%
\end{pgfscope}%
\begin{pgfscope}%
\pgfpathrectangle{\pgfqpoint{0.754762in}{1.346088in}}{\pgfqpoint{0.072381in}{0.036815in}}%
\pgfusepath{clip}%
\pgfsetrectcap%
\pgfsetroundjoin%
\pgfsetlinewidth{0.501875pt}%
\definecolor{currentstroke}{rgb}{0.862745,0.223529,0.466667}%
\pgfsetstrokecolor{currentstroke}%
\pgfsetdash{}{0pt}%
\pgfpathmoveto{\pgfqpoint{0.754762in}{1.376564in}}%
\pgfpathlineto{\pgfqpoint{0.771100in}{1.392902in}}%
\pgfusepath{stroke}%
\end{pgfscope}%
\begin{pgfscope}%
\pgfpathrectangle{\pgfqpoint{1.135714in}{1.278787in}}{\pgfqpoint{0.072381in}{0.066058in}}%
\pgfusepath{clip}%
\pgfsetrectcap%
\pgfsetroundjoin%
\pgfsetlinewidth{0.501875pt}%
\definecolor{currentstroke}{rgb}{0.862745,0.223529,0.466667}%
\pgfsetstrokecolor{currentstroke}%
\pgfsetdash{}{0pt}%
\pgfpathmoveto{\pgfqpoint{1.198095in}{1.268787in}}%
\pgfpathlineto{\pgfqpoint{1.208095in}{1.278787in}}%
\pgfusepath{stroke}%
\end{pgfscope}%
\begin{pgfscope}%
\pgfpathrectangle{\pgfqpoint{1.135714in}{1.278787in}}{\pgfqpoint{0.072381in}{0.066058in}}%
\pgfusepath{clip}%
\pgfsetrectcap%
\pgfsetroundjoin%
\pgfsetlinewidth{0.501875pt}%
\definecolor{currentstroke}{rgb}{0.862745,0.223529,0.466667}%
\pgfsetstrokecolor{currentstroke}%
\pgfsetdash{}{0pt}%
\pgfpathmoveto{\pgfqpoint{1.180952in}{1.268787in}}%
\pgfpathlineto{\pgfqpoint{1.208095in}{1.295930in}}%
\pgfusepath{stroke}%
\end{pgfscope}%
\begin{pgfscope}%
\pgfpathrectangle{\pgfqpoint{1.135714in}{1.278787in}}{\pgfqpoint{0.072381in}{0.066058in}}%
\pgfusepath{clip}%
\pgfsetrectcap%
\pgfsetroundjoin%
\pgfsetlinewidth{0.501875pt}%
\definecolor{currentstroke}{rgb}{0.862745,0.223529,0.466667}%
\pgfsetstrokecolor{currentstroke}%
\pgfsetdash{}{0pt}%
\pgfpathmoveto{\pgfqpoint{1.163810in}{1.268787in}}%
\pgfpathlineto{\pgfqpoint{1.208095in}{1.313073in}}%
\pgfusepath{stroke}%
\end{pgfscope}%
\begin{pgfscope}%
\pgfpathrectangle{\pgfqpoint{1.135714in}{1.278787in}}{\pgfqpoint{0.072381in}{0.066058in}}%
\pgfusepath{clip}%
\pgfsetrectcap%
\pgfsetroundjoin%
\pgfsetlinewidth{0.501875pt}%
\definecolor{currentstroke}{rgb}{0.862745,0.223529,0.466667}%
\pgfsetstrokecolor{currentstroke}%
\pgfsetdash{}{0pt}%
\pgfpathmoveto{\pgfqpoint{1.146667in}{1.268787in}}%
\pgfpathlineto{\pgfqpoint{1.208095in}{1.330215in}}%
\pgfusepath{stroke}%
\end{pgfscope}%
\begin{pgfscope}%
\pgfpathrectangle{\pgfqpoint{1.135714in}{1.278787in}}{\pgfqpoint{0.072381in}{0.066058in}}%
\pgfusepath{clip}%
\pgfsetrectcap%
\pgfsetroundjoin%
\pgfsetlinewidth{0.501875pt}%
\definecolor{currentstroke}{rgb}{0.862745,0.223529,0.466667}%
\pgfsetstrokecolor{currentstroke}%
\pgfsetdash{}{0pt}%
\pgfpathmoveto{\pgfqpoint{1.135714in}{1.274977in}}%
\pgfpathlineto{\pgfqpoint{1.208095in}{1.347358in}}%
\pgfusepath{stroke}%
\end{pgfscope}%
\begin{pgfscope}%
\pgfpathrectangle{\pgfqpoint{1.135714in}{1.278787in}}{\pgfqpoint{0.072381in}{0.066058in}}%
\pgfusepath{clip}%
\pgfsetrectcap%
\pgfsetroundjoin%
\pgfsetlinewidth{0.501875pt}%
\definecolor{currentstroke}{rgb}{0.862745,0.223529,0.466667}%
\pgfsetstrokecolor{currentstroke}%
\pgfsetdash{}{0pt}%
\pgfpathmoveto{\pgfqpoint{1.135714in}{1.292120in}}%
\pgfpathlineto{\pgfqpoint{1.198439in}{1.354845in}}%
\pgfusepath{stroke}%
\end{pgfscope}%
\begin{pgfscope}%
\pgfpathrectangle{\pgfqpoint{1.135714in}{1.278787in}}{\pgfqpoint{0.072381in}{0.066058in}}%
\pgfusepath{clip}%
\pgfsetrectcap%
\pgfsetroundjoin%
\pgfsetlinewidth{0.501875pt}%
\definecolor{currentstroke}{rgb}{0.862745,0.223529,0.466667}%
\pgfsetstrokecolor{currentstroke}%
\pgfsetdash{}{0pt}%
\pgfpathmoveto{\pgfqpoint{1.135714in}{1.309263in}}%
\pgfpathlineto{\pgfqpoint{1.181296in}{1.354845in}}%
\pgfusepath{stroke}%
\end{pgfscope}%
\begin{pgfscope}%
\pgfpathrectangle{\pgfqpoint{1.135714in}{1.278787in}}{\pgfqpoint{0.072381in}{0.066058in}}%
\pgfusepath{clip}%
\pgfsetrectcap%
\pgfsetroundjoin%
\pgfsetlinewidth{0.501875pt}%
\definecolor{currentstroke}{rgb}{0.862745,0.223529,0.466667}%
\pgfsetstrokecolor{currentstroke}%
\pgfsetdash{}{0pt}%
\pgfpathmoveto{\pgfqpoint{1.135714in}{1.326406in}}%
\pgfpathlineto{\pgfqpoint{1.164153in}{1.354845in}}%
\pgfusepath{stroke}%
\end{pgfscope}%
\begin{pgfscope}%
\pgfpathrectangle{\pgfqpoint{1.135714in}{1.278787in}}{\pgfqpoint{0.072381in}{0.066058in}}%
\pgfusepath{clip}%
\pgfsetrectcap%
\pgfsetroundjoin%
\pgfsetlinewidth{0.501875pt}%
\definecolor{currentstroke}{rgb}{0.862745,0.223529,0.466667}%
\pgfsetstrokecolor{currentstroke}%
\pgfsetdash{}{0pt}%
\pgfpathmoveto{\pgfqpoint{1.135714in}{1.343549in}}%
\pgfpathlineto{\pgfqpoint{1.147010in}{1.354845in}}%
\pgfusepath{stroke}%
\end{pgfscope}%
\begin{pgfscope}%
\pgfsetbuttcap%
\pgfsetmiterjoin%
\definecolor{currentfill}{rgb}{0.031373,0.564706,0.600000}%
\pgfsetfillcolor{currentfill}%
\pgfsetlinewidth{1.003750pt}%
\definecolor{currentstroke}{rgb}{0.031373,0.564706,0.600000}%
\pgfsetstrokecolor{currentstroke}%
\pgfsetdash{}{0pt}%
\pgfpathmoveto{\pgfqpoint{0.219801in}{0.207778in}}%
\pgfpathlineto{\pgfqpoint{0.342023in}{0.207778in}}%
\pgfpathlineto{\pgfqpoint{0.342023in}{0.285556in}}%
\pgfpathlineto{\pgfqpoint{0.219801in}{0.285556in}}%
\pgfpathlineto{\pgfqpoint{0.219801in}{0.207778in}}%
\pgfpathclose%
\pgfusepath{stroke,fill}%
\end{pgfscope}%
\begin{pgfscope}%
\definecolor{textcolor}{rgb}{0.101961,0.101961,0.098039}%
\pgfsetstrokecolor{textcolor}%
\pgfsetfillcolor{textcolor}%
\pgftext[x=0.375356in,y=0.207778in,left,base]{\color{textcolor}\rmfamily\fontsize{8.000000}{9.600000}\selectfont \(\displaystyle \RotTerm{\mathbf{R}_{\mathrm{QK}}}\)}%
\end{pgfscope}%
\begin{pgfscope}%
\pgfsetbuttcap%
\pgfsetmiterjoin%
\definecolor{currentfill}{rgb}{0.862745,0.223529,0.466667}%
\pgfsetfillcolor{currentfill}%
\pgfsetlinewidth{1.003750pt}%
\definecolor{currentstroke}{rgb}{0.862745,0.223529,0.466667}%
\pgfsetstrokecolor{currentstroke}%
\pgfsetdash{}{0pt}%
\pgfpathmoveto{\pgfqpoint{0.728210in}{0.207778in}}%
\pgfpathlineto{\pgfqpoint{0.850432in}{0.207778in}}%
\pgfpathlineto{\pgfqpoint{0.850432in}{0.285556in}}%
\pgfpathlineto{\pgfqpoint{0.728210in}{0.285556in}}%
\pgfpathlineto{\pgfqpoint{0.728210in}{0.207778in}}%
\pgfpathclose%
\pgfusepath{stroke,fill}%
\end{pgfscope}%
\begin{pgfscope}%
\definecolor{textcolor}{rgb}{0.101961,0.101961,0.098039}%
\pgfsetstrokecolor{textcolor}%
\pgfsetfillcolor{textcolor}%
\pgftext[x=0.883765in,y=0.207778in,left,base]{\color{textcolor}\rmfamily\fontsize{8.000000}{9.600000}\selectfont \(\displaystyle \RescaleTerm{\mathbf{D}_{\mathrm{QK}}}\)}%
\end{pgfscope}%
\begin{pgfscope}%
\pgfsetbuttcap%
\pgfsetmiterjoin%
\definecolor{currentfill}{rgb}{0.300000,0.300000,0.300000}%
\pgfsetfillcolor{currentfill}%
\pgfsetlinewidth{1.003750pt}%
\definecolor{currentstroke}{rgb}{0.300000,0.300000,0.300000}%
\pgfsetstrokecolor{currentstroke}%
\pgfsetdash{}{0pt}%
\pgfpathmoveto{\pgfqpoint{0.221346in}{0.052222in}}%
\pgfpathlineto{\pgfqpoint{0.343568in}{0.052222in}}%
\pgfpathlineto{\pgfqpoint{0.343568in}{0.130000in}}%
\pgfpathlineto{\pgfqpoint{0.221346in}{0.130000in}}%
\pgfpathlineto{\pgfqpoint{0.221346in}{0.052222in}}%
\pgfpathclose%
\pgfusepath{stroke,fill}%
\end{pgfscope}%
\begin{pgfscope}%
\definecolor{textcolor}{rgb}{0.101961,0.101961,0.098039}%
\pgfsetstrokecolor{textcolor}%
\pgfsetfillcolor{textcolor}%
\pgftext[x=0.376901in,y=0.052222in,left,base]{\color{textcolor}\rmfamily\fontsize{8.000000}{9.600000}\selectfont \(\displaystyle \langle\)H,C\(\displaystyle \rangle\)}%
\end{pgfscope}%
\begin{pgfscope}%
\pgfsetbuttcap%
\pgfsetmiterjoin%
\definecolor{currentfill}{rgb}{0.685000,0.685000,0.685000}%
\pgfsetfillcolor{currentfill}%
\pgfsetlinewidth{1.003750pt}%
\definecolor{currentstroke}{rgb}{0.685000,0.685000,0.685000}%
\pgfsetstrokecolor{currentstroke}%
\pgfsetdash{}{0pt}%
\pgfpathmoveto{\pgfqpoint{0.739722in}{0.052222in}}%
\pgfpathlineto{\pgfqpoint{0.861944in}{0.052222in}}%
\pgfpathlineto{\pgfqpoint{0.861944in}{0.130000in}}%
\pgfpathlineto{\pgfqpoint{0.739722in}{0.130000in}}%
\pgfpathlineto{\pgfqpoint{0.739722in}{0.052222in}}%
\pgfpathclose%
\pgfusepath{stroke,fill}%
\end{pgfscope}%
\begin{pgfscope}%
\definecolor{textcolor}{rgb}{0.101961,0.101961,0.098039}%
\pgfsetstrokecolor{textcolor}%
\pgfsetfillcolor{textcolor}%
\pgftext[x=0.895278in,y=0.052222in,left,base]{\color{textcolor}\rmfamily\fontsize{8.000000}{9.600000}\selectfont \(\displaystyle \langle\)T,C\(\displaystyle \rangle\)}%
\end{pgfscope}%
\end{pgfpicture}%
\makeatother%
\endgroup%

%% file: figures/pgf/fig_val_structure.pgf
\providecommand{\mathdefault}[1]{#1}%
\begingroup%
\makeatletter%
\begin{pgfpicture}%
\pgfpathrectangle{\pgfpointorigin}{\pgfqpoint{1.370000in}{1.640000in}}%
\pgfusepath{use as bounding box, clip}%
\begin{pgfscope}%
\pgfsetbuttcap%
\pgfsetmiterjoin%
\definecolor{currentfill}{rgb}{1.000000,1.000000,1.000000}%
\pgfsetfillcolor{currentfill}%
\pgfsetlinewidth{0.000000pt}%
\definecolor{currentstroke}{rgb}{1.000000,1.000000,1.000000}%
\pgfsetstrokecolor{currentstroke}%
\pgfsetdash{}{0pt}%
\pgfpathmoveto{\pgfqpoint{0.000000in}{0.000000in}}%
\pgfpathlineto{\pgfqpoint{1.370000in}{0.000000in}}%
\pgfpathlineto{\pgfqpoint{1.370000in}{1.640000in}}%
\pgfpathlineto{\pgfqpoint{0.000000in}{1.640000in}}%
\pgfpathlineto{\pgfqpoint{0.000000in}{0.000000in}}%
\pgfpathclose%
\pgfusepath{fill}%
\end{pgfscope}%
\begin{pgfscope}%
\pgfsetbuttcap%
\pgfsetmiterjoin%
\definecolor{currentfill}{rgb}{1.000000,1.000000,1.000000}%
\pgfsetfillcolor{currentfill}%
\pgfsetlinewidth{0.000000pt}%
\definecolor{currentstroke}{rgb}{0.000000,0.000000,0.000000}%
\pgfsetstrokecolor{currentstroke}%
\pgfsetstrokeopacity{0.000000}%
\pgfsetdash{}{0pt}%
\pgfpathmoveto{\pgfqpoint{0.450000in}{0.780000in}}%
\pgfpathlineto{\pgfqpoint{1.250000in}{0.780000in}}%
\pgfpathlineto{\pgfqpoint{1.250000in}{1.580000in}}%
\pgfpathlineto{\pgfqpoint{0.450000in}{1.580000in}}%
\pgfpathlineto{\pgfqpoint{0.450000in}{0.780000in}}%
\pgfpathclose%
\pgfusepath{fill}%
\end{pgfscope}%
\begin{pgfscope}%
\pgfpathrectangle{\pgfqpoint{0.450000in}{0.780000in}}{\pgfqpoint{0.800000in}{0.800000in}}%
\pgfusepath{clip}%
\pgfsetrectcap%
\pgfsetroundjoin%
\pgfsetlinewidth{0.501875pt}%
\definecolor{currentstroke}{rgb}{0.886275,0.882353,0.866667}%
\pgfsetstrokecolor{currentstroke}%
\pgfsetdash{}{0pt}%
\pgfpathmoveto{\pgfqpoint{0.450000in}{0.780000in}}%
\pgfpathlineto{\pgfqpoint{1.250000in}{0.780000in}}%
\pgfusepath{stroke}%
\end{pgfscope}%
\begin{pgfscope}%
\pgfsetbuttcap%
\pgfsetroundjoin%
\definecolor{currentfill}{rgb}{0.321569,0.317647,0.305882}%
\pgfsetfillcolor{currentfill}%
\pgfsetlinewidth{0.602250pt}%
\definecolor{currentstroke}{rgb}{0.321569,0.317647,0.305882}%
\pgfsetstrokecolor{currentstroke}%
\pgfsetdash{}{0pt}%
\pgfsys@defobject{currentmarker}{\pgfqpoint{-0.034722in}{0.000000in}}{\pgfqpoint{-0.000000in}{0.000000in}}{%
\pgfpathmoveto{\pgfqpoint{-0.000000in}{0.000000in}}%
\pgfpathlineto{\pgfqpoint{-0.034722in}{0.000000in}}%
\pgfusepath{stroke,fill}%
}%
\begin{pgfscope}%
\pgfsys@transformshift{0.450000in}{0.780000in}%
\pgfsys@useobject{currentmarker}{}%
\end{pgfscope}%
\end{pgfscope}%
\begin{pgfscope}%
\definecolor{textcolor}{rgb}{0.101961,0.101961,0.098039}%
\pgfsetstrokecolor{textcolor}%
\pgfsetfillcolor{textcolor}%
\pgftext[x=0.311111in, y=0.742112in, left, base]{\color{textcolor}\rmfamily\fontsize{8.000000}{9.600000}\selectfont 0}%
\end{pgfscope}%
\begin{pgfscope}%
\pgfpathrectangle{\pgfqpoint{0.450000in}{0.780000in}}{\pgfqpoint{0.800000in}{0.800000in}}%
\pgfusepath{clip}%
\pgfsetrectcap%
\pgfsetroundjoin%
\pgfsetlinewidth{0.501875pt}%
\definecolor{currentstroke}{rgb}{0.886275,0.882353,0.866667}%
\pgfsetstrokecolor{currentstroke}%
\pgfsetdash{}{0pt}%
\pgfpathmoveto{\pgfqpoint{0.450000in}{1.180000in}}%
\pgfpathlineto{\pgfqpoint{1.250000in}{1.180000in}}%
\pgfusepath{stroke}%
\end{pgfscope}%
\begin{pgfscope}%
\pgfsetbuttcap%
\pgfsetroundjoin%
\definecolor{currentfill}{rgb}{0.321569,0.317647,0.305882}%
\pgfsetfillcolor{currentfill}%
\pgfsetlinewidth{0.602250pt}%
\definecolor{currentstroke}{rgb}{0.321569,0.317647,0.305882}%
\pgfsetstrokecolor{currentstroke}%
\pgfsetdash{}{0pt}%
\pgfsys@defobject{currentmarker}{\pgfqpoint{-0.034722in}{0.000000in}}{\pgfqpoint{-0.000000in}{0.000000in}}{%
\pgfpathmoveto{\pgfqpoint{-0.000000in}{0.000000in}}%
\pgfpathlineto{\pgfqpoint{-0.034722in}{0.000000in}}%
\pgfusepath{stroke,fill}%
}%
\begin{pgfscope}%
\pgfsys@transformshift{0.450000in}{1.180000in}%
\pgfsys@useobject{currentmarker}{}%
\end{pgfscope}%
\end{pgfscope}%
\begin{pgfscope}%
\definecolor{textcolor}{rgb}{0.101961,0.101961,0.098039}%
\pgfsetstrokecolor{textcolor}%
\pgfsetfillcolor{textcolor}%
\pgftext[x=0.255556in, y=1.142112in, left, base]{\color{textcolor}\rmfamily\fontsize{8.000000}{9.600000}\selectfont 50}%
\end{pgfscope}%
\begin{pgfscope}%
\pgfpathrectangle{\pgfqpoint{0.450000in}{0.780000in}}{\pgfqpoint{0.800000in}{0.800000in}}%
\pgfusepath{clip}%
\pgfsetrectcap%
\pgfsetroundjoin%
\pgfsetlinewidth{0.501875pt}%
\definecolor{currentstroke}{rgb}{0.886275,0.882353,0.866667}%
\pgfsetstrokecolor{currentstroke}%
\pgfsetdash{}{0pt}%
\pgfpathmoveto{\pgfqpoint{0.450000in}{1.580000in}}%
\pgfpathlineto{\pgfqpoint{1.250000in}{1.580000in}}%
\pgfusepath{stroke}%
\end{pgfscope}%
\begin{pgfscope}%
\pgfsetbuttcap%
\pgfsetroundjoin%
\definecolor{currentfill}{rgb}{0.321569,0.317647,0.305882}%
\pgfsetfillcolor{currentfill}%
\pgfsetlinewidth{0.602250pt}%
\definecolor{currentstroke}{rgb}{0.321569,0.317647,0.305882}%
\pgfsetstrokecolor{currentstroke}%
\pgfsetdash{}{0pt}%
\pgfsys@defobject{currentmarker}{\pgfqpoint{-0.034722in}{0.000000in}}{\pgfqpoint{-0.000000in}{0.000000in}}{%
\pgfpathmoveto{\pgfqpoint{-0.000000in}{0.000000in}}%
\pgfpathlineto{\pgfqpoint{-0.034722in}{0.000000in}}%
\pgfusepath{stroke,fill}%
}%
\begin{pgfscope}%
\pgfsys@transformshift{0.450000in}{1.580000in}%
\pgfsys@useobject{currentmarker}{}%
\end{pgfscope}%
\end{pgfscope}%
\begin{pgfscope}%
\definecolor{textcolor}{rgb}{0.101961,0.101961,0.098039}%
\pgfsetstrokecolor{textcolor}%
\pgfsetfillcolor{textcolor}%
\pgftext[x=0.200000in, y=1.542112in, left, base]{\color{textcolor}\rmfamily\fontsize{8.000000}{9.600000}\selectfont 100}%
\end{pgfscope}%
\begin{pgfscope}%
\definecolor{textcolor}{rgb}{0.101961,0.101961,0.098039}%
\pgfsetstrokecolor{textcolor}%
\pgfsetfillcolor{textcolor}%
\pgftext[x=0.144444in,y=1.180000in,,bottom,rotate=90.000000]{\color{textcolor}\rmfamily\fontsize{9.000000}{10.800000}\selectfont recovered (\%)}%
\end{pgfscope}%
\begin{pgfscope}%
\pgfsetbuttcap%
\pgfsetmiterjoin%
\definecolor{currentfill}{rgb}{0.031373,0.564706,0.600000}%
\pgfsetfillcolor{currentfill}%
\pgfsetlinewidth{0.000000pt}%
\definecolor{currentstroke}{rgb}{0.000000,0.000000,0.000000}%
\pgfsetstrokecolor{currentstroke}%
\pgfsetstrokeopacity{0.000000}%
\pgfsetdash{}{0pt}%
\pgfpathmoveto{\pgfqpoint{0.473784in}{0.780000in}}%
\pgfpathlineto{\pgfqpoint{0.514865in}{0.780000in}}%
\pgfpathlineto{\pgfqpoint{0.514865in}{1.021790in}}%
\pgfpathlineto{\pgfqpoint{0.473784in}{1.021790in}}%
\pgfpathlineto{\pgfqpoint{0.473784in}{0.780000in}}%
\pgfpathclose%
\pgfusepath{fill}%
\end{pgfscope}%
\begin{pgfscope}%
\pgfsetbuttcap%
\pgfsetmiterjoin%
\definecolor{currentfill}{rgb}{0.862745,0.223529,0.466667}%
\pgfsetfillcolor{currentfill}%
\pgfsetlinewidth{0.000000pt}%
\definecolor{currentstroke}{rgb}{0.000000,0.000000,0.000000}%
\pgfsetstrokecolor{currentstroke}%
\pgfsetstrokeopacity{0.000000}%
\pgfsetdash{}{0pt}%
\pgfpathmoveto{\pgfqpoint{0.514865in}{0.780000in}}%
\pgfpathlineto{\pgfqpoint{0.555946in}{0.780000in}}%
\pgfpathlineto{\pgfqpoint{0.555946in}{1.019443in}}%
\pgfpathlineto{\pgfqpoint{0.514865in}{1.019443in}}%
\pgfpathlineto{\pgfqpoint{0.514865in}{0.780000in}}%
\pgfpathclose%
\pgfusepath{fill}%
\end{pgfscope}%
\begin{pgfscope}%
\pgfsetbuttcap%
\pgfsetmiterjoin%
\definecolor{currentfill}{rgb}{0.031373,0.564706,0.600000}%
\pgfsetfillcolor{currentfill}%
\pgfsetlinewidth{0.000000pt}%
\definecolor{currentstroke}{rgb}{0.000000,0.000000,0.000000}%
\pgfsetstrokecolor{currentstroke}%
\pgfsetstrokeopacity{0.000000}%
\pgfsetdash{}{0pt}%
\pgfpathmoveto{\pgfqpoint{0.581892in}{0.780000in}}%
\pgfpathlineto{\pgfqpoint{0.622973in}{0.780000in}}%
\pgfpathlineto{\pgfqpoint{0.622973in}{1.489560in}}%
\pgfpathlineto{\pgfqpoint{0.581892in}{1.489560in}}%
\pgfpathlineto{\pgfqpoint{0.581892in}{0.780000in}}%
\pgfpathclose%
\pgfusepath{fill}%
\end{pgfscope}%
\begin{pgfscope}%
\pgfsetbuttcap%
\pgfsetmiterjoin%
\definecolor{currentfill}{rgb}{0.862745,0.223529,0.466667}%
\pgfsetfillcolor{currentfill}%
\pgfsetlinewidth{0.000000pt}%
\definecolor{currentstroke}{rgb}{0.000000,0.000000,0.000000}%
\pgfsetstrokecolor{currentstroke}%
\pgfsetstrokeopacity{0.000000}%
\pgfsetdash{}{0pt}%
\pgfpathmoveto{\pgfqpoint{0.622973in}{0.780000in}}%
\pgfpathlineto{\pgfqpoint{0.664054in}{0.780000in}}%
\pgfpathlineto{\pgfqpoint{0.664054in}{1.490744in}}%
\pgfpathlineto{\pgfqpoint{0.622973in}{1.490744in}}%
\pgfpathlineto{\pgfqpoint{0.622973in}{0.780000in}}%
\pgfpathclose%
\pgfusepath{fill}%
\end{pgfscope}%
\begin{pgfscope}%
\pgfsetbuttcap%
\pgfsetmiterjoin%
\definecolor{currentfill}{rgb}{0.031373,0.564706,0.600000}%
\pgfsetfillcolor{currentfill}%
\pgfsetlinewidth{0.000000pt}%
\definecolor{currentstroke}{rgb}{0.000000,0.000000,0.000000}%
\pgfsetstrokecolor{currentstroke}%
\pgfsetstrokeopacity{0.000000}%
\pgfsetdash{}{0pt}%
\pgfpathmoveto{\pgfqpoint{0.754865in}{0.780000in}}%
\pgfpathlineto{\pgfqpoint{0.795946in}{0.780000in}}%
\pgfpathlineto{\pgfqpoint{0.795946in}{1.127276in}}%
\pgfpathlineto{\pgfqpoint{0.754865in}{1.127276in}}%
\pgfpathlineto{\pgfqpoint{0.754865in}{0.780000in}}%
\pgfpathclose%
\pgfusepath{fill}%
\end{pgfscope}%
\begin{pgfscope}%
\pgfsetbuttcap%
\pgfsetmiterjoin%
\definecolor{currentfill}{rgb}{0.862745,0.223529,0.466667}%
\pgfsetfillcolor{currentfill}%
\pgfsetlinewidth{0.000000pt}%
\definecolor{currentstroke}{rgb}{0.000000,0.000000,0.000000}%
\pgfsetstrokecolor{currentstroke}%
\pgfsetstrokeopacity{0.000000}%
\pgfsetdash{}{0pt}%
\pgfpathmoveto{\pgfqpoint{0.795946in}{0.780000in}}%
\pgfpathlineto{\pgfqpoint{0.837027in}{0.780000in}}%
\pgfpathlineto{\pgfqpoint{0.837027in}{1.141506in}}%
\pgfpathlineto{\pgfqpoint{0.795946in}{1.141506in}}%
\pgfpathlineto{\pgfqpoint{0.795946in}{0.780000in}}%
\pgfpathclose%
\pgfusepath{fill}%
\end{pgfscope}%
\begin{pgfscope}%
\pgfsetbuttcap%
\pgfsetmiterjoin%
\definecolor{currentfill}{rgb}{0.031373,0.564706,0.600000}%
\pgfsetfillcolor{currentfill}%
\pgfsetlinewidth{0.000000pt}%
\definecolor{currentstroke}{rgb}{0.000000,0.000000,0.000000}%
\pgfsetstrokecolor{currentstroke}%
\pgfsetstrokeopacity{0.000000}%
\pgfsetdash{}{0pt}%
\pgfpathmoveto{\pgfqpoint{0.862973in}{0.780000in}}%
\pgfpathlineto{\pgfqpoint{0.904054in}{0.780000in}}%
\pgfpathlineto{\pgfqpoint{0.904054in}{1.467331in}}%
\pgfpathlineto{\pgfqpoint{0.862973in}{1.467331in}}%
\pgfpathlineto{\pgfqpoint{0.862973in}{0.780000in}}%
\pgfpathclose%
\pgfusepath{fill}%
\end{pgfscope}%
\begin{pgfscope}%
\pgfsetbuttcap%
\pgfsetmiterjoin%
\definecolor{currentfill}{rgb}{0.862745,0.223529,0.466667}%
\pgfsetfillcolor{currentfill}%
\pgfsetlinewidth{0.000000pt}%
\definecolor{currentstroke}{rgb}{0.000000,0.000000,0.000000}%
\pgfsetstrokecolor{currentstroke}%
\pgfsetstrokeopacity{0.000000}%
\pgfsetdash{}{0pt}%
\pgfpathmoveto{\pgfqpoint{0.904054in}{0.780000in}}%
\pgfpathlineto{\pgfqpoint{0.945135in}{0.780000in}}%
\pgfpathlineto{\pgfqpoint{0.945135in}{1.484487in}}%
\pgfpathlineto{\pgfqpoint{0.904054in}{1.484487in}}%
\pgfpathlineto{\pgfqpoint{0.904054in}{0.780000in}}%
\pgfpathclose%
\pgfusepath{fill}%
\end{pgfscope}%
\begin{pgfscope}%
\pgfsetbuttcap%
\pgfsetmiterjoin%
\definecolor{currentfill}{rgb}{0.031373,0.564706,0.600000}%
\pgfsetfillcolor{currentfill}%
\pgfsetlinewidth{0.000000pt}%
\definecolor{currentstroke}{rgb}{0.000000,0.000000,0.000000}%
\pgfsetstrokecolor{currentstroke}%
\pgfsetstrokeopacity{0.000000}%
\pgfsetdash{}{0pt}%
\pgfpathmoveto{\pgfqpoint{1.035946in}{0.780000in}}%
\pgfpathlineto{\pgfqpoint{1.077027in}{0.780000in}}%
\pgfpathlineto{\pgfqpoint{1.077027in}{1.102509in}}%
\pgfpathlineto{\pgfqpoint{1.035946in}{1.102509in}}%
\pgfpathlineto{\pgfqpoint{1.035946in}{0.780000in}}%
\pgfpathclose%
\pgfusepath{fill}%
\end{pgfscope}%
\begin{pgfscope}%
\pgfsetbuttcap%
\pgfsetmiterjoin%
\definecolor{currentfill}{rgb}{0.862745,0.223529,0.466667}%
\pgfsetfillcolor{currentfill}%
\pgfsetlinewidth{0.000000pt}%
\definecolor{currentstroke}{rgb}{0.000000,0.000000,0.000000}%
\pgfsetstrokecolor{currentstroke}%
\pgfsetstrokeopacity{0.000000}%
\pgfsetdash{}{0pt}%
\pgfpathmoveto{\pgfqpoint{1.077027in}{0.780000in}}%
\pgfpathlineto{\pgfqpoint{1.118108in}{0.780000in}}%
\pgfpathlineto{\pgfqpoint{1.118108in}{1.109292in}}%
\pgfpathlineto{\pgfqpoint{1.077027in}{1.109292in}}%
\pgfpathlineto{\pgfqpoint{1.077027in}{0.780000in}}%
\pgfpathclose%
\pgfusepath{fill}%
\end{pgfscope}%
\begin{pgfscope}%
\pgfsetbuttcap%
\pgfsetmiterjoin%
\definecolor{currentfill}{rgb}{0.031373,0.564706,0.600000}%
\pgfsetfillcolor{currentfill}%
\pgfsetlinewidth{0.000000pt}%
\definecolor{currentstroke}{rgb}{0.000000,0.000000,0.000000}%
\pgfsetstrokecolor{currentstroke}%
\pgfsetstrokeopacity{0.000000}%
\pgfsetdash{}{0pt}%
\pgfpathmoveto{\pgfqpoint{1.144054in}{0.780000in}}%
\pgfpathlineto{\pgfqpoint{1.185135in}{0.780000in}}%
\pgfpathlineto{\pgfqpoint{1.185135in}{1.326614in}}%
\pgfpathlineto{\pgfqpoint{1.144054in}{1.326614in}}%
\pgfpathlineto{\pgfqpoint{1.144054in}{0.780000in}}%
\pgfpathclose%
\pgfusepath{fill}%
\end{pgfscope}%
\begin{pgfscope}%
\pgfsetbuttcap%
\pgfsetmiterjoin%
\definecolor{currentfill}{rgb}{0.862745,0.223529,0.466667}%
\pgfsetfillcolor{currentfill}%
\pgfsetlinewidth{0.000000pt}%
\definecolor{currentstroke}{rgb}{0.000000,0.000000,0.000000}%
\pgfsetstrokecolor{currentstroke}%
\pgfsetstrokeopacity{0.000000}%
\pgfsetdash{}{0pt}%
\pgfpathmoveto{\pgfqpoint{1.185135in}{0.780000in}}%
\pgfpathlineto{\pgfqpoint{1.226216in}{0.780000in}}%
\pgfpathlineto{\pgfqpoint{1.226216in}{1.337883in}}%
\pgfpathlineto{\pgfqpoint{1.185135in}{1.337883in}}%
\pgfpathlineto{\pgfqpoint{1.185135in}{0.780000in}}%
\pgfpathclose%
\pgfusepath{fill}%
\end{pgfscope}%
\begin{pgfscope}%
\pgfsetrectcap%
\pgfsetmiterjoin%
\pgfsetlinewidth{0.602250pt}%
\definecolor{currentstroke}{rgb}{0.321569,0.317647,0.305882}%
\pgfsetstrokecolor{currentstroke}%
\pgfsetdash{}{0pt}%
\pgfpathmoveto{\pgfqpoint{0.450000in}{0.780000in}}%
\pgfpathlineto{\pgfqpoint{0.450000in}{1.580000in}}%
\pgfusepath{stroke}%
\end{pgfscope}%
\begin{pgfscope}%
\pgfsetrectcap%
\pgfsetmiterjoin%
\pgfsetlinewidth{0.602250pt}%
\definecolor{currentstroke}{rgb}{0.321569,0.317647,0.305882}%
\pgfsetstrokecolor{currentstroke}%
\pgfsetdash{}{0pt}%
\pgfpathmoveto{\pgfqpoint{0.450000in}{0.780000in}}%
\pgfpathlineto{\pgfqpoint{1.250000in}{0.780000in}}%
\pgfusepath{stroke}%
\end{pgfscope}%
\begin{pgfscope}%
\definecolor{textcolor}{rgb}{0.101961,0.101961,0.098039}%
\pgfsetstrokecolor{textcolor}%
\pgfsetfillcolor{textcolor}%
\pgftext[x=0.514865in,y=0.731389in,right,top,rotate=60.000000]{\color{textcolor}\rmfamily\fontsize{8.000000}{9.600000}\selectfont LA}%
\end{pgfscope}%
\begin{pgfscope}%
\definecolor{textcolor}{rgb}{0.101961,0.101961,0.098039}%
\pgfsetstrokecolor{textcolor}%
\pgfsetfillcolor{textcolor}%
\pgftext[x=0.622973in,y=0.731389in,right,top,rotate=60.000000]{\color{textcolor}\rmfamily\fontsize{8.000000}{9.600000}\selectfont DN}%
\end{pgfscope}%
\begin{pgfscope}%
\definecolor{textcolor}{rgb}{0.101961,0.101961,0.098039}%
\pgfsetstrokecolor{textcolor}%
\pgfsetfillcolor{textcolor}%
\pgftext[x=0.795946in,y=0.731389in,right,top,rotate=60.000000]{\color{textcolor}\rmfamily\fontsize{8.000000}{9.600000}\selectfont M2}%
\end{pgfscope}%
\begin{pgfscope}%
\definecolor{textcolor}{rgb}{0.101961,0.101961,0.098039}%
\pgfsetstrokecolor{textcolor}%
\pgfsetfillcolor{textcolor}%
\pgftext[x=0.904054in,y=0.731389in,right,top,rotate=60.000000]{\color{textcolor}\rmfamily\fontsize{8.000000}{9.600000}\selectfont GDN}%
\end{pgfscope}%
\begin{pgfscope}%
\definecolor{textcolor}{rgb}{0.101961,0.101961,0.098039}%
\pgfsetstrokecolor{textcolor}%
\pgfsetfillcolor{textcolor}%
\pgftext[x=1.077027in,y=0.731389in,right,top,rotate=60.000000]{\color{textcolor}\rmfamily\fontsize{8.000000}{9.600000}\selectfont GLA}%
\end{pgfscope}%
\begin{pgfscope}%
\definecolor{textcolor}{rgb}{0.101961,0.101961,0.098039}%
\pgfsetstrokecolor{textcolor}%
\pgfsetfillcolor{textcolor}%
\pgftext[x=1.185135in,y=0.731389in,right,top,rotate=60.000000]{\color{textcolor}\rmfamily\fontsize{8.000000}{9.600000}\selectfont KDA}%
\end{pgfscope}%
\begin{pgfscope}%
\pgfsetbuttcap%
\pgfsetmiterjoin%
\definecolor{currentfill}{rgb}{0.031373,0.564706,0.600000}%
\pgfsetfillcolor{currentfill}%
\pgfsetlinewidth{1.003750pt}%
\definecolor{currentstroke}{rgb}{0.031373,0.564706,0.600000}%
\pgfsetstrokecolor{currentstroke}%
\pgfsetdash{}{0pt}%
\pgfpathmoveto{\pgfqpoint{0.299082in}{0.207778in}}%
\pgfpathlineto{\pgfqpoint{0.421304in}{0.207778in}}%
\pgfpathlineto{\pgfqpoint{0.421304in}{0.285556in}}%
\pgfpathlineto{\pgfqpoint{0.299082in}{0.285556in}}%
\pgfpathlineto{\pgfqpoint{0.299082in}{0.207778in}}%
\pgfpathclose%
\pgfusepath{stroke,fill}%
\end{pgfscope}%
\begin{pgfscope}%
\definecolor{textcolor}{rgb}{0.101961,0.101961,0.098039}%
\pgfsetstrokecolor{textcolor}%
\pgfsetfillcolor{textcolor}%
\pgftext[x=0.454637in,y=0.207778in,left,base]{\color{textcolor}\rmfamily\fontsize{8.000000}{9.600000}\selectfont \(\displaystyle \RotTerm{\mathbf{R}_{\mathrm{V}}}\)}%
\end{pgfscope}%
\begin{pgfscope}%
\pgfsetbuttcap%
\pgfsetmiterjoin%
\definecolor{currentfill}{rgb}{0.862745,0.223529,0.466667}%
\pgfsetfillcolor{currentfill}%
\pgfsetlinewidth{1.003750pt}%
\definecolor{currentstroke}{rgb}{0.862745,0.223529,0.466667}%
\pgfsetstrokecolor{currentstroke}%
\pgfsetdash{}{0pt}%
\pgfpathmoveto{\pgfqpoint{0.728210in}{0.207778in}}%
\pgfpathlineto{\pgfqpoint{0.850432in}{0.207778in}}%
\pgfpathlineto{\pgfqpoint{0.850432in}{0.285556in}}%
\pgfpathlineto{\pgfqpoint{0.728210in}{0.285556in}}%
\pgfpathlineto{\pgfqpoint{0.728210in}{0.207778in}}%
\pgfpathclose%
\pgfusepath{stroke,fill}%
\end{pgfscope}%
\begin{pgfscope}%
\definecolor{textcolor}{rgb}{0.101961,0.101961,0.098039}%
\pgfsetstrokecolor{textcolor}%
\pgfsetfillcolor{textcolor}%
\pgftext[x=0.883765in,y=0.207778in,left,base]{\color{textcolor}\rmfamily\fontsize{8.000000}{9.600000}\selectfont \(\displaystyle \RescaleTerm{\mathbf{D}_{\mathrm{V}}}\)}%
\end{pgfscope}%
\end{pgfpicture}%
\makeatother%
\endgroup%

%% file: figures/pgf/fig_val_norm.pgf
\providecommand{\mathdefault}[1]{#1}%
\begingroup%
\makeatletter%
\begin{pgfpicture}%
\pgfpathrectangle{\pgfpointorigin}{\pgfqpoint{1.370000in}{1.640000in}}%
\pgfusepath{use as bounding box, clip}%
\begin{pgfscope}%
\pgfsetbuttcap%
\pgfsetmiterjoin%
\definecolor{currentfill}{rgb}{1.000000,1.000000,1.000000}%
\pgfsetfillcolor{currentfill}%
\pgfsetlinewidth{0.000000pt}%
\definecolor{currentstroke}{rgb}{1.000000,1.000000,1.000000}%
\pgfsetstrokecolor{currentstroke}%
\pgfsetdash{}{0pt}%
\pgfpathmoveto{\pgfqpoint{0.000000in}{0.000000in}}%
\pgfpathlineto{\pgfqpoint{1.370000in}{0.000000in}}%
\pgfpathlineto{\pgfqpoint{1.370000in}{1.640000in}}%
\pgfpathlineto{\pgfqpoint{0.000000in}{1.640000in}}%
\pgfpathlineto{\pgfqpoint{0.000000in}{0.000000in}}%
\pgfpathclose%
\pgfusepath{fill}%
\end{pgfscope}%
\begin{pgfscope}%
\pgfsetbuttcap%
\pgfsetmiterjoin%
\definecolor{currentfill}{rgb}{1.000000,1.000000,1.000000}%
\pgfsetfillcolor{currentfill}%
\pgfsetlinewidth{0.000000pt}%
\definecolor{currentstroke}{rgb}{0.000000,0.000000,0.000000}%
\pgfsetstrokecolor{currentstroke}%
\pgfsetstrokeopacity{0.000000}%
\pgfsetdash{}{0pt}%
\pgfpathmoveto{\pgfqpoint{0.450000in}{0.780000in}}%
\pgfpathlineto{\pgfqpoint{1.250000in}{0.780000in}}%
\pgfpathlineto{\pgfqpoint{1.250000in}{1.580000in}}%
\pgfpathlineto{\pgfqpoint{0.450000in}{1.580000in}}%
\pgfpathlineto{\pgfqpoint{0.450000in}{0.780000in}}%
\pgfpathclose%
\pgfusepath{fill}%
\end{pgfscope}%
\begin{pgfscope}%
\definecolor{textcolor}{rgb}{0.101961,0.101961,0.098039}%
\pgfsetstrokecolor{textcolor}%
\pgfsetfillcolor{textcolor}%
\pgftext[x=0.514865in,y=0.731389in,right,top,rotate=60.000000]{\color{textcolor}\rmfamily\fontsize{8.000000}{9.600000}\selectfont LA+\(\displaystyle \beta\)}%
\end{pgfscope}%
\begin{pgfscope}%
\definecolor{textcolor}{rgb}{0.101961,0.101961,0.098039}%
\pgfsetstrokecolor{textcolor}%
\pgfsetfillcolor{textcolor}%
\pgftext[x=0.622973in,y=0.731389in,right,top,rotate=60.000000]{\color{textcolor}\rmfamily\fontsize{8.000000}{9.600000}\selectfont LA+\(\displaystyle \beta^\dagger\)}%
\end{pgfscope}%
\begin{pgfscope}%
\definecolor{textcolor}{rgb}{0.101961,0.101961,0.098039}%
\pgfsetstrokecolor{textcolor}%
\pgfsetfillcolor{textcolor}%
\pgftext[x=0.795946in,y=0.731389in,right,top,rotate=60.000000]{\color{textcolor}\rmfamily\fontsize{8.000000}{9.600000}\selectfont M2+\(\displaystyle \beta\)}%
\end{pgfscope}%
\begin{pgfscope}%
\definecolor{textcolor}{rgb}{0.101961,0.101961,0.098039}%
\pgfsetstrokecolor{textcolor}%
\pgfsetfillcolor{textcolor}%
\pgftext[x=0.904054in,y=0.731389in,right,top,rotate=60.000000]{\color{textcolor}\rmfamily\fontsize{8.000000}{9.600000}\selectfont M2+\(\displaystyle \beta^\dagger\)}%
\end{pgfscope}%
\begin{pgfscope}%
\definecolor{textcolor}{rgb}{0.101961,0.101961,0.098039}%
\pgfsetstrokecolor{textcolor}%
\pgfsetfillcolor{textcolor}%
\pgftext[x=1.077027in,y=0.731389in,right,top,rotate=60.000000]{\color{textcolor}\rmfamily\fontsize{8.000000}{9.600000}\selectfont GLA+\(\displaystyle \beta\)}%
\end{pgfscope}%
\begin{pgfscope}%
\definecolor{textcolor}{rgb}{0.101961,0.101961,0.098039}%
\pgfsetstrokecolor{textcolor}%
\pgfsetfillcolor{textcolor}%
\pgftext[x=1.185135in,y=0.731389in,right,top,rotate=60.000000]{\color{textcolor}\rmfamily\fontsize{8.000000}{9.600000}\selectfont GLA+\(\displaystyle \beta^\dagger\)}%
\end{pgfscope}%
\begin{pgfscope}%
\pgfpathrectangle{\pgfqpoint{0.450000in}{0.780000in}}{\pgfqpoint{0.800000in}{0.800000in}}%
\pgfusepath{clip}%
\pgfsetrectcap%
\pgfsetroundjoin%
\pgfsetlinewidth{0.501875pt}%
\definecolor{currentstroke}{rgb}{0.886275,0.882353,0.866667}%
\pgfsetstrokecolor{currentstroke}%
\pgfsetdash{}{0pt}%
\pgfpathmoveto{\pgfqpoint{0.450000in}{0.808301in}}%
\pgfpathlineto{\pgfqpoint{1.250000in}{0.808301in}}%
\pgfusepath{stroke}%
\end{pgfscope}%
\begin{pgfscope}%
\pgfsetbuttcap%
\pgfsetroundjoin%
\definecolor{currentfill}{rgb}{0.321569,0.317647,0.305882}%
\pgfsetfillcolor{currentfill}%
\pgfsetlinewidth{0.602250pt}%
\definecolor{currentstroke}{rgb}{0.321569,0.317647,0.305882}%
\pgfsetstrokecolor{currentstroke}%
\pgfsetdash{}{0pt}%
\pgfsys@defobject{currentmarker}{\pgfqpoint{-0.034722in}{0.000000in}}{\pgfqpoint{-0.000000in}{0.000000in}}{%
\pgfpathmoveto{\pgfqpoint{-0.000000in}{0.000000in}}%
\pgfpathlineto{\pgfqpoint{-0.034722in}{0.000000in}}%
\pgfusepath{stroke,fill}%
}%
\begin{pgfscope}%
\pgfsys@transformshift{0.450000in}{0.808301in}%
\pgfsys@useobject{currentmarker}{}%
\end{pgfscope}%
\end{pgfscope}%
\begin{pgfscope}%
\definecolor{textcolor}{rgb}{0.101961,0.101961,0.098039}%
\pgfsetstrokecolor{textcolor}%
\pgfsetfillcolor{textcolor}%
\pgftext[x=0.172222in, y=0.770412in, left, base]{\color{textcolor}\rmfamily\fontsize{8.000000}{9.600000}\selectfont 0.01}%
\end{pgfscope}%
\begin{pgfscope}%
\pgfpathrectangle{\pgfqpoint{0.450000in}{0.780000in}}{\pgfqpoint{0.800000in}{0.800000in}}%
\pgfusepath{clip}%
\pgfsetrectcap%
\pgfsetroundjoin%
\pgfsetlinewidth{0.501875pt}%
\definecolor{currentstroke}{rgb}{0.886275,0.882353,0.866667}%
\pgfsetstrokecolor{currentstroke}%
\pgfsetdash{}{0pt}%
\pgfpathmoveto{\pgfqpoint{0.450000in}{1.126470in}}%
\pgfpathlineto{\pgfqpoint{1.250000in}{1.126470in}}%
\pgfusepath{stroke}%
\end{pgfscope}%
\begin{pgfscope}%
\pgfsetbuttcap%
\pgfsetroundjoin%
\definecolor{currentfill}{rgb}{0.321569,0.317647,0.305882}%
\pgfsetfillcolor{currentfill}%
\pgfsetlinewidth{0.602250pt}%
\definecolor{currentstroke}{rgb}{0.321569,0.317647,0.305882}%
\pgfsetstrokecolor{currentstroke}%
\pgfsetdash{}{0pt}%
\pgfsys@defobject{currentmarker}{\pgfqpoint{-0.034722in}{0.000000in}}{\pgfqpoint{-0.000000in}{0.000000in}}{%
\pgfpathmoveto{\pgfqpoint{-0.000000in}{0.000000in}}%
\pgfpathlineto{\pgfqpoint{-0.034722in}{0.000000in}}%
\pgfusepath{stroke,fill}%
}%
\begin{pgfscope}%
\pgfsys@transformshift{0.450000in}{1.126470in}%
\pgfsys@useobject{currentmarker}{}%
\end{pgfscope}%
\end{pgfscope}%
\begin{pgfscope}%
\definecolor{textcolor}{rgb}{0.101961,0.101961,0.098039}%
\pgfsetstrokecolor{textcolor}%
\pgfsetfillcolor{textcolor}%
\pgftext[x=0.227778in, y=1.088582in, left, base]{\color{textcolor}\rmfamily\fontsize{8.000000}{9.600000}\selectfont 0.1}%
\end{pgfscope}%
\begin{pgfscope}%
\pgfpathrectangle{\pgfqpoint{0.450000in}{0.780000in}}{\pgfqpoint{0.800000in}{0.800000in}}%
\pgfusepath{clip}%
\pgfsetrectcap%
\pgfsetroundjoin%
\pgfsetlinewidth{0.501875pt}%
\definecolor{currentstroke}{rgb}{0.886275,0.882353,0.866667}%
\pgfsetstrokecolor{currentstroke}%
\pgfsetdash{}{0pt}%
\pgfpathmoveto{\pgfqpoint{0.450000in}{1.444639in}}%
\pgfpathlineto{\pgfqpoint{1.250000in}{1.444639in}}%
\pgfusepath{stroke}%
\end{pgfscope}%
\begin{pgfscope}%
\pgfsetbuttcap%
\pgfsetroundjoin%
\definecolor{currentfill}{rgb}{0.321569,0.317647,0.305882}%
\pgfsetfillcolor{currentfill}%
\pgfsetlinewidth{0.602250pt}%
\definecolor{currentstroke}{rgb}{0.321569,0.317647,0.305882}%
\pgfsetstrokecolor{currentstroke}%
\pgfsetdash{}{0pt}%
\pgfsys@defobject{currentmarker}{\pgfqpoint{-0.034722in}{0.000000in}}{\pgfqpoint{-0.000000in}{0.000000in}}{%
\pgfpathmoveto{\pgfqpoint{-0.000000in}{0.000000in}}%
\pgfpathlineto{\pgfqpoint{-0.034722in}{0.000000in}}%
\pgfusepath{stroke,fill}%
}%
\begin{pgfscope}%
\pgfsys@transformshift{0.450000in}{1.444639in}%
\pgfsys@useobject{currentmarker}{}%
\end{pgfscope}%
\end{pgfscope}%
\begin{pgfscope}%
\definecolor{textcolor}{rgb}{0.101961,0.101961,0.098039}%
\pgfsetstrokecolor{textcolor}%
\pgfsetfillcolor{textcolor}%
\pgftext[x=0.311111in, y=1.406751in, left, base]{\color{textcolor}\rmfamily\fontsize{8.000000}{9.600000}\selectfont 1}%
\end{pgfscope}%
\begin{pgfscope}%
\definecolor{textcolor}{rgb}{0.101961,0.101961,0.098039}%
\pgfsetstrokecolor{textcolor}%
\pgfsetfillcolor{textcolor}%
\pgftext[x=0.116667in,y=1.180000in,,bottom,rotate=90.000000]{\color{textcolor}\rmfamily\fontsize{9.000000}{10.800000}\selectfont \(\displaystyle \Delta\)NLL}%
\end{pgfscope}%
\begin{pgfscope}%
\pgfpathrectangle{\pgfqpoint{0.450000in}{0.780000in}}{\pgfqpoint{0.800000in}{0.800000in}}%
\pgfusepath{clip}%
\pgfsetrectcap%
\pgfsetroundjoin%
\pgfsetlinewidth{0.903375pt}%
\definecolor{currentstroke}{rgb}{0.321569,0.317647,0.305882}%
\pgfsetstrokecolor{currentstroke}%
\pgfsetdash{}{0pt}%
\pgfpathmoveto{\pgfqpoint{0.514865in}{1.199474in}}%
\pgfpathlineto{\pgfqpoint{0.514865in}{1.357180in}}%
\pgfusepath{stroke}%
\end{pgfscope}%
\begin{pgfscope}%
\pgfpathrectangle{\pgfqpoint{0.450000in}{0.780000in}}{\pgfqpoint{0.800000in}{0.800000in}}%
\pgfusepath{clip}%
\pgfsetrectcap%
\pgfsetroundjoin%
\pgfsetlinewidth{0.903375pt}%
\definecolor{currentstroke}{rgb}{0.321569,0.317647,0.305882}%
\pgfsetstrokecolor{currentstroke}%
\pgfsetdash{}{0pt}%
\pgfpathmoveto{\pgfqpoint{0.622973in}{1.402251in}}%
\pgfpathlineto{\pgfqpoint{0.622973in}{1.523973in}}%
\pgfusepath{stroke}%
\end{pgfscope}%
\begin{pgfscope}%
\pgfpathrectangle{\pgfqpoint{0.450000in}{0.780000in}}{\pgfqpoint{0.800000in}{0.800000in}}%
\pgfusepath{clip}%
\pgfsetrectcap%
\pgfsetroundjoin%
\pgfsetlinewidth{0.903375pt}%
\definecolor{currentstroke}{rgb}{0.321569,0.317647,0.305882}%
\pgfsetstrokecolor{currentstroke}%
\pgfsetdash{}{0pt}%
\pgfpathmoveto{\pgfqpoint{0.795946in}{0.836027in}}%
\pgfpathlineto{\pgfqpoint{0.795946in}{1.033958in}}%
\pgfusepath{stroke}%
\end{pgfscope}%
\begin{pgfscope}%
\pgfpathrectangle{\pgfqpoint{0.450000in}{0.780000in}}{\pgfqpoint{0.800000in}{0.800000in}}%
\pgfusepath{clip}%
\pgfsetrectcap%
\pgfsetroundjoin%
\pgfsetlinewidth{0.903375pt}%
\definecolor{currentstroke}{rgb}{0.321569,0.317647,0.305882}%
\pgfsetstrokecolor{currentstroke}%
\pgfsetdash{}{0pt}%
\pgfpathmoveto{\pgfqpoint{0.904054in}{1.111643in}}%
\pgfpathlineto{\pgfqpoint{0.904054in}{1.184202in}}%
\pgfusepath{stroke}%
\end{pgfscope}%
\begin{pgfscope}%
\pgfpathrectangle{\pgfqpoint{0.450000in}{0.780000in}}{\pgfqpoint{0.800000in}{0.800000in}}%
\pgfusepath{clip}%
\pgfsetrectcap%
\pgfsetroundjoin%
\pgfsetlinewidth{0.903375pt}%
\definecolor{currentstroke}{rgb}{0.321569,0.317647,0.305882}%
\pgfsetstrokecolor{currentstroke}%
\pgfsetdash{}{0pt}%
\pgfpathmoveto{\pgfqpoint{1.077027in}{0.868537in}}%
\pgfpathlineto{\pgfqpoint{1.077027in}{1.022388in}}%
\pgfusepath{stroke}%
\end{pgfscope}%
\begin{pgfscope}%
\pgfpathrectangle{\pgfqpoint{0.450000in}{0.780000in}}{\pgfqpoint{0.800000in}{0.800000in}}%
\pgfusepath{clip}%
\pgfsetrectcap%
\pgfsetroundjoin%
\pgfsetlinewidth{0.903375pt}%
\definecolor{currentstroke}{rgb}{0.321569,0.317647,0.305882}%
\pgfsetstrokecolor{currentstroke}%
\pgfsetdash{}{0pt}%
\pgfpathmoveto{\pgfqpoint{1.185135in}{1.216728in}}%
\pgfpathlineto{\pgfqpoint{1.185135in}{1.256248in}}%
\pgfusepath{stroke}%
\end{pgfscope}%
\begin{pgfscope}%
\pgfsetrectcap%
\pgfsetmiterjoin%
\pgfsetlinewidth{0.602250pt}%
\definecolor{currentstroke}{rgb}{0.321569,0.317647,0.305882}%
\pgfsetstrokecolor{currentstroke}%
\pgfsetdash{}{0pt}%
\pgfpathmoveto{\pgfqpoint{0.450000in}{0.780000in}}%
\pgfpathlineto{\pgfqpoint{0.450000in}{1.580000in}}%
\pgfusepath{stroke}%
\end{pgfscope}%
\begin{pgfscope}%
\pgfsetrectcap%
\pgfsetmiterjoin%
\pgfsetlinewidth{0.602250pt}%
\definecolor{currentstroke}{rgb}{0.321569,0.317647,0.305882}%
\pgfsetstrokecolor{currentstroke}%
\pgfsetdash{}{0pt}%
\pgfpathmoveto{\pgfqpoint{0.450000in}{0.780000in}}%
\pgfpathlineto{\pgfqpoint{1.250000in}{0.780000in}}%
\pgfusepath{stroke}%
\end{pgfscope}%
\begin{pgfscope}%
\pgfpathrectangle{\pgfqpoint{0.450000in}{0.780000in}}{\pgfqpoint{0.800000in}{0.800000in}}%
\pgfusepath{clip}%
\pgfsetbuttcap%
\pgfsetroundjoin%
\definecolor{currentfill}{rgb}{1.000000,1.000000,1.000000}%
\pgfsetfillcolor{currentfill}%
\pgfsetlinewidth{0.803000pt}%
\definecolor{currentstroke}{rgb}{0.321569,0.317647,0.305882}%
\pgfsetstrokecolor{currentstroke}%
\pgfsetdash{}{0pt}%
\pgfsys@defobject{currentmarker}{\pgfqpoint{-0.020833in}{-0.020833in}}{\pgfqpoint{0.020833in}{0.020833in}}{%
\pgfpathmoveto{\pgfqpoint{0.000000in}{-0.020833in}}%
\pgfpathcurveto{\pgfqpoint{0.005525in}{-0.020833in}}{\pgfqpoint{0.010825in}{-0.018638in}}{\pgfqpoint{0.014731in}{-0.014731in}}%
\pgfpathcurveto{\pgfqpoint{0.018638in}{-0.010825in}}{\pgfqpoint{0.020833in}{-0.005525in}}{\pgfqpoint{0.020833in}{0.000000in}}%
\pgfpathcurveto{\pgfqpoint{0.020833in}{0.005525in}}{\pgfqpoint{0.018638in}{0.010825in}}{\pgfqpoint{0.014731in}{0.014731in}}%
\pgfpathcurveto{\pgfqpoint{0.010825in}{0.018638in}}{\pgfqpoint{0.005525in}{0.020833in}}{\pgfqpoint{0.000000in}{0.020833in}}%
\pgfpathcurveto{\pgfqpoint{-0.005525in}{0.020833in}}{\pgfqpoint{-0.010825in}{0.018638in}}{\pgfqpoint{-0.014731in}{0.014731in}}%
\pgfpathcurveto{\pgfqpoint{-0.018638in}{0.010825in}}{\pgfqpoint{-0.020833in}{0.005525in}}{\pgfqpoint{-0.020833in}{0.000000in}}%
\pgfpathcurveto{\pgfqpoint{-0.020833in}{-0.005525in}}{\pgfqpoint{-0.018638in}{-0.010825in}}{\pgfqpoint{-0.014731in}{-0.014731in}}%
\pgfpathcurveto{\pgfqpoint{-0.010825in}{-0.018638in}}{\pgfqpoint{-0.005525in}{-0.020833in}}{\pgfqpoint{0.000000in}{-0.020833in}}%
\pgfpathlineto{\pgfqpoint{0.000000in}{-0.020833in}}%
\pgfpathclose%
\pgfusepath{stroke,fill}%
}%
\begin{pgfscope}%
\pgfsys@transformshift{0.514865in}{1.357180in}%
\pgfsys@useobject{currentmarker}{}%
\end{pgfscope}%
\end{pgfscope}%
\begin{pgfscope}%
\pgfpathrectangle{\pgfqpoint{0.450000in}{0.780000in}}{\pgfqpoint{0.800000in}{0.800000in}}%
\pgfusepath{clip}%
\pgfsetbuttcap%
\pgfsetroundjoin%
\definecolor{currentfill}{rgb}{0.031373,0.564706,0.600000}%
\pgfsetfillcolor{currentfill}%
\pgfsetlinewidth{0.803000pt}%
\definecolor{currentstroke}{rgb}{0.031373,0.564706,0.600000}%
\pgfsetstrokecolor{currentstroke}%
\pgfsetdash{}{0pt}%
\pgfsys@defobject{currentmarker}{\pgfqpoint{-0.020833in}{-0.020833in}}{\pgfqpoint{0.020833in}{0.020833in}}{%
\pgfpathmoveto{\pgfqpoint{0.000000in}{-0.020833in}}%
\pgfpathcurveto{\pgfqpoint{0.005525in}{-0.020833in}}{\pgfqpoint{0.010825in}{-0.018638in}}{\pgfqpoint{0.014731in}{-0.014731in}}%
\pgfpathcurveto{\pgfqpoint{0.018638in}{-0.010825in}}{\pgfqpoint{0.020833in}{-0.005525in}}{\pgfqpoint{0.020833in}{0.000000in}}%
\pgfpathcurveto{\pgfqpoint{0.020833in}{0.005525in}}{\pgfqpoint{0.018638in}{0.010825in}}{\pgfqpoint{0.014731in}{0.014731in}}%
\pgfpathcurveto{\pgfqpoint{0.010825in}{0.018638in}}{\pgfqpoint{0.005525in}{0.020833in}}{\pgfqpoint{0.000000in}{0.020833in}}%
\pgfpathcurveto{\pgfqpoint{-0.005525in}{0.020833in}}{\pgfqpoint{-0.010825in}{0.018638in}}{\pgfqpoint{-0.014731in}{0.014731in}}%
\pgfpathcurveto{\pgfqpoint{-0.018638in}{0.010825in}}{\pgfqpoint{-0.020833in}{0.005525in}}{\pgfqpoint{-0.020833in}{0.000000in}}%
\pgfpathcurveto{\pgfqpoint{-0.020833in}{-0.005525in}}{\pgfqpoint{-0.018638in}{-0.010825in}}{\pgfqpoint{-0.014731in}{-0.014731in}}%
\pgfpathcurveto{\pgfqpoint{-0.010825in}{-0.018638in}}{\pgfqpoint{-0.005525in}{-0.020833in}}{\pgfqpoint{0.000000in}{-0.020833in}}%
\pgfpathlineto{\pgfqpoint{0.000000in}{-0.020833in}}%
\pgfpathclose%
\pgfusepath{stroke,fill}%
}%
\begin{pgfscope}%
\pgfsys@transformshift{0.514865in}{1.209567in}%
\pgfsys@useobject{currentmarker}{}%
\end{pgfscope}%
\end{pgfscope}%
\begin{pgfscope}%
\pgfpathrectangle{\pgfqpoint{0.450000in}{0.780000in}}{\pgfqpoint{0.800000in}{0.800000in}}%
\pgfusepath{clip}%
\pgfsetbuttcap%
\pgfsetmiterjoin%
\definecolor{currentfill}{rgb}{0.862745,0.223529,0.466667}%
\pgfsetfillcolor{currentfill}%
\pgfsetlinewidth{0.803000pt}%
\definecolor{currentstroke}{rgb}{0.862745,0.223529,0.466667}%
\pgfsetstrokecolor{currentstroke}%
\pgfsetdash{}{0pt}%
\pgfsys@defobject{currentmarker}{\pgfqpoint{-0.020833in}{-0.020833in}}{\pgfqpoint{0.020833in}{0.020833in}}{%
\pgfpathmoveto{\pgfqpoint{-0.020833in}{-0.020833in}}%
\pgfpathlineto{\pgfqpoint{0.020833in}{-0.020833in}}%
\pgfpathlineto{\pgfqpoint{0.020833in}{0.020833in}}%
\pgfpathlineto{\pgfqpoint{-0.020833in}{0.020833in}}%
\pgfpathlineto{\pgfqpoint{-0.020833in}{-0.020833in}}%
\pgfpathclose%
\pgfusepath{stroke,fill}%
}%
\begin{pgfscope}%
\pgfsys@transformshift{0.514865in}{1.199474in}%
\pgfsys@useobject{currentmarker}{}%
\end{pgfscope}%
\end{pgfscope}%
\begin{pgfscope}%
\pgfpathrectangle{\pgfqpoint{0.450000in}{0.780000in}}{\pgfqpoint{0.800000in}{0.800000in}}%
\pgfusepath{clip}%
\pgfsetbuttcap%
\pgfsetroundjoin%
\definecolor{currentfill}{rgb}{1.000000,1.000000,1.000000}%
\pgfsetfillcolor{currentfill}%
\pgfsetlinewidth{0.803000pt}%
\definecolor{currentstroke}{rgb}{0.321569,0.317647,0.305882}%
\pgfsetstrokecolor{currentstroke}%
\pgfsetdash{}{0pt}%
\pgfsys@defobject{currentmarker}{\pgfqpoint{-0.020833in}{-0.020833in}}{\pgfqpoint{0.020833in}{0.020833in}}{%
\pgfpathmoveto{\pgfqpoint{0.000000in}{-0.020833in}}%
\pgfpathcurveto{\pgfqpoint{0.005525in}{-0.020833in}}{\pgfqpoint{0.010825in}{-0.018638in}}{\pgfqpoint{0.014731in}{-0.014731in}}%
\pgfpathcurveto{\pgfqpoint{0.018638in}{-0.010825in}}{\pgfqpoint{0.020833in}{-0.005525in}}{\pgfqpoint{0.020833in}{0.000000in}}%
\pgfpathcurveto{\pgfqpoint{0.020833in}{0.005525in}}{\pgfqpoint{0.018638in}{0.010825in}}{\pgfqpoint{0.014731in}{0.014731in}}%
\pgfpathcurveto{\pgfqpoint{0.010825in}{0.018638in}}{\pgfqpoint{0.005525in}{0.020833in}}{\pgfqpoint{0.000000in}{0.020833in}}%
\pgfpathcurveto{\pgfqpoint{-0.005525in}{0.020833in}}{\pgfqpoint{-0.010825in}{0.018638in}}{\pgfqpoint{-0.014731in}{0.014731in}}%
\pgfpathcurveto{\pgfqpoint{-0.018638in}{0.010825in}}{\pgfqpoint{-0.020833in}{0.005525in}}{\pgfqpoint{-0.020833in}{0.000000in}}%
\pgfpathcurveto{\pgfqpoint{-0.020833in}{-0.005525in}}{\pgfqpoint{-0.018638in}{-0.010825in}}{\pgfqpoint{-0.014731in}{-0.014731in}}%
\pgfpathcurveto{\pgfqpoint{-0.010825in}{-0.018638in}}{\pgfqpoint{-0.005525in}{-0.020833in}}{\pgfqpoint{0.000000in}{-0.020833in}}%
\pgfpathlineto{\pgfqpoint{0.000000in}{-0.020833in}}%
\pgfpathclose%
\pgfusepath{stroke,fill}%
}%
\begin{pgfscope}%
\pgfsys@transformshift{0.622973in}{1.523973in}%
\pgfsys@useobject{currentmarker}{}%
\end{pgfscope}%
\end{pgfscope}%
\begin{pgfscope}%
\pgfpathrectangle{\pgfqpoint{0.450000in}{0.780000in}}{\pgfqpoint{0.800000in}{0.800000in}}%
\pgfusepath{clip}%
\pgfsetbuttcap%
\pgfsetroundjoin%
\definecolor{currentfill}{rgb}{0.031373,0.564706,0.600000}%
\pgfsetfillcolor{currentfill}%
\pgfsetlinewidth{0.803000pt}%
\definecolor{currentstroke}{rgb}{0.031373,0.564706,0.600000}%
\pgfsetstrokecolor{currentstroke}%
\pgfsetdash{}{0pt}%
\pgfsys@defobject{currentmarker}{\pgfqpoint{-0.020833in}{-0.020833in}}{\pgfqpoint{0.020833in}{0.020833in}}{%
\pgfpathmoveto{\pgfqpoint{0.000000in}{-0.020833in}}%
\pgfpathcurveto{\pgfqpoint{0.005525in}{-0.020833in}}{\pgfqpoint{0.010825in}{-0.018638in}}{\pgfqpoint{0.014731in}{-0.014731in}}%
\pgfpathcurveto{\pgfqpoint{0.018638in}{-0.010825in}}{\pgfqpoint{0.020833in}{-0.005525in}}{\pgfqpoint{0.020833in}{0.000000in}}%
\pgfpathcurveto{\pgfqpoint{0.020833in}{0.005525in}}{\pgfqpoint{0.018638in}{0.010825in}}{\pgfqpoint{0.014731in}{0.014731in}}%
\pgfpathcurveto{\pgfqpoint{0.010825in}{0.018638in}}{\pgfqpoint{0.005525in}{0.020833in}}{\pgfqpoint{0.000000in}{0.020833in}}%
\pgfpathcurveto{\pgfqpoint{-0.005525in}{0.020833in}}{\pgfqpoint{-0.010825in}{0.018638in}}{\pgfqpoint{-0.014731in}{0.014731in}}%
\pgfpathcurveto{\pgfqpoint{-0.018638in}{0.010825in}}{\pgfqpoint{-0.020833in}{0.005525in}}{\pgfqpoint{-0.020833in}{0.000000in}}%
\pgfpathcurveto{\pgfqpoint{-0.020833in}{-0.005525in}}{\pgfqpoint{-0.018638in}{-0.010825in}}{\pgfqpoint{-0.014731in}{-0.014731in}}%
\pgfpathcurveto{\pgfqpoint{-0.010825in}{-0.018638in}}{\pgfqpoint{-0.005525in}{-0.020833in}}{\pgfqpoint{0.000000in}{-0.020833in}}%
\pgfpathlineto{\pgfqpoint{0.000000in}{-0.020833in}}%
\pgfpathclose%
\pgfusepath{stroke,fill}%
}%
\begin{pgfscope}%
\pgfsys@transformshift{0.622973in}{1.447896in}%
\pgfsys@useobject{currentmarker}{}%
\end{pgfscope}%
\end{pgfscope}%
\begin{pgfscope}%
\pgfpathrectangle{\pgfqpoint{0.450000in}{0.780000in}}{\pgfqpoint{0.800000in}{0.800000in}}%
\pgfusepath{clip}%
\pgfsetbuttcap%
\pgfsetmiterjoin%
\definecolor{currentfill}{rgb}{0.862745,0.223529,0.466667}%
\pgfsetfillcolor{currentfill}%
\pgfsetlinewidth{0.803000pt}%
\definecolor{currentstroke}{rgb}{0.862745,0.223529,0.466667}%
\pgfsetstrokecolor{currentstroke}%
\pgfsetdash{}{0pt}%
\pgfsys@defobject{currentmarker}{\pgfqpoint{-0.020833in}{-0.020833in}}{\pgfqpoint{0.020833in}{0.020833in}}{%
\pgfpathmoveto{\pgfqpoint{-0.020833in}{-0.020833in}}%
\pgfpathlineto{\pgfqpoint{0.020833in}{-0.020833in}}%
\pgfpathlineto{\pgfqpoint{0.020833in}{0.020833in}}%
\pgfpathlineto{\pgfqpoint{-0.020833in}{0.020833in}}%
\pgfpathlineto{\pgfqpoint{-0.020833in}{-0.020833in}}%
\pgfpathclose%
\pgfusepath{stroke,fill}%
}%
\begin{pgfscope}%
\pgfsys@transformshift{0.622973in}{1.402251in}%
\pgfsys@useobject{currentmarker}{}%
\end{pgfscope}%
\end{pgfscope}%
\begin{pgfscope}%
\pgfpathrectangle{\pgfqpoint{0.450000in}{0.780000in}}{\pgfqpoint{0.800000in}{0.800000in}}%
\pgfusepath{clip}%
\pgfsetbuttcap%
\pgfsetroundjoin%
\definecolor{currentfill}{rgb}{1.000000,1.000000,1.000000}%
\pgfsetfillcolor{currentfill}%
\pgfsetlinewidth{0.803000pt}%
\definecolor{currentstroke}{rgb}{0.321569,0.317647,0.305882}%
\pgfsetstrokecolor{currentstroke}%
\pgfsetdash{}{0pt}%
\pgfsys@defobject{currentmarker}{\pgfqpoint{-0.020833in}{-0.020833in}}{\pgfqpoint{0.020833in}{0.020833in}}{%
\pgfpathmoveto{\pgfqpoint{0.000000in}{-0.020833in}}%
\pgfpathcurveto{\pgfqpoint{0.005525in}{-0.020833in}}{\pgfqpoint{0.010825in}{-0.018638in}}{\pgfqpoint{0.014731in}{-0.014731in}}%
\pgfpathcurveto{\pgfqpoint{0.018638in}{-0.010825in}}{\pgfqpoint{0.020833in}{-0.005525in}}{\pgfqpoint{0.020833in}{0.000000in}}%
\pgfpathcurveto{\pgfqpoint{0.020833in}{0.005525in}}{\pgfqpoint{0.018638in}{0.010825in}}{\pgfqpoint{0.014731in}{0.014731in}}%
\pgfpathcurveto{\pgfqpoint{0.010825in}{0.018638in}}{\pgfqpoint{0.005525in}{0.020833in}}{\pgfqpoint{0.000000in}{0.020833in}}%
\pgfpathcurveto{\pgfqpoint{-0.005525in}{0.020833in}}{\pgfqpoint{-0.010825in}{0.018638in}}{\pgfqpoint{-0.014731in}{0.014731in}}%
\pgfpathcurveto{\pgfqpoint{-0.018638in}{0.010825in}}{\pgfqpoint{-0.020833in}{0.005525in}}{\pgfqpoint{-0.020833in}{0.000000in}}%
\pgfpathcurveto{\pgfqpoint{-0.020833in}{-0.005525in}}{\pgfqpoint{-0.018638in}{-0.010825in}}{\pgfqpoint{-0.014731in}{-0.014731in}}%
\pgfpathcurveto{\pgfqpoint{-0.010825in}{-0.018638in}}{\pgfqpoint{-0.005525in}{-0.020833in}}{\pgfqpoint{0.000000in}{-0.020833in}}%
\pgfpathlineto{\pgfqpoint{0.000000in}{-0.020833in}}%
\pgfpathclose%
\pgfusepath{stroke,fill}%
}%
\begin{pgfscope}%
\pgfsys@transformshift{0.795946in}{1.033958in}%
\pgfsys@useobject{currentmarker}{}%
\end{pgfscope}%
\end{pgfscope}%
\begin{pgfscope}%
\pgfpathrectangle{\pgfqpoint{0.450000in}{0.780000in}}{\pgfqpoint{0.800000in}{0.800000in}}%
\pgfusepath{clip}%
\pgfsetbuttcap%
\pgfsetroundjoin%
\definecolor{currentfill}{rgb}{0.031373,0.564706,0.600000}%
\pgfsetfillcolor{currentfill}%
\pgfsetlinewidth{0.803000pt}%
\definecolor{currentstroke}{rgb}{0.031373,0.564706,0.600000}%
\pgfsetstrokecolor{currentstroke}%
\pgfsetdash{}{0pt}%
\pgfsys@defobject{currentmarker}{\pgfqpoint{-0.020833in}{-0.020833in}}{\pgfqpoint{0.020833in}{0.020833in}}{%
\pgfpathmoveto{\pgfqpoint{0.000000in}{-0.020833in}}%
\pgfpathcurveto{\pgfqpoint{0.005525in}{-0.020833in}}{\pgfqpoint{0.010825in}{-0.018638in}}{\pgfqpoint{0.014731in}{-0.014731in}}%
\pgfpathcurveto{\pgfqpoint{0.018638in}{-0.010825in}}{\pgfqpoint{0.020833in}{-0.005525in}}{\pgfqpoint{0.020833in}{0.000000in}}%
\pgfpathcurveto{\pgfqpoint{0.020833in}{0.005525in}}{\pgfqpoint{0.018638in}{0.010825in}}{\pgfqpoint{0.014731in}{0.014731in}}%
\pgfpathcurveto{\pgfqpoint{0.010825in}{0.018638in}}{\pgfqpoint{0.005525in}{0.020833in}}{\pgfqpoint{0.000000in}{0.020833in}}%
\pgfpathcurveto{\pgfqpoint{-0.005525in}{0.020833in}}{\pgfqpoint{-0.010825in}{0.018638in}}{\pgfqpoint{-0.014731in}{0.014731in}}%
\pgfpathcurveto{\pgfqpoint{-0.018638in}{0.010825in}}{\pgfqpoint{-0.020833in}{0.005525in}}{\pgfqpoint{-0.020833in}{0.000000in}}%
\pgfpathcurveto{\pgfqpoint{-0.020833in}{-0.005525in}}{\pgfqpoint{-0.018638in}{-0.010825in}}{\pgfqpoint{-0.014731in}{-0.014731in}}%
\pgfpathcurveto{\pgfqpoint{-0.010825in}{-0.018638in}}{\pgfqpoint{-0.005525in}{-0.020833in}}{\pgfqpoint{0.000000in}{-0.020833in}}%
\pgfpathlineto{\pgfqpoint{0.000000in}{-0.020833in}}%
\pgfpathclose%
\pgfusepath{stroke,fill}%
}%
\begin{pgfscope}%
\pgfsys@transformshift{0.795946in}{0.857240in}%
\pgfsys@useobject{currentmarker}{}%
\end{pgfscope}%
\end{pgfscope}%
\begin{pgfscope}%
\pgfpathrectangle{\pgfqpoint{0.450000in}{0.780000in}}{\pgfqpoint{0.800000in}{0.800000in}}%
\pgfusepath{clip}%
\pgfsetbuttcap%
\pgfsetmiterjoin%
\definecolor{currentfill}{rgb}{0.862745,0.223529,0.466667}%
\pgfsetfillcolor{currentfill}%
\pgfsetlinewidth{0.803000pt}%
\definecolor{currentstroke}{rgb}{0.862745,0.223529,0.466667}%
\pgfsetstrokecolor{currentstroke}%
\pgfsetdash{}{0pt}%
\pgfsys@defobject{currentmarker}{\pgfqpoint{-0.020833in}{-0.020833in}}{\pgfqpoint{0.020833in}{0.020833in}}{%
\pgfpathmoveto{\pgfqpoint{-0.020833in}{-0.020833in}}%
\pgfpathlineto{\pgfqpoint{0.020833in}{-0.020833in}}%
\pgfpathlineto{\pgfqpoint{0.020833in}{0.020833in}}%
\pgfpathlineto{\pgfqpoint{-0.020833in}{0.020833in}}%
\pgfpathlineto{\pgfqpoint{-0.020833in}{-0.020833in}}%
\pgfpathclose%
\pgfusepath{stroke,fill}%
}%
\begin{pgfscope}%
\pgfsys@transformshift{0.795946in}{0.836027in}%
\pgfsys@useobject{currentmarker}{}%
\end{pgfscope}%
\end{pgfscope}%
\begin{pgfscope}%
\pgfpathrectangle{\pgfqpoint{0.450000in}{0.780000in}}{\pgfqpoint{0.800000in}{0.800000in}}%
\pgfusepath{clip}%
\pgfsetbuttcap%
\pgfsetroundjoin%
\definecolor{currentfill}{rgb}{1.000000,1.000000,1.000000}%
\pgfsetfillcolor{currentfill}%
\pgfsetlinewidth{0.803000pt}%
\definecolor{currentstroke}{rgb}{0.321569,0.317647,0.305882}%
\pgfsetstrokecolor{currentstroke}%
\pgfsetdash{}{0pt}%
\pgfsys@defobject{currentmarker}{\pgfqpoint{-0.020833in}{-0.020833in}}{\pgfqpoint{0.020833in}{0.020833in}}{%
\pgfpathmoveto{\pgfqpoint{0.000000in}{-0.020833in}}%
\pgfpathcurveto{\pgfqpoint{0.005525in}{-0.020833in}}{\pgfqpoint{0.010825in}{-0.018638in}}{\pgfqpoint{0.014731in}{-0.014731in}}%
\pgfpathcurveto{\pgfqpoint{0.018638in}{-0.010825in}}{\pgfqpoint{0.020833in}{-0.005525in}}{\pgfqpoint{0.020833in}{0.000000in}}%
\pgfpathcurveto{\pgfqpoint{0.020833in}{0.005525in}}{\pgfqpoint{0.018638in}{0.010825in}}{\pgfqpoint{0.014731in}{0.014731in}}%
\pgfpathcurveto{\pgfqpoint{0.010825in}{0.018638in}}{\pgfqpoint{0.005525in}{0.020833in}}{\pgfqpoint{0.000000in}{0.020833in}}%
\pgfpathcurveto{\pgfqpoint{-0.005525in}{0.020833in}}{\pgfqpoint{-0.010825in}{0.018638in}}{\pgfqpoint{-0.014731in}{0.014731in}}%
\pgfpathcurveto{\pgfqpoint{-0.018638in}{0.010825in}}{\pgfqpoint{-0.020833in}{0.005525in}}{\pgfqpoint{-0.020833in}{0.000000in}}%
\pgfpathcurveto{\pgfqpoint{-0.020833in}{-0.005525in}}{\pgfqpoint{-0.018638in}{-0.010825in}}{\pgfqpoint{-0.014731in}{-0.014731in}}%
\pgfpathcurveto{\pgfqpoint{-0.010825in}{-0.018638in}}{\pgfqpoint{-0.005525in}{-0.020833in}}{\pgfqpoint{0.000000in}{-0.020833in}}%
\pgfpathlineto{\pgfqpoint{0.000000in}{-0.020833in}}%
\pgfpathclose%
\pgfusepath{stroke,fill}%
}%
\begin{pgfscope}%
\pgfsys@transformshift{0.904054in}{1.184202in}%
\pgfsys@useobject{currentmarker}{}%
\end{pgfscope}%
\end{pgfscope}%
\begin{pgfscope}%
\pgfpathrectangle{\pgfqpoint{0.450000in}{0.780000in}}{\pgfqpoint{0.800000in}{0.800000in}}%
\pgfusepath{clip}%
\pgfsetbuttcap%
\pgfsetroundjoin%
\definecolor{currentfill}{rgb}{0.031373,0.564706,0.600000}%
\pgfsetfillcolor{currentfill}%
\pgfsetlinewidth{0.803000pt}%
\definecolor{currentstroke}{rgb}{0.031373,0.564706,0.600000}%
\pgfsetstrokecolor{currentstroke}%
\pgfsetdash{}{0pt}%
\pgfsys@defobject{currentmarker}{\pgfqpoint{-0.020833in}{-0.020833in}}{\pgfqpoint{0.020833in}{0.020833in}}{%
\pgfpathmoveto{\pgfqpoint{0.000000in}{-0.020833in}}%
\pgfpathcurveto{\pgfqpoint{0.005525in}{-0.020833in}}{\pgfqpoint{0.010825in}{-0.018638in}}{\pgfqpoint{0.014731in}{-0.014731in}}%
\pgfpathcurveto{\pgfqpoint{0.018638in}{-0.010825in}}{\pgfqpoint{0.020833in}{-0.005525in}}{\pgfqpoint{0.020833in}{0.000000in}}%
\pgfpathcurveto{\pgfqpoint{0.020833in}{0.005525in}}{\pgfqpoint{0.018638in}{0.010825in}}{\pgfqpoint{0.014731in}{0.014731in}}%
\pgfpathcurveto{\pgfqpoint{0.010825in}{0.018638in}}{\pgfqpoint{0.005525in}{0.020833in}}{\pgfqpoint{0.000000in}{0.020833in}}%
\pgfpathcurveto{\pgfqpoint{-0.005525in}{0.020833in}}{\pgfqpoint{-0.010825in}{0.018638in}}{\pgfqpoint{-0.014731in}{0.014731in}}%
\pgfpathcurveto{\pgfqpoint{-0.018638in}{0.010825in}}{\pgfqpoint{-0.020833in}{0.005525in}}{\pgfqpoint{-0.020833in}{0.000000in}}%
\pgfpathcurveto{\pgfqpoint{-0.020833in}{-0.005525in}}{\pgfqpoint{-0.018638in}{-0.010825in}}{\pgfqpoint{-0.014731in}{-0.014731in}}%
\pgfpathcurveto{\pgfqpoint{-0.010825in}{-0.018638in}}{\pgfqpoint{-0.005525in}{-0.020833in}}{\pgfqpoint{0.000000in}{-0.020833in}}%
\pgfpathlineto{\pgfqpoint{0.000000in}{-0.020833in}}%
\pgfpathclose%
\pgfusepath{stroke,fill}%
}%
\begin{pgfscope}%
\pgfsys@transformshift{0.904054in}{1.117236in}%
\pgfsys@useobject{currentmarker}{}%
\end{pgfscope}%
\end{pgfscope}%
\begin{pgfscope}%
\pgfpathrectangle{\pgfqpoint{0.450000in}{0.780000in}}{\pgfqpoint{0.800000in}{0.800000in}}%
\pgfusepath{clip}%
\pgfsetbuttcap%
\pgfsetmiterjoin%
\definecolor{currentfill}{rgb}{0.862745,0.223529,0.466667}%
\pgfsetfillcolor{currentfill}%
\pgfsetlinewidth{0.803000pt}%
\definecolor{currentstroke}{rgb}{0.862745,0.223529,0.466667}%
\pgfsetstrokecolor{currentstroke}%
\pgfsetdash{}{0pt}%
\pgfsys@defobject{currentmarker}{\pgfqpoint{-0.020833in}{-0.020833in}}{\pgfqpoint{0.020833in}{0.020833in}}{%
\pgfpathmoveto{\pgfqpoint{-0.020833in}{-0.020833in}}%
\pgfpathlineto{\pgfqpoint{0.020833in}{-0.020833in}}%
\pgfpathlineto{\pgfqpoint{0.020833in}{0.020833in}}%
\pgfpathlineto{\pgfqpoint{-0.020833in}{0.020833in}}%
\pgfpathlineto{\pgfqpoint{-0.020833in}{-0.020833in}}%
\pgfpathclose%
\pgfusepath{stroke,fill}%
}%
\begin{pgfscope}%
\pgfsys@transformshift{0.904054in}{1.111643in}%
\pgfsys@useobject{currentmarker}{}%
\end{pgfscope}%
\end{pgfscope}%
\begin{pgfscope}%
\pgfpathrectangle{\pgfqpoint{0.450000in}{0.780000in}}{\pgfqpoint{0.800000in}{0.800000in}}%
\pgfusepath{clip}%
\pgfsetbuttcap%
\pgfsetroundjoin%
\definecolor{currentfill}{rgb}{1.000000,1.000000,1.000000}%
\pgfsetfillcolor{currentfill}%
\pgfsetlinewidth{0.803000pt}%
\definecolor{currentstroke}{rgb}{0.321569,0.317647,0.305882}%
\pgfsetstrokecolor{currentstroke}%
\pgfsetdash{}{0pt}%
\pgfsys@defobject{currentmarker}{\pgfqpoint{-0.020833in}{-0.020833in}}{\pgfqpoint{0.020833in}{0.020833in}}{%
\pgfpathmoveto{\pgfqpoint{0.000000in}{-0.020833in}}%
\pgfpathcurveto{\pgfqpoint{0.005525in}{-0.020833in}}{\pgfqpoint{0.010825in}{-0.018638in}}{\pgfqpoint{0.014731in}{-0.014731in}}%
\pgfpathcurveto{\pgfqpoint{0.018638in}{-0.010825in}}{\pgfqpoint{0.020833in}{-0.005525in}}{\pgfqpoint{0.020833in}{0.000000in}}%
\pgfpathcurveto{\pgfqpoint{0.020833in}{0.005525in}}{\pgfqpoint{0.018638in}{0.010825in}}{\pgfqpoint{0.014731in}{0.014731in}}%
\pgfpathcurveto{\pgfqpoint{0.010825in}{0.018638in}}{\pgfqpoint{0.005525in}{0.020833in}}{\pgfqpoint{0.000000in}{0.020833in}}%
\pgfpathcurveto{\pgfqpoint{-0.005525in}{0.020833in}}{\pgfqpoint{-0.010825in}{0.018638in}}{\pgfqpoint{-0.014731in}{0.014731in}}%
\pgfpathcurveto{\pgfqpoint{-0.018638in}{0.010825in}}{\pgfqpoint{-0.020833in}{0.005525in}}{\pgfqpoint{-0.020833in}{0.000000in}}%
\pgfpathcurveto{\pgfqpoint{-0.020833in}{-0.005525in}}{\pgfqpoint{-0.018638in}{-0.010825in}}{\pgfqpoint{-0.014731in}{-0.014731in}}%
\pgfpathcurveto{\pgfqpoint{-0.010825in}{-0.018638in}}{\pgfqpoint{-0.005525in}{-0.020833in}}{\pgfqpoint{0.000000in}{-0.020833in}}%
\pgfpathlineto{\pgfqpoint{0.000000in}{-0.020833in}}%
\pgfpathclose%
\pgfusepath{stroke,fill}%
}%
\begin{pgfscope}%
\pgfsys@transformshift{1.077027in}{1.022388in}%
\pgfsys@useobject{currentmarker}{}%
\end{pgfscope}%
\end{pgfscope}%
\begin{pgfscope}%
\pgfpathrectangle{\pgfqpoint{0.450000in}{0.780000in}}{\pgfqpoint{0.800000in}{0.800000in}}%
\pgfusepath{clip}%
\pgfsetbuttcap%
\pgfsetroundjoin%
\definecolor{currentfill}{rgb}{0.031373,0.564706,0.600000}%
\pgfsetfillcolor{currentfill}%
\pgfsetlinewidth{0.803000pt}%
\definecolor{currentstroke}{rgb}{0.031373,0.564706,0.600000}%
\pgfsetstrokecolor{currentstroke}%
\pgfsetdash{}{0pt}%
\pgfsys@defobject{currentmarker}{\pgfqpoint{-0.020833in}{-0.020833in}}{\pgfqpoint{0.020833in}{0.020833in}}{%
\pgfpathmoveto{\pgfqpoint{0.000000in}{-0.020833in}}%
\pgfpathcurveto{\pgfqpoint{0.005525in}{-0.020833in}}{\pgfqpoint{0.010825in}{-0.018638in}}{\pgfqpoint{0.014731in}{-0.014731in}}%
\pgfpathcurveto{\pgfqpoint{0.018638in}{-0.010825in}}{\pgfqpoint{0.020833in}{-0.005525in}}{\pgfqpoint{0.020833in}{0.000000in}}%
\pgfpathcurveto{\pgfqpoint{0.020833in}{0.005525in}}{\pgfqpoint{0.018638in}{0.010825in}}{\pgfqpoint{0.014731in}{0.014731in}}%
\pgfpathcurveto{\pgfqpoint{0.010825in}{0.018638in}}{\pgfqpoint{0.005525in}{0.020833in}}{\pgfqpoint{0.000000in}{0.020833in}}%
\pgfpathcurveto{\pgfqpoint{-0.005525in}{0.020833in}}{\pgfqpoint{-0.010825in}{0.018638in}}{\pgfqpoint{-0.014731in}{0.014731in}}%
\pgfpathcurveto{\pgfqpoint{-0.018638in}{0.010825in}}{\pgfqpoint{-0.020833in}{0.005525in}}{\pgfqpoint{-0.020833in}{0.000000in}}%
\pgfpathcurveto{\pgfqpoint{-0.020833in}{-0.005525in}}{\pgfqpoint{-0.018638in}{-0.010825in}}{\pgfqpoint{-0.014731in}{-0.014731in}}%
\pgfpathcurveto{\pgfqpoint{-0.010825in}{-0.018638in}}{\pgfqpoint{-0.005525in}{-0.020833in}}{\pgfqpoint{0.000000in}{-0.020833in}}%
\pgfpathlineto{\pgfqpoint{0.000000in}{-0.020833in}}%
\pgfpathclose%
\pgfusepath{stroke,fill}%
}%
\begin{pgfscope}%
\pgfsys@transformshift{1.077027in}{0.875787in}%
\pgfsys@useobject{currentmarker}{}%
\end{pgfscope}%
\end{pgfscope}%
\begin{pgfscope}%
\pgfpathrectangle{\pgfqpoint{0.450000in}{0.780000in}}{\pgfqpoint{0.800000in}{0.800000in}}%
\pgfusepath{clip}%
\pgfsetbuttcap%
\pgfsetmiterjoin%
\definecolor{currentfill}{rgb}{0.862745,0.223529,0.466667}%
\pgfsetfillcolor{currentfill}%
\pgfsetlinewidth{0.803000pt}%
\definecolor{currentstroke}{rgb}{0.862745,0.223529,0.466667}%
\pgfsetstrokecolor{currentstroke}%
\pgfsetdash{}{0pt}%
\pgfsys@defobject{currentmarker}{\pgfqpoint{-0.020833in}{-0.020833in}}{\pgfqpoint{0.020833in}{0.020833in}}{%
\pgfpathmoveto{\pgfqpoint{-0.020833in}{-0.020833in}}%
\pgfpathlineto{\pgfqpoint{0.020833in}{-0.020833in}}%
\pgfpathlineto{\pgfqpoint{0.020833in}{0.020833in}}%
\pgfpathlineto{\pgfqpoint{-0.020833in}{0.020833in}}%
\pgfpathlineto{\pgfqpoint{-0.020833in}{-0.020833in}}%
\pgfpathclose%
\pgfusepath{stroke,fill}%
}%
\begin{pgfscope}%
\pgfsys@transformshift{1.077027in}{0.868537in}%
\pgfsys@useobject{currentmarker}{}%
\end{pgfscope}%
\end{pgfscope}%
\begin{pgfscope}%
\pgfpathrectangle{\pgfqpoint{0.450000in}{0.780000in}}{\pgfqpoint{0.800000in}{0.800000in}}%
\pgfusepath{clip}%
\pgfsetbuttcap%
\pgfsetroundjoin%
\definecolor{currentfill}{rgb}{1.000000,1.000000,1.000000}%
\pgfsetfillcolor{currentfill}%
\pgfsetlinewidth{0.803000pt}%
\definecolor{currentstroke}{rgb}{0.321569,0.317647,0.305882}%
\pgfsetstrokecolor{currentstroke}%
\pgfsetdash{}{0pt}%
\pgfsys@defobject{currentmarker}{\pgfqpoint{-0.020833in}{-0.020833in}}{\pgfqpoint{0.020833in}{0.020833in}}{%
\pgfpathmoveto{\pgfqpoint{0.000000in}{-0.020833in}}%
\pgfpathcurveto{\pgfqpoint{0.005525in}{-0.020833in}}{\pgfqpoint{0.010825in}{-0.018638in}}{\pgfqpoint{0.014731in}{-0.014731in}}%
\pgfpathcurveto{\pgfqpoint{0.018638in}{-0.010825in}}{\pgfqpoint{0.020833in}{-0.005525in}}{\pgfqpoint{0.020833in}{0.000000in}}%
\pgfpathcurveto{\pgfqpoint{0.020833in}{0.005525in}}{\pgfqpoint{0.018638in}{0.010825in}}{\pgfqpoint{0.014731in}{0.014731in}}%
\pgfpathcurveto{\pgfqpoint{0.010825in}{0.018638in}}{\pgfqpoint{0.005525in}{0.020833in}}{\pgfqpoint{0.000000in}{0.020833in}}%
\pgfpathcurveto{\pgfqpoint{-0.005525in}{0.020833in}}{\pgfqpoint{-0.010825in}{0.018638in}}{\pgfqpoint{-0.014731in}{0.014731in}}%
\pgfpathcurveto{\pgfqpoint{-0.018638in}{0.010825in}}{\pgfqpoint{-0.020833in}{0.005525in}}{\pgfqpoint{-0.020833in}{0.000000in}}%
\pgfpathcurveto{\pgfqpoint{-0.020833in}{-0.005525in}}{\pgfqpoint{-0.018638in}{-0.010825in}}{\pgfqpoint{-0.014731in}{-0.014731in}}%
\pgfpathcurveto{\pgfqpoint{-0.010825in}{-0.018638in}}{\pgfqpoint{-0.005525in}{-0.020833in}}{\pgfqpoint{0.000000in}{-0.020833in}}%
\pgfpathlineto{\pgfqpoint{0.000000in}{-0.020833in}}%
\pgfpathclose%
\pgfusepath{stroke,fill}%
}%
\begin{pgfscope}%
\pgfsys@transformshift{1.185135in}{1.256248in}%
\pgfsys@useobject{currentmarker}{}%
\end{pgfscope}%
\end{pgfscope}%
\begin{pgfscope}%
\pgfpathrectangle{\pgfqpoint{0.450000in}{0.780000in}}{\pgfqpoint{0.800000in}{0.800000in}}%
\pgfusepath{clip}%
\pgfsetbuttcap%
\pgfsetroundjoin%
\definecolor{currentfill}{rgb}{0.031373,0.564706,0.600000}%
\pgfsetfillcolor{currentfill}%
\pgfsetlinewidth{0.803000pt}%
\definecolor{currentstroke}{rgb}{0.031373,0.564706,0.600000}%
\pgfsetstrokecolor{currentstroke}%
\pgfsetdash{}{0pt}%
\pgfsys@defobject{currentmarker}{\pgfqpoint{-0.020833in}{-0.020833in}}{\pgfqpoint{0.020833in}{0.020833in}}{%
\pgfpathmoveto{\pgfqpoint{0.000000in}{-0.020833in}}%
\pgfpathcurveto{\pgfqpoint{0.005525in}{-0.020833in}}{\pgfqpoint{0.010825in}{-0.018638in}}{\pgfqpoint{0.014731in}{-0.014731in}}%
\pgfpathcurveto{\pgfqpoint{0.018638in}{-0.010825in}}{\pgfqpoint{0.020833in}{-0.005525in}}{\pgfqpoint{0.020833in}{0.000000in}}%
\pgfpathcurveto{\pgfqpoint{0.020833in}{0.005525in}}{\pgfqpoint{0.018638in}{0.010825in}}{\pgfqpoint{0.014731in}{0.014731in}}%
\pgfpathcurveto{\pgfqpoint{0.010825in}{0.018638in}}{\pgfqpoint{0.005525in}{0.020833in}}{\pgfqpoint{0.000000in}{0.020833in}}%
\pgfpathcurveto{\pgfqpoint{-0.005525in}{0.020833in}}{\pgfqpoint{-0.010825in}{0.018638in}}{\pgfqpoint{-0.014731in}{0.014731in}}%
\pgfpathcurveto{\pgfqpoint{-0.018638in}{0.010825in}}{\pgfqpoint{-0.020833in}{0.005525in}}{\pgfqpoint{-0.020833in}{0.000000in}}%
\pgfpathcurveto{\pgfqpoint{-0.020833in}{-0.005525in}}{\pgfqpoint{-0.018638in}{-0.010825in}}{\pgfqpoint{-0.014731in}{-0.014731in}}%
\pgfpathcurveto{\pgfqpoint{-0.010825in}{-0.018638in}}{\pgfqpoint{-0.005525in}{-0.020833in}}{\pgfqpoint{0.000000in}{-0.020833in}}%
\pgfpathlineto{\pgfqpoint{0.000000in}{-0.020833in}}%
\pgfpathclose%
\pgfusepath{stroke,fill}%
}%
\begin{pgfscope}%
\pgfsys@transformshift{1.185135in}{1.219839in}%
\pgfsys@useobject{currentmarker}{}%
\end{pgfscope}%
\end{pgfscope}%
\begin{pgfscope}%
\pgfpathrectangle{\pgfqpoint{0.450000in}{0.780000in}}{\pgfqpoint{0.800000in}{0.800000in}}%
\pgfusepath{clip}%
\pgfsetbuttcap%
\pgfsetmiterjoin%
\definecolor{currentfill}{rgb}{0.862745,0.223529,0.466667}%
\pgfsetfillcolor{currentfill}%
\pgfsetlinewidth{0.803000pt}%
\definecolor{currentstroke}{rgb}{0.862745,0.223529,0.466667}%
\pgfsetstrokecolor{currentstroke}%
\pgfsetdash{}{0pt}%
\pgfsys@defobject{currentmarker}{\pgfqpoint{-0.020833in}{-0.020833in}}{\pgfqpoint{0.020833in}{0.020833in}}{%
\pgfpathmoveto{\pgfqpoint{-0.020833in}{-0.020833in}}%
\pgfpathlineto{\pgfqpoint{0.020833in}{-0.020833in}}%
\pgfpathlineto{\pgfqpoint{0.020833in}{0.020833in}}%
\pgfpathlineto{\pgfqpoint{-0.020833in}{0.020833in}}%
\pgfpathlineto{\pgfqpoint{-0.020833in}{-0.020833in}}%
\pgfpathclose%
\pgfusepath{stroke,fill}%
}%
\begin{pgfscope}%
\pgfsys@transformshift{1.185135in}{1.216728in}%
\pgfsys@useobject{currentmarker}{}%
\end{pgfscope}%
\end{pgfscope}%
\begin{pgfscope}%
\pgfsetbuttcap%
\pgfsetroundjoin%
\definecolor{currentfill}{rgb}{0.031373,0.564706,0.600000}%
\pgfsetfillcolor{currentfill}%
\pgfsetlinewidth{1.003750pt}%
\definecolor{currentstroke}{rgb}{0.031373,0.564706,0.600000}%
\pgfsetstrokecolor{currentstroke}%
\pgfsetdash{}{0pt}%
\pgfsys@defobject{currentmarker}{\pgfqpoint{-0.020833in}{-0.020833in}}{\pgfqpoint{0.020833in}{0.020833in}}{%
\pgfpathmoveto{\pgfqpoint{0.000000in}{-0.020833in}}%
\pgfpathcurveto{\pgfqpoint{0.005525in}{-0.020833in}}{\pgfqpoint{0.010825in}{-0.018638in}}{\pgfqpoint{0.014731in}{-0.014731in}}%
\pgfpathcurveto{\pgfqpoint{0.018638in}{-0.010825in}}{\pgfqpoint{0.020833in}{-0.005525in}}{\pgfqpoint{0.020833in}{0.000000in}}%
\pgfpathcurveto{\pgfqpoint{0.020833in}{0.005525in}}{\pgfqpoint{0.018638in}{0.010825in}}{\pgfqpoint{0.014731in}{0.014731in}}%
\pgfpathcurveto{\pgfqpoint{0.010825in}{0.018638in}}{\pgfqpoint{0.005525in}{0.020833in}}{\pgfqpoint{0.000000in}{0.020833in}}%
\pgfpathcurveto{\pgfqpoint{-0.005525in}{0.020833in}}{\pgfqpoint{-0.010825in}{0.018638in}}{\pgfqpoint{-0.014731in}{0.014731in}}%
\pgfpathcurveto{\pgfqpoint{-0.018638in}{0.010825in}}{\pgfqpoint{-0.020833in}{0.005525in}}{\pgfqpoint{-0.020833in}{0.000000in}}%
\pgfpathcurveto{\pgfqpoint{-0.020833in}{-0.005525in}}{\pgfqpoint{-0.018638in}{-0.010825in}}{\pgfqpoint{-0.014731in}{-0.014731in}}%
\pgfpathcurveto{\pgfqpoint{-0.010825in}{-0.018638in}}{\pgfqpoint{-0.005525in}{-0.020833in}}{\pgfqpoint{0.000000in}{-0.020833in}}%
\pgfpathlineto{\pgfqpoint{0.000000in}{-0.020833in}}%
\pgfpathclose%
\pgfusepath{stroke,fill}%
}%
\begin{pgfscope}%
\pgfsys@transformshift{0.268371in}{0.246667in}%
\pgfsys@useobject{currentmarker}{}%
\end{pgfscope}%
\end{pgfscope}%
\begin{pgfscope}%
\definecolor{textcolor}{rgb}{0.101961,0.101961,0.098039}%
\pgfsetstrokecolor{textcolor}%
\pgfsetfillcolor{textcolor}%
\pgftext[x=0.362815in,y=0.207778in,left,base]{\color{textcolor}\rmfamily\fontsize{8.000000}{9.600000}\selectfont \(\displaystyle +\)\(\displaystyle \RotTerm{\mathbf{R}_{\mathrm{V}}}\)}%
\end{pgfscope}%
\begin{pgfscope}%
\pgfsetbuttcap%
\pgfsetmiterjoin%
\definecolor{currentfill}{rgb}{0.862745,0.223529,0.466667}%
\pgfsetfillcolor{currentfill}%
\pgfsetlinewidth{1.003750pt}%
\definecolor{currentstroke}{rgb}{0.862745,0.223529,0.466667}%
\pgfsetstrokecolor{currentstroke}%
\pgfsetdash{}{0pt}%
\pgfsys@defobject{currentmarker}{\pgfqpoint{-0.020833in}{-0.020833in}}{\pgfqpoint{0.020833in}{0.020833in}}{%
\pgfpathmoveto{\pgfqpoint{-0.020833in}{-0.020833in}}%
\pgfpathlineto{\pgfqpoint{0.020833in}{-0.020833in}}%
\pgfpathlineto{\pgfqpoint{0.020833in}{0.020833in}}%
\pgfpathlineto{\pgfqpoint{-0.020833in}{0.020833in}}%
\pgfpathlineto{\pgfqpoint{-0.020833in}{-0.020833in}}%
\pgfpathclose%
\pgfusepath{stroke,fill}%
}%
\begin{pgfscope}%
\pgfsys@transformshift{0.789321in}{0.246667in}%
\pgfsys@useobject{currentmarker}{}%
\end{pgfscope}%
\end{pgfscope}%
\begin{pgfscope}%
\definecolor{textcolor}{rgb}{0.101961,0.101961,0.098039}%
\pgfsetstrokecolor{textcolor}%
\pgfsetfillcolor{textcolor}%
\pgftext[x=0.883765in,y=0.207778in,left,base]{\color{textcolor}\rmfamily\fontsize{8.000000}{9.600000}\selectfont \(\displaystyle +\)\(\displaystyle \RescaleTerm{\mathbf{D}_{\mathrm{V}}}\)}%
\end{pgfscope}%
\begin{pgfscope}%
\pgfsetbuttcap%
\pgfsetroundjoin%
\definecolor{currentfill}{rgb}{1.000000,1.000000,1.000000}%
\pgfsetfillcolor{currentfill}%
\pgfsetlinewidth{0.803000pt}%
\definecolor{currentstroke}{rgb}{0.321569,0.317647,0.305882}%
\pgfsetstrokecolor{currentstroke}%
\pgfsetdash{}{0pt}%
\pgfsys@defobject{currentmarker}{\pgfqpoint{-0.020833in}{-0.020833in}}{\pgfqpoint{0.020833in}{0.020833in}}{%
\pgfpathmoveto{\pgfqpoint{0.000000in}{-0.020833in}}%
\pgfpathcurveto{\pgfqpoint{0.005525in}{-0.020833in}}{\pgfqpoint{0.010825in}{-0.018638in}}{\pgfqpoint{0.014731in}{-0.014731in}}%
\pgfpathcurveto{\pgfqpoint{0.018638in}{-0.010825in}}{\pgfqpoint{0.020833in}{-0.005525in}}{\pgfqpoint{0.020833in}{0.000000in}}%
\pgfpathcurveto{\pgfqpoint{0.020833in}{0.005525in}}{\pgfqpoint{0.018638in}{0.010825in}}{\pgfqpoint{0.014731in}{0.014731in}}%
\pgfpathcurveto{\pgfqpoint{0.010825in}{0.018638in}}{\pgfqpoint{0.005525in}{0.020833in}}{\pgfqpoint{0.000000in}{0.020833in}}%
\pgfpathcurveto{\pgfqpoint{-0.005525in}{0.020833in}}{\pgfqpoint{-0.010825in}{0.018638in}}{\pgfqpoint{-0.014731in}{0.014731in}}%
\pgfpathcurveto{\pgfqpoint{-0.018638in}{0.010825in}}{\pgfqpoint{-0.020833in}{0.005525in}}{\pgfqpoint{-0.020833in}{0.000000in}}%
\pgfpathcurveto{\pgfqpoint{-0.020833in}{-0.005525in}}{\pgfqpoint{-0.018638in}{-0.010825in}}{\pgfqpoint{-0.014731in}{-0.014731in}}%
\pgfpathcurveto{\pgfqpoint{-0.010825in}{-0.018638in}}{\pgfqpoint{-0.005525in}{-0.020833in}}{\pgfqpoint{0.000000in}{-0.020833in}}%
\pgfpathlineto{\pgfqpoint{0.000000in}{-0.020833in}}%
\pgfpathclose%
\pgfusepath{stroke,fill}%
}%
\begin{pgfscope}%
\pgfsys@transformshift{0.550223in}{0.096667in}%
\pgfsys@useobject{currentmarker}{}%
\end{pgfscope}%
\end{pgfscope}%
\begin{pgfscope}%
\definecolor{textcolor}{rgb}{0.101961,0.101961,0.098039}%
\pgfsetstrokecolor{textcolor}%
\pgfsetfillcolor{textcolor}%
\pgftext[x=0.644667in,y=0.057778in,left,base]{\color{textcolor}\rmfamily\fontsize{8.000000}{9.600000}\selectfont naive}%
\end{pgfscope}%
\end{pgfpicture}%
\makeatother%
\endgroup%

%% file: figures/pgf/fig_geometry_value.pgf
\providecommand{\mathdefault}[1]{#1}%
\begingroup%
\makeatletter%
\begin{pgfpicture}%
\pgfpathrectangle{\pgfpointorigin}{\pgfqpoint{1.370000in}{1.640000in}}%
\pgfusepath{use as bounding box, clip}%
\begin{pgfscope}%
\pgfsetbuttcap%
\pgfsetmiterjoin%
\definecolor{currentfill}{rgb}{1.000000,1.000000,1.000000}%
\pgfsetfillcolor{currentfill}%
\pgfsetlinewidth{0.000000pt}%
\definecolor{currentstroke}{rgb}{1.000000,1.000000,1.000000}%
\pgfsetstrokecolor{currentstroke}%
\pgfsetdash{}{0pt}%
\pgfpathmoveto{\pgfqpoint{0.000000in}{0.000000in}}%
\pgfpathlineto{\pgfqpoint{1.370000in}{0.000000in}}%
\pgfpathlineto{\pgfqpoint{1.370000in}{1.640000in}}%
\pgfpathlineto{\pgfqpoint{0.000000in}{1.640000in}}%
\pgfpathlineto{\pgfqpoint{0.000000in}{0.000000in}}%
\pgfpathclose%
\pgfusepath{fill}%
\end{pgfscope}%
\begin{pgfscope}%
\pgfsetbuttcap%
\pgfsetmiterjoin%
\definecolor{currentfill}{rgb}{1.000000,1.000000,1.000000}%
\pgfsetfillcolor{currentfill}%
\pgfsetlinewidth{0.000000pt}%
\definecolor{currentstroke}{rgb}{0.000000,0.000000,0.000000}%
\pgfsetstrokecolor{currentstroke}%
\pgfsetstrokeopacity{0.000000}%
\pgfsetdash{}{0pt}%
\pgfpathmoveto{\pgfqpoint{0.450000in}{0.780000in}}%
\pgfpathlineto{\pgfqpoint{1.250000in}{0.780000in}}%
\pgfpathlineto{\pgfqpoint{1.250000in}{1.580000in}}%
\pgfpathlineto{\pgfqpoint{0.450000in}{1.580000in}}%
\pgfpathlineto{\pgfqpoint{0.450000in}{0.780000in}}%
\pgfpathclose%
\pgfusepath{fill}%
\end{pgfscope}%
\begin{pgfscope}%
\definecolor{textcolor}{rgb}{0.101961,0.101961,0.098039}%
\pgfsetstrokecolor{textcolor}%
\pgfsetfillcolor{textcolor}%
\pgftext[x=0.564286in,y=0.731389in,right,top,rotate=45.000000]{\color{textcolor}\rmfamily\fontsize{8.000000}{9.600000}\selectfont LA}%
\end{pgfscope}%
\begin{pgfscope}%
\definecolor{textcolor}{rgb}{0.101961,0.101961,0.098039}%
\pgfsetstrokecolor{textcolor}%
\pgfsetfillcolor{textcolor}%
\pgftext[x=0.754762in,y=0.731389in,right,top,rotate=45.000000]{\color{textcolor}\rmfamily\fontsize{8.000000}{9.600000}\selectfont LA+\(\displaystyle \beta\)}%
\end{pgfscope}%
\begin{pgfscope}%
\definecolor{textcolor}{rgb}{0.101961,0.101961,0.098039}%
\pgfsetstrokecolor{textcolor}%
\pgfsetfillcolor{textcolor}%
\pgftext[x=0.945238in,y=0.731389in,right,top,rotate=45.000000]{\color{textcolor}\rmfamily\fontsize{8.000000}{9.600000}\selectfont M2}%
\end{pgfscope}%
\begin{pgfscope}%
\definecolor{textcolor}{rgb}{0.101961,0.101961,0.098039}%
\pgfsetstrokecolor{textcolor}%
\pgfsetfillcolor{textcolor}%
\pgftext[x=1.135714in,y=0.731389in,right,top,rotate=45.000000]{\color{textcolor}\rmfamily\fontsize{8.000000}{9.600000}\selectfont M2+\(\displaystyle \beta\)}%
\end{pgfscope}%
\begin{pgfscope}%
\pgfpathrectangle{\pgfqpoint{0.450000in}{0.780000in}}{\pgfqpoint{0.800000in}{0.800000in}}%
\pgfusepath{clip}%
\pgfsetrectcap%
\pgfsetroundjoin%
\pgfsetlinewidth{0.501875pt}%
\definecolor{currentstroke}{rgb}{0.886275,0.882353,0.866667}%
\pgfsetstrokecolor{currentstroke}%
\pgfsetdash{}{0pt}%
\pgfpathmoveto{\pgfqpoint{0.450000in}{0.780000in}}%
\pgfpathlineto{\pgfqpoint{1.250000in}{0.780000in}}%
\pgfusepath{stroke}%
\end{pgfscope}%
\begin{pgfscope}%
\pgfsetbuttcap%
\pgfsetroundjoin%
\definecolor{currentfill}{rgb}{0.321569,0.317647,0.305882}%
\pgfsetfillcolor{currentfill}%
\pgfsetlinewidth{0.602250pt}%
\definecolor{currentstroke}{rgb}{0.321569,0.317647,0.305882}%
\pgfsetstrokecolor{currentstroke}%
\pgfsetdash{}{0pt}%
\pgfsys@defobject{currentmarker}{\pgfqpoint{-0.034722in}{0.000000in}}{\pgfqpoint{-0.000000in}{0.000000in}}{%
\pgfpathmoveto{\pgfqpoint{-0.000000in}{0.000000in}}%
\pgfpathlineto{\pgfqpoint{-0.034722in}{0.000000in}}%
\pgfusepath{stroke,fill}%
}%
\begin{pgfscope}%
\pgfsys@transformshift{0.450000in}{0.780000in}%
\pgfsys@useobject{currentmarker}{}%
\end{pgfscope}%
\end{pgfscope}%
\begin{pgfscope}%
\definecolor{textcolor}{rgb}{0.101961,0.101961,0.098039}%
\pgfsetstrokecolor{textcolor}%
\pgfsetfillcolor{textcolor}%
\pgftext[x=0.311111in, y=0.742112in, left, base]{\color{textcolor}\rmfamily\fontsize{8.000000}{9.600000}\selectfont 0}%
\end{pgfscope}%
\begin{pgfscope}%
\pgfpathrectangle{\pgfqpoint{0.450000in}{0.780000in}}{\pgfqpoint{0.800000in}{0.800000in}}%
\pgfusepath{clip}%
\pgfsetrectcap%
\pgfsetroundjoin%
\pgfsetlinewidth{0.501875pt}%
\definecolor{currentstroke}{rgb}{0.886275,0.882353,0.866667}%
\pgfsetstrokecolor{currentstroke}%
\pgfsetdash{}{0pt}%
\pgfpathmoveto{\pgfqpoint{0.450000in}{1.180000in}}%
\pgfpathlineto{\pgfqpoint{1.250000in}{1.180000in}}%
\pgfusepath{stroke}%
\end{pgfscope}%
\begin{pgfscope}%
\pgfsetbuttcap%
\pgfsetroundjoin%
\definecolor{currentfill}{rgb}{0.321569,0.317647,0.305882}%
\pgfsetfillcolor{currentfill}%
\pgfsetlinewidth{0.602250pt}%
\definecolor{currentstroke}{rgb}{0.321569,0.317647,0.305882}%
\pgfsetstrokecolor{currentstroke}%
\pgfsetdash{}{0pt}%
\pgfsys@defobject{currentmarker}{\pgfqpoint{-0.034722in}{0.000000in}}{\pgfqpoint{-0.000000in}{0.000000in}}{%
\pgfpathmoveto{\pgfqpoint{-0.000000in}{0.000000in}}%
\pgfpathlineto{\pgfqpoint{-0.034722in}{0.000000in}}%
\pgfusepath{stroke,fill}%
}%
\begin{pgfscope}%
\pgfsys@transformshift{0.450000in}{1.180000in}%
\pgfsys@useobject{currentmarker}{}%
\end{pgfscope}%
\end{pgfscope}%
\begin{pgfscope}%
\definecolor{textcolor}{rgb}{0.101961,0.101961,0.098039}%
\pgfsetstrokecolor{textcolor}%
\pgfsetfillcolor{textcolor}%
\pgftext[x=0.255556in, y=1.142112in, left, base]{\color{textcolor}\rmfamily\fontsize{8.000000}{9.600000}\selectfont 50}%
\end{pgfscope}%
\begin{pgfscope}%
\pgfpathrectangle{\pgfqpoint{0.450000in}{0.780000in}}{\pgfqpoint{0.800000in}{0.800000in}}%
\pgfusepath{clip}%
\pgfsetrectcap%
\pgfsetroundjoin%
\pgfsetlinewidth{0.501875pt}%
\definecolor{currentstroke}{rgb}{0.886275,0.882353,0.866667}%
\pgfsetstrokecolor{currentstroke}%
\pgfsetdash{}{0pt}%
\pgfpathmoveto{\pgfqpoint{0.450000in}{1.580000in}}%
\pgfpathlineto{\pgfqpoint{1.250000in}{1.580000in}}%
\pgfusepath{stroke}%
\end{pgfscope}%
\begin{pgfscope}%
\pgfsetbuttcap%
\pgfsetroundjoin%
\definecolor{currentfill}{rgb}{0.321569,0.317647,0.305882}%
\pgfsetfillcolor{currentfill}%
\pgfsetlinewidth{0.602250pt}%
\definecolor{currentstroke}{rgb}{0.321569,0.317647,0.305882}%
\pgfsetstrokecolor{currentstroke}%
\pgfsetdash{}{0pt}%
\pgfsys@defobject{currentmarker}{\pgfqpoint{-0.034722in}{0.000000in}}{\pgfqpoint{-0.000000in}{0.000000in}}{%
\pgfpathmoveto{\pgfqpoint{-0.000000in}{0.000000in}}%
\pgfpathlineto{\pgfqpoint{-0.034722in}{0.000000in}}%
\pgfusepath{stroke,fill}%
}%
\begin{pgfscope}%
\pgfsys@transformshift{0.450000in}{1.580000in}%
\pgfsys@useobject{currentmarker}{}%
\end{pgfscope}%
\end{pgfscope}%
\begin{pgfscope}%
\definecolor{textcolor}{rgb}{0.101961,0.101961,0.098039}%
\pgfsetstrokecolor{textcolor}%
\pgfsetfillcolor{textcolor}%
\pgftext[x=0.200000in, y=1.542112in, left, base]{\color{textcolor}\rmfamily\fontsize{8.000000}{9.600000}\selectfont 100}%
\end{pgfscope}%
\begin{pgfscope}%
\definecolor{textcolor}{rgb}{0.101961,0.101961,0.098039}%
\pgfsetstrokecolor{textcolor}%
\pgfsetfillcolor{textcolor}%
\pgftext[x=0.144444in,y=1.180000in,,bottom,rotate=90.000000]{\color{textcolor}\rmfamily\fontsize{9.000000}{10.800000}\selectfont recovered (\%)}%
\end{pgfscope}%
\begin{pgfscope}%
\pgfpathrectangle{\pgfqpoint{0.450000in}{0.780000in}}{\pgfqpoint{0.800000in}{0.800000in}}%
\pgfusepath{clip}%
\pgfsetrectcap%
\pgfsetroundjoin%
\pgfsetlinewidth{0.602250pt}%
\definecolor{currentstroke}{rgb}{0.321569,0.317647,0.305882}%
\pgfsetstrokecolor{currentstroke}%
\pgfsetdash{}{0pt}%
\pgfpathmoveto{\pgfqpoint{0.450000in}{0.780000in}}%
\pgfpathlineto{\pgfqpoint{1.250000in}{0.780000in}}%
\pgfusepath{stroke}%
\end{pgfscope}%
\begin{pgfscope}%
\pgfpathrectangle{\pgfqpoint{0.450000in}{0.780000in}}{\pgfqpoint{0.800000in}{0.800000in}}%
\pgfusepath{clip}%
\pgfsetbuttcap%
\pgfsetmiterjoin%
\definecolor{currentfill}{rgb}{0.564118,0.804118,0.820000}%
\pgfsetfillcolor{currentfill}%
\pgfsetlinewidth{0.000000pt}%
\definecolor{currentstroke}{rgb}{0.000000,0.000000,0.000000}%
\pgfsetstrokecolor{currentstroke}%
\pgfsetstrokeopacity{0.000000}%
\pgfsetdash{}{0pt}%
\pgfpathmoveto{\pgfqpoint{0.491905in}{0.781117in}}%
\pgfpathlineto{\pgfqpoint{0.564286in}{0.781117in}}%
\pgfpathlineto{\pgfqpoint{0.564286in}{1.021790in}}%
\pgfpathlineto{\pgfqpoint{0.491905in}{1.021790in}}%
\pgfpathlineto{\pgfqpoint{0.491905in}{0.781117in}}%
\pgfpathclose%
\pgfusepath{fill}%
\end{pgfscope}%
\begin{pgfscope}%
\pgfpathrectangle{\pgfqpoint{0.450000in}{0.780000in}}{\pgfqpoint{0.800000in}{0.800000in}}%
\pgfusepath{clip}%
\pgfsetbuttcap%
\pgfsetmiterjoin%
\definecolor{currentfill}{rgb}{0.938235,0.650588,0.760000}%
\pgfsetfillcolor{currentfill}%
\pgfsetlinewidth{0.000000pt}%
\definecolor{currentstroke}{rgb}{0.000000,0.000000,0.000000}%
\pgfsetstrokecolor{currentstroke}%
\pgfsetstrokeopacity{0.000000}%
\pgfsetdash{}{0pt}%
\pgfpathmoveto{\pgfqpoint{0.564286in}{0.781883in}}%
\pgfpathlineto{\pgfqpoint{0.636667in}{0.781883in}}%
\pgfpathlineto{\pgfqpoint{0.636667in}{1.019443in}}%
\pgfpathlineto{\pgfqpoint{0.564286in}{1.019443in}}%
\pgfpathlineto{\pgfqpoint{0.564286in}{0.781883in}}%
\pgfpathclose%
\pgfusepath{fill}%
\end{pgfscope}%
\begin{pgfscope}%
\pgfpathrectangle{\pgfqpoint{0.450000in}{0.780000in}}{\pgfqpoint{0.800000in}{0.800000in}}%
\pgfusepath{clip}%
\pgfsetbuttcap%
\pgfsetmiterjoin%
\definecolor{currentfill}{rgb}{0.564118,0.804118,0.820000}%
\pgfsetfillcolor{currentfill}%
\pgfsetlinewidth{0.000000pt}%
\definecolor{currentstroke}{rgb}{0.000000,0.000000,0.000000}%
\pgfsetstrokecolor{currentstroke}%
\pgfsetstrokeopacity{0.000000}%
\pgfsetdash{}{0pt}%
\pgfpathmoveto{\pgfqpoint{0.682381in}{0.770706in}}%
\pgfpathlineto{\pgfqpoint{0.754762in}{0.770706in}}%
\pgfpathlineto{\pgfqpoint{0.754762in}{1.305120in}}%
\pgfpathlineto{\pgfqpoint{0.682381in}{1.305120in}}%
\pgfpathlineto{\pgfqpoint{0.682381in}{0.770706in}}%
\pgfpathclose%
\pgfusepath{fill}%
\end{pgfscope}%
\begin{pgfscope}%
\pgfpathrectangle{\pgfqpoint{0.450000in}{0.780000in}}{\pgfqpoint{0.800000in}{0.800000in}}%
\pgfusepath{clip}%
\pgfsetbuttcap%
\pgfsetmiterjoin%
\definecolor{currentfill}{rgb}{0.938235,0.650588,0.760000}%
\pgfsetfillcolor{currentfill}%
\pgfsetlinewidth{0.000000pt}%
\definecolor{currentstroke}{rgb}{0.000000,0.000000,0.000000}%
\pgfsetstrokecolor{currentstroke}%
\pgfsetstrokeopacity{0.000000}%
\pgfsetdash{}{0pt}%
\pgfpathmoveto{\pgfqpoint{0.754762in}{0.781635in}}%
\pgfpathlineto{\pgfqpoint{0.827143in}{0.781635in}}%
\pgfpathlineto{\pgfqpoint{0.827143in}{1.324483in}}%
\pgfpathlineto{\pgfqpoint{0.754762in}{1.324483in}}%
\pgfpathlineto{\pgfqpoint{0.754762in}{0.781635in}}%
\pgfpathclose%
\pgfusepath{fill}%
\end{pgfscope}%
\begin{pgfscope}%
\pgfpathrectangle{\pgfqpoint{0.450000in}{0.780000in}}{\pgfqpoint{0.800000in}{0.800000in}}%
\pgfusepath{clip}%
\pgfsetbuttcap%
\pgfsetmiterjoin%
\definecolor{currentfill}{rgb}{0.564118,0.804118,0.820000}%
\pgfsetfillcolor{currentfill}%
\pgfsetlinewidth{0.000000pt}%
\definecolor{currentstroke}{rgb}{0.000000,0.000000,0.000000}%
\pgfsetstrokecolor{currentstroke}%
\pgfsetstrokeopacity{0.000000}%
\pgfsetdash{}{0pt}%
\pgfpathmoveto{\pgfqpoint{0.872857in}{0.789739in}}%
\pgfpathlineto{\pgfqpoint{0.945238in}{0.789739in}}%
\pgfpathlineto{\pgfqpoint{0.945238in}{1.127276in}}%
\pgfpathlineto{\pgfqpoint{0.872857in}{1.127276in}}%
\pgfpathlineto{\pgfqpoint{0.872857in}{0.789739in}}%
\pgfpathclose%
\pgfusepath{fill}%
\end{pgfscope}%
\begin{pgfscope}%
\pgfpathrectangle{\pgfqpoint{0.450000in}{0.780000in}}{\pgfqpoint{0.800000in}{0.800000in}}%
\pgfusepath{clip}%
\pgfsetbuttcap%
\pgfsetmiterjoin%
\definecolor{currentfill}{rgb}{0.938235,0.650588,0.760000}%
\pgfsetfillcolor{currentfill}%
\pgfsetlinewidth{0.000000pt}%
\definecolor{currentstroke}{rgb}{0.000000,0.000000,0.000000}%
\pgfsetstrokecolor{currentstroke}%
\pgfsetstrokeopacity{0.000000}%
\pgfsetdash{}{0pt}%
\pgfpathmoveto{\pgfqpoint{0.945238in}{0.791040in}}%
\pgfpathlineto{\pgfqpoint{1.017619in}{0.791040in}}%
\pgfpathlineto{\pgfqpoint{1.017619in}{1.141506in}}%
\pgfpathlineto{\pgfqpoint{0.945238in}{1.141506in}}%
\pgfpathlineto{\pgfqpoint{0.945238in}{0.791040in}}%
\pgfpathclose%
\pgfusepath{fill}%
\end{pgfscope}%
\begin{pgfscope}%
\pgfpathrectangle{\pgfqpoint{0.450000in}{0.780000in}}{\pgfqpoint{0.800000in}{0.800000in}}%
\pgfusepath{clip}%
\pgfsetbuttcap%
\pgfsetmiterjoin%
\definecolor{currentfill}{rgb}{0.564118,0.804118,0.820000}%
\pgfsetfillcolor{currentfill}%
\pgfsetlinewidth{0.000000pt}%
\definecolor{currentstroke}{rgb}{0.000000,0.000000,0.000000}%
\pgfsetstrokecolor{currentstroke}%
\pgfsetstrokeopacity{0.000000}%
\pgfsetdash{}{0pt}%
\pgfpathmoveto{\pgfqpoint{1.063333in}{0.819628in}}%
\pgfpathlineto{\pgfqpoint{1.135714in}{0.819628in}}%
\pgfpathlineto{\pgfqpoint{1.135714in}{1.357326in}}%
\pgfpathlineto{\pgfqpoint{1.063333in}{1.357326in}}%
\pgfpathlineto{\pgfqpoint{1.063333in}{0.819628in}}%
\pgfpathclose%
\pgfusepath{fill}%
\end{pgfscope}%
\begin{pgfscope}%
\pgfpathrectangle{\pgfqpoint{0.450000in}{0.780000in}}{\pgfqpoint{0.800000in}{0.800000in}}%
\pgfusepath{clip}%
\pgfsetbuttcap%
\pgfsetmiterjoin%
\definecolor{currentfill}{rgb}{0.938235,0.650588,0.760000}%
\pgfsetfillcolor{currentfill}%
\pgfsetlinewidth{0.000000pt}%
\definecolor{currentstroke}{rgb}{0.000000,0.000000,0.000000}%
\pgfsetstrokecolor{currentstroke}%
\pgfsetstrokeopacity{0.000000}%
\pgfsetdash{}{0pt}%
\pgfpathmoveto{\pgfqpoint{1.135714in}{0.792760in}}%
\pgfpathlineto{\pgfqpoint{1.208095in}{0.792760in}}%
\pgfpathlineto{\pgfqpoint{1.208095in}{1.389016in}}%
\pgfpathlineto{\pgfqpoint{1.135714in}{1.389016in}}%
\pgfpathlineto{\pgfqpoint{1.135714in}{0.792760in}}%
\pgfpathclose%
\pgfusepath{fill}%
\end{pgfscope}%
\begin{pgfscope}%
\pgfpathrectangle{\pgfqpoint{0.450000in}{0.780000in}}{\pgfqpoint{0.800000in}{0.800000in}}%
\pgfusepath{clip}%
\pgfsetbuttcap%
\pgfsetmiterjoin%
\definecolor{currentfill}{rgb}{0.031373,0.564706,0.600000}%
\pgfsetfillcolor{currentfill}%
\pgfsetlinewidth{0.000000pt}%
\definecolor{currentstroke}{rgb}{0.000000,0.000000,0.000000}%
\pgfsetstrokecolor{currentstroke}%
\pgfsetstrokeopacity{0.000000}%
\pgfsetdash{}{0pt}%
\pgfpathmoveto{\pgfqpoint{0.491905in}{0.780000in}}%
\pgfpathlineto{\pgfqpoint{0.564286in}{0.780000in}}%
\pgfpathlineto{\pgfqpoint{0.564286in}{0.781117in}}%
\pgfpathlineto{\pgfqpoint{0.491905in}{0.781117in}}%
\pgfpathlineto{\pgfqpoint{0.491905in}{0.780000in}}%
\pgfpathclose%
\pgfusepath{fill}%
\end{pgfscope}%
\begin{pgfscope}%
\pgfpathrectangle{\pgfqpoint{0.450000in}{0.780000in}}{\pgfqpoint{0.800000in}{0.800000in}}%
\pgfusepath{clip}%
\pgfsetbuttcap%
\pgfsetmiterjoin%
\definecolor{currentfill}{rgb}{0.862745,0.223529,0.466667}%
\pgfsetfillcolor{currentfill}%
\pgfsetlinewidth{0.000000pt}%
\definecolor{currentstroke}{rgb}{0.000000,0.000000,0.000000}%
\pgfsetstrokecolor{currentstroke}%
\pgfsetstrokeopacity{0.000000}%
\pgfsetdash{}{0pt}%
\pgfpathmoveto{\pgfqpoint{0.564286in}{0.780000in}}%
\pgfpathlineto{\pgfqpoint{0.636667in}{0.780000in}}%
\pgfpathlineto{\pgfqpoint{0.636667in}{0.781883in}}%
\pgfpathlineto{\pgfqpoint{0.564286in}{0.781883in}}%
\pgfpathlineto{\pgfqpoint{0.564286in}{0.780000in}}%
\pgfpathclose%
\pgfusepath{fill}%
\end{pgfscope}%
\begin{pgfscope}%
\pgfpathrectangle{\pgfqpoint{0.450000in}{0.780000in}}{\pgfqpoint{0.800000in}{0.800000in}}%
\pgfusepath{clip}%
\pgfsetbuttcap%
\pgfsetmiterjoin%
\definecolor{currentfill}{rgb}{0.031373,0.564706,0.600000}%
\pgfsetfillcolor{currentfill}%
\pgfsetlinewidth{0.000000pt}%
\definecolor{currentstroke}{rgb}{0.000000,0.000000,0.000000}%
\pgfsetstrokecolor{currentstroke}%
\pgfsetstrokeopacity{0.000000}%
\pgfsetdash{}{0pt}%
\pgfpathmoveto{\pgfqpoint{0.682381in}{0.780000in}}%
\pgfpathlineto{\pgfqpoint{0.754762in}{0.780000in}}%
\pgfpathlineto{\pgfqpoint{0.754762in}{0.770706in}}%
\pgfpathlineto{\pgfqpoint{0.682381in}{0.770706in}}%
\pgfpathlineto{\pgfqpoint{0.682381in}{0.780000in}}%
\pgfpathclose%
\pgfusepath{fill}%
\end{pgfscope}%
\begin{pgfscope}%
\pgfpathrectangle{\pgfqpoint{0.450000in}{0.780000in}}{\pgfqpoint{0.800000in}{0.800000in}}%
\pgfusepath{clip}%
\pgfsetbuttcap%
\pgfsetmiterjoin%
\definecolor{currentfill}{rgb}{0.862745,0.223529,0.466667}%
\pgfsetfillcolor{currentfill}%
\pgfsetlinewidth{0.000000pt}%
\definecolor{currentstroke}{rgb}{0.000000,0.000000,0.000000}%
\pgfsetstrokecolor{currentstroke}%
\pgfsetstrokeopacity{0.000000}%
\pgfsetdash{}{0pt}%
\pgfpathmoveto{\pgfqpoint{0.754762in}{0.780000in}}%
\pgfpathlineto{\pgfqpoint{0.827143in}{0.780000in}}%
\pgfpathlineto{\pgfqpoint{0.827143in}{0.781635in}}%
\pgfpathlineto{\pgfqpoint{0.754762in}{0.781635in}}%
\pgfpathlineto{\pgfqpoint{0.754762in}{0.780000in}}%
\pgfpathclose%
\pgfusepath{fill}%
\end{pgfscope}%
\begin{pgfscope}%
\pgfpathrectangle{\pgfqpoint{0.450000in}{0.780000in}}{\pgfqpoint{0.800000in}{0.800000in}}%
\pgfusepath{clip}%
\pgfsetbuttcap%
\pgfsetmiterjoin%
\definecolor{currentfill}{rgb}{0.031373,0.564706,0.600000}%
\pgfsetfillcolor{currentfill}%
\pgfsetlinewidth{0.000000pt}%
\definecolor{currentstroke}{rgb}{0.000000,0.000000,0.000000}%
\pgfsetstrokecolor{currentstroke}%
\pgfsetstrokeopacity{0.000000}%
\pgfsetdash{}{0pt}%
\pgfpathmoveto{\pgfqpoint{0.872857in}{0.780000in}}%
\pgfpathlineto{\pgfqpoint{0.945238in}{0.780000in}}%
\pgfpathlineto{\pgfqpoint{0.945238in}{0.789739in}}%
\pgfpathlineto{\pgfqpoint{0.872857in}{0.789739in}}%
\pgfpathlineto{\pgfqpoint{0.872857in}{0.780000in}}%
\pgfpathclose%
\pgfusepath{fill}%
\end{pgfscope}%
\begin{pgfscope}%
\pgfpathrectangle{\pgfqpoint{0.450000in}{0.780000in}}{\pgfqpoint{0.800000in}{0.800000in}}%
\pgfusepath{clip}%
\pgfsetbuttcap%
\pgfsetmiterjoin%
\definecolor{currentfill}{rgb}{0.862745,0.223529,0.466667}%
\pgfsetfillcolor{currentfill}%
\pgfsetlinewidth{0.000000pt}%
\definecolor{currentstroke}{rgb}{0.000000,0.000000,0.000000}%
\pgfsetstrokecolor{currentstroke}%
\pgfsetstrokeopacity{0.000000}%
\pgfsetdash{}{0pt}%
\pgfpathmoveto{\pgfqpoint{0.945238in}{0.780000in}}%
\pgfpathlineto{\pgfqpoint{1.017619in}{0.780000in}}%
\pgfpathlineto{\pgfqpoint{1.017619in}{0.791040in}}%
\pgfpathlineto{\pgfqpoint{0.945238in}{0.791040in}}%
\pgfpathlineto{\pgfqpoint{0.945238in}{0.780000in}}%
\pgfpathclose%
\pgfusepath{fill}%
\end{pgfscope}%
\begin{pgfscope}%
\pgfpathrectangle{\pgfqpoint{0.450000in}{0.780000in}}{\pgfqpoint{0.800000in}{0.800000in}}%
\pgfusepath{clip}%
\pgfsetbuttcap%
\pgfsetmiterjoin%
\definecolor{currentfill}{rgb}{0.031373,0.564706,0.600000}%
\pgfsetfillcolor{currentfill}%
\pgfsetlinewidth{0.000000pt}%
\definecolor{currentstroke}{rgb}{0.000000,0.000000,0.000000}%
\pgfsetstrokecolor{currentstroke}%
\pgfsetstrokeopacity{0.000000}%
\pgfsetdash{}{0pt}%
\pgfpathmoveto{\pgfqpoint{1.063333in}{0.780000in}}%
\pgfpathlineto{\pgfqpoint{1.135714in}{0.780000in}}%
\pgfpathlineto{\pgfqpoint{1.135714in}{0.819628in}}%
\pgfpathlineto{\pgfqpoint{1.063333in}{0.819628in}}%
\pgfpathlineto{\pgfqpoint{1.063333in}{0.780000in}}%
\pgfpathclose%
\pgfusepath{fill}%
\end{pgfscope}%
\begin{pgfscope}%
\pgfpathrectangle{\pgfqpoint{0.450000in}{0.780000in}}{\pgfqpoint{0.800000in}{0.800000in}}%
\pgfusepath{clip}%
\pgfsetbuttcap%
\pgfsetmiterjoin%
\definecolor{currentfill}{rgb}{0.862745,0.223529,0.466667}%
\pgfsetfillcolor{currentfill}%
\pgfsetlinewidth{0.000000pt}%
\definecolor{currentstroke}{rgb}{0.000000,0.000000,0.000000}%
\pgfsetstrokecolor{currentstroke}%
\pgfsetstrokeopacity{0.000000}%
\pgfsetdash{}{0pt}%
\pgfpathmoveto{\pgfqpoint{1.135714in}{0.780000in}}%
\pgfpathlineto{\pgfqpoint{1.208095in}{0.780000in}}%
\pgfpathlineto{\pgfqpoint{1.208095in}{0.792760in}}%
\pgfpathlineto{\pgfqpoint{1.135714in}{0.792760in}}%
\pgfpathlineto{\pgfqpoint{1.135714in}{0.780000in}}%
\pgfpathclose%
\pgfusepath{fill}%
\end{pgfscope}%
\begin{pgfscope}%
\pgfsetrectcap%
\pgfsetmiterjoin%
\pgfsetlinewidth{0.602250pt}%
\definecolor{currentstroke}{rgb}{0.321569,0.317647,0.305882}%
\pgfsetstrokecolor{currentstroke}%
\pgfsetdash{}{0pt}%
\pgfpathmoveto{\pgfqpoint{0.450000in}{0.780000in}}%
\pgfpathlineto{\pgfqpoint{0.450000in}{1.580000in}}%
\pgfusepath{stroke}%
\end{pgfscope}%
\begin{pgfscope}%
\pgfsetrectcap%
\pgfsetmiterjoin%
\pgfsetlinewidth{0.602250pt}%
\definecolor{currentstroke}{rgb}{0.321569,0.317647,0.305882}%
\pgfsetstrokecolor{currentstroke}%
\pgfsetdash{}{0pt}%
\pgfpathmoveto{\pgfqpoint{0.450000in}{0.780000in}}%
\pgfpathlineto{\pgfqpoint{1.250000in}{0.780000in}}%
\pgfusepath{stroke}%
\end{pgfscope}%
\begin{pgfscope}%
\pgfsetbuttcap%
\pgfsetmiterjoin%
\definecolor{currentfill}{rgb}{0.031373,0.564706,0.600000}%
\pgfsetfillcolor{currentfill}%
\pgfsetlinewidth{1.003750pt}%
\definecolor{currentstroke}{rgb}{0.031373,0.564706,0.600000}%
\pgfsetstrokecolor{currentstroke}%
\pgfsetdash{}{0pt}%
\pgfpathmoveto{\pgfqpoint{0.299082in}{0.207778in}}%
\pgfpathlineto{\pgfqpoint{0.421304in}{0.207778in}}%
\pgfpathlineto{\pgfqpoint{0.421304in}{0.285556in}}%
\pgfpathlineto{\pgfqpoint{0.299082in}{0.285556in}}%
\pgfpathlineto{\pgfqpoint{0.299082in}{0.207778in}}%
\pgfpathclose%
\pgfusepath{stroke,fill}%
\end{pgfscope}%
\begin{pgfscope}%
\definecolor{textcolor}{rgb}{0.101961,0.101961,0.098039}%
\pgfsetstrokecolor{textcolor}%
\pgfsetfillcolor{textcolor}%
\pgftext[x=0.454637in,y=0.207778in,left,base]{\color{textcolor}\rmfamily\fontsize{8.000000}{9.600000}\selectfont \(\displaystyle \RotTerm{\mathbf{R}_{\mathrm{V}}}\)}%
\end{pgfscope}%
\begin{pgfscope}%
\pgfsetbuttcap%
\pgfsetmiterjoin%
\definecolor{currentfill}{rgb}{0.862745,0.223529,0.466667}%
\pgfsetfillcolor{currentfill}%
\pgfsetlinewidth{1.003750pt}%
\definecolor{currentstroke}{rgb}{0.862745,0.223529,0.466667}%
\pgfsetstrokecolor{currentstroke}%
\pgfsetdash{}{0pt}%
\pgfpathmoveto{\pgfqpoint{0.728210in}{0.207778in}}%
\pgfpathlineto{\pgfqpoint{0.850432in}{0.207778in}}%
\pgfpathlineto{\pgfqpoint{0.850432in}{0.285556in}}%
\pgfpathlineto{\pgfqpoint{0.728210in}{0.285556in}}%
\pgfpathlineto{\pgfqpoint{0.728210in}{0.207778in}}%
\pgfpathclose%
\pgfusepath{stroke,fill}%
\end{pgfscope}%
\begin{pgfscope}%
\definecolor{textcolor}{rgb}{0.101961,0.101961,0.098039}%
\pgfsetstrokecolor{textcolor}%
\pgfsetfillcolor{textcolor}%
\pgftext[x=0.883765in,y=0.207778in,left,base]{\color{textcolor}\rmfamily\fontsize{8.000000}{9.600000}\selectfont \(\displaystyle \RescaleTerm{\mathbf{D}_{\mathrm{V}}}\)}%
\end{pgfscope}%
\begin{pgfscope}%
\pgfsetbuttcap%
\pgfsetmiterjoin%
\definecolor{currentfill}{rgb}{0.300000,0.300000,0.300000}%
\pgfsetfillcolor{currentfill}%
\pgfsetlinewidth{1.003750pt}%
\definecolor{currentstroke}{rgb}{0.300000,0.300000,0.300000}%
\pgfsetstrokecolor{currentstroke}%
\pgfsetdash{}{0pt}%
\pgfpathmoveto{\pgfqpoint{0.234734in}{0.052222in}}%
\pgfpathlineto{\pgfqpoint{0.356956in}{0.052222in}}%
\pgfpathlineto{\pgfqpoint{0.356956in}{0.130000in}}%
\pgfpathlineto{\pgfqpoint{0.234734in}{0.130000in}}%
\pgfpathlineto{\pgfqpoint{0.234734in}{0.052222in}}%
\pgfpathclose%
\pgfusepath{stroke,fill}%
\end{pgfscope}%
\begin{pgfscope}%
\definecolor{textcolor}{rgb}{0.101961,0.101961,0.098039}%
\pgfsetstrokecolor{textcolor}%
\pgfsetfillcolor{textcolor}%
\pgftext[x=0.390290in,y=0.052222in,left,base]{\color{textcolor}\rmfamily\fontsize{8.000000}{9.600000}\selectfont \(\displaystyle \langle\)T,C\(\displaystyle \rangle\)}%
\end{pgfscope}%
\begin{pgfscope}%
\pgfsetbuttcap%
\pgfsetmiterjoin%
\definecolor{currentfill}{rgb}{0.685000,0.685000,0.685000}%
\pgfsetfillcolor{currentfill}%
\pgfsetlinewidth{1.003750pt}%
\definecolor{currentstroke}{rgb}{0.685000,0.685000,0.685000}%
\pgfsetstrokecolor{currentstroke}%
\pgfsetdash{}{0pt}%
\pgfpathmoveto{\pgfqpoint{0.732555in}{0.052222in}}%
\pgfpathlineto{\pgfqpoint{0.854777in}{0.052222in}}%
\pgfpathlineto{\pgfqpoint{0.854777in}{0.130000in}}%
\pgfpathlineto{\pgfqpoint{0.732555in}{0.130000in}}%
\pgfpathlineto{\pgfqpoint{0.732555in}{0.052222in}}%
\pgfpathclose%
\pgfusepath{stroke,fill}%
\end{pgfscope}%
\begin{pgfscope}%
\definecolor{textcolor}{rgb}{0.101961,0.101961,0.098039}%
\pgfsetstrokecolor{textcolor}%
\pgfsetfillcolor{textcolor}%
\pgftext[x=0.888111in,y=0.052222in,left,base]{\color{textcolor}\rmfamily\fontsize{8.000000}{9.600000}\selectfont \(\displaystyle \langle\)T,T\(\displaystyle \rangle\)}%
\end{pgfscope}%
\end{pgfpicture}%
\makeatother%
\endgroup%

%% file: figures/pgf/fig_storage_diff.pgf
\providecommand{\mathdefault}[1]{#1}%
\begingroup%
\makeatletter%
\begin{pgfpicture}%
\pgfpathrectangle{\pgfpointorigin}{\pgfqpoint{1.370000in}{1.640000in}}%
\pgfusepath{use as bounding box, clip}%
\begin{pgfscope}%
\pgfsetbuttcap%
\pgfsetmiterjoin%
\definecolor{currentfill}{rgb}{1.000000,1.000000,1.000000}%
\pgfsetfillcolor{currentfill}%
\pgfsetlinewidth{0.000000pt}%
\definecolor{currentstroke}{rgb}{1.000000,1.000000,1.000000}%
\pgfsetstrokecolor{currentstroke}%
\pgfsetdash{}{0pt}%
\pgfpathmoveto{\pgfqpoint{0.000000in}{0.000000in}}%
\pgfpathlineto{\pgfqpoint{1.370000in}{0.000000in}}%
\pgfpathlineto{\pgfqpoint{1.370000in}{1.640000in}}%
\pgfpathlineto{\pgfqpoint{0.000000in}{1.640000in}}%
\pgfpathlineto{\pgfqpoint{0.000000in}{0.000000in}}%
\pgfpathclose%
\pgfusepath{fill}%
\end{pgfscope}%
\begin{pgfscope}%
\pgfsetbuttcap%
\pgfsetmiterjoin%
\definecolor{currentfill}{rgb}{1.000000,1.000000,1.000000}%
\pgfsetfillcolor{currentfill}%
\pgfsetlinewidth{0.000000pt}%
\definecolor{currentstroke}{rgb}{0.000000,0.000000,0.000000}%
\pgfsetstrokecolor{currentstroke}%
\pgfsetstrokeopacity{0.000000}%
\pgfsetdash{}{0pt}%
\pgfpathmoveto{\pgfqpoint{0.450000in}{0.780000in}}%
\pgfpathlineto{\pgfqpoint{1.250000in}{0.780000in}}%
\pgfpathlineto{\pgfqpoint{1.250000in}{1.580000in}}%
\pgfpathlineto{\pgfqpoint{0.450000in}{1.580000in}}%
\pgfpathlineto{\pgfqpoint{0.450000in}{0.780000in}}%
\pgfpathclose%
\pgfusepath{fill}%
\end{pgfscope}%
\begin{pgfscope}%
\definecolor{textcolor}{rgb}{0.101961,0.101961,0.098039}%
\pgfsetstrokecolor{textcolor}%
\pgfsetfillcolor{textcolor}%
\pgftext[x=0.538889in,y=0.731389in,right,top,rotate=60.000000]{\color{textcolor}\rmfamily\fontsize{8.000000}{9.600000}\selectfont LA}%
\end{pgfscope}%
\begin{pgfscope}%
\definecolor{textcolor}{rgb}{0.101961,0.101961,0.098039}%
\pgfsetstrokecolor{textcolor}%
\pgfsetfillcolor{textcolor}%
\pgftext[x=0.775926in,y=0.731389in,right,top,rotate=60.000000]{\color{textcolor}\rmfamily\fontsize{8.000000}{9.600000}\selectfont M2}%
\end{pgfscope}%
\begin{pgfscope}%
\definecolor{textcolor}{rgb}{0.101961,0.101961,0.098039}%
\pgfsetstrokecolor{textcolor}%
\pgfsetfillcolor{textcolor}%
\pgftext[x=0.924074in,y=0.731389in,right,top,rotate=60.000000]{\color{textcolor}\rmfamily\fontsize{8.000000}{9.600000}\selectfont GDN}%
\end{pgfscope}%
\begin{pgfscope}%
\definecolor{textcolor}{rgb}{0.101961,0.101961,0.098039}%
\pgfsetstrokecolor{textcolor}%
\pgfsetfillcolor{textcolor}%
\pgftext[x=1.161111in,y=0.731389in,right,top,rotate=60.000000]{\color{textcolor}\rmfamily\fontsize{8.000000}{9.600000}\selectfont GLA}%
\end{pgfscope}%
\begin{pgfscope}%
\pgfpathrectangle{\pgfqpoint{0.450000in}{0.780000in}}{\pgfqpoint{0.800000in}{0.800000in}}%
\pgfusepath{clip}%
\pgfsetrectcap%
\pgfsetroundjoin%
\pgfsetlinewidth{0.501875pt}%
\definecolor{currentstroke}{rgb}{0.886275,0.882353,0.866667}%
\pgfsetstrokecolor{currentstroke}%
\pgfsetdash{}{0pt}%
\pgfpathmoveto{\pgfqpoint{0.450000in}{0.818095in}}%
\pgfpathlineto{\pgfqpoint{1.250000in}{0.818095in}}%
\pgfusepath{stroke}%
\end{pgfscope}%
\begin{pgfscope}%
\pgfsetbuttcap%
\pgfsetroundjoin%
\definecolor{currentfill}{rgb}{0.321569,0.317647,0.305882}%
\pgfsetfillcolor{currentfill}%
\pgfsetlinewidth{0.602250pt}%
\definecolor{currentstroke}{rgb}{0.321569,0.317647,0.305882}%
\pgfsetstrokecolor{currentstroke}%
\pgfsetdash{}{0pt}%
\pgfsys@defobject{currentmarker}{\pgfqpoint{-0.034722in}{0.000000in}}{\pgfqpoint{-0.000000in}{0.000000in}}{%
\pgfpathmoveto{\pgfqpoint{-0.000000in}{0.000000in}}%
\pgfpathlineto{\pgfqpoint{-0.034722in}{0.000000in}}%
\pgfusepath{stroke,fill}%
}%
\begin{pgfscope}%
\pgfsys@transformshift{0.450000in}{0.818095in}%
\pgfsys@useobject{currentmarker}{}%
\end{pgfscope}%
\end{pgfscope}%
\begin{pgfscope}%
\definecolor{textcolor}{rgb}{0.101961,0.101961,0.098039}%
\pgfsetstrokecolor{textcolor}%
\pgfsetfillcolor{textcolor}%
\pgftext[x=0.311111in, y=0.780207in, left, base]{\color{textcolor}\rmfamily\fontsize{8.000000}{9.600000}\selectfont 0}%
\end{pgfscope}%
\begin{pgfscope}%
\pgfpathrectangle{\pgfqpoint{0.450000in}{0.780000in}}{\pgfqpoint{0.800000in}{0.800000in}}%
\pgfusepath{clip}%
\pgfsetrectcap%
\pgfsetroundjoin%
\pgfsetlinewidth{0.501875pt}%
\definecolor{currentstroke}{rgb}{0.886275,0.882353,0.866667}%
\pgfsetstrokecolor{currentstroke}%
\pgfsetdash{}{0pt}%
\pgfpathmoveto{\pgfqpoint{0.450000in}{1.199048in}}%
\pgfpathlineto{\pgfqpoint{1.250000in}{1.199048in}}%
\pgfusepath{stroke}%
\end{pgfscope}%
\begin{pgfscope}%
\pgfsetbuttcap%
\pgfsetroundjoin%
\definecolor{currentfill}{rgb}{0.321569,0.317647,0.305882}%
\pgfsetfillcolor{currentfill}%
\pgfsetlinewidth{0.602250pt}%
\definecolor{currentstroke}{rgb}{0.321569,0.317647,0.305882}%
\pgfsetstrokecolor{currentstroke}%
\pgfsetdash{}{0pt}%
\pgfsys@defobject{currentmarker}{\pgfqpoint{-0.034722in}{0.000000in}}{\pgfqpoint{-0.000000in}{0.000000in}}{%
\pgfpathmoveto{\pgfqpoint{-0.000000in}{0.000000in}}%
\pgfpathlineto{\pgfqpoint{-0.034722in}{0.000000in}}%
\pgfusepath{stroke,fill}%
}%
\begin{pgfscope}%
\pgfsys@transformshift{0.450000in}{1.199048in}%
\pgfsys@useobject{currentmarker}{}%
\end{pgfscope}%
\end{pgfscope}%
\begin{pgfscope}%
\definecolor{textcolor}{rgb}{0.101961,0.101961,0.098039}%
\pgfsetstrokecolor{textcolor}%
\pgfsetfillcolor{textcolor}%
\pgftext[x=0.255556in, y=1.161159in, left, base]{\color{textcolor}\rmfamily\fontsize{8.000000}{9.600000}\selectfont 35}%
\end{pgfscope}%
\begin{pgfscope}%
\pgfpathrectangle{\pgfqpoint{0.450000in}{0.780000in}}{\pgfqpoint{0.800000in}{0.800000in}}%
\pgfusepath{clip}%
\pgfsetrectcap%
\pgfsetroundjoin%
\pgfsetlinewidth{0.501875pt}%
\definecolor{currentstroke}{rgb}{0.886275,0.882353,0.866667}%
\pgfsetstrokecolor{currentstroke}%
\pgfsetdash{}{0pt}%
\pgfpathmoveto{\pgfqpoint{0.450000in}{1.580000in}}%
\pgfpathlineto{\pgfqpoint{1.250000in}{1.580000in}}%
\pgfusepath{stroke}%
\end{pgfscope}%
\begin{pgfscope}%
\pgfsetbuttcap%
\pgfsetroundjoin%
\definecolor{currentfill}{rgb}{0.321569,0.317647,0.305882}%
\pgfsetfillcolor{currentfill}%
\pgfsetlinewidth{0.602250pt}%
\definecolor{currentstroke}{rgb}{0.321569,0.317647,0.305882}%
\pgfsetstrokecolor{currentstroke}%
\pgfsetdash{}{0pt}%
\pgfsys@defobject{currentmarker}{\pgfqpoint{-0.034722in}{0.000000in}}{\pgfqpoint{-0.000000in}{0.000000in}}{%
\pgfpathmoveto{\pgfqpoint{-0.000000in}{0.000000in}}%
\pgfpathlineto{\pgfqpoint{-0.034722in}{0.000000in}}%
\pgfusepath{stroke,fill}%
}%
\begin{pgfscope}%
\pgfsys@transformshift{0.450000in}{1.580000in}%
\pgfsys@useobject{currentmarker}{}%
\end{pgfscope}%
\end{pgfscope}%
\begin{pgfscope}%
\definecolor{textcolor}{rgb}{0.101961,0.101961,0.098039}%
\pgfsetstrokecolor{textcolor}%
\pgfsetfillcolor{textcolor}%
\pgftext[x=0.255556in, y=1.542112in, left, base]{\color{textcolor}\rmfamily\fontsize{8.000000}{9.600000}\selectfont 70}%
\end{pgfscope}%
\begin{pgfscope}%
\definecolor{textcolor}{rgb}{0.101961,0.101961,0.098039}%
\pgfsetstrokecolor{textcolor}%
\pgfsetfillcolor{textcolor}%
\pgftext[x=0.200000in,y=1.180000in,,bottom,rotate=90.000000]{\color{textcolor}\rmfamily\fontsize{9.000000}{10.800000}\selectfont lost (\%)}%
\end{pgfscope}%
\begin{pgfscope}%
\pgfpathrectangle{\pgfqpoint{0.450000in}{0.780000in}}{\pgfqpoint{0.800000in}{0.800000in}}%
\pgfusepath{clip}%
\pgfsetrectcap%
\pgfsetroundjoin%
\pgfsetlinewidth{0.602250pt}%
\definecolor{currentstroke}{rgb}{0.321569,0.317647,0.305882}%
\pgfsetstrokecolor{currentstroke}%
\pgfsetdash{}{0pt}%
\pgfpathmoveto{\pgfqpoint{0.450000in}{0.818095in}}%
\pgfpathlineto{\pgfqpoint{1.250000in}{0.818095in}}%
\pgfusepath{stroke}%
\end{pgfscope}%
\begin{pgfscope}%
\pgfpathrectangle{\pgfqpoint{0.450000in}{0.780000in}}{\pgfqpoint{0.800000in}{0.800000in}}%
\pgfusepath{clip}%
\pgfsetbuttcap%
\pgfsetmiterjoin%
\definecolor{currentfill}{rgb}{0.486275,0.113725,0.435294}%
\pgfsetfillcolor{currentfill}%
\pgfsetlinewidth{0.000000pt}%
\definecolor{currentstroke}{rgb}{0.000000,0.000000,0.000000}%
\pgfsetstrokecolor{currentstroke}%
\pgfsetstrokeopacity{0.000000}%
\pgfsetdash{}{0pt}%
\pgfpathmoveto{\pgfqpoint{0.479630in}{0.818095in}}%
\pgfpathlineto{\pgfqpoint{0.519136in}{0.818095in}}%
\pgfpathlineto{\pgfqpoint{0.519136in}{1.514310in}}%
\pgfpathlineto{\pgfqpoint{0.479630in}{1.514310in}}%
\pgfpathlineto{\pgfqpoint{0.479630in}{0.818095in}}%
\pgfpathclose%
\pgfusepath{fill}%
\end{pgfscope}%
\begin{pgfscope}%
\pgfpathrectangle{\pgfqpoint{0.450000in}{0.780000in}}{\pgfqpoint{0.800000in}{0.800000in}}%
\pgfusepath{clip}%
\pgfsetbuttcap%
\pgfsetmiterjoin%
\definecolor{currentfill}{rgb}{0.031373,0.564706,0.600000}%
\pgfsetfillcolor{currentfill}%
\pgfsetlinewidth{0.000000pt}%
\definecolor{currentstroke}{rgb}{0.000000,0.000000,0.000000}%
\pgfsetstrokecolor{currentstroke}%
\pgfsetstrokeopacity{0.000000}%
\pgfsetdash{}{0pt}%
\pgfpathmoveto{\pgfqpoint{0.519136in}{0.818095in}}%
\pgfpathlineto{\pgfqpoint{0.558642in}{0.818095in}}%
\pgfpathlineto{\pgfqpoint{0.558642in}{1.090373in}}%
\pgfpathlineto{\pgfqpoint{0.519136in}{1.090373in}}%
\pgfpathlineto{\pgfqpoint{0.519136in}{0.818095in}}%
\pgfpathclose%
\pgfusepath{fill}%
\end{pgfscope}%
\begin{pgfscope}%
\pgfpathrectangle{\pgfqpoint{0.450000in}{0.780000in}}{\pgfqpoint{0.800000in}{0.800000in}}%
\pgfusepath{clip}%
\pgfsetbuttcap%
\pgfsetmiterjoin%
\definecolor{currentfill}{rgb}{0.941176,0.454902,0.431373}%
\pgfsetfillcolor{currentfill}%
\pgfsetlinewidth{0.000000pt}%
\definecolor{currentstroke}{rgb}{0.000000,0.000000,0.000000}%
\pgfsetstrokecolor{currentstroke}%
\pgfsetstrokeopacity{0.000000}%
\pgfsetdash{}{0pt}%
\pgfpathmoveto{\pgfqpoint{0.558642in}{0.818095in}}%
\pgfpathlineto{\pgfqpoint{0.598148in}{0.818095in}}%
\pgfpathlineto{\pgfqpoint{0.598148in}{1.013507in}}%
\pgfpathlineto{\pgfqpoint{0.558642in}{1.013507in}}%
\pgfpathlineto{\pgfqpoint{0.558642in}{0.818095in}}%
\pgfpathclose%
\pgfusepath{fill}%
\end{pgfscope}%
\begin{pgfscope}%
\pgfpathrectangle{\pgfqpoint{0.450000in}{0.780000in}}{\pgfqpoint{0.800000in}{0.800000in}}%
\pgfusepath{clip}%
\pgfsetbuttcap%
\pgfsetmiterjoin%
\definecolor{currentfill}{rgb}{0.486275,0.113725,0.435294}%
\pgfsetfillcolor{currentfill}%
\pgfsetlinewidth{0.000000pt}%
\definecolor{currentstroke}{rgb}{0.000000,0.000000,0.000000}%
\pgfsetstrokecolor{currentstroke}%
\pgfsetstrokeopacity{0.000000}%
\pgfsetdash{}{0pt}%
\pgfpathmoveto{\pgfqpoint{0.716667in}{0.818095in}}%
\pgfpathlineto{\pgfqpoint{0.756173in}{0.818095in}}%
\pgfpathlineto{\pgfqpoint{0.756173in}{0.925149in}}%
\pgfpathlineto{\pgfqpoint{0.716667in}{0.925149in}}%
\pgfpathlineto{\pgfqpoint{0.716667in}{0.818095in}}%
\pgfpathclose%
\pgfusepath{fill}%
\end{pgfscope}%
\begin{pgfscope}%
\pgfpathrectangle{\pgfqpoint{0.450000in}{0.780000in}}{\pgfqpoint{0.800000in}{0.800000in}}%
\pgfusepath{clip}%
\pgfsetbuttcap%
\pgfsetmiterjoin%
\definecolor{currentfill}{rgb}{0.031373,0.564706,0.600000}%
\pgfsetfillcolor{currentfill}%
\pgfsetlinewidth{0.000000pt}%
\definecolor{currentstroke}{rgb}{0.000000,0.000000,0.000000}%
\pgfsetstrokecolor{currentstroke}%
\pgfsetstrokeopacity{0.000000}%
\pgfsetdash{}{0pt}%
\pgfpathmoveto{\pgfqpoint{0.756173in}{0.818095in}}%
\pgfpathlineto{\pgfqpoint{0.795679in}{0.818095in}}%
\pgfpathlineto{\pgfqpoint{0.795679in}{0.865385in}}%
\pgfpathlineto{\pgfqpoint{0.756173in}{0.865385in}}%
\pgfpathlineto{\pgfqpoint{0.756173in}{0.818095in}}%
\pgfpathclose%
\pgfusepath{fill}%
\end{pgfscope}%
\begin{pgfscope}%
\pgfpathrectangle{\pgfqpoint{0.450000in}{0.780000in}}{\pgfqpoint{0.800000in}{0.800000in}}%
\pgfusepath{clip}%
\pgfsetbuttcap%
\pgfsetmiterjoin%
\definecolor{currentfill}{rgb}{0.941176,0.454902,0.431373}%
\pgfsetfillcolor{currentfill}%
\pgfsetlinewidth{0.000000pt}%
\definecolor{currentstroke}{rgb}{0.000000,0.000000,0.000000}%
\pgfsetstrokecolor{currentstroke}%
\pgfsetstrokeopacity{0.000000}%
\pgfsetdash{}{0pt}%
\pgfpathmoveto{\pgfqpoint{0.795679in}{0.818095in}}%
\pgfpathlineto{\pgfqpoint{0.835185in}{0.818095in}}%
\pgfpathlineto{\pgfqpoint{0.835185in}{0.843584in}}%
\pgfpathlineto{\pgfqpoint{0.795679in}{0.843584in}}%
\pgfpathlineto{\pgfqpoint{0.795679in}{0.818095in}}%
\pgfpathclose%
\pgfusepath{fill}%
\end{pgfscope}%
\begin{pgfscope}%
\pgfpathrectangle{\pgfqpoint{0.450000in}{0.780000in}}{\pgfqpoint{0.800000in}{0.800000in}}%
\pgfusepath{clip}%
\pgfsetbuttcap%
\pgfsetmiterjoin%
\definecolor{currentfill}{rgb}{0.486275,0.113725,0.435294}%
\pgfsetfillcolor{currentfill}%
\pgfsetlinewidth{0.000000pt}%
\definecolor{currentstroke}{rgb}{0.000000,0.000000,0.000000}%
\pgfsetstrokecolor{currentstroke}%
\pgfsetstrokeopacity{0.000000}%
\pgfsetdash{}{0pt}%
\pgfpathmoveto{\pgfqpoint{0.864815in}{0.818095in}}%
\pgfpathlineto{\pgfqpoint{0.904321in}{0.818095in}}%
\pgfpathlineto{\pgfqpoint{0.904321in}{0.836464in}}%
\pgfpathlineto{\pgfqpoint{0.864815in}{0.836464in}}%
\pgfpathlineto{\pgfqpoint{0.864815in}{0.818095in}}%
\pgfpathclose%
\pgfusepath{fill}%
\end{pgfscope}%
\begin{pgfscope}%
\pgfpathrectangle{\pgfqpoint{0.450000in}{0.780000in}}{\pgfqpoint{0.800000in}{0.800000in}}%
\pgfusepath{clip}%
\pgfsetbuttcap%
\pgfsetmiterjoin%
\definecolor{currentfill}{rgb}{0.031373,0.564706,0.600000}%
\pgfsetfillcolor{currentfill}%
\pgfsetlinewidth{0.000000pt}%
\definecolor{currentstroke}{rgb}{0.000000,0.000000,0.000000}%
\pgfsetstrokecolor{currentstroke}%
\pgfsetstrokeopacity{0.000000}%
\pgfsetdash{}{0pt}%
\pgfpathmoveto{\pgfqpoint{0.904321in}{0.818095in}}%
\pgfpathlineto{\pgfqpoint{0.943827in}{0.818095in}}%
\pgfpathlineto{\pgfqpoint{0.943827in}{0.824694in}}%
\pgfpathlineto{\pgfqpoint{0.904321in}{0.824694in}}%
\pgfpathlineto{\pgfqpoint{0.904321in}{0.818095in}}%
\pgfpathclose%
\pgfusepath{fill}%
\end{pgfscope}%
\begin{pgfscope}%
\pgfpathrectangle{\pgfqpoint{0.450000in}{0.780000in}}{\pgfqpoint{0.800000in}{0.800000in}}%
\pgfusepath{clip}%
\pgfsetbuttcap%
\pgfsetmiterjoin%
\definecolor{currentfill}{rgb}{0.941176,0.454902,0.431373}%
\pgfsetfillcolor{currentfill}%
\pgfsetlinewidth{0.000000pt}%
\definecolor{currentstroke}{rgb}{0.000000,0.000000,0.000000}%
\pgfsetstrokecolor{currentstroke}%
\pgfsetstrokeopacity{0.000000}%
\pgfsetdash{}{0pt}%
\pgfpathmoveto{\pgfqpoint{0.943827in}{0.818095in}}%
\pgfpathlineto{\pgfqpoint{0.983333in}{0.818095in}}%
\pgfpathlineto{\pgfqpoint{0.983333in}{0.817541in}}%
\pgfpathlineto{\pgfqpoint{0.943827in}{0.817541in}}%
\pgfpathlineto{\pgfqpoint{0.943827in}{0.818095in}}%
\pgfpathclose%
\pgfusepath{fill}%
\end{pgfscope}%
\begin{pgfscope}%
\pgfpathrectangle{\pgfqpoint{0.450000in}{0.780000in}}{\pgfqpoint{0.800000in}{0.800000in}}%
\pgfusepath{clip}%
\pgfsetbuttcap%
\pgfsetmiterjoin%
\definecolor{currentfill}{rgb}{0.486275,0.113725,0.435294}%
\pgfsetfillcolor{currentfill}%
\pgfsetlinewidth{0.000000pt}%
\definecolor{currentstroke}{rgb}{0.000000,0.000000,0.000000}%
\pgfsetstrokecolor{currentstroke}%
\pgfsetstrokeopacity{0.000000}%
\pgfsetdash{}{0pt}%
\pgfpathmoveto{\pgfqpoint{1.101852in}{0.818095in}}%
\pgfpathlineto{\pgfqpoint{1.141358in}{0.818095in}}%
\pgfpathlineto{\pgfqpoint{1.141358in}{0.812897in}}%
\pgfpathlineto{\pgfqpoint{1.101852in}{0.812897in}}%
\pgfpathlineto{\pgfqpoint{1.101852in}{0.818095in}}%
\pgfpathclose%
\pgfusepath{fill}%
\end{pgfscope}%
\begin{pgfscope}%
\pgfpathrectangle{\pgfqpoint{0.450000in}{0.780000in}}{\pgfqpoint{0.800000in}{0.800000in}}%
\pgfusepath{clip}%
\pgfsetbuttcap%
\pgfsetmiterjoin%
\definecolor{currentfill}{rgb}{0.031373,0.564706,0.600000}%
\pgfsetfillcolor{currentfill}%
\pgfsetlinewidth{0.000000pt}%
\definecolor{currentstroke}{rgb}{0.000000,0.000000,0.000000}%
\pgfsetstrokecolor{currentstroke}%
\pgfsetstrokeopacity{0.000000}%
\pgfsetdash{}{0pt}%
\pgfpathmoveto{\pgfqpoint{1.141358in}{0.818095in}}%
\pgfpathlineto{\pgfqpoint{1.180864in}{0.818095in}}%
\pgfpathlineto{\pgfqpoint{1.180864in}{0.833739in}}%
\pgfpathlineto{\pgfqpoint{1.141358in}{0.833739in}}%
\pgfpathlineto{\pgfqpoint{1.141358in}{0.818095in}}%
\pgfpathclose%
\pgfusepath{fill}%
\end{pgfscope}%
\begin{pgfscope}%
\pgfpathrectangle{\pgfqpoint{0.450000in}{0.780000in}}{\pgfqpoint{0.800000in}{0.800000in}}%
\pgfusepath{clip}%
\pgfsetbuttcap%
\pgfsetmiterjoin%
\definecolor{currentfill}{rgb}{0.941176,0.454902,0.431373}%
\pgfsetfillcolor{currentfill}%
\pgfsetlinewidth{0.000000pt}%
\definecolor{currentstroke}{rgb}{0.000000,0.000000,0.000000}%
\pgfsetstrokecolor{currentstroke}%
\pgfsetstrokeopacity{0.000000}%
\pgfsetdash{}{0pt}%
\pgfpathmoveto{\pgfqpoint{1.180864in}{0.818095in}}%
\pgfpathlineto{\pgfqpoint{1.220370in}{0.818095in}}%
\pgfpathlineto{\pgfqpoint{1.220370in}{0.826118in}}%
\pgfpathlineto{\pgfqpoint{1.180864in}{0.826118in}}%
\pgfpathlineto{\pgfqpoint{1.180864in}{0.818095in}}%
\pgfpathclose%
\pgfusepath{fill}%
\end{pgfscope}%
\begin{pgfscope}%
\pgfsetrectcap%
\pgfsetmiterjoin%
\pgfsetlinewidth{0.602250pt}%
\definecolor{currentstroke}{rgb}{0.321569,0.317647,0.305882}%
\pgfsetstrokecolor{currentstroke}%
\pgfsetdash{}{0pt}%
\pgfpathmoveto{\pgfqpoint{0.450000in}{0.780000in}}%
\pgfpathlineto{\pgfqpoint{0.450000in}{1.580000in}}%
\pgfusepath{stroke}%
\end{pgfscope}%
\begin{pgfscope}%
\pgfsetrectcap%
\pgfsetmiterjoin%
\pgfsetlinewidth{0.602250pt}%
\definecolor{currentstroke}{rgb}{0.321569,0.317647,0.305882}%
\pgfsetstrokecolor{currentstroke}%
\pgfsetdash{}{0pt}%
\pgfpathmoveto{\pgfqpoint{0.450000in}{0.780000in}}%
\pgfpathlineto{\pgfqpoint{1.250000in}{0.780000in}}%
\pgfusepath{stroke}%
\end{pgfscope}%
\begin{pgfscope}%
\pgfsetbuttcap%
\pgfsetmiterjoin%
\definecolor{currentfill}{rgb}{0.486275,0.113725,0.435294}%
\pgfsetfillcolor{currentfill}%
\pgfsetlinewidth{1.003750pt}%
\definecolor{currentstroke}{rgb}{0.486275,0.113725,0.435294}%
\pgfsetstrokecolor{currentstroke}%
\pgfsetdash{}{0pt}%
\pgfpathmoveto{\pgfqpoint{-0.024454in}{0.202222in}}%
\pgfpathlineto{\pgfqpoint{0.097768in}{0.202222in}}%
\pgfpathlineto{\pgfqpoint{0.097768in}{0.280000in}}%
\pgfpathlineto{\pgfqpoint{-0.024454in}{0.280000in}}%
\pgfpathlineto{\pgfqpoint{-0.024454in}{0.202222in}}%
\pgfpathclose%
\pgfusepath{stroke,fill}%
\end{pgfscope}%
\begin{pgfscope}%
\definecolor{textcolor}{rgb}{0.101961,0.101961,0.098039}%
\pgfsetstrokecolor{textcolor}%
\pgfsetfillcolor{textcolor}%
\pgftext[x=0.131101in,y=0.202222in,left,base]{\color{textcolor}\rmfamily\fontsize{8.000000}{9.600000}\selectfont \(\displaystyle \langle\)H,C\(\displaystyle \rangle\)}%
\end{pgfscope}%
\begin{pgfscope}%
\pgfsetbuttcap%
\pgfsetmiterjoin%
\definecolor{currentfill}{rgb}{0.031373,0.564706,0.600000}%
\pgfsetfillcolor{currentfill}%
\pgfsetlinewidth{1.003750pt}%
\definecolor{currentstroke}{rgb}{0.031373,0.564706,0.600000}%
\pgfsetstrokecolor{currentstroke}%
\pgfsetdash{}{0pt}%
\pgfpathmoveto{\pgfqpoint{0.493922in}{0.202222in}}%
\pgfpathlineto{\pgfqpoint{0.616144in}{0.202222in}}%
\pgfpathlineto{\pgfqpoint{0.616144in}{0.280000in}}%
\pgfpathlineto{\pgfqpoint{0.493922in}{0.280000in}}%
\pgfpathlineto{\pgfqpoint{0.493922in}{0.202222in}}%
\pgfpathclose%
\pgfusepath{stroke,fill}%
\end{pgfscope}%
\begin{pgfscope}%
\definecolor{textcolor}{rgb}{0.101961,0.101961,0.098039}%
\pgfsetstrokecolor{textcolor}%
\pgfsetfillcolor{textcolor}%
\pgftext[x=0.649478in,y=0.202222in,left,base]{\color{textcolor}\rmfamily\fontsize{8.000000}{9.600000}\selectfont \(\displaystyle \langle\)T,C\(\displaystyle \rangle\)}%
\end{pgfscope}%
\begin{pgfscope}%
\pgfsetbuttcap%
\pgfsetmiterjoin%
\definecolor{currentfill}{rgb}{0.941176,0.454902,0.431373}%
\pgfsetfillcolor{currentfill}%
\pgfsetlinewidth{1.003750pt}%
\definecolor{currentstroke}{rgb}{0.941176,0.454902,0.431373}%
\pgfsetstrokecolor{currentstroke}%
\pgfsetdash{}{0pt}%
\pgfpathmoveto{\pgfqpoint{0.991743in}{0.202222in}}%
\pgfpathlineto{\pgfqpoint{1.113966in}{0.202222in}}%
\pgfpathlineto{\pgfqpoint{1.113966in}{0.280000in}}%
\pgfpathlineto{\pgfqpoint{0.991743in}{0.280000in}}%
\pgfpathlineto{\pgfqpoint{0.991743in}{0.202222in}}%
\pgfpathclose%
\pgfusepath{stroke,fill}%
\end{pgfscope}%
\begin{pgfscope}%
\definecolor{textcolor}{rgb}{0.101961,0.101961,0.098039}%
\pgfsetstrokecolor{textcolor}%
\pgfsetfillcolor{textcolor}%
\pgftext[x=1.147299in,y=0.202222in,left,base]{\color{textcolor}\rmfamily\fontsize{8.000000}{9.600000}\selectfont \(\displaystyle \langle\)T,T\(\displaystyle \rangle\)}%
\end{pgfscope}%
\end{pgfpicture}%
\makeatother%
\endgroup%

%% file: figures/pgf/fig_abs_structure.pgf
\providecommand{\mathdefault}[1]{#1}%
\begingroup%
\makeatletter%
\begin{pgfpicture}%
\pgfpathrectangle{\pgfpointorigin}{\pgfqpoint{1.370000in}{1.480000in}}%
\pgfusepath{use as bounding box, clip}%
\begin{pgfscope}%
\pgfsetbuttcap%
\pgfsetmiterjoin%
\definecolor{currentfill}{rgb}{1.000000,1.000000,1.000000}%
\pgfsetfillcolor{currentfill}%
\pgfsetlinewidth{0.000000pt}%
\definecolor{currentstroke}{rgb}{1.000000,1.000000,1.000000}%
\pgfsetstrokecolor{currentstroke}%
\pgfsetdash{}{0pt}%
\pgfpathmoveto{\pgfqpoint{0.000000in}{0.000000in}}%
\pgfpathlineto{\pgfqpoint{1.370000in}{0.000000in}}%
\pgfpathlineto{\pgfqpoint{1.370000in}{1.480000in}}%
\pgfpathlineto{\pgfqpoint{0.000000in}{1.480000in}}%
\pgfpathlineto{\pgfqpoint{0.000000in}{0.000000in}}%
\pgfpathclose%
\pgfusepath{fill}%
\end{pgfscope}%
\begin{pgfscope}%
\pgfsetbuttcap%
\pgfsetmiterjoin%
\definecolor{currentfill}{rgb}{1.000000,1.000000,1.000000}%
\pgfsetfillcolor{currentfill}%
\pgfsetlinewidth{0.000000pt}%
\definecolor{currentstroke}{rgb}{0.000000,0.000000,0.000000}%
\pgfsetstrokecolor{currentstroke}%
\pgfsetstrokeopacity{0.000000}%
\pgfsetdash{}{0pt}%
\pgfpathmoveto{\pgfqpoint{0.450000in}{0.620000in}}%
\pgfpathlineto{\pgfqpoint{1.250000in}{0.620000in}}%
\pgfpathlineto{\pgfqpoint{1.250000in}{1.420000in}}%
\pgfpathlineto{\pgfqpoint{0.450000in}{1.420000in}}%
\pgfpathlineto{\pgfqpoint{0.450000in}{0.620000in}}%
\pgfpathclose%
\pgfusepath{fill}%
\end{pgfscope}%
\begin{pgfscope}%
\pgfpathrectangle{\pgfqpoint{0.450000in}{0.620000in}}{\pgfqpoint{0.800000in}{0.800000in}}%
\pgfusepath{clip}%
\pgfsetrectcap%
\pgfsetroundjoin%
\pgfsetlinewidth{0.501875pt}%
\definecolor{currentstroke}{rgb}{0.886275,0.882353,0.866667}%
\pgfsetstrokecolor{currentstroke}%
\pgfsetdash{}{0pt}%
\pgfpathmoveto{\pgfqpoint{0.450000in}{0.620000in}}%
\pgfpathlineto{\pgfqpoint{1.250000in}{0.620000in}}%
\pgfusepath{stroke}%
\end{pgfscope}%
\begin{pgfscope}%
\pgfsetbuttcap%
\pgfsetroundjoin%
\definecolor{currentfill}{rgb}{0.321569,0.317647,0.305882}%
\pgfsetfillcolor{currentfill}%
\pgfsetlinewidth{0.602250pt}%
\definecolor{currentstroke}{rgb}{0.321569,0.317647,0.305882}%
\pgfsetstrokecolor{currentstroke}%
\pgfsetdash{}{0pt}%
\pgfsys@defobject{currentmarker}{\pgfqpoint{-0.034722in}{0.000000in}}{\pgfqpoint{-0.000000in}{0.000000in}}{%
\pgfpathmoveto{\pgfqpoint{-0.000000in}{0.000000in}}%
\pgfpathlineto{\pgfqpoint{-0.034722in}{0.000000in}}%
\pgfusepath{stroke,fill}%
}%
\begin{pgfscope}%
\pgfsys@transformshift{0.450000in}{0.620000in}%
\pgfsys@useobject{currentmarker}{}%
\end{pgfscope}%
\end{pgfscope}%
\begin{pgfscope}%
\definecolor{textcolor}{rgb}{0.101961,0.101961,0.098039}%
\pgfsetstrokecolor{textcolor}%
\pgfsetfillcolor{textcolor}%
\pgftext[x=0.311111in, y=0.582112in, left, base]{\color{textcolor}\rmfamily\fontsize{8.000000}{9.600000}\selectfont 0}%
\end{pgfscope}%
\begin{pgfscope}%
\pgfpathrectangle{\pgfqpoint{0.450000in}{0.620000in}}{\pgfqpoint{0.800000in}{0.800000in}}%
\pgfusepath{clip}%
\pgfsetrectcap%
\pgfsetroundjoin%
\pgfsetlinewidth{0.501875pt}%
\definecolor{currentstroke}{rgb}{0.886275,0.882353,0.866667}%
\pgfsetstrokecolor{currentstroke}%
\pgfsetdash{}{0pt}%
\pgfpathmoveto{\pgfqpoint{0.450000in}{1.020000in}}%
\pgfpathlineto{\pgfqpoint{1.250000in}{1.020000in}}%
\pgfusepath{stroke}%
\end{pgfscope}%
\begin{pgfscope}%
\pgfsetbuttcap%
\pgfsetroundjoin%
\definecolor{currentfill}{rgb}{0.321569,0.317647,0.305882}%
\pgfsetfillcolor{currentfill}%
\pgfsetlinewidth{0.602250pt}%
\definecolor{currentstroke}{rgb}{0.321569,0.317647,0.305882}%
\pgfsetstrokecolor{currentstroke}%
\pgfsetdash{}{0pt}%
\pgfsys@defobject{currentmarker}{\pgfqpoint{-0.034722in}{0.000000in}}{\pgfqpoint{-0.000000in}{0.000000in}}{%
\pgfpathmoveto{\pgfqpoint{-0.000000in}{0.000000in}}%
\pgfpathlineto{\pgfqpoint{-0.034722in}{0.000000in}}%
\pgfusepath{stroke,fill}%
}%
\begin{pgfscope}%
\pgfsys@transformshift{0.450000in}{1.020000in}%
\pgfsys@useobject{currentmarker}{}%
\end{pgfscope}%
\end{pgfscope}%
\begin{pgfscope}%
\definecolor{textcolor}{rgb}{0.101961,0.101961,0.098039}%
\pgfsetstrokecolor{textcolor}%
\pgfsetfillcolor{textcolor}%
\pgftext[x=0.255556in, y=0.982112in, left, base]{\color{textcolor}\rmfamily\fontsize{8.000000}{9.600000}\selectfont 30}%
\end{pgfscope}%
\begin{pgfscope}%
\pgfpathrectangle{\pgfqpoint{0.450000in}{0.620000in}}{\pgfqpoint{0.800000in}{0.800000in}}%
\pgfusepath{clip}%
\pgfsetrectcap%
\pgfsetroundjoin%
\pgfsetlinewidth{0.501875pt}%
\definecolor{currentstroke}{rgb}{0.886275,0.882353,0.866667}%
\pgfsetstrokecolor{currentstroke}%
\pgfsetdash{}{0pt}%
\pgfpathmoveto{\pgfqpoint{0.450000in}{1.420000in}}%
\pgfpathlineto{\pgfqpoint{1.250000in}{1.420000in}}%
\pgfusepath{stroke}%
\end{pgfscope}%
\begin{pgfscope}%
\pgfsetbuttcap%
\pgfsetroundjoin%
\definecolor{currentfill}{rgb}{0.321569,0.317647,0.305882}%
\pgfsetfillcolor{currentfill}%
\pgfsetlinewidth{0.602250pt}%
\definecolor{currentstroke}{rgb}{0.321569,0.317647,0.305882}%
\pgfsetstrokecolor{currentstroke}%
\pgfsetdash{}{0pt}%
\pgfsys@defobject{currentmarker}{\pgfqpoint{-0.034722in}{0.000000in}}{\pgfqpoint{-0.000000in}{0.000000in}}{%
\pgfpathmoveto{\pgfqpoint{-0.000000in}{0.000000in}}%
\pgfpathlineto{\pgfqpoint{-0.034722in}{0.000000in}}%
\pgfusepath{stroke,fill}%
}%
\begin{pgfscope}%
\pgfsys@transformshift{0.450000in}{1.420000in}%
\pgfsys@useobject{currentmarker}{}%
\end{pgfscope}%
\end{pgfscope}%
\begin{pgfscope}%
\definecolor{textcolor}{rgb}{0.101961,0.101961,0.098039}%
\pgfsetstrokecolor{textcolor}%
\pgfsetfillcolor{textcolor}%
\pgftext[x=0.255556in, y=1.382112in, left, base]{\color{textcolor}\rmfamily\fontsize{8.000000}{9.600000}\selectfont 60}%
\end{pgfscope}%
\begin{pgfscope}%
\definecolor{textcolor}{rgb}{0.101961,0.101961,0.098039}%
\pgfsetstrokecolor{textcolor}%
\pgfsetfillcolor{textcolor}%
\pgftext[x=0.200000in,y=1.020000in,,bottom,rotate=90.000000]{\color{textcolor}\rmfamily\fontsize{9.000000}{10.800000}\selectfont recovered (\%)}%
\end{pgfscope}%
\begin{pgfscope}%
\pgfsetbuttcap%
\pgfsetmiterjoin%
\definecolor{currentfill}{rgb}{0.901961,0.623529,0.000000}%
\pgfsetfillcolor{currentfill}%
\pgfsetlinewidth{0.000000pt}%
\definecolor{currentstroke}{rgb}{0.000000,0.000000,0.000000}%
\pgfsetstrokecolor{currentstroke}%
\pgfsetstrokeopacity{0.000000}%
\pgfsetdash{}{0pt}%
\pgfpathmoveto{\pgfqpoint{0.494324in}{0.620000in}}%
\pgfpathlineto{\pgfqpoint{0.535405in}{0.620000in}}%
\pgfpathlineto{\pgfqpoint{0.535405in}{0.632351in}}%
\pgfpathlineto{\pgfqpoint{0.494324in}{0.632351in}}%
\pgfpathlineto{\pgfqpoint{0.494324in}{0.620000in}}%
\pgfpathclose%
\pgfusepath{fill}%
\end{pgfscope}%
\begin{pgfscope}%
\pgfsetbuttcap%
\pgfsetmiterjoin%
\definecolor{currentfill}{rgb}{0.901961,0.623529,0.000000}%
\pgfsetfillcolor{currentfill}%
\pgfsetlinewidth{0.000000pt}%
\definecolor{currentstroke}{rgb}{0.000000,0.000000,0.000000}%
\pgfsetstrokecolor{currentstroke}%
\pgfsetstrokeopacity{0.000000}%
\pgfsetdash{}{0pt}%
\pgfpathmoveto{\pgfqpoint{0.602432in}{0.620000in}}%
\pgfpathlineto{\pgfqpoint{0.643514in}{0.620000in}}%
\pgfpathlineto{\pgfqpoint{0.643514in}{1.225479in}}%
\pgfpathlineto{\pgfqpoint{0.602432in}{1.225479in}}%
\pgfpathlineto{\pgfqpoint{0.602432in}{0.620000in}}%
\pgfpathclose%
\pgfusepath{fill}%
\end{pgfscope}%
\begin{pgfscope}%
\pgfsetbuttcap%
\pgfsetmiterjoin%
\definecolor{currentfill}{rgb}{0.901961,0.623529,0.000000}%
\pgfsetfillcolor{currentfill}%
\pgfsetlinewidth{0.000000pt}%
\definecolor{currentstroke}{rgb}{0.000000,0.000000,0.000000}%
\pgfsetstrokecolor{currentstroke}%
\pgfsetstrokeopacity{0.000000}%
\pgfsetdash{}{0pt}%
\pgfpathmoveto{\pgfqpoint{0.775405in}{0.620000in}}%
\pgfpathlineto{\pgfqpoint{0.816486in}{0.620000in}}%
\pgfpathlineto{\pgfqpoint{0.816486in}{0.914919in}}%
\pgfpathlineto{\pgfqpoint{0.775405in}{0.914919in}}%
\pgfpathlineto{\pgfqpoint{0.775405in}{0.620000in}}%
\pgfpathclose%
\pgfusepath{fill}%
\end{pgfscope}%
\begin{pgfscope}%
\pgfsetbuttcap%
\pgfsetmiterjoin%
\definecolor{currentfill}{rgb}{0.901961,0.623529,0.000000}%
\pgfsetfillcolor{currentfill}%
\pgfsetlinewidth{0.000000pt}%
\definecolor{currentstroke}{rgb}{0.000000,0.000000,0.000000}%
\pgfsetstrokecolor{currentstroke}%
\pgfsetstrokeopacity{0.000000}%
\pgfsetdash{}{0pt}%
\pgfpathmoveto{\pgfqpoint{0.883514in}{0.620000in}}%
\pgfpathlineto{\pgfqpoint{0.924595in}{0.620000in}}%
\pgfpathlineto{\pgfqpoint{0.924595in}{1.350900in}}%
\pgfpathlineto{\pgfqpoint{0.883514in}{1.350900in}}%
\pgfpathlineto{\pgfqpoint{0.883514in}{0.620000in}}%
\pgfpathclose%
\pgfusepath{fill}%
\end{pgfscope}%
\begin{pgfscope}%
\pgfsetbuttcap%
\pgfsetmiterjoin%
\definecolor{currentfill}{rgb}{0.901961,0.623529,0.000000}%
\pgfsetfillcolor{currentfill}%
\pgfsetlinewidth{0.000000pt}%
\definecolor{currentstroke}{rgb}{0.000000,0.000000,0.000000}%
\pgfsetstrokecolor{currentstroke}%
\pgfsetstrokeopacity{0.000000}%
\pgfsetdash{}{0pt}%
\pgfpathmoveto{\pgfqpoint{1.056486in}{0.620000in}}%
\pgfpathlineto{\pgfqpoint{1.097568in}{0.620000in}}%
\pgfpathlineto{\pgfqpoint{1.097568in}{0.720133in}}%
\pgfpathlineto{\pgfqpoint{1.056486in}{0.720133in}}%
\pgfpathlineto{\pgfqpoint{1.056486in}{0.620000in}}%
\pgfpathclose%
\pgfusepath{fill}%
\end{pgfscope}%
\begin{pgfscope}%
\pgfsetbuttcap%
\pgfsetmiterjoin%
\definecolor{currentfill}{rgb}{0.901961,0.623529,0.000000}%
\pgfsetfillcolor{currentfill}%
\pgfsetlinewidth{0.000000pt}%
\definecolor{currentstroke}{rgb}{0.000000,0.000000,0.000000}%
\pgfsetstrokecolor{currentstroke}%
\pgfsetstrokeopacity{0.000000}%
\pgfsetdash{}{0pt}%
\pgfpathmoveto{\pgfqpoint{1.164595in}{0.620000in}}%
\pgfpathlineto{\pgfqpoint{1.205676in}{0.620000in}}%
\pgfpathlineto{\pgfqpoint{1.205676in}{0.877691in}}%
\pgfpathlineto{\pgfqpoint{1.164595in}{0.877691in}}%
\pgfpathlineto{\pgfqpoint{1.164595in}{0.620000in}}%
\pgfpathclose%
\pgfusepath{fill}%
\end{pgfscope}%
\begin{pgfscope}%
\pgfsetrectcap%
\pgfsetmiterjoin%
\pgfsetlinewidth{0.602250pt}%
\definecolor{currentstroke}{rgb}{0.321569,0.317647,0.305882}%
\pgfsetstrokecolor{currentstroke}%
\pgfsetdash{}{0pt}%
\pgfpathmoveto{\pgfqpoint{0.450000in}{0.620000in}}%
\pgfpathlineto{\pgfqpoint{0.450000in}{1.420000in}}%
\pgfusepath{stroke}%
\end{pgfscope}%
\begin{pgfscope}%
\pgfsetrectcap%
\pgfsetmiterjoin%
\pgfsetlinewidth{0.602250pt}%
\definecolor{currentstroke}{rgb}{0.321569,0.317647,0.305882}%
\pgfsetstrokecolor{currentstroke}%
\pgfsetdash{}{0pt}%
\pgfpathmoveto{\pgfqpoint{0.450000in}{0.620000in}}%
\pgfpathlineto{\pgfqpoint{1.250000in}{0.620000in}}%
\pgfusepath{stroke}%
\end{pgfscope}%
\begin{pgfscope}%
\definecolor{textcolor}{rgb}{0.101961,0.101961,0.098039}%
\pgfsetstrokecolor{textcolor}%
\pgfsetfillcolor{textcolor}%
\pgftext[x=0.514865in,y=0.571389in,right,top,rotate=60.000000]{\color{textcolor}\rmfamily\fontsize{8.000000}{9.600000}\selectfont LA+\(\displaystyle \beta\)}%
\end{pgfscope}%
\begin{pgfscope}%
\definecolor{textcolor}{rgb}{0.101961,0.101961,0.098039}%
\pgfsetstrokecolor{textcolor}%
\pgfsetfillcolor{textcolor}%
\pgftext[x=0.622973in,y=0.571389in,right,top,rotate=60.000000]{\color{textcolor}\rmfamily\fontsize{8.000000}{9.600000}\selectfont DN}%
\end{pgfscope}%
\begin{pgfscope}%
\definecolor{textcolor}{rgb}{0.101961,0.101961,0.098039}%
\pgfsetstrokecolor{textcolor}%
\pgfsetfillcolor{textcolor}%
\pgftext[x=0.795946in,y=0.571389in,right,top,rotate=60.000000]{\color{textcolor}\rmfamily\fontsize{8.000000}{9.600000}\selectfont M2+\(\displaystyle \beta\)}%
\end{pgfscope}%
\begin{pgfscope}%
\definecolor{textcolor}{rgb}{0.101961,0.101961,0.098039}%
\pgfsetstrokecolor{textcolor}%
\pgfsetfillcolor{textcolor}%
\pgftext[x=0.904054in,y=0.571389in,right,top,rotate=60.000000]{\color{textcolor}\rmfamily\fontsize{8.000000}{9.600000}\selectfont GDN}%
\end{pgfscope}%
\begin{pgfscope}%
\definecolor{textcolor}{rgb}{0.101961,0.101961,0.098039}%
\pgfsetstrokecolor{textcolor}%
\pgfsetfillcolor{textcolor}%
\pgftext[x=1.077027in,y=0.571389in,right,top,rotate=60.000000]{\color{textcolor}\rmfamily\fontsize{8.000000}{9.600000}\selectfont GLA+\(\displaystyle \beta\)}%
\end{pgfscope}%
\begin{pgfscope}%
\definecolor{textcolor}{rgb}{0.101961,0.101961,0.098039}%
\pgfsetstrokecolor{textcolor}%
\pgfsetfillcolor{textcolor}%
\pgftext[x=1.185135in,y=0.571389in,right,top,rotate=60.000000]{\color{textcolor}\rmfamily\fontsize{8.000000}{9.600000}\selectfont KDA}%
\end{pgfscope}%
\begin{pgfscope}%
\pgfsetbuttcap%
\pgfsetmiterjoin%
\definecolor{currentfill}{rgb}{0.901961,0.623529,0.000000}%
\pgfsetfillcolor{currentfill}%
\pgfsetlinewidth{1.003750pt}%
\definecolor{currentstroke}{rgb}{0.901961,0.623529,0.000000}%
\pgfsetstrokecolor{currentstroke}%
\pgfsetdash{}{0pt}%
\pgfpathmoveto{\pgfqpoint{0.507560in}{0.013210in}}%
\pgfpathlineto{\pgfqpoint{0.629782in}{0.013210in}}%
\pgfpathlineto{\pgfqpoint{0.629782in}{0.090987in}}%
\pgfpathlineto{\pgfqpoint{0.507560in}{0.090987in}}%
\pgfpathlineto{\pgfqpoint{0.507560in}{0.013210in}}%
\pgfpathclose%
\pgfusepath{stroke,fill}%
\end{pgfscope}%
\begin{pgfscope}%
\definecolor{textcolor}{rgb}{0.101961,0.101961,0.098039}%
\pgfsetstrokecolor{textcolor}%
\pgfsetfillcolor{textcolor}%
\pgftext[x=0.663115in,y=0.013210in,left,base]{\color{textcolor}\rmfamily\fontsize{8.000000}{9.600000}\selectfont \(\displaystyle \WriteGateTerm{\mat B_{[i]}^\xi}\)}%
\end{pgfscope}%
\end{pgfpicture}%
\makeatother%
\endgroup%

%% file: figures/pgf/fig_decay_side.pgf
\providecommand{\mathdefault}[1]{#1}%
\begingroup%
\makeatletter%
\begin{pgfpicture}%
\pgfpathrectangle{\pgfpointorigin}{\pgfqpoint{1.370000in}{1.480000in}}%
\pgfusepath{use as bounding box, clip}%
\begin{pgfscope}%
\pgfsetbuttcap%
\pgfsetmiterjoin%
\definecolor{currentfill}{rgb}{1.000000,1.000000,1.000000}%
\pgfsetfillcolor{currentfill}%
\pgfsetlinewidth{0.000000pt}%
\definecolor{currentstroke}{rgb}{1.000000,1.000000,1.000000}%
\pgfsetstrokecolor{currentstroke}%
\pgfsetdash{}{0pt}%
\pgfpathmoveto{\pgfqpoint{0.000000in}{0.000000in}}%
\pgfpathlineto{\pgfqpoint{1.370000in}{0.000000in}}%
\pgfpathlineto{\pgfqpoint{1.370000in}{1.480000in}}%
\pgfpathlineto{\pgfqpoint{0.000000in}{1.480000in}}%
\pgfpathlineto{\pgfqpoint{0.000000in}{0.000000in}}%
\pgfpathclose%
\pgfusepath{fill}%
\end{pgfscope}%
\begin{pgfscope}%
\pgfsetbuttcap%
\pgfsetmiterjoin%
\definecolor{currentfill}{rgb}{1.000000,1.000000,1.000000}%
\pgfsetfillcolor{currentfill}%
\pgfsetlinewidth{0.000000pt}%
\definecolor{currentstroke}{rgb}{0.000000,0.000000,0.000000}%
\pgfsetstrokecolor{currentstroke}%
\pgfsetstrokeopacity{0.000000}%
\pgfsetdash{}{0pt}%
\pgfpathmoveto{\pgfqpoint{0.450000in}{0.620000in}}%
\pgfpathlineto{\pgfqpoint{1.250000in}{0.620000in}}%
\pgfpathlineto{\pgfqpoint{1.250000in}{1.420000in}}%
\pgfpathlineto{\pgfqpoint{0.450000in}{1.420000in}}%
\pgfpathlineto{\pgfqpoint{0.450000in}{0.620000in}}%
\pgfpathclose%
\pgfusepath{fill}%
\end{pgfscope}%
\begin{pgfscope}%
\definecolor{textcolor}{rgb}{0.101961,0.101961,0.098039}%
\pgfsetstrokecolor{textcolor}%
\pgfsetfillcolor{textcolor}%
\pgftext[x=0.550000in,y=0.571389in,right,top,rotate=60.000000]{\color{textcolor}\rmfamily\fontsize{8.000000}{9.600000}\selectfont M2+\(\displaystyle \beta\)}%
\end{pgfscope}%
\begin{pgfscope}%
\definecolor{textcolor}{rgb}{0.101961,0.101961,0.098039}%
\pgfsetstrokecolor{textcolor}%
\pgfsetfillcolor{textcolor}%
\pgftext[x=0.716667in,y=0.571389in,right,top,rotate=60.000000]{\color{textcolor}\rmfamily\fontsize{8.000000}{9.600000}\selectfont GDN}%
\end{pgfscope}%
\begin{pgfscope}%
\definecolor{textcolor}{rgb}{0.101961,0.101961,0.098039}%
\pgfsetstrokecolor{textcolor}%
\pgfsetfillcolor{textcolor}%
\pgftext[x=0.983333in,y=0.571389in,right,top,rotate=60.000000]{\color{textcolor}\rmfamily\fontsize{8.000000}{9.600000}\selectfont GLA+\(\displaystyle \beta\)}%
\end{pgfscope}%
\begin{pgfscope}%
\definecolor{textcolor}{rgb}{0.101961,0.101961,0.098039}%
\pgfsetstrokecolor{textcolor}%
\pgfsetfillcolor{textcolor}%
\pgftext[x=1.150000in,y=0.571389in,right,top,rotate=60.000000]{\color{textcolor}\rmfamily\fontsize{8.000000}{9.600000}\selectfont KDA}%
\end{pgfscope}%
\begin{pgfscope}%
\pgfpathrectangle{\pgfqpoint{0.450000in}{0.620000in}}{\pgfqpoint{0.800000in}{0.800000in}}%
\pgfusepath{clip}%
\pgfsetrectcap%
\pgfsetroundjoin%
\pgfsetlinewidth{0.501875pt}%
\definecolor{currentstroke}{rgb}{0.886275,0.882353,0.866667}%
\pgfsetstrokecolor{currentstroke}%
\pgfsetdash{}{0pt}%
\pgfpathmoveto{\pgfqpoint{0.450000in}{0.620000in}}%
\pgfpathlineto{\pgfqpoint{1.250000in}{0.620000in}}%
\pgfusepath{stroke}%
\end{pgfscope}%
\begin{pgfscope}%
\pgfsetbuttcap%
\pgfsetroundjoin%
\definecolor{currentfill}{rgb}{0.321569,0.317647,0.305882}%
\pgfsetfillcolor{currentfill}%
\pgfsetlinewidth{0.602250pt}%
\definecolor{currentstroke}{rgb}{0.321569,0.317647,0.305882}%
\pgfsetstrokecolor{currentstroke}%
\pgfsetdash{}{0pt}%
\pgfsys@defobject{currentmarker}{\pgfqpoint{-0.034722in}{0.000000in}}{\pgfqpoint{-0.000000in}{0.000000in}}{%
\pgfpathmoveto{\pgfqpoint{-0.000000in}{0.000000in}}%
\pgfpathlineto{\pgfqpoint{-0.034722in}{0.000000in}}%
\pgfusepath{stroke,fill}%
}%
\begin{pgfscope}%
\pgfsys@transformshift{0.450000in}{0.620000in}%
\pgfsys@useobject{currentmarker}{}%
\end{pgfscope}%
\end{pgfscope}%
\begin{pgfscope}%
\definecolor{textcolor}{rgb}{0.101961,0.101961,0.098039}%
\pgfsetstrokecolor{textcolor}%
\pgfsetfillcolor{textcolor}%
\pgftext[x=0.311111in, y=0.582112in, left, base]{\color{textcolor}\rmfamily\fontsize{8.000000}{9.600000}\selectfont 0}%
\end{pgfscope}%
\begin{pgfscope}%
\pgfpathrectangle{\pgfqpoint{0.450000in}{0.620000in}}{\pgfqpoint{0.800000in}{0.800000in}}%
\pgfusepath{clip}%
\pgfsetrectcap%
\pgfsetroundjoin%
\pgfsetlinewidth{0.501875pt}%
\definecolor{currentstroke}{rgb}{0.886275,0.882353,0.866667}%
\pgfsetstrokecolor{currentstroke}%
\pgfsetdash{}{0pt}%
\pgfpathmoveto{\pgfqpoint{0.450000in}{1.020000in}}%
\pgfpathlineto{\pgfqpoint{1.250000in}{1.020000in}}%
\pgfusepath{stroke}%
\end{pgfscope}%
\begin{pgfscope}%
\pgfsetbuttcap%
\pgfsetroundjoin%
\definecolor{currentfill}{rgb}{0.321569,0.317647,0.305882}%
\pgfsetfillcolor{currentfill}%
\pgfsetlinewidth{0.602250pt}%
\definecolor{currentstroke}{rgb}{0.321569,0.317647,0.305882}%
\pgfsetstrokecolor{currentstroke}%
\pgfsetdash{}{0pt}%
\pgfsys@defobject{currentmarker}{\pgfqpoint{-0.034722in}{0.000000in}}{\pgfqpoint{-0.000000in}{0.000000in}}{%
\pgfpathmoveto{\pgfqpoint{-0.000000in}{0.000000in}}%
\pgfpathlineto{\pgfqpoint{-0.034722in}{0.000000in}}%
\pgfusepath{stroke,fill}%
}%
\begin{pgfscope}%
\pgfsys@transformshift{0.450000in}{1.020000in}%
\pgfsys@useobject{currentmarker}{}%
\end{pgfscope}%
\end{pgfscope}%
\begin{pgfscope}%
\definecolor{textcolor}{rgb}{0.101961,0.101961,0.098039}%
\pgfsetstrokecolor{textcolor}%
\pgfsetfillcolor{textcolor}%
\pgftext[x=0.255556in, y=0.982112in, left, base]{\color{textcolor}\rmfamily\fontsize{8.000000}{9.600000}\selectfont 15}%
\end{pgfscope}%
\begin{pgfscope}%
\pgfpathrectangle{\pgfqpoint{0.450000in}{0.620000in}}{\pgfqpoint{0.800000in}{0.800000in}}%
\pgfusepath{clip}%
\pgfsetrectcap%
\pgfsetroundjoin%
\pgfsetlinewidth{0.501875pt}%
\definecolor{currentstroke}{rgb}{0.886275,0.882353,0.866667}%
\pgfsetstrokecolor{currentstroke}%
\pgfsetdash{}{0pt}%
\pgfpathmoveto{\pgfqpoint{0.450000in}{1.420000in}}%
\pgfpathlineto{\pgfqpoint{1.250000in}{1.420000in}}%
\pgfusepath{stroke}%
\end{pgfscope}%
\begin{pgfscope}%
\pgfsetbuttcap%
\pgfsetroundjoin%
\definecolor{currentfill}{rgb}{0.321569,0.317647,0.305882}%
\pgfsetfillcolor{currentfill}%
\pgfsetlinewidth{0.602250pt}%
\definecolor{currentstroke}{rgb}{0.321569,0.317647,0.305882}%
\pgfsetstrokecolor{currentstroke}%
\pgfsetdash{}{0pt}%
\pgfsys@defobject{currentmarker}{\pgfqpoint{-0.034722in}{0.000000in}}{\pgfqpoint{-0.000000in}{0.000000in}}{%
\pgfpathmoveto{\pgfqpoint{-0.000000in}{0.000000in}}%
\pgfpathlineto{\pgfqpoint{-0.034722in}{0.000000in}}%
\pgfusepath{stroke,fill}%
}%
\begin{pgfscope}%
\pgfsys@transformshift{0.450000in}{1.420000in}%
\pgfsys@useobject{currentmarker}{}%
\end{pgfscope}%
\end{pgfscope}%
\begin{pgfscope}%
\definecolor{textcolor}{rgb}{0.101961,0.101961,0.098039}%
\pgfsetstrokecolor{textcolor}%
\pgfsetfillcolor{textcolor}%
\pgftext[x=0.255556in, y=1.382112in, left, base]{\color{textcolor}\rmfamily\fontsize{8.000000}{9.600000}\selectfont 30}%
\end{pgfscope}%
\begin{pgfscope}%
\definecolor{textcolor}{rgb}{0.101961,0.101961,0.098039}%
\pgfsetstrokecolor{textcolor}%
\pgfsetfillcolor{textcolor}%
\pgftext[x=0.200000in,y=1.020000in,,bottom,rotate=90.000000]{\color{textcolor}\rmfamily\fontsize{9.000000}{10.800000}\selectfont key vs.\ value (\%)}%
\end{pgfscope}%
\begin{pgfscope}%
\pgfpathrectangle{\pgfqpoint{0.450000in}{0.620000in}}{\pgfqpoint{0.800000in}{0.800000in}}%
\pgfusepath{clip}%
\pgfsetbuttcap%
\pgfsetmiterjoin%
\definecolor{currentfill}{rgb}{0.337255,0.705882,0.913725}%
\pgfsetfillcolor{currentfill}%
\pgfsetlinewidth{0.000000pt}%
\definecolor{currentstroke}{rgb}{0.000000,0.000000,0.000000}%
\pgfsetstrokecolor{currentstroke}%
\pgfsetstrokeopacity{0.000000}%
\pgfsetdash{}{0pt}%
\pgfpathmoveto{\pgfqpoint{0.500000in}{0.620000in}}%
\pgfpathlineto{\pgfqpoint{0.600000in}{0.620000in}}%
\pgfpathlineto{\pgfqpoint{0.600000in}{1.346094in}}%
\pgfpathlineto{\pgfqpoint{0.500000in}{1.346094in}}%
\pgfpathlineto{\pgfqpoint{0.500000in}{0.620000in}}%
\pgfpathclose%
\pgfusepath{fill}%
\end{pgfscope}%
\begin{pgfscope}%
\pgfpathrectangle{\pgfqpoint{0.450000in}{0.620000in}}{\pgfqpoint{0.800000in}{0.800000in}}%
\pgfusepath{clip}%
\pgfsetbuttcap%
\pgfsetmiterjoin%
\definecolor{currentfill}{rgb}{0.337255,0.705882,0.913725}%
\pgfsetfillcolor{currentfill}%
\pgfsetlinewidth{0.000000pt}%
\definecolor{currentstroke}{rgb}{0.000000,0.000000,0.000000}%
\pgfsetstrokecolor{currentstroke}%
\pgfsetstrokeopacity{0.000000}%
\pgfsetdash{}{0pt}%
\pgfpathmoveto{\pgfqpoint{0.666667in}{0.620000in}}%
\pgfpathlineto{\pgfqpoint{0.766667in}{0.620000in}}%
\pgfpathlineto{\pgfqpoint{0.766667in}{1.183928in}}%
\pgfpathlineto{\pgfqpoint{0.666667in}{1.183928in}}%
\pgfpathlineto{\pgfqpoint{0.666667in}{0.620000in}}%
\pgfpathclose%
\pgfusepath{fill}%
\end{pgfscope}%
\begin{pgfscope}%
\pgfpathrectangle{\pgfqpoint{0.450000in}{0.620000in}}{\pgfqpoint{0.800000in}{0.800000in}}%
\pgfusepath{clip}%
\pgfsetbuttcap%
\pgfsetmiterjoin%
\definecolor{currentfill}{rgb}{0.000000,0.447059,0.698039}%
\pgfsetfillcolor{currentfill}%
\pgfsetlinewidth{0.000000pt}%
\definecolor{currentstroke}{rgb}{0.000000,0.000000,0.000000}%
\pgfsetstrokecolor{currentstroke}%
\pgfsetstrokeopacity{0.000000}%
\pgfsetdash{}{0pt}%
\pgfpathmoveto{\pgfqpoint{0.933333in}{0.620000in}}%
\pgfpathlineto{\pgfqpoint{1.033333in}{0.620000in}}%
\pgfpathlineto{\pgfqpoint{1.033333in}{0.649543in}}%
\pgfpathlineto{\pgfqpoint{0.933333in}{0.649543in}}%
\pgfpathlineto{\pgfqpoint{0.933333in}{0.620000in}}%
\pgfpathclose%
\pgfusepath{fill}%
\end{pgfscope}%
\begin{pgfscope}%
\pgfpathrectangle{\pgfqpoint{0.450000in}{0.620000in}}{\pgfqpoint{0.800000in}{0.800000in}}%
\pgfusepath{clip}%
\pgfsetbuttcap%
\pgfsetmiterjoin%
\definecolor{currentfill}{rgb}{0.000000,0.447059,0.698039}%
\pgfsetfillcolor{currentfill}%
\pgfsetlinewidth{0.000000pt}%
\definecolor{currentstroke}{rgb}{0.000000,0.000000,0.000000}%
\pgfsetstrokecolor{currentstroke}%
\pgfsetstrokeopacity{0.000000}%
\pgfsetdash{}{0pt}%
\pgfpathmoveto{\pgfqpoint{1.100000in}{0.620000in}}%
\pgfpathlineto{\pgfqpoint{1.200000in}{0.620000in}}%
\pgfpathlineto{\pgfqpoint{1.200000in}{0.651417in}}%
\pgfpathlineto{\pgfqpoint{1.100000in}{0.651417in}}%
\pgfpathlineto{\pgfqpoint{1.100000in}{0.620000in}}%
\pgfpathclose%
\pgfusepath{fill}%
\end{pgfscope}%
\begin{pgfscope}%
\pgfsetrectcap%
\pgfsetmiterjoin%
\pgfsetlinewidth{0.602250pt}%
\definecolor{currentstroke}{rgb}{0.321569,0.317647,0.305882}%
\pgfsetstrokecolor{currentstroke}%
\pgfsetdash{}{0pt}%
\pgfpathmoveto{\pgfqpoint{0.450000in}{0.620000in}}%
\pgfpathlineto{\pgfqpoint{0.450000in}{1.420000in}}%
\pgfusepath{stroke}%
\end{pgfscope}%
\begin{pgfscope}%
\pgfsetrectcap%
\pgfsetmiterjoin%
\pgfsetlinewidth{0.602250pt}%
\definecolor{currentstroke}{rgb}{0.321569,0.317647,0.305882}%
\pgfsetstrokecolor{currentstroke}%
\pgfsetdash{}{0pt}%
\pgfpathmoveto{\pgfqpoint{0.450000in}{0.620000in}}%
\pgfpathlineto{\pgfqpoint{1.250000in}{0.620000in}}%
\pgfusepath{stroke}%
\end{pgfscope}%
\begin{pgfscope}%
\pgfsetbuttcap%
\pgfsetmiterjoin%
\definecolor{currentfill}{rgb}{0.337255,0.705882,0.913725}%
\pgfsetfillcolor{currentfill}%
\pgfsetlinewidth{1.003750pt}%
\definecolor{currentstroke}{rgb}{0.337255,0.705882,0.913725}%
\pgfsetstrokecolor{currentstroke}%
\pgfsetdash{}{0pt}%
\pgfpathmoveto{\pgfqpoint{0.216115in}{0.035201in}}%
\pgfpathlineto{\pgfqpoint{0.338337in}{0.035201in}}%
\pgfpathlineto{\pgfqpoint{0.338337in}{0.112979in}}%
\pgfpathlineto{\pgfqpoint{0.216115in}{0.112979in}}%
\pgfpathlineto{\pgfqpoint{0.216115in}{0.035201in}}%
\pgfpathclose%
\pgfusepath{stroke,fill}%
\end{pgfscope}%
\begin{pgfscope}%
\definecolor{textcolor}{rgb}{0.101961,0.101961,0.098039}%
\pgfsetstrokecolor{textcolor}%
\pgfsetfillcolor{textcolor}%
\pgftext[x=0.371670in,y=0.035201in,left,base]{\color{textcolor}\rmfamily\fontsize{8.000000}{9.600000}\selectfont \(\displaystyle \DecayTerm{(\Dtail_{[i]})^\eta}\)}%
\end{pgfscope}%
\begin{pgfscope}%
\pgfsetbuttcap%
\pgfsetmiterjoin%
\definecolor{currentfill}{rgb}{0.000000,0.447059,0.698039}%
\pgfsetfillcolor{currentfill}%
\pgfsetlinewidth{1.003750pt}%
\definecolor{currentstroke}{rgb}{0.000000,0.447059,0.698039}%
\pgfsetstrokecolor{currentstroke}%
\pgfsetdash{}{0pt}%
\pgfpathmoveto{\pgfqpoint{0.806413in}{0.035201in}}%
\pgfpathlineto{\pgfqpoint{0.928635in}{0.035201in}}%
\pgfpathlineto{\pgfqpoint{0.928635in}{0.112979in}}%
\pgfpathlineto{\pgfqpoint{0.806413in}{0.112979in}}%
\pgfpathlineto{\pgfqpoint{0.806413in}{0.035201in}}%
\pgfpathclose%
\pgfusepath{stroke,fill}%
\end{pgfscope}%
\begin{pgfscope}%
\definecolor{textcolor}{rgb}{0.101961,0.101961,0.098039}%
\pgfsetstrokecolor{textcolor}%
\pgfsetfillcolor{textcolor}%
\pgftext[x=0.961968in,y=0.035201in,left,base]{\color{textcolor}\rmfamily\fontsize{8.000000}{9.600000}\selectfont \(\displaystyle \DecayTerm{\mat E_{[i]}^\eta}\)}%
\end{pgfscope}%
\end{pgfpicture}%
\makeatother%
\endgroup%

%% file: figures/pgf/fig_prefill.pgf
\providecommand{\mathdefault}[1]{#1}%
\begingroup%
\makeatletter%
\begin{pgfpicture}%
\pgfpathrectangle{\pgfpointorigin}{\pgfqpoint{1.290000in}{1.380000in}}%
\pgfusepath{use as bounding box, clip}%
\begin{pgfscope}%
\pgfsetbuttcap%
\pgfsetmiterjoin%
\definecolor{currentfill}{rgb}{1.000000,1.000000,1.000000}%
\pgfsetfillcolor{currentfill}%
\pgfsetlinewidth{0.000000pt}%
\definecolor{currentstroke}{rgb}{1.000000,1.000000,1.000000}%
\pgfsetstrokecolor{currentstroke}%
\pgfsetdash{}{0pt}%
\pgfpathmoveto{\pgfqpoint{0.000000in}{0.000000in}}%
\pgfpathlineto{\pgfqpoint{1.290000in}{0.000000in}}%
\pgfpathlineto{\pgfqpoint{1.290000in}{1.380000in}}%
\pgfpathlineto{\pgfqpoint{0.000000in}{1.380000in}}%
\pgfpathlineto{\pgfqpoint{0.000000in}{0.000000in}}%
\pgfpathclose%
\pgfusepath{fill}%
\end{pgfscope}%
\begin{pgfscope}%
\pgfsetbuttcap%
\pgfsetmiterjoin%
\definecolor{currentfill}{rgb}{1.000000,1.000000,1.000000}%
\pgfsetfillcolor{currentfill}%
\pgfsetlinewidth{0.000000pt}%
\definecolor{currentstroke}{rgb}{0.000000,0.000000,0.000000}%
\pgfsetstrokecolor{currentstroke}%
\pgfsetstrokeopacity{0.000000}%
\pgfsetdash{}{0pt}%
\pgfpathmoveto{\pgfqpoint{0.460000in}{0.520000in}}%
\pgfpathlineto{\pgfqpoint{1.260000in}{0.520000in}}%
\pgfpathlineto{\pgfqpoint{1.260000in}{1.320000in}}%
\pgfpathlineto{\pgfqpoint{0.460000in}{1.320000in}}%
\pgfpathlineto{\pgfqpoint{0.460000in}{0.520000in}}%
\pgfpathclose%
\pgfusepath{fill}%
\end{pgfscope}%
\begin{pgfscope}%
\pgfpathrectangle{\pgfqpoint{0.460000in}{0.520000in}}{\pgfqpoint{0.800000in}{0.800000in}}%
\pgfusepath{clip}%
\pgfsetbuttcap%
\pgfsetroundjoin%
\pgfsetlinewidth{0.602250pt}%
\definecolor{currentstroke}{rgb}{0.321569,0.317647,0.305882}%
\pgfsetstrokecolor{currentstroke}%
\pgfsetdash{{0.600000pt}{0.990000pt}}{0.000000pt}%
\pgfpathmoveto{\pgfqpoint{0.460000in}{0.920000in}}%
\pgfpathlineto{\pgfqpoint{1.260000in}{0.920000in}}%
\pgfusepath{stroke}%
\end{pgfscope}%
\begin{pgfscope}%
\pgfpathrectangle{\pgfqpoint{0.460000in}{0.520000in}}{\pgfqpoint{0.800000in}{0.800000in}}%
\pgfusepath{clip}%
\pgfsetrectcap%
\pgfsetroundjoin%
\pgfsetlinewidth{0.501875pt}%
\definecolor{currentstroke}{rgb}{0.886275,0.882353,0.866667}%
\pgfsetstrokecolor{currentstroke}%
\pgfsetdash{}{0pt}%
\pgfpathmoveto{\pgfqpoint{0.496364in}{0.520000in}}%
\pgfpathlineto{\pgfqpoint{0.496364in}{1.320000in}}%
\pgfusepath{stroke}%
\end{pgfscope}%
\begin{pgfscope}%
\pgfsetbuttcap%
\pgfsetroundjoin%
\definecolor{currentfill}{rgb}{0.321569,0.317647,0.305882}%
\pgfsetfillcolor{currentfill}%
\pgfsetlinewidth{0.602250pt}%
\definecolor{currentstroke}{rgb}{0.321569,0.317647,0.305882}%
\pgfsetstrokecolor{currentstroke}%
\pgfsetdash{}{0pt}%
\pgfsys@defobject{currentmarker}{\pgfqpoint{0.000000in}{-0.034722in}}{\pgfqpoint{0.000000in}{0.000000in}}{%
\pgfpathmoveto{\pgfqpoint{0.000000in}{0.000000in}}%
\pgfpathlineto{\pgfqpoint{0.000000in}{-0.034722in}}%
\pgfusepath{stroke,fill}%
}%
\begin{pgfscope}%
\pgfsys@transformshift{0.496364in}{0.520000in}%
\pgfsys@useobject{currentmarker}{}%
\end{pgfscope}%
\end{pgfscope}%
\begin{pgfscope}%
\definecolor{textcolor}{rgb}{0.101961,0.101961,0.098039}%
\pgfsetstrokecolor{textcolor}%
\pgfsetfillcolor{textcolor}%
\pgftext[x=0.496364in,y=0.436667in,,top]{\color{textcolor}{\rmfamily\fontsize{8.000000}{9.600000}\selectfont\catcode`\^=\active\def^{\ifmmode\sp\else\^{}\fi}\catcode`\%=\active\def
\end{pgfscope}%
\begin{pgfscope}%
\pgfpathrectangle{\pgfqpoint{0.460000in}{0.520000in}}{\pgfqpoint{0.800000in}{0.800000in}}%
\pgfusepath{clip}%
\pgfsetrectcap%
\pgfsetroundjoin%
\pgfsetlinewidth{0.501875pt}%
\definecolor{currentstroke}{rgb}{0.886275,0.882353,0.866667}%
\pgfsetstrokecolor{currentstroke}%
\pgfsetdash{}{0pt}%
\pgfpathmoveto{\pgfqpoint{0.787273in}{0.520000in}}%
\pgfpathlineto{\pgfqpoint{0.787273in}{1.320000in}}%
\pgfusepath{stroke}%
\end{pgfscope}%
\begin{pgfscope}%
\pgfsetbuttcap%
\pgfsetroundjoin%
\definecolor{currentfill}{rgb}{0.321569,0.317647,0.305882}%
\pgfsetfillcolor{currentfill}%
\pgfsetlinewidth{0.602250pt}%
\definecolor{currentstroke}{rgb}{0.321569,0.317647,0.305882}%
\pgfsetstrokecolor{currentstroke}%
\pgfsetdash{}{0pt}%
\pgfsys@defobject{currentmarker}{\pgfqpoint{0.000000in}{-0.034722in}}{\pgfqpoint{0.000000in}{0.000000in}}{%
\pgfpathmoveto{\pgfqpoint{0.000000in}{0.000000in}}%
\pgfpathlineto{\pgfqpoint{0.000000in}{-0.034722in}}%
\pgfusepath{stroke,fill}%
}%
\begin{pgfscope}%
\pgfsys@transformshift{0.787273in}{0.520000in}%
\pgfsys@useobject{currentmarker}{}%
\end{pgfscope}%
\end{pgfscope}%
\begin{pgfscope}%
\definecolor{textcolor}{rgb}{0.101961,0.101961,0.098039}%
\pgfsetstrokecolor{textcolor}%
\pgfsetfillcolor{textcolor}%
\pgftext[x=0.787273in,y=0.436667in,,top]{\color{textcolor}{\rmfamily\fontsize{8.000000}{9.600000}\selectfont\catcode`\^=\active\def^{\ifmmode\sp\else\^{}\fi}\catcode`\%=\active\def
\end{pgfscope}%
\begin{pgfscope}%
\pgfpathrectangle{\pgfqpoint{0.460000in}{0.520000in}}{\pgfqpoint{0.800000in}{0.800000in}}%
\pgfusepath{clip}%
\pgfsetrectcap%
\pgfsetroundjoin%
\pgfsetlinewidth{0.501875pt}%
\definecolor{currentstroke}{rgb}{0.886275,0.882353,0.866667}%
\pgfsetstrokecolor{currentstroke}%
\pgfsetdash{}{0pt}%
\pgfpathmoveto{\pgfqpoint{1.078182in}{0.520000in}}%
\pgfpathlineto{\pgfqpoint{1.078182in}{1.320000in}}%
\pgfusepath{stroke}%
\end{pgfscope}%
\begin{pgfscope}%
\pgfsetbuttcap%
\pgfsetroundjoin%
\definecolor{currentfill}{rgb}{0.321569,0.317647,0.305882}%
\pgfsetfillcolor{currentfill}%
\pgfsetlinewidth{0.602250pt}%
\definecolor{currentstroke}{rgb}{0.321569,0.317647,0.305882}%
\pgfsetstrokecolor{currentstroke}%
\pgfsetdash{}{0pt}%
\pgfsys@defobject{currentmarker}{\pgfqpoint{0.000000in}{-0.034722in}}{\pgfqpoint{0.000000in}{0.000000in}}{%
\pgfpathmoveto{\pgfqpoint{0.000000in}{0.000000in}}%
\pgfpathlineto{\pgfqpoint{0.000000in}{-0.034722in}}%
\pgfusepath{stroke,fill}%
}%
\begin{pgfscope}%
\pgfsys@transformshift{1.078182in}{0.520000in}%
\pgfsys@useobject{currentmarker}{}%
\end{pgfscope}%
\end{pgfscope}%
\begin{pgfscope}%
\definecolor{textcolor}{rgb}{0.101961,0.101961,0.098039}%
\pgfsetstrokecolor{textcolor}%
\pgfsetfillcolor{textcolor}%
\pgftext[x=1.078182in,y=0.436667in,,top]{\color{textcolor}{\rmfamily\fontsize{8.000000}{9.600000}\selectfont\catcode`\^=\active\def^{\ifmmode\sp\else\^{}\fi}\catcode`\%=\active\def
\end{pgfscope}%
\begin{pgfscope}%
\pgfpathrectangle{\pgfqpoint{0.460000in}{0.520000in}}{\pgfqpoint{0.800000in}{0.800000in}}%
\pgfusepath{clip}%
\pgfsetrectcap%
\pgfsetroundjoin%
\pgfsetlinewidth{0.501875pt}%
\definecolor{currentstroke}{rgb}{0.886275,0.882353,0.866667}%
\pgfsetstrokecolor{currentstroke}%
\pgfsetdash{}{0pt}%
\pgfpathmoveto{\pgfqpoint{1.223636in}{0.520000in}}%
\pgfpathlineto{\pgfqpoint{1.223636in}{1.320000in}}%
\pgfusepath{stroke}%
\end{pgfscope}%
\begin{pgfscope}%
\pgfsetbuttcap%
\pgfsetroundjoin%
\definecolor{currentfill}{rgb}{0.321569,0.317647,0.305882}%
\pgfsetfillcolor{currentfill}%
\pgfsetlinewidth{0.602250pt}%
\definecolor{currentstroke}{rgb}{0.321569,0.317647,0.305882}%
\pgfsetstrokecolor{currentstroke}%
\pgfsetdash{}{0pt}%
\pgfsys@defobject{currentmarker}{\pgfqpoint{0.000000in}{-0.034722in}}{\pgfqpoint{0.000000in}{0.000000in}}{%
\pgfpathmoveto{\pgfqpoint{0.000000in}{0.000000in}}%
\pgfpathlineto{\pgfqpoint{0.000000in}{-0.034722in}}%
\pgfusepath{stroke,fill}%
}%
\begin{pgfscope}%
\pgfsys@transformshift{1.223636in}{0.520000in}%
\pgfsys@useobject{currentmarker}{}%
\end{pgfscope}%
\end{pgfscope}%
\begin{pgfscope}%
\definecolor{textcolor}{rgb}{0.101961,0.101961,0.098039}%
\pgfsetstrokecolor{textcolor}%
\pgfsetfillcolor{textcolor}%
\pgftext[x=1.223636in,y=0.436667in,,top]{\color{textcolor}{\rmfamily\fontsize{8.000000}{9.600000}\selectfont\catcode`\^=\active\def^{\ifmmode\sp\else\^{}\fi}\catcode`\%=\active\def
\end{pgfscope}%
\begin{pgfscope}%
\definecolor{textcolor}{rgb}{0.101961,0.101961,0.098039}%
\pgfsetstrokecolor{textcolor}%
\pgfsetfillcolor{textcolor}%
\pgftext[x=0.860000in,y=0.309058in,,top]{\color{textcolor}{\rmfamily\fontsize{9.000000}{10.800000}\selectfont\catcode`\^=\active\def^{\ifmmode\sp\else\^{}\fi}\catcode`\%=\active\def
\end{pgfscope}%
\begin{pgfscope}%
\pgfpathrectangle{\pgfqpoint{0.460000in}{0.520000in}}{\pgfqpoint{0.800000in}{0.800000in}}%
\pgfusepath{clip}%
\pgfsetrectcap%
\pgfsetroundjoin%
\pgfsetlinewidth{0.501875pt}%
\definecolor{currentstroke}{rgb}{0.886275,0.882353,0.866667}%
\pgfsetstrokecolor{currentstroke}%
\pgfsetdash{}{0pt}%
\pgfpathmoveto{\pgfqpoint{0.460000in}{0.520000in}}%
\pgfpathlineto{\pgfqpoint{1.260000in}{0.520000in}}%
\pgfusepath{stroke}%
\end{pgfscope}%
\begin{pgfscope}%
\pgfsetbuttcap%
\pgfsetroundjoin%
\definecolor{currentfill}{rgb}{0.321569,0.317647,0.305882}%
\pgfsetfillcolor{currentfill}%
\pgfsetlinewidth{0.602250pt}%
\definecolor{currentstroke}{rgb}{0.321569,0.317647,0.305882}%
\pgfsetstrokecolor{currentstroke}%
\pgfsetdash{}{0pt}%
\pgfsys@defobject{currentmarker}{\pgfqpoint{-0.034722in}{0.000000in}}{\pgfqpoint{-0.000000in}{0.000000in}}{%
\pgfpathmoveto{\pgfqpoint{-0.000000in}{0.000000in}}%
\pgfpathlineto{\pgfqpoint{-0.034722in}{0.000000in}}%
\pgfusepath{stroke,fill}%
}%
\begin{pgfscope}%
\pgfsys@transformshift{0.460000in}{0.520000in}%
\pgfsys@useobject{currentmarker}{}%
\end{pgfscope}%
\end{pgfscope}%
\begin{pgfscope}%
\definecolor{textcolor}{rgb}{0.101961,0.101961,0.098039}%
\pgfsetstrokecolor{textcolor}%
\pgfsetfillcolor{textcolor}%
\pgftext[x=0.293333in, y=0.482112in, left, base]{\color{textcolor}{\rmfamily\fontsize{8.000000}{9.600000}\selectfont\catcode`\^=\active\def^{\ifmmode\sp\else\^{}\fi}\catcode`\%=\active\def
\end{pgfscope}%
\begin{pgfscope}%
\pgfpathrectangle{\pgfqpoint{0.460000in}{0.520000in}}{\pgfqpoint{0.800000in}{0.800000in}}%
\pgfusepath{clip}%
\pgfsetrectcap%
\pgfsetroundjoin%
\pgfsetlinewidth{0.501875pt}%
\definecolor{currentstroke}{rgb}{0.886275,0.882353,0.866667}%
\pgfsetstrokecolor{currentstroke}%
\pgfsetdash{}{0pt}%
\pgfpathmoveto{\pgfqpoint{0.460000in}{0.920000in}}%
\pgfpathlineto{\pgfqpoint{1.260000in}{0.920000in}}%
\pgfusepath{stroke}%
\end{pgfscope}%
\begin{pgfscope}%
\pgfsetbuttcap%
\pgfsetroundjoin%
\definecolor{currentfill}{rgb}{0.321569,0.317647,0.305882}%
\pgfsetfillcolor{currentfill}%
\pgfsetlinewidth{0.602250pt}%
\definecolor{currentstroke}{rgb}{0.321569,0.317647,0.305882}%
\pgfsetstrokecolor{currentstroke}%
\pgfsetdash{}{0pt}%
\pgfsys@defobject{currentmarker}{\pgfqpoint{-0.034722in}{0.000000in}}{\pgfqpoint{-0.000000in}{0.000000in}}{%
\pgfpathmoveto{\pgfqpoint{-0.000000in}{0.000000in}}%
\pgfpathlineto{\pgfqpoint{-0.034722in}{0.000000in}}%
\pgfusepath{stroke,fill}%
}%
\begin{pgfscope}%
\pgfsys@transformshift{0.460000in}{0.920000in}%
\pgfsys@useobject{currentmarker}{}%
\end{pgfscope}%
\end{pgfscope}%
\begin{pgfscope}%
\definecolor{textcolor}{rgb}{0.101961,0.101961,0.098039}%
\pgfsetstrokecolor{textcolor}%
\pgfsetfillcolor{textcolor}%
\pgftext[x=0.321111in, y=0.882112in, left, base]{\color{textcolor}{\rmfamily\fontsize{8.000000}{9.600000}\selectfont\catcode`\^=\active\def^{\ifmmode\sp\else\^{}\fi}\catcode`\%=\active\def
\end{pgfscope}%
\begin{pgfscope}%
\pgfpathrectangle{\pgfqpoint{0.460000in}{0.520000in}}{\pgfqpoint{0.800000in}{0.800000in}}%
\pgfusepath{clip}%
\pgfsetrectcap%
\pgfsetroundjoin%
\pgfsetlinewidth{0.501875pt}%
\definecolor{currentstroke}{rgb}{0.886275,0.882353,0.866667}%
\pgfsetstrokecolor{currentstroke}%
\pgfsetdash{}{0pt}%
\pgfpathmoveto{\pgfqpoint{0.460000in}{1.320000in}}%
\pgfpathlineto{\pgfqpoint{1.260000in}{1.320000in}}%
\pgfusepath{stroke}%
\end{pgfscope}%
\begin{pgfscope}%
\pgfsetbuttcap%
\pgfsetroundjoin%
\definecolor{currentfill}{rgb}{0.321569,0.317647,0.305882}%
\pgfsetfillcolor{currentfill}%
\pgfsetlinewidth{0.602250pt}%
\definecolor{currentstroke}{rgb}{0.321569,0.317647,0.305882}%
\pgfsetstrokecolor{currentstroke}%
\pgfsetdash{}{0pt}%
\pgfsys@defobject{currentmarker}{\pgfqpoint{-0.034722in}{0.000000in}}{\pgfqpoint{-0.000000in}{0.000000in}}{%
\pgfpathmoveto{\pgfqpoint{-0.000000in}{0.000000in}}%
\pgfpathlineto{\pgfqpoint{-0.034722in}{0.000000in}}%
\pgfusepath{stroke,fill}%
}%
\begin{pgfscope}%
\pgfsys@transformshift{0.460000in}{1.320000in}%
\pgfsys@useobject{currentmarker}{}%
\end{pgfscope}%
\end{pgfscope}%
\begin{pgfscope}%
\definecolor{textcolor}{rgb}{0.101961,0.101961,0.098039}%
\pgfsetstrokecolor{textcolor}%
\pgfsetfillcolor{textcolor}%
\pgftext[x=0.237778in, y=1.282112in, left, base]{\color{textcolor}{\rmfamily\fontsize{8.000000}{9.600000}\selectfont\catcode`\^=\active\def^{\ifmmode\sp\else\^{}\fi}\catcode`\%=\active\def
\end{pgfscope}%
\begin{pgfscope}%
\definecolor{textcolor}{rgb}{0.101961,0.101961,0.098039}%
\pgfsetstrokecolor{textcolor}%
\pgfsetfillcolor{textcolor}%
\pgftext[x=0.182222in,y=0.920000in,,bottom,rotate=90.000000]{\color{textcolor}{\rmfamily\fontsize{9.000000}{10.800000}\selectfont\catcode`\^=\active\def^{\ifmmode\sp\else\^{}\fi}\catcode`\%=\active\def
\end{pgfscope}%
\begin{pgfscope}%
\pgfpathrectangle{\pgfqpoint{0.460000in}{0.520000in}}{\pgfqpoint{0.800000in}{0.800000in}}%
\pgfusepath{clip}%
\pgfsetrectcap%
\pgfsetroundjoin%
\pgfsetlinewidth{1.003750pt}%
\definecolor{currentstroke}{rgb}{0.031373,0.564706,0.600000}%
\pgfsetstrokecolor{currentstroke}%
\pgfsetdash{}{0pt}%
\pgfpathmoveto{\pgfqpoint{0.496364in}{0.877079in}}%
\pgfpathlineto{\pgfqpoint{0.787273in}{0.881817in}}%
\pgfpathlineto{\pgfqpoint{0.932727in}{0.909144in}}%
\pgfpathlineto{\pgfqpoint{1.078182in}{0.919512in}}%
\pgfpathlineto{\pgfqpoint{1.223636in}{0.931325in}}%
\pgfusepath{stroke}%
\end{pgfscope}%
\begin{pgfscope}%
\pgfpathrectangle{\pgfqpoint{0.460000in}{0.520000in}}{\pgfqpoint{0.800000in}{0.800000in}}%
\pgfusepath{clip}%
\pgfsetbuttcap%
\pgfsetroundjoin%
\definecolor{currentfill}{rgb}{0.031373,0.564706,0.600000}%
\pgfsetfillcolor{currentfill}%
\pgfsetlinewidth{1.003750pt}%
\definecolor{currentstroke}{rgb}{0.031373,0.564706,0.600000}%
\pgfsetstrokecolor{currentstroke}%
\pgfsetdash{}{0pt}%
\pgfsys@defobject{currentmarker}{\pgfqpoint{-0.019444in}{-0.019444in}}{\pgfqpoint{0.019444in}{0.019444in}}{%
\pgfpathmoveto{\pgfqpoint{0.000000in}{-0.019444in}}%
\pgfpathcurveto{\pgfqpoint{0.005157in}{-0.019444in}}{\pgfqpoint{0.010103in}{-0.017396in}}{\pgfqpoint{0.013749in}{-0.013749in}}%
\pgfpathcurveto{\pgfqpoint{0.017396in}{-0.010103in}}{\pgfqpoint{0.019444in}{-0.005157in}}{\pgfqpoint{0.019444in}{0.000000in}}%
\pgfpathcurveto{\pgfqpoint{0.019444in}{0.005157in}}{\pgfqpoint{0.017396in}{0.010103in}}{\pgfqpoint{0.013749in}{0.013749in}}%
\pgfpathcurveto{\pgfqpoint{0.010103in}{0.017396in}}{\pgfqpoint{0.005157in}{0.019444in}}{\pgfqpoint{0.000000in}{0.019444in}}%
\pgfpathcurveto{\pgfqpoint{-0.005157in}{0.019444in}}{\pgfqpoint{-0.010103in}{0.017396in}}{\pgfqpoint{-0.013749in}{0.013749in}}%
\pgfpathcurveto{\pgfqpoint{-0.017396in}{0.010103in}}{\pgfqpoint{-0.019444in}{0.005157in}}{\pgfqpoint{-0.019444in}{0.000000in}}%
\pgfpathcurveto{\pgfqpoint{-0.019444in}{-0.005157in}}{\pgfqpoint{-0.017396in}{-0.010103in}}{\pgfqpoint{-0.013749in}{-0.013749in}}%
\pgfpathcurveto{\pgfqpoint{-0.010103in}{-0.017396in}}{\pgfqpoint{-0.005157in}{-0.019444in}}{\pgfqpoint{0.000000in}{-0.019444in}}%
\pgfpathlineto{\pgfqpoint{0.000000in}{-0.019444in}}%
\pgfpathclose%
\pgfusepath{stroke,fill}%
}%
\begin{pgfscope}%
\pgfsys@transformshift{0.496364in}{0.877079in}%
\pgfsys@useobject{currentmarker}{}%
\end{pgfscope}%
\begin{pgfscope}%
\pgfsys@transformshift{0.787273in}{0.881817in}%
\pgfsys@useobject{currentmarker}{}%
\end{pgfscope}%
\begin{pgfscope}%
\pgfsys@transformshift{0.932727in}{0.909144in}%
\pgfsys@useobject{currentmarker}{}%
\end{pgfscope}%
\begin{pgfscope}%
\pgfsys@transformshift{1.078182in}{0.919512in}%
\pgfsys@useobject{currentmarker}{}%
\end{pgfscope}%
\begin{pgfscope}%
\pgfsys@transformshift{1.223636in}{0.931325in}%
\pgfsys@useobject{currentmarker}{}%
\end{pgfscope}%
\end{pgfscope}%
\begin{pgfscope}%
\pgfpathrectangle{\pgfqpoint{0.460000in}{0.520000in}}{\pgfqpoint{0.800000in}{0.800000in}}%
\pgfusepath{clip}%
\pgfsetrectcap%
\pgfsetroundjoin%
\pgfsetlinewidth{1.003750pt}%
\definecolor{currentstroke}{rgb}{0.862745,0.223529,0.466667}%
\pgfsetstrokecolor{currentstroke}%
\pgfsetdash{}{0pt}%
\pgfpathmoveto{\pgfqpoint{0.496364in}{0.919636in}}%
\pgfpathlineto{\pgfqpoint{0.787273in}{0.916566in}}%
\pgfpathlineto{\pgfqpoint{0.932727in}{0.918198in}}%
\pgfpathlineto{\pgfqpoint{1.078182in}{0.911657in}}%
\pgfpathlineto{\pgfqpoint{1.223636in}{0.917587in}}%
\pgfusepath{stroke}%
\end{pgfscope}%
\begin{pgfscope}%
\pgfpathrectangle{\pgfqpoint{0.460000in}{0.520000in}}{\pgfqpoint{0.800000in}{0.800000in}}%
\pgfusepath{clip}%
\pgfsetbuttcap%
\pgfsetroundjoin%
\definecolor{currentfill}{rgb}{0.862745,0.223529,0.466667}%
\pgfsetfillcolor{currentfill}%
\pgfsetlinewidth{1.003750pt}%
\definecolor{currentstroke}{rgb}{0.862745,0.223529,0.466667}%
\pgfsetstrokecolor{currentstroke}%
\pgfsetdash{}{0pt}%
\pgfsys@defobject{currentmarker}{\pgfqpoint{-0.019444in}{-0.019444in}}{\pgfqpoint{0.019444in}{0.019444in}}{%
\pgfpathmoveto{\pgfqpoint{0.000000in}{-0.019444in}}%
\pgfpathcurveto{\pgfqpoint{0.005157in}{-0.019444in}}{\pgfqpoint{0.010103in}{-0.017396in}}{\pgfqpoint{0.013749in}{-0.013749in}}%
\pgfpathcurveto{\pgfqpoint{0.017396in}{-0.010103in}}{\pgfqpoint{0.019444in}{-0.005157in}}{\pgfqpoint{0.019444in}{0.000000in}}%
\pgfpathcurveto{\pgfqpoint{0.019444in}{0.005157in}}{\pgfqpoint{0.017396in}{0.010103in}}{\pgfqpoint{0.013749in}{0.013749in}}%
\pgfpathcurveto{\pgfqpoint{0.010103in}{0.017396in}}{\pgfqpoint{0.005157in}{0.019444in}}{\pgfqpoint{0.000000in}{0.019444in}}%
\pgfpathcurveto{\pgfqpoint{-0.005157in}{0.019444in}}{\pgfqpoint{-0.010103in}{0.017396in}}{\pgfqpoint{-0.013749in}{0.013749in}}%
\pgfpathcurveto{\pgfqpoint{-0.017396in}{0.010103in}}{\pgfqpoint{-0.019444in}{0.005157in}}{\pgfqpoint{-0.019444in}{0.000000in}}%
\pgfpathcurveto{\pgfqpoint{-0.019444in}{-0.005157in}}{\pgfqpoint{-0.017396in}{-0.010103in}}{\pgfqpoint{-0.013749in}{-0.013749in}}%
\pgfpathcurveto{\pgfqpoint{-0.010103in}{-0.017396in}}{\pgfqpoint{-0.005157in}{-0.019444in}}{\pgfqpoint{0.000000in}{-0.019444in}}%
\pgfpathlineto{\pgfqpoint{0.000000in}{-0.019444in}}%
\pgfpathclose%
\pgfusepath{stroke,fill}%
}%
\begin{pgfscope}%
\pgfsys@transformshift{0.496364in}{0.919636in}%
\pgfsys@useobject{currentmarker}{}%
\end{pgfscope}%
\begin{pgfscope}%
\pgfsys@transformshift{0.787273in}{0.916566in}%
\pgfsys@useobject{currentmarker}{}%
\end{pgfscope}%
\begin{pgfscope}%
\pgfsys@transformshift{0.932727in}{0.918198in}%
\pgfsys@useobject{currentmarker}{}%
\end{pgfscope}%
\begin{pgfscope}%
\pgfsys@transformshift{1.078182in}{0.911657in}%
\pgfsys@useobject{currentmarker}{}%
\end{pgfscope}%
\begin{pgfscope}%
\pgfsys@transformshift{1.223636in}{0.917587in}%
\pgfsys@useobject{currentmarker}{}%
\end{pgfscope}%
\end{pgfscope}%
\begin{pgfscope}%
\pgfpathrectangle{\pgfqpoint{0.460000in}{0.520000in}}{\pgfqpoint{0.800000in}{0.800000in}}%
\pgfusepath{clip}%
\pgfsetrectcap%
\pgfsetroundjoin%
\pgfsetlinewidth{1.003750pt}%
\definecolor{currentstroke}{rgb}{0.015686,0.321569,0.458824}%
\pgfsetstrokecolor{currentstroke}%
\pgfsetdash{}{0pt}%
\pgfpathmoveto{\pgfqpoint{0.496364in}{0.905048in}}%
\pgfpathlineto{\pgfqpoint{0.787273in}{0.927522in}}%
\pgfpathlineto{\pgfqpoint{0.932727in}{0.920189in}}%
\pgfusepath{stroke}%
\end{pgfscope}%
\begin{pgfscope}%
\pgfpathrectangle{\pgfqpoint{0.460000in}{0.520000in}}{\pgfqpoint{0.800000in}{0.800000in}}%
\pgfusepath{clip}%
\pgfsetbuttcap%
\pgfsetroundjoin%
\definecolor{currentfill}{rgb}{0.015686,0.321569,0.458824}%
\pgfsetfillcolor{currentfill}%
\pgfsetlinewidth{1.003750pt}%
\definecolor{currentstroke}{rgb}{0.015686,0.321569,0.458824}%
\pgfsetstrokecolor{currentstroke}%
\pgfsetdash{}{0pt}%
\pgfsys@defobject{currentmarker}{\pgfqpoint{-0.019444in}{-0.019444in}}{\pgfqpoint{0.019444in}{0.019444in}}{%
\pgfpathmoveto{\pgfqpoint{0.000000in}{-0.019444in}}%
\pgfpathcurveto{\pgfqpoint{0.005157in}{-0.019444in}}{\pgfqpoint{0.010103in}{-0.017396in}}{\pgfqpoint{0.013749in}{-0.013749in}}%
\pgfpathcurveto{\pgfqpoint{0.017396in}{-0.010103in}}{\pgfqpoint{0.019444in}{-0.005157in}}{\pgfqpoint{0.019444in}{0.000000in}}%
\pgfpathcurveto{\pgfqpoint{0.019444in}{0.005157in}}{\pgfqpoint{0.017396in}{0.010103in}}{\pgfqpoint{0.013749in}{0.013749in}}%
\pgfpathcurveto{\pgfqpoint{0.010103in}{0.017396in}}{\pgfqpoint{0.005157in}{0.019444in}}{\pgfqpoint{0.000000in}{0.019444in}}%
\pgfpathcurveto{\pgfqpoint{-0.005157in}{0.019444in}}{\pgfqpoint{-0.010103in}{0.017396in}}{\pgfqpoint{-0.013749in}{0.013749in}}%
\pgfpathcurveto{\pgfqpoint{-0.017396in}{0.010103in}}{\pgfqpoint{-0.019444in}{0.005157in}}{\pgfqpoint{-0.019444in}{0.000000in}}%
\pgfpathcurveto{\pgfqpoint{-0.019444in}{-0.005157in}}{\pgfqpoint{-0.017396in}{-0.010103in}}{\pgfqpoint{-0.013749in}{-0.013749in}}%
\pgfpathcurveto{\pgfqpoint{-0.010103in}{-0.017396in}}{\pgfqpoint{-0.005157in}{-0.019444in}}{\pgfqpoint{0.000000in}{-0.019444in}}%
\pgfpathlineto{\pgfqpoint{0.000000in}{-0.019444in}}%
\pgfpathclose%
\pgfusepath{stroke,fill}%
}%
\begin{pgfscope}%
\pgfsys@transformshift{0.496364in}{0.905048in}%
\pgfsys@useobject{currentmarker}{}%
\end{pgfscope}%
\begin{pgfscope}%
\pgfsys@transformshift{0.787273in}{0.927522in}%
\pgfsys@useobject{currentmarker}{}%
\end{pgfscope}%
\begin{pgfscope}%
\pgfsys@transformshift{0.932727in}{0.920189in}%
\pgfsys@useobject{currentmarker}{}%
\end{pgfscope}%
\end{pgfscope}%
\begin{pgfscope}%
\pgfsetrectcap%
\pgfsetmiterjoin%
\pgfsetlinewidth{0.602250pt}%
\definecolor{currentstroke}{rgb}{0.321569,0.317647,0.305882}%
\pgfsetstrokecolor{currentstroke}%
\pgfsetdash{}{0pt}%
\pgfpathmoveto{\pgfqpoint{0.460000in}{0.520000in}}%
\pgfpathlineto{\pgfqpoint{0.460000in}{1.320000in}}%
\pgfusepath{stroke}%
\end{pgfscope}%
\begin{pgfscope}%
\pgfsetrectcap%
\pgfsetmiterjoin%
\pgfsetlinewidth{0.602250pt}%
\definecolor{currentstroke}{rgb}{0.321569,0.317647,0.305882}%
\pgfsetstrokecolor{currentstroke}%
\pgfsetdash{}{0pt}%
\pgfpathmoveto{\pgfqpoint{0.460000in}{0.520000in}}%
\pgfpathlineto{\pgfqpoint{1.260000in}{0.520000in}}%
\pgfusepath{stroke}%
\end{pgfscope}%
\begin{pgfscope}%
\pgfsetrectcap%
\pgfsetroundjoin%
\pgfsetlinewidth{1.003750pt}%
\definecolor{currentstroke}{rgb}{0.031373,0.564706,0.600000}%
\pgfsetstrokecolor{currentstroke}%
\pgfsetdash{}{0pt}%
\pgfpathmoveto{\pgfqpoint{0.044946in}{0.076667in}}%
\pgfpathlineto{\pgfqpoint{0.106057in}{0.076667in}}%
\pgfpathlineto{\pgfqpoint{0.167168in}{0.076667in}}%
\pgfusepath{stroke}%
\end{pgfscope}%
\begin{pgfscope}%
\pgfsetbuttcap%
\pgfsetroundjoin%
\definecolor{currentfill}{rgb}{0.031373,0.564706,0.600000}%
\pgfsetfillcolor{currentfill}%
\pgfsetlinewidth{1.003750pt}%
\definecolor{currentstroke}{rgb}{0.031373,0.564706,0.600000}%
\pgfsetstrokecolor{currentstroke}%
\pgfsetdash{}{0pt}%
\pgfsys@defobject{currentmarker}{\pgfqpoint{-0.019444in}{-0.019444in}}{\pgfqpoint{0.019444in}{0.019444in}}{%
\pgfpathmoveto{\pgfqpoint{0.000000in}{-0.019444in}}%
\pgfpathcurveto{\pgfqpoint{0.005157in}{-0.019444in}}{\pgfqpoint{0.010103in}{-0.017396in}}{\pgfqpoint{0.013749in}{-0.013749in}}%
\pgfpathcurveto{\pgfqpoint{0.017396in}{-0.010103in}}{\pgfqpoint{0.019444in}{-0.005157in}}{\pgfqpoint{0.019444in}{0.000000in}}%
\pgfpathcurveto{\pgfqpoint{0.019444in}{0.005157in}}{\pgfqpoint{0.017396in}{0.010103in}}{\pgfqpoint{0.013749in}{0.013749in}}%
\pgfpathcurveto{\pgfqpoint{0.010103in}{0.017396in}}{\pgfqpoint{0.005157in}{0.019444in}}{\pgfqpoint{0.000000in}{0.019444in}}%
\pgfpathcurveto{\pgfqpoint{-0.005157in}{0.019444in}}{\pgfqpoint{-0.010103in}{0.017396in}}{\pgfqpoint{-0.013749in}{0.013749in}}%
\pgfpathcurveto{\pgfqpoint{-0.017396in}{0.010103in}}{\pgfqpoint{-0.019444in}{0.005157in}}{\pgfqpoint{-0.019444in}{0.000000in}}%
\pgfpathcurveto{\pgfqpoint{-0.019444in}{-0.005157in}}{\pgfqpoint{-0.017396in}{-0.010103in}}{\pgfqpoint{-0.013749in}{-0.013749in}}%
\pgfpathcurveto{\pgfqpoint{-0.010103in}{-0.017396in}}{\pgfqpoint{-0.005157in}{-0.019444in}}{\pgfqpoint{0.000000in}{-0.019444in}}%
\pgfpathlineto{\pgfqpoint{0.000000in}{-0.019444in}}%
\pgfpathclose%
\pgfusepath{stroke,fill}%
}%
\begin{pgfscope}%
\pgfsys@transformshift{0.106057in}{0.076667in}%
\pgfsys@useobject{currentmarker}{}%
\end{pgfscope}%
\end{pgfscope}%
\begin{pgfscope}%
\definecolor{textcolor}{rgb}{0.101961,0.101961,0.098039}%
\pgfsetstrokecolor{textcolor}%
\pgfsetfillcolor{textcolor}%
\pgftext[x=0.200501in,y=0.037778in,left,base]{\color{textcolor}{\rmfamily\fontsize{8.000000}{9.600000}\selectfont\catcode`\^=\active\def^{\ifmmode\sp\else\^{}\fi}\catcode`\%=\active\def
\end{pgfscope}%
\begin{pgfscope}%
\pgfsetrectcap%
\pgfsetroundjoin%
\pgfsetlinewidth{1.003750pt}%
\definecolor{currentstroke}{rgb}{0.862745,0.223529,0.466667}%
\pgfsetstrokecolor{currentstroke}%
\pgfsetdash{}{0pt}%
\pgfpathmoveto{\pgfqpoint{0.502389in}{0.076667in}}%
\pgfpathlineto{\pgfqpoint{0.563500in}{0.076667in}}%
\pgfpathlineto{\pgfqpoint{0.624612in}{0.076667in}}%
\pgfusepath{stroke}%
\end{pgfscope}%
\begin{pgfscope}%
\pgfsetbuttcap%
\pgfsetroundjoin%
\definecolor{currentfill}{rgb}{0.862745,0.223529,0.466667}%
\pgfsetfillcolor{currentfill}%
\pgfsetlinewidth{1.003750pt}%
\definecolor{currentstroke}{rgb}{0.862745,0.223529,0.466667}%
\pgfsetstrokecolor{currentstroke}%
\pgfsetdash{}{0pt}%
\pgfsys@defobject{currentmarker}{\pgfqpoint{-0.019444in}{-0.019444in}}{\pgfqpoint{0.019444in}{0.019444in}}{%
\pgfpathmoveto{\pgfqpoint{0.000000in}{-0.019444in}}%
\pgfpathcurveto{\pgfqpoint{0.005157in}{-0.019444in}}{\pgfqpoint{0.010103in}{-0.017396in}}{\pgfqpoint{0.013749in}{-0.013749in}}%
\pgfpathcurveto{\pgfqpoint{0.017396in}{-0.010103in}}{\pgfqpoint{0.019444in}{-0.005157in}}{\pgfqpoint{0.019444in}{0.000000in}}%
\pgfpathcurveto{\pgfqpoint{0.019444in}{0.005157in}}{\pgfqpoint{0.017396in}{0.010103in}}{\pgfqpoint{0.013749in}{0.013749in}}%
\pgfpathcurveto{\pgfqpoint{0.010103in}{0.017396in}}{\pgfqpoint{0.005157in}{0.019444in}}{\pgfqpoint{0.000000in}{0.019444in}}%
\pgfpathcurveto{\pgfqpoint{-0.005157in}{0.019444in}}{\pgfqpoint{-0.010103in}{0.017396in}}{\pgfqpoint{-0.013749in}{0.013749in}}%
\pgfpathcurveto{\pgfqpoint{-0.017396in}{0.010103in}}{\pgfqpoint{-0.019444in}{0.005157in}}{\pgfqpoint{-0.019444in}{0.000000in}}%
\pgfpathcurveto{\pgfqpoint{-0.019444in}{-0.005157in}}{\pgfqpoint{-0.017396in}{-0.010103in}}{\pgfqpoint{-0.013749in}{-0.013749in}}%
\pgfpathcurveto{\pgfqpoint{-0.010103in}{-0.017396in}}{\pgfqpoint{-0.005157in}{-0.019444in}}{\pgfqpoint{0.000000in}{-0.019444in}}%
\pgfpathlineto{\pgfqpoint{0.000000in}{-0.019444in}}%
\pgfpathclose%
\pgfusepath{stroke,fill}%
}%
\begin{pgfscope}%
\pgfsys@transformshift{0.563500in}{0.076667in}%
\pgfsys@useobject{currentmarker}{}%
\end{pgfscope}%
\end{pgfscope}%
\begin{pgfscope}%
\definecolor{textcolor}{rgb}{0.101961,0.101961,0.098039}%
\pgfsetstrokecolor{textcolor}%
\pgfsetfillcolor{textcolor}%
\pgftext[x=0.657945in,y=0.037778in,left,base]{\color{textcolor}{\rmfamily\fontsize{8.000000}{9.600000}\selectfont\catcode`\^=\active\def^{\ifmmode\sp\else\^{}\fi}\catcode`\%=\active\def
\end{pgfscope}%
\begin{pgfscope}%
\pgfsetrectcap%
\pgfsetroundjoin%
\pgfsetlinewidth{1.003750pt}%
\definecolor{currentstroke}{rgb}{0.015686,0.321569,0.458824}%
\pgfsetstrokecolor{currentstroke}%
\pgfsetdash{}{0pt}%
\pgfpathmoveto{\pgfqpoint{0.959833in}{0.076667in}}%
\pgfpathlineto{\pgfqpoint{1.020944in}{0.076667in}}%
\pgfpathlineto{\pgfqpoint{1.082055in}{0.076667in}}%
\pgfusepath{stroke}%
\end{pgfscope}%
\begin{pgfscope}%
\pgfsetbuttcap%
\pgfsetroundjoin%
\definecolor{currentfill}{rgb}{0.015686,0.321569,0.458824}%
\pgfsetfillcolor{currentfill}%
\pgfsetlinewidth{1.003750pt}%
\definecolor{currentstroke}{rgb}{0.015686,0.321569,0.458824}%
\pgfsetstrokecolor{currentstroke}%
\pgfsetdash{}{0pt}%
\pgfsys@defobject{currentmarker}{\pgfqpoint{-0.019444in}{-0.019444in}}{\pgfqpoint{0.019444in}{0.019444in}}{%
\pgfpathmoveto{\pgfqpoint{0.000000in}{-0.019444in}}%
\pgfpathcurveto{\pgfqpoint{0.005157in}{-0.019444in}}{\pgfqpoint{0.010103in}{-0.017396in}}{\pgfqpoint{0.013749in}{-0.013749in}}%
\pgfpathcurveto{\pgfqpoint{0.017396in}{-0.010103in}}{\pgfqpoint{0.019444in}{-0.005157in}}{\pgfqpoint{0.019444in}{0.000000in}}%
\pgfpathcurveto{\pgfqpoint{0.019444in}{0.005157in}}{\pgfqpoint{0.017396in}{0.010103in}}{\pgfqpoint{0.013749in}{0.013749in}}%
\pgfpathcurveto{\pgfqpoint{0.010103in}{0.017396in}}{\pgfqpoint{0.005157in}{0.019444in}}{\pgfqpoint{0.000000in}{0.019444in}}%
\pgfpathcurveto{\pgfqpoint{-0.005157in}{0.019444in}}{\pgfqpoint{-0.010103in}{0.017396in}}{\pgfqpoint{-0.013749in}{0.013749in}}%
\pgfpathcurveto{\pgfqpoint{-0.017396in}{0.010103in}}{\pgfqpoint{-0.019444in}{0.005157in}}{\pgfqpoint{-0.019444in}{0.000000in}}%
\pgfpathcurveto{\pgfqpoint{-0.019444in}{-0.005157in}}{\pgfqpoint{-0.017396in}{-0.010103in}}{\pgfqpoint{-0.013749in}{-0.013749in}}%
\pgfpathcurveto{\pgfqpoint{-0.010103in}{-0.017396in}}{\pgfqpoint{-0.005157in}{-0.019444in}}{\pgfqpoint{0.000000in}{-0.019444in}}%
\pgfpathlineto{\pgfqpoint{0.000000in}{-0.019444in}}%
\pgfpathclose%
\pgfusepath{stroke,fill}%
}%
\begin{pgfscope}%
\pgfsys@transformshift{1.020944in}{0.076667in}%
\pgfsys@useobject{currentmarker}{}%
\end{pgfscope}%
\end{pgfscope}%
\begin{pgfscope}%
\definecolor{textcolor}{rgb}{0.101961,0.101961,0.098039}%
\pgfsetstrokecolor{textcolor}%
\pgfsetfillcolor{textcolor}%
\pgftext[x=1.115388in,y=0.037778in,left,base]{\color{textcolor}{\rmfamily\fontsize{8.000000}{9.600000}\selectfont\catcode`\^=\active\def^{\ifmmode\sp\else\^{}\fi}\catcode`\%=\active\def
\end{pgfscope}%
\end{pgfpicture}%
\makeatother%
\endgroup%

%% file: figures/pgf/fig_kernels.pgf
\providecommand{\mathdefault}[1]{#1}%
\begingroup%
\makeatletter%
\begin{pgfpicture}%
\pgfpathrectangle{\pgfpointorigin}{\pgfqpoint{1.290000in}{1.380000in}}%
\pgfusepath{use as bounding box, clip}%
\begin{pgfscope}%
\pgfsetbuttcap%
\pgfsetmiterjoin%
\definecolor{currentfill}{rgb}{1.000000,1.000000,1.000000}%
\pgfsetfillcolor{currentfill}%
\pgfsetlinewidth{0.000000pt}%
\definecolor{currentstroke}{rgb}{1.000000,1.000000,1.000000}%
\pgfsetstrokecolor{currentstroke}%
\pgfsetdash{}{0pt}%
\pgfpathmoveto{\pgfqpoint{0.000000in}{0.000000in}}%
\pgfpathlineto{\pgfqpoint{1.290000in}{0.000000in}}%
\pgfpathlineto{\pgfqpoint{1.290000in}{1.380000in}}%
\pgfpathlineto{\pgfqpoint{0.000000in}{1.380000in}}%
\pgfpathlineto{\pgfqpoint{0.000000in}{0.000000in}}%
\pgfpathclose%
\pgfusepath{fill}%
\end{pgfscope}%
\begin{pgfscope}%
\pgfsetbuttcap%
\pgfsetmiterjoin%
\definecolor{currentfill}{rgb}{1.000000,1.000000,1.000000}%
\pgfsetfillcolor{currentfill}%
\pgfsetlinewidth{0.000000pt}%
\definecolor{currentstroke}{rgb}{0.000000,0.000000,0.000000}%
\pgfsetstrokecolor{currentstroke}%
\pgfsetstrokeopacity{0.000000}%
\pgfsetdash{}{0pt}%
\pgfpathmoveto{\pgfqpoint{0.460000in}{0.520000in}}%
\pgfpathlineto{\pgfqpoint{1.260000in}{0.520000in}}%
\pgfpathlineto{\pgfqpoint{1.260000in}{1.320000in}}%
\pgfpathlineto{\pgfqpoint{0.460000in}{1.320000in}}%
\pgfpathlineto{\pgfqpoint{0.460000in}{0.520000in}}%
\pgfpathclose%
\pgfusepath{fill}%
\end{pgfscope}%
\begin{pgfscope}%
\definecolor{textcolor}{rgb}{0.101961,0.101961,0.098039}%
\pgfsetstrokecolor{textcolor}%
\pgfsetfillcolor{textcolor}%
\pgftext[x=0.537419in,y=0.471389in,right,top,rotate=60.000000]{\color{textcolor}\rmfamily\fontsize{8.000000}{9.600000}\selectfont LA}%
\end{pgfscope}%
\begin{pgfscope}%
\definecolor{textcolor}{rgb}{0.101961,0.101961,0.098039}%
\pgfsetstrokecolor{textcolor}%
\pgfsetfillcolor{textcolor}%
\pgftext[x=0.666452in,y=0.471389in,right,top,rotate=60.000000]{\color{textcolor}\rmfamily\fontsize{8.000000}{9.600000}\selectfont M2}%
\end{pgfscope}%
\begin{pgfscope}%
\definecolor{textcolor}{rgb}{0.101961,0.101961,0.098039}%
\pgfsetstrokecolor{textcolor}%
\pgfsetfillcolor{textcolor}%
\pgftext[x=0.795484in,y=0.471389in,right,top,rotate=60.000000]{\color{textcolor}\rmfamily\fontsize{8.000000}{9.600000}\selectfont GLA}%
\end{pgfscope}%
\begin{pgfscope}%
\definecolor{textcolor}{rgb}{0.101961,0.101961,0.098039}%
\pgfsetstrokecolor{textcolor}%
\pgfsetfillcolor{textcolor}%
\pgftext[x=0.924516in,y=0.471389in,right,top,rotate=60.000000]{\color{textcolor}\rmfamily\fontsize{8.000000}{9.600000}\selectfont DN}%
\end{pgfscope}%
\begin{pgfscope}%
\definecolor{textcolor}{rgb}{0.101961,0.101961,0.098039}%
\pgfsetstrokecolor{textcolor}%
\pgfsetfillcolor{textcolor}%
\pgftext[x=1.053548in,y=0.471389in,right,top,rotate=60.000000]{\color{textcolor}\rmfamily\fontsize{8.000000}{9.600000}\selectfont GDN}%
\end{pgfscope}%
\begin{pgfscope}%
\definecolor{textcolor}{rgb}{0.101961,0.101961,0.098039}%
\pgfsetstrokecolor{textcolor}%
\pgfsetfillcolor{textcolor}%
\pgftext[x=1.182581in,y=0.471389in,right,top,rotate=60.000000]{\color{textcolor}\rmfamily\fontsize{8.000000}{9.600000}\selectfont KDA}%
\end{pgfscope}%
\begin{pgfscope}%
\pgfpathrectangle{\pgfqpoint{0.460000in}{0.520000in}}{\pgfqpoint{0.800000in}{0.800000in}}%
\pgfusepath{clip}%
\pgfsetrectcap%
\pgfsetroundjoin%
\pgfsetlinewidth{0.501875pt}%
\definecolor{currentstroke}{rgb}{0.886275,0.882353,0.866667}%
\pgfsetstrokecolor{currentstroke}%
\pgfsetdash{}{0pt}%
\pgfpathmoveto{\pgfqpoint{0.460000in}{0.520000in}}%
\pgfpathlineto{\pgfqpoint{1.260000in}{0.520000in}}%
\pgfusepath{stroke}%
\end{pgfscope}%
\begin{pgfscope}%
\pgfsetbuttcap%
\pgfsetroundjoin%
\definecolor{currentfill}{rgb}{0.321569,0.317647,0.305882}%
\pgfsetfillcolor{currentfill}%
\pgfsetlinewidth{0.602250pt}%
\definecolor{currentstroke}{rgb}{0.321569,0.317647,0.305882}%
\pgfsetstrokecolor{currentstroke}%
\pgfsetdash{}{0pt}%
\pgfsys@defobject{currentmarker}{\pgfqpoint{-0.034722in}{0.000000in}}{\pgfqpoint{-0.000000in}{0.000000in}}{%
\pgfpathmoveto{\pgfqpoint{-0.000000in}{0.000000in}}%
\pgfpathlineto{\pgfqpoint{-0.034722in}{0.000000in}}%
\pgfusepath{stroke,fill}%
}%
\begin{pgfscope}%
\pgfsys@transformshift{0.460000in}{0.520000in}%
\pgfsys@useobject{currentmarker}{}%
\end{pgfscope}%
\end{pgfscope}%
\begin{pgfscope}%
\definecolor{textcolor}{rgb}{0.101961,0.101961,0.098039}%
\pgfsetstrokecolor{textcolor}%
\pgfsetfillcolor{textcolor}%
\pgftext[x=0.321111in, y=0.482112in, left, base]{\color{textcolor}\rmfamily\fontsize{8.000000}{9.600000}\selectfont 0}%
\end{pgfscope}%
\begin{pgfscope}%
\pgfpathrectangle{\pgfqpoint{0.460000in}{0.520000in}}{\pgfqpoint{0.800000in}{0.800000in}}%
\pgfusepath{clip}%
\pgfsetrectcap%
\pgfsetroundjoin%
\pgfsetlinewidth{0.501875pt}%
\definecolor{currentstroke}{rgb}{0.886275,0.882353,0.866667}%
\pgfsetstrokecolor{currentstroke}%
\pgfsetdash{}{0pt}%
\pgfpathmoveto{\pgfqpoint{0.460000in}{0.710476in}}%
\pgfpathlineto{\pgfqpoint{1.260000in}{0.710476in}}%
\pgfusepath{stroke}%
\end{pgfscope}%
\begin{pgfscope}%
\pgfsetbuttcap%
\pgfsetroundjoin%
\definecolor{currentfill}{rgb}{0.321569,0.317647,0.305882}%
\pgfsetfillcolor{currentfill}%
\pgfsetlinewidth{0.602250pt}%
\definecolor{currentstroke}{rgb}{0.321569,0.317647,0.305882}%
\pgfsetstrokecolor{currentstroke}%
\pgfsetdash{}{0pt}%
\pgfsys@defobject{currentmarker}{\pgfqpoint{-0.034722in}{0.000000in}}{\pgfqpoint{-0.000000in}{0.000000in}}{%
\pgfpathmoveto{\pgfqpoint{-0.000000in}{0.000000in}}%
\pgfpathlineto{\pgfqpoint{-0.034722in}{0.000000in}}%
\pgfusepath{stroke,fill}%
}%
\begin{pgfscope}%
\pgfsys@transformshift{0.460000in}{0.710476in}%
\pgfsys@useobject{currentmarker}{}%
\end{pgfscope}%
\end{pgfscope}%
\begin{pgfscope}%
\definecolor{textcolor}{rgb}{0.101961,0.101961,0.098039}%
\pgfsetstrokecolor{textcolor}%
\pgfsetfillcolor{textcolor}%
\pgftext[x=0.321111in, y=0.672588in, left, base]{\color{textcolor}\rmfamily\fontsize{8.000000}{9.600000}\selectfont 1}%
\end{pgfscope}%
\begin{pgfscope}%
\pgfpathrectangle{\pgfqpoint{0.460000in}{0.520000in}}{\pgfqpoint{0.800000in}{0.800000in}}%
\pgfusepath{clip}%
\pgfsetrectcap%
\pgfsetroundjoin%
\pgfsetlinewidth{0.501875pt}%
\definecolor{currentstroke}{rgb}{0.886275,0.882353,0.866667}%
\pgfsetstrokecolor{currentstroke}%
\pgfsetdash{}{0pt}%
\pgfpathmoveto{\pgfqpoint{0.460000in}{0.900952in}}%
\pgfpathlineto{\pgfqpoint{1.260000in}{0.900952in}}%
\pgfusepath{stroke}%
\end{pgfscope}%
\begin{pgfscope}%
\pgfsetbuttcap%
\pgfsetroundjoin%
\definecolor{currentfill}{rgb}{0.321569,0.317647,0.305882}%
\pgfsetfillcolor{currentfill}%
\pgfsetlinewidth{0.602250pt}%
\definecolor{currentstroke}{rgb}{0.321569,0.317647,0.305882}%
\pgfsetstrokecolor{currentstroke}%
\pgfsetdash{}{0pt}%
\pgfsys@defobject{currentmarker}{\pgfqpoint{-0.034722in}{0.000000in}}{\pgfqpoint{-0.000000in}{0.000000in}}{%
\pgfpathmoveto{\pgfqpoint{-0.000000in}{0.000000in}}%
\pgfpathlineto{\pgfqpoint{-0.034722in}{0.000000in}}%
\pgfusepath{stroke,fill}%
}%
\begin{pgfscope}%
\pgfsys@transformshift{0.460000in}{0.900952in}%
\pgfsys@useobject{currentmarker}{}%
\end{pgfscope}%
\end{pgfscope}%
\begin{pgfscope}%
\definecolor{textcolor}{rgb}{0.101961,0.101961,0.098039}%
\pgfsetstrokecolor{textcolor}%
\pgfsetfillcolor{textcolor}%
\pgftext[x=0.321111in, y=0.863064in, left, base]{\color{textcolor}\rmfamily\fontsize{8.000000}{9.600000}\selectfont 2}%
\end{pgfscope}%
\begin{pgfscope}%
\pgfpathrectangle{\pgfqpoint{0.460000in}{0.520000in}}{\pgfqpoint{0.800000in}{0.800000in}}%
\pgfusepath{clip}%
\pgfsetrectcap%
\pgfsetroundjoin%
\pgfsetlinewidth{0.501875pt}%
\definecolor{currentstroke}{rgb}{0.886275,0.882353,0.866667}%
\pgfsetstrokecolor{currentstroke}%
\pgfsetdash{}{0pt}%
\pgfpathmoveto{\pgfqpoint{0.460000in}{1.091429in}}%
\pgfpathlineto{\pgfqpoint{1.260000in}{1.091429in}}%
\pgfusepath{stroke}%
\end{pgfscope}%
\begin{pgfscope}%
\pgfsetbuttcap%
\pgfsetroundjoin%
\definecolor{currentfill}{rgb}{0.321569,0.317647,0.305882}%
\pgfsetfillcolor{currentfill}%
\pgfsetlinewidth{0.602250pt}%
\definecolor{currentstroke}{rgb}{0.321569,0.317647,0.305882}%
\pgfsetstrokecolor{currentstroke}%
\pgfsetdash{}{0pt}%
\pgfsys@defobject{currentmarker}{\pgfqpoint{-0.034722in}{0.000000in}}{\pgfqpoint{-0.000000in}{0.000000in}}{%
\pgfpathmoveto{\pgfqpoint{-0.000000in}{0.000000in}}%
\pgfpathlineto{\pgfqpoint{-0.034722in}{0.000000in}}%
\pgfusepath{stroke,fill}%
}%
\begin{pgfscope}%
\pgfsys@transformshift{0.460000in}{1.091429in}%
\pgfsys@useobject{currentmarker}{}%
\end{pgfscope}%
\end{pgfscope}%
\begin{pgfscope}%
\definecolor{textcolor}{rgb}{0.101961,0.101961,0.098039}%
\pgfsetstrokecolor{textcolor}%
\pgfsetfillcolor{textcolor}%
\pgftext[x=0.321111in, y=1.053540in, left, base]{\color{textcolor}\rmfamily\fontsize{8.000000}{9.600000}\selectfont 3}%
\end{pgfscope}%
\begin{pgfscope}%
\pgfpathrectangle{\pgfqpoint{0.460000in}{0.520000in}}{\pgfqpoint{0.800000in}{0.800000in}}%
\pgfusepath{clip}%
\pgfsetrectcap%
\pgfsetroundjoin%
\pgfsetlinewidth{0.501875pt}%
\definecolor{currentstroke}{rgb}{0.886275,0.882353,0.866667}%
\pgfsetstrokecolor{currentstroke}%
\pgfsetdash{}{0pt}%
\pgfpathmoveto{\pgfqpoint{0.460000in}{1.281905in}}%
\pgfpathlineto{\pgfqpoint{1.260000in}{1.281905in}}%
\pgfusepath{stroke}%
\end{pgfscope}%
\begin{pgfscope}%
\pgfsetbuttcap%
\pgfsetroundjoin%
\definecolor{currentfill}{rgb}{0.321569,0.317647,0.305882}%
\pgfsetfillcolor{currentfill}%
\pgfsetlinewidth{0.602250pt}%
\definecolor{currentstroke}{rgb}{0.321569,0.317647,0.305882}%
\pgfsetstrokecolor{currentstroke}%
\pgfsetdash{}{0pt}%
\pgfsys@defobject{currentmarker}{\pgfqpoint{-0.034722in}{0.000000in}}{\pgfqpoint{-0.000000in}{0.000000in}}{%
\pgfpathmoveto{\pgfqpoint{-0.000000in}{0.000000in}}%
\pgfpathlineto{\pgfqpoint{-0.034722in}{0.000000in}}%
\pgfusepath{stroke,fill}%
}%
\begin{pgfscope}%
\pgfsys@transformshift{0.460000in}{1.281905in}%
\pgfsys@useobject{currentmarker}{}%
\end{pgfscope}%
\end{pgfscope}%
\begin{pgfscope}%
\definecolor{textcolor}{rgb}{0.101961,0.101961,0.098039}%
\pgfsetstrokecolor{textcolor}%
\pgfsetfillcolor{textcolor}%
\pgftext[x=0.321111in, y=1.244017in, left, base]{\color{textcolor}\rmfamily\fontsize{8.000000}{9.600000}\selectfont 4}%
\end{pgfscope}%
\begin{pgfscope}%
\definecolor{textcolor}{rgb}{0.101961,0.101961,0.098039}%
\pgfsetstrokecolor{textcolor}%
\pgfsetfillcolor{textcolor}%
\pgftext[x=0.265556in,y=0.920000in,,bottom,rotate=90.000000]{\color{textcolor}\rmfamily\fontsize{9.000000}{10.800000}\selectfont relative overhead}%
\end{pgfscope}%
\begin{pgfscope}%
\pgfpathrectangle{\pgfqpoint{0.460000in}{0.520000in}}{\pgfqpoint{0.800000in}{0.800000in}}%
\pgfusepath{clip}%
\pgfsetbuttcap%
\pgfsetroundjoin%
\pgfsetlinewidth{0.803000pt}%
\definecolor{currentstroke}{rgb}{0.321569,0.317647,0.305882}%
\pgfsetstrokecolor{currentstroke}%
\pgfsetdash{{2.960000pt}{1.280000pt}}{0.000000pt}%
\pgfpathmoveto{\pgfqpoint{0.460000in}{0.710476in}}%
\pgfpathlineto{\pgfqpoint{1.260000in}{0.710476in}}%
\pgfusepath{stroke}%
\end{pgfscope}%
\begin{pgfscope}%
\pgfpathrectangle{\pgfqpoint{0.460000in}{0.520000in}}{\pgfqpoint{0.800000in}{0.800000in}}%
\pgfusepath{clip}%
\pgfsetbuttcap%
\pgfsetmiterjoin%
\definecolor{currentfill}{rgb}{0.031373,0.564706,0.600000}%
\pgfsetfillcolor{currentfill}%
\pgfsetlinewidth{0.000000pt}%
\definecolor{currentstroke}{rgb}{0.000000,0.000000,0.000000}%
\pgfsetstrokecolor{currentstroke}%
\pgfsetstrokeopacity{0.000000}%
\pgfsetdash{}{0pt}%
\pgfpathmoveto{\pgfqpoint{0.497419in}{0.520000in}}%
\pgfpathlineto{\pgfqpoint{0.577419in}{0.520000in}}%
\pgfpathlineto{\pgfqpoint{0.577419in}{0.782050in}}%
\pgfpathlineto{\pgfqpoint{0.497419in}{0.782050in}}%
\pgfpathlineto{\pgfqpoint{0.497419in}{0.520000in}}%
\pgfpathclose%
\pgfusepath{fill}%
\end{pgfscope}%
\begin{pgfscope}%
\pgfpathrectangle{\pgfqpoint{0.460000in}{0.520000in}}{\pgfqpoint{0.800000in}{0.800000in}}%
\pgfusepath{clip}%
\pgfsetbuttcap%
\pgfsetmiterjoin%
\definecolor{currentfill}{rgb}{0.031373,0.564706,0.600000}%
\pgfsetfillcolor{currentfill}%
\pgfsetlinewidth{0.000000pt}%
\definecolor{currentstroke}{rgb}{0.000000,0.000000,0.000000}%
\pgfsetstrokecolor{currentstroke}%
\pgfsetstrokeopacity{0.000000}%
\pgfsetdash{}{0pt}%
\pgfpathmoveto{\pgfqpoint{0.626452in}{0.520000in}}%
\pgfpathlineto{\pgfqpoint{0.706452in}{0.520000in}}%
\pgfpathlineto{\pgfqpoint{0.706452in}{0.789185in}}%
\pgfpathlineto{\pgfqpoint{0.626452in}{0.789185in}}%
\pgfpathlineto{\pgfqpoint{0.626452in}{0.520000in}}%
\pgfpathclose%
\pgfusepath{fill}%
\end{pgfscope}%
\begin{pgfscope}%
\pgfpathrectangle{\pgfqpoint{0.460000in}{0.520000in}}{\pgfqpoint{0.800000in}{0.800000in}}%
\pgfusepath{clip}%
\pgfsetbuttcap%
\pgfsetmiterjoin%
\definecolor{currentfill}{rgb}{0.031373,0.564706,0.600000}%
\pgfsetfillcolor{currentfill}%
\pgfsetlinewidth{0.000000pt}%
\definecolor{currentstroke}{rgb}{0.000000,0.000000,0.000000}%
\pgfsetstrokecolor{currentstroke}%
\pgfsetstrokeopacity{0.000000}%
\pgfsetdash{}{0pt}%
\pgfpathmoveto{\pgfqpoint{0.755484in}{0.520000in}}%
\pgfpathlineto{\pgfqpoint{0.835484in}{0.520000in}}%
\pgfpathlineto{\pgfqpoint{0.835484in}{0.778620in}}%
\pgfpathlineto{\pgfqpoint{0.755484in}{0.778620in}}%
\pgfpathlineto{\pgfqpoint{0.755484in}{0.520000in}}%
\pgfpathclose%
\pgfusepath{fill}%
\end{pgfscope}%
\begin{pgfscope}%
\pgfpathrectangle{\pgfqpoint{0.460000in}{0.520000in}}{\pgfqpoint{0.800000in}{0.800000in}}%
\pgfusepath{clip}%
\pgfsetbuttcap%
\pgfsetmiterjoin%
\definecolor{currentfill}{rgb}{0.031373,0.564706,0.600000}%
\pgfsetfillcolor{currentfill}%
\pgfsetlinewidth{0.000000pt}%
\definecolor{currentstroke}{rgb}{0.000000,0.000000,0.000000}%
\pgfsetstrokecolor{currentstroke}%
\pgfsetstrokeopacity{0.000000}%
\pgfsetdash{}{0pt}%
\pgfpathmoveto{\pgfqpoint{0.884516in}{0.520000in}}%
\pgfpathlineto{\pgfqpoint{0.964516in}{0.520000in}}%
\pgfpathlineto{\pgfqpoint{0.964516in}{1.128212in}}%
\pgfpathlineto{\pgfqpoint{0.884516in}{1.128212in}}%
\pgfpathlineto{\pgfqpoint{0.884516in}{0.520000in}}%
\pgfpathclose%
\pgfusepath{fill}%
\end{pgfscope}%
\begin{pgfscope}%
\pgfpathrectangle{\pgfqpoint{0.460000in}{0.520000in}}{\pgfqpoint{0.800000in}{0.800000in}}%
\pgfusepath{clip}%
\pgfsetbuttcap%
\pgfsetmiterjoin%
\definecolor{currentfill}{rgb}{0.031373,0.564706,0.600000}%
\pgfsetfillcolor{currentfill}%
\pgfsetlinewidth{0.000000pt}%
\definecolor{currentstroke}{rgb}{0.000000,0.000000,0.000000}%
\pgfsetstrokecolor{currentstroke}%
\pgfsetstrokeopacity{0.000000}%
\pgfsetdash{}{0pt}%
\pgfpathmoveto{\pgfqpoint{1.013548in}{0.520000in}}%
\pgfpathlineto{\pgfqpoint{1.093548in}{0.520000in}}%
\pgfpathlineto{\pgfqpoint{1.093548in}{1.072188in}}%
\pgfpathlineto{\pgfqpoint{1.013548in}{1.072188in}}%
\pgfpathlineto{\pgfqpoint{1.013548in}{0.520000in}}%
\pgfpathclose%
\pgfusepath{fill}%
\end{pgfscope}%
\begin{pgfscope}%
\pgfpathrectangle{\pgfqpoint{0.460000in}{0.520000in}}{\pgfqpoint{0.800000in}{0.800000in}}%
\pgfusepath{clip}%
\pgfsetbuttcap%
\pgfsetmiterjoin%
\definecolor{currentfill}{rgb}{0.031373,0.564706,0.600000}%
\pgfsetfillcolor{currentfill}%
\pgfsetlinewidth{0.000000pt}%
\definecolor{currentstroke}{rgb}{0.000000,0.000000,0.000000}%
\pgfsetstrokecolor{currentstroke}%
\pgfsetstrokeopacity{0.000000}%
\pgfsetdash{}{0pt}%
\pgfpathmoveto{\pgfqpoint{1.142581in}{0.520000in}}%
\pgfpathlineto{\pgfqpoint{1.222581in}{0.520000in}}%
\pgfpathlineto{\pgfqpoint{1.222581in}{0.872329in}}%
\pgfpathlineto{\pgfqpoint{1.142581in}{0.872329in}}%
\pgfpathlineto{\pgfqpoint{1.142581in}{0.520000in}}%
\pgfpathclose%
\pgfusepath{fill}%
\end{pgfscope}%
\begin{pgfscope}%
\pgfsetrectcap%
\pgfsetmiterjoin%
\pgfsetlinewidth{0.602250pt}%
\definecolor{currentstroke}{rgb}{0.321569,0.317647,0.305882}%
\pgfsetstrokecolor{currentstroke}%
\pgfsetdash{}{0pt}%
\pgfpathmoveto{\pgfqpoint{0.460000in}{0.520000in}}%
\pgfpathlineto{\pgfqpoint{0.460000in}{1.320000in}}%
\pgfusepath{stroke}%
\end{pgfscope}%
\begin{pgfscope}%
\pgfsetrectcap%
\pgfsetmiterjoin%
\pgfsetlinewidth{0.602250pt}%
\definecolor{currentstroke}{rgb}{0.321569,0.317647,0.305882}%
\pgfsetstrokecolor{currentstroke}%
\pgfsetdash{}{0pt}%
\pgfpathmoveto{\pgfqpoint{0.460000in}{0.520000in}}%
\pgfpathlineto{\pgfqpoint{1.260000in}{0.520000in}}%
\pgfusepath{stroke}%
\end{pgfscope}%
\begin{pgfscope}%
\pgfpathrectangle{\pgfqpoint{0.460000in}{0.520000in}}{\pgfqpoint{0.800000in}{0.800000in}}%
\pgfusepath{clip}%
\pgfsetrectcap%
\pgfsetroundjoin%
\pgfsetlinewidth{0.702625pt}%
\definecolor{currentstroke}{rgb}{0.101961,0.101961,0.098039}%
\pgfsetstrokecolor{currentstroke}%
\pgfsetdash{}{0pt}%
\pgfpathmoveto{\pgfqpoint{0.537419in}{0.776768in}}%
\pgfpathlineto{\pgfqpoint{0.537419in}{0.786091in}}%
\pgfusepath{stroke}%
\end{pgfscope}%
\begin{pgfscope}%
\pgfpathrectangle{\pgfqpoint{0.460000in}{0.520000in}}{\pgfqpoint{0.800000in}{0.800000in}}%
\pgfusepath{clip}%
\pgfsetrectcap%
\pgfsetroundjoin%
\pgfsetlinewidth{0.702625pt}%
\definecolor{currentstroke}{rgb}{0.101961,0.101961,0.098039}%
\pgfsetstrokecolor{currentstroke}%
\pgfsetdash{}{0pt}%
\pgfpathmoveto{\pgfqpoint{0.666452in}{0.781889in}}%
\pgfpathlineto{\pgfqpoint{0.666452in}{0.791161in}}%
\pgfusepath{stroke}%
\end{pgfscope}%
\begin{pgfscope}%
\pgfpathrectangle{\pgfqpoint{0.460000in}{0.520000in}}{\pgfqpoint{0.800000in}{0.800000in}}%
\pgfusepath{clip}%
\pgfsetrectcap%
\pgfsetroundjoin%
\pgfsetlinewidth{0.702625pt}%
\definecolor{currentstroke}{rgb}{0.101961,0.101961,0.098039}%
\pgfsetstrokecolor{currentstroke}%
\pgfsetdash{}{0pt}%
\pgfpathmoveto{\pgfqpoint{0.795484in}{0.777972in}}%
\pgfpathlineto{\pgfqpoint{0.795484in}{0.779690in}}%
\pgfusepath{stroke}%
\end{pgfscope}%
\begin{pgfscope}%
\pgfpathrectangle{\pgfqpoint{0.460000in}{0.520000in}}{\pgfqpoint{0.800000in}{0.800000in}}%
\pgfusepath{clip}%
\pgfsetrectcap%
\pgfsetroundjoin%
\pgfsetlinewidth{0.702625pt}%
\definecolor{currentstroke}{rgb}{0.101961,0.101961,0.098039}%
\pgfsetstrokecolor{currentstroke}%
\pgfsetdash{}{0pt}%
\pgfpathmoveto{\pgfqpoint{0.924516in}{1.126821in}}%
\pgfpathlineto{\pgfqpoint{0.924516in}{1.132258in}}%
\pgfusepath{stroke}%
\end{pgfscope}%
\begin{pgfscope}%
\pgfpathrectangle{\pgfqpoint{0.460000in}{0.520000in}}{\pgfqpoint{0.800000in}{0.800000in}}%
\pgfusepath{clip}%
\pgfsetrectcap%
\pgfsetroundjoin%
\pgfsetlinewidth{0.702625pt}%
\definecolor{currentstroke}{rgb}{0.101961,0.101961,0.098039}%
\pgfsetstrokecolor{currentstroke}%
\pgfsetdash{}{0pt}%
\pgfpathmoveto{\pgfqpoint{1.053548in}{1.070230in}}%
\pgfpathlineto{\pgfqpoint{1.053548in}{1.076809in}}%
\pgfusepath{stroke}%
\end{pgfscope}%
\begin{pgfscope}%
\pgfpathrectangle{\pgfqpoint{0.460000in}{0.520000in}}{\pgfqpoint{0.800000in}{0.800000in}}%
\pgfusepath{clip}%
\pgfsetrectcap%
\pgfsetroundjoin%
\pgfsetlinewidth{0.702625pt}%
\definecolor{currentstroke}{rgb}{0.101961,0.101961,0.098039}%
\pgfsetstrokecolor{currentstroke}%
\pgfsetdash{}{0pt}%
\pgfpathmoveto{\pgfqpoint{1.182581in}{0.872150in}}%
\pgfpathlineto{\pgfqpoint{1.182581in}{0.872994in}}%
\pgfusepath{stroke}%
\end{pgfscope}%
\end{pgfpicture}%
\makeatother%
\endgroup%

%% file: figures/pgf/fig_crossover.pgf
\providecommand{\mathdefault}[1]{#1}%
\begingroup%
\makeatletter%
\begin{pgfpicture}%
\pgfpathrectangle{\pgfpointorigin}{\pgfqpoint{1.290000in}{1.380000in}}%
\pgfusepath{use as bounding box, clip}%
\begin{pgfscope}%
\pgfsetbuttcap%
\pgfsetmiterjoin%
\definecolor{currentfill}{rgb}{1.000000,1.000000,1.000000}%
\pgfsetfillcolor{currentfill}%
\pgfsetlinewidth{0.000000pt}%
\definecolor{currentstroke}{rgb}{1.000000,1.000000,1.000000}%
\pgfsetstrokecolor{currentstroke}%
\pgfsetdash{}{0pt}%
\pgfpathmoveto{\pgfqpoint{0.000000in}{0.000000in}}%
\pgfpathlineto{\pgfqpoint{1.290000in}{0.000000in}}%
\pgfpathlineto{\pgfqpoint{1.290000in}{1.380000in}}%
\pgfpathlineto{\pgfqpoint{0.000000in}{1.380000in}}%
\pgfpathlineto{\pgfqpoint{0.000000in}{0.000000in}}%
\pgfpathclose%
\pgfusepath{fill}%
\end{pgfscope}%
\begin{pgfscope}%
\pgfsetbuttcap%
\pgfsetmiterjoin%
\definecolor{currentfill}{rgb}{1.000000,1.000000,1.000000}%
\pgfsetfillcolor{currentfill}%
\pgfsetlinewidth{0.000000pt}%
\definecolor{currentstroke}{rgb}{0.000000,0.000000,0.000000}%
\pgfsetstrokecolor{currentstroke}%
\pgfsetstrokeopacity{0.000000}%
\pgfsetdash{}{0pt}%
\pgfpathmoveto{\pgfqpoint{0.460000in}{0.520000in}}%
\pgfpathlineto{\pgfqpoint{1.260000in}{0.520000in}}%
\pgfpathlineto{\pgfqpoint{1.260000in}{1.320000in}}%
\pgfpathlineto{\pgfqpoint{0.460000in}{1.320000in}}%
\pgfpathlineto{\pgfqpoint{0.460000in}{0.520000in}}%
\pgfpathclose%
\pgfusepath{fill}%
\end{pgfscope}%
\begin{pgfscope}%
\pgfpathrectangle{\pgfqpoint{0.460000in}{0.520000in}}{\pgfqpoint{0.800000in}{0.800000in}}%
\pgfusepath{clip}%
\pgfsetbuttcap%
\pgfsetroundjoin%
\pgfsetlinewidth{0.602250pt}%
\definecolor{currentstroke}{rgb}{0.321569,0.317647,0.305882}%
\pgfsetstrokecolor{currentstroke}%
\pgfsetdash{{0.600000pt}{0.990000pt}}{0.000000pt}%
\pgfpathmoveto{\pgfqpoint{0.460000in}{0.786667in}}%
\pgfpathlineto{\pgfqpoint{1.260000in}{0.786667in}}%
\pgfusepath{stroke}%
\end{pgfscope}%
\begin{pgfscope}%
\pgfpathrectangle{\pgfqpoint{0.460000in}{0.520000in}}{\pgfqpoint{0.800000in}{0.800000in}}%
\pgfusepath{clip}%
\pgfsetrectcap%
\pgfsetroundjoin%
\pgfsetlinewidth{0.501875pt}%
\definecolor{currentstroke}{rgb}{0.886275,0.882353,0.866667}%
\pgfsetstrokecolor{currentstroke}%
\pgfsetdash{}{0pt}%
\pgfpathmoveto{\pgfqpoint{0.546818in}{0.520000in}}%
\pgfpathlineto{\pgfqpoint{0.546818in}{1.320000in}}%
\pgfusepath{stroke}%
\end{pgfscope}%
\begin{pgfscope}%
\pgfsetbuttcap%
\pgfsetroundjoin%
\definecolor{currentfill}{rgb}{0.321569,0.317647,0.305882}%
\pgfsetfillcolor{currentfill}%
\pgfsetlinewidth{0.602250pt}%
\definecolor{currentstroke}{rgb}{0.321569,0.317647,0.305882}%
\pgfsetstrokecolor{currentstroke}%
\pgfsetdash{}{0pt}%
\pgfsys@defobject{currentmarker}{\pgfqpoint{0.000000in}{-0.034722in}}{\pgfqpoint{0.000000in}{0.000000in}}{%
\pgfpathmoveto{\pgfqpoint{0.000000in}{0.000000in}}%
\pgfpathlineto{\pgfqpoint{0.000000in}{-0.034722in}}%
\pgfusepath{stroke,fill}%
}%
\begin{pgfscope}%
\pgfsys@transformshift{0.546818in}{0.520000in}%
\pgfsys@useobject{currentmarker}{}%
\end{pgfscope}%
\end{pgfscope}%
\begin{pgfscope}%
\definecolor{textcolor}{rgb}{0.101961,0.101961,0.098039}%
\pgfsetstrokecolor{textcolor}%
\pgfsetfillcolor{textcolor}%
\pgftext[x=0.546818in,y=0.436667in,,top]{\color{textcolor}{\rmfamily\fontsize{8.000000}{9.600000}\selectfont\catcode`\^=\active\def^{\ifmmode\sp\else\^{}\fi}\catcode`\%=\active\def
\end{pgfscope}%
\begin{pgfscope}%
\pgfpathrectangle{\pgfqpoint{0.460000in}{0.520000in}}{\pgfqpoint{0.800000in}{0.800000in}}%
\pgfusepath{clip}%
\pgfsetrectcap%
\pgfsetroundjoin%
\pgfsetlinewidth{0.501875pt}%
\definecolor{currentstroke}{rgb}{0.886275,0.882353,0.866667}%
\pgfsetstrokecolor{currentstroke}%
\pgfsetdash{}{0pt}%
\pgfpathmoveto{\pgfqpoint{0.756000in}{0.520000in}}%
\pgfpathlineto{\pgfqpoint{0.756000in}{1.320000in}}%
\pgfusepath{stroke}%
\end{pgfscope}%
\begin{pgfscope}%
\pgfsetbuttcap%
\pgfsetroundjoin%
\definecolor{currentfill}{rgb}{0.321569,0.317647,0.305882}%
\pgfsetfillcolor{currentfill}%
\pgfsetlinewidth{0.602250pt}%
\definecolor{currentstroke}{rgb}{0.321569,0.317647,0.305882}%
\pgfsetstrokecolor{currentstroke}%
\pgfsetdash{}{0pt}%
\pgfsys@defobject{currentmarker}{\pgfqpoint{0.000000in}{-0.034722in}}{\pgfqpoint{0.000000in}{0.000000in}}{%
\pgfpathmoveto{\pgfqpoint{0.000000in}{0.000000in}}%
\pgfpathlineto{\pgfqpoint{0.000000in}{-0.034722in}}%
\pgfusepath{stroke,fill}%
}%
\begin{pgfscope}%
\pgfsys@transformshift{0.756000in}{0.520000in}%
\pgfsys@useobject{currentmarker}{}%
\end{pgfscope}%
\end{pgfscope}%
\begin{pgfscope}%
\definecolor{textcolor}{rgb}{0.101961,0.101961,0.098039}%
\pgfsetstrokecolor{textcolor}%
\pgfsetfillcolor{textcolor}%
\pgftext[x=0.756000in,y=0.436667in,,top]{\color{textcolor}{\rmfamily\fontsize{8.000000}{9.600000}\selectfont\catcode`\^=\active\def^{\ifmmode\sp\else\^{}\fi}\catcode`\%=\active\def
\end{pgfscope}%
\begin{pgfscope}%
\pgfpathrectangle{\pgfqpoint{0.460000in}{0.520000in}}{\pgfqpoint{0.800000in}{0.800000in}}%
\pgfusepath{clip}%
\pgfsetrectcap%
\pgfsetroundjoin%
\pgfsetlinewidth{0.501875pt}%
\definecolor{currentstroke}{rgb}{0.886275,0.882353,0.866667}%
\pgfsetstrokecolor{currentstroke}%
\pgfsetdash{}{0pt}%
\pgfpathmoveto{\pgfqpoint{0.965181in}{0.520000in}}%
\pgfpathlineto{\pgfqpoint{0.965181in}{1.320000in}}%
\pgfusepath{stroke}%
\end{pgfscope}%
\begin{pgfscope}%
\pgfsetbuttcap%
\pgfsetroundjoin%
\definecolor{currentfill}{rgb}{0.321569,0.317647,0.305882}%
\pgfsetfillcolor{currentfill}%
\pgfsetlinewidth{0.602250pt}%
\definecolor{currentstroke}{rgb}{0.321569,0.317647,0.305882}%
\pgfsetstrokecolor{currentstroke}%
\pgfsetdash{}{0pt}%
\pgfsys@defobject{currentmarker}{\pgfqpoint{0.000000in}{-0.034722in}}{\pgfqpoint{0.000000in}{0.000000in}}{%
\pgfpathmoveto{\pgfqpoint{0.000000in}{0.000000in}}%
\pgfpathlineto{\pgfqpoint{0.000000in}{-0.034722in}}%
\pgfusepath{stroke,fill}%
}%
\begin{pgfscope}%
\pgfsys@transformshift{0.965181in}{0.520000in}%
\pgfsys@useobject{currentmarker}{}%
\end{pgfscope}%
\end{pgfscope}%
\begin{pgfscope}%
\definecolor{textcolor}{rgb}{0.101961,0.101961,0.098039}%
\pgfsetstrokecolor{textcolor}%
\pgfsetfillcolor{textcolor}%
\pgftext[x=0.965181in,y=0.436667in,,top]{\color{textcolor}{\rmfamily\fontsize{8.000000}{9.600000}\selectfont\catcode`\^=\active\def^{\ifmmode\sp\else\^{}\fi}\catcode`\%=\active\def
\end{pgfscope}%
\begin{pgfscope}%
\pgfpathrectangle{\pgfqpoint{0.460000in}{0.520000in}}{\pgfqpoint{0.800000in}{0.800000in}}%
\pgfusepath{clip}%
\pgfsetrectcap%
\pgfsetroundjoin%
\pgfsetlinewidth{0.501875pt}%
\definecolor{currentstroke}{rgb}{0.886275,0.882353,0.866667}%
\pgfsetstrokecolor{currentstroke}%
\pgfsetdash{}{0pt}%
\pgfpathmoveto{\pgfqpoint{1.174363in}{0.520000in}}%
\pgfpathlineto{\pgfqpoint{1.174363in}{1.320000in}}%
\pgfusepath{stroke}%
\end{pgfscope}%
\begin{pgfscope}%
\pgfsetbuttcap%
\pgfsetroundjoin%
\definecolor{currentfill}{rgb}{0.321569,0.317647,0.305882}%
\pgfsetfillcolor{currentfill}%
\pgfsetlinewidth{0.602250pt}%
\definecolor{currentstroke}{rgb}{0.321569,0.317647,0.305882}%
\pgfsetstrokecolor{currentstroke}%
\pgfsetdash{}{0pt}%
\pgfsys@defobject{currentmarker}{\pgfqpoint{0.000000in}{-0.034722in}}{\pgfqpoint{0.000000in}{0.000000in}}{%
\pgfpathmoveto{\pgfqpoint{0.000000in}{0.000000in}}%
\pgfpathlineto{\pgfqpoint{0.000000in}{-0.034722in}}%
\pgfusepath{stroke,fill}%
}%
\begin{pgfscope}%
\pgfsys@transformshift{1.174363in}{0.520000in}%
\pgfsys@useobject{currentmarker}{}%
\end{pgfscope}%
\end{pgfscope}%
\begin{pgfscope}%
\definecolor{textcolor}{rgb}{0.101961,0.101961,0.098039}%
\pgfsetstrokecolor{textcolor}%
\pgfsetfillcolor{textcolor}%
\pgftext[x=1.174363in,y=0.436667in,,top]{\color{textcolor}{\rmfamily\fontsize{8.000000}{9.600000}\selectfont\catcode`\^=\active\def^{\ifmmode\sp\else\^{}\fi}\catcode`\%=\active\def
\end{pgfscope}%
\begin{pgfscope}%
\definecolor{textcolor}{rgb}{0.101961,0.101961,0.098039}%
\pgfsetstrokecolor{textcolor}%
\pgfsetfillcolor{textcolor}%
\pgftext[x=0.860000in,y=0.309058in,,top]{\color{textcolor}{\rmfamily\fontsize{9.000000}{10.800000}\selectfont\catcode`\^=\active\def^{\ifmmode\sp\else\^{}\fi}\catcode`\%=\active\def
\end{pgfscope}%
\begin{pgfscope}%
\pgfpathrectangle{\pgfqpoint{0.460000in}{0.520000in}}{\pgfqpoint{0.800000in}{0.800000in}}%
\pgfusepath{clip}%
\pgfsetrectcap%
\pgfsetroundjoin%
\pgfsetlinewidth{0.501875pt}%
\definecolor{currentstroke}{rgb}{0.886275,0.882353,0.866667}%
\pgfsetstrokecolor{currentstroke}%
\pgfsetdash{}{0pt}%
\pgfpathmoveto{\pgfqpoint{0.460000in}{0.520000in}}%
\pgfpathlineto{\pgfqpoint{1.260000in}{0.520000in}}%
\pgfusepath{stroke}%
\end{pgfscope}%
\begin{pgfscope}%
\pgfsetbuttcap%
\pgfsetroundjoin%
\definecolor{currentfill}{rgb}{0.321569,0.317647,0.305882}%
\pgfsetfillcolor{currentfill}%
\pgfsetlinewidth{0.602250pt}%
\definecolor{currentstroke}{rgb}{0.321569,0.317647,0.305882}%
\pgfsetstrokecolor{currentstroke}%
\pgfsetdash{}{0pt}%
\pgfsys@defobject{currentmarker}{\pgfqpoint{-0.034722in}{0.000000in}}{\pgfqpoint{-0.000000in}{0.000000in}}{%
\pgfpathmoveto{\pgfqpoint{-0.000000in}{0.000000in}}%
\pgfpathlineto{\pgfqpoint{-0.034722in}{0.000000in}}%
\pgfusepath{stroke,fill}%
}%
\begin{pgfscope}%
\pgfsys@transformshift{0.460000in}{0.520000in}%
\pgfsys@useobject{currentmarker}{}%
\end{pgfscope}%
\end{pgfscope}%
\begin{pgfscope}%
\definecolor{textcolor}{rgb}{0.101961,0.101961,0.098039}%
\pgfsetstrokecolor{textcolor}%
\pgfsetfillcolor{textcolor}%
\pgftext[x=0.293333in, y=0.482112in, left, base]{\color{textcolor}{\rmfamily\fontsize{8.000000}{9.600000}\selectfont\catcode`\^=\active\def^{\ifmmode\sp\else\^{}\fi}\catcode`\%=\active\def
\end{pgfscope}%
\begin{pgfscope}%
\pgfpathrectangle{\pgfqpoint{0.460000in}{0.520000in}}{\pgfqpoint{0.800000in}{0.800000in}}%
\pgfusepath{clip}%
\pgfsetrectcap%
\pgfsetroundjoin%
\pgfsetlinewidth{0.501875pt}%
\definecolor{currentstroke}{rgb}{0.886275,0.882353,0.866667}%
\pgfsetstrokecolor{currentstroke}%
\pgfsetdash{}{0pt}%
\pgfpathmoveto{\pgfqpoint{0.460000in}{0.786667in}}%
\pgfpathlineto{\pgfqpoint{1.260000in}{0.786667in}}%
\pgfusepath{stroke}%
\end{pgfscope}%
\begin{pgfscope}%
\pgfsetbuttcap%
\pgfsetroundjoin%
\definecolor{currentfill}{rgb}{0.321569,0.317647,0.305882}%
\pgfsetfillcolor{currentfill}%
\pgfsetlinewidth{0.602250pt}%
\definecolor{currentstroke}{rgb}{0.321569,0.317647,0.305882}%
\pgfsetstrokecolor{currentstroke}%
\pgfsetdash{}{0pt}%
\pgfsys@defobject{currentmarker}{\pgfqpoint{-0.034722in}{0.000000in}}{\pgfqpoint{-0.000000in}{0.000000in}}{%
\pgfpathmoveto{\pgfqpoint{-0.000000in}{0.000000in}}%
\pgfpathlineto{\pgfqpoint{-0.034722in}{0.000000in}}%
\pgfusepath{stroke,fill}%
}%
\begin{pgfscope}%
\pgfsys@transformshift{0.460000in}{0.786667in}%
\pgfsys@useobject{currentmarker}{}%
\end{pgfscope}%
\end{pgfscope}%
\begin{pgfscope}%
\definecolor{textcolor}{rgb}{0.101961,0.101961,0.098039}%
\pgfsetstrokecolor{textcolor}%
\pgfsetfillcolor{textcolor}%
\pgftext[x=0.321111in, y=0.748778in, left, base]{\color{textcolor}{\rmfamily\fontsize{8.000000}{9.600000}\selectfont\catcode`\^=\active\def^{\ifmmode\sp\else\^{}\fi}\catcode`\%=\active\def
\end{pgfscope}%
\begin{pgfscope}%
\pgfpathrectangle{\pgfqpoint{0.460000in}{0.520000in}}{\pgfqpoint{0.800000in}{0.800000in}}%
\pgfusepath{clip}%
\pgfsetrectcap%
\pgfsetroundjoin%
\pgfsetlinewidth{0.501875pt}%
\definecolor{currentstroke}{rgb}{0.886275,0.882353,0.866667}%
\pgfsetstrokecolor{currentstroke}%
\pgfsetdash{}{0pt}%
\pgfpathmoveto{\pgfqpoint{0.460000in}{1.053333in}}%
\pgfpathlineto{\pgfqpoint{1.260000in}{1.053333in}}%
\pgfusepath{stroke}%
\end{pgfscope}%
\begin{pgfscope}%
\pgfsetbuttcap%
\pgfsetroundjoin%
\definecolor{currentfill}{rgb}{0.321569,0.317647,0.305882}%
\pgfsetfillcolor{currentfill}%
\pgfsetlinewidth{0.602250pt}%
\definecolor{currentstroke}{rgb}{0.321569,0.317647,0.305882}%
\pgfsetstrokecolor{currentstroke}%
\pgfsetdash{}{0pt}%
\pgfsys@defobject{currentmarker}{\pgfqpoint{-0.034722in}{0.000000in}}{\pgfqpoint{-0.000000in}{0.000000in}}{%
\pgfpathmoveto{\pgfqpoint{-0.000000in}{0.000000in}}%
\pgfpathlineto{\pgfqpoint{-0.034722in}{0.000000in}}%
\pgfusepath{stroke,fill}%
}%
\begin{pgfscope}%
\pgfsys@transformshift{0.460000in}{1.053333in}%
\pgfsys@useobject{currentmarker}{}%
\end{pgfscope}%
\end{pgfscope}%
\begin{pgfscope}%
\definecolor{textcolor}{rgb}{0.101961,0.101961,0.098039}%
\pgfsetstrokecolor{textcolor}%
\pgfsetfillcolor{textcolor}%
\pgftext[x=0.237778in, y=1.015445in, left, base]{\color{textcolor}{\rmfamily\fontsize{8.000000}{9.600000}\selectfont\catcode`\^=\active\def^{\ifmmode\sp\else\^{}\fi}\catcode`\%=\active\def
\end{pgfscope}%
\begin{pgfscope}%
\definecolor{textcolor}{rgb}{0.101961,0.101961,0.098039}%
\pgfsetstrokecolor{textcolor}%
\pgfsetfillcolor{textcolor}%
\pgftext[x=0.182222in,y=0.920000in,,bottom,rotate=90.000000]{\color{textcolor}{\rmfamily\fontsize{9.000000}{10.800000}\selectfont\catcode`\^=\active\def^{\ifmmode\sp\else\^{}\fi}\catcode`\%=\active\def
\end{pgfscope}%
\begin{pgfscope}%
\pgfpathrectangle{\pgfqpoint{0.460000in}{0.520000in}}{\pgfqpoint{0.800000in}{0.800000in}}%
\pgfusepath{clip}%
\pgfsetrectcap%
\pgfsetroundjoin%
\pgfsetlinewidth{1.003750pt}%
\definecolor{currentstroke}{rgb}{0.015686,0.321569,0.458824}%
\pgfsetstrokecolor{currentstroke}%
\pgfsetdash{}{0pt}%
\pgfpathmoveto{\pgfqpoint{0.546818in}{1.124292in}}%
\pgfpathlineto{\pgfqpoint{0.756000in}{0.929843in}}%
\pgfpathlineto{\pgfqpoint{0.965181in}{0.939613in}}%
\pgfpathlineto{\pgfqpoint{1.174363in}{0.917543in}}%
\pgfusepath{stroke}%
\end{pgfscope}%
\begin{pgfscope}%
\pgfpathrectangle{\pgfqpoint{0.460000in}{0.520000in}}{\pgfqpoint{0.800000in}{0.800000in}}%
\pgfusepath{clip}%
\pgfsetbuttcap%
\pgfsetmiterjoin%
\definecolor{currentfill}{rgb}{0.015686,0.321569,0.458824}%
\pgfsetfillcolor{currentfill}%
\pgfsetlinewidth{1.003750pt}%
\definecolor{currentstroke}{rgb}{0.015686,0.321569,0.458824}%
\pgfsetstrokecolor{currentstroke}%
\pgfsetdash{}{0pt}%
\pgfsys@defobject{currentmarker}{\pgfqpoint{-0.020833in}{-0.020833in}}{\pgfqpoint{0.020833in}{0.020833in}}{%
\pgfpathmoveto{\pgfqpoint{-0.020833in}{-0.020833in}}%
\pgfpathlineto{\pgfqpoint{0.020833in}{-0.020833in}}%
\pgfpathlineto{\pgfqpoint{0.020833in}{0.020833in}}%
\pgfpathlineto{\pgfqpoint{-0.020833in}{0.020833in}}%
\pgfpathlineto{\pgfqpoint{-0.020833in}{-0.020833in}}%
\pgfpathclose%
\pgfusepath{stroke,fill}%
}%
\begin{pgfscope}%
\pgfsys@transformshift{0.546818in}{1.124292in}%
\pgfsys@useobject{currentmarker}{}%
\end{pgfscope}%
\begin{pgfscope}%
\pgfsys@transformshift{0.756000in}{0.929843in}%
\pgfsys@useobject{currentmarker}{}%
\end{pgfscope}%
\begin{pgfscope}%
\pgfsys@transformshift{0.965181in}{0.939613in}%
\pgfsys@useobject{currentmarker}{}%
\end{pgfscope}%
\begin{pgfscope}%
\pgfsys@transformshift{1.174363in}{0.917543in}%
\pgfsys@useobject{currentmarker}{}%
\end{pgfscope}%
\end{pgfscope}%
\begin{pgfscope}%
\pgfpathrectangle{\pgfqpoint{0.460000in}{0.520000in}}{\pgfqpoint{0.800000in}{0.800000in}}%
\pgfusepath{clip}%
\pgfsetrectcap%
\pgfsetroundjoin%
\pgfsetlinewidth{1.003750pt}%
\definecolor{currentstroke}{rgb}{0.031373,0.564706,0.600000}%
\pgfsetstrokecolor{currentstroke}%
\pgfsetdash{}{0pt}%
\pgfpathmoveto{\pgfqpoint{0.546818in}{1.013215in}}%
\pgfpathlineto{\pgfqpoint{0.756000in}{0.852527in}}%
\pgfpathlineto{\pgfqpoint{0.965181in}{0.668544in}}%
\pgfpathlineto{\pgfqpoint{1.174363in}{0.578847in}}%
\pgfusepath{stroke}%
\end{pgfscope}%
\begin{pgfscope}%
\pgfpathrectangle{\pgfqpoint{0.460000in}{0.520000in}}{\pgfqpoint{0.800000in}{0.800000in}}%
\pgfusepath{clip}%
\pgfsetbuttcap%
\pgfsetmiterjoin%
\definecolor{currentfill}{rgb}{0.031373,0.564706,0.600000}%
\pgfsetfillcolor{currentfill}%
\pgfsetlinewidth{1.003750pt}%
\definecolor{currentstroke}{rgb}{0.031373,0.564706,0.600000}%
\pgfsetstrokecolor{currentstroke}%
\pgfsetdash{}{0pt}%
\pgfsys@defobject{currentmarker}{\pgfqpoint{-0.020833in}{-0.020833in}}{\pgfqpoint{0.020833in}{0.020833in}}{%
\pgfpathmoveto{\pgfqpoint{-0.020833in}{-0.020833in}}%
\pgfpathlineto{\pgfqpoint{0.020833in}{-0.020833in}}%
\pgfpathlineto{\pgfqpoint{0.020833in}{0.020833in}}%
\pgfpathlineto{\pgfqpoint{-0.020833in}{0.020833in}}%
\pgfpathlineto{\pgfqpoint{-0.020833in}{-0.020833in}}%
\pgfpathclose%
\pgfusepath{stroke,fill}%
}%
\begin{pgfscope}%
\pgfsys@transformshift{0.546818in}{1.013215in}%
\pgfsys@useobject{currentmarker}{}%
\end{pgfscope}%
\begin{pgfscope}%
\pgfsys@transformshift{0.756000in}{0.852527in}%
\pgfsys@useobject{currentmarker}{}%
\end{pgfscope}%
\begin{pgfscope}%
\pgfsys@transformshift{0.965181in}{0.668544in}%
\pgfsys@useobject{currentmarker}{}%
\end{pgfscope}%
\begin{pgfscope}%
\pgfsys@transformshift{1.174363in}{0.578847in}%
\pgfsys@useobject{currentmarker}{}%
\end{pgfscope}%
\end{pgfscope}%
\begin{pgfscope}%
\pgfsetrectcap%
\pgfsetmiterjoin%
\pgfsetlinewidth{0.602250pt}%
\definecolor{currentstroke}{rgb}{0.321569,0.317647,0.305882}%
\pgfsetstrokecolor{currentstroke}%
\pgfsetdash{}{0pt}%
\pgfpathmoveto{\pgfqpoint{0.460000in}{0.520000in}}%
\pgfpathlineto{\pgfqpoint{0.460000in}{1.320000in}}%
\pgfusepath{stroke}%
\end{pgfscope}%
\begin{pgfscope}%
\pgfsetrectcap%
\pgfsetmiterjoin%
\pgfsetlinewidth{0.602250pt}%
\definecolor{currentstroke}{rgb}{0.321569,0.317647,0.305882}%
\pgfsetstrokecolor{currentstroke}%
\pgfsetdash{}{0pt}%
\pgfpathmoveto{\pgfqpoint{0.460000in}{0.520000in}}%
\pgfpathlineto{\pgfqpoint{1.260000in}{0.520000in}}%
\pgfusepath{stroke}%
\end{pgfscope}%
\begin{pgfscope}%
\pgfpathrectangle{\pgfqpoint{0.460000in}{0.520000in}}{\pgfqpoint{0.800000in}{0.800000in}}%
\pgfusepath{clip}%
\pgfsetbuttcap%
\pgfsetmiterjoin%
\definecolor{currentfill}{rgb}{0.000000,0.000000,0.000000}%
\pgfsetfillcolor{currentfill}%
\pgfsetfillopacity{0.000000}%
\pgfsetlinewidth{0.903375pt}%
\definecolor{currentstroke}{rgb}{0.015686,0.321569,0.458824}%
\pgfsetstrokecolor{currentstroke}%
\pgfsetdash{}{0pt}%
\pgfsys@defobject{currentmarker}{\pgfqpoint{-0.038194in}{-0.038194in}}{\pgfqpoint{0.038194in}{0.038194in}}{%
\pgfpathmoveto{\pgfqpoint{-0.038194in}{-0.038194in}}%
\pgfpathlineto{\pgfqpoint{0.038194in}{-0.038194in}}%
\pgfpathlineto{\pgfqpoint{0.038194in}{0.038194in}}%
\pgfpathlineto{\pgfqpoint{-0.038194in}{0.038194in}}%
\pgfpathlineto{\pgfqpoint{-0.038194in}{-0.038194in}}%
\pgfpathclose%
\pgfusepath{stroke,fill}%
}%
\begin{pgfscope}%
\pgfsys@transformshift{0.965181in}{0.939613in}%
\pgfsys@useobject{currentmarker}{}%
\end{pgfscope}%
\end{pgfscope}%
\begin{pgfscope}%
\pgfpathrectangle{\pgfqpoint{0.460000in}{0.520000in}}{\pgfqpoint{0.800000in}{0.800000in}}%
\pgfusepath{clip}%
\pgfsetbuttcap%
\pgfsetmiterjoin%
\definecolor{currentfill}{rgb}{0.000000,0.000000,0.000000}%
\pgfsetfillcolor{currentfill}%
\pgfsetfillopacity{0.000000}%
\pgfsetlinewidth{0.903375pt}%
\definecolor{currentstroke}{rgb}{0.031373,0.564706,0.600000}%
\pgfsetstrokecolor{currentstroke}%
\pgfsetdash{}{0pt}%
\pgfsys@defobject{currentmarker}{\pgfqpoint{-0.038194in}{-0.038194in}}{\pgfqpoint{0.038194in}{0.038194in}}{%
\pgfpathmoveto{\pgfqpoint{-0.038194in}{-0.038194in}}%
\pgfpathlineto{\pgfqpoint{0.038194in}{-0.038194in}}%
\pgfpathlineto{\pgfqpoint{0.038194in}{0.038194in}}%
\pgfpathlineto{\pgfqpoint{-0.038194in}{0.038194in}}%
\pgfpathlineto{\pgfqpoint{-0.038194in}{-0.038194in}}%
\pgfpathclose%
\pgfusepath{stroke,fill}%
}%
\begin{pgfscope}%
\pgfsys@transformshift{0.756000in}{0.852527in}%
\pgfsys@useobject{currentmarker}{}%
\end{pgfscope}%
\end{pgfscope}%
\begin{pgfscope}%
\pgfsetbuttcap%
\pgfsetmiterjoin%
\definecolor{currentfill}{rgb}{0.000000,0.000000,0.000000}%
\pgfsetfillcolor{currentfill}%
\pgfsetfillopacity{0.000000}%
\pgfsetlinewidth{1.003750pt}%
\definecolor{currentstroke}{rgb}{0.321569,0.317647,0.305882}%
\pgfsetstrokecolor{currentstroke}%
\pgfsetdash{}{0pt}%
\pgfsys@defobject{currentmarker}{\pgfqpoint{-0.034722in}{-0.034722in}}{\pgfqpoint{0.034722in}{0.034722in}}{%
\pgfpathmoveto{\pgfqpoint{-0.034722in}{-0.034722in}}%
\pgfpathlineto{\pgfqpoint{0.034722in}{-0.034722in}}%
\pgfpathlineto{\pgfqpoint{0.034722in}{0.034722in}}%
\pgfpathlineto{\pgfqpoint{-0.034722in}{0.034722in}}%
\pgfpathlineto{\pgfqpoint{-0.034722in}{-0.034722in}}%
\pgfpathclose%
\pgfusepath{stroke,fill}%
}%
\begin{pgfscope}%
\pgfsys@transformshift{0.627558in}{1.214444in}%
\pgfsys@useobject{currentmarker}{}%
\end{pgfscope}%
\end{pgfscope}%
\begin{pgfscope}%
\definecolor{textcolor}{rgb}{0.101961,0.101961,0.098039}%
\pgfsetstrokecolor{textcolor}%
\pgfsetfillcolor{textcolor}%
\pgftext[x=0.716447in,y=1.175556in,left,base]{\color{textcolor}{\rmfamily\fontsize{8.000000}{9.600000}\selectfont\catcode`\^=\active\def^{\ifmmode\sp\else\^{}\fi}\catcode`\%=\active\def
\end{pgfscope}%
\begin{pgfscope}%
\pgfsetrectcap%
\pgfsetroundjoin%
\pgfsetlinewidth{1.003750pt}%
\definecolor{currentstroke}{rgb}{0.015686,0.321569,0.458824}%
\pgfsetstrokecolor{currentstroke}%
\pgfsetdash{}{0pt}%
\pgfpathmoveto{\pgfqpoint{-0.020588in}{0.076667in}}%
\pgfpathlineto{\pgfqpoint{0.040523in}{0.076667in}}%
\pgfpathlineto{\pgfqpoint{0.101635in}{0.076667in}}%
\pgfusepath{stroke}%
\end{pgfscope}%
\begin{pgfscope}%
\pgfsetbuttcap%
\pgfsetmiterjoin%
\definecolor{currentfill}{rgb}{0.015686,0.321569,0.458824}%
\pgfsetfillcolor{currentfill}%
\pgfsetlinewidth{1.003750pt}%
\definecolor{currentstroke}{rgb}{0.015686,0.321569,0.458824}%
\pgfsetstrokecolor{currentstroke}%
\pgfsetdash{}{0pt}%
\pgfsys@defobject{currentmarker}{\pgfqpoint{-0.020833in}{-0.020833in}}{\pgfqpoint{0.020833in}{0.020833in}}{%
\pgfpathmoveto{\pgfqpoint{-0.020833in}{-0.020833in}}%
\pgfpathlineto{\pgfqpoint{0.020833in}{-0.020833in}}%
\pgfpathlineto{\pgfqpoint{0.020833in}{0.020833in}}%
\pgfpathlineto{\pgfqpoint{-0.020833in}{0.020833in}}%
\pgfpathlineto{\pgfqpoint{-0.020833in}{-0.020833in}}%
\pgfpathclose%
\pgfusepath{stroke,fill}%
}%
\begin{pgfscope}%
\pgfsys@transformshift{0.040523in}{0.076667in}%
\pgfsys@useobject{currentmarker}{}%
\end{pgfscope}%
\end{pgfscope}%
\begin{pgfscope}%
\definecolor{textcolor}{rgb}{0.101961,0.101961,0.098039}%
\pgfsetstrokecolor{textcolor}%
\pgfsetfillcolor{textcolor}%
\pgftext[x=0.134968in,y=0.037778in,left,base]{\color{textcolor}{\rmfamily\fontsize{8.000000}{9.600000}\selectfont\catcode`\^=\active\def^{\ifmmode\sp\else\^{}\fi}\catcode`\%=\active\def
\end{pgfscope}%
\begin{pgfscope}%
\pgfsetrectcap%
\pgfsetroundjoin%
\pgfsetlinewidth{1.003750pt}%
\definecolor{currentstroke}{rgb}{0.031373,0.564706,0.600000}%
\pgfsetstrokecolor{currentstroke}%
\pgfsetdash{}{0pt}%
\pgfpathmoveto{\pgfqpoint{0.718959in}{0.076667in}}%
\pgfpathlineto{\pgfqpoint{0.780070in}{0.076667in}}%
\pgfpathlineto{\pgfqpoint{0.841181in}{0.076667in}}%
\pgfusepath{stroke}%
\end{pgfscope}%
\begin{pgfscope}%
\pgfsetbuttcap%
\pgfsetmiterjoin%
\definecolor{currentfill}{rgb}{0.031373,0.564706,0.600000}%
\pgfsetfillcolor{currentfill}%
\pgfsetlinewidth{1.003750pt}%
\definecolor{currentstroke}{rgb}{0.031373,0.564706,0.600000}%
\pgfsetstrokecolor{currentstroke}%
\pgfsetdash{}{0pt}%
\pgfsys@defobject{currentmarker}{\pgfqpoint{-0.020833in}{-0.020833in}}{\pgfqpoint{0.020833in}{0.020833in}}{%
\pgfpathmoveto{\pgfqpoint{-0.020833in}{-0.020833in}}%
\pgfpathlineto{\pgfqpoint{0.020833in}{-0.020833in}}%
\pgfpathlineto{\pgfqpoint{0.020833in}{0.020833in}}%
\pgfpathlineto{\pgfqpoint{-0.020833in}{0.020833in}}%
\pgfpathlineto{\pgfqpoint{-0.020833in}{-0.020833in}}%
\pgfpathclose%
\pgfusepath{stroke,fill}%
}%
\begin{pgfscope}%
\pgfsys@transformshift{0.780070in}{0.076667in}%
\pgfsys@useobject{currentmarker}{}%
\end{pgfscope}%
\end{pgfscope}%
\begin{pgfscope}%
\definecolor{textcolor}{rgb}{0.101961,0.101961,0.098039}%
\pgfsetstrokecolor{textcolor}%
\pgfsetfillcolor{textcolor}%
\pgftext[x=0.874514in,y=0.037778in,left,base]{\color{textcolor}{\rmfamily\fontsize{8.000000}{9.600000}\selectfont\catcode`\^=\active\def^{\ifmmode\sp\else\^{}\fi}\catcode`\%=\active\def
\end{pgfscope}%
\end{pgfpicture}%
\makeatother%
\endgroup%

%% file: main_material/5_conclusion.tex
\section{Conclusion}
\label{sec:discussion}
In this work, we introduce LinQuant, a framework that exploits seven functional invariances for low-precision linear recurrences, and characterize their applicability across a wide range of recurrent mixers. These consist of {mixer-external changes of basis and mixer-internal absorptions}, namely alternative allocations of decay and write factors between operands. Our controlled study shows their benefits depend on recurrence structure, QK normalization, scale granularity, and the basis used to quantize the state. Notably, changes of basis and write-gate allocation are particularly effective under coarse scales. {These findings transfer to production Mamba-2 and Nemotron-H models, where LinQuant improves quantization quality. Moreover, on the evaluated Mamba-2 models, query--key rotation adds no measurable prefill latency.} Although our implementation focuses on integer prefill, the introduced invariances provide a broader design space (e.g. floating-point) for future low-precision implementations.\looseness-1

%% file: supplementary_material/A_extended_related_work.tex
\section{Extended related work}
\label{app:related-work}

\textbf{Linear recurrent and hybrid language models.}
A growing line of work replaces softmax attention with recurrent sequence mixers whose state size \citep{ess} and per-token decode cost do not grow with context length. This is achieved by implicitly summarizing past tokens in a fixed-size hidden state, instead of explicitly in a KV-cache~\citep{linear_attention,deltarule,gla}. Selective state-space models such as Mamba and Mamba-2 introduce input-dependent recurrent dynamics and hardware-efficient implementations~\citep{mamba,dao2024transformers}.

More expressive recurrences modify how information is retained and written into the hidden state. DeltaNet corrects stored associations by removing existing associations from its state~\citep{deltanet}, while Gated DeltaNet combines this correction mechanism with scalar forgetting~\citep{gdn}, and KDA uses channelwise decay~\citep{kda}. Preconditioned DeltaNet further incorporates curvature-aware updates~\citep{pdn}, while {DeltaProduct enhances} the state transition operator~\citep{deltaproduct}. In a recent development, large-scale hybrid models combine recurrent and softmax attention layers~\citep{kda,blakeman2025nemotron}.

Our work studies how the structure of these different recurrences enables and constrains functional invariances. We then leverage these invariances for post-training quantization (PTQ).

\textbf{PTQ of softmax Transformers.}
Weight-only PTQ methods such as GPTQ and AWQ reduce quantization error through reconstruction-based compensation or channel protection~\citep{gptq,awq}. For joint weight and activation quantization, SmoothQuant transfers activation range into weights~\citep{smoothquant}, while QServe combines quantization with efficient serving~\citep{qserve}. Rotation-based methods redistribute concentrated energy across channels: QuIP\# uses Hadamard incoherence processing for weight quantization~\citep{quip}, QuaRot extends rotations to weights, activations, and the KV cache~\citep{quarot}, and SpinQuant learns rotations from calibration data~\citep{spinquant}. LinQuant extends the study of functional invariances to linear recurrences. 

\textbf{Quantization of recurrent language models.}
Quamba develops an 8-bit PTQ recipe around sensitive selective-scan features and activation outliers~\citep{chiang2024quamba}. MambaQuant uses variance-aligned rotations~\citep{mambaquant}, while Quamba2 extends PTQ across Mamba and Mamba-2 at multiple precisions~\citep{quamba2}. 

SSDi8 targets low-precision execution inside the chunkwise Mamba-2 computation~\citep{kimssdi8}. It reuses quantized keys and queries, absorbs decay into values, and carries an INT8 state, all to maintain a persistent INT8 computational path. This decay placement is the value-side endpoint of our scalar-decay absorption family. Our production pipeline builds on SSDi8 by adding state-axis rotations and rescalings (Section~\ref{sec:experiments:production}).

\textbf{Recurrent-state compression beyond quantization.}
\citet{chahine2025curious} prune the hidden state during training by leveraging the free choice of key-axis basis, which we identify as invariance \textbf{(1)}. Similarly, \citet{nazari2026key} leverage the free choice of basis to prune the state of pretrained linear recurrences. Low-rank and sparse recurrent connectivity has also been studied outside language modeling~\citep{tumma}. In LinQuant, we instead retain state dimensions and use equivalent representations to improve quantization, addressing a complementary source of recurrent efficiency.

%% file: supplementary_material/B_experimental_setup.tex
\section{Experimental setup}
\label{app:experimental-setup}
We give the setup and timing methodology for the controlled study and production setting. Results are reported in Appendices~\ref{app:zoo-results} and~\ref{app:production-ablations}.

\subsection{Controlled study}
\label{app:zoo-details}
\input{tables/mixer_zoo}

\paragraph{Models and training.}
The 340M zoo contains fifteen recurrence variants, or cells: LA, Mamba-2, and GLA with and without write gating and QK normalization, plus DN, GDN, and KDA with QK normalization. All models use the FLA implementation~\citep{fla} and share a 24-layer architecture with a SwiGLU MLP, an output gate on the mixer, and tied embeddings. At 340M parameters, the model width is 1024 with 8 heads and {\(D_k=D_v=128\)}. At 1.3B, the width is 2048 with 16 heads and the same head dimensions.

Each cell is an instance of Equation~\eqref{eq:zoo-recurrence} with the decay and write rule of Table~\ref{tab:mixer-zoo}. Appendix~\ref{app:quantized-forms} gives the quantized chunkwise forms. QK normalization applies \(\ell_2\) normalization to \(\bb q_t\) and \(\bb k_t\) independently within each head.

Every model is trained from scratch on SlimPajama using FLA and a Llama tokenizer, with 15B tokens for the 340M models and 100B for the 1.3B models. All runs use sequence length 2048 and AdamW with peak learning rate \(4\times10^{-4}\), 1024 warm-up steps, and cosine decay to one tenth of the peak. We clip gradients at 1.

\paragraph{Quantization.}

Quantization is symmetric, with static scales calibrated by the maximum absolute activation over eight WikiText-2 training windows of 2048 tokens. {The state write, state read, and local-score contractions use INT8 operands with INT32 accumulation {(except the 16-token diagonal blocks of the local score under diagonal decay, which compute the decay-weighted dot products in FP32 using dequantized queries and keys)}; the intra-chunk output uses FP16 arithmetic on {INT8-quantized values}.} An FP32 state is carried between chunks, and reads consume an INT8 copy with one scale per head and value channel, calibrated on the carry. We call this copy the \emph{stored state}, and vary operand precision and stored-state precision separately. INT6 and INT4 configurations {restrict the integer range of the operands} within the same boundaries and kernels. The stored state remains INT8 unless stated otherwise. Chunk size is 64, with 16-token sub-blocks for diagonal decay, as in FLA.

\paragraph{Evaluation.}
We evaluate perplexity on WikiText-2 test at sequence length 2048. Transformation parameters are selected using cross-entropy on separate WikiText-2 samples. Excess NLL is \(\log(\mathrm{PPL}_{\mathrm{quant}}/\mathrm{PPL}_{\mathrm{FP}})\), and NLL recovered is the difference between the naive quantizer's NLL and that of the transformed configuration. We evaluate zero-shot accuracy with the Language Model Evaluation Harness~\citep{lmeval} on LAMBADA, HellaSwag, PIQA, ARC-Easy, ARC-Challenge, and WinoGrande.



\paragraph{Transformation selection.}

We use the diagonal rescaling parameterizations of Section~\ref{sec:change-of-basis} and sweep \(\kappa\in\{\tfrac14,\tfrac12,\tfrac34,1\}\), including both \(\RescaleTerm{\mat D_{QK}}\) alone and its composition with \(\RotTerm{\mat R_{QK}}\) on linear-write cells. The full \(\RescaleTerm{\mat D_V}\) sweep is evaluated under shared value scales. 

For decaying cells, we sweep \(\eta\in\{0,\tfrac14,\tfrac12,\tfrac34,1\}\). Linear write-factor absorption uses \(\xi\in\{1,\tfrac12,0\}\), while delta cells use \((\sigma,\tau)\in\{1,\tfrac12,0\}^2\).



We form the \emph{stack} by combining transformations whose selected settings improve held-out cross-entropy by at least \(0.003\) nats over the naive quantizer. We use canonical state storage unless stated otherwise, and retain the naive quantizer if no transformation qualifies.

\paragraph{Timing.}
\label{app:zoo-timing-setup}
We compare the transformation stack with canonical \(\RotTerm{\mat R_V}\) against naive INT8 on the A5000, A100, and H100 across batch sizes, with \(T=2048\), \(H=8\), and \(D_k=D_v=128\). We report ratios of median call times after autotuning and warm-up. Both calls include gate processing, operand quantization, and recurrent computation, including preparation for delta cells. Forward rotations performed before the call are excluded.

\subsection{Production setting}
\label{app:experimental-details}
We evaluate pretrained Mamba-2 and Nemotron-H models using SSDi8's W4A8 projection and INT8-recurrence pipeline. Unless otherwise stated, the matched baseline uses value-side decay absorption, static per-\((H,D_v)\) value scales, per-group INT8 keys and queries, and an INT8 fixed-point state pass.

\textbf{Calibration.} The controlled study calibrates activations on WikiText-2 (Appendix~\ref{app:zoo-details}). {In production, static activation scales and \(\RescaleTerm{\mat D_{QK}}\) are calibrated on \(512\) Pile sequences of \(512\) tokens for Mamba-2 and \(2048\) tokens for Nemotron-H. The activation scales are measured on the FP16 model with the projection-level Hadamard transformations applied; production weight quantization then uses \(128\) WikiText-2 training samples with GPTQ after fusing these transformations.} Each layer uses maximum absolute activation measurements with the same scale axes as the quantized kernel (Algorithm~\ref{alg:linquant}). In the recurrent core, we observe the raw operands \(\mat V,\mat K,\mat Q\), decay-weighted operands \(\DecayTerm{\Dtail_{[i]}}\odot\mat V_{[i]},\DecayTerm{\Dtail_{[i]}}\odot\mat K_{[i]}\), local state contribution \(\mat H\), and chunk product \(\mat Z=\mat Q\mat K^{\!\top}\).

For \(\RescaleTerm{\mat D_{QK}}\), we measure channel maxima of the Mamba-2 key and query streams, conventionally denoted \(\mat B\) and \(\mat C\), after convolution and SiLU. Rescaling is applied at this same point during inference. Moving it before the nonlinear preprocessing would not preserve the function.

\textbf{Evaluation.} We evaluate WikiText-2 test perplexity at sequence length \(2048\), using the standard tokenizer for each model: EleutherAI/gpt-neox-20b for Mamba-2 1.3B and 2.7B, and NVIDIA SentencePiece for Mamba-2 8B. Downstream evaluation uses the Language Model Evaluation Harness~\citep{lmeval} on SSDi8's six zero-shot tasks: LAMBADA, HellaSwag, PIQA, ARC-Easy, ARC-Challenge, and WinoGrande. For Mamba-2, rows with static-\((H,D_v)\) value-side scales run without SSDi8's output-projection mean correction, as do the tables marked ``no compensation''; the remaining Mamba-2 rows apply it with step factor \(0.15\), and the Nemotron-H rows with \(0.1\).

\paragraph{Timing.}
\label{app:timing-methodology}
Unless otherwise indicated, latency measurements use an NVIDIA RTX A5000 with batch size \(B=8\) and sequence length {\(T=2048\)}. For component measurements, we instantiate a single Mamba-2 mixer layer with random weights and time each kernel or sub-operation using CUDA events. Each run uses \(30\) warm-up forward passes for Triton autotuning and allocation, followed by \(200\) measured forward passes. We synchronize the stream after each measured forward pass, sum calls belonging to the same stage, and report the median stage latency in microseconds. 

%% file: tables/mixer_zoo.tex
\begin{table}[ht]
\centering
\small
\setlength{\tabcolsep}{3pt}
\renewcommand{\arraystretch}{1.45}
\caption{State-update rules for recurrent models in the mixer-zoo design space. Colors identify \DecayTerm{decay} and \WriteGateTerm{write gating}.}
\label{tab:mixer-zoo}
\begin{tabular}{ll}
\toprule
\multicolumn{1}{c}{model} & \multicolumn{1}{c}{update}\\
\midrule
Linear attention~\citep{linear_attention}
    & $\mat S_t = \mat S_{t-1} + \bb v_t \bb k_t^\top$ \\

Mamba-2~\citep{dao2024transformers}
    & $\mat S_t = \DecayTerm{\alpha_t}\, \mat S_{t-1} + \bb v_t \bb k_t^\top$ \\

Gated Linear Attention (GLA)~\citep{gla}
    & $\mat S_t = \mat S_{t-1}\, \DecayTerm{\operatorname{diag}(\bb\alpha_t)} + \bb v_t \bb k_t^\top$ \\

DeltaNet~\citep{deltarule,deltanet}
    & $\mat S_t = \mat S_{t-1}\, \bigl(\mat I - \WriteGateTerm{\beta_t} \bb k_t \bb k_t^\top\bigr) + \WriteGateTerm{\beta_t}\, \bb v_t \bb k_t^\top$ \\

Gated DeltaNet~\citep{gdn}
    & $\mat S_t = \DecayTerm{\alpha_t}\, \mat S_{t-1}\, \bigl(\mat I - \WriteGateTerm{\beta_t} \bb k_t \bb k_t^\top\bigr) + \WriteGateTerm{\beta_t}\, \bb v_t \bb k_t^\top$ \\

Kimi Delta Attention (KDA)~\citep{kda}
    & $\mat S_t = \mat S_{t-1}\, \DecayTerm{\operatorname{diag}(\bb\alpha_t)}\, \bigl(\mat I - \WriteGateTerm{\beta_t} \bb k_t \bb k_t^\top\bigr) + \WriteGateTerm{\beta_t}\, \bb v_t \bb k_t^\top$ \\
\bottomrule
\end{tabular}
\end{table}

%% file: supplementary_material/C_invariances.tex
\section{The seven invariances: statements and proofs}
\label{app:invariance-proofs}

We prove the invariances of Section~\ref{sec:method} for Equation~\eqref{eq:zoo-recurrence} in exact arithmetic, with positive decay and write gates and consistently transformed initial states. Table~\ref{tab:predictions} summarizes their applicability.

\paragraph{(1/7) Key-axis rotation \(\RotTerm{\mat R_{QK}}\).}
Let \(\RotTerm{\mat R}=\RotTerm{\mat R_{QK}}\) be orthogonal and commute with every \(\DecayTerm{\mat\Lambda_t}\). With \(\bb q'_t=\RotTerm{\mat R}\bb q_t\), \(\bb k'_t=\RotTerm{\mat R}\bb k_t\), and \(\mat S'_{t-1}=\mat S_{t-1}\RotTerm{\mat R\T}\), the update and readout satisfy
\begin{align*}
\mat S'_t
&=\bigl(\mat S_{t-1}\DecayTerm{\mat\Lambda_t}(\mat I-\delta\WriteGateTerm{\beta_t}\bb k_t\bb k_t\T)+\WriteGateTerm{\beta_t}\bb v_t\bb k_t\T\bigr)\RotTerm{\mat R\T}
=\mat S_t\RotTerm{\mat R\T},\\
\mat S'_t\bb q'_t&=\mat S_t\bb q_t.
\end{align*}
Commutation is automatic for absent or scalar decay. For fixed diagonal decay, \(\RotTerm{\mat R}\) must preserve the common decay eigenspaces. Transforming the decay itself would generally make it dense. Local rotations of both decay-weighted score operands remain valid and leave the carried state unchanged. A global key rotation preserves the state quantizer's scale axes but can change its calibrated maxima.

\paragraph{(2/7) Key-axis rescaling \(\RescaleTerm{\mat D_{QK}}\).}
Let \(\RescaleTerm{\mat D}=\RescaleTerm{\mat D_{QK}}\succ0\) be diagonal, with \(\bb q'_t=\RescaleTerm{\mat D}\bb q_t\), \(\bb k'_t=\RescaleTerm{\mat D^{-1}}\bb k_t\), and \(\mat S'_{t-1}=\mat S_{t-1}\RescaleTerm{\mat D^{-1}}\). Since \(\RescaleTerm{\mat D}\) commutes with diagonal decay, a linear write gives
\begin{equation*}
\mat S'_t=(\mat S_{t-1}\DecayTerm{\mat\Lambda_t}+\WriteGateTerm{\beta_t}\bb v_t\bb k_t\T)\RescaleTerm{\mat D^{-1}}
=\mat S_t\RescaleTerm{\mat D^{-1}},\qquad
\mat S'_t\bb q'_t=\mat S_t\bb q_t.
\end{equation*}
For a delta write, matching the erasure terms requires
\(\RescaleTerm{\mat D^{-2}}\bb k_t\bb k_t\T\RescaleTerm{\mat D^{-1}}=\bb k_t\bb k_t\T\RescaleTerm{\mat D^{-1}}\), or \(\RescaleTerm{\mat D^2}\bb k_t=\bb k_t\), excluding a general rescaling. The transformation acts after any QK normalization.

\paragraph{(3/7--4/7) Value-axis rotation and rescaling.}
For any invertible \(\mat T\), set \(\bb v'_t=\mat T\bb v_t\) and \(\mat S'_{t-1}=\mat T\mat S_{t-1}\). For either write rule and any diagonal decay,
\begin{equation*}
\mat S'_t=\mat T\bigl(\mat S_{t-1}\DecayTerm{\mat\Lambda_t}(\mat I-\delta\WriteGateTerm{\beta_t}\bb k_t\bb k_t\T)+\WriteGateTerm{\beta_t}\bb v_t\bb k_t\T\bigr)
=\mat T\mat S_t,\qquad \mat T^{-1}\bb o'_t=\bb o_t.
\end{equation*}
Choose \(\mat T=\RotTerm{\mat R_V}\) or \(\mat T=\RescaleTerm{\mat D_V}\). Canonical storage undoes \(\mat T\) on each chunk's write and local output. Delta pseudo-values additionally require the incoming-state correction in the transformed value basis. With per-channel scales, diagonal rescaling cancels because \((d_pv_p)/(d_ps_p)=v_p/s_p\), where \(s_p\) is the channel scale. Consistently transformed scales therefore leave quantization unchanged, apart from finite-precision rounding.

\paragraph{(5/7--7/7) Decay and linear-write absorptions.}
All three follow from the token-wise identity
\begin{equation*}
 c_{[i]}^r\bb v_{[i]}^r(\bb k_{[i]}^r)\T
=\bigl((c_{[i]}^r)^\eta\bb v_{[i]}^r\bigr)
 \bigl((c_{[i]}^r)^{1-\eta}\bb k_{[i]}^r\bigr)\T,
\qquad c_{[i]}^r>0.
\end{equation*}
Apply it to each chunk-write term with \(c_{[i]}^r\) equal to the scalar tail decay, the diagonal-decay magnitude \(\DecayTerm{e_{[i]}^r}\), or the write gate \(\WriteGateTerm{\beta_{[i]}^r}\). Other factors retain their placement. In particular, the relative channel-dependent decay remains on the key. For delta-rule decay absorption, use the corresponding pseudo-values in place of gated values.

\paragraph{(7/7) Delta write-factor absorption.}
Using the interval weights of Section~\ref{sec:background-chunkwise}, define the strictly lower-triangular Gram matrix
\begin{equation*}
 C_{[i],rs}=\mathbf 1_{s<r}\sum_n K_{[i],rn}K_{[i],sn}\DecayTerm{\gamma_{[i],n}^{s\to r}}.
\end{equation*}
Scalar decay is broadcast across channels. Without decay, \(\DecayTerm{\gamma}\equiv1\). The gate allocation is a diagonal similarity transformation:
\begin{align*}
\mat L_{[i]}^{(\sigma)}
&=\WriteGateTerm{\mat B_{[i]}^\sigma}\mat C_{[i]}\WriteGateTerm{\mat B_{[i]}^{1-\sigma}}
=\WriteGateTerm{\mat B_{[i]}^{\sigma-1}}\mat L_{[i]}^{(1)}\WriteGateTerm{\mat B_{[i]}^{1-\sigma}},\\
(\mat I+\mat L_{[i]}^{(\sigma)})^{-1}\WriteGateTerm{\mat B_{[i]}^\sigma}
&=\WriteGateTerm{\mat B_{[i]}^{\sigma-1}}(\mat I+\mat L_{[i]}^{(1)})^{-1}\WriteGateTerm{\mat B_{[i]}}.
\end{align*}
Writing \(\mat F_{[i]}:=\mat F_{[i]}^{(1)}\), the complete pseudo-values satisfy
\begin{equation*}
\mat F_{[i]}^{(\sigma)}
=(\mat I+\mat L_{[i]}^{(\sigma)})^{-1}\WriteGateTerm{\mat B_{[i]}^\sigma}
 \bigl(\mat V_{[i]}-(\DecayTerm{\Dpref_{[i]}}\odot\mat K_{[i]})\mat S_{[i]}\T\bigr)
=\WriteGateTerm{\mat B_{[i]}^{\sigma-1}}\mat F_{[i]}.
\end{equation*}
For the transformed operands of (7/7), the remaining gate powers sum to \(1-\sigma\), giving
\begin{align*}
\bigl(\mat V_{[i]}^{(\sigma,\tau)}\bigr)\T
 (\DecayTerm{\Dtail_{[i]}}\odot\mat K_{[i]}^{(\sigma,\tau)})
&=\bigl(\mat F_{[i]}^{(\sigma)}\bigr)\T\WriteGateTerm{\mat B_{[i]}^{1-\sigma}}
 (\DecayTerm{\Dtail_{[i]}}\odot\mat K_{[i]})\\
&=\mat F_{[i]}\T(\DecayTerm{\Dtail_{[i]}}\odot\mat K_{[i]}).
\end{align*}
The same token-wise cancellation preserves the local output. The incoming-state read is unchanged. At \((\sigma,\tau)=(1,1)\), this recovers the standard WY form~\citep{deltanet,gdn,fla}.

\begingroup
\paragraph{Aside: channel permutations.}
As an aside, we note that channel permutations within a head also preserve the recurrence. On the key axis, the rotation argument applies with a permutation matrix \(\mat P\) and consistently permuted decay \(\mat\Lambda'_t=\mat P\mat\Lambda_t\mat P\T\), which remains diagonal. This holds for both linear and delta writes. Value-axis permutations follow from the equivariance above, with the inverse applied to the output. We do not explore permutations in our study: under our per-tensor, per-head, and per-channel quantizers, they only reorder quantized entries and their associated scales, leaving quantization error unchanged. They can matter under groupwise quantization, where channel order determines which entries share a scale~\citep{zhang2026dampdecayawaremixedprecisionrecurrentstate}.
\par
\endgroup

%% file: supplementary_material/D_kernels.tex
\section{Low-precision realizations and kernel implementation}
\label{app:pseudocode}
This appendix describes the quantized chunkwise forms, the controlled and production kernels, and the placement and fusion of transformations. Figure~\ref{fig:architecture-overview} provides an architectural overview of these transformation locations.

\subsection{Quantized chunkwise forms}
\label{app:quantized-forms}
Section~\ref{sec:method:quantization} gives the scalar-decay linear recurrence underlying Mamba-2. The other five families retain its state write, state pass, state read, local score, and intra-chunk output, but form their operands differently. {Throughout this subsection, \(\mathcal Q\) denotes quantization at the indicated boundary, with scale factors implicit. The write, read, and local-score contractions use {INT8 operands} with INT32 accumulation {(except the 16-token diagonal blocks of the local score under diagonal decay, which compute the decay-weighted dot products in FP32 using dequantized queries and keys)}, with the corresponding scales applied to the accumulated products.} The intra-chunk output uses FP16 arithmetic on {INT8-quantized values}. 

\textbf{Decay factors.}
We use the interval weights \(\DecayTerm{\gamma_{[i]}^{s\to r}}\) and stacked prefix and tail weights \(\DecayTerm{\Dpref_{[i]}},\DecayTerm{\Dtail_{[i]}}\) from Section~\ref{sec:background-chunkwise}. 

\textbf{No decay, linear write (LA).} Setting \(\DecayTerm{\gamma}\equiv1\) removes all decay factors. With an optional write gate, the chunk computation becomes
\begin{align*}
    \widehat{\mat H}_{[i]}&=\mathcal Q(\WriteGateTerm{\mat B_{[i]}}\mat V_{[i]})^\top\mathcal Q(\mat K_{[i]}),\\
    \mat S_{[i+1]}&=\mat S_{[i]}+\widehat{\mat H}_{[i]},\qquad \widehat{\mat S}_{[i+1]}=\mathcal Q(\mat S_{[i+1]}),\\
    \widehat{\mat O}^{\mathrm{inter}}_{[i]}&=\mathcal Q(\mat Q_{[i]})\widehat{\mat S}_{[i]}^\top,\\
    \widehat{\mat A}_{[i]}&=\big(\mathcal Q(\mat Q_{[i]})\mathcal Q(\mat K_{[i]})^\top\big)\odot\mat M,\\
    \widehat{\mat O}^{\mathrm{intra}}_{[i]}&=\widehat{\mat A}_{[i]}\mathcal Q(\WriteGateTerm{\mat B_{[i]}}\mat V_{[i]}).
\end{align*}

\textbf{Diagonal decay, linear write (GLA).} For positive decay gates, the masked product in Section~\ref{sec:background-chunkwise} expands as
\begin{equation}
    \left[\tril\!\left((\DecayTerm{\Dpref_{[i]}}\odot\mat Q_{[i]})
    \left(\frac{\mat K_{[i]}}{\DecayTerm{\Dpref_{[i]}}}\right)^\top\right)\right]_{rs}
    =M_{rs}\sum_n Q_{[i],rn}K_{[i],sn}\DecayTerm{\gamma_{[i],n}^{s\to r}}.
\label{eq:local-score}
\end{equation}
The right-hand side also applies when decay gates are zero. Quantizing the contraction operands for positive gates gives
\begin{equation}
\begin{aligned}
    \widehat{\mat H}_{[i]}&=\mathcal Q(\WriteGateTerm{\mat B_{[i]}}\mat V_{[i]})^\top\mathcal Q(\DecayTerm{\Dtail_{[i]}}\odot\mat K_{[i]}),\\
    \mat S_{[i+1]}&=\mat S_{[i]}\DecayTerm{\diag(\gamma_{[i],1}^C,\ldots,\gamma_{[i],D_k}^C)}+\widehat{\mat H}_{[i]},\qquad \widehat{\mat S}_{[i+1]}=\mathcal Q(\mat S_{[i+1]}),\\
    \widehat{\mat O}^{\mathrm{inter}}_{[i]}&=\mathcal Q(\DecayTerm{\Dpref_{[i]}}\odot\mat Q_{[i]})\widehat{\mat S}_{[i]}^\top,\\
    \widehat A_{[i],rs}&=\sum_n\mathcal Q\big(Q_{[i],rn}\DecayTerm{\gamma_{[i],n}^r/\gamma_{[i],n}^a}\big)\,\mathcal Q\big(K_{[i],sn}\DecayTerm{\gamma_{[i],n}^a/\gamma_{[i],n}^s}\big),\\
    \widehat{\mat O}^{\mathrm{intra}}_{[i]}&=\widehat{\mat A}_{[i]}\mathcal Q(\WriteGateTerm{\mat B_{[i]}}\mat V_{[i]}).
\end{aligned}
\label{eq:quantized-chunk-gla}
\end{equation}
where the displayed local score applies to causal pairs \(s\le r\). Entries above the causal diagonal are zero. The decay lies inside the reduction over channels, as in Equation~\eqref{eq:local-score}. For cross-block scores, {the kernel anchors the two decay factors at the first position \(a\) of the block containing \(r\), so that both \(\DecayTerm{\gamma_{[i],n}^{r}/\gamma_{[i],n}^{a}}\) and \(\DecayTerm{\gamma_{[i],n}^{a}/\gamma_{[i],n}^{s}}\) are at most one}, then quantizes and multiplies the anchored operands. Within a block, {it uses quantized queries and keys} and applies \(\DecayTerm{\gamma_{[i],n}^{s\to r}}\) inside the FP32 sum. A query--key rotation can act on the anchored operands in registers, but cannot generally be applied once to the unweighted \(\mat Q\) and \(\mat K\) (Section~\ref{sec:method}).

\textbf{Delta write (DeltaNet, GDN, KDA).} Delta cells first construct pseudo-values that account for interactions among writes and with the incoming state. The Gram matrix, the triangular solve, and the WY intermediates are computed in FP32; only the incoming-state correction uses {an INT8-quantized copy of} \(\mat W_{[i]}\) against the INT8 copy of the incoming state. For absent or scalar decay, let \(\DecayTerm{\mat\Gamma_{[i]}}\) denote the causal decay mask. The compensated write-factor placement \((\sigma,\tau)\) gives
\begin{align*}
    \mat G_{[i]}&=\tril\!\Big(\big((\WriteGateTerm{\mat B_{[i]}^{\sigma}}\mat K_{[i]})\,(\WriteGateTerm{\mat B_{[i]}^{1-\sigma}}\mat K_{[i]})^\top\big)\odot\DecayTerm{\mat\Gamma_{[i]}},-1\Big),
    \qquad \mat T_{[i]}=(\mat I+\mat G_{[i]})^{-1},\\
    \mat W_{[i]}&=\mat T_{[i]}\WriteGateTerm{\mat B_{[i]}^{\sigma}}(\DecayTerm{\Dpref_{[i]}}\odot\mat K_{[i]}),
    \qquad \mat U_{[i]}=\mat T_{[i]}\,\WriteGateTerm{\mat B_{[i]}^{\sigma}}\mat V_{[i]},\\
    \widehat{\mat F}_{[i]}^{(\sigma)}&=\mat U_{[i]}-\mathcal Q(\mat W_{[i]})\widehat{\mat S}_{[i]}^\top,\\
    \mat V_{[i]}^{(\sigma,\tau)}&=\WriteGateTerm{\mat B_{[i]}^{(1-\sigma)(1-\tau)}}\widehat{\mat F}_{[i]}^{(\sigma)},
    \qquad \mat K_{[i]}^{(\sigma,\tau)}=\WriteGateTerm{\mat B_{[i]}^{(1-\sigma)\tau}}\mat K_{[i]},
\end{align*}
Here \(\mat U_{[i]},\mat W_{[i]}\) are WY intermediates, and \(\widehat{\mat F}_{[i]}^{(\sigma)}\) approximates the complete pseudo-values \(\mat F_{[i]}^{(\sigma)}\) using the quantized incoming-state correction. The transformed write operands \(\mat V_{[i]}^{(\sigma,\tau)},\mat K_{[i]}^{(\sigma,\tau)}\) enter the chunk step:
\begin{align*}
    \widehat{\mat H}_{[i]}&=\mathcal Q(\mat V_{[i]}^{(\sigma,\tau)})^\top\mathcal Q(\DecayTerm{\Dtail_{[i]}}\odot\mat K_{[i]}^{(\sigma,\tau)}),\\
    \mat S_{[i+1]}&=\mat S_{[i]}\DecayTerm{\mat P_{[i]}}+\widehat{\mat H}_{[i]},\qquad \widehat{\mat S}_{[i+1]}=\mathcal Q(\mat S_{[i+1]}),\\
    \widehat{\mat O}^{\mathrm{inter}}_{[i]}&={\DecayTerm{\Dpref_{[i]}}\odot\big(\mathcal Q(\mat Q_{[i]})\widehat{\mat S}_{[i]}^\top\big)},\\
    \widehat{\mat A}_{[i]}&=\big(\mathcal Q(\mat Q_{[i]})\mathcal Q(\mat K_{[i]}^{(\sigma,\tau)})^\top\big)\odot\DecayTerm{\mat\Gamma_{[i]}},\\
    \widehat{\mat O}^{\mathrm{intra}}_{[i]}&=\widehat{\mat A}_{[i]}\mathcal Q(\mat V_{[i]}^{(\sigma,\tau)}),
\end{align*}
where \(\DecayTerm{\mat P_{[i]}}\) is \(\mat I\) for DeltaNet, \(\DecayTerm{\gamma_{[i]}^C\mat I}\) for GDN, and \(\DecayTerm{\diag(\gamma_{[i],1}^C,\ldots,\gamma_{[i],D_k}^C)}\) for KDA. For KDA, both the Gram and local-score contractions place the channel-dependent decay inside the reduction, the local score using the anchored construction of Equation~\eqref{eq:quantized-chunk-gla} in place of a scalar mask applied after the product{, and the state read quantizes the decay-weighted query \(\mathcal Q(\DecayTerm{\Dpref_{[i]}}\odot\mat Q_{[i]})\)}. At \((\sigma,\tau)=(1,1)\), with no decay and quantization removed, \(\mat V_{[i]}^{(\sigma,\tau)}=\mat F_{[i]}\) recovers the pseudo-values in Section~\ref{sec:background-chunkwise}. Setting \(\sigma=\tfrac12\) instead assigns \(\WriteGateTerm{\mat B_{[i]}^{1/2}}\) to each Gram operand.

\textbf{Absorptions.} For a linear write with scalar decay, the tail weights \(\DecayTerm{\Dtail_{[i]}}\) and write gate \(\WriteGateTerm{\mat B_{[i]}}\) can each be split between the two operands before quantization. With elementwise powers of the tail vector and broadcasting across channels, the write is
\begin{equation*}
\mathcal Q\!\left(\DecayTerm{(\Dtail_{[i]})^{\eta}}\odot(\WriteGateTerm{\mat B_{[i]}^{\xi}}\mat V_{[i]})\right)^\top
\mathcal Q\!\left(\DecayTerm{(\Dtail_{[i]})^{1-\eta}}\odot(\WriteGateTerm{\mat B_{[i]}^{1-\xi}}\mat K_{[i]})\right).
\end{equation*}
Here \(\eta=1\) and \(\xi=1\) put the respective factors on the value, while zero puts them on the key. Under diagonal decay, only the scalar maximum \(\DecayTerm{e_{[i]}^r}=\max_n\DecayTerm{(\Dtail_{[i]})_{rn}}\) moves. The relative channel-dependent decay remains on the key. Delta cells use pseudo-values for decay absorption and the compensated \((\sigma,\tau)\) construction for write strength, as proved in Appendix~\ref{app:invariance-proofs}.

\subsection{Kernel implementation}

\paragraph{Controlled study.}
\label{app:zoo-kernels}
{Diagonal rescalings are applied to the operands before quantization, and absorptions redistribute token-wise factors already used by the write. Algorithm~\ref{alg:kernel} shows one chunk step for scalar or no decay with canonical state storage. Unused axis transformations are set to the identity. Matrices stack tokens as rows, so the rotations of Section~\ref{sec:change-of-basis} act by right multiplication with \(\RotTerm{\mat R\T}\).}

{Here, \(\mathcal Q_b\) quantizes operands to \(b\)-bit integers and \(\operatorname{deq}\) restores the associated scales. Scale factors not shown explicitly are folded into the surrounding operations. The full-precision carry is \(\mat S\), and \(\hat{\mat S}_{[i]}\) is its INT8 read copy. The tail weights are \(\DecayTerm{\bb g}\), the prefix weights \(\DecayTerm{\bb p}\), the causal decay mask \(\DecayTerm{\mat\Gamma}\), and the chunk decay \(\DecayTerm{\gamma^C}\).}

\input{figures/alg_kernel}

{The initial rescalings and rotations are fused with quantization; the output is multiplied by \(\RescaleTerm{\mat D_V^{-1}}\) to undo value rescaling. The local path and state write quantize their operands separately. The chosen write-gate allocation is used in the local score, intra-chunk output, and state write; the delta-rule operands from Algorithm~\ref{alg:delta} already include it. The state read, local score, and state write use INT8 operands with INT32 accumulation, while the intra-chunk output uses FP16 arithmetic on quantized values. The state is carried in FP32, with an INT8 copy used by the next chunk.}

{Decay weights are unity for LA and DeltaNet, scalars for Mamba-2 and GDN, and channel-dependent vectors for GLA and KDA. Under diagonal decay, only the scalar magnitude \(e_r=\max_n g_{rn}\) is redistributed: \(\bb w_k=\bb g\odot e^{-\eta}\) and \(\bb w_v=e^\eta\). Prefix decay is applied to the query before quantization in the state read, and the chunk decay becomes \(\diag(\gamma^C_n)\). The local score uses the construction in Appendix~\ref{app:quantized-forms}, including its FP32 computation within each 16-token block. Query--key rotations act only on the decay-weighted operands of Equation~\eqref{eq:quantized-chunk-gla}, before quantization.}

Canonical storage under \(\RotTerm{\mat R_V}\) adds matrix products inside the call. Each chunk's rotated write is multiplied by \(\RotTerm{\mat R_V^{\top}}\) before it joins the FP32 carry, and the local output is returned to the original basis. The state read stays canonical. {For canonical storage, delta cells right-multiply the incoming-state correction \(\mathcal Q(\mat W_{[i]})\widehat{\mat S}_{[i]}^\top\) by \(\mat R_V^\top\) before subtracting it from \(\mat U_{[i]}\), so both terms use the same rotated value basis.} {With rotated storage, the inverse rotations of the local output and state write are replaced by one \(\mat O\gets\mat O\mat R_V\) at the end of the chunk. This incurs the accuracy penalty discussed in Appendix~\ref{app:zoo-storage}.} The canonical delta kernels hold the full state and the \(D_v\times D_v\) rotation operand per program. 

\begin{algorithm}[H]
\caption{WY preparation with compensated write-factor placement \((\sigma,\tau)\), per head and chunk. \(\DecayTerm{\mat M_<}\) is the strictly lower causal mask, including scalar decay when present. Diagonal decay is applied inside the Gram reduction as described above. The Gram matrix, solve, and WY intermediates are FP32; \(\widehat{\mat S}_{\rm in}\) is the INT8-quantized incoming state and \(\mat W_{[i]}\) is the only quantized operand of the preparation. The resulting transformed write operands are used in the chunk step.}
\label{alg:delta}
\footnotesize
\begin{algorithmic}[1]
\State {\(\mat K^{\sigma}_{[i]}\gets\WriteGateTerm{\mat B_{[i]}^{\sigma}}\mat K_{[i]}\), \(\mat K^{1-\sigma}_{[i]}\gets\WriteGateTerm{\mat B_{[i]}^{1-\sigma}}\mat K_{[i]}\)} \Comment{Gram factors}
\State {\(\mat G_{[i]}\gets(\mat K^{\sigma}_{[i]}\,(\mat K^{1-\sigma}_{[i]})\T)\odot\DecayTerm{\mat M_<}\)} \Comment{Gram matrix, fp32}
\State {\(\mat T_{[i]}\gets(\mat I+\mat G_{[i]})^{-1}\)} \Comment{unit lower-triangular solve, fp32}
\State {\(\mat W_{[i]}\gets\mat T_{[i]}(\mat K^{\sigma}_{[i]}\odot\DecayTerm{\Dpref_{[i]}})\)}, \(\mat U_{[i]}\gets\mat T_{[i]}\,\WriteGateTerm{\mat B_{[i]}^{\sigma}}\mat V_{[i]}\) \Comment{WY intermediates, fp32}
\State {\(\mat C_{[i]}\gets\mathcal Q_b(\mat W_{[i]})\,\widehat{\mat S}_{\rm in}\T\)} \Comment{{incoming-state correction}}
\State {Canonical \(\mat R_V\): \(\mat C_{[i]}\gets\mat C_{[i]}\mat R_V\T\)} \Comment{{match the rotated value basis of \(\mat U_{[i]}\)}}
\State {\(\widehat{\mat F}_{[i]}^{(\sigma)}\gets\mat U_{[i]}-\mat C_{[i]}\)} \Comment{{complete pseudo-values}}
\State \(\mat K_{[i]}^{(\sigma,\tau)}\gets\WriteGateTerm{\mat B_{[i]}^{(1-\sigma)\tau}}\mat K_{[i]}\), \(\mat V_{[i]}^{(\sigma,\tau)}\gets\WriteGateTerm{\mat B_{[i]}^{(1-\sigma)(1-\tau)}}\widehat{\mat F}_{[i]}^{(\sigma)}\) \Comment{transformed write operands}
\State \Return \(\mat K_{[i]}^{(\sigma,\tau)},\mat V_{[i]}^{(\sigma,\tau)}\) for Algorithm~\ref{alg:kernel}
\end{algorithmic}
\end{algorithm}

\paragraph{Production Mamba-2.}
The production implementation adds state-axis transformations to SSDi8's dataflow, as summarized in Algorithm~\ref{alg:linquant}. {The added \(\RescaleTerm{\mat D_{QK}}\) rescaling is applied as a per-channel multiply inside the fused split-and-quantize pass over the convolution output, followed by \(\RotTerm{\mat R_{QK}}\) in the same pass.} Neither transformation requires a separate kernel launch or memory pass in this implementation.

\begin{algorithm}[H]
\caption{The Mamba-2 kernel of LinQuant: SSDi8's dataflow with \(\RotTerm{\mat R_{QK}}\) and \(\RescaleTerm{\mat D_{QK}}\). Keys and queries are shared by the \(G\) groups and read by the \(H\) heads. \(g(h)\) denotes the group of head \(h\). \(\DecayTerm{\bb w}\) denotes the within-chunk decay weights (tail factors for the write, prefix factors for the read), and \(\DecayTerm{\gamma_{[i]}^C}\) denotes the chunk decay.}
\label{alg:linquant}
\footnotesize
\begin{algorithmic}[1]
\Statex \textbf{Once per layer} (the fused pass over the convolution output)
\State {\(\mat K\gets\mat K\RescaleTerm{\mat D_{QK}^{-1}}\RotTerm{\mat R_{QK}\T}\), \(\mat Q\gets\mat Q\RescaleTerm{\mat D_{QK}}\RotTerm{\mat R_{QK}\T}\).} \(\hat{\mat K}\gets\mathcal Q_8(\mat K,\bb s_K)\), \(\hat{\mat Q}\gets\mathcal Q_8(\mat Q,\bb s_Q)\) \Comment{per group. {The quantized queries and keys are reused by three stages}}
\Statex \textbf{Per chunk \([i]\) and head \(h\)}
\State \(\hat{\mat V}\gets\mathcal Q_8(\DecayTerm{\bb w_{\mathrm{tail}}}\odot\mat V,\bb s_V)\), tile by tile in registers \Comment{decay on the value. Static per-\((H,D_v)\) scale or dynamic per tile. Key side: \(\hat{\mat K}\gets\mathcal Q_8(\DecayTerm{\bb w_{\mathrm{tail}}}\odot\mat K)\) from an fp re-read instead}
\State \(\mat H\gets\hat{\mat V}\T\hat{\mat K}_{g(h)}\) \Comment{write: INT8\(\times\)INT8, INT32 accumulation, one rescale}
\State \(\hat{\mat S}_{[i+1]}\gets\operatorname{fx}(\DecayTerm{\gamma_{[i]}^C}\hat{\mat S}_{[i]})+\mathcal Q_8(\mat H,\bb s_S)\) \Comment{state pass in INT8 fixed point, per-\((H,D_v)\) scale}
\State \(\hat{\mat Z}\gets\mathcal Q_8(\hat{\mat Q}_{g(h)}\hat{\mat K}_{g(h)}\T)\) \Comment{local score: INT32 accumulation, requantized per group}
\State \(\mat O\gets\DecayTerm{\bb w_{\mathrm{pref}}}\odot\operatorname{deq}(\hat{\mat Q}_{g(h)}\hat{\mat S}_{[i]}\T)+(\operatorname{deq}(\hat{\mat Z})\odot\DecayTerm{\mat\Gamma})\mat V\) \Comment{read on INT8. Intra-chunk output fp16, the scale of \(\hat{\mat Z}\) folded into \(\DecayTerm{\mat\Gamma}\)}
\end{algorithmic}
\end{algorithm}

\paragraph{Transformation placement and rotation fusion.}
\label{app:fusing-rotations}
We implement the online rotations using structured signed Hadamard transforms. Forward rotations are fused into kernels that already materialize the activations. The query--key rotation joins the split-and-quantize kernel after convolution. The value rotation joins value quantization. The inverse value rotation must restore the original basis before an incompatible downstream operation. In our Mamba-2 implementation, the gated RMSNorm after the recurrence normalizes across the full hidden dimension. We therefore apply the per-head inverse rotation at the end of the recurrence-output kernel, before this normalization.

\paragraph{Applying decay in registers.}
\label{app:decay-absorption-kernel}
Keys are shared by groups, while decay weights can differ across heads. A direct implementation of key-side absorption could therefore materialize a larger, head-indexed tensor of weighted keys. Our write kernel avoids this intermediate tensor. Each program loads its group's key tile and its head's decay weights, then forms the weighted entries in registers:
\begin{equation*}
    k'_{r,n} = \DecayTerm{w_{h,r}}\,k_{g(h),r,n},
\end{equation*}
The weighted tile is quantized and clipped before the INT8 tensor-core contraction. Programs for heads in the same group read the same underlying key tile and apply their own weights, so head-specific values never need to be stored as an expanded floating-point tensor.


%% file: figures/alg_kernel.tex
\begingroup
\setlength{\intextsep}{8pt}
\begin{algorithm}[H]
\caption{{One INT8 chunk step with LinQuant transformations (per head).}}
\label{alg:kernel}
\small
\begin{algorithmic}[1]
\State once: {\(\mat Q\gets\mat Q\RescaleTerm{\mat D_{QK}}\), \(\mat K\gets\mat K\RescaleTerm{\mat D_{QK}^{-1}}\), \(\mat V\gets\mat V\RescaleTerm{\mat D_V}\)} \Comment{{rescale operands}}
\State once: {\(\mat Q\gets\mat Q\RotTerm{\mat R_{QK}\T}\), \(\mat K\gets\mat K\RotTerm{\mat R_{QK}\T}\), \(\mat V\gets\mat V\RotTerm{\mat R_V\T}\)} \Comment{{rotate operands}}
\If{\(\delta=0\)}
\State {\((\mat K,\mat V)\gets(\WriteGateTerm{\mat B_{[i]}^{1-\xi}}\mat K,\ \WriteGateTerm{\mat B_{[i]}^{\xi}}\mat V)\)} \Comment{{write-gate absorption}}
\Else
\State \((\mat K,\mat V)\gets(\mat K_{[i]}^{(\sigma,\tau)},\mat V_{[i]}^{(\sigma,\tau)})\) from Algorithm~\ref{alg:delta}
\EndIf
\State \(\bb w_k\gets\DecayTerm{\bb g^{1-\eta}}\), \(\bb w_v\gets\DecayTerm{\bb g^\eta}\) \Comment{{decay absorption}}
\State \(\mat O\gets\DecayTerm{\bb p}\odot\operatorname{deq}\big(\mathcal Q_b(\mat Q)\hat{\mat S}_{[i]}\T\big)\) \Comment{{state read}}
\State \(\mat A\gets\operatorname{deq}\big(\mathcal Q_b(\mat Q)\mathcal Q_b(\mat K)\T\big)\odot\DecayTerm{\mat\Gamma}\) \Comment{{local score}}
\State {\(\mat O_{\rm loc}\gets\mat A\,\mathcal Q_b(\mat V)\)} \Comment{{intra-chunk output}}
\State {\(\mat O_{\rm loc}\gets\mat O_{\rm loc}\mat R_V\)} \Comment{{restore value basis}}
\State {\(\mat O\gets\mat O+\mat O_{\rm loc}\)} \Comment{{combine outputs}}
\State \(\hat{\mat K},\hat{\mat V}\gets\mathcal Q_b(\bb w_k\odot\mat K),\ \mathcal Q_b(\bb w_v\odot\mat V)\) \Comment{{quantize write operands}}
\State \(\mat H\gets\operatorname{deq}(\hat{\mat V}\T\hat{\mat K})\) \Comment{{state write}}
\State {\(\mat H\gets\RotTerm{\mat R_V\T}\mat H\)} \Comment{{restore value basis}}
\State \(\mat S\gets\mat S{\DecayTerm{\gamma^C}}+\mat H\). \(\hat{\mat S}_{[i+1]}\gets\mathcal Q_8(\mat S)\) \Comment{{carry and read copy}}
\end{algorithmic}
\end{algorithm}
\endgroup

%% file: supplementary_material/E_zoo_results.tex
\section{Controlled-study results}
\label{app:zoo-results}
This appendix reports the controlled-study results for Section~\ref{sec:experiments:quality}, using the setup in Appendix~\ref{app:zoo-details}. We organize them by key-axis transformations (1/7--2/7), value-axis transformations (3/7--4/7), and absorptions (5/7--7/7), followed by compact quality summaries, precision controls, and downstream accuracy. The complete numerical grids and per-task breakdowns are retained in the supplementary source.

\subsection{Individual transformations}

\paragraph{Key-axis transformations (1/7)-(2/7).}

\begin{figure}[H]
    \centering
    \input{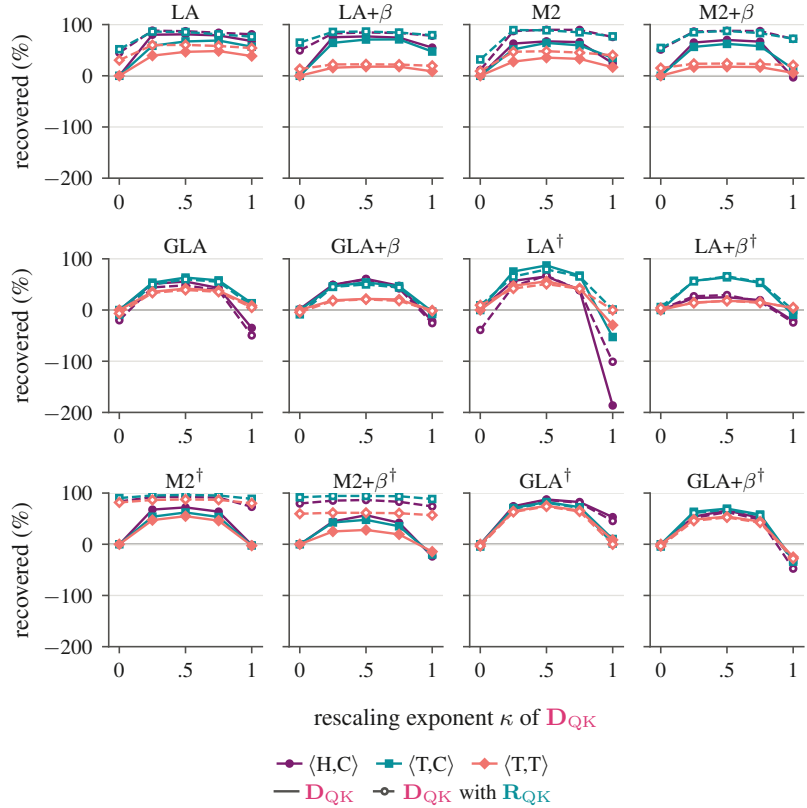}
    \caption{Response to the exponent \(\kappa\) of \(\RescaleTerm{\mat D_{QK}}\), alone (solid) and composed with \(\RotTerm{\mat R_{QK}}\) (dashed, hollow), per linear cell and regime, {INT6-quantized operands}, INT8 stored state: percent of the naive excess NLL recovered. At \(\kappa=0\) the dashed curve is \(\RotTerm{\mat R_{QK}}\) alone.}
    \label{fig:response_sqk}
\end{figure}

Figure~\ref{fig:response_sqk} shows the response to query--key rescaling, with separate panels for cells and curves for scale regimes. The query-key rescaling sweep tests both range balancing and its composition with rotation.

\paragraph{Value-axis transformations (3/7)-(4/7).}
\label{app:zoo-storage}

\begin{figure}[H]
    \centering
    \input{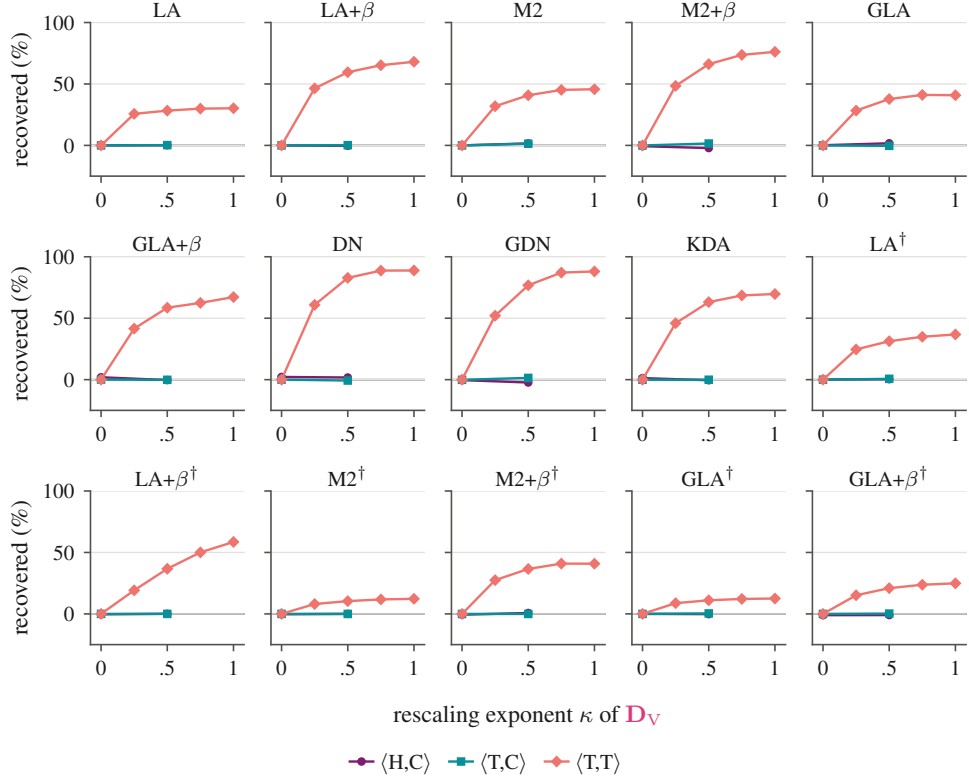}
    \caption{Response to the exponent \(\kappa\) of \(\RescaleTerm{\mat D_V}\), percent of the naive excess NLL recovered: per-channel scales absorb a consistent diagonal rescaling, whereas shared scales allow the exponent to change the error. The full grid \(\kappa\in\{\tfrac14,\tfrac12,\tfrac34,1\}\) was run only under coarse scales \(\langle\mathrm{T,T}\rangle\).}
    \label{fig:response_sv}
\end{figure}

Figure~\ref{fig:response_sv} shows the value-rescaling sweep, with separate panels for cells and curves for scale regimes. The value-rescaling sweep isolates the role of scale granularity: consistent per-channel scales absorb the rescaling, while shared scales allow it to change quantization error.

\begin{figure}[H]
\centering
\input{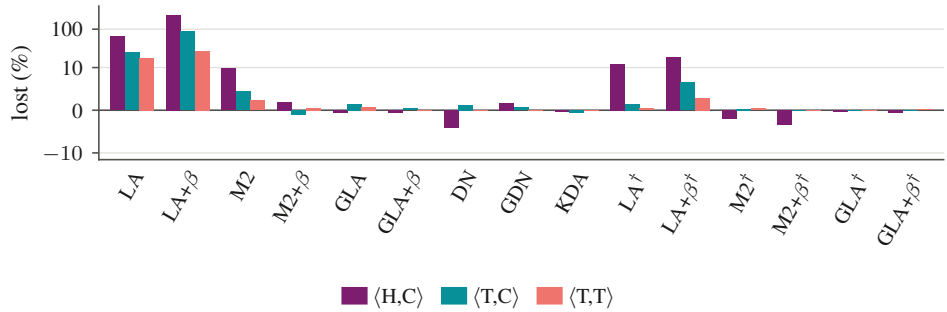}
\caption{Full storage-basis control for the 340M zoo. Excess NLL of rotated minus canonical storage under \(\RotTerm{\mat R_V}\), as a percentage of each model's naive excess NLL, per scale regime, {INT6-quantized operands}. Positive values favor canonical storage. Values use rounded per-configuration perplexities.}
\label{fig:storage_full}
\end{figure}

\paragraph{State-storage basis.}
Figure~\ref{fig:storage_full} compares canonical and rotated storage for every cell at INT6. The stored state uses one scale per head and value channel. A value rotation mixes these channels before quantization, changing the range covered by each scale. The largest benefit from canonical storage occurs in LA. Decaying and delta-rule cells are much less sensitive, although the sign and size of the difference depend on the configuration. 

\paragraph{Canonical value-rotation overhead.}
\label{app:zoo-timing}
Table~\ref{tab:zoo-rv-overhead} compares the transformation stack with canonical \(\RotTerm{\mat R_V}\) against naive INT8 across GPUs at a fixed batch size, complementing Figure~\ref{fig:overhead_rv}. The timing methodology and boundaries are given in Appendix~\ref{app:zoo-timing-setup}.

\begin{table}[ht]
\centering
\small
\setlength{\tabcolsep}{4pt}
\caption{Canonical value-rotation overhead across GPUs: median INT8 kernel time for the transformation stack with $\RotTerm{\mat R_V}$ relative to naive INT8 ($1\times$: equal time). $B=H=8$, $T=2048$, and $D_k=D_v=128$; methodology is in Appendix~\ref{app:zoo-timing-setup}.}
\label{tab:zoo-rv-overhead}
\begin{adjustbox}{max width=\textwidth}
\begin{tabular}{lrrr}
\toprule
Recurrence & RTX A5000 & A100-SXM4 & H100 \\
\midrule
LA & 1.61 & 1.24 & 1.60 \\
Mamba-2 & 1.42 & 1.42 & 1.60 \\
GLA & 1.36 & 1.72 & 1.59 \\
DN & 3.37 & 10.56 & 2.43 \\
GDN & 2.84 & 7.90 & 2.27 \\
KDA & 1.88 & 4.99 & 1.77 \\
\bottomrule
\end{tabular}
\end{adjustbox}
\end{table}

\paragraph{Absorptions (5/7)-(7/7).}

\begin{figure}[H]
    \centering
    \input{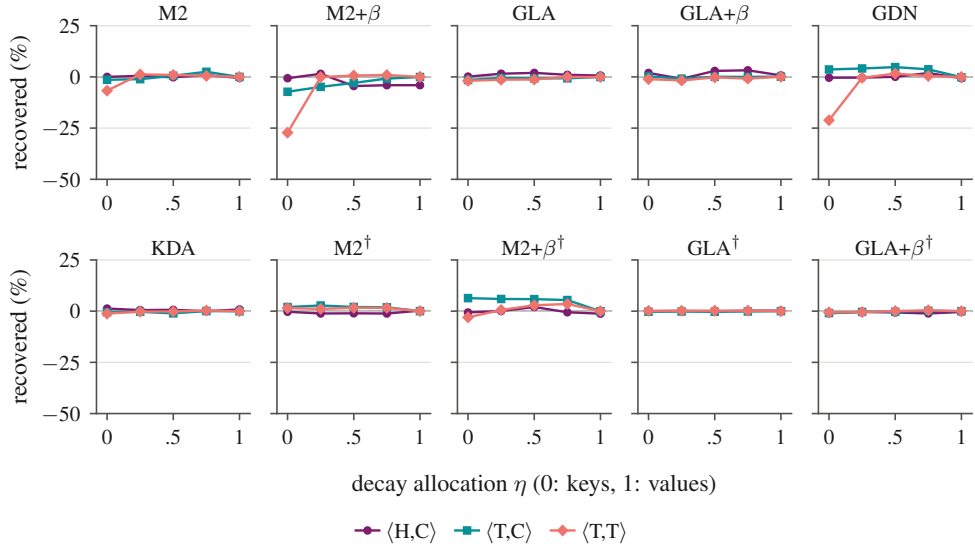}
    \caption{Response to the decay allocation \(\eta\) (decay on the key at 0, on the value at 1) for the decaying cells, INT6 operands and INT8 stored state: percent of the naive excess NLL recovered, one curve per scale regime. Placing scalar decay on the keys raises the excess by 21--27\,\% on M2\(+\WriteGateTerm{\beta}\) and GDN under \(\langle\mathrm{T,T}\rangle\). Diagonal decay and the per-channel-value regimes stay within a few percent.}
    \label{fig:response_eta}
\end{figure}

The absorption sweeps compare where token-wise factors enter the quantized computation. Scalar-decay placement has relatively small effects in the deployment-oriented regimes{ (Figure~\ref{fig:response_eta})}. For gated linear writes, {the value-side endpoint \(\xi=1\) performs best under \(\langle\mathrm{T,C}\rangle\), while the other regimes can favor an interior split}. For the delta cells under coarse scales, the selection procedure moves the compensation to the keys (\(\tau=1\)) and chooses \(\sigma\in\{0,\tfrac12\}\), with small differences between placements. The individual curves make clear how these conclusions depend on the cell and scale regime.

\begin{figure}[H]
    \centering
    \input{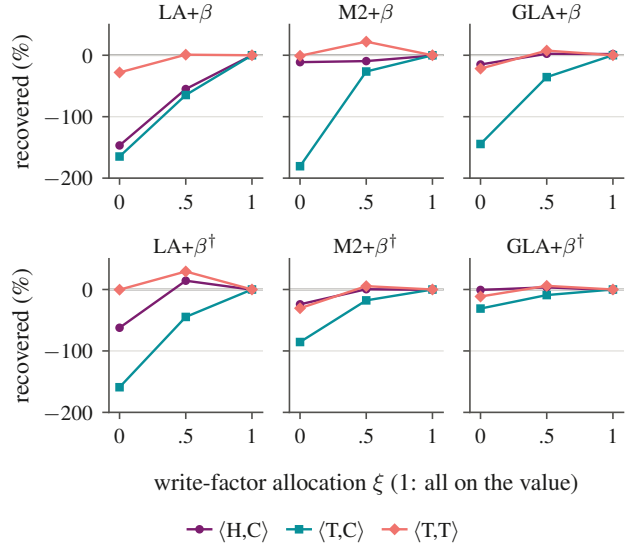}
    \caption{Write-factor allocation on the six gated linear models, with INT6 operands and INT8 stored state: \(\WriteGateTerm{\mat B_{[i]}^\xi}\) multiplies values and \(\WriteGateTerm{\mat B_{[i]}^{1-\xi}}\) multiplies keys. Percent of the naive excess NLL recovered. Each curve shows one scale regime. \(\dagger\) marks models without QK normalization.}
    \label{fig:response_rho}
\end{figure}

\begin{figure}[ht]
    \centering
    \input{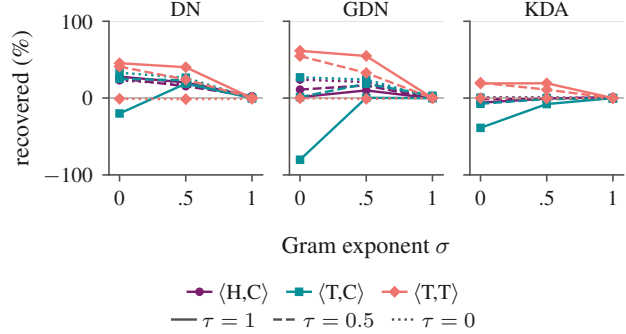}
    \caption{Delta write-factor allocation with INT6 operands and INT8 stored state. Curves vary the Gram exponent \(\sigma\) at fixed \(\tau\), which allocates the remaining factor between the transformed write operands: solid \(\tau=1\), dashed \(\tau=1/2\), dotted \(\tau=0\). Percent of the naive excess NLL recovered. Each colour denotes one scale regime.}
    \label{fig:response_sigtau}
\end{figure}

\begin{figure}[ht]
    \centering
    \includegraphics[width=0.46\linewidth]{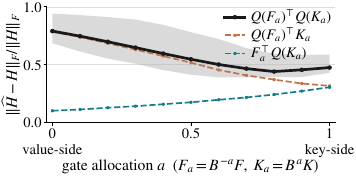}
    \caption{Write error of DeltaNet~340M against the gate allocation \(a\) (\(\sigma=1-a\), \(\tau=1\); \(F_a=B^{-a}F\), \(K_a=B^{a}K\) via the matched solve), with INT6 static per-tensor write scales recalibrated per allocation on the calibration split. Relative Frobenius error of the quantized write against the full-precision \(H=F^{\top}K\), quantizing both operands (solid), the corrected values only, or the keys only (dashed). Per-head medians over 256 held-out WikiText-2 chunks at layers 3, 9, 15 and 21; the curve is the median over all heads and the band the interquartile range. {Here, \(a=0\) is value-side absorption and \(a=1\) is key-side absorption.}}
    \label{fig:write_error_sweep}
\end{figure}

\subsection{Combined transformations and precision controls}
\begin{figure}[ht]
\centering
\input{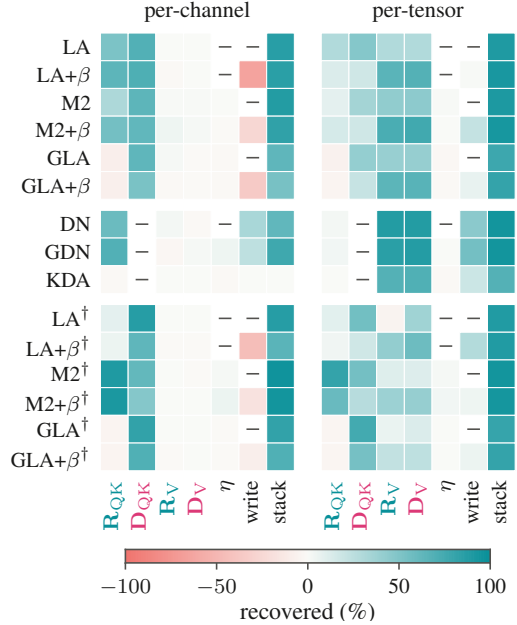}
\caption{Percentage of the naive quantizer's excess NLL recovered per cell and invariance at its selected setting (negative: the invariance increases the loss), shared keys with per-channel values (left) and with per-tensor values (right).}
\label{fig:heatmap}
\end{figure}

Figure~\ref{fig:heatmap} compares the two shared-key regimes at INT6. Key scales are held fixed and only value scales change, from per channel on the left to per tensor on the right. State-axis transformations can help in either panel, whereas the larger value-axis gains emerge when values share a scale.

\paragraph{Scale granularity and combined transformations.}
\label{app:zoo-tables}
Figures~\ref{fig:key_geometry} and~\ref{fig:val_geometry} isolate the effects of coarsening key and value scales. Tables~\ref{tab:zoo-quality-340m} and~\ref{tab:zoo-quality-1p3b} summarize INT6 quality at both model sizes, retaining every recurrence reported in the corresponding full grids. Coarse value scales particularly affect the delta-rule models; value-axis transformations recover much of this loss. Combining transformations can improve further, although the selected stack does not outperform every individual setting in every configuration.

Each individual invariance uses its held-out-selected parameters. ``Best single'' is the lowest reported test perplexity among these individual settings, including either storage basis for value rotation but excluding compositions. It is a descriptive oracle comparison, not a selection rule. The stack retains the original held-out selection and canonical storage.

\begin{table}[ht]
\centering
\small
\setlength{\tabcolsep}{4pt}
\caption{INT6 WikiText-2 perplexity for the 340M zoo. Keys use per-tensor scales; value scales are per channel $\langle\mathrm{T,C}\rangle$ or per tensor $\langle\mathrm{T,T}\rangle$. Best single is the descriptive comparison defined in the text; the stack is selected on held-out data and uses canonical storage. A dash denotes no reported promoted stack. $\dagger$: no QK normalization.}
\label{tab:zoo-quality-340m}
\begin{adjustbox}{max width=\textwidth}
\begin{tabular}{lrrrrrrr}
\toprule
 & & \multicolumn{3}{c}{$\langle\mathrm{T,C}\rangle$} & \multicolumn{3}{c}{$\langle\mathrm{T,T}\rangle$} \\
\cmidrule(lr){3-5}\cmidrule(lr){6-8}
Recurrence & FP & Naive & Best single & Stack & Naive & Best single & Stack \\
\midrule
LA & 17.117 & 25.706 & 19.428 & 18.097 & 30.694 & 23.151 & 18.138 \\
LA+$\WriteGateTerm{\beta}$ & 16.688 & 19.631 & 17.499 & 17.089 & 28.381 & 19.773 & 17.388 \\
Mamba-2 & 15.553 & 16.029 & 15.722 & 15.605 & 16.446 & 16.036 & 15.633 \\
Mamba-2+$\WriteGateTerm{\beta}$ & 15.420 & 15.595 & 15.485 & 15.449 & 16.230 & 15.609 & 15.473 \\
GLA & 15.104 & 15.651 & 15.302 & 15.302 & 16.048 & 15.646 & 15.309 \\
GLA+$\WriteGateTerm{\beta}$ & 15.308 & 15.540 & 15.418 & 15.415 & 16.046 & 15.547 & 15.440 \\
DeltaNet & 15.678 & 15.925 & 15.778 & 15.769 & 20.323 & 16.138 & 15.871 \\
GDN & 15.004 & 15.168 & 15.055 & 15.042 & 17.064 & 15.236 & 15.103 \\
KDA & 14.489 & 14.846 & 14.843 & -- & 15.837 & 14.885 & 14.906 \\
\midrule
LA$^{\dagger}$ & 17.230 & 24.916 & 18.079 & 18.071 & 31.582 & 22.486 & 18.336 \\
LA+$\WriteGateTerm{\beta}$$^{\dagger}$ & 16.980 & 30.719 & 21.042 & 20.738 & 100.247 & 35.441 & 20.429 \\
Mamba-2$^{\dagger}$ & 16.721 & 20.842 & 17.086 & 16.848 & 21.479 & 17.506 & 16.924 \\
Mamba-2+$\WriteGateTerm{\beta}$$^{\dagger}$ & 16.189 & 17.665 & 16.306 & 16.264 & 18.844 & 17.213 & 16.326 \\
GLA$^{\dagger}$ & 16.070 & 20.282 & 16.740 & 16.739 & 20.977 & 17.168 & 16.907 \\
GLA+$\WriteGateTerm{\beta}$$^{\dagger}$ & 15.891 & 19.222 & 16.834 & 16.838 & 20.523 & 17.873 & 16.647 \\
\bottomrule
\end{tabular}
\end{adjustbox}
\end{table}

\begin{table}[ht]
\centering
\small
\setlength{\tabcolsep}{4pt}
\caption{INT6 WikiText-2 perplexity for the 1.3B zoo. Keys use per-tensor scales; value scales are per channel $\langle\mathrm{T,C}\rangle$ or per tensor $\langle\mathrm{T,T}\rangle$. Best single is the descriptive comparison defined in the text; the stack is selected on held-out data and uses canonical storage. A dash denotes no reported promoted stack.}
\label{tab:zoo-quality-1p3b}
\begin{adjustbox}{max width=\textwidth}
\begin{tabular}{lrrrrrrr}
\toprule
 & & \multicolumn{3}{c}{$\langle\mathrm{T,C}\rangle$} & \multicolumn{3}{c}{$\langle\mathrm{T,T}\rangle$} \\
\cmidrule(lr){3-5}\cmidrule(lr){6-8}
Recurrence & FP & Naive & Best single & Stack & Naive & Best single & Stack \\
\midrule
LA+$\WriteGateTerm{\beta}$ & 10.276 & 12.101 & 10.768 & 10.625 & 148.173 & 12.588 & 10.842 \\
Mamba-2+$\WriteGateTerm{\beta}$ & 9.755 & 9.839 & 9.805 & 9.789 & 12.687 & 9.895 & 9.802 \\
GLA+$\WriteGateTerm{\beta}$ & 9.189 & 9.307 & 9.267 & 9.260 & 10.115 & 9.316 & 9.263 \\
DeltaNet & 9.917 & 10.050 & 9.979 & 9.973 & 46.639 & 10.406 & 10.111 \\
GDN & 9.727 & 9.784 & 9.751 & 9.752 & 11.401 & 9.867 & 9.788 \\
KDA & 9.460 & 9.603 & 9.593 & -- & 10.741 & 9.642 & 9.653 \\
\bottomrule
\end{tabular}
\end{adjustbox}
\end{table}

\paragraph{Precision sensitivity.}
Table~\ref{tab:recovery} contrasts INT8 and INT4. With per-channel value scales, several normalized decaying and delta-rule models are already close to full precision at INT8, leaving little room for a promoted transformation. At INT4, the stack can reduce perplexity substantially but does not consistently recover full-precision quality. The gap is particularly large with coarse value scales in the 1.3B zoo.

\begin{table}[ht]
\centering
\small
\setlength{\tabcolsep}{4pt}
\caption{Precision sensitivity on WikiText-2. INT8 uses per-channel value scales; INT4 includes both value-scale policies. Keys use per-tensor scales throughout. The stack is held-out-selected; dashes denote no reported promoted stack. Quantized settings retain the FP32 carry and INT8 read copy. $\dagger$: no QK normalization.}
\label{tab:recovery}
\begin{adjustbox}{max width=\textwidth}
\begin{tabular}{lrrrrrrr}
\toprule
 & & \multicolumn{2}{c}{INT8, $\langle\mathrm{T,C}\rangle$} & \multicolumn{2}{c}{INT4, $\langle\mathrm{T,C}\rangle$} & \multicolumn{2}{c}{INT4, $\langle\mathrm{T,T}\rangle$} \\
\cmidrule(lr){3-4}\cmidrule(lr){5-6}\cmidrule(lr){7-8}
Recurrence & FP & Naive & Stack & Naive & Stack & Naive & Stack \\
\midrule
\multicolumn{8}{@{}l}{\emph{340M}} \\
LA & 17.117 & 19.482 & 17.265 & 73.226 & 24.736 & 498.768 & 32.750 \\
LA+$\WriteGateTerm{\beta}$ & 16.688 & 17.541 & 16.798 & 83.120 & 22.495 & 726.570 & 73.118 \\
Mamba-2 & 15.553 & 15.612 & 15.561 & 25.812 & 16.427 & 85.441 & 17.141 \\
Mamba-2+$\WriteGateTerm{\beta}$ & 15.420 & 15.443 & -- & 21.059 & 16.070 & 55.010 & 16.748 \\
GLA & 15.104 & 15.124 & -- & 32.431 & 20.587 & 88.884 & 21.214 \\
GLA+$\WriteGateTerm{\beta}$ & 15.308 & 15.322 & -- & 24.896 & 18.766 & 70.122 & 19.410 \\
DeltaNet & 15.678 & 15.698 & -- & 24.523 & 17.685 & 168.077 & 20.072 \\
GDN & 15.004 & 15.019 & -- & 20.540 & 16.098 & 82.814 & 17.421 \\
KDA & 14.489 & 14.511 & -- & 28.732 & -- & 111.449 & 30.205 \\
LA$^{\dagger}$ & 17.230 & 17.797 & 17.388 & 314.039 & 37.950 & 436.488 & 66.355 \\
LA+$\WriteGateTerm{\beta}$$^{\dagger}$ & 16.980 & 18.744 & 17.902 & 515.217 & 69.456 & 663.576 & 94.058 \\
Mamba-2$^{\dagger}$ & 16.721 & 16.911 & 16.735 & 316.762 & 23.555 & 362.199 & 25.470 \\
Mamba-2+$\WriteGateTerm{\beta}$$^{\dagger}$ & 16.189 & 16.266 & 16.194 & 202.732 & 18.344 & 285.469 & 22.110 \\
GLA$^{\dagger}$ & 16.070 & 16.282 & 16.099 & 173.810 & 46.248 & 263.196 & 54.814 \\
GLA+$\WriteGateTerm{\beta}$$^{\dagger}$ & 15.891 & 16.009 & 15.935 & 172.781 & 42.062 & 244.264 & 45.162 \\
\midrule
\multicolumn{8}{@{}l}{\emph{1.3B}} \\
LA+$\WriteGateTerm{\beta}$ & 10.276 & 10.763 & 10.349 & 60.518 & 13.987 & 4185.731 & 44.904 \\
Mamba-2+$\WriteGateTerm{\beta}$ & 9.755 & 9.774 & -- & 11.767 & 10.185 & 295.661 & 10.971 \\
GLA+$\WriteGateTerm{\beta}$ & 9.189 & 9.205 & -- & 12.509 & 10.786 & 160.470 & 11.423 \\
DeltaNet & 9.917 & 9.934 & -- & 13.753 & 11.322 & 957.275 & 15.235 \\
GDN & 9.727 & 9.731 & -- & 11.916 & 10.329 & 424.986 & 16.016 \\
KDA & 9.460 & 9.467 & -- & 20.150 & 19.877 & 352.929 & 25.376 \\
\bottomrule
\end{tabular}
\end{adjustbox}
\end{table}

The write-scale control in Table~\ref{tab:int4-write-scales} shows no improvement from dynamic scaling in the tested configurations. These experiments use {4-bit integer ranges} within INT8 kernels; deployment below INT8 therefore requires both a more accurate quantization scheme and an efficient sub-byte implementation.

\input{tables/zoo_int4_write_scales}

\paragraph{State precision and write-scale controls.}
\label{app:boundaries}
Table~\ref{tab:boundaries} separates state-read error from error fed back through the carry. Keeping the read state in floating point reduces error most strongly in accumulating recurrences. Carrying the requantized state into subsequent updates adds error beyond that of an INT8 read copy. The INT6 comparison retains all 340M recurrences at a fixed scale policy.

\begin{table}[ht]
\centering
\small
\setlength{\tabcolsep}{4pt}
\caption{State-precision control at INT6 for all 340M recurrences, with per-head keys, per-channel values, and no invariance transforms. FP state leaves the state unquantized; INT8 read quantizes a read copy while retaining an FP32 carry; INT8 carry feeds the requantized state into subsequent updates. Entries are WikiText-2 perplexities. $\dagger$: no QK normalization.}
\label{tab:boundaries}
\begin{adjustbox}{max width=\textwidth}
\begin{tabular}{lrrrr}
\toprule
Recurrence & FP model & FP state & INT8 read & INT8 carry \\
\midrule
LA & 17.117 & 19.954 & 22.349 & 23.963 \\
LA+$\WriteGateTerm{\beta}$ & 16.688 & 17.373 & 18.209 & 19.118 \\
Mamba-2 & 15.553 & 15.763 & 15.795 & 15.914 \\
Mamba-2+$\WriteGateTerm{\beta}$ & 15.420 & 15.518 & 15.524 & 15.545 \\
GLA & 15.104 & 15.307 & 15.312 & 15.360 \\
GLA+$\WriteGateTerm{\beta}$ & 15.308 & 15.457 & 15.456 & 15.566 \\
DeltaNet & 15.678 & 15.870 & 15.867 & 16.012 \\
GDN & 15.004 & 15.120 & 15.122 & 15.153 \\
KDA & 14.489 & 14.760 & 14.764 & 14.861 \\
LA$^{\dagger}$ & 17.230 & 18.294 & 18.502 & 19.443 \\
LA+$\WriteGateTerm{\beta}$$^{\dagger}$ & 16.980 & 20.621 & 21.446 & 22.277 \\
Mamba-2$^{\dagger}$ & 16.721 & 17.106 & 17.115 & 17.328 \\
Mamba-2+$\WriteGateTerm{\beta}$$^{\dagger}$ & 16.189 & 16.384 & 16.391 & 16.577 \\
GLA$^{\dagger}$ & 16.070 & 18.087 & 18.091 & 18.326 \\
GLA+$\WriteGateTerm{\beta}$$^{\dagger}$ & 15.891 & 16.709 & 16.703 & 16.818 \\
\bottomrule
\end{tabular}
\end{adjustbox}
\end{table}

Refining write-key scales alone does not resolve the error from coarse read keys (Table~\ref{tab:zoo-controls}). At INT6 it recovers part of the loss in linear-write models, with much smaller effects in delta-rule models. INT32 local-score accumulation and the FP16 intra-chunk output product remain fixed in these controls.

\begin{table}[ht]
\centering
\small
\setlength{\tabcolsep}{4pt}
\caption{Write-key scale control at INT6 for all 340M recurrences, with an FP32 carry and INT8 read copy. Read keys keep per-tensor scales. Shared uses the same per-tensor keys for reads and writes; fine write uses a separate per-channel write-key quantizer. No invariance transforms are applied. Entries are WikiText-2 perplexities; $\dagger$ marks no QK normalization.}
\label{tab:zoo-controls}
\begin{adjustbox}{max width=\textwidth}
\begin{tabular}{lrrrrr}
\toprule
 & & \multicolumn{2}{c}{Per-channel values} & \multicolumn{2}{c}{Per-tensor values} \\
\cmidrule(lr){3-4}\cmidrule(lr){5-6}
Recurrence & FP & Shared & Fine write & Shared & Fine write \\
\midrule
LA & 17.117 & 25.706 & 22.422 & 30.694 & 26.380 \\
LA+$\WriteGateTerm{\beta}$ & 16.688 & 19.631 & 18.312 & 28.381 & 26.888 \\
Mamba-2 & 15.553 & 16.029 & 15.796 & 16.446 & 16.242 \\
Mamba-2+$\WriteGateTerm{\beta}$ & 15.420 & 15.595 & 15.536 & 16.230 & 16.161 \\
GLA & 15.104 & 15.651 & 15.325 & 16.048 & 15.678 \\
GLA+$\WriteGateTerm{\beta}$ & 15.308 & 15.540 & 15.463 & 16.046 & 15.952 \\
DeltaNet & 15.678 & 15.925 & 15.916 & 20.323 & 20.261 \\
GDN & 15.004 & 15.168 & 15.145 & 17.064 & 17.059 \\
KDA & 14.489 & 14.846 & 14.787 & 15.837 & 15.766 \\
LA$^{\dagger}$ & 17.230 & 24.916 & 20.597 & 31.582 & 25.976 \\
LA+$\WriteGateTerm{\beta}$$^{\dagger}$ & 16.980 & 30.719 & 29.583 & 100.247 & 92.325 \\
Mamba-2$^{\dagger}$ & 16.721 & 20.842 & 20.215 & 21.479 & 20.817 \\
Mamba-2+$\WriteGateTerm{\beta}$$^{\dagger}$ & 16.189 & 17.665 & 17.380 & 18.844 & 18.460 \\
GLA$^{\dagger}$ & 16.070 & 20.282 & 19.788 & 20.977 & 20.411 \\
GLA+$\WriteGateTerm{\beta}$$^{\dagger}$ & 15.891 & 19.222 & 18.182 & 20.523 & 19.410 \\
\bottomrule
\end{tabular}
\end{adjustbox}
\end{table}

\subsection{Downstream accuracy}
Tables~\ref{tab:zeroshot} and~\ref{tab:zeroshot-1p3b} retain LAMBADA and the six-task average for every reported recurrence. The stack largely closes the gap to full precision for 340M LA and gated LA. At 1.3B, gated LA recovers most of the LAMBADA accuracy lost to naive quantization, although a gap remains. The other normalized recurrent models change little under this INT8 setting.

\begin{table}[ht]
\centering
\small
\setlength{\tabcolsep}{4pt}
\caption{Zero-shot accuracy for the 340M zoo with INT8 operands and $\langle\mathrm{T,C}\rangle$ scales. The six-task average includes LAMBADA, HellaSwag, ARC-e, ARC-c, PIQA, and WinoGrande, with normalized accuracy for HellaSwag and ARC-c. A dash denotes an unreported stack evaluation. $\dagger$: no QK normalization.}
\label{tab:zeroshot}
\begin{adjustbox}{max width=\textwidth}
\begin{tabular}{lrrrrrr}
\toprule
 & \multicolumn{3}{c}{LAMBADA} & \multicolumn{3}{c}{Six-task average} \\
\cmidrule(lr){2-4}\cmidrule(lr){5-7}
Recurrence & FP & Naive & Stack & FP & Naive & Stack \\
\midrule
LA & 0.267 & 0.042 & 0.258 & 0.398 & 0.356 & 0.395 \\
LA+$\WriteGateTerm{\beta}$ & 0.283 & 0.095 & 0.268 & 0.400 & 0.367 & 0.398 \\
Mamba-2 & 0.324 & 0.310 & 0.316 & 0.410 & 0.409 & 0.409 \\
Mamba-2+$\WriteGateTerm{\beta}$ & 0.316 & 0.307 & -- & 0.413 & 0.412 & -- \\
GLA & 0.335 & 0.331 & -- & 0.419 & 0.417 & -- \\
GLA+$\WriteGateTerm{\beta}$ & 0.332 & 0.328 & -- & 0.416 & 0.415 & -- \\
DeltaNet & 0.273 & 0.283 & -- & 0.402 & 0.406 & -- \\
GDN & 0.361 & 0.349 & -- & 0.418 & 0.413 & -- \\
KDA & 0.341 & 0.339 & -- & 0.413 & 0.413 & -- \\
LA$^{\dagger}$ & 0.266 & 0.242 & 0.250 & 0.397 & 0.391 & 0.394 \\
LA+$\WriteGateTerm{\beta}$$^{\dagger}$ & 0.273 & 0.150 & 0.222 & 0.400 & 0.372 & 0.380 \\
Mamba-2$^{\dagger}$ & 0.290 & 0.283 & 0.286 & 0.402 & 0.401 & 0.401 \\
Mamba-2+$\WriteGateTerm{\beta}$$^{\dagger}$ & 0.304 & 0.299 & 0.298 & 0.405 & 0.405 & 0.404 \\
GLA$^{\dagger}$ & 0.302 & 0.301 & 0.302 & 0.406 & 0.406 & 0.405 \\
GLA+$\WriteGateTerm{\beta}$$^{\dagger}$ & 0.314 & 0.311 & 0.311 & 0.406 & 0.406 & 0.405 \\
\bottomrule
\end{tabular}
\end{adjustbox}
\end{table}

\begin{table}[ht]
\centering
\small
\setlength{\tabcolsep}{4pt}
\caption{Zero-shot accuracy for the 1.3B zoo with INT8 operands and $\langle\mathrm{T,C}\rangle$ scales. The six-task average includes LAMBADA, HellaSwag, ARC-e, ARC-c, PIQA, and WinoGrande, with normalized accuracy for HellaSwag and ARC-c. A dash denotes an unreported stack evaluation.}
\label{tab:zeroshot-1p3b}
\begin{adjustbox}{max width=\textwidth}
\begin{tabular}{lrrrrrr}
\toprule
 & \multicolumn{3}{c}{LAMBADA} & \multicolumn{3}{c}{Six-task average} \\
\cmidrule(lr){2-4}\cmidrule(lr){5-7}
Recurrence & FP & Naive & Stack & FP & Naive & Stack \\
\midrule
LA+$\WriteGateTerm{\beta}$ & 0.465 & 0.169 & 0.408 & 0.507 & 0.447 & 0.493 \\
Mamba-2+$\WriteGateTerm{\beta}$ & 0.480 & 0.469 & -- & 0.522 & 0.521 & -- \\
GLA+$\WriteGateTerm{\beta}$ & 0.531 & 0.518 & -- & 0.544 & 0.545 & -- \\
DeltaNet & 0.454 & 0.458 & -- & 0.505 & 0.508 & -- \\
GDN & 0.511 & 0.505 & -- & 0.535 & 0.532 & -- \\
KDA & 0.522 & 0.514 & -- & 0.540 & 0.538 & -- \\
\bottomrule
\end{tabular}
\end{adjustbox}
\end{table}

%% file: tables/zoo_int4_write_scales.tex
\begin{table}[ht]
\centering
\small
\caption{WikiText-2 perplexity for the static and dynamic per-chunk write-scale controls at INT4, with per-head read-key scales and INT8 stored states (340M models).}
\label{tab:int4-write-scales}
\begin{tabular}{lrr}
\toprule
Model & Static write scales & Dynamic write scales \\
\midrule
LA & 22.7 & 26.2 \\
Mamba-2 & 16.1 & 16.4 \\
GDN & 18.4 & 18.8 \\
\bottomrule
\end{tabular}
\end{table}

%% file: supplementary_material/F_production_results.tex
\section{Production results}
\label{app:production-ablations}
This appendix reports the production results for Section~\ref{sec:experiments:production}. Setup and timing methodology are given in Appendix~\ref{app:experimental-details}. We compare with the official SSDi8 release, then examine query--key transformations, value rotation, and decay allocation. Compact quality and downstream summaries follow.

\subsection{Ablations}

\paragraph{Comparison with the official SSDi8 release.}
\label{app:ablations}
Our implementation closely matches the official SSDi8 release in perplexity and prefill latency across the three Mamba-2 sizes (Table~\ref{tab:ssdi8-vs-valstatic}). Both use value-side absorption, static per-\((H,D_v)\) scales, and no recurrent rotation.

Small numerical differences remain in the fixed-point decay gate and FP32 rounding of the fused quantizers. For Nemotron-H, where no official SSDi8 implementation is available, the matched baseline uses our implementation of this setting.

\begin{table}[ht]
\centering
\caption{Mamba-2 baseline validation on WikiText-2. Quantized rows use W4A8 projections, value-side static-\((H,D_v)\) absorption, no rotation, and no compensation. Full-model prefill is measured on the A5000 at \(B=8\), \(T=2048\).}
\label{tab:ssdi8-vs-valstatic}
\small
\setlength{\tabcolsep}{6.0pt}
\renewcommand{\arraystretch}{1.05}
\begin{tabular}{lccc}
\toprule
Method & 1.3B & 2.7B & 8B \\
\midrule
\multicolumn{4}{l}{\emph{WikiText-2 perplexity}} \\
FP16           & 10.42 & 9.06 & 7.25 \\
SSDi8 W4A8 (official implementation)  & 10.923 & 9.430 & 7.629 \\
Value-side baseline (our re-implementation)  & 10.928 & 9.441 & 7.620 \\
\midrule
\multicolumn{4}{l}{\emph{Prefill latency (ms)}} \\
SSDi8 W4A8 (official implementation) & 616 & 1184 & 2563 \\
Value-side baseline (our re-implementation) & 609 & 1163 & 2542 \\
\bottomrule
\end{tabular}
\end{table}

\paragraph{Query--key rotation and diagonal rescaling.}
\label{app:production-qk}
Query--key rotation lowers perplexity at both decay endpoints and under both scale policies, with the largest gains at 8B (Table~\ref{tab:wt2-absorption-rotation-ablation-rand}). The endpoints remain close in quality after rotation.

The exponent sweep favors \(\kappa=\tfrac12\), with \(\tfrac14\) close behind (Table~\ref{tab:production-rescaling-controls}). Incorporating rescaling into the existing kernel leaves measured layer latency unchanged, whereas a separate elementwise operation adds overhead.

\input{tables/production_rescaling_controls}

\paragraph{Value rotation.}
\label{app:value-rotation}
{With full-precision projections and the production pipeline's per-channel value scales, storing the state in a value-rotated basis increases perplexity from 10.496 to 10.533 (Table~\ref{tab:production-value-storage}). Canonical storage gives 10.497, essentially matching the unrotated baseline.} 

\input{tables/production_value_storage}

\begin{figure}[H]
    \centering
    \includegraphics[height=1.7in]{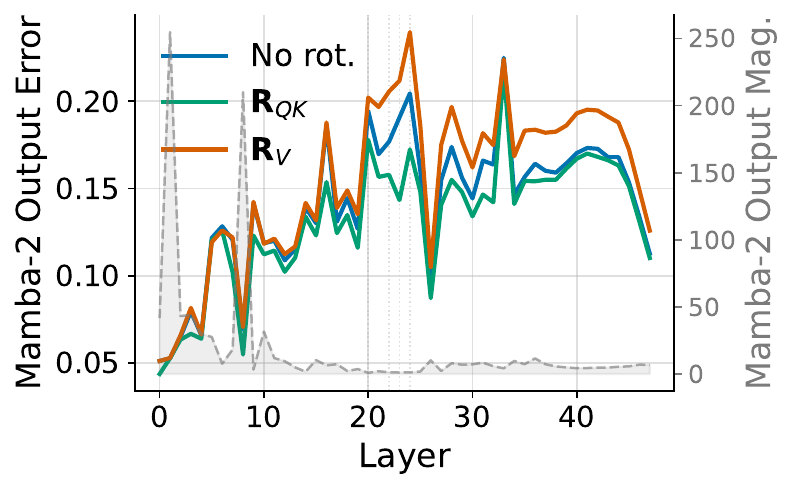}
    \caption{Per-layer relative error against the FP16 recurrence output, alongside its magnitude, for three rotation settings with rotated state storage. Value-rotation error grows in later layers with small recurrence outputs. Section~\ref{app:value-rotation} discusses the interaction with RMSNorm.}
    \label{fig:error_and_magnitude}
\end{figure}

Figure~\ref{fig:error_and_magnitude} shows that value-rotation error grows in later layers with small recurrence outputs. This suggests a role for the RMSNorm between the recurrence and output projection: normalizing a small output can amplify its quantization error before it enters the residual stream. Canonical storage avoids the measured perplexity penalty but still requires inverse transforms. Since it provides no quality gain under the production scale policy, LinQuant leaves the value axis unchanged.

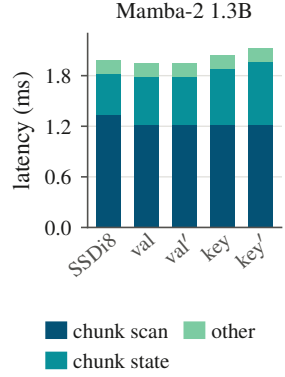
\begin{figure}[H]
    \centering
    \input{figures/pgf/ssd_breakdown_1.3b.pgf}
    \caption{Per-stage Mamba-2 1.3B recurrence latency at $B=8$, $T=2048$: {the SSDi8 setting (our matched value-side baseline)} and key-/value-side configurations with $\RotTerm{\mat R_{QK}}$ and static or dynamic ($'$) scales.}
    \label{fig:ssd_breakdown_1}
\end{figure}

\paragraph{Decay allocation.}
\label{app:production-decay}
We compare decay placement along two axes: quantization quality and execution cost. {Table~\ref{tab:wt2-absorption-rotation-ablation-rand} compares key- and value-side configurations under static and dynamic scales, with and without query--key rotation. Static value-side runs omit the output-projection mean correction used by the other configurations (Appendix~\ref{app:experimental-details}), so the static comparison does not isolate decay placement.} Figure~\ref{fig:ssd_breakdown_1} isolates the affected kernel stage, and Figure~\ref{fig:crossover} varies the operand dimensions.

\begin{table}[ht]
\centering
\caption{Mamba-2 W4A8 absorption endpoint and rotation ablation, WikiText-2 perplexity, mean\,$\pm$\,std over four calibration sets, random Hadamard $\RotTerm{\mat R_{QK}}$, without $\RescaleTerm{\mat D_{QK}}$. {Static value-side runs omit output-projection mean correction; the other configurations include it.}}
\label{tab:wt2-absorption-rotation-ablation-rand}
\scriptsize
\setlength{\tabcolsep}{3.0pt}
\renewcommand{\arraystretch}{1.05}
\begin{adjustbox}{max width=\textwidth}
\begin{tabular}{llcccccc}
\toprule
\multirow{2}{*}{Decay on} & \multirow{2}{*}{Granularity}
& \multicolumn{2}{c}{1.3B}
& \multicolumn{2}{c}{2.7B}
& \multicolumn{2}{c}{8B} \\
\cmidrule(lr){3-4}\cmidrule(lr){5-6}\cmidrule(lr){7-8}
&
& None & \(\RotTerm{\mat R_{QK}}\)
& None & \(\RotTerm{\mat R_{QK}}\)
& None & \(\RotTerm{\mat R_{QK}}\) \\
\midrule
Value & static-$(H,D_v)$
& 10.928\,{\scriptsize$\pm$0.004} & 10.860\,{\scriptsize$\pm$0.007}
& 9.441\,{\scriptsize$\pm$0.017} & 9.379\,{\scriptsize$\pm$0.011}
& 7.620\,{\scriptsize$\pm$0.009} & 7.456\,{\scriptsize$\pm$0.009} \\
Value & dynamic-$(H,1)$
& 10.894\,{\scriptsize$\pm$0.005} & 10.829\,{\scriptsize$\pm$0.012}
& 9.433\,{\scriptsize$\pm$0.006} & 9.371\,{\scriptsize$\pm$0.012}
& 7.642\,{\scriptsize$\pm$0.013} & 7.447\,{\scriptsize$\pm$0.008} \\
\midrule
Key & static-$(H,D_k)$
& 10.925\,{\scriptsize$\pm$0.003} & 10.849\,{\scriptsize$\pm$0.008}
& 9.424\,{\scriptsize$\pm$0.016} & 9.367\,{\scriptsize$\pm$0.011}
& 7.601\,{\scriptsize$\pm$0.009} & 7.455\,{\scriptsize$\pm$0.006} \\
Key & dynamic-$(H,1)$
& 10.884\,{\scriptsize$\pm$0.008} & 10.828\,{\scriptsize$\pm$0.010}
& 9.424\,{\scriptsize$\pm$0.011} & 9.360\,{\scriptsize$\pm$0.004}
& 7.634\,{\scriptsize$\pm$0.009} & 7.444\,{\scriptsize$\pm$0.009} \\
\bottomrule
\end{tabular}
\end{adjustbox}
\end{table}

The endpoints have similar perplexity under the tested scale policies (Table~\ref{tab:wt2-absorption-rotation-ablation-rand}). Without rotation, key-side absorption is slightly better in each comparison. Static scales are per channel on the absorbed operand, while dynamic scales are per \((\mathrm{chunk},H,1)\) and computed inside the kernel. These results support either placement on quality grounds, consistent with the limited sensitivity to scalar-decay placement at INT8 in the zoo.

Key-side absorption need not materialize a head-expanded tensor. Each thread block loads its group's keys and applies head-specific decay in registers (Appendix~\ref{app:decay-absorption-kernel}). The remaining difference is in the state write, which reloads the absorbed operand in floating point before weighting and quantizing it. Figure~\ref{fig:ssd_breakdown_1} shows that the other stages are unchanged.

\begin{figure}[ht]
    \centering
    \begin{subfigure}[t]{0.45\linewidth}\centering\input{figures/pgf/sweep_batch_1.3b.pgf}\end{subfigure}
    \begin{subfigure}[t]{0.45\linewidth}\centering\input{figures/pgf/sweep_seqlen_1.3b.pgf}\end{subfigure}
    \caption{Mamba-2 1.3B recurrence latency against batch size at $T=2048$ (left) and against sequence length at $B=32$ (right) for FP16, the official SSDi8 implementation, and the value- and key-side configurations with static scales and $\RotTerm{\mat R_{QK}}$. SSDi8 and the value-side configuration have similar latency once launch overhead no longer dominates.}
    \label{fig:sweep_1p3b}
\end{figure}
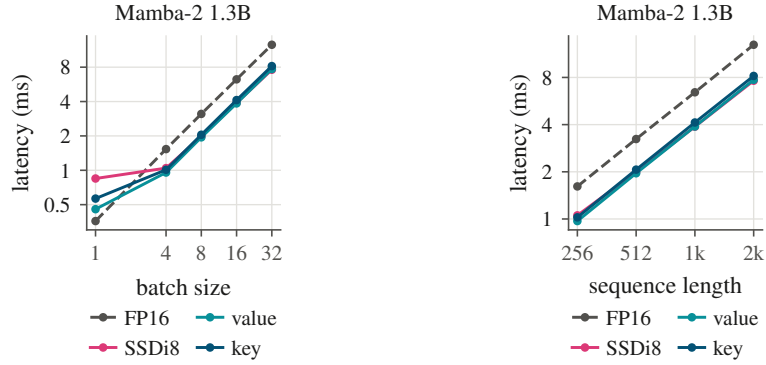

Value-side absorption is faster for the tested Mamba-2 geometry, where \(D_k=128\) and \(D_v=64\) (Figure~\ref{fig:ssd_breakdown_1}). Dynamic scales add cost on the key side because finding the maximum requires another pass over the weighted operand. Figure~\ref{fig:sweep_1p3b} extends the comparison across batch size and sequence length.

The geometry sweep in Figure~\ref{fig:crossover} shows that wider values can reverse the preferred endpoint at \(D_k=64\), while the key-side penalty persists across the tested widths at \(D_k=128\). Operand dimensions and the kernel's accumulation and rescaling layout therefore both influence the cost of decay placement.

\subsection{Quality and downstream summaries}
\label{app:full-results}
Table~\ref{tab:production-full} reports the static-scale configurations used in the main comparison, retaining calibration variability for Mamba-2 and both decay endpoints for the combined transformation. Dynamic-scale endpoint comparisons remain in Table~\ref{tab:wt2-absorption-rotation-ablation-rand}. The complete configuration grid and per-task breakdowns are retained in the supplementary source.

\begin{table}[ht]
\centering
\small
\setlength{\tabcolsep}{4pt}
\caption{Production quality with static scales and W4A8 projections. Perplexity is mean\,$\pm$\,standard deviation {over four calibration sets for Mamba-2 (rows without a $\pm$ are single runs) and a single run for Nemotron-H.} Avg. is the six-task average from the default calibration run. Value and Key denote the decay-absorption endpoint, with per-channel scales on that operand. The official SSDi8 implementation is available only for Mamba-2.}
\label{tab:production-full}
\begin{adjustbox}{max width=\textwidth}
\begin{tabular}{lrr}
\toprule
Configuration & WikiText-2 PPL & Avg. \\
\midrule
\multicolumn{3}{@{}l}{\emph{Mamba-2 1.3B}} \\
FP16 & 10.420 & 0.595 \\
SSDi8 (official) & 10.923\,{\scriptsize$\pm$0.007} & 0.588 \\
Value: $\RotTerm{R_{QK}}$ & 10.860\,{\scriptsize$\pm$0.007} & 0.590 \\
Value: $\RotTerm{R_{QK}}+\RescaleTerm{D_{QK}}$ & 10.849\,{\scriptsize$\pm$0.011} & 0.588 \\
Key: $\RotTerm{R_{QK}}+\RescaleTerm{D_{QK}}$ & 10.830\,{\scriptsize$\pm$0.007} & 0.589 \\
\midrule
\multicolumn{3}{@{}l}{\emph{Mamba-2 2.7B}} \\
FP16 & 9.060 & 0.638 \\
SSDi8 (official) & 9.430 & 0.626 \\
Value: $\RotTerm{R_{QK}}$ & 9.379\,{\scriptsize$\pm$0.011} & 0.629 \\
Value: $\RotTerm{R_{QK}}+\RescaleTerm{D_{QK}}$ & 9.383\,{\scriptsize$\pm$0.007} & 0.627 \\
Key: $\RotTerm{R_{QK}}+\RescaleTerm{D_{QK}}$ & 9.372\,{\scriptsize$\pm$0.003} & 0.630 \\
\midrule
\multicolumn{3}{@{}l}{\emph{Mamba-2 8B}} \\
FP16 & 7.250 & 0.707 \\
SSDi8 (official) & 7.629 & 0.696 \\
Value: $\RotTerm{R_{QK}}$ & 7.456\,{\scriptsize$\pm$0.009} & 0.703 \\
Value: $\RotTerm{R_{QK}}+\RescaleTerm{D_{QK}}$ & 7.435\,{\scriptsize$\pm$0.008} & 0.699 \\
Key: $\RotTerm{R_{QK}}+\RescaleTerm{D_{QK}}$ & 7.437\,{\scriptsize$\pm$0.004} & 0.697 \\
\midrule
\multicolumn{3}{@{}l}{\emph{Nemotron-H Base}} \\
FP16 & 6.030 & 0.758 \\
Value: $\RotTerm{R_{QK}}$ & 6.186 & 0.751 \\
Value: $\RotTerm{R_{QK}}+\RescaleTerm{D_{QK}}$ & 6.178 & 0.753 \\
Key: $\RotTerm{R_{QK}}+\RescaleTerm{D_{QK}}$ & 6.175 & 0.753 \\
\midrule
\multicolumn{3}{@{}l}{\emph{Nemotron-H Reasoning}} \\
FP16 & 8.423 & 0.731 \\
Value: $\RotTerm{R_{QK}}$ & 8.666 & 0.727 \\
Value: $\RotTerm{R_{QK}}+\RescaleTerm{D_{QK}}$ & 8.650 & 0.728 \\
Key: $\RotTerm{R_{QK}}+\RescaleTerm{D_{QK}}$ & 8.670 & 0.728 \\
\bottomrule
\end{tabular}
\end{adjustbox}
\end{table}

Table~\ref{tab:production-zeroshot} summarizes LAMBADA and the six-task average for the rescaled configurations. The two decay endpoints have similar downstream accuracy. Perplexity improvements do not imply uniform gains in task accuracy, as the comparison with the unrescaled configurations in Table~\ref{tab:production-full} shows.

\begin{table}[ht]
\centering
\small
\setlength{\tabcolsep}{4pt}
\caption{Downstream accuracy for the static production configurations with $\RotTerm{\mat R_{QK}}+\RescaleTerm{\mat D_{QK}}$. Both endpoints use per-channel scales on the absorbed operand. Mamba-2 uses the default calibration set. The six-task average follows Table~\ref{tab:production-full}.}
\label{tab:production-zeroshot}
\begin{adjustbox}{max width=\textwidth}
\begin{tabular}{lrrrr}
\toprule
 & \multicolumn{2}{c}{LAMBADA} & \multicolumn{2}{c}{Six-task average} \\
\cmidrule(lr){2-3}\cmidrule(lr){4-5}
Model & Value & Key & Value & Key \\
\midrule
Mamba-2 1.3B & 0.644 & 0.645 & 0.588 & 0.589 \\
Mamba-2 2.7B & 0.675 & 0.684 & 0.627 & 0.630 \\
Mamba-2 8B & 0.706 & 0.704 & 0.699 & 0.697 \\
Nemotron-H Base & 0.729 & 0.727 & 0.753 & 0.753 \\
Nemotron-H Reasoning & 0.660 & 0.660 & 0.728 & 0.728 \\
\bottomrule
\end{tabular}
\end{adjustbox}
\end{table}

\paragraph{Hadamard signs.}
Table~\ref{tab:production-signs} isolates deterministic versus random Hadamard signs under static value-side absorption. Random signs improve five of the six query--key rotation comparisons shown; the exception is the small WikiText-2 difference at 8B. The main experiments use random signs. Additional endpoint and scale combinations are retained in the full source grids.

\makeatletter
\setlength{\@fptop}{0pt}
\makeatother
\begin{table}[ht]
\centering
\small
\setlength{\tabcolsep}{4pt}
\caption{Hadamard-sign control with W4A8 projections, static value-side decay absorption, and no diagonal rescaling. Entries are single-calibration-run perplexities, distinct from the multi-calibration means above. Det. and Rand. apply deterministic and random-signed query--key Hadamard rotations.}
\label{tab:production-signs}
\begin{adjustbox}{max width=\textwidth}
\begin{tabular}{llrrr}
\toprule
Model & Dataset & No rotation & Det. & Rand. \\
\midrule
Mamba-2 1.3B & WikiText-2 & 10.930 & 10.882 & 10.857 \\
Mamba-2 1.3B & Pile & 7.415 & 7.397 & 7.382 \\
Mamba-2 2.7B & WikiText-2 & 9.420 & 9.400 & 9.374 \\
Mamba-2 2.7B & Pile & 6.572 & 6.557 & 6.555 \\
Mamba-2 8B & WikiText-2 & 7.619 & 7.463 & 7.467 \\
Mamba-2 8B & Pile & 6.293 & 6.213 & 6.206 \\
\bottomrule
\end{tabular}
\end{adjustbox}
\end{table}

%% file: tables/production_rescaling_controls.tex
\begin{table}[ht]
\centering
\small
\caption{Production query--key rescaling controls. Top: WikiText-2 perplexity with static value-side decay absorption. Bottom: Mamba-2 1.3B layer latency at \(B=1\), relative to the baseline without rescaling. A dash denotes an unreported absolute time.}
\label{tab:production-rescaling-controls}
\begin{tabular}{lrr}
\toprule
Rescaling exponent \(\kappa\) & 1.3B & 8B \\
\midrule
\(1/4\) & 10.858 & 7.435 \\
\(1/2\) & 10.849 & 7.435 \\
\(3/4\) & 10.866 & 7.441 \\
\midrule
Implementation & Latency (ms) & Relative latency \\
\midrule
Without rescaling & 1.498 & 1.00 \\
Fused rescaling & 1.498 & 1.00 \\
Separate elementwise rescaling & -- & 1.04 \\
\bottomrule
\end{tabular}
\end{table}

%% file: tables/production_value_storage.tex
\begin{table}[ht]
\centering
\small
\caption{{Value-rotation diagnostics on Mamba-2 1.3B: WikiText-2 perplexity with full-precision projections and an INT8 fixed-point state. Canonical storage is evaluated by fake quantization.}}
\label{tab:production-value-storage}
\begin{tabular}{lr}
\toprule
Transformation and state basis & FP projections \\
\midrule
No rotation & 10.496 \\
\(\RotTerm{\mat R_V}\), rotated storage & 10.533 \\
\(\RotTerm{\mat R_V}\), canonical storage & 10.497 \\
\bottomrule
\end{tabular}
\end{table}

%% file: figures/pgf/ssd_breakdown_1.3b.pgf
\begingroup%
\makeatletter%
\begin{pgfpicture}%
\pgfpathrectangle{\pgfpointorigin}{\pgfqpoint{1.460000in}{2.030000in}}%
\pgfusepath{use as bounding box, clip}%
\begin{pgfscope}%
\pgfsetbuttcap%
\pgfsetmiterjoin%
\definecolor{currentfill}{rgb}{1.000000,1.000000,1.000000}%
\pgfsetfillcolor{currentfill}%
\pgfsetlinewidth{0.000000pt}%
\definecolor{currentstroke}{rgb}{1.000000,1.000000,1.000000}%
\pgfsetstrokecolor{currentstroke}%
\pgfsetdash{}{0pt}%
\pgfpathmoveto{\pgfqpoint{0.000000in}{0.000000in}}%
\pgfpathlineto{\pgfqpoint{1.460000in}{0.000000in}}%
\pgfpathlineto{\pgfqpoint{1.460000in}{2.030000in}}%
\pgfpathlineto{\pgfqpoint{0.000000in}{2.030000in}}%
\pgfpathlineto{\pgfqpoint{0.000000in}{0.000000in}}%
\pgfpathclose%
\pgfusepath{fill}%
\end{pgfscope}%
\begin{pgfscope}%
\pgfsetbuttcap%
\pgfsetmiterjoin%
\definecolor{currentfill}{rgb}{1.000000,1.000000,1.000000}%
\pgfsetfillcolor{currentfill}%
\pgfsetlinewidth{0.000000pt}%
\definecolor{currentstroke}{rgb}{0.000000,0.000000,0.000000}%
\pgfsetstrokecolor{currentstroke}%
\pgfsetstrokeopacity{0.000000}%
\pgfsetdash{}{0pt}%
\pgfpathmoveto{\pgfqpoint{0.400000in}{0.800000in}}%
\pgfpathlineto{\pgfqpoint{1.410000in}{0.800000in}}%
\pgfpathlineto{\pgfqpoint{1.410000in}{1.810000in}}%
\pgfpathlineto{\pgfqpoint{0.400000in}{1.810000in}}%
\pgfpathlineto{\pgfqpoint{0.400000in}{0.800000in}}%
\pgfpathclose%
\pgfusepath{fill}%
\end{pgfscope}%
\begin{pgfscope}%
\definecolor{textcolor}{rgb}{0.321569,0.317647,0.305882}%
\pgfsetstrokecolor{textcolor}%
\pgfsetfillcolor{textcolor}%
\pgftext[x=0.510083in,y=0.779167in,right,top,rotate=40.000000]{\color{textcolor}\rmfamily\fontsize{8.000000}{9.600000}\selectfont SSDi8}%
\end{pgfscope}%
\begin{pgfscope}%
\definecolor{textcolor}{rgb}{0.321569,0.317647,0.305882}%
\pgfsetstrokecolor{textcolor}%
\pgfsetfillcolor{textcolor}%
\pgftext[x=0.707542in,y=0.779167in,right,top,rotate=40.000000]{\color{textcolor}\rmfamily\fontsize{8.000000}{9.600000}\selectfont val}%
\end{pgfscope}%
\begin{pgfscope}%
\definecolor{textcolor}{rgb}{0.321569,0.317647,0.305882}%
\pgfsetstrokecolor{textcolor}%
\pgfsetfillcolor{textcolor}%
\pgftext[x=0.905000in,y=0.779167in,right,top,rotate=40.000000]{\color{textcolor}\rmfamily\fontsize{8.000000}{9.600000}\selectfont val\(\displaystyle '\)}%
\end{pgfscope}%
\begin{pgfscope}%
\definecolor{textcolor}{rgb}{0.321569,0.317647,0.305882}%
\pgfsetstrokecolor{textcolor}%
\pgfsetfillcolor{textcolor}%
\pgftext[x=1.102458in,y=0.779167in,right,top,rotate=40.000000]{\color{textcolor}\rmfamily\fontsize{8.000000}{9.600000}\selectfont key}%
\end{pgfscope}%
\begin{pgfscope}%
\definecolor{textcolor}{rgb}{0.321569,0.317647,0.305882}%
\pgfsetstrokecolor{textcolor}%
\pgfsetfillcolor{textcolor}%
\pgftext[x=1.299917in,y=0.779167in,right,top,rotate=40.000000]{\color{textcolor}\rmfamily\fontsize{8.000000}{9.600000}\selectfont key\(\displaystyle '\)}%
\end{pgfscope}%
\begin{pgfscope}%
\pgfpathrectangle{\pgfqpoint{0.400000in}{0.800000in}}{\pgfqpoint{1.010000in}{1.010000in}}%
\pgfusepath{clip}%
\pgfsetrectcap%
\pgfsetroundjoin%
\pgfsetlinewidth{0.501875pt}%
\definecolor{currentstroke}{rgb}{0.886275,0.882353,0.866667}%
\pgfsetstrokecolor{currentstroke}%
\pgfsetdash{}{0pt}%
\pgfpathmoveto{\pgfqpoint{0.400000in}{0.800000in}}%
\pgfpathlineto{\pgfqpoint{1.410000in}{0.800000in}}%
\pgfusepath{stroke}%
\end{pgfscope}%
\begin{pgfscope}%
\pgfsetbuttcap%
\pgfsetroundjoin%
\definecolor{currentfill}{rgb}{0.321569,0.317647,0.305882}%
\pgfsetfillcolor{currentfill}%
\pgfsetlinewidth{0.602250pt}%
\definecolor{currentstroke}{rgb}{0.321569,0.317647,0.305882}%
\pgfsetstrokecolor{currentstroke}%
\pgfsetdash{}{0pt}%
\pgfsys@defobject{currentmarker}{\pgfqpoint{-0.034722in}{0.000000in}}{\pgfqpoint{-0.000000in}{0.000000in}}{%
\pgfpathmoveto{\pgfqpoint{-0.000000in}{0.000000in}}%
\pgfpathlineto{\pgfqpoint{-0.034722in}{0.000000in}}%
\pgfusepath{stroke,fill}%
}%
\begin{pgfscope}%
\pgfsys@transformshift{0.400000in}{0.800000in}%
\pgfsys@useobject{currentmarker}{}%
\end{pgfscope}%
\end{pgfscope}%
\begin{pgfscope}%
\definecolor{textcolor}{rgb}{0.101961,0.101961,0.098039}%
\pgfsetstrokecolor{textcolor}%
\pgfsetfillcolor{textcolor}%
\pgftext[x=0.177778in, y=0.762112in, left, base]{\color{textcolor}\rmfamily\fontsize{8.000000}{9.600000}\selectfont 0.0}%
\end{pgfscope}%
\begin{pgfscope}%
\pgfpathrectangle{\pgfqpoint{0.400000in}{0.800000in}}{\pgfqpoint{1.010000in}{1.010000in}}%
\pgfusepath{clip}%
\pgfsetrectcap%
\pgfsetroundjoin%
\pgfsetlinewidth{0.501875pt}%
\definecolor{currentstroke}{rgb}{0.886275,0.882353,0.866667}%
\pgfsetstrokecolor{currentstroke}%
\pgfsetdash{}{0pt}%
\pgfpathmoveto{\pgfqpoint{0.400000in}{1.063537in}}%
\pgfpathlineto{\pgfqpoint{1.410000in}{1.063537in}}%
\pgfusepath{stroke}%
\end{pgfscope}%
\begin{pgfscope}%
\pgfsetbuttcap%
\pgfsetroundjoin%
\definecolor{currentfill}{rgb}{0.321569,0.317647,0.305882}%
\pgfsetfillcolor{currentfill}%
\pgfsetlinewidth{0.602250pt}%
\definecolor{currentstroke}{rgb}{0.321569,0.317647,0.305882}%
\pgfsetstrokecolor{currentstroke}%
\pgfsetdash{}{0pt}%
\pgfsys@defobject{currentmarker}{\pgfqpoint{-0.034722in}{0.000000in}}{\pgfqpoint{-0.000000in}{0.000000in}}{%
\pgfpathmoveto{\pgfqpoint{-0.000000in}{0.000000in}}%
\pgfpathlineto{\pgfqpoint{-0.034722in}{0.000000in}}%
\pgfusepath{stroke,fill}%
}%
\begin{pgfscope}%
\pgfsys@transformshift{0.400000in}{1.063537in}%
\pgfsys@useobject{currentmarker}{}%
\end{pgfscope}%
\end{pgfscope}%
\begin{pgfscope}%
\definecolor{textcolor}{rgb}{0.101961,0.101961,0.098039}%
\pgfsetstrokecolor{textcolor}%
\pgfsetfillcolor{textcolor}%
\pgftext[x=0.177778in, y=1.025649in, left, base]{\color{textcolor}\rmfamily\fontsize{8.000000}{9.600000}\selectfont 0.6}%
\end{pgfscope}%
\begin{pgfscope}%
\pgfpathrectangle{\pgfqpoint{0.400000in}{0.800000in}}{\pgfqpoint{1.010000in}{1.010000in}}%
\pgfusepath{clip}%
\pgfsetrectcap%
\pgfsetroundjoin%
\pgfsetlinewidth{0.501875pt}%
\definecolor{currentstroke}{rgb}{0.886275,0.882353,0.866667}%
\pgfsetstrokecolor{currentstroke}%
\pgfsetdash{}{0pt}%
\pgfpathmoveto{\pgfqpoint{0.400000in}{1.327075in}}%
\pgfpathlineto{\pgfqpoint{1.410000in}{1.327075in}}%
\pgfusepath{stroke}%
\end{pgfscope}%
\begin{pgfscope}%
\pgfsetbuttcap%
\pgfsetroundjoin%
\definecolor{currentfill}{rgb}{0.321569,0.317647,0.305882}%
\pgfsetfillcolor{currentfill}%
\pgfsetlinewidth{0.602250pt}%
\definecolor{currentstroke}{rgb}{0.321569,0.317647,0.305882}%
\pgfsetstrokecolor{currentstroke}%
\pgfsetdash{}{0pt}%
\pgfsys@defobject{currentmarker}{\pgfqpoint{-0.034722in}{0.000000in}}{\pgfqpoint{-0.000000in}{0.000000in}}{%
\pgfpathmoveto{\pgfqpoint{-0.000000in}{0.000000in}}%
\pgfpathlineto{\pgfqpoint{-0.034722in}{0.000000in}}%
\pgfusepath{stroke,fill}%
}%
\begin{pgfscope}%
\pgfsys@transformshift{0.400000in}{1.327075in}%
\pgfsys@useobject{currentmarker}{}%
\end{pgfscope}%
\end{pgfscope}%
\begin{pgfscope}%
\definecolor{textcolor}{rgb}{0.101961,0.101961,0.098039}%
\pgfsetstrokecolor{textcolor}%
\pgfsetfillcolor{textcolor}%
\pgftext[x=0.177778in, y=1.289187in, left, base]{\color{textcolor}\rmfamily\fontsize{8.000000}{9.600000}\selectfont 1.2}%
\end{pgfscope}%
\begin{pgfscope}%
\pgfpathrectangle{\pgfqpoint{0.400000in}{0.800000in}}{\pgfqpoint{1.010000in}{1.010000in}}%
\pgfusepath{clip}%
\pgfsetrectcap%
\pgfsetroundjoin%
\pgfsetlinewidth{0.501875pt}%
\definecolor{currentstroke}{rgb}{0.886275,0.882353,0.866667}%
\pgfsetstrokecolor{currentstroke}%
\pgfsetdash{}{0pt}%
\pgfpathmoveto{\pgfqpoint{0.400000in}{1.590612in}}%
\pgfpathlineto{\pgfqpoint{1.410000in}{1.590612in}}%
\pgfusepath{stroke}%
\end{pgfscope}%
\begin{pgfscope}%
\pgfsetbuttcap%
\pgfsetroundjoin%
\definecolor{currentfill}{rgb}{0.321569,0.317647,0.305882}%
\pgfsetfillcolor{currentfill}%
\pgfsetlinewidth{0.602250pt}%
\definecolor{currentstroke}{rgb}{0.321569,0.317647,0.305882}%
\pgfsetstrokecolor{currentstroke}%
\pgfsetdash{}{0pt}%
\pgfsys@defobject{currentmarker}{\pgfqpoint{-0.034722in}{0.000000in}}{\pgfqpoint{-0.000000in}{0.000000in}}{%
\pgfpathmoveto{\pgfqpoint{-0.000000in}{0.000000in}}%
\pgfpathlineto{\pgfqpoint{-0.034722in}{0.000000in}}%
\pgfusepath{stroke,fill}%
}%
\begin{pgfscope}%
\pgfsys@transformshift{0.400000in}{1.590612in}%
\pgfsys@useobject{currentmarker}{}%
\end{pgfscope}%
\end{pgfscope}%
\begin{pgfscope}%
\definecolor{textcolor}{rgb}{0.101961,0.101961,0.098039}%
\pgfsetstrokecolor{textcolor}%
\pgfsetfillcolor{textcolor}%
\pgftext[x=0.177778in, y=1.552724in, left, base]{\color{textcolor}\rmfamily\fontsize{8.000000}{9.600000}\selectfont 1.8}%
\end{pgfscope}%
\begin{pgfscope}%
\definecolor{textcolor}{rgb}{0.101961,0.101961,0.098039}%
\pgfsetstrokecolor{textcolor}%
\pgfsetfillcolor{textcolor}%
\pgftext[x=0.122222in,y=1.305000in,,bottom,rotate=90.000000]{\color{textcolor}\rmfamily\fontsize{9.000000}{10.800000}\selectfont latency (ms)}%
\end{pgfscope}%
\begin{pgfscope}%
\pgfpathrectangle{\pgfqpoint{0.400000in}{0.800000in}}{\pgfqpoint{1.010000in}{1.010000in}}%
\pgfusepath{clip}%
\pgfsetbuttcap%
\pgfsetmiterjoin%
\definecolor{currentfill}{rgb}{0.015686,0.321569,0.458824}%
\pgfsetfillcolor{currentfill}%
\pgfsetlinewidth{0.000000pt}%
\definecolor{currentstroke}{rgb}{0.000000,0.000000,0.000000}%
\pgfsetstrokecolor{currentstroke}%
\pgfsetstrokeopacity{0.000000}%
\pgfsetdash{}{0pt}%
\pgfpathmoveto{\pgfqpoint{0.445909in}{0.800000in}}%
\pgfpathlineto{\pgfqpoint{0.574257in}{0.800000in}}%
\pgfpathlineto{\pgfqpoint{0.574257in}{1.389860in}}%
\pgfpathlineto{\pgfqpoint{0.445909in}{1.389860in}}%
\pgfpathlineto{\pgfqpoint{0.445909in}{0.800000in}}%
\pgfpathclose%
\pgfusepath{fill}%
\end{pgfscope}%
\begin{pgfscope}%
\pgfpathrectangle{\pgfqpoint{0.400000in}{0.800000in}}{\pgfqpoint{1.010000in}{1.010000in}}%
\pgfusepath{clip}%
\pgfsetbuttcap%
\pgfsetmiterjoin%
\definecolor{currentfill}{rgb}{0.015686,0.321569,0.458824}%
\pgfsetfillcolor{currentfill}%
\pgfsetlinewidth{0.000000pt}%
\definecolor{currentstroke}{rgb}{0.000000,0.000000,0.000000}%
\pgfsetstrokecolor{currentstroke}%
\pgfsetstrokeopacity{0.000000}%
\pgfsetdash{}{0pt}%
\pgfpathmoveto{\pgfqpoint{0.643368in}{0.800000in}}%
\pgfpathlineto{\pgfqpoint{0.771716in}{0.800000in}}%
\pgfpathlineto{\pgfqpoint{0.771716in}{1.332458in}}%
\pgfpathlineto{\pgfqpoint{0.643368in}{1.332458in}}%
\pgfpathlineto{\pgfqpoint{0.643368in}{0.800000in}}%
\pgfpathclose%
\pgfusepath{fill}%
\end{pgfscope}%
\begin{pgfscope}%
\pgfpathrectangle{\pgfqpoint{0.400000in}{0.800000in}}{\pgfqpoint{1.010000in}{1.010000in}}%
\pgfusepath{clip}%
\pgfsetbuttcap%
\pgfsetmiterjoin%
\definecolor{currentfill}{rgb}{0.015686,0.321569,0.458824}%
\pgfsetfillcolor{currentfill}%
\pgfsetlinewidth{0.000000pt}%
\definecolor{currentstroke}{rgb}{0.000000,0.000000,0.000000}%
\pgfsetstrokecolor{currentstroke}%
\pgfsetstrokeopacity{0.000000}%
\pgfsetdash{}{0pt}%
\pgfpathmoveto{\pgfqpoint{0.840826in}{0.800000in}}%
\pgfpathlineto{\pgfqpoint{0.969174in}{0.800000in}}%
\pgfpathlineto{\pgfqpoint{0.969174in}{1.332367in}}%
\pgfpathlineto{\pgfqpoint{0.840826in}{1.332367in}}%
\pgfpathlineto{\pgfqpoint{0.840826in}{0.800000in}}%
\pgfpathclose%
\pgfusepath{fill}%
\end{pgfscope}%
\begin{pgfscope}%
\pgfpathrectangle{\pgfqpoint{0.400000in}{0.800000in}}{\pgfqpoint{1.010000in}{1.010000in}}%
\pgfusepath{clip}%
\pgfsetbuttcap%
\pgfsetmiterjoin%
\definecolor{currentfill}{rgb}{0.015686,0.321569,0.458824}%
\pgfsetfillcolor{currentfill}%
\pgfsetlinewidth{0.000000pt}%
\definecolor{currentstroke}{rgb}{0.000000,0.000000,0.000000}%
\pgfsetstrokecolor{currentstroke}%
\pgfsetstrokeopacity{0.000000}%
\pgfsetdash{}{0pt}%
\pgfpathmoveto{\pgfqpoint{1.038284in}{0.800000in}}%
\pgfpathlineto{\pgfqpoint{1.166632in}{0.800000in}}%
\pgfpathlineto{\pgfqpoint{1.166632in}{1.333210in}}%
\pgfpathlineto{\pgfqpoint{1.038284in}{1.333210in}}%
\pgfpathlineto{\pgfqpoint{1.038284in}{0.800000in}}%
\pgfpathclose%
\pgfusepath{fill}%
\end{pgfscope}%
\begin{pgfscope}%
\pgfpathrectangle{\pgfqpoint{0.400000in}{0.800000in}}{\pgfqpoint{1.010000in}{1.010000in}}%
\pgfusepath{clip}%
\pgfsetbuttcap%
\pgfsetmiterjoin%
\definecolor{currentfill}{rgb}{0.015686,0.321569,0.458824}%
\pgfsetfillcolor{currentfill}%
\pgfsetlinewidth{0.000000pt}%
\definecolor{currentstroke}{rgb}{0.000000,0.000000,0.000000}%
\pgfsetstrokecolor{currentstroke}%
\pgfsetstrokeopacity{0.000000}%
\pgfsetdash{}{0pt}%
\pgfpathmoveto{\pgfqpoint{1.235743in}{0.800000in}}%
\pgfpathlineto{\pgfqpoint{1.364091in}{0.800000in}}%
\pgfpathlineto{\pgfqpoint{1.364091in}{1.332781in}}%
\pgfpathlineto{\pgfqpoint{1.235743in}{1.332781in}}%
\pgfpathlineto{\pgfqpoint{1.235743in}{0.800000in}}%
\pgfpathclose%
\pgfusepath{fill}%
\end{pgfscope}%
\begin{pgfscope}%
\pgfpathrectangle{\pgfqpoint{0.400000in}{0.800000in}}{\pgfqpoint{1.010000in}{1.010000in}}%
\pgfusepath{clip}%
\pgfsetbuttcap%
\pgfsetmiterjoin%
\definecolor{currentfill}{rgb}{0.031373,0.564706,0.600000}%
\pgfsetfillcolor{currentfill}%
\pgfsetlinewidth{0.000000pt}%
\definecolor{currentstroke}{rgb}{0.000000,0.000000,0.000000}%
\pgfsetstrokecolor{currentstroke}%
\pgfsetstrokeopacity{0.000000}%
\pgfsetdash{}{0pt}%
\pgfpathmoveto{\pgfqpoint{0.445909in}{1.389860in}}%
\pgfpathlineto{\pgfqpoint{0.574257in}{1.389860in}}%
\pgfpathlineto{\pgfqpoint{0.574257in}{1.600106in}}%
\pgfpathlineto{\pgfqpoint{0.445909in}{1.600106in}}%
\pgfpathlineto{\pgfqpoint{0.445909in}{1.389860in}}%
\pgfpathclose%
\pgfusepath{fill}%
\end{pgfscope}%
\begin{pgfscope}%
\pgfpathrectangle{\pgfqpoint{0.400000in}{0.800000in}}{\pgfqpoint{1.010000in}{1.010000in}}%
\pgfusepath{clip}%
\pgfsetbuttcap%
\pgfsetmiterjoin%
\definecolor{currentfill}{rgb}{0.031373,0.564706,0.600000}%
\pgfsetfillcolor{currentfill}%
\pgfsetlinewidth{0.000000pt}%
\definecolor{currentstroke}{rgb}{0.000000,0.000000,0.000000}%
\pgfsetstrokecolor{currentstroke}%
\pgfsetstrokeopacity{0.000000}%
\pgfsetdash{}{0pt}%
\pgfpathmoveto{\pgfqpoint{0.643368in}{1.332458in}}%
\pgfpathlineto{\pgfqpoint{0.771716in}{1.332458in}}%
\pgfpathlineto{\pgfqpoint{0.771716in}{1.585229in}}%
\pgfpathlineto{\pgfqpoint{0.643368in}{1.585229in}}%
\pgfpathlineto{\pgfqpoint{0.643368in}{1.332458in}}%
\pgfpathclose%
\pgfusepath{fill}%
\end{pgfscope}%
\begin{pgfscope}%
\pgfpathrectangle{\pgfqpoint{0.400000in}{0.800000in}}{\pgfqpoint{1.010000in}{1.010000in}}%
\pgfusepath{clip}%
\pgfsetbuttcap%
\pgfsetmiterjoin%
\definecolor{currentfill}{rgb}{0.031373,0.564706,0.600000}%
\pgfsetfillcolor{currentfill}%
\pgfsetlinewidth{0.000000pt}%
\definecolor{currentstroke}{rgb}{0.000000,0.000000,0.000000}%
\pgfsetstrokecolor{currentstroke}%
\pgfsetstrokeopacity{0.000000}%
\pgfsetdash{}{0pt}%
\pgfpathmoveto{\pgfqpoint{0.840826in}{1.332367in}}%
\pgfpathlineto{\pgfqpoint{0.969174in}{1.332367in}}%
\pgfpathlineto{\pgfqpoint{0.969174in}{1.585644in}}%
\pgfpathlineto{\pgfqpoint{0.840826in}{1.585644in}}%
\pgfpathlineto{\pgfqpoint{0.840826in}{1.332367in}}%
\pgfpathclose%
\pgfusepath{fill}%
\end{pgfscope}%
\begin{pgfscope}%
\pgfpathrectangle{\pgfqpoint{0.400000in}{0.800000in}}{\pgfqpoint{1.010000in}{1.010000in}}%
\pgfusepath{clip}%
\pgfsetbuttcap%
\pgfsetmiterjoin%
\definecolor{currentfill}{rgb}{0.031373,0.564706,0.600000}%
\pgfsetfillcolor{currentfill}%
\pgfsetlinewidth{0.000000pt}%
\definecolor{currentstroke}{rgb}{0.000000,0.000000,0.000000}%
\pgfsetstrokecolor{currentstroke}%
\pgfsetstrokeopacity{0.000000}%
\pgfsetdash{}{0pt}%
\pgfpathmoveto{\pgfqpoint{1.038284in}{1.333210in}}%
\pgfpathlineto{\pgfqpoint{1.166632in}{1.333210in}}%
\pgfpathlineto{\pgfqpoint{1.166632in}{1.627697in}}%
\pgfpathlineto{\pgfqpoint{1.038284in}{1.627697in}}%
\pgfpathlineto{\pgfqpoint{1.038284in}{1.333210in}}%
\pgfpathclose%
\pgfusepath{fill}%
\end{pgfscope}%
\begin{pgfscope}%
\pgfpathrectangle{\pgfqpoint{0.400000in}{0.800000in}}{\pgfqpoint{1.010000in}{1.010000in}}%
\pgfusepath{clip}%
\pgfsetbuttcap%
\pgfsetmiterjoin%
\definecolor{currentfill}{rgb}{0.031373,0.564706,0.600000}%
\pgfsetfillcolor{currentfill}%
\pgfsetlinewidth{0.000000pt}%
\definecolor{currentstroke}{rgb}{0.000000,0.000000,0.000000}%
\pgfsetstrokecolor{currentstroke}%
\pgfsetstrokeopacity{0.000000}%
\pgfsetdash{}{0pt}%
\pgfpathmoveto{\pgfqpoint{1.235743in}{1.332781in}}%
\pgfpathlineto{\pgfqpoint{1.364091in}{1.332781in}}%
\pgfpathlineto{\pgfqpoint{1.364091in}{1.664206in}}%
\pgfpathlineto{\pgfqpoint{1.235743in}{1.664206in}}%
\pgfpathlineto{\pgfqpoint{1.235743in}{1.332781in}}%
\pgfpathclose%
\pgfusepath{fill}%
\end{pgfscope}%
\begin{pgfscope}%
\pgfpathrectangle{\pgfqpoint{0.400000in}{0.800000in}}{\pgfqpoint{1.010000in}{1.010000in}}%
\pgfusepath{clip}%
\pgfsetbuttcap%
\pgfsetmiterjoin%
\definecolor{currentfill}{rgb}{0.486275,0.796078,0.635294}%
\pgfsetfillcolor{currentfill}%
\pgfsetlinewidth{0.000000pt}%
\definecolor{currentstroke}{rgb}{0.000000,0.000000,0.000000}%
\pgfsetstrokecolor{currentstroke}%
\pgfsetstrokeopacity{0.000000}%
\pgfsetdash{}{0pt}%
\pgfpathmoveto{\pgfqpoint{0.445909in}{1.600106in}}%
\pgfpathlineto{\pgfqpoint{0.574257in}{1.600106in}}%
\pgfpathlineto{\pgfqpoint{0.574257in}{1.668971in}}%
\pgfpathlineto{\pgfqpoint{0.445909in}{1.668971in}}%
\pgfpathlineto{\pgfqpoint{0.445909in}{1.600106in}}%
\pgfpathclose%
\pgfusepath{fill}%
\end{pgfscope}%
\begin{pgfscope}%
\pgfpathrectangle{\pgfqpoint{0.400000in}{0.800000in}}{\pgfqpoint{1.010000in}{1.010000in}}%
\pgfusepath{clip}%
\pgfsetbuttcap%
\pgfsetmiterjoin%
\definecolor{currentfill}{rgb}{0.486275,0.796078,0.635294}%
\pgfsetfillcolor{currentfill}%
\pgfsetlinewidth{0.000000pt}%
\definecolor{currentstroke}{rgb}{0.000000,0.000000,0.000000}%
\pgfsetstrokecolor{currentstroke}%
\pgfsetstrokeopacity{0.000000}%
\pgfsetdash{}{0pt}%
\pgfpathmoveto{\pgfqpoint{0.643368in}{1.585229in}}%
\pgfpathlineto{\pgfqpoint{0.771716in}{1.585229in}}%
\pgfpathlineto{\pgfqpoint{0.771716in}{1.653538in}}%
\pgfpathlineto{\pgfqpoint{0.643368in}{1.653538in}}%
\pgfpathlineto{\pgfqpoint{0.643368in}{1.585229in}}%
\pgfpathclose%
\pgfusepath{fill}%
\end{pgfscope}%
\begin{pgfscope}%
\pgfpathrectangle{\pgfqpoint{0.400000in}{0.800000in}}{\pgfqpoint{1.010000in}{1.010000in}}%
\pgfusepath{clip}%
\pgfsetbuttcap%
\pgfsetmiterjoin%
\definecolor{currentfill}{rgb}{0.486275,0.796078,0.635294}%
\pgfsetfillcolor{currentfill}%
\pgfsetlinewidth{0.000000pt}%
\definecolor{currentstroke}{rgb}{0.000000,0.000000,0.000000}%
\pgfsetstrokecolor{currentstroke}%
\pgfsetstrokeopacity{0.000000}%
\pgfsetdash{}{0pt}%
\pgfpathmoveto{\pgfqpoint{0.840826in}{1.585644in}}%
\pgfpathlineto{\pgfqpoint{0.969174in}{1.585644in}}%
\pgfpathlineto{\pgfqpoint{0.969174in}{1.653896in}}%
\pgfpathlineto{\pgfqpoint{0.840826in}{1.653896in}}%
\pgfpathlineto{\pgfqpoint{0.840826in}{1.585644in}}%
\pgfpathclose%
\pgfusepath{fill}%
\end{pgfscope}%
\begin{pgfscope}%
\pgfpathrectangle{\pgfqpoint{0.400000in}{0.800000in}}{\pgfqpoint{1.010000in}{1.010000in}}%
\pgfusepath{clip}%
\pgfsetbuttcap%
\pgfsetmiterjoin%
\definecolor{currentfill}{rgb}{0.486275,0.796078,0.635294}%
\pgfsetfillcolor{currentfill}%
\pgfsetlinewidth{0.000000pt}%
\definecolor{currentstroke}{rgb}{0.000000,0.000000,0.000000}%
\pgfsetstrokecolor{currentstroke}%
\pgfsetstrokeopacity{0.000000}%
\pgfsetdash{}{0pt}%
\pgfpathmoveto{\pgfqpoint{1.038284in}{1.627697in}}%
\pgfpathlineto{\pgfqpoint{1.166632in}{1.627697in}}%
\pgfpathlineto{\pgfqpoint{1.166632in}{1.699365in}}%
\pgfpathlineto{\pgfqpoint{1.038284in}{1.699365in}}%
\pgfpathlineto{\pgfqpoint{1.038284in}{1.627697in}}%
\pgfpathclose%
\pgfusepath{fill}%
\end{pgfscope}%
\begin{pgfscope}%
\pgfpathrectangle{\pgfqpoint{0.400000in}{0.800000in}}{\pgfqpoint{1.010000in}{1.010000in}}%
\pgfusepath{clip}%
\pgfsetbuttcap%
\pgfsetmiterjoin%
\definecolor{currentfill}{rgb}{0.486275,0.796078,0.635294}%
\pgfsetfillcolor{currentfill}%
\pgfsetlinewidth{0.000000pt}%
\definecolor{currentstroke}{rgb}{0.000000,0.000000,0.000000}%
\pgfsetstrokecolor{currentstroke}%
\pgfsetstrokeopacity{0.000000}%
\pgfsetdash{}{0pt}%
\pgfpathmoveto{\pgfqpoint{1.235743in}{1.664206in}}%
\pgfpathlineto{\pgfqpoint{1.364091in}{1.664206in}}%
\pgfpathlineto{\pgfqpoint{1.364091in}{1.735185in}}%
\pgfpathlineto{\pgfqpoint{1.235743in}{1.735185in}}%
\pgfpathlineto{\pgfqpoint{1.235743in}{1.664206in}}%
\pgfpathclose%
\pgfusepath{fill}%
\end{pgfscope}%
\begin{pgfscope}%
\pgfsetrectcap%
\pgfsetmiterjoin%
\pgfsetlinewidth{0.602250pt}%
\definecolor{currentstroke}{rgb}{0.321569,0.317647,0.305882}%
\pgfsetstrokecolor{currentstroke}%
\pgfsetdash{}{0pt}%
\pgfpathmoveto{\pgfqpoint{0.400000in}{0.800000in}}%
\pgfpathlineto{\pgfqpoint{0.400000in}{1.810000in}}%
\pgfusepath{stroke}%
\end{pgfscope}%
\begin{pgfscope}%
\pgfsetrectcap%
\pgfsetmiterjoin%
\pgfsetlinewidth{0.602250pt}%
\definecolor{currentstroke}{rgb}{0.321569,0.317647,0.305882}%
\pgfsetstrokecolor{currentstroke}%
\pgfsetdash{}{0pt}%
\pgfpathmoveto{\pgfqpoint{0.400000in}{0.800000in}}%
\pgfpathlineto{\pgfqpoint{1.410000in}{0.800000in}}%
\pgfusepath{stroke}%
\end{pgfscope}%
\begin{pgfscope}%
\definecolor{textcolor}{rgb}{0.101961,0.101961,0.098039}%
\pgfsetstrokecolor{textcolor}%
\pgfsetfillcolor{textcolor}%
\pgftext[x=0.905000in,y=1.893333in,,base]{\color{textcolor}\rmfamily\fontsize{8.500000}{10.200000}\selectfont Mamba-2 1.3B}%
\end{pgfscope}%
\begin{pgfscope}%
\pgfsetbuttcap%
\pgfsetmiterjoin%
\definecolor{currentfill}{rgb}{0.015686,0.321569,0.458824}%
\pgfsetfillcolor{currentfill}%
\pgfsetlinewidth{1.003750pt}%
\definecolor{currentstroke}{rgb}{0.015686,0.321569,0.458824}%
\pgfsetstrokecolor{currentstroke}%
\pgfsetdash{}{0pt}%
\pgfpathmoveto{\pgfqpoint{0.186559in}{0.225887in}}%
\pgfpathlineto{\pgfqpoint{0.286559in}{0.225887in}}%
\pgfpathlineto{\pgfqpoint{0.286559in}{0.303665in}}%
\pgfpathlineto{\pgfqpoint{0.186559in}{0.303665in}}%
\pgfpathlineto{\pgfqpoint{0.186559in}{0.225887in}}%
\pgfpathclose%
\pgfusepath{stroke,fill}%
\end{pgfscope}%
\begin{pgfscope}%
\definecolor{textcolor}{rgb}{0.101961,0.101961,0.098039}%
\pgfsetstrokecolor{textcolor}%
\pgfsetfillcolor{textcolor}%
\pgftext[x=0.325448in,y=0.225887in,left,base]{\color{textcolor}\rmfamily\fontsize{8.000000}{9.600000}\selectfont chunk scan}%
\end{pgfscope}%
\begin{pgfscope}%
\pgfsetbuttcap%
\pgfsetmiterjoin%
\definecolor{currentfill}{rgb}{0.031373,0.564706,0.600000}%
\pgfsetfillcolor{currentfill}%
\pgfsetlinewidth{1.003750pt}%
\definecolor{currentstroke}{rgb}{0.031373,0.564706,0.600000}%
\pgfsetstrokecolor{currentstroke}%
\pgfsetdash{}{0pt}%
\pgfpathmoveto{\pgfqpoint{0.186559in}{0.068499in}}%
\pgfpathlineto{\pgfqpoint{0.286559in}{0.068499in}}%
\pgfpathlineto{\pgfqpoint{0.286559in}{0.146277in}}%
\pgfpathlineto{\pgfqpoint{0.186559in}{0.146277in}}%
\pgfpathlineto{\pgfqpoint{0.186559in}{0.068499in}}%
\pgfpathclose%
\pgfusepath{stroke,fill}%
\end{pgfscope}%
\begin{pgfscope}%
\definecolor{textcolor}{rgb}{0.101961,0.101961,0.098039}%
\pgfsetstrokecolor{textcolor}%
\pgfsetfillcolor{textcolor}%
\pgftext[x=0.325448in,y=0.068499in,left,base]{\color{textcolor}\rmfamily\fontsize{8.000000}{9.600000}\selectfont chunk state}%
\end{pgfscope}%
\begin{pgfscope}%
\pgfsetbuttcap%
\pgfsetmiterjoin%
\definecolor{currentfill}{rgb}{0.486275,0.796078,0.635294}%
\pgfsetfillcolor{currentfill}%
\pgfsetlinewidth{1.003750pt}%
\definecolor{currentstroke}{rgb}{0.486275,0.796078,0.635294}%
\pgfsetstrokecolor{currentstroke}%
\pgfsetdash{}{0pt}%
\pgfpathmoveto{\pgfqpoint{0.906221in}{0.225887in}}%
\pgfpathlineto{\pgfqpoint{1.006221in}{0.225887in}}%
\pgfpathlineto{\pgfqpoint{1.006221in}{0.303665in}}%
\pgfpathlineto{\pgfqpoint{0.906221in}{0.303665in}}%
\pgfpathlineto{\pgfqpoint{0.906221in}{0.225887in}}%
\pgfpathclose%
\pgfusepath{stroke,fill}%
\end{pgfscope}%
\begin{pgfscope}%
\definecolor{textcolor}{rgb}{0.101961,0.101961,0.098039}%
\pgfsetstrokecolor{textcolor}%
\pgfsetfillcolor{textcolor}%
\pgftext[x=1.045110in,y=0.225887in,left,base]{\color{textcolor}\rmfamily\fontsize{8.000000}{9.600000}\selectfont other}%
\end{pgfscope}%
\end{pgfpicture}%
\makeatother%
\endgroup%

%% file: figures/pgf/sweep_batch_1.3b.pgf
\begingroup%
\makeatletter%
\begin{pgfpicture}%
\pgfpathrectangle{\pgfpointorigin}{\pgfqpoint{1.460000in}{1.950000in}}%
\pgfusepath{use as bounding box, clip}%
\begin{pgfscope}%
\pgfsetbuttcap%
\pgfsetmiterjoin%
\definecolor{currentfill}{rgb}{1.000000,1.000000,1.000000}%
\pgfsetfillcolor{currentfill}%
\pgfsetlinewidth{0.000000pt}%
\definecolor{currentstroke}{rgb}{1.000000,1.000000,1.000000}%
\pgfsetstrokecolor{currentstroke}%
\pgfsetdash{}{0pt}%
\pgfpathmoveto{\pgfqpoint{0.000000in}{0.000000in}}%
\pgfpathlineto{\pgfqpoint{1.460000in}{0.000000in}}%
\pgfpathlineto{\pgfqpoint{1.460000in}{1.950000in}}%
\pgfpathlineto{\pgfqpoint{0.000000in}{1.950000in}}%
\pgfpathlineto{\pgfqpoint{0.000000in}{0.000000in}}%
\pgfpathclose%
\pgfusepath{fill}%
\end{pgfscope}%
\begin{pgfscope}%
\pgfsetbuttcap%
\pgfsetmiterjoin%
\definecolor{currentfill}{rgb}{1.000000,1.000000,1.000000}%
\pgfsetfillcolor{currentfill}%
\pgfsetlinewidth{0.000000pt}%
\definecolor{currentstroke}{rgb}{0.000000,0.000000,0.000000}%
\pgfsetstrokecolor{currentstroke}%
\pgfsetstrokeopacity{0.000000}%
\pgfsetdash{}{0pt}%
\pgfpathmoveto{\pgfqpoint{0.400000in}{0.720000in}}%
\pgfpathlineto{\pgfqpoint{1.410000in}{0.720000in}}%
\pgfpathlineto{\pgfqpoint{1.410000in}{1.730000in}}%
\pgfpathlineto{\pgfqpoint{0.400000in}{1.730000in}}%
\pgfpathlineto{\pgfqpoint{0.400000in}{0.720000in}}%
\pgfpathclose%
\pgfusepath{fill}%
\end{pgfscope}%
\begin{pgfscope}%
\pgfpathrectangle{\pgfqpoint{0.400000in}{0.720000in}}{\pgfqpoint{1.010000in}{1.010000in}}%
\pgfusepath{clip}%
\pgfsetrectcap%
\pgfsetroundjoin%
\pgfsetlinewidth{0.501875pt}%
\definecolor{currentstroke}{rgb}{0.886275,0.882353,0.866667}%
\pgfsetstrokecolor{currentstroke}%
\pgfsetdash{}{0pt}%
\pgfpathmoveto{\pgfqpoint{0.445909in}{0.720000in}}%
\pgfpathlineto{\pgfqpoint{0.445909in}{1.730000in}}%
\pgfusepath{stroke}%
\end{pgfscope}%
\begin{pgfscope}%
\pgfsetbuttcap%
\pgfsetroundjoin%
\definecolor{currentfill}{rgb}{0.321569,0.317647,0.305882}%
\pgfsetfillcolor{currentfill}%
\pgfsetlinewidth{0.602250pt}%
\definecolor{currentstroke}{rgb}{0.321569,0.317647,0.305882}%
\pgfsetstrokecolor{currentstroke}%
\pgfsetdash{}{0pt}%
\pgfsys@defobject{currentmarker}{\pgfqpoint{0.000000in}{-0.034722in}}{\pgfqpoint{0.000000in}{0.000000in}}{%
\pgfpathmoveto{\pgfqpoint{0.000000in}{0.000000in}}%
\pgfpathlineto{\pgfqpoint{0.000000in}{-0.034722in}}%
\pgfusepath{stroke,fill}%
}%
\begin{pgfscope}%
\pgfsys@transformshift{0.445909in}{0.720000in}%
\pgfsys@useobject{currentmarker}{}%
\end{pgfscope}%
\end{pgfscope}%
\begin{pgfscope}%
\definecolor{textcolor}{rgb}{0.321569,0.317647,0.305882}%
\pgfsetstrokecolor{textcolor}%
\pgfsetfillcolor{textcolor}%
\pgftext[x=0.445909in,y=0.636667in,,top]{\color{textcolor}\rmfamily\fontsize{8.000000}{9.600000}\selectfont 1}%
\end{pgfscope}%
\begin{pgfscope}%
\pgfpathrectangle{\pgfqpoint{0.400000in}{0.720000in}}{\pgfqpoint{1.010000in}{1.010000in}}%
\pgfusepath{clip}%
\pgfsetrectcap%
\pgfsetroundjoin%
\pgfsetlinewidth{0.501875pt}%
\definecolor{currentstroke}{rgb}{0.886275,0.882353,0.866667}%
\pgfsetstrokecolor{currentstroke}%
\pgfsetdash{}{0pt}%
\pgfpathmoveto{\pgfqpoint{0.813182in}{0.720000in}}%
\pgfpathlineto{\pgfqpoint{0.813182in}{1.730000in}}%
\pgfusepath{stroke}%
\end{pgfscope}%
\begin{pgfscope}%
\pgfsetbuttcap%
\pgfsetroundjoin%
\definecolor{currentfill}{rgb}{0.321569,0.317647,0.305882}%
\pgfsetfillcolor{currentfill}%
\pgfsetlinewidth{0.602250pt}%
\definecolor{currentstroke}{rgb}{0.321569,0.317647,0.305882}%
\pgfsetstrokecolor{currentstroke}%
\pgfsetdash{}{0pt}%
\pgfsys@defobject{currentmarker}{\pgfqpoint{0.000000in}{-0.034722in}}{\pgfqpoint{0.000000in}{0.000000in}}{%
\pgfpathmoveto{\pgfqpoint{0.000000in}{0.000000in}}%
\pgfpathlineto{\pgfqpoint{0.000000in}{-0.034722in}}%
\pgfusepath{stroke,fill}%
}%
\begin{pgfscope}%
\pgfsys@transformshift{0.813182in}{0.720000in}%
\pgfsys@useobject{currentmarker}{}%
\end{pgfscope}%
\end{pgfscope}%
\begin{pgfscope}%
\definecolor{textcolor}{rgb}{0.321569,0.317647,0.305882}%
\pgfsetstrokecolor{textcolor}%
\pgfsetfillcolor{textcolor}%
\pgftext[x=0.813182in,y=0.636667in,,top]{\color{textcolor}\rmfamily\fontsize{8.000000}{9.600000}\selectfont 4}%
\end{pgfscope}%
\begin{pgfscope}%
\pgfpathrectangle{\pgfqpoint{0.400000in}{0.720000in}}{\pgfqpoint{1.010000in}{1.010000in}}%
\pgfusepath{clip}%
\pgfsetrectcap%
\pgfsetroundjoin%
\pgfsetlinewidth{0.501875pt}%
\definecolor{currentstroke}{rgb}{0.886275,0.882353,0.866667}%
\pgfsetstrokecolor{currentstroke}%
\pgfsetdash{}{0pt}%
\pgfpathmoveto{\pgfqpoint{0.996818in}{0.720000in}}%
\pgfpathlineto{\pgfqpoint{0.996818in}{1.730000in}}%
\pgfusepath{stroke}%
\end{pgfscope}%
\begin{pgfscope}%
\pgfsetbuttcap%
\pgfsetroundjoin%
\definecolor{currentfill}{rgb}{0.321569,0.317647,0.305882}%
\pgfsetfillcolor{currentfill}%
\pgfsetlinewidth{0.602250pt}%
\definecolor{currentstroke}{rgb}{0.321569,0.317647,0.305882}%
\pgfsetstrokecolor{currentstroke}%
\pgfsetdash{}{0pt}%
\pgfsys@defobject{currentmarker}{\pgfqpoint{0.000000in}{-0.034722in}}{\pgfqpoint{0.000000in}{0.000000in}}{%
\pgfpathmoveto{\pgfqpoint{0.000000in}{0.000000in}}%
\pgfpathlineto{\pgfqpoint{0.000000in}{-0.034722in}}%
\pgfusepath{stroke,fill}%
}%
\begin{pgfscope}%
\pgfsys@transformshift{0.996818in}{0.720000in}%
\pgfsys@useobject{currentmarker}{}%
\end{pgfscope}%
\end{pgfscope}%
\begin{pgfscope}%
\definecolor{textcolor}{rgb}{0.321569,0.317647,0.305882}%
\pgfsetstrokecolor{textcolor}%
\pgfsetfillcolor{textcolor}%
\pgftext[x=0.996818in,y=0.636667in,,top]{\color{textcolor}\rmfamily\fontsize{8.000000}{9.600000}\selectfont 8}%
\end{pgfscope}%
\begin{pgfscope}%
\pgfpathrectangle{\pgfqpoint{0.400000in}{0.720000in}}{\pgfqpoint{1.010000in}{1.010000in}}%
\pgfusepath{clip}%
\pgfsetrectcap%
\pgfsetroundjoin%
\pgfsetlinewidth{0.501875pt}%
\definecolor{currentstroke}{rgb}{0.886275,0.882353,0.866667}%
\pgfsetstrokecolor{currentstroke}%
\pgfsetdash{}{0pt}%
\pgfpathmoveto{\pgfqpoint{1.180455in}{0.720000in}}%
\pgfpathlineto{\pgfqpoint{1.180455in}{1.730000in}}%
\pgfusepath{stroke}%
\end{pgfscope}%
\begin{pgfscope}%
\pgfsetbuttcap%
\pgfsetroundjoin%
\definecolor{currentfill}{rgb}{0.321569,0.317647,0.305882}%
\pgfsetfillcolor{currentfill}%
\pgfsetlinewidth{0.602250pt}%
\definecolor{currentstroke}{rgb}{0.321569,0.317647,0.305882}%
\pgfsetstrokecolor{currentstroke}%
\pgfsetdash{}{0pt}%
\pgfsys@defobject{currentmarker}{\pgfqpoint{0.000000in}{-0.034722in}}{\pgfqpoint{0.000000in}{0.000000in}}{%
\pgfpathmoveto{\pgfqpoint{0.000000in}{0.000000in}}%
\pgfpathlineto{\pgfqpoint{0.000000in}{-0.034722in}}%
\pgfusepath{stroke,fill}%
}%
\begin{pgfscope}%
\pgfsys@transformshift{1.180455in}{0.720000in}%
\pgfsys@useobject{currentmarker}{}%
\end{pgfscope}%
\end{pgfscope}%
\begin{pgfscope}%
\definecolor{textcolor}{rgb}{0.321569,0.317647,0.305882}%
\pgfsetstrokecolor{textcolor}%
\pgfsetfillcolor{textcolor}%
\pgftext[x=1.180455in,y=0.636667in,,top]{\color{textcolor}\rmfamily\fontsize{8.000000}{9.600000}\selectfont 16}%
\end{pgfscope}%
\begin{pgfscope}%
\pgfpathrectangle{\pgfqpoint{0.400000in}{0.720000in}}{\pgfqpoint{1.010000in}{1.010000in}}%
\pgfusepath{clip}%
\pgfsetrectcap%
\pgfsetroundjoin%
\pgfsetlinewidth{0.501875pt}%
\definecolor{currentstroke}{rgb}{0.886275,0.882353,0.866667}%
\pgfsetstrokecolor{currentstroke}%
\pgfsetdash{}{0pt}%
\pgfpathmoveto{\pgfqpoint{1.364091in}{0.720000in}}%
\pgfpathlineto{\pgfqpoint{1.364091in}{1.730000in}}%
\pgfusepath{stroke}%
\end{pgfscope}%
\begin{pgfscope}%
\pgfsetbuttcap%
\pgfsetroundjoin%
\definecolor{currentfill}{rgb}{0.321569,0.317647,0.305882}%
\pgfsetfillcolor{currentfill}%
\pgfsetlinewidth{0.602250pt}%
\definecolor{currentstroke}{rgb}{0.321569,0.317647,0.305882}%
\pgfsetstrokecolor{currentstroke}%
\pgfsetdash{}{0pt}%
\pgfsys@defobject{currentmarker}{\pgfqpoint{0.000000in}{-0.034722in}}{\pgfqpoint{0.000000in}{0.000000in}}{%
\pgfpathmoveto{\pgfqpoint{0.000000in}{0.000000in}}%
\pgfpathlineto{\pgfqpoint{0.000000in}{-0.034722in}}%
\pgfusepath{stroke,fill}%
}%
\begin{pgfscope}%
\pgfsys@transformshift{1.364091in}{0.720000in}%
\pgfsys@useobject{currentmarker}{}%
\end{pgfscope}%
\end{pgfscope}%
\begin{pgfscope}%
\definecolor{textcolor}{rgb}{0.321569,0.317647,0.305882}%
\pgfsetstrokecolor{textcolor}%
\pgfsetfillcolor{textcolor}%
\pgftext[x=1.364091in,y=0.636667in,,top]{\color{textcolor}\rmfamily\fontsize{8.000000}{9.600000}\selectfont 32}%
\end{pgfscope}%
\begin{pgfscope}%
\pgfpathrectangle{\pgfqpoint{0.400000in}{0.720000in}}{\pgfqpoint{1.010000in}{1.010000in}}%
\pgfusepath{clip}%
\pgfsetrectcap%
\pgfsetroundjoin%
\pgfsetlinewidth{0.501875pt}%
\definecolor{currentstroke}{rgb}{0.886275,0.882353,0.866667}%
\pgfsetstrokecolor{currentstroke}%
\pgfsetdash{}{0pt}%
\pgfpathmoveto{\pgfqpoint{0.400000in}{0.851348in}}%
\pgfpathlineto{\pgfqpoint{1.410000in}{0.851348in}}%
\pgfusepath{stroke}%
\end{pgfscope}%
\begin{pgfscope}%
\pgfsetbuttcap%
\pgfsetroundjoin%
\definecolor{currentfill}{rgb}{0.321569,0.317647,0.305882}%
\pgfsetfillcolor{currentfill}%
\pgfsetlinewidth{0.602250pt}%
\definecolor{currentstroke}{rgb}{0.321569,0.317647,0.305882}%
\pgfsetstrokecolor{currentstroke}%
\pgfsetdash{}{0pt}%
\pgfsys@defobject{currentmarker}{\pgfqpoint{-0.034722in}{0.000000in}}{\pgfqpoint{-0.000000in}{0.000000in}}{%
\pgfpathmoveto{\pgfqpoint{-0.000000in}{0.000000in}}%
\pgfpathlineto{\pgfqpoint{-0.034722in}{0.000000in}}%
\pgfusepath{stroke,fill}%
}%
\begin{pgfscope}%
\pgfsys@transformshift{0.400000in}{0.851348in}%
\pgfsys@useobject{currentmarker}{}%
\end{pgfscope}%
\end{pgfscope}%
\begin{pgfscope}%
\definecolor{textcolor}{rgb}{0.101961,0.101961,0.098039}%
\pgfsetstrokecolor{textcolor}%
\pgfsetfillcolor{textcolor}%
\pgftext[x=0.177778in, y=0.813460in, left, base]{\color{textcolor}\rmfamily\fontsize{8.000000}{9.600000}\selectfont 0.5}%
\end{pgfscope}%
\begin{pgfscope}%
\pgfpathrectangle{\pgfqpoint{0.400000in}{0.720000in}}{\pgfqpoint{1.010000in}{1.010000in}}%
\pgfusepath{clip}%
\pgfsetrectcap%
\pgfsetroundjoin%
\pgfsetlinewidth{0.501875pt}%
\definecolor{currentstroke}{rgb}{0.886275,0.882353,0.866667}%
\pgfsetstrokecolor{currentstroke}%
\pgfsetdash{}{0pt}%
\pgfpathmoveto{\pgfqpoint{0.400000in}{1.030364in}}%
\pgfpathlineto{\pgfqpoint{1.410000in}{1.030364in}}%
\pgfusepath{stroke}%
\end{pgfscope}%
\begin{pgfscope}%
\pgfsetbuttcap%
\pgfsetroundjoin%
\definecolor{currentfill}{rgb}{0.321569,0.317647,0.305882}%
\pgfsetfillcolor{currentfill}%
\pgfsetlinewidth{0.602250pt}%
\definecolor{currentstroke}{rgb}{0.321569,0.317647,0.305882}%
\pgfsetstrokecolor{currentstroke}%
\pgfsetdash{}{0pt}%
\pgfsys@defobject{currentmarker}{\pgfqpoint{-0.034722in}{0.000000in}}{\pgfqpoint{-0.000000in}{0.000000in}}{%
\pgfpathmoveto{\pgfqpoint{-0.000000in}{0.000000in}}%
\pgfpathlineto{\pgfqpoint{-0.034722in}{0.000000in}}%
\pgfusepath{stroke,fill}%
}%
\begin{pgfscope}%
\pgfsys@transformshift{0.400000in}{1.030364in}%
\pgfsys@useobject{currentmarker}{}%
\end{pgfscope}%
\end{pgfscope}%
\begin{pgfscope}%
\definecolor{textcolor}{rgb}{0.101961,0.101961,0.098039}%
\pgfsetstrokecolor{textcolor}%
\pgfsetfillcolor{textcolor}%
\pgftext[x=0.261111in, y=0.992476in, left, base]{\color{textcolor}\rmfamily\fontsize{8.000000}{9.600000}\selectfont 1}%
\end{pgfscope}%
\begin{pgfscope}%
\pgfpathrectangle{\pgfqpoint{0.400000in}{0.720000in}}{\pgfqpoint{1.010000in}{1.010000in}}%
\pgfusepath{clip}%
\pgfsetrectcap%
\pgfsetroundjoin%
\pgfsetlinewidth{0.501875pt}%
\definecolor{currentstroke}{rgb}{0.886275,0.882353,0.866667}%
\pgfsetstrokecolor{currentstroke}%
\pgfsetdash{}{0pt}%
\pgfpathmoveto{\pgfqpoint{0.400000in}{1.209380in}}%
\pgfpathlineto{\pgfqpoint{1.410000in}{1.209380in}}%
\pgfusepath{stroke}%
\end{pgfscope}%
\begin{pgfscope}%
\pgfsetbuttcap%
\pgfsetroundjoin%
\definecolor{currentfill}{rgb}{0.321569,0.317647,0.305882}%
\pgfsetfillcolor{currentfill}%
\pgfsetlinewidth{0.602250pt}%
\definecolor{currentstroke}{rgb}{0.321569,0.317647,0.305882}%
\pgfsetstrokecolor{currentstroke}%
\pgfsetdash{}{0pt}%
\pgfsys@defobject{currentmarker}{\pgfqpoint{-0.034722in}{0.000000in}}{\pgfqpoint{-0.000000in}{0.000000in}}{%
\pgfpathmoveto{\pgfqpoint{-0.000000in}{0.000000in}}%
\pgfpathlineto{\pgfqpoint{-0.034722in}{0.000000in}}%
\pgfusepath{stroke,fill}%
}%
\begin{pgfscope}%
\pgfsys@transformshift{0.400000in}{1.209380in}%
\pgfsys@useobject{currentmarker}{}%
\end{pgfscope}%
\end{pgfscope}%
\begin{pgfscope}%
\definecolor{textcolor}{rgb}{0.101961,0.101961,0.098039}%
\pgfsetstrokecolor{textcolor}%
\pgfsetfillcolor{textcolor}%
\pgftext[x=0.261111in, y=1.171492in, left, base]{\color{textcolor}\rmfamily\fontsize{8.000000}{9.600000}\selectfont 2}%
\end{pgfscope}%
\begin{pgfscope}%
\pgfpathrectangle{\pgfqpoint{0.400000in}{0.720000in}}{\pgfqpoint{1.010000in}{1.010000in}}%
\pgfusepath{clip}%
\pgfsetrectcap%
\pgfsetroundjoin%
\pgfsetlinewidth{0.501875pt}%
\definecolor{currentstroke}{rgb}{0.886275,0.882353,0.866667}%
\pgfsetstrokecolor{currentstroke}%
\pgfsetdash{}{0pt}%
\pgfpathmoveto{\pgfqpoint{0.400000in}{1.388397in}}%
\pgfpathlineto{\pgfqpoint{1.410000in}{1.388397in}}%
\pgfusepath{stroke}%
\end{pgfscope}%
\begin{pgfscope}%
\pgfsetbuttcap%
\pgfsetroundjoin%
\definecolor{currentfill}{rgb}{0.321569,0.317647,0.305882}%
\pgfsetfillcolor{currentfill}%
\pgfsetlinewidth{0.602250pt}%
\definecolor{currentstroke}{rgb}{0.321569,0.317647,0.305882}%
\pgfsetstrokecolor{currentstroke}%
\pgfsetdash{}{0pt}%
\pgfsys@defobject{currentmarker}{\pgfqpoint{-0.034722in}{0.000000in}}{\pgfqpoint{-0.000000in}{0.000000in}}{%
\pgfpathmoveto{\pgfqpoint{-0.000000in}{0.000000in}}%
\pgfpathlineto{\pgfqpoint{-0.034722in}{0.000000in}}%
\pgfusepath{stroke,fill}%
}%
\begin{pgfscope}%
\pgfsys@transformshift{0.400000in}{1.388397in}%
\pgfsys@useobject{currentmarker}{}%
\end{pgfscope}%
\end{pgfscope}%
\begin{pgfscope}%
\definecolor{textcolor}{rgb}{0.101961,0.101961,0.098039}%
\pgfsetstrokecolor{textcolor}%
\pgfsetfillcolor{textcolor}%
\pgftext[x=0.261111in, y=1.350509in, left, base]{\color{textcolor}\rmfamily\fontsize{8.000000}{9.600000}\selectfont 4}%
\end{pgfscope}%
\begin{pgfscope}%
\pgfpathrectangle{\pgfqpoint{0.400000in}{0.720000in}}{\pgfqpoint{1.010000in}{1.010000in}}%
\pgfusepath{clip}%
\pgfsetrectcap%
\pgfsetroundjoin%
\pgfsetlinewidth{0.501875pt}%
\definecolor{currentstroke}{rgb}{0.886275,0.882353,0.866667}%
\pgfsetstrokecolor{currentstroke}%
\pgfsetdash{}{0pt}%
\pgfpathmoveto{\pgfqpoint{0.400000in}{1.567413in}}%
\pgfpathlineto{\pgfqpoint{1.410000in}{1.567413in}}%
\pgfusepath{stroke}%
\end{pgfscope}%
\begin{pgfscope}%
\pgfsetbuttcap%
\pgfsetroundjoin%
\definecolor{currentfill}{rgb}{0.321569,0.317647,0.305882}%
\pgfsetfillcolor{currentfill}%
\pgfsetlinewidth{0.602250pt}%
\definecolor{currentstroke}{rgb}{0.321569,0.317647,0.305882}%
\pgfsetstrokecolor{currentstroke}%
\pgfsetdash{}{0pt}%
\pgfsys@defobject{currentmarker}{\pgfqpoint{-0.034722in}{0.000000in}}{\pgfqpoint{-0.000000in}{0.000000in}}{%
\pgfpathmoveto{\pgfqpoint{-0.000000in}{0.000000in}}%
\pgfpathlineto{\pgfqpoint{-0.034722in}{0.000000in}}%
\pgfusepath{stroke,fill}%
}%
\begin{pgfscope}%
\pgfsys@transformshift{0.400000in}{1.567413in}%
\pgfsys@useobject{currentmarker}{}%
\end{pgfscope}%
\end{pgfscope}%
\begin{pgfscope}%
\definecolor{textcolor}{rgb}{0.101961,0.101961,0.098039}%
\pgfsetstrokecolor{textcolor}%
\pgfsetfillcolor{textcolor}%
\pgftext[x=0.261111in, y=1.529525in, left, base]{\color{textcolor}\rmfamily\fontsize{8.000000}{9.600000}\selectfont 8}%
\end{pgfscope}%
\begin{pgfscope}%
\definecolor{textcolor}{rgb}{0.101961,0.101961,0.098039}%
\pgfsetstrokecolor{textcolor}%
\pgfsetfillcolor{textcolor}%
\pgftext[x=0.122222in,y=1.225000in,,bottom,rotate=90.000000]{\color{textcolor}\rmfamily\fontsize{9.000000}{10.800000}\selectfont latency (ms)}%
\end{pgfscope}%
\begin{pgfscope}%
\pgfpathrectangle{\pgfqpoint{0.400000in}{0.720000in}}{\pgfqpoint{1.010000in}{1.010000in}}%
\pgfusepath{clip}%
\pgfsetbuttcap%
\pgfsetroundjoin%
\pgfsetlinewidth{1.104125pt}%
\definecolor{currentstroke}{rgb}{0.321569,0.317647,0.305882}%
\pgfsetstrokecolor{currentstroke}%
\pgfsetdash{{4.070000pt}{1.760000pt}}{0.000000pt}%
\pgfpathmoveto{\pgfqpoint{0.445909in}{0.765909in}}%
\pgfpathlineto{\pgfqpoint{0.813182in}{1.140698in}}%
\pgfpathlineto{\pgfqpoint{0.996818in}{1.323979in}}%
\pgfpathlineto{\pgfqpoint{1.180455in}{1.504015in}}%
\pgfpathlineto{\pgfqpoint{1.364091in}{1.684091in}}%
\pgfusepath{stroke}%
\end{pgfscope}%
\begin{pgfscope}%
\pgfpathrectangle{\pgfqpoint{0.400000in}{0.720000in}}{\pgfqpoint{1.010000in}{1.010000in}}%
\pgfusepath{clip}%
\pgfsetbuttcap%
\pgfsetroundjoin%
\definecolor{currentfill}{rgb}{0.321569,0.317647,0.305882}%
\pgfsetfillcolor{currentfill}%
\pgfsetlinewidth{1.003750pt}%
\definecolor{currentstroke}{rgb}{0.321569,0.317647,0.305882}%
\pgfsetstrokecolor{currentstroke}%
\pgfsetdash{}{0pt}%
\pgfsys@defobject{currentmarker}{\pgfqpoint{-0.015278in}{-0.015278in}}{\pgfqpoint{0.015278in}{0.015278in}}{%
\pgfpathmoveto{\pgfqpoint{0.000000in}{-0.015278in}}%
\pgfpathcurveto{\pgfqpoint{0.004052in}{-0.015278in}}{\pgfqpoint{0.007938in}{-0.013668in}}{\pgfqpoint{0.010803in}{-0.010803in}}%
\pgfpathcurveto{\pgfqpoint{0.013668in}{-0.007938in}}{\pgfqpoint{0.015278in}{-0.004052in}}{\pgfqpoint{0.015278in}{0.000000in}}%
\pgfpathcurveto{\pgfqpoint{0.015278in}{0.004052in}}{\pgfqpoint{0.013668in}{0.007938in}}{\pgfqpoint{0.010803in}{0.010803in}}%
\pgfpathcurveto{\pgfqpoint{0.007938in}{0.013668in}}{\pgfqpoint{0.004052in}{0.015278in}}{\pgfqpoint{0.000000in}{0.015278in}}%
\pgfpathcurveto{\pgfqpoint{-0.004052in}{0.015278in}}{\pgfqpoint{-0.007938in}{0.013668in}}{\pgfqpoint{-0.010803in}{0.010803in}}%
\pgfpathcurveto{\pgfqpoint{-0.013668in}{0.007938in}}{\pgfqpoint{-0.015278in}{0.004052in}}{\pgfqpoint{-0.015278in}{0.000000in}}%
\pgfpathcurveto{\pgfqpoint{-0.015278in}{-0.004052in}}{\pgfqpoint{-0.013668in}{-0.007938in}}{\pgfqpoint{-0.010803in}{-0.010803in}}%
\pgfpathcurveto{\pgfqpoint{-0.007938in}{-0.013668in}}{\pgfqpoint{-0.004052in}{-0.015278in}}{\pgfqpoint{0.000000in}{-0.015278in}}%
\pgfpathlineto{\pgfqpoint{0.000000in}{-0.015278in}}%
\pgfpathclose%
\pgfusepath{stroke,fill}%
}%
\begin{pgfscope}%
\pgfsys@transformshift{0.445909in}{0.765909in}%
\pgfsys@useobject{currentmarker}{}%
\end{pgfscope}%
\begin{pgfscope}%
\pgfsys@transformshift{0.813182in}{1.140698in}%
\pgfsys@useobject{currentmarker}{}%
\end{pgfscope}%
\begin{pgfscope}%
\pgfsys@transformshift{0.996818in}{1.323979in}%
\pgfsys@useobject{currentmarker}{}%
\end{pgfscope}%
\begin{pgfscope}%
\pgfsys@transformshift{1.180455in}{1.504015in}%
\pgfsys@useobject{currentmarker}{}%
\end{pgfscope}%
\begin{pgfscope}%
\pgfsys@transformshift{1.364091in}{1.684091in}%
\pgfsys@useobject{currentmarker}{}%
\end{pgfscope}%
\end{pgfscope}%
\begin{pgfscope}%
\pgfpathrectangle{\pgfqpoint{0.400000in}{0.720000in}}{\pgfqpoint{1.010000in}{1.010000in}}%
\pgfusepath{clip}%
\pgfsetrectcap%
\pgfsetroundjoin%
\pgfsetlinewidth{1.104125pt}%
\definecolor{currentstroke}{rgb}{0.862745,0.223529,0.466667}%
\pgfsetstrokecolor{currentstroke}%
\pgfsetdash{}{0pt}%
\pgfpathmoveto{\pgfqpoint{0.445909in}{0.987520in}}%
\pgfpathlineto{\pgfqpoint{0.813182in}{1.041687in}}%
\pgfpathlineto{\pgfqpoint{0.996818in}{1.206576in}}%
\pgfpathlineto{\pgfqpoint{1.180455in}{1.381374in}}%
\pgfpathlineto{\pgfqpoint{1.364091in}{1.554037in}}%
\pgfusepath{stroke}%
\end{pgfscope}%
\begin{pgfscope}%
\pgfpathrectangle{\pgfqpoint{0.400000in}{0.720000in}}{\pgfqpoint{1.010000in}{1.010000in}}%
\pgfusepath{clip}%
\pgfsetbuttcap%
\pgfsetroundjoin%
\definecolor{currentfill}{rgb}{0.862745,0.223529,0.466667}%
\pgfsetfillcolor{currentfill}%
\pgfsetlinewidth{1.003750pt}%
\definecolor{currentstroke}{rgb}{0.862745,0.223529,0.466667}%
\pgfsetstrokecolor{currentstroke}%
\pgfsetdash{}{0pt}%
\pgfsys@defobject{currentmarker}{\pgfqpoint{-0.015278in}{-0.015278in}}{\pgfqpoint{0.015278in}{0.015278in}}{%
\pgfpathmoveto{\pgfqpoint{0.000000in}{-0.015278in}}%
\pgfpathcurveto{\pgfqpoint{0.004052in}{-0.015278in}}{\pgfqpoint{0.007938in}{-0.013668in}}{\pgfqpoint{0.010803in}{-0.010803in}}%
\pgfpathcurveto{\pgfqpoint{0.013668in}{-0.007938in}}{\pgfqpoint{0.015278in}{-0.004052in}}{\pgfqpoint{0.015278in}{0.000000in}}%
\pgfpathcurveto{\pgfqpoint{0.015278in}{0.004052in}}{\pgfqpoint{0.013668in}{0.007938in}}{\pgfqpoint{0.010803in}{0.010803in}}%
\pgfpathcurveto{\pgfqpoint{0.007938in}{0.013668in}}{\pgfqpoint{0.004052in}{0.015278in}}{\pgfqpoint{0.000000in}{0.015278in}}%
\pgfpathcurveto{\pgfqpoint{-0.004052in}{0.015278in}}{\pgfqpoint{-0.007938in}{0.013668in}}{\pgfqpoint{-0.010803in}{0.010803in}}%
\pgfpathcurveto{\pgfqpoint{-0.013668in}{0.007938in}}{\pgfqpoint{-0.015278in}{0.004052in}}{\pgfqpoint{-0.015278in}{0.000000in}}%
\pgfpathcurveto{\pgfqpoint{-0.015278in}{-0.004052in}}{\pgfqpoint{-0.013668in}{-0.007938in}}{\pgfqpoint{-0.010803in}{-0.010803in}}%
\pgfpathcurveto{\pgfqpoint{-0.007938in}{-0.013668in}}{\pgfqpoint{-0.004052in}{-0.015278in}}{\pgfqpoint{0.000000in}{-0.015278in}}%
\pgfpathlineto{\pgfqpoint{0.000000in}{-0.015278in}}%
\pgfpathclose%
\pgfusepath{stroke,fill}%
}%
\begin{pgfscope}%
\pgfsys@transformshift{0.445909in}{0.987520in}%
\pgfsys@useobject{currentmarker}{}%
\end{pgfscope}%
\begin{pgfscope}%
\pgfsys@transformshift{0.813182in}{1.041687in}%
\pgfsys@useobject{currentmarker}{}%
\end{pgfscope}%
\begin{pgfscope}%
\pgfsys@transformshift{0.996818in}{1.206576in}%
\pgfsys@useobject{currentmarker}{}%
\end{pgfscope}%
\begin{pgfscope}%
\pgfsys@transformshift{1.180455in}{1.381374in}%
\pgfsys@useobject{currentmarker}{}%
\end{pgfscope}%
\begin{pgfscope}%
\pgfsys@transformshift{1.364091in}{1.554037in}%
\pgfsys@useobject{currentmarker}{}%
\end{pgfscope}%
\end{pgfscope}%
\begin{pgfscope}%
\pgfpathrectangle{\pgfqpoint{0.400000in}{0.720000in}}{\pgfqpoint{1.010000in}{1.010000in}}%
\pgfusepath{clip}%
\pgfsetrectcap%
\pgfsetroundjoin%
\pgfsetlinewidth{1.104125pt}%
\definecolor{currentstroke}{rgb}{0.031373,0.564706,0.600000}%
\pgfsetstrokecolor{currentstroke}%
\pgfsetdash{}{0pt}%
\pgfpathmoveto{\pgfqpoint{0.445909in}{0.827766in}}%
\pgfpathlineto{\pgfqpoint{0.813182in}{1.018674in}}%
\pgfpathlineto{\pgfqpoint{0.996818in}{1.201948in}}%
\pgfpathlineto{\pgfqpoint{1.180455in}{1.378784in}}%
\pgfpathlineto{\pgfqpoint{1.364091in}{1.557684in}}%
\pgfusepath{stroke}%
\end{pgfscope}%
\begin{pgfscope}%
\pgfpathrectangle{\pgfqpoint{0.400000in}{0.720000in}}{\pgfqpoint{1.010000in}{1.010000in}}%
\pgfusepath{clip}%
\pgfsetbuttcap%
\pgfsetroundjoin%
\definecolor{currentfill}{rgb}{0.031373,0.564706,0.600000}%
\pgfsetfillcolor{currentfill}%
\pgfsetlinewidth{1.003750pt}%
\definecolor{currentstroke}{rgb}{0.031373,0.564706,0.600000}%
\pgfsetstrokecolor{currentstroke}%
\pgfsetdash{}{0pt}%
\pgfsys@defobject{currentmarker}{\pgfqpoint{-0.015278in}{-0.015278in}}{\pgfqpoint{0.015278in}{0.015278in}}{%
\pgfpathmoveto{\pgfqpoint{0.000000in}{-0.015278in}}%
\pgfpathcurveto{\pgfqpoint{0.004052in}{-0.015278in}}{\pgfqpoint{0.007938in}{-0.013668in}}{\pgfqpoint{0.010803in}{-0.010803in}}%
\pgfpathcurveto{\pgfqpoint{0.013668in}{-0.007938in}}{\pgfqpoint{0.015278in}{-0.004052in}}{\pgfqpoint{0.015278in}{0.000000in}}%
\pgfpathcurveto{\pgfqpoint{0.015278in}{0.004052in}}{\pgfqpoint{0.013668in}{0.007938in}}{\pgfqpoint{0.010803in}{0.010803in}}%
\pgfpathcurveto{\pgfqpoint{0.007938in}{0.013668in}}{\pgfqpoint{0.004052in}{0.015278in}}{\pgfqpoint{0.000000in}{0.015278in}}%
\pgfpathcurveto{\pgfqpoint{-0.004052in}{0.015278in}}{\pgfqpoint{-0.007938in}{0.013668in}}{\pgfqpoint{-0.010803in}{0.010803in}}%
\pgfpathcurveto{\pgfqpoint{-0.013668in}{0.007938in}}{\pgfqpoint{-0.015278in}{0.004052in}}{\pgfqpoint{-0.015278in}{0.000000in}}%
\pgfpathcurveto{\pgfqpoint{-0.015278in}{-0.004052in}}{\pgfqpoint{-0.013668in}{-0.007938in}}{\pgfqpoint{-0.010803in}{-0.010803in}}%
\pgfpathcurveto{\pgfqpoint{-0.007938in}{-0.013668in}}{\pgfqpoint{-0.004052in}{-0.015278in}}{\pgfqpoint{0.000000in}{-0.015278in}}%
\pgfpathlineto{\pgfqpoint{0.000000in}{-0.015278in}}%
\pgfpathclose%
\pgfusepath{stroke,fill}%
}%
\begin{pgfscope}%
\pgfsys@transformshift{0.445909in}{0.827766in}%
\pgfsys@useobject{currentmarker}{}%
\end{pgfscope}%
\begin{pgfscope}%
\pgfsys@transformshift{0.813182in}{1.018674in}%
\pgfsys@useobject{currentmarker}{}%
\end{pgfscope}%
\begin{pgfscope}%
\pgfsys@transformshift{0.996818in}{1.201948in}%
\pgfsys@useobject{currentmarker}{}%
\end{pgfscope}%
\begin{pgfscope}%
\pgfsys@transformshift{1.180455in}{1.378784in}%
\pgfsys@useobject{currentmarker}{}%
\end{pgfscope}%
\begin{pgfscope}%
\pgfsys@transformshift{1.364091in}{1.557684in}%
\pgfsys@useobject{currentmarker}{}%
\end{pgfscope}%
\end{pgfscope}%
\begin{pgfscope}%
\pgfpathrectangle{\pgfqpoint{0.400000in}{0.720000in}}{\pgfqpoint{1.010000in}{1.010000in}}%
\pgfusepath{clip}%
\pgfsetrectcap%
\pgfsetroundjoin%
\pgfsetlinewidth{1.104125pt}%
\definecolor{currentstroke}{rgb}{0.015686,0.321569,0.458824}%
\pgfsetstrokecolor{currentstroke}%
\pgfsetdash{}{0pt}%
\pgfpathmoveto{\pgfqpoint{0.445909in}{0.882982in}}%
\pgfpathlineto{\pgfqpoint{0.813182in}{1.033024in}}%
\pgfpathlineto{\pgfqpoint{0.996818in}{1.215455in}}%
\pgfpathlineto{\pgfqpoint{1.180455in}{1.396121in}}%
\pgfpathlineto{\pgfqpoint{1.364091in}{1.573161in}}%
\pgfusepath{stroke}%
\end{pgfscope}%
\begin{pgfscope}%
\pgfpathrectangle{\pgfqpoint{0.400000in}{0.720000in}}{\pgfqpoint{1.010000in}{1.010000in}}%
\pgfusepath{clip}%
\pgfsetbuttcap%
\pgfsetroundjoin%
\definecolor{currentfill}{rgb}{0.015686,0.321569,0.458824}%
\pgfsetfillcolor{currentfill}%
\pgfsetlinewidth{1.003750pt}%
\definecolor{currentstroke}{rgb}{0.015686,0.321569,0.458824}%
\pgfsetstrokecolor{currentstroke}%
\pgfsetdash{}{0pt}%
\pgfsys@defobject{currentmarker}{\pgfqpoint{-0.015278in}{-0.015278in}}{\pgfqpoint{0.015278in}{0.015278in}}{%
\pgfpathmoveto{\pgfqpoint{0.000000in}{-0.015278in}}%
\pgfpathcurveto{\pgfqpoint{0.004052in}{-0.015278in}}{\pgfqpoint{0.007938in}{-0.013668in}}{\pgfqpoint{0.010803in}{-0.010803in}}%
\pgfpathcurveto{\pgfqpoint{0.013668in}{-0.007938in}}{\pgfqpoint{0.015278in}{-0.004052in}}{\pgfqpoint{0.015278in}{0.000000in}}%
\pgfpathcurveto{\pgfqpoint{0.015278in}{0.004052in}}{\pgfqpoint{0.013668in}{0.007938in}}{\pgfqpoint{0.010803in}{0.010803in}}%
\pgfpathcurveto{\pgfqpoint{0.007938in}{0.013668in}}{\pgfqpoint{0.004052in}{0.015278in}}{\pgfqpoint{0.000000in}{0.015278in}}%
\pgfpathcurveto{\pgfqpoint{-0.004052in}{0.015278in}}{\pgfqpoint{-0.007938in}{0.013668in}}{\pgfqpoint{-0.010803in}{0.010803in}}%
\pgfpathcurveto{\pgfqpoint{-0.013668in}{0.007938in}}{\pgfqpoint{-0.015278in}{0.004052in}}{\pgfqpoint{-0.015278in}{0.000000in}}%
\pgfpathcurveto{\pgfqpoint{-0.015278in}{-0.004052in}}{\pgfqpoint{-0.013668in}{-0.007938in}}{\pgfqpoint{-0.010803in}{-0.010803in}}%
\pgfpathcurveto{\pgfqpoint{-0.007938in}{-0.013668in}}{\pgfqpoint{-0.004052in}{-0.015278in}}{\pgfqpoint{0.000000in}{-0.015278in}}%
\pgfpathlineto{\pgfqpoint{0.000000in}{-0.015278in}}%
\pgfpathclose%
\pgfusepath{stroke,fill}%
}%
\begin{pgfscope}%
\pgfsys@transformshift{0.445909in}{0.882982in}%
\pgfsys@useobject{currentmarker}{}%
\end{pgfscope}%
\begin{pgfscope}%
\pgfsys@transformshift{0.813182in}{1.033024in}%
\pgfsys@useobject{currentmarker}{}%
\end{pgfscope}%
\begin{pgfscope}%
\pgfsys@transformshift{0.996818in}{1.215455in}%
\pgfsys@useobject{currentmarker}{}%
\end{pgfscope}%
\begin{pgfscope}%
\pgfsys@transformshift{1.180455in}{1.396121in}%
\pgfsys@useobject{currentmarker}{}%
\end{pgfscope}%
\begin{pgfscope}%
\pgfsys@transformshift{1.364091in}{1.573161in}%
\pgfsys@useobject{currentmarker}{}%
\end{pgfscope}%
\end{pgfscope}%
\begin{pgfscope}%
\pgfsetrectcap%
\pgfsetmiterjoin%
\pgfsetlinewidth{0.602250pt}%
\definecolor{currentstroke}{rgb}{0.321569,0.317647,0.305882}%
\pgfsetstrokecolor{currentstroke}%
\pgfsetdash{}{0pt}%
\pgfpathmoveto{\pgfqpoint{0.400000in}{0.720000in}}%
\pgfpathlineto{\pgfqpoint{0.400000in}{1.730000in}}%
\pgfusepath{stroke}%
\end{pgfscope}%
\begin{pgfscope}%
\pgfsetrectcap%
\pgfsetmiterjoin%
\pgfsetlinewidth{0.602250pt}%
\definecolor{currentstroke}{rgb}{0.321569,0.317647,0.305882}%
\pgfsetstrokecolor{currentstroke}%
\pgfsetdash{}{0pt}%
\pgfpathmoveto{\pgfqpoint{0.400000in}{0.720000in}}%
\pgfpathlineto{\pgfqpoint{1.410000in}{0.720000in}}%
\pgfusepath{stroke}%
\end{pgfscope}%
\begin{pgfscope}%
\definecolor{textcolor}{rgb}{0.101961,0.101961,0.098039}%
\pgfsetstrokecolor{textcolor}%
\pgfsetfillcolor{textcolor}%
\pgftext[x=0.905000in,y=1.813333in,,base]{\color{textcolor}\rmfamily\fontsize{8.500000}{10.200000}\selectfont Mamba-2 1.3B}%
\end{pgfscope}%
\begin{pgfscope}%
\definecolor{textcolor}{rgb}{0.101961,0.101961,0.098039}%
\pgfsetstrokecolor{textcolor}%
\pgfsetfillcolor{textcolor}%
\pgftext[x=0.905000in,y=0.420000in,,]{\color{textcolor}\rmfamily\fontsize{9.000000}{10.800000}\selectfont batch size}%
\end{pgfscope}%
\begin{pgfscope}%
\pgfsetbuttcap%
\pgfsetroundjoin%
\pgfsetlinewidth{1.104125pt}%
\definecolor{currentstroke}{rgb}{0.321569,0.317647,0.305882}%
\pgfsetstrokecolor{currentstroke}%
\pgfsetdash{{4.070000pt}{1.760000pt}}{0.000000pt}%
\pgfpathmoveto{\pgfqpoint{0.418725in}{0.264776in}}%
\pgfpathlineto{\pgfqpoint{0.485391in}{0.264776in}}%
\pgfpathlineto{\pgfqpoint{0.552058in}{0.264776in}}%
\pgfusepath{stroke}%
\end{pgfscope}%
\begin{pgfscope}%
\pgfsetbuttcap%
\pgfsetroundjoin%
\definecolor{currentfill}{rgb}{0.321569,0.317647,0.305882}%
\pgfsetfillcolor{currentfill}%
\pgfsetlinewidth{1.003750pt}%
\definecolor{currentstroke}{rgb}{0.321569,0.317647,0.305882}%
\pgfsetstrokecolor{currentstroke}%
\pgfsetdash{}{0pt}%
\pgfsys@defobject{currentmarker}{\pgfqpoint{-0.015278in}{-0.015278in}}{\pgfqpoint{0.015278in}{0.015278in}}{%
\pgfpathmoveto{\pgfqpoint{0.000000in}{-0.015278in}}%
\pgfpathcurveto{\pgfqpoint{0.004052in}{-0.015278in}}{\pgfqpoint{0.007938in}{-0.013668in}}{\pgfqpoint{0.010803in}{-0.010803in}}%
\pgfpathcurveto{\pgfqpoint{0.013668in}{-0.007938in}}{\pgfqpoint{0.015278in}{-0.004052in}}{\pgfqpoint{0.015278in}{0.000000in}}%
\pgfpathcurveto{\pgfqpoint{0.015278in}{0.004052in}}{\pgfqpoint{0.013668in}{0.007938in}}{\pgfqpoint{0.010803in}{0.010803in}}%
\pgfpathcurveto{\pgfqpoint{0.007938in}{0.013668in}}{\pgfqpoint{0.004052in}{0.015278in}}{\pgfqpoint{0.000000in}{0.015278in}}%
\pgfpathcurveto{\pgfqpoint{-0.004052in}{0.015278in}}{\pgfqpoint{-0.007938in}{0.013668in}}{\pgfqpoint{-0.010803in}{0.010803in}}%
\pgfpathcurveto{\pgfqpoint{-0.013668in}{0.007938in}}{\pgfqpoint{-0.015278in}{0.004052in}}{\pgfqpoint{-0.015278in}{0.000000in}}%
\pgfpathcurveto{\pgfqpoint{-0.015278in}{-0.004052in}}{\pgfqpoint{-0.013668in}{-0.007938in}}{\pgfqpoint{-0.010803in}{-0.010803in}}%
\pgfpathcurveto{\pgfqpoint{-0.007938in}{-0.013668in}}{\pgfqpoint{-0.004052in}{-0.015278in}}{\pgfqpoint{0.000000in}{-0.015278in}}%
\pgfpathlineto{\pgfqpoint{0.000000in}{-0.015278in}}%
\pgfpathclose%
\pgfusepath{stroke,fill}%
}%
\begin{pgfscope}%
\pgfsys@transformshift{0.485391in}{0.264776in}%
\pgfsys@useobject{currentmarker}{}%
\end{pgfscope}%
\end{pgfscope}%
\begin{pgfscope}%
\definecolor{textcolor}{rgb}{0.101961,0.101961,0.098039}%
\pgfsetstrokecolor{textcolor}%
\pgfsetfillcolor{textcolor}%
\pgftext[x=0.596503in,y=0.225887in,left,base]{\color{textcolor}\rmfamily\fontsize{8.000000}{9.600000}\selectfont FP16}%
\end{pgfscope}%
\begin{pgfscope}%
\pgfsetrectcap%
\pgfsetroundjoin%
\pgfsetlinewidth{1.104125pt}%
\definecolor{currentstroke}{rgb}{0.862745,0.223529,0.466667}%
\pgfsetstrokecolor{currentstroke}%
\pgfsetdash{}{0pt}%
\pgfpathmoveto{\pgfqpoint{0.418725in}{0.107388in}}%
\pgfpathlineto{\pgfqpoint{0.485391in}{0.107388in}}%
\pgfpathlineto{\pgfqpoint{0.552058in}{0.107388in}}%
\pgfusepath{stroke}%
\end{pgfscope}%
\begin{pgfscope}%
\pgfsetbuttcap%
\pgfsetroundjoin%
\definecolor{currentfill}{rgb}{0.862745,0.223529,0.466667}%
\pgfsetfillcolor{currentfill}%
\pgfsetlinewidth{1.003750pt}%
\definecolor{currentstroke}{rgb}{0.862745,0.223529,0.466667}%
\pgfsetstrokecolor{currentstroke}%
\pgfsetdash{}{0pt}%
\pgfsys@defobject{currentmarker}{\pgfqpoint{-0.015278in}{-0.015278in}}{\pgfqpoint{0.015278in}{0.015278in}}{%
\pgfpathmoveto{\pgfqpoint{0.000000in}{-0.015278in}}%
\pgfpathcurveto{\pgfqpoint{0.004052in}{-0.015278in}}{\pgfqpoint{0.007938in}{-0.013668in}}{\pgfqpoint{0.010803in}{-0.010803in}}%
\pgfpathcurveto{\pgfqpoint{0.013668in}{-0.007938in}}{\pgfqpoint{0.015278in}{-0.004052in}}{\pgfqpoint{0.015278in}{0.000000in}}%
\pgfpathcurveto{\pgfqpoint{0.015278in}{0.004052in}}{\pgfqpoint{0.013668in}{0.007938in}}{\pgfqpoint{0.010803in}{0.010803in}}%
\pgfpathcurveto{\pgfqpoint{0.007938in}{0.013668in}}{\pgfqpoint{0.004052in}{0.015278in}}{\pgfqpoint{0.000000in}{0.015278in}}%
\pgfpathcurveto{\pgfqpoint{-0.004052in}{0.015278in}}{\pgfqpoint{-0.007938in}{0.013668in}}{\pgfqpoint{-0.010803in}{0.010803in}}%
\pgfpathcurveto{\pgfqpoint{-0.013668in}{0.007938in}}{\pgfqpoint{-0.015278in}{0.004052in}}{\pgfqpoint{-0.015278in}{0.000000in}}%
\pgfpathcurveto{\pgfqpoint{-0.015278in}{-0.004052in}}{\pgfqpoint{-0.013668in}{-0.007938in}}{\pgfqpoint{-0.010803in}{-0.010803in}}%
\pgfpathcurveto{\pgfqpoint{-0.007938in}{-0.013668in}}{\pgfqpoint{-0.004052in}{-0.015278in}}{\pgfqpoint{0.000000in}{-0.015278in}}%
\pgfpathlineto{\pgfqpoint{0.000000in}{-0.015278in}}%
\pgfpathclose%
\pgfusepath{stroke,fill}%
}%
\begin{pgfscope}%
\pgfsys@transformshift{0.485391in}{0.107388in}%
\pgfsys@useobject{currentmarker}{}%
\end{pgfscope}%
\end{pgfscope}%
\begin{pgfscope}%
\definecolor{textcolor}{rgb}{0.101961,0.101961,0.098039}%
\pgfsetstrokecolor{textcolor}%
\pgfsetfillcolor{textcolor}%
\pgftext[x=0.596503in,y=0.068499in,left,base]{\color{textcolor}\rmfamily\fontsize{8.000000}{9.600000}\selectfont SSDi8}%
\end{pgfscope}%
\begin{pgfscope}%
\pgfsetrectcap%
\pgfsetroundjoin%
\pgfsetlinewidth{1.104125pt}%
\definecolor{currentstroke}{rgb}{0.031373,0.564706,0.600000}%
\pgfsetstrokecolor{currentstroke}%
\pgfsetdash{}{0pt}%
\pgfpathmoveto{\pgfqpoint{0.975611in}{0.264776in}}%
\pgfpathlineto{\pgfqpoint{1.042277in}{0.264776in}}%
\pgfpathlineto{\pgfqpoint{1.108944in}{0.264776in}}%
\pgfusepath{stroke}%
\end{pgfscope}%
\begin{pgfscope}%
\pgfsetbuttcap%
\pgfsetroundjoin%
\definecolor{currentfill}{rgb}{0.031373,0.564706,0.600000}%
\pgfsetfillcolor{currentfill}%
\pgfsetlinewidth{1.003750pt}%
\definecolor{currentstroke}{rgb}{0.031373,0.564706,0.600000}%
\pgfsetstrokecolor{currentstroke}%
\pgfsetdash{}{0pt}%
\pgfsys@defobject{currentmarker}{\pgfqpoint{-0.015278in}{-0.015278in}}{\pgfqpoint{0.015278in}{0.015278in}}{%
\pgfpathmoveto{\pgfqpoint{0.000000in}{-0.015278in}}%
\pgfpathcurveto{\pgfqpoint{0.004052in}{-0.015278in}}{\pgfqpoint{0.007938in}{-0.013668in}}{\pgfqpoint{0.010803in}{-0.010803in}}%
\pgfpathcurveto{\pgfqpoint{0.013668in}{-0.007938in}}{\pgfqpoint{0.015278in}{-0.004052in}}{\pgfqpoint{0.015278in}{0.000000in}}%
\pgfpathcurveto{\pgfqpoint{0.015278in}{0.004052in}}{\pgfqpoint{0.013668in}{0.007938in}}{\pgfqpoint{0.010803in}{0.010803in}}%
\pgfpathcurveto{\pgfqpoint{0.007938in}{0.013668in}}{\pgfqpoint{0.004052in}{0.015278in}}{\pgfqpoint{0.000000in}{0.015278in}}%
\pgfpathcurveto{\pgfqpoint{-0.004052in}{0.015278in}}{\pgfqpoint{-0.007938in}{0.013668in}}{\pgfqpoint{-0.010803in}{0.010803in}}%
\pgfpathcurveto{\pgfqpoint{-0.013668in}{0.007938in}}{\pgfqpoint{-0.015278in}{0.004052in}}{\pgfqpoint{-0.015278in}{0.000000in}}%
\pgfpathcurveto{\pgfqpoint{-0.015278in}{-0.004052in}}{\pgfqpoint{-0.013668in}{-0.007938in}}{\pgfqpoint{-0.010803in}{-0.010803in}}%
\pgfpathcurveto{\pgfqpoint{-0.007938in}{-0.013668in}}{\pgfqpoint{-0.004052in}{-0.015278in}}{\pgfqpoint{0.000000in}{-0.015278in}}%
\pgfpathlineto{\pgfqpoint{0.000000in}{-0.015278in}}%
\pgfpathclose%
\pgfusepath{stroke,fill}%
}%
\begin{pgfscope}%
\pgfsys@transformshift{1.042277in}{0.264776in}%
\pgfsys@useobject{currentmarker}{}%
\end{pgfscope}%
\end{pgfscope}%
\begin{pgfscope}%
\definecolor{textcolor}{rgb}{0.101961,0.101961,0.098039}%
\pgfsetstrokecolor{textcolor}%
\pgfsetfillcolor{textcolor}%
\pgftext[x=1.153389in,y=0.225887in,left,base]{\color{textcolor}\rmfamily\fontsize{8.000000}{9.600000}\selectfont value}%
\end{pgfscope}%
\begin{pgfscope}%
\pgfsetrectcap%
\pgfsetroundjoin%
\pgfsetlinewidth{1.104125pt}%
\definecolor{currentstroke}{rgb}{0.015686,0.321569,0.458824}%
\pgfsetstrokecolor{currentstroke}%
\pgfsetdash{}{0pt}%
\pgfpathmoveto{\pgfqpoint{0.975611in}{0.107388in}}%
\pgfpathlineto{\pgfqpoint{1.042277in}{0.107388in}}%
\pgfpathlineto{\pgfqpoint{1.108944in}{0.107388in}}%
\pgfusepath{stroke}%
\end{pgfscope}%
\begin{pgfscope}%
\pgfsetbuttcap%
\pgfsetroundjoin%
\definecolor{currentfill}{rgb}{0.015686,0.321569,0.458824}%
\pgfsetfillcolor{currentfill}%
\pgfsetlinewidth{1.003750pt}%
\definecolor{currentstroke}{rgb}{0.015686,0.321569,0.458824}%
\pgfsetstrokecolor{currentstroke}%
\pgfsetdash{}{0pt}%
\pgfsys@defobject{currentmarker}{\pgfqpoint{-0.015278in}{-0.015278in}}{\pgfqpoint{0.015278in}{0.015278in}}{%
\pgfpathmoveto{\pgfqpoint{0.000000in}{-0.015278in}}%
\pgfpathcurveto{\pgfqpoint{0.004052in}{-0.015278in}}{\pgfqpoint{0.007938in}{-0.013668in}}{\pgfqpoint{0.010803in}{-0.010803in}}%
\pgfpathcurveto{\pgfqpoint{0.013668in}{-0.007938in}}{\pgfqpoint{0.015278in}{-0.004052in}}{\pgfqpoint{0.015278in}{0.000000in}}%
\pgfpathcurveto{\pgfqpoint{0.015278in}{0.004052in}}{\pgfqpoint{0.013668in}{0.007938in}}{\pgfqpoint{0.010803in}{0.010803in}}%
\pgfpathcurveto{\pgfqpoint{0.007938in}{0.013668in}}{\pgfqpoint{0.004052in}{0.015278in}}{\pgfqpoint{0.000000in}{0.015278in}}%
\pgfpathcurveto{\pgfqpoint{-0.004052in}{0.015278in}}{\pgfqpoint{-0.007938in}{0.013668in}}{\pgfqpoint{-0.010803in}{0.010803in}}%
\pgfpathcurveto{\pgfqpoint{-0.013668in}{0.007938in}}{\pgfqpoint{-0.015278in}{0.004052in}}{\pgfqpoint{-0.015278in}{0.000000in}}%
\pgfpathcurveto{\pgfqpoint{-0.015278in}{-0.004052in}}{\pgfqpoint{-0.013668in}{-0.007938in}}{\pgfqpoint{-0.010803in}{-0.010803in}}%
\pgfpathcurveto{\pgfqpoint{-0.007938in}{-0.013668in}}{\pgfqpoint{-0.004052in}{-0.015278in}}{\pgfqpoint{0.000000in}{-0.015278in}}%
\pgfpathlineto{\pgfqpoint{0.000000in}{-0.015278in}}%
\pgfpathclose%
\pgfusepath{stroke,fill}%
}%
\begin{pgfscope}%
\pgfsys@transformshift{1.042277in}{0.107388in}%
\pgfsys@useobject{currentmarker}{}%
\end{pgfscope}%
\end{pgfscope}%
\begin{pgfscope}%
\definecolor{textcolor}{rgb}{0.101961,0.101961,0.098039}%
\pgfsetstrokecolor{textcolor}%
\pgfsetfillcolor{textcolor}%
\pgftext[x=1.153389in,y=0.068499in,left,base]{\color{textcolor}\rmfamily\fontsize{8.000000}{9.600000}\selectfont key}%
\end{pgfscope}%
\end{pgfpicture}%
\makeatother%
\endgroup%

%% file: figures/pgf/sweep_seqlen_1.3b.pgf
\begingroup%
\makeatletter%
\begin{pgfpicture}%
\pgfpathrectangle{\pgfpointorigin}{\pgfqpoint{1.460000in}{1.950000in}}%
\pgfusepath{use as bounding box, clip}%
\begin{pgfscope}%
\pgfsetbuttcap%
\pgfsetmiterjoin%
\definecolor{currentfill}{rgb}{1.000000,1.000000,1.000000}%
\pgfsetfillcolor{currentfill}%
\pgfsetlinewidth{0.000000pt}%
\definecolor{currentstroke}{rgb}{1.000000,1.000000,1.000000}%
\pgfsetstrokecolor{currentstroke}%
\pgfsetdash{}{0pt}%
\pgfpathmoveto{\pgfqpoint{0.000000in}{0.000000in}}%
\pgfpathlineto{\pgfqpoint{1.460000in}{0.000000in}}%
\pgfpathlineto{\pgfqpoint{1.460000in}{1.950000in}}%
\pgfpathlineto{\pgfqpoint{0.000000in}{1.950000in}}%
\pgfpathlineto{\pgfqpoint{0.000000in}{0.000000in}}%
\pgfpathclose%
\pgfusepath{fill}%
\end{pgfscope}%
\begin{pgfscope}%
\pgfsetbuttcap%
\pgfsetmiterjoin%
\definecolor{currentfill}{rgb}{1.000000,1.000000,1.000000}%
\pgfsetfillcolor{currentfill}%
\pgfsetlinewidth{0.000000pt}%
\definecolor{currentstroke}{rgb}{0.000000,0.000000,0.000000}%
\pgfsetstrokecolor{currentstroke}%
\pgfsetstrokeopacity{0.000000}%
\pgfsetdash{}{0pt}%
\pgfpathmoveto{\pgfqpoint{0.400000in}{0.720000in}}%
\pgfpathlineto{\pgfqpoint{1.410000in}{0.720000in}}%
\pgfpathlineto{\pgfqpoint{1.410000in}{1.730000in}}%
\pgfpathlineto{\pgfqpoint{0.400000in}{1.730000in}}%
\pgfpathlineto{\pgfqpoint{0.400000in}{0.720000in}}%
\pgfpathclose%
\pgfusepath{fill}%
\end{pgfscope}%
\begin{pgfscope}%
\pgfpathrectangle{\pgfqpoint{0.400000in}{0.720000in}}{\pgfqpoint{1.010000in}{1.010000in}}%
\pgfusepath{clip}%
\pgfsetrectcap%
\pgfsetroundjoin%
\pgfsetlinewidth{0.501875pt}%
\definecolor{currentstroke}{rgb}{0.886275,0.882353,0.866667}%
\pgfsetstrokecolor{currentstroke}%
\pgfsetdash{}{0pt}%
\pgfpathmoveto{\pgfqpoint{0.445909in}{0.720000in}}%
\pgfpathlineto{\pgfqpoint{0.445909in}{1.730000in}}%
\pgfusepath{stroke}%
\end{pgfscope}%
\begin{pgfscope}%
\pgfsetbuttcap%
\pgfsetroundjoin%
\definecolor{currentfill}{rgb}{0.321569,0.317647,0.305882}%
\pgfsetfillcolor{currentfill}%
\pgfsetlinewidth{0.602250pt}%
\definecolor{currentstroke}{rgb}{0.321569,0.317647,0.305882}%
\pgfsetstrokecolor{currentstroke}%
\pgfsetdash{}{0pt}%
\pgfsys@defobject{currentmarker}{\pgfqpoint{0.000000in}{-0.034722in}}{\pgfqpoint{0.000000in}{0.000000in}}{%
\pgfpathmoveto{\pgfqpoint{0.000000in}{0.000000in}}%
\pgfpathlineto{\pgfqpoint{0.000000in}{-0.034722in}}%
\pgfusepath{stroke,fill}%
}%
\begin{pgfscope}%
\pgfsys@transformshift{0.445909in}{0.720000in}%
\pgfsys@useobject{currentmarker}{}%
\end{pgfscope}%
\end{pgfscope}%
\begin{pgfscope}%
\definecolor{textcolor}{rgb}{0.321569,0.317647,0.305882}%
\pgfsetstrokecolor{textcolor}%
\pgfsetfillcolor{textcolor}%
\pgftext[x=0.445909in,y=0.636667in,,top]{\color{textcolor}\rmfamily\fontsize{8.000000}{9.600000}\selectfont 256}%
\end{pgfscope}%
\begin{pgfscope}%
\pgfpathrectangle{\pgfqpoint{0.400000in}{0.720000in}}{\pgfqpoint{1.010000in}{1.010000in}}%
\pgfusepath{clip}%
\pgfsetrectcap%
\pgfsetroundjoin%
\pgfsetlinewidth{0.501875pt}%
\definecolor{currentstroke}{rgb}{0.886275,0.882353,0.866667}%
\pgfsetstrokecolor{currentstroke}%
\pgfsetdash{}{0pt}%
\pgfpathmoveto{\pgfqpoint{0.751970in}{0.720000in}}%
\pgfpathlineto{\pgfqpoint{0.751970in}{1.730000in}}%
\pgfusepath{stroke}%
\end{pgfscope}%
\begin{pgfscope}%
\pgfsetbuttcap%
\pgfsetroundjoin%
\definecolor{currentfill}{rgb}{0.321569,0.317647,0.305882}%
\pgfsetfillcolor{currentfill}%
\pgfsetlinewidth{0.602250pt}%
\definecolor{currentstroke}{rgb}{0.321569,0.317647,0.305882}%
\pgfsetstrokecolor{currentstroke}%
\pgfsetdash{}{0pt}%
\pgfsys@defobject{currentmarker}{\pgfqpoint{0.000000in}{-0.034722in}}{\pgfqpoint{0.000000in}{0.000000in}}{%
\pgfpathmoveto{\pgfqpoint{0.000000in}{0.000000in}}%
\pgfpathlineto{\pgfqpoint{0.000000in}{-0.034722in}}%
\pgfusepath{stroke,fill}%
}%
\begin{pgfscope}%
\pgfsys@transformshift{0.751970in}{0.720000in}%
\pgfsys@useobject{currentmarker}{}%
\end{pgfscope}%
\end{pgfscope}%
\begin{pgfscope}%
\definecolor{textcolor}{rgb}{0.321569,0.317647,0.305882}%
\pgfsetstrokecolor{textcolor}%
\pgfsetfillcolor{textcolor}%
\pgftext[x=0.751970in,y=0.636667in,,top]{\color{textcolor}\rmfamily\fontsize{8.000000}{9.600000}\selectfont 512}%
\end{pgfscope}%
\begin{pgfscope}%
\pgfpathrectangle{\pgfqpoint{0.400000in}{0.720000in}}{\pgfqpoint{1.010000in}{1.010000in}}%
\pgfusepath{clip}%
\pgfsetrectcap%
\pgfsetroundjoin%
\pgfsetlinewidth{0.501875pt}%
\definecolor{currentstroke}{rgb}{0.886275,0.882353,0.866667}%
\pgfsetstrokecolor{currentstroke}%
\pgfsetdash{}{0pt}%
\pgfpathmoveto{\pgfqpoint{1.058030in}{0.720000in}}%
\pgfpathlineto{\pgfqpoint{1.058030in}{1.730000in}}%
\pgfusepath{stroke}%
\end{pgfscope}%
\begin{pgfscope}%
\pgfsetbuttcap%
\pgfsetroundjoin%
\definecolor{currentfill}{rgb}{0.321569,0.317647,0.305882}%
\pgfsetfillcolor{currentfill}%
\pgfsetlinewidth{0.602250pt}%
\definecolor{currentstroke}{rgb}{0.321569,0.317647,0.305882}%
\pgfsetstrokecolor{currentstroke}%
\pgfsetdash{}{0pt}%
\pgfsys@defobject{currentmarker}{\pgfqpoint{0.000000in}{-0.034722in}}{\pgfqpoint{0.000000in}{0.000000in}}{%
\pgfpathmoveto{\pgfqpoint{0.000000in}{0.000000in}}%
\pgfpathlineto{\pgfqpoint{0.000000in}{-0.034722in}}%
\pgfusepath{stroke,fill}%
}%
\begin{pgfscope}%
\pgfsys@transformshift{1.058030in}{0.720000in}%
\pgfsys@useobject{currentmarker}{}%
\end{pgfscope}%
\end{pgfscope}%
\begin{pgfscope}%
\definecolor{textcolor}{rgb}{0.321569,0.317647,0.305882}%
\pgfsetstrokecolor{textcolor}%
\pgfsetfillcolor{textcolor}%
\pgftext[x=1.058030in,y=0.636667in,,top]{\color{textcolor}\rmfamily\fontsize{8.000000}{9.600000}\selectfont 1k}%
\end{pgfscope}%
\begin{pgfscope}%
\pgfpathrectangle{\pgfqpoint{0.400000in}{0.720000in}}{\pgfqpoint{1.010000in}{1.010000in}}%
\pgfusepath{clip}%
\pgfsetrectcap%
\pgfsetroundjoin%
\pgfsetlinewidth{0.501875pt}%
\definecolor{currentstroke}{rgb}{0.886275,0.882353,0.866667}%
\pgfsetstrokecolor{currentstroke}%
\pgfsetdash{}{0pt}%
\pgfpathmoveto{\pgfqpoint{1.364091in}{0.720000in}}%
\pgfpathlineto{\pgfqpoint{1.364091in}{1.730000in}}%
\pgfusepath{stroke}%
\end{pgfscope}%
\begin{pgfscope}%
\pgfsetbuttcap%
\pgfsetroundjoin%
\definecolor{currentfill}{rgb}{0.321569,0.317647,0.305882}%
\pgfsetfillcolor{currentfill}%
\pgfsetlinewidth{0.602250pt}%
\definecolor{currentstroke}{rgb}{0.321569,0.317647,0.305882}%
\pgfsetstrokecolor{currentstroke}%
\pgfsetdash{}{0pt}%
\pgfsys@defobject{currentmarker}{\pgfqpoint{0.000000in}{-0.034722in}}{\pgfqpoint{0.000000in}{0.000000in}}{%
\pgfpathmoveto{\pgfqpoint{0.000000in}{0.000000in}}%
\pgfpathlineto{\pgfqpoint{0.000000in}{-0.034722in}}%
\pgfusepath{stroke,fill}%
}%
\begin{pgfscope}%
\pgfsys@transformshift{1.364091in}{0.720000in}%
\pgfsys@useobject{currentmarker}{}%
\end{pgfscope}%
\end{pgfscope}%
\begin{pgfscope}%
\definecolor{textcolor}{rgb}{0.321569,0.317647,0.305882}%
\pgfsetstrokecolor{textcolor}%
\pgfsetfillcolor{textcolor}%
\pgftext[x=1.364091in,y=0.636667in,,top]{\color{textcolor}\rmfamily\fontsize{8.000000}{9.600000}\selectfont 2k}%
\end{pgfscope}%
\begin{pgfscope}%
\pgfpathrectangle{\pgfqpoint{0.400000in}{0.720000in}}{\pgfqpoint{1.010000in}{1.010000in}}%
\pgfusepath{clip}%
\pgfsetrectcap%
\pgfsetroundjoin%
\pgfsetlinewidth{0.501875pt}%
\definecolor{currentstroke}{rgb}{0.886275,0.882353,0.866667}%
\pgfsetstrokecolor{currentstroke}%
\pgfsetdash{}{0pt}%
\pgfpathmoveto{\pgfqpoint{0.400000in}{0.776748in}}%
\pgfpathlineto{\pgfqpoint{1.410000in}{0.776748in}}%
\pgfusepath{stroke}%
\end{pgfscope}%
\begin{pgfscope}%
\pgfsetbuttcap%
\pgfsetroundjoin%
\definecolor{currentfill}{rgb}{0.321569,0.317647,0.305882}%
\pgfsetfillcolor{currentfill}%
\pgfsetlinewidth{0.602250pt}%
\definecolor{currentstroke}{rgb}{0.321569,0.317647,0.305882}%
\pgfsetstrokecolor{currentstroke}%
\pgfsetdash{}{0pt}%
\pgfsys@defobject{currentmarker}{\pgfqpoint{-0.034722in}{0.000000in}}{\pgfqpoint{-0.000000in}{0.000000in}}{%
\pgfpathmoveto{\pgfqpoint{-0.000000in}{0.000000in}}%
\pgfpathlineto{\pgfqpoint{-0.034722in}{0.000000in}}%
\pgfusepath{stroke,fill}%
}%
\begin{pgfscope}%
\pgfsys@transformshift{0.400000in}{0.776748in}%
\pgfsys@useobject{currentmarker}{}%
\end{pgfscope}%
\end{pgfscope}%
\begin{pgfscope}%
\definecolor{textcolor}{rgb}{0.101961,0.101961,0.098039}%
\pgfsetstrokecolor{textcolor}%
\pgfsetfillcolor{textcolor}%
\pgftext[x=0.261111in, y=0.738860in, left, base]{\color{textcolor}\rmfamily\fontsize{8.000000}{9.600000}\selectfont 1}%
\end{pgfscope}%
\begin{pgfscope}%
\pgfpathrectangle{\pgfqpoint{0.400000in}{0.720000in}}{\pgfqpoint{1.010000in}{1.010000in}}%
\pgfusepath{clip}%
\pgfsetrectcap%
\pgfsetroundjoin%
\pgfsetlinewidth{0.501875pt}%
\definecolor{currentstroke}{rgb}{0.886275,0.882353,0.866667}%
\pgfsetstrokecolor{currentstroke}%
\pgfsetdash{}{0pt}%
\pgfpathmoveto{\pgfqpoint{0.400000in}{1.022478in}}%
\pgfpathlineto{\pgfqpoint{1.410000in}{1.022478in}}%
\pgfusepath{stroke}%
\end{pgfscope}%
\begin{pgfscope}%
\pgfsetbuttcap%
\pgfsetroundjoin%
\definecolor{currentfill}{rgb}{0.321569,0.317647,0.305882}%
\pgfsetfillcolor{currentfill}%
\pgfsetlinewidth{0.602250pt}%
\definecolor{currentstroke}{rgb}{0.321569,0.317647,0.305882}%
\pgfsetstrokecolor{currentstroke}%
\pgfsetdash{}{0pt}%
\pgfsys@defobject{currentmarker}{\pgfqpoint{-0.034722in}{0.000000in}}{\pgfqpoint{-0.000000in}{0.000000in}}{%
\pgfpathmoveto{\pgfqpoint{-0.000000in}{0.000000in}}%
\pgfpathlineto{\pgfqpoint{-0.034722in}{0.000000in}}%
\pgfusepath{stroke,fill}%
}%
\begin{pgfscope}%
\pgfsys@transformshift{0.400000in}{1.022478in}%
\pgfsys@useobject{currentmarker}{}%
\end{pgfscope}%
\end{pgfscope}%
\begin{pgfscope}%
\definecolor{textcolor}{rgb}{0.101961,0.101961,0.098039}%
\pgfsetstrokecolor{textcolor}%
\pgfsetfillcolor{textcolor}%
\pgftext[x=0.261111in, y=0.984590in, left, base]{\color{textcolor}\rmfamily\fontsize{8.000000}{9.600000}\selectfont 2}%
\end{pgfscope}%
\begin{pgfscope}%
\pgfpathrectangle{\pgfqpoint{0.400000in}{0.720000in}}{\pgfqpoint{1.010000in}{1.010000in}}%
\pgfusepath{clip}%
\pgfsetrectcap%
\pgfsetroundjoin%
\pgfsetlinewidth{0.501875pt}%
\definecolor{currentstroke}{rgb}{0.886275,0.882353,0.866667}%
\pgfsetstrokecolor{currentstroke}%
\pgfsetdash{}{0pt}%
\pgfpathmoveto{\pgfqpoint{0.400000in}{1.268208in}}%
\pgfpathlineto{\pgfqpoint{1.410000in}{1.268208in}}%
\pgfusepath{stroke}%
\end{pgfscope}%
\begin{pgfscope}%
\pgfsetbuttcap%
\pgfsetroundjoin%
\definecolor{currentfill}{rgb}{0.321569,0.317647,0.305882}%
\pgfsetfillcolor{currentfill}%
\pgfsetlinewidth{0.602250pt}%
\definecolor{currentstroke}{rgb}{0.321569,0.317647,0.305882}%
\pgfsetstrokecolor{currentstroke}%
\pgfsetdash{}{0pt}%
\pgfsys@defobject{currentmarker}{\pgfqpoint{-0.034722in}{0.000000in}}{\pgfqpoint{-0.000000in}{0.000000in}}{%
\pgfpathmoveto{\pgfqpoint{-0.000000in}{0.000000in}}%
\pgfpathlineto{\pgfqpoint{-0.034722in}{0.000000in}}%
\pgfusepath{stroke,fill}%
}%
\begin{pgfscope}%
\pgfsys@transformshift{0.400000in}{1.268208in}%
\pgfsys@useobject{currentmarker}{}%
\end{pgfscope}%
\end{pgfscope}%
\begin{pgfscope}%
\definecolor{textcolor}{rgb}{0.101961,0.101961,0.098039}%
\pgfsetstrokecolor{textcolor}%
\pgfsetfillcolor{textcolor}%
\pgftext[x=0.261111in, y=1.230320in, left, base]{\color{textcolor}\rmfamily\fontsize{8.000000}{9.600000}\selectfont 4}%
\end{pgfscope}%
\begin{pgfscope}%
\pgfpathrectangle{\pgfqpoint{0.400000in}{0.720000in}}{\pgfqpoint{1.010000in}{1.010000in}}%
\pgfusepath{clip}%
\pgfsetrectcap%
\pgfsetroundjoin%
\pgfsetlinewidth{0.501875pt}%
\definecolor{currentstroke}{rgb}{0.886275,0.882353,0.866667}%
\pgfsetstrokecolor{currentstroke}%
\pgfsetdash{}{0pt}%
\pgfpathmoveto{\pgfqpoint{0.400000in}{1.513938in}}%
\pgfpathlineto{\pgfqpoint{1.410000in}{1.513938in}}%
\pgfusepath{stroke}%
\end{pgfscope}%
\begin{pgfscope}%
\pgfsetbuttcap%
\pgfsetroundjoin%
\definecolor{currentfill}{rgb}{0.321569,0.317647,0.305882}%
\pgfsetfillcolor{currentfill}%
\pgfsetlinewidth{0.602250pt}%
\definecolor{currentstroke}{rgb}{0.321569,0.317647,0.305882}%
\pgfsetstrokecolor{currentstroke}%
\pgfsetdash{}{0pt}%
\pgfsys@defobject{currentmarker}{\pgfqpoint{-0.034722in}{0.000000in}}{\pgfqpoint{-0.000000in}{0.000000in}}{%
\pgfpathmoveto{\pgfqpoint{-0.000000in}{0.000000in}}%
\pgfpathlineto{\pgfqpoint{-0.034722in}{0.000000in}}%
\pgfusepath{stroke,fill}%
}%
\begin{pgfscope}%
\pgfsys@transformshift{0.400000in}{1.513938in}%
\pgfsys@useobject{currentmarker}{}%
\end{pgfscope}%
\end{pgfscope}%
\begin{pgfscope}%
\definecolor{textcolor}{rgb}{0.101961,0.101961,0.098039}%
\pgfsetstrokecolor{textcolor}%
\pgfsetfillcolor{textcolor}%
\pgftext[x=0.261111in, y=1.476050in, left, base]{\color{textcolor}\rmfamily\fontsize{8.000000}{9.600000}\selectfont 8}%
\end{pgfscope}%
\begin{pgfscope}%
\definecolor{textcolor}{rgb}{0.101961,0.101961,0.098039}%
\pgfsetstrokecolor{textcolor}%
\pgfsetfillcolor{textcolor}%
\pgftext[x=0.205556in,y=1.225000in,,bottom,rotate=90.000000]{\color{textcolor}\rmfamily\fontsize{9.000000}{10.800000}\selectfont latency (ms)}%
\end{pgfscope}%
\begin{pgfscope}%
\pgfpathrectangle{\pgfqpoint{0.400000in}{0.720000in}}{\pgfqpoint{1.010000in}{1.010000in}}%
\pgfusepath{clip}%
\pgfsetbuttcap%
\pgfsetroundjoin%
\pgfsetlinewidth{1.104125pt}%
\definecolor{currentstroke}{rgb}{0.321569,0.317647,0.305882}%
\pgfsetstrokecolor{currentstroke}%
\pgfsetdash{{4.070000pt}{1.760000pt}}{0.000000pt}%
\pgfpathmoveto{\pgfqpoint{0.445909in}{0.946329in}}%
\pgfpathlineto{\pgfqpoint{0.751970in}{1.193158in}}%
\pgfpathlineto{\pgfqpoint{1.058030in}{1.436551in}}%
\pgfpathlineto{\pgfqpoint{1.364091in}{1.684091in}}%
\pgfusepath{stroke}%
\end{pgfscope}%
\begin{pgfscope}%
\pgfpathrectangle{\pgfqpoint{0.400000in}{0.720000in}}{\pgfqpoint{1.010000in}{1.010000in}}%
\pgfusepath{clip}%
\pgfsetbuttcap%
\pgfsetroundjoin%
\definecolor{currentfill}{rgb}{0.321569,0.317647,0.305882}%
\pgfsetfillcolor{currentfill}%
\pgfsetlinewidth{1.003750pt}%
\definecolor{currentstroke}{rgb}{0.321569,0.317647,0.305882}%
\pgfsetstrokecolor{currentstroke}%
\pgfsetdash{}{0pt}%
\pgfsys@defobject{currentmarker}{\pgfqpoint{-0.015278in}{-0.015278in}}{\pgfqpoint{0.015278in}{0.015278in}}{%
\pgfpathmoveto{\pgfqpoint{0.000000in}{-0.015278in}}%
\pgfpathcurveto{\pgfqpoint{0.004052in}{-0.015278in}}{\pgfqpoint{0.007938in}{-0.013668in}}{\pgfqpoint{0.010803in}{-0.010803in}}%
\pgfpathcurveto{\pgfqpoint{0.013668in}{-0.007938in}}{\pgfqpoint{0.015278in}{-0.004052in}}{\pgfqpoint{0.015278in}{0.000000in}}%
\pgfpathcurveto{\pgfqpoint{0.015278in}{0.004052in}}{\pgfqpoint{0.013668in}{0.007938in}}{\pgfqpoint{0.010803in}{0.010803in}}%
\pgfpathcurveto{\pgfqpoint{0.007938in}{0.013668in}}{\pgfqpoint{0.004052in}{0.015278in}}{\pgfqpoint{0.000000in}{0.015278in}}%
\pgfpathcurveto{\pgfqpoint{-0.004052in}{0.015278in}}{\pgfqpoint{-0.007938in}{0.013668in}}{\pgfqpoint{-0.010803in}{0.010803in}}%
\pgfpathcurveto{\pgfqpoint{-0.013668in}{0.007938in}}{\pgfqpoint{-0.015278in}{0.004052in}}{\pgfqpoint{-0.015278in}{0.000000in}}%
\pgfpathcurveto{\pgfqpoint{-0.015278in}{-0.004052in}}{\pgfqpoint{-0.013668in}{-0.007938in}}{\pgfqpoint{-0.010803in}{-0.010803in}}%
\pgfpathcurveto{\pgfqpoint{-0.007938in}{-0.013668in}}{\pgfqpoint{-0.004052in}{-0.015278in}}{\pgfqpoint{0.000000in}{-0.015278in}}%
\pgfpathlineto{\pgfqpoint{0.000000in}{-0.015278in}}%
\pgfpathclose%
\pgfusepath{stroke,fill}%
}%
\begin{pgfscope}%
\pgfsys@transformshift{0.445909in}{0.946329in}%
\pgfsys@useobject{currentmarker}{}%
\end{pgfscope}%
\begin{pgfscope}%
\pgfsys@transformshift{0.751970in}{1.193158in}%
\pgfsys@useobject{currentmarker}{}%
\end{pgfscope}%
\begin{pgfscope}%
\pgfsys@transformshift{1.058030in}{1.436551in}%
\pgfsys@useobject{currentmarker}{}%
\end{pgfscope}%
\begin{pgfscope}%
\pgfsys@transformshift{1.364091in}{1.684091in}%
\pgfsys@useobject{currentmarker}{}%
\end{pgfscope}%
\end{pgfscope}%
\begin{pgfscope}%
\pgfpathrectangle{\pgfqpoint{0.400000in}{0.720000in}}{\pgfqpoint{1.010000in}{1.010000in}}%
\pgfusepath{clip}%
\pgfsetrectcap%
\pgfsetroundjoin%
\pgfsetlinewidth{1.104125pt}%
\definecolor{currentstroke}{rgb}{0.862745,0.223529,0.466667}%
\pgfsetstrokecolor{currentstroke}%
\pgfsetdash{}{0pt}%
\pgfpathmoveto{\pgfqpoint{0.445909in}{0.795565in}}%
\pgfpathlineto{\pgfqpoint{0.751970in}{1.021237in}}%
\pgfpathlineto{\pgfqpoint{1.058030in}{1.257687in}}%
\pgfpathlineto{\pgfqpoint{1.364091in}{1.496852in}}%
\pgfusepath{stroke}%
\end{pgfscope}%
\begin{pgfscope}%
\pgfpathrectangle{\pgfqpoint{0.400000in}{0.720000in}}{\pgfqpoint{1.010000in}{1.010000in}}%
\pgfusepath{clip}%
\pgfsetbuttcap%
\pgfsetroundjoin%
\definecolor{currentfill}{rgb}{0.862745,0.223529,0.466667}%
\pgfsetfillcolor{currentfill}%
\pgfsetlinewidth{1.003750pt}%
\definecolor{currentstroke}{rgb}{0.862745,0.223529,0.466667}%
\pgfsetstrokecolor{currentstroke}%
\pgfsetdash{}{0pt}%
\pgfsys@defobject{currentmarker}{\pgfqpoint{-0.015278in}{-0.015278in}}{\pgfqpoint{0.015278in}{0.015278in}}{%
\pgfpathmoveto{\pgfqpoint{0.000000in}{-0.015278in}}%
\pgfpathcurveto{\pgfqpoint{0.004052in}{-0.015278in}}{\pgfqpoint{0.007938in}{-0.013668in}}{\pgfqpoint{0.010803in}{-0.010803in}}%
\pgfpathcurveto{\pgfqpoint{0.013668in}{-0.007938in}}{\pgfqpoint{0.015278in}{-0.004052in}}{\pgfqpoint{0.015278in}{0.000000in}}%
\pgfpathcurveto{\pgfqpoint{0.015278in}{0.004052in}}{\pgfqpoint{0.013668in}{0.007938in}}{\pgfqpoint{0.010803in}{0.010803in}}%
\pgfpathcurveto{\pgfqpoint{0.007938in}{0.013668in}}{\pgfqpoint{0.004052in}{0.015278in}}{\pgfqpoint{0.000000in}{0.015278in}}%
\pgfpathcurveto{\pgfqpoint{-0.004052in}{0.015278in}}{\pgfqpoint{-0.007938in}{0.013668in}}{\pgfqpoint{-0.010803in}{0.010803in}}%
\pgfpathcurveto{\pgfqpoint{-0.013668in}{0.007938in}}{\pgfqpoint{-0.015278in}{0.004052in}}{\pgfqpoint{-0.015278in}{0.000000in}}%
\pgfpathcurveto{\pgfqpoint{-0.015278in}{-0.004052in}}{\pgfqpoint{-0.013668in}{-0.007938in}}{\pgfqpoint{-0.010803in}{-0.010803in}}%
\pgfpathcurveto{\pgfqpoint{-0.007938in}{-0.013668in}}{\pgfqpoint{-0.004052in}{-0.015278in}}{\pgfqpoint{0.000000in}{-0.015278in}}%
\pgfpathlineto{\pgfqpoint{0.000000in}{-0.015278in}}%
\pgfpathclose%
\pgfusepath{stroke,fill}%
}%
\begin{pgfscope}%
\pgfsys@transformshift{0.445909in}{0.795565in}%
\pgfsys@useobject{currentmarker}{}%
\end{pgfscope}%
\begin{pgfscope}%
\pgfsys@transformshift{0.751970in}{1.021237in}%
\pgfsys@useobject{currentmarker}{}%
\end{pgfscope}%
\begin{pgfscope}%
\pgfsys@transformshift{1.058030in}{1.257687in}%
\pgfsys@useobject{currentmarker}{}%
\end{pgfscope}%
\begin{pgfscope}%
\pgfsys@transformshift{1.364091in}{1.496852in}%
\pgfsys@useobject{currentmarker}{}%
\end{pgfscope}%
\end{pgfscope}%
\begin{pgfscope}%
\pgfpathrectangle{\pgfqpoint{0.400000in}{0.720000in}}{\pgfqpoint{1.010000in}{1.010000in}}%
\pgfusepath{clip}%
\pgfsetrectcap%
\pgfsetroundjoin%
\pgfsetlinewidth{1.104125pt}%
\definecolor{currentstroke}{rgb}{0.031373,0.564706,0.600000}%
\pgfsetstrokecolor{currentstroke}%
\pgfsetdash{}{0pt}%
\pgfpathmoveto{\pgfqpoint{0.445909in}{0.765909in}}%
\pgfpathlineto{\pgfqpoint{0.751970in}{1.014841in}}%
\pgfpathlineto{\pgfqpoint{1.058030in}{1.258394in}}%
\pgfpathlineto{\pgfqpoint{1.364091in}{1.501516in}}%
\pgfusepath{stroke}%
\end{pgfscope}%
\begin{pgfscope}%
\pgfpathrectangle{\pgfqpoint{0.400000in}{0.720000in}}{\pgfqpoint{1.010000in}{1.010000in}}%
\pgfusepath{clip}%
\pgfsetbuttcap%
\pgfsetroundjoin%
\definecolor{currentfill}{rgb}{0.031373,0.564706,0.600000}%
\pgfsetfillcolor{currentfill}%
\pgfsetlinewidth{1.003750pt}%
\definecolor{currentstroke}{rgb}{0.031373,0.564706,0.600000}%
\pgfsetstrokecolor{currentstroke}%
\pgfsetdash{}{0pt}%
\pgfsys@defobject{currentmarker}{\pgfqpoint{-0.015278in}{-0.015278in}}{\pgfqpoint{0.015278in}{0.015278in}}{%
\pgfpathmoveto{\pgfqpoint{0.000000in}{-0.015278in}}%
\pgfpathcurveto{\pgfqpoint{0.004052in}{-0.015278in}}{\pgfqpoint{0.007938in}{-0.013668in}}{\pgfqpoint{0.010803in}{-0.010803in}}%
\pgfpathcurveto{\pgfqpoint{0.013668in}{-0.007938in}}{\pgfqpoint{0.015278in}{-0.004052in}}{\pgfqpoint{0.015278in}{0.000000in}}%
\pgfpathcurveto{\pgfqpoint{0.015278in}{0.004052in}}{\pgfqpoint{0.013668in}{0.007938in}}{\pgfqpoint{0.010803in}{0.010803in}}%
\pgfpathcurveto{\pgfqpoint{0.007938in}{0.013668in}}{\pgfqpoint{0.004052in}{0.015278in}}{\pgfqpoint{0.000000in}{0.015278in}}%
\pgfpathcurveto{\pgfqpoint{-0.004052in}{0.015278in}}{\pgfqpoint{-0.007938in}{0.013668in}}{\pgfqpoint{-0.010803in}{0.010803in}}%
\pgfpathcurveto{\pgfqpoint{-0.013668in}{0.007938in}}{\pgfqpoint{-0.015278in}{0.004052in}}{\pgfqpoint{-0.015278in}{0.000000in}}%
\pgfpathcurveto{\pgfqpoint{-0.015278in}{-0.004052in}}{\pgfqpoint{-0.013668in}{-0.007938in}}{\pgfqpoint{-0.010803in}{-0.010803in}}%
\pgfpathcurveto{\pgfqpoint{-0.007938in}{-0.013668in}}{\pgfqpoint{-0.004052in}{-0.015278in}}{\pgfqpoint{0.000000in}{-0.015278in}}%
\pgfpathlineto{\pgfqpoint{0.000000in}{-0.015278in}}%
\pgfpathclose%
\pgfusepath{stroke,fill}%
}%
\begin{pgfscope}%
\pgfsys@transformshift{0.445909in}{0.765909in}%
\pgfsys@useobject{currentmarker}{}%
\end{pgfscope}%
\begin{pgfscope}%
\pgfsys@transformshift{0.751970in}{1.014841in}%
\pgfsys@useobject{currentmarker}{}%
\end{pgfscope}%
\begin{pgfscope}%
\pgfsys@transformshift{1.058030in}{1.258394in}%
\pgfsys@useobject{currentmarker}{}%
\end{pgfscope}%
\begin{pgfscope}%
\pgfsys@transformshift{1.364091in}{1.501516in}%
\pgfsys@useobject{currentmarker}{}%
\end{pgfscope}%
\end{pgfscope}%
\begin{pgfscope}%
\pgfpathrectangle{\pgfqpoint{0.400000in}{0.720000in}}{\pgfqpoint{1.010000in}{1.010000in}}%
\pgfusepath{clip}%
\pgfsetrectcap%
\pgfsetroundjoin%
\pgfsetlinewidth{1.104125pt}%
\definecolor{currentstroke}{rgb}{0.015686,0.321569,0.458824}%
\pgfsetstrokecolor{currentstroke}%
\pgfsetdash{}{0pt}%
\pgfpathmoveto{\pgfqpoint{0.445909in}{0.785256in}}%
\pgfpathlineto{\pgfqpoint{0.751970in}{1.033779in}}%
\pgfpathlineto{\pgfqpoint{1.058030in}{1.279256in}}%
\pgfpathlineto{\pgfqpoint{1.364091in}{1.521703in}}%
\pgfusepath{stroke}%
\end{pgfscope}%
\begin{pgfscope}%
\pgfpathrectangle{\pgfqpoint{0.400000in}{0.720000in}}{\pgfqpoint{1.010000in}{1.010000in}}%
\pgfusepath{clip}%
\pgfsetbuttcap%
\pgfsetroundjoin%
\definecolor{currentfill}{rgb}{0.015686,0.321569,0.458824}%
\pgfsetfillcolor{currentfill}%
\pgfsetlinewidth{1.003750pt}%
\definecolor{currentstroke}{rgb}{0.015686,0.321569,0.458824}%
\pgfsetstrokecolor{currentstroke}%
\pgfsetdash{}{0pt}%
\pgfsys@defobject{currentmarker}{\pgfqpoint{-0.015278in}{-0.015278in}}{\pgfqpoint{0.015278in}{0.015278in}}{%
\pgfpathmoveto{\pgfqpoint{0.000000in}{-0.015278in}}%
\pgfpathcurveto{\pgfqpoint{0.004052in}{-0.015278in}}{\pgfqpoint{0.007938in}{-0.013668in}}{\pgfqpoint{0.010803in}{-0.010803in}}%
\pgfpathcurveto{\pgfqpoint{0.013668in}{-0.007938in}}{\pgfqpoint{0.015278in}{-0.004052in}}{\pgfqpoint{0.015278in}{0.000000in}}%
\pgfpathcurveto{\pgfqpoint{0.015278in}{0.004052in}}{\pgfqpoint{0.013668in}{0.007938in}}{\pgfqpoint{0.010803in}{0.010803in}}%
\pgfpathcurveto{\pgfqpoint{0.007938in}{0.013668in}}{\pgfqpoint{0.004052in}{0.015278in}}{\pgfqpoint{0.000000in}{0.015278in}}%
\pgfpathcurveto{\pgfqpoint{-0.004052in}{0.015278in}}{\pgfqpoint{-0.007938in}{0.013668in}}{\pgfqpoint{-0.010803in}{0.010803in}}%
\pgfpathcurveto{\pgfqpoint{-0.013668in}{0.007938in}}{\pgfqpoint{-0.015278in}{0.004052in}}{\pgfqpoint{-0.015278in}{0.000000in}}%
\pgfpathcurveto{\pgfqpoint{-0.015278in}{-0.004052in}}{\pgfqpoint{-0.013668in}{-0.007938in}}{\pgfqpoint{-0.010803in}{-0.010803in}}%
\pgfpathcurveto{\pgfqpoint{-0.007938in}{-0.013668in}}{\pgfqpoint{-0.004052in}{-0.015278in}}{\pgfqpoint{0.000000in}{-0.015278in}}%
\pgfpathlineto{\pgfqpoint{0.000000in}{-0.015278in}}%
\pgfpathclose%
\pgfusepath{stroke,fill}%
}%
\begin{pgfscope}%
\pgfsys@transformshift{0.445909in}{0.785256in}%
\pgfsys@useobject{currentmarker}{}%
\end{pgfscope}%
\begin{pgfscope}%
\pgfsys@transformshift{0.751970in}{1.033779in}%
\pgfsys@useobject{currentmarker}{}%
\end{pgfscope}%
\begin{pgfscope}%
\pgfsys@transformshift{1.058030in}{1.279256in}%
\pgfsys@useobject{currentmarker}{}%
\end{pgfscope}%
\begin{pgfscope}%
\pgfsys@transformshift{1.364091in}{1.521703in}%
\pgfsys@useobject{currentmarker}{}%
\end{pgfscope}%
\end{pgfscope}%
\begin{pgfscope}%
\pgfsetrectcap%
\pgfsetmiterjoin%
\pgfsetlinewidth{0.602250pt}%
\definecolor{currentstroke}{rgb}{0.321569,0.317647,0.305882}%
\pgfsetstrokecolor{currentstroke}%
\pgfsetdash{}{0pt}%
\pgfpathmoveto{\pgfqpoint{0.400000in}{0.720000in}}%
\pgfpathlineto{\pgfqpoint{0.400000in}{1.730000in}}%
\pgfusepath{stroke}%
\end{pgfscope}%
\begin{pgfscope}%
\pgfsetrectcap%
\pgfsetmiterjoin%
\pgfsetlinewidth{0.602250pt}%
\definecolor{currentstroke}{rgb}{0.321569,0.317647,0.305882}%
\pgfsetstrokecolor{currentstroke}%
\pgfsetdash{}{0pt}%
\pgfpathmoveto{\pgfqpoint{0.400000in}{0.720000in}}%
\pgfpathlineto{\pgfqpoint{1.410000in}{0.720000in}}%
\pgfusepath{stroke}%
\end{pgfscope}%
\begin{pgfscope}%
\definecolor{textcolor}{rgb}{0.101961,0.101961,0.098039}%
\pgfsetstrokecolor{textcolor}%
\pgfsetfillcolor{textcolor}%
\pgftext[x=0.905000in,y=1.813333in,,base]{\color{textcolor}\rmfamily\fontsize{8.500000}{10.200000}\selectfont Mamba-2 1.3B}%
\end{pgfscope}%
\begin{pgfscope}%
\definecolor{textcolor}{rgb}{0.101961,0.101961,0.098039}%
\pgfsetstrokecolor{textcolor}%
\pgfsetfillcolor{textcolor}%
\pgftext[x=0.905000in,y=0.420000in,,]{\color{textcolor}\rmfamily\fontsize{9.000000}{10.800000}\selectfont sequence length}%
\end{pgfscope}%
\begin{pgfscope}%
\pgfsetbuttcap%
\pgfsetroundjoin%
\pgfsetlinewidth{1.104125pt}%
\definecolor{currentstroke}{rgb}{0.321569,0.317647,0.305882}%
\pgfsetstrokecolor{currentstroke}%
\pgfsetdash{{4.070000pt}{1.760000pt}}{0.000000pt}%
\pgfpathmoveto{\pgfqpoint{0.418725in}{0.264776in}}%
\pgfpathlineto{\pgfqpoint{0.485391in}{0.264776in}}%
\pgfpathlineto{\pgfqpoint{0.552058in}{0.264776in}}%
\pgfusepath{stroke}%
\end{pgfscope}%
\begin{pgfscope}%
\pgfsetbuttcap%
\pgfsetroundjoin%
\definecolor{currentfill}{rgb}{0.321569,0.317647,0.305882}%
\pgfsetfillcolor{currentfill}%
\pgfsetlinewidth{1.003750pt}%
\definecolor{currentstroke}{rgb}{0.321569,0.317647,0.305882}%
\pgfsetstrokecolor{currentstroke}%
\pgfsetdash{}{0pt}%
\pgfsys@defobject{currentmarker}{\pgfqpoint{-0.015278in}{-0.015278in}}{\pgfqpoint{0.015278in}{0.015278in}}{%
\pgfpathmoveto{\pgfqpoint{0.000000in}{-0.015278in}}%
\pgfpathcurveto{\pgfqpoint{0.004052in}{-0.015278in}}{\pgfqpoint{0.007938in}{-0.013668in}}{\pgfqpoint{0.010803in}{-0.010803in}}%
\pgfpathcurveto{\pgfqpoint{0.013668in}{-0.007938in}}{\pgfqpoint{0.015278in}{-0.004052in}}{\pgfqpoint{0.015278in}{0.000000in}}%
\pgfpathcurveto{\pgfqpoint{0.015278in}{0.004052in}}{\pgfqpoint{0.013668in}{0.007938in}}{\pgfqpoint{0.010803in}{0.010803in}}%
\pgfpathcurveto{\pgfqpoint{0.007938in}{0.013668in}}{\pgfqpoint{0.004052in}{0.015278in}}{\pgfqpoint{0.000000in}{0.015278in}}%
\pgfpathcurveto{\pgfqpoint{-0.004052in}{0.015278in}}{\pgfqpoint{-0.007938in}{0.013668in}}{\pgfqpoint{-0.010803in}{0.010803in}}%
\pgfpathcurveto{\pgfqpoint{-0.013668in}{0.007938in}}{\pgfqpoint{-0.015278in}{0.004052in}}{\pgfqpoint{-0.015278in}{0.000000in}}%
\pgfpathcurveto{\pgfqpoint{-0.015278in}{-0.004052in}}{\pgfqpoint{-0.013668in}{-0.007938in}}{\pgfqpoint{-0.010803in}{-0.010803in}}%
\pgfpathcurveto{\pgfqpoint{-0.007938in}{-0.013668in}}{\pgfqpoint{-0.004052in}{-0.015278in}}{\pgfqpoint{0.000000in}{-0.015278in}}%
\pgfpathlineto{\pgfqpoint{0.000000in}{-0.015278in}}%
\pgfpathclose%
\pgfusepath{stroke,fill}%
}%
\begin{pgfscope}%
\pgfsys@transformshift{0.485391in}{0.264776in}%
\pgfsys@useobject{currentmarker}{}%
\end{pgfscope}%
\end{pgfscope}%
\begin{pgfscope}%
\definecolor{textcolor}{rgb}{0.101961,0.101961,0.098039}%
\pgfsetstrokecolor{textcolor}%
\pgfsetfillcolor{textcolor}%
\pgftext[x=0.596503in,y=0.225887in,left,base]{\color{textcolor}\rmfamily\fontsize{8.000000}{9.600000}\selectfont FP16}%
\end{pgfscope}%
\begin{pgfscope}%
\pgfsetrectcap%
\pgfsetroundjoin%
\pgfsetlinewidth{1.104125pt}%
\definecolor{currentstroke}{rgb}{0.862745,0.223529,0.466667}%
\pgfsetstrokecolor{currentstroke}%
\pgfsetdash{}{0pt}%
\pgfpathmoveto{\pgfqpoint{0.418725in}{0.107388in}}%
\pgfpathlineto{\pgfqpoint{0.485391in}{0.107388in}}%
\pgfpathlineto{\pgfqpoint{0.552058in}{0.107388in}}%
\pgfusepath{stroke}%
\end{pgfscope}%
\begin{pgfscope}%
\pgfsetbuttcap%
\pgfsetroundjoin%
\definecolor{currentfill}{rgb}{0.862745,0.223529,0.466667}%
\pgfsetfillcolor{currentfill}%
\pgfsetlinewidth{1.003750pt}%
\definecolor{currentstroke}{rgb}{0.862745,0.223529,0.466667}%
\pgfsetstrokecolor{currentstroke}%
\pgfsetdash{}{0pt}%
\pgfsys@defobject{currentmarker}{\pgfqpoint{-0.015278in}{-0.015278in}}{\pgfqpoint{0.015278in}{0.015278in}}{%
\pgfpathmoveto{\pgfqpoint{0.000000in}{-0.015278in}}%
\pgfpathcurveto{\pgfqpoint{0.004052in}{-0.015278in}}{\pgfqpoint{0.007938in}{-0.013668in}}{\pgfqpoint{0.010803in}{-0.010803in}}%
\pgfpathcurveto{\pgfqpoint{0.013668in}{-0.007938in}}{\pgfqpoint{0.015278in}{-0.004052in}}{\pgfqpoint{0.015278in}{0.000000in}}%
\pgfpathcurveto{\pgfqpoint{0.015278in}{0.004052in}}{\pgfqpoint{0.013668in}{0.007938in}}{\pgfqpoint{0.010803in}{0.010803in}}%
\pgfpathcurveto{\pgfqpoint{0.007938in}{0.013668in}}{\pgfqpoint{0.004052in}{0.015278in}}{\pgfqpoint{0.000000in}{0.015278in}}%
\pgfpathcurveto{\pgfqpoint{-0.004052in}{0.015278in}}{\pgfqpoint{-0.007938in}{0.013668in}}{\pgfqpoint{-0.010803in}{0.010803in}}%
\pgfpathcurveto{\pgfqpoint{-0.013668in}{0.007938in}}{\pgfqpoint{-0.015278in}{0.004052in}}{\pgfqpoint{-0.015278in}{0.000000in}}%
\pgfpathcurveto{\pgfqpoint{-0.015278in}{-0.004052in}}{\pgfqpoint{-0.013668in}{-0.007938in}}{\pgfqpoint{-0.010803in}{-0.010803in}}%
\pgfpathcurveto{\pgfqpoint{-0.007938in}{-0.013668in}}{\pgfqpoint{-0.004052in}{-0.015278in}}{\pgfqpoint{0.000000in}{-0.015278in}}%
\pgfpathlineto{\pgfqpoint{0.000000in}{-0.015278in}}%
\pgfpathclose%
\pgfusepath{stroke,fill}%
}%
\begin{pgfscope}%
\pgfsys@transformshift{0.485391in}{0.107388in}%
\pgfsys@useobject{currentmarker}{}%
\end{pgfscope}%
\end{pgfscope}%
\begin{pgfscope}%
\definecolor{textcolor}{rgb}{0.101961,0.101961,0.098039}%
\pgfsetstrokecolor{textcolor}%
\pgfsetfillcolor{textcolor}%
\pgftext[x=0.596503in,y=0.068499in,left,base]{\color{textcolor}\rmfamily\fontsize{8.000000}{9.600000}\selectfont SSDi8}%
\end{pgfscope}%
\begin{pgfscope}%
\pgfsetrectcap%
\pgfsetroundjoin%
\pgfsetlinewidth{1.104125pt}%
\definecolor{currentstroke}{rgb}{0.031373,0.564706,0.600000}%
\pgfsetstrokecolor{currentstroke}%
\pgfsetdash{}{0pt}%
\pgfpathmoveto{\pgfqpoint{0.975611in}{0.264776in}}%
\pgfpathlineto{\pgfqpoint{1.042277in}{0.264776in}}%
\pgfpathlineto{\pgfqpoint{1.108944in}{0.264776in}}%
\pgfusepath{stroke}%
\end{pgfscope}%
\begin{pgfscope}%
\pgfsetbuttcap%
\pgfsetroundjoin%
\definecolor{currentfill}{rgb}{0.031373,0.564706,0.600000}%
\pgfsetfillcolor{currentfill}%
\pgfsetlinewidth{1.003750pt}%
\definecolor{currentstroke}{rgb}{0.031373,0.564706,0.600000}%
\pgfsetstrokecolor{currentstroke}%
\pgfsetdash{}{0pt}%
\pgfsys@defobject{currentmarker}{\pgfqpoint{-0.015278in}{-0.015278in}}{\pgfqpoint{0.015278in}{0.015278in}}{%
\pgfpathmoveto{\pgfqpoint{0.000000in}{-0.015278in}}%
\pgfpathcurveto{\pgfqpoint{0.004052in}{-0.015278in}}{\pgfqpoint{0.007938in}{-0.013668in}}{\pgfqpoint{0.010803in}{-0.010803in}}%
\pgfpathcurveto{\pgfqpoint{0.013668in}{-0.007938in}}{\pgfqpoint{0.015278in}{-0.004052in}}{\pgfqpoint{0.015278in}{0.000000in}}%
\pgfpathcurveto{\pgfqpoint{0.015278in}{0.004052in}}{\pgfqpoint{0.013668in}{0.007938in}}{\pgfqpoint{0.010803in}{0.010803in}}%
\pgfpathcurveto{\pgfqpoint{0.007938in}{0.013668in}}{\pgfqpoint{0.004052in}{0.015278in}}{\pgfqpoint{0.000000in}{0.015278in}}%
\pgfpathcurveto{\pgfqpoint{-0.004052in}{0.015278in}}{\pgfqpoint{-0.007938in}{0.013668in}}{\pgfqpoint{-0.010803in}{0.010803in}}%
\pgfpathcurveto{\pgfqpoint{-0.013668in}{0.007938in}}{\pgfqpoint{-0.015278in}{0.004052in}}{\pgfqpoint{-0.015278in}{0.000000in}}%
\pgfpathcurveto{\pgfqpoint{-0.015278in}{-0.004052in}}{\pgfqpoint{-0.013668in}{-0.007938in}}{\pgfqpoint{-0.010803in}{-0.010803in}}%
\pgfpathcurveto{\pgfqpoint{-0.007938in}{-0.013668in}}{\pgfqpoint{-0.004052in}{-0.015278in}}{\pgfqpoint{0.000000in}{-0.015278in}}%
\pgfpathlineto{\pgfqpoint{0.000000in}{-0.015278in}}%
\pgfpathclose%
\pgfusepath{stroke,fill}%
}%
\begin{pgfscope}%
\pgfsys@transformshift{1.042277in}{0.264776in}%
\pgfsys@useobject{currentmarker}{}%
\end{pgfscope}%
\end{pgfscope}%
\begin{pgfscope}%
\definecolor{textcolor}{rgb}{0.101961,0.101961,0.098039}%
\pgfsetstrokecolor{textcolor}%
\pgfsetfillcolor{textcolor}%
\pgftext[x=1.153389in,y=0.225887in,left,base]{\color{textcolor}\rmfamily\fontsize{8.000000}{9.600000}\selectfont value}%
\end{pgfscope}%
\begin{pgfscope}%
\pgfsetrectcap%
\pgfsetroundjoin%
\pgfsetlinewidth{1.104125pt}%
\definecolor{currentstroke}{rgb}{0.015686,0.321569,0.458824}%
\pgfsetstrokecolor{currentstroke}%
\pgfsetdash{}{0pt}%
\pgfpathmoveto{\pgfqpoint{0.975611in}{0.107388in}}%
\pgfpathlineto{\pgfqpoint{1.042277in}{0.107388in}}%
\pgfpathlineto{\pgfqpoint{1.108944in}{0.107388in}}%
\pgfusepath{stroke}%
\end{pgfscope}%
\begin{pgfscope}%
\pgfsetbuttcap%
\pgfsetroundjoin%
\definecolor{currentfill}{rgb}{0.015686,0.321569,0.458824}%
\pgfsetfillcolor{currentfill}%
\pgfsetlinewidth{1.003750pt}%
\definecolor{currentstroke}{rgb}{0.015686,0.321569,0.458824}%
\pgfsetstrokecolor{currentstroke}%
\pgfsetdash{}{0pt}%
\pgfsys@defobject{currentmarker}{\pgfqpoint{-0.015278in}{-0.015278in}}{\pgfqpoint{0.015278in}{0.015278in}}{%
\pgfpathmoveto{\pgfqpoint{0.000000in}{-0.015278in}}%
\pgfpathcurveto{\pgfqpoint{0.004052in}{-0.015278in}}{\pgfqpoint{0.007938in}{-0.013668in}}{\pgfqpoint{0.010803in}{-0.010803in}}%
\pgfpathcurveto{\pgfqpoint{0.013668in}{-0.007938in}}{\pgfqpoint{0.015278in}{-0.004052in}}{\pgfqpoint{0.015278in}{0.000000in}}%
\pgfpathcurveto{\pgfqpoint{0.015278in}{0.004052in}}{\pgfqpoint{0.013668in}{0.007938in}}{\pgfqpoint{0.010803in}{0.010803in}}%
\pgfpathcurveto{\pgfqpoint{0.007938in}{0.013668in}}{\pgfqpoint{0.004052in}{0.015278in}}{\pgfqpoint{0.000000in}{0.015278in}}%
\pgfpathcurveto{\pgfqpoint{-0.004052in}{0.015278in}}{\pgfqpoint{-0.007938in}{0.013668in}}{\pgfqpoint{-0.010803in}{0.010803in}}%
\pgfpathcurveto{\pgfqpoint{-0.013668in}{0.007938in}}{\pgfqpoint{-0.015278in}{0.004052in}}{\pgfqpoint{-0.015278in}{0.000000in}}%
\pgfpathcurveto{\pgfqpoint{-0.015278in}{-0.004052in}}{\pgfqpoint{-0.013668in}{-0.007938in}}{\pgfqpoint{-0.010803in}{-0.010803in}}%
\pgfpathcurveto{\pgfqpoint{-0.007938in}{-0.013668in}}{\pgfqpoint{-0.004052in}{-0.015278in}}{\pgfqpoint{0.000000in}{-0.015278in}}%
\pgfpathlineto{\pgfqpoint{0.000000in}{-0.015278in}}%
\pgfpathclose%
\pgfusepath{stroke,fill}%
}%
\begin{pgfscope}%
\pgfsys@transformshift{1.042277in}{0.107388in}%
\pgfsys@useobject{currentmarker}{}%
\end{pgfscope}%
\end{pgfscope}%
\begin{pgfscope}%
\definecolor{textcolor}{rgb}{0.101961,0.101961,0.098039}%
\pgfsetstrokecolor{textcolor}%
\pgfsetfillcolor{textcolor}%
\pgftext[x=1.153389in,y=0.068499in,left,base]{\color{textcolor}\rmfamily\fontsize{8.000000}{9.600000}\selectfont key}%
\end{pgfscope}%
\end{pgfpicture}%
\makeatother%
\endgroup%